\documentclass[acmsmall,screen]{acmart}

\usepackage{tcolorbox}
\usepackage{fontawesome}
\usepackage{float}
\usepackage[normalem]{ulem}
\usepackage{geometry}
\usepackage{multirow,bigstrut}
\usepackage{colortbl}
\usepackage{longtable}
\usepackage{changepage}
\usepackage[printonlyused]{acronym} % Acronyms with \ac
\usepackage{bookmark}
\usepackage{natbib}
\usepackage{placeins}

\newcolumntype{P}[1]{>{\centering\arraybackslash}p{#1}}

\definecolor{Gray1}{gray}{0.95}
\definecolor{Gray2}{gray}{0.89}
\definecolor{ILCblue}{RGB}{68,119,170}
\definecolor{ISpink}{RGB}{170,51,119}
\definecolor{SSyellow}{RGB}{204,187,68}
\definecolor{ODgreen}{RGB}{34,136,51}

\definecolor{RelGreen}{RGB}{154,203,63}
\definecolor{RelYellow}{RGB}{253,234,81}
\definecolor{RelOrange}{RGB}{253,184,41}
\definecolor{RelBlue}{RGB}{58,187,201}
\definecolor{RelGray}{RGB}{94,89,89}

\hypersetup{
  pdfinfo={
  pdfproducer={},
  Title={},
  Subject={},
  Author={},
  }
}

\makeatletter
\AtBeginDocument{%
  \renewcommand*{\AC@hyperlink}[2]{%
    \begingroup
      \hypersetup{hidelinks}%
      \hyperlink{#1}{#2}%
    \endgroup
  }%
}
\makeatother

\AtBeginDocument{%
  \providecommand\BibTeX{{%
    \normalfont B\kern-0.5em{\scshape i\kern-0.25em b}\kern-0.8em\TeX}}}

\begin{document}

%\DeclareUrlCommand\ULurl@{%
%  \def\UrlFont{\ttfamily\color{blue}}%
%  \def\UrlLeft{\uline\bgroup}%
%  \def\UrlRight{\egroup}}
%\def\ULurl@#1{\hyper@linkurl{\ULurl@@{#1}}{#1}}
%\DeclareRobustCommand*\ULurl{\hyper@normalise\ULurl@}

\title[Metrics Reloaded]{Metrics Reloaded: Recommendations for image analysis validation}


%------------------------------------
% Core team
%------------------------------------

\author{Lena Maier-Hein}
% \authornote{The complete list of affiliations can be found in Appendix~\ref{app:authors}.}
\authornote{\textbf{Corresponding authors:} Lena Maier-Hein, l.maier-hein@dkfz-heidelberg.de; Annika Reinke: a.reinke@dkfz-heidelberg.de; Paul F. Jäger: p.jaeger@dkfz-heidelberg.de.}
\authornote{\textbf{Shared first authors:} Lena Maier-Hein and Annika Reinke}

\email{l.maier-hein@dkfz.de}
\affiliation{%
  \institution{German Cancer Research Center (DKFZ) Heidelberg, Division of Intelligent Medical Systems and HI Helmholtz Imaging}
  % \department{Div. Intelligent Medical Systems and HI Helmholtz Imaging}
  % \city{}
  \country{Germany}
}
\affiliation{%
  \institution{Faculty of Mathematics and Computer Science and Medical Faculty, Heidelberg University}
  % \department{Faculty of Mathematics and Computer Science and Medical Faculty}
  % \city{Heidelberg}
  \country{Heidelberg, Germany}
}
\affiliation{%
  \institution{National Center for Tumor Diseases (NCT), NCT Heidelberg, a partnership between DKFZ and University Medical Center Heidelberg}
  % \city{Heidelberg}
  \country{Germany}
}

\author{Annika Reinke}
\authornotemark[1]
\authornotemark[2]
\affiliation{%
  \institution{German Cancer Research Center (DKFZ) Heidelberg, Division of Intelligent Medical Systems and HI Helmholtz Imaging}
  % \department{Div. Intelligent Medical Systems and HI Helmholtz Imaging}
  % \city{Heidelberg}
  \country{Germany}
}
\affiliation{%
  \institution{Faculty of Mathematics and Computer Science, Heidelberg University}
  % \department{Faculty of Mathematics and Computer Science}
  % \city{Heidelberg}
  \country{Heidelberg, Germany}
}

\author{Patrick Godau}
\affiliation{%
  \institution{German Cancer Research Center (DKFZ) Heidelberg, Division of Intelligent Medical Systems}
  % \department{Div. Intelligent Medical Systems}
  % \city{Heidelberg}
  \country{Germany}
}
\affiliation{%
  \institution{Faculty of Mathematics and Computer Science, Heidelberg University}
  % \department{Faculty of Mathematics and Computer Science}
  % \city{Heidelberg}
  \country{Heidelberg, Germany}
}
\affiliation{%
  \institution{National Center for Tumor Diseases (NCT), NCT Heidelberg, a partnership between DKFZ and University Medical Center Heidelberg}
  % \city{Heidelberg}
  \country{Germany}
}

\author{Minu D. Tizabi}
\affiliation{%
  \institution{German Cancer Research Center (DKFZ) Heidelberg, Division of Intelligent Medical Systems}
  % \department{Div. Intelligent Medical Systems and HI Helmholtz Imaging}
  % \city{Heidelberg}
  \country{Germany}
}
\affiliation{%
  \institution{National Center for Tumor Diseases (NCT), NCT Heidelberg, a partnership between DKFZ and University Medical Center Heidelberg}
  % \city{Heidelberg}
  \country{Germany}
}

%------------------------------------
% Group Leads (alphabetic order)
%------------------------------------
\author{Florian Buettner}
\affiliation{%
  \institution{German Cancer Consortium (DKTK), partner site Frankfurt/Mainz, a partnership between DKFZ and UCT Frankfurt-Marburg}
  % \city{Heidelberg}
  \country{Germany}
}
\affiliation{%
  \institution{German Cancer Research Center (DKFZ) Heidelberg}
  % \city{Heidelberg}
  \country{Germany}
}
\affiliation{%
 \institution{Goethe University Frankfurt, Department of Medicine}
  % \city{Frankfurt}
  \country{Germany}
}
\affiliation{%
 \institution{Goethe University Frankfurt, Department of Informatics}
  % \city{Frankfurt}
  \country{Germany}
}
\affiliation{%
  \institution{Frankfurt Cancer Insititute}
  % \city{Frankfurt}
  \country{Germany}
}

\author{Evangelia Christodoulou}
\affiliation{%
  \institution{German Cancer Research Center (DKFZ) Heidelberg, Division of Intelligent Medical Systems}
  % \department{Div. Intelligent Medical Systems}
  % \city{Heidelberg}
  \country{Germany}
}

\author{Ben Glocker}
\affiliation{%
  \institution{Department of Computing, Imperial College London}
  % \department{Department of Computing}
  % \city{London}
  \country{London, UK}}

\author{Fabian Isensee}
\affiliation{%
  \institution{German Cancer Research Center (DKFZ) Heidelberg, Division of Medical Image Computing and HI Applied Computer Vision Lab}
  % \department{HI Applied Computer Vision Lab, Div. of Medical Image Computing}
  % \city{Heidelberg}
  \country{Germany}
}

\author{Jens Kleesiek}
\affiliation{%
  \institution{Institute for AI in Medicine, University Medicine Essen}
  % \department{Institute for AI in Medicine}
  % \city{Essen}
  \country{Essen, Germany}
}

\author{Michal Kozubek}
\affiliation{%
  \institution{Centre for Biomedical Image Analysis and Faculty of Informatics, Masaryk University}
  % \department{Centre for Biomedical Image Analysis and Faculty of Informatics}
  % \city{Brno}
  \country{Brno, Czech Republic}
}

\author{Mauricio Reyes}
\affiliation{%
  \institution{ARTORG Center for Biomedical Engineering Research, University of Bern}
  \country{Bern, Switzerland}
}
\affiliation{%
  \institution{Department of Radiation Oncology, University Hospital Bern, University of Bern}
  \country{Bern, Switzerland}
}

\author{Michael A. Riegler}
\affiliation{%
  \institution{Simula Metropolitan Center for Digital Engineering}
  % \city{Oslo}
  \country{Oslo, Norway}
}
\affiliation{%
  \institution{UiT The Arctic University of Norway}
  % \city{Oslo}
  \country{romsø, Norway}
}

\author{Manuel Wiesenfarth}
\affiliation{%
  \institution{German Cancer Research Center (DKFZ) Heidelberg, Division of Biostatistics}
  % \department{Div. Biostatistics}
  % \city{Heidelberg}
  \country{Germany}
}

%------------------------------------
% Implementation leads (alphabetic order)
%------------------------------------
\author{A. Emre Kavur}
\affiliation{%
  \institution{German Cancer Research Center (DKFZ) Heidelberg, Division of Intelligent Medical Systems, Division of Medical Image Computing, HI Applied Computer Vision Lab}
  % \department{HI Applied Computer Vision Lab, Div. of Medical Image Computing and Div. Intelligent Medical Systems}
  % \city{Heidelberg}
  \country{Germany}
}

\author{Carole H. Sudre}
\affiliation{%
 \institution{MRC Unit for Lifelong Health and Ageing at UCL and Centre for Medical Image Computing, Department of Computer Science, University College London}
 % \department{MRC Unit for Lifelong Health and Ageing at UCL and Centre for Medical Image Computing, Department of Computer Science}
 % \city{London}
 \country{London, UK}}
 \affiliation{%
 \institution{School of Biomedical Engineering and Imaging Science, King's College London}
 % \department{School of Biomedical Engineering and Imaging Science}
 % \city{London}
 \country{London, UK}}

%------------------------------------
% Extended Core Team (alphabetic order)
%------------------------------------
\author{Michael Baumgartner}
\affiliation{%
  \institution{German Cancer Research Center (DKFZ) Heidelberg, Division of Medical Image Computing}
  % \department{Div. Medical Image Computing}
  % \city{Heidelberg}
  \country{Germany}
}

\author{Matthias Eisenmann}
\affiliation{%
  \institution{German Cancer Research Center (DKFZ) Heidelberg, Division of Intelligent Medical Systems}
  % \department{Div. Intelligent Medical Systems and HI Helmholtz Imaging}
  % \city{Heidelberg}
  \country{Germany}
}

\author{Doreen Heckmann-Nötzel}
\affiliation{%
  \institution{German Cancer Research Center (DKFZ) Heidelberg, Division of Intelligent Medical Systems}
  % \department{Div. Intelligent Medical Systems}
  % \city{Heidelberg}
  \country{Germany}
}
\affiliation{%
  \institution{National Center for Tumor Diseases (NCT), NCT Heidelberg, a partnership between DKFZ and University Medical Center Heidelberg}
  % \city{Heidelberg}
  \country{Germany}
}

\author{Tim Rädsch}
\affiliation{%
  \institution{German Cancer Research Center (DKFZ) Heidelberg, Division of Intelligent Medical Systems and HI Helmholtz Imaging}
  % \department{Div. Intelligent Medical Systems and HI Helmholtz Imaging}
  % \city{Heidelberg}
  \country{Germany}
}

%------------------------------------
% Committee (alphabetic order)
%------------------------------------
\author{Laura Acion}
\affiliation{%
  \institution{Instituto de Cálculo, CONICET -- Universidad de Buenos Aires}
  % \department{Instituto de Cálculo}
  % \city{Buenos Aires}
  \country{Buenos Aires, Argentina}
}

\author{Michela Antonelli}
\affiliation{%
 \institution{School of Biomedical Engineering and Imaging Science, King's College London}
 % \department{School of Biomedical Engineering and Imaging Science}
 % \city{London}
 \country{London, UK}}
\affiliation{%
 \institution{Centre for Medical Image Computing, University College London}
 % \department{Centre for Medical Image Computing}
 % \city{London}
 \country{London, UK}}

\author{Tal Arbel}
\affiliation{%
  \institution{Centre for Intelligent Machines and MILA (Québec Artificial Intelligence Institute), McGill University}
  % \department{Centre for Intelligent Machines and MILA (Québec Artificial Intelligence Institute)}
  % \city{Montreal}
  \country{Montréal, Canada}}

\author{Spyridon Bakas}
\affiliation{%
  \institution{Division of Computational Pathology, Dept of Pathology \& Laboratory Medicine, Indiana University School of Medicine, IU Health Information and Translational Sciences Building}
  \country{Indianapolis, USA}}
\affiliation{%
  \institution{Center for Biomedical Image Computing and Analytics (CBICA), University of Pennsylvania, Richards Medical Research Laboratories FL7}
  \country{Philadelphia, PA, USA}}

\author{Arriel Benis}
\affiliation{%
  \institution{Department of Digital Medical Technologies, Holon Institute of Technology}
  % \department{Department of Digital Medical Technologies}
  % \city{Holon}
  \country{Holon, Israel}
}
\affiliation{%
  \institution{European Federation for Medical Informatics}
  % \city{Le Mont-sur-Lausanne}
  \country{Le Mont-sur-Lausanne, Switzerland}
}

\author{Matthew B. Blaschko}
\affiliation{%
  \institution{Center for Processing Speech and Images, Department of Electrical Engineering, KU Leuven}
  % \department{Center for Processing Speech and Images, Department of Electrical Engineering}
  % \city{Leuven}
  \country{Kasteelpark Arenberg 10 - box 2441, 3001 Leuven, Belgium}
}

\author{M. Jorge Cardoso}
\affiliation{%
 \institution{School of Biomedical Engineering and Imaging Science, King's College London}
 % \department{School of Biomedical Engineering and Imaging Science}
 % \city{London}
 \country{London, UK}}
  
\author{Veronika Cheplygina}
\affiliation{%
  \institution{Department of Computer Science, IT University of Copenhagen}
  % \department{Department of Computer Science}
  % \city{Copenhagen}
  \country{Copenhagen, Denmark}}

\author{Beth A. Cimini}
\affiliation{%
  \institution{Imaging Platform, Broad Institute of MIT and Harvard}
  % \department{Imaging Platform}
  % \city{Cambridge, Massachusetts}
  \country{Cambridge, MA, USA}}

\author{Gary S. Collins}
\affiliation{%
  \institution{Centre for Statistics in Medicine, University of Oxford}
  % \department{Centre for Statistics in Medicine}
  % \city{Oxford}
  \country{Oxford, UK}}
  
\author{Keyvan Farahani}
\affiliation{%
  \institution{Center for Biomedical Informatics and Information Technology, National Cancer Institute}
  % \department{Center for Biomedical Informatics and Information Technology}
  \country{Bethesda, MD, USA}}
  
\author{Luciana Ferrer}
\affiliation{%
  % \department{Instituto de Investigacion en Ciencias de la Computacion (ICC)}
\institution{Instituto de Investigación en Ciencias de la Computación (ICC), CONICET-UBA}
  \country{Ciudad Universitaria, Ciudad Autónoma de Buenos Aires, Argentina}}
  
\author{Adrian Galdran}
\affiliation{%
  \institution{Universitat Pompeu Fabra}
  % \city{Barcelona}
  \country{Barcelona, Spain}}
\affiliation{%
  \institution{University of Adelaide}
  % \city{Adelaide}
  \country{Adelaide, Australia}}  

\author{Bram van Ginneken}
\affiliation{%
  \institution{Fraunhofer MEVIS}
  % \city{Bremen}
  \country{Bremen, Germany}
}
\affiliation{%
  \institution{Radboud Institute for Health Sciences, Radboud University Medical Center}
  % \department{Radboud Institute for Health Sciences}
  % \city{Nijmegen}
  \country{Nijmegen, The Netherlands}
}
  
\author{Robert Haase}
\affiliation{%
  \institution{Now with: Center for Scalable Data Analytics and Artificial Intelligence (ScaDS.AI), Leipzig University}
  \country{Leipzig, Germany}
}
\affiliation{%
  \institution{Technische Universität (TU) Dresden, DFG Cluster of Excellence "Physics of Life"}
  % \department{Technische Universität (TU) Dresden}
  % \city{Dresden}
  \country{Dresden, Germany}
}
\affiliation{%
  \institution{Center for Systems Biology}
  % \city{Dresden}
  \country{Dresden, Germany}
}

\author{Daniel A. Hashimoto}
\affiliation{%
  \institution{Department of Surgery, Perelman School of Medicine}
  % \department{Department of Surgery}
  % \city{Philadelphia, PA}
  \country{Philadelphia, PA, USA}}
\affiliation{%
  \institution{General Robotics Automation Sensing and Perception Laboratory, School of Engineering and Applied Science, University of Pennsylvania}
  % \department{General Robotics Automation Sensing and Perception Laboratory, School of Engineering and Applied Science}
  % \city{Philadelphia, PA}
  \country{Philadelphia, PA, USA}}
  
\author{Michael M. Hoffman}
\affiliation{%
  \institution{Princess Margaret Cancer Centre, University Health Network}
  \country{Toronto, Canada}}
\affiliation{%
  \institution{Department of Medical Biophysics, University of Toronto}
  \country{Toronto, Canada}}
\affiliation{%
  \institution{Department of Computer Science, University of Toronto}
  \country{Toronto, Canada}}
\affiliation{%
  \institution{Vector Institute for Artificial Intelligence}
  \country{Toronto, Canada}}

\author{Merel Huisman}
\affiliation{%
  \institution{Department of Radiology and Nuclear Medicine, Radboud University Medical Center}
  % \department{Department of Radiology}
  % \city{Nijmegen}
  \country{Nijmegen, The Netherlands}
}

\author{Pierre Jannin}
\affiliation{%
  \institution{Laboratoire Traitement du Signal et de l’Image – UMR\_S 1099, Université de Rennes 1}
  % \department{Laboratoire Traitement du Signal et de l’Image – UMR\_S 1099}
  % \city{Rennes}
  \country{Rennes, France}
}
\affiliation{%
  \institution{INSERM}
  % \city{Rennes}
  \country{Paris Cedex, France}
}

\author{Charles E. Kahn}
\affiliation{%
  \institution{Department of Radiology and Institute for Biomedical Informatics, University of Pennsylvania}
  % \department{Department of Radiology and Institute for Biomedical Informatics}
  % \city{Philadelphia, Pennsylvania}
  \country{Philadelphia, PA, USA}
}

\author{Dagmar Kainmueller}
\affiliation{%
  \institution{Max-Delbrück Center for Molecular Medicine in the Helmholtz Association (MDC), Biomedical Image Analysis and HI Helmholtz Imaging}
  \country{Berlin, Germany}
}
\affiliation{%
  \institution{University of Potsdam, Digital Engineering Faculty}
  \country{Potsdam, Germany}
}

\author{Bernhard Kainz}
\affiliation{%
  \institution{Department of Computing, Faculty of Engineering, Imperial College London}
  % \department{Department of Computing, Faculty of Engineering}
  % \city{London}
  \country{London, UK}
}
\affiliation{%
  \institution{Department AIBE, Friedrich-Alexander-Universität (FAU)}
  % \department{Department AIBE}
  % \city{Erlangen-Nürnberg}
  \country{Erlangen-Nürnberg, Germany}
}

\author{Alexandros Karargyris}
\affiliation{%
  \institution{IHU Strasbourg}
  % \city{Strasbourg}
  \country{Strasbourg, France}
}

\author{Alan Karthikesalingam}
\affiliation{%
  \institution{Google Health DeepMind}
  % \city{London}
  \country{London, UK}
}

\author{Hannes Kenngott}
\affiliation{%
  \institution{Department of General, Visceral and Transplantation Surgery, Heidelberg University Hospital}
  % \department{}
  % \city{Heidelberg}
  \country{Heidelberg, Germany}
}

\author{Florian Kofler}
\affiliation{%
  \institution{Helmholtz AI}
  % \city{München}
  \country{München, Germany}
}

\author{Annette Kopp-Schneider}
\affiliation{%
  \institution{German Cancer Research Center (DKFZ) Heidelberg, Division of Biostatistics}
  % \department{Div. Biostatistics}
  % \city{Heidelberg}
  \country{Germany}
}

\author{Anna Kreshuk}
\affiliation{%
  \institution{Cell Biology and Biophysics Unit, European Molecular Biology Laboratory (EMBL)}
  % \department{Cell Biology and Biophysics Unit}
  % \city{Heidelberg}
  \country{Heidelberg, Germany}
}

\author{Tahsin Kurc}
\affiliation{%
  \institution{Department of Biomedical Informatics, Stony Brook University}
  % \department{Department of Biomedical Informatics}
  % \city{Stony Brook, New York}
  \country{Stony Brook, NY, USA}
}

\author{Bennett A. Landman}
\affiliation{%
  \institution{Electrical Engineering, Vanderbilt University}
  % \department{Electrical Engineering}
  % \city{Nashville, Tennessee}
  \country{Nashville, TN, USA}
}

\author{Geert Litjens}
\affiliation{%
  \institution{Department of Pathology, Radboud University Medical Center}
  % \department{Department of Pathology}
  % \city{Nijmegen}
  \country{Nijmegen, The Netherlands}
}

\author{Amin Madani}
\affiliation{%
  \institution{Department of Surgery, University Health Network}
  % \department{Department of Surgery}
  % \city{Toronto}
  \country{Philadelphia, PA, Canada}
}

\author{Klaus Maier-Hein}
\affiliation{%
  \institution{German Cancer Research Center (DKFZ) Heidelberg, Division of Medical Image Computing and HI Helmholtz Imaging,}
  % \department{Div. Medical Image Computing and HI Helmholtz Imaging}
  % \city{Heidelberg}
  \country{Germany}
}
\affiliation{%
  \institution{Pattern Analysis and Learning Group, Department of Radiation Oncology, Heidelberg University Hospital}
  % \department{Pattern Analysis and Learning Group, Department of Radiation Oncology}
  % \city{Heidelberg}
  \country{Heidelberg, Germany}
}

\author{Anne L. Martel}
\affiliation{%
  \institution{Physical Sciences, Sunnybrook Research Institute}
  % \department{Physical Sciences}
  % \city{Toronto}
  \country{Toronto, Canada}
}
\affiliation{%
  \institution{Department of Medical Biophysics, University of Toronto}
  % \department{Department of Medical Biophysics}
  % \city{Toronto}
  \country{Toronto, Canada}
}

\author{Peter Mattson}
\affiliation{%
  \institution{Google}
  % \city{Mountain View}
  \country{Mountain View, CA 94043, USA}
}

\author{Erik Meijering}
\affiliation{%
  \institution{School of Computer Science and Engineering, University of New South Wales}
  % \department{School of Computer Science and Engineering}
  % \city{New South Wales}
  \country{Sydney, Kensington, Australia}
}

\author{Bjoern Menze}
\affiliation{%
  \institution{Department of Quantitative Biomedicine, University of Zurich}
  % \department{Department of Quantitative Biomedicine}
  % \city{Zurich}
  \country{Zurich, Switzerland}
}

\author{Karel G.M. Moons}
\affiliation{%
  \institution{Julius Center for Health Sciences and Primary Care, UMC Utrecht, Utrecht University}
  % \department{Julius Center for Health Sciences and Primary Care}
  % \city{Utrecht}
  \country{Utrecht, The Netherlands}
}

\author{Henning Müller}
\affiliation{%
  \institution{Information Systems Institute, University of Applied Sciences Western Switzerland (HES-SO)}
  % \department{Information Systems Institute}
  % \city{Sierre}
  \country{Sierre, Switzerland}
}
\affiliation{%
  \institution{Medical Faculty, University of Geneva}
  % \department{Medical Faculty}
  % \city{Geneva}
  \country{Geneva, Switzerland}
}

\author{Brennan Nichyporuk}
\affiliation{%
  \institution{MILA (Québec Artificial Intelligence Institute)}
  % \city{Montreal}
  \country{Montréal, Canada}}
  
\author{Felix Nickel}
\affiliation{%
  \institution{Department of General, Visceral and Thoracic Surgery, University Medical Center Hamburg-Eppendorf}
  % \department{Department of General, Visceral and Thoracic Surgery}
  % \city{Hamburg}
  \country{Hamburg, Germany}
}

\author{Jens Petersen}
\affiliation{%
  \institution{German Cancer Research Center (DKFZ) Heidelberg, Division of Medical Image Computing}
  % \department{Div. Medical Image Computing}
  % \city{Heidelberg}
  \country{Germany}
}

\author{Nasir Rajpoot}
\affiliation{%
  \institution{Tissue Image Analytics Laboratory, Department of Computer Science, University of Warwick}
  % \department{Tissue Image Analytics Laboratory, Department of Computer Science}
  % \city{Coventry, West Midlands}
  \country{Coventry, UK}
}

\author{Nicola Rieke}
\affiliation{%
  \institution{NVIDIA GmbH}
  % \city{Munich}
  \country{München, Germany}
}

\author{Julio Saez-Rodriguez}
\affiliation{%
  \institution{Institute for Computational Biomedicine, Heidelberg University}
  % \department{Institute for Computational Biomedicine}
  % \city{Heidelberg}
  \country{Heidelberg, Germany}
}
\affiliation{%
    % \department{Faculty of Medicine}
  \institution{Faculty of Medicine, Heidelberg University Hospital}
  % \city{Heidelberg}
  \country{Heidelberg, Germany}
}
% \affiliation{%
%   \institution{BioQuant}
%   % \city{Heidelberg}
%   \country{Heidelberg, Germany}
% }

\author{Clara I. Sánchez}
\affiliation{%
  \institution{Informatics Institute, Faculty of Science, University of Amsterdam}
  % \department{Informatics Institute, Faculty of Science}
  % \city{Amsterdam}
  \country{Amsterdam, The Netherlands}
}

\author{Shravya Shetty}
\affiliation{%
  \institution{Google Health, Google}
  % \department{Google Health}
  % \city{Palo Alto}
  \country{Palo Alto, CA, USA}
}

\author{Maarten van Smeden}
\affiliation{%
  \institution{Julius Center for Health Sciences and Primary Care, University Medical Center Utrecht}
  % \department{Julius Center for Health Sciences and Primary Care}
  % \city{Utrecht}
  \country{Utrecht, The Netherlands}
}

\author{Ronald M. Summers}
\affiliation{%
  \institution{National Institutes of Health Clinical Center}
  % \department{Radiology and Imaging Sciences, Clinical Center}
  % \city{Bethesda, Maryland}
  \country{Bethesda, MD, USA}
}

\author{Abdel A. Taha}
\affiliation{%
\institution{Institute of Information Systems Engineering, TU Wien}
% \department{Institute of Information Systems Engineering}
  % \city{Vienna}
  \country{ Vienna, Austria}
}

\author{Aleksei Tiulpin}
\affiliation{%
\institution{Research Unit of Health Sciences and Technology, Faculty of Medicine, University of Oulu}
  \country{Oulu, Finland}
}
\affiliation{%
\institution{Neurocenter Oulu, Oulu University Hospital}
  \country{Oulu, Finland}
}

\author{Sotirios A. Tsaftaris}
\affiliation{%
  \institution{School of Engineering, The University of Edinburgh}
  % \department{School of Engineering}
  % \city{Edinburgh}
  \country{Edinburgh, Scotland}
}

\author{Ben Van Calster}
\affiliation{%
  \institution{Department of Development and Regeneration and EPI-centre, KU Leuven}
  % \department{Department of Development and Regeneration and EPI-centre}
  % \city{Leuven}
  \country{Leuven, Belgium}
}
\affiliation{%
  \institution{Department of Biomedical Data Sciences, Leiden University Medical Center}
  % \department{Department of Biomedical Data Sciences}
  % \city{Leiden}
  \country{Leiden, The Netherlands}
}

\author{Gaël Varoquaux}
\affiliation{%
  \institution{Parietal project team, INRIA Saclay-Île de France}
  % \department{Parietal project team}
  % \city{Palaiseau}
  \country{Palaiseau, France}
}

%------------------------------------
% Senior author 
%------------------------------------
\author{Paul F. Jäger}
\authornotemark[1]
\affiliation{%
  \institution{German Cancer Research Center (DKFZ) Heidelberg, Interactive Machine Learning Group and HI Helmholtz Imaging}
  % \department{Interactive Machine Learning Group and HI Helmholtz Imaging}
  % \city{Heidelberg}
  \country{Germany}
}

\renewcommand{\shortauthors}{Maier-Hein/Reinke et al.}

\begin{abstract}
\newpage
\textbf{Abstract:} 

Increasing evidence shows that flaws in machine learning (ML) algorithm validation are an underestimated global problem. Particularly in automatic biomedical image analysis, chosen performance metrics often do not reflect the domain interest, thus failing to adequately measure scientific progress and hindering translation of ML techniques into practice. To overcome this, we created \textit{Metrics Reloaded}, a comprehensive framework guiding researchers in the problem-aware selection of metrics. It was developed by a large international consortium in a multi-stage Delphi process and is based on the novel concept of a \textit{problem fingerprint} -- a structured representation of the given problem that captures all aspects that are relevant for metric selection, from the domain interest to the properties of the target structure(s), data set and algorithm output. Based on the problem fingerprint, users are guided through the process of choosing and applying appropriate validation metrics while being made aware of potential pitfalls. \textit{Metrics Reloaded} targets image analysis problems that can be interpreted as a classification task at image, object or pixel level, namely \textit{image-level classification},
\textit{object detection},
\textit{semantic segmentation}, and \textit{instance segmentation} tasks. To improve the user experience, we implemented the framework in the \textit{Metrics Reloaded} online tool. Following the convergence of ML methodology across application domains, \textit{Metrics Reloaded} fosters the convergence of validation methodology. Its applicability is demonstrated for various biomedical use cases.

\end{abstract}


\maketitle
\setlength{\parskip}{0.5em}
%\newpage
%\addtocontents{toc}{\protect\setcounter{tocdepth}{-1}}
%\tableofcontents
%\addtocontents{toc}{\protect\setcounter{tocdepth}{2}}

\newpage
\section*{Main}
\addcontentsline{toc}{chapter}{\protect\numberline{}Main}

Automatic image processing with \ac{ML} is  gaining increasing traction in biological and medical imaging research and practice. Research has predominantly focused on the development of new image processing algorithms. The critical issue of reliable and objective performance assessment of these algorithms, however, remains largely unexplored.
While the suitability of new medical treatments is typically directly assessed via well-interpretable and clinically meaningful measures, such as survival and complication rates, algorithm performance in image processing is commonly assessed with validation metrics\footnote{Not to be confused with distance metrics in the pure mathematical sense.} that should serve as proxies for the domain interest. In consequence, the impact of validation metrics cannot be overstated; first, they are the basis for deciding on the practical (e.g. clinical) suitability of a method and are thus \textit{a key component for translation into biomedical practice}. In fact, validation that is not conducted according to relevant metrics could be one major reason for why many \ac{AI} developments in medical imaging fail to reach clinical practice \cite{kelly2019key, shah2019making}. In other words, the numbers presented in journals and conference proceedings do not reflect how successful a system will be when applied in practice. Second, \textit{metrics guide the scientific progress in the field}; flawed metric use can lead to  entirely futile resource investment and infeasible research directions while obscuring true scientific advancements. 

\begin{figure}[H]
    \centering
    \includegraphics[width=.9\linewidth]{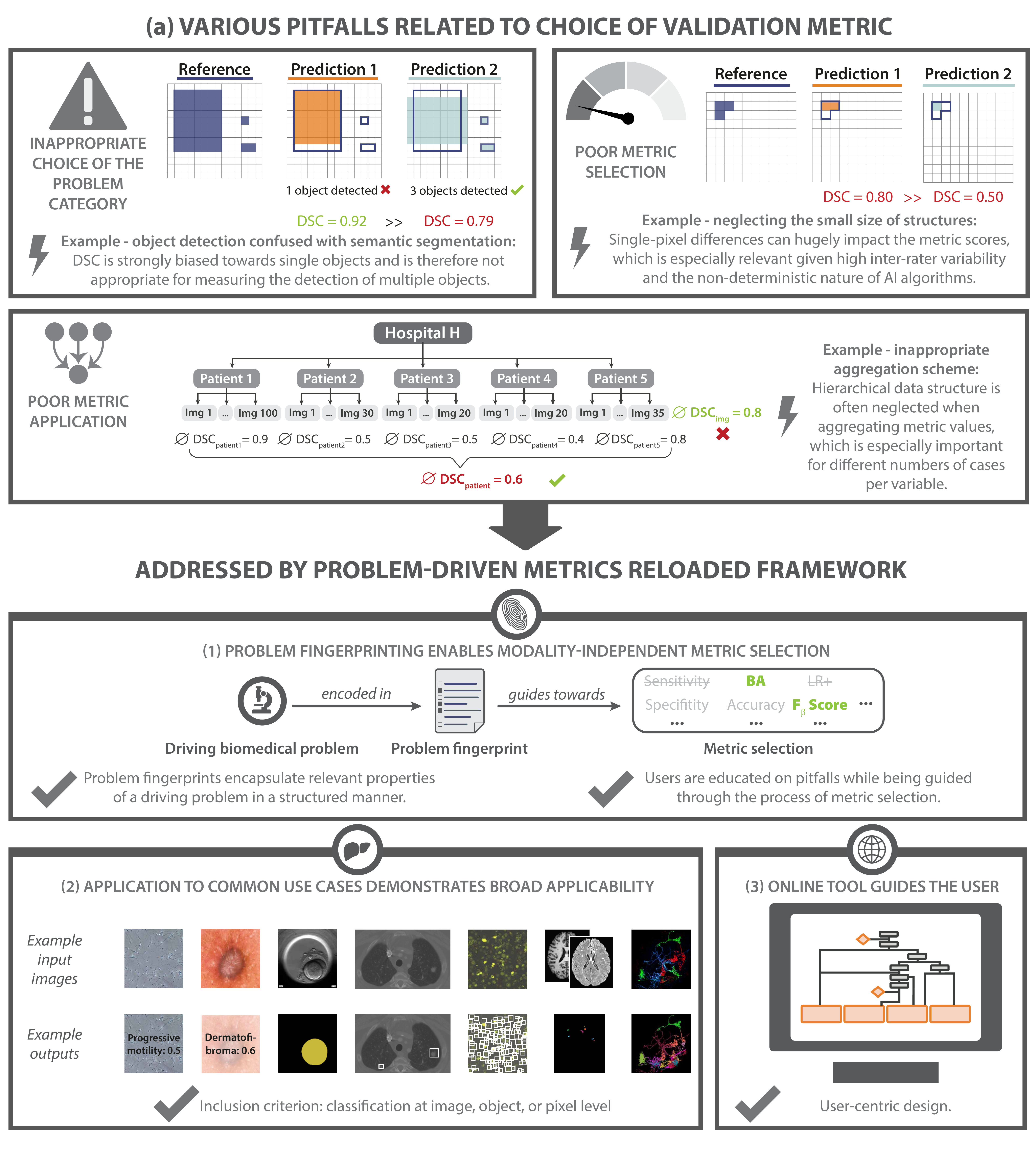}
    \caption{\textbf{Contributions of the \textit{Metrics Reloaded} framework.} \textbf{a)} Motivation: Common problems related to metrics typically arise from (top left) inappropriate choice of the problem category (here: object detection confused with semantic segmentation), (top right) poor metric selection (here: neglecting the small size of structures) and (bottom) poor metric application (here: inappropriate aggregation scheme). Pitfalls are highlighted by lightning bolts, $\varnothing$ refers to the average \textit{\ac{DSC}} values. Green metric values correspond to a good metric value, whereas red values correspond to a poor value. Green check marks indicate desirable behavior of metrics, red crosses indicate undesirable behavior. \textbf{b)} \textit{Metrics Reloaded} addresses these pitfalls. (1) To enable the selection of metrics that match the domain interest, the framework is based on the new concept of \textit{problem fingerprinting}, i.e., the generation of a structured representation of the given biomedical problem that captures all properties that are relevant for metric selection. Based on the problem fingerprint, \textit{Metrics Reloaded} guides the user through the process of metric selection and application while raising awareness of relevant pitfalls. (2) An instantiation of the framework for common biomedical use cases demonstrates its broad applicability. (3) A publicly available online tool facilitates application of the framework.}
    \label{fig:metrics_matter_overview}
\end{figure}

Despite the importance of metrics, an increasing body of work shows that the metrics used in common practice often do not adequately reflect the underlying biomedical problems, diminishing the validity of the investigated methods~\cite{honauer2015hci, correia2006video, kofler2021DICE, gooding2018comparative, vaassen2020evaluation, konukoglu2012discriminative, margolin2014evaluate, maier2018rankings, Tran2022}. This especially holds true for challenges, internationally respected competitions that over the last few years have become the de facto standard for comparative performance assessment of image processing methods. These challenges are often published in prestigious journals~\cite{chenouard2014objective, sage2015quantitative, ulman2017objective} and receive tremendous attention from both the scientific community and industry. Among a number of shortcomings in design and quality control that were recently unveiled by a multi-center initiative~\cite{maier2018rankings}, the choice of inappropriate metrics stood out as a core problem. 
Compared to other areas of AI research, choosing the right metric is particularly challenging in image processing because the suitability of a metric depends on various factors.
As a foundation for the present work, we identified three core categories related to pitfalls in metric selection (see Fig.~\ref{fig:metrics_matter_overview}a):
\begin{description}
\item[Inappropriate choice of the problem category:] The chosen metrics do not always reflect the biomedical need. For
example, object detection problems are often framed as segmentation tasks, resulting in the use of metrics that do not account for the potentially critical localization of all objects in the scene~\citep{carass2020evaluating,jager2020challenges} (Fig.~\ref{fig:metrics_matter_overview}a, top left).
\item[Poor metric selection:] Certain characteristics of a given biomedical problem render particular metrics inadequate. Mathematical metric properties are often neglected, for example, when using the \ac{DSC} in the presence of particularly small structures (Fig.~\ref{fig:metrics_matter_overview}a, top right). 
\item[Poor metric application:] Even if a metric is well-suited for a given problem in principle, pitfalls can occur when applying that metric to a specific data set. For example, a common flaw pertains to ignoring hierarchical data structure, as in data from multiple hospitals or a variable number of images per patient (Fig.~\ref{fig:metrics_matter_overview}a, bottom), when aggregating metric values. 
\end{description}
\newpage

\newpage
These problems are magnified by the fact that common practice often grows historically, and poor standards may be propagated between generations of scientists and in prominent publications.
To dismantle such historically grown poor practices and leverage distributed knowledge from various subfields of image processing, we established the multidisciplinary \textit{Metrics Reloaded}\footnote{We thank the Intelligent Medical Systems (IMSY) lab members Nina Sautter, Patricia Vieten and Tim Adler for the suggestion of the name, inspired by the \textit{Matrix} movies.} consortium. This consortium comprises international experts from the fields of medical image analysis, biological image analysis, medical guideline development, general ML, different medical disciplines, statistics and epidemiology, representing a large number of biomedical imaging initiatives and societies.

\textit{The mission of \textit{Metrics Reloaded} is to foster reliable algorithm validation through problem-aware, standardized choice of metrics with the long-term goal of (1) enabling the reliable tracking of scientific progress and (2) aiding to bridge the current chasm between \ac{ML} research and translation into biomedical imaging practice.}  

\begin{figure}[H]
    \centering
    \includegraphics[width=0.9\linewidth]{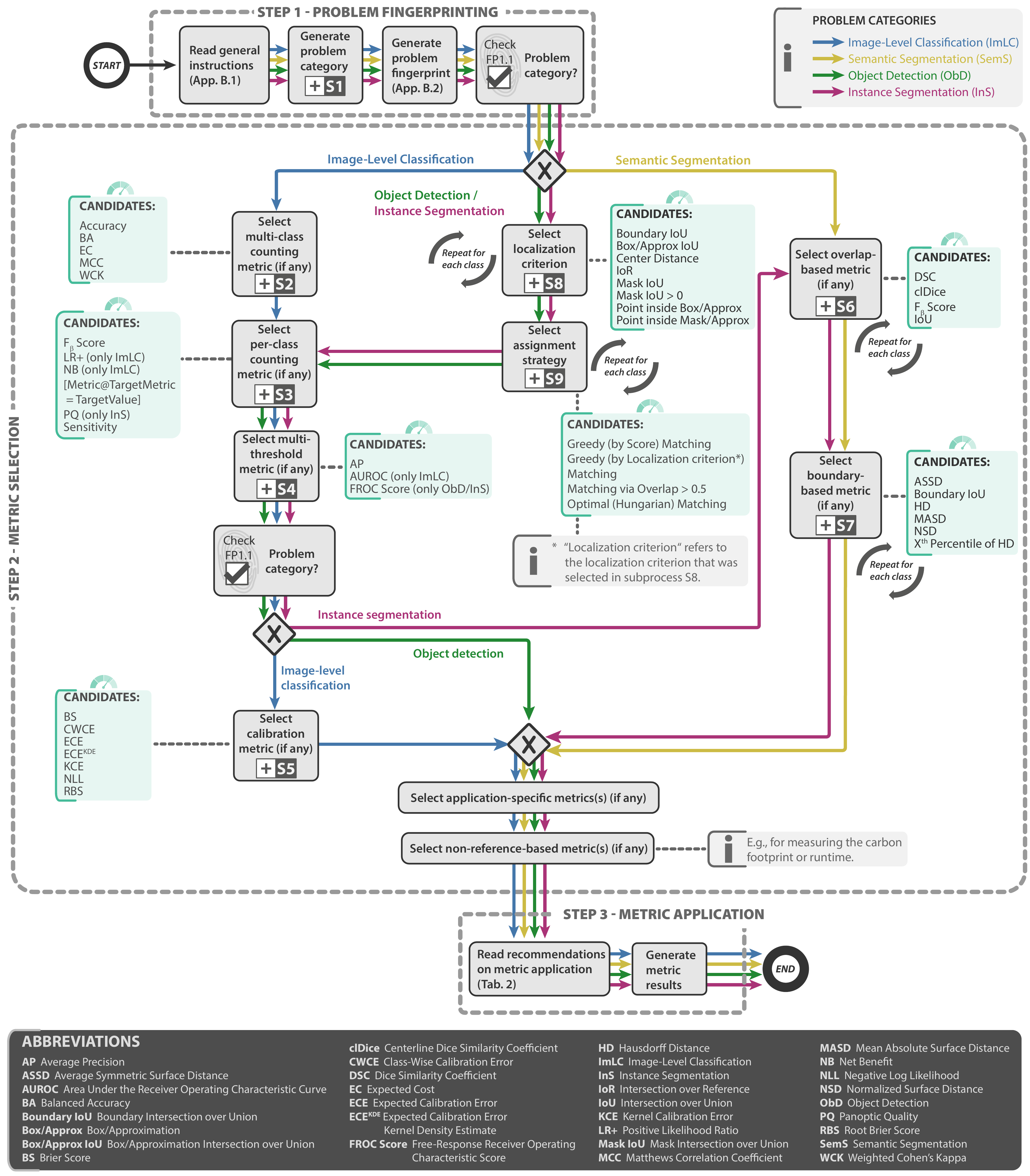}
    \caption{\textbf{\textit{Metrics Reloaded} recommendation framework from a user perspective.} In \textbf{step 1 - problem fingerprinting}, the given biomedical image analysis problem is mapped to the appropriate image \textit{problem category}, namely \textcolor{ILCblue}{\textit{image-level classification (ImLC)}}, \textcolor{SSyellow}{\textit{semantic segmentation (SemS)}},  \textcolor{ODgreen}{\textit{object detection (ObD)}}, or \textcolor{ISpink}{\textit{instance segmentation (InS)}} (Fig.~\ref{fig:problem_categories}). The problem category and further characteristics of the given biomedical problem relevant for metric selection are then captured in a \textit{problem fingerprint} (Fig.~\ref{fig:fingerprint}). In \textbf{step 2 - metric selection}, the user follows the respective coloured path of the chosen problem category (\textcolor{ILCblue}{ImLC $\rightarrow$}, \textcolor{SSyellow}{SemS $\rightarrow$},  \textcolor{ODgreen}{ObD $\rightarrow$}, or \textcolor{ISpink}{InS $\rightarrow$}) to select a suitable pool of metrics from the \textit{Metrics Reloaded} pools shown in green. When a tree branches, the fingerprint items determine which exact path to take. Finally, in \textbf{step 3 - metric application}, the user is supported in applying the metrics to a given data set. During the traversal of the decision tree, the user goes through \textit{subprocesses}, indicated by the $\boxplus$-symbol, which are provided in Extended Data Figs.~\ref{fig:category_mapping}~-~\ref{fig:metric-selection:s9-assignment-strategy} and represent relevant steps in the metric selection process. Ambiguities related to metric selection are resolved via \textit{decision guides} (Suppl. Note~\ref{app:decision_guides}) that help users make an educated decision when multiple options are possible. A comprehensive textual description of the recommendations for all four problem categories as well as for the selection of corresponding calibration metrics (if any) is provided in Suppl. Note~\ref{app:ilc} - Suppl. Note~\ref{app:calibration}. An overview of the symbols used in the process diagram is provided in Fig.~\ref{app:symbols}. Condensed versions of the mappings for every category can be found in Suppl. Note~\ref{app:ilc} for image-level classification, Suppl. Note~\ref{app:ss} for semantic segmentation, Suppl. Note~\ref{app:od} for object detection, and Suppl. Note~\ref{app:is} for instance segmentation.} 
    \label{fig:meta-mapping}
\end{figure}

\begin{figure}[H]
    \centering
    \includegraphics[width=1\linewidth]{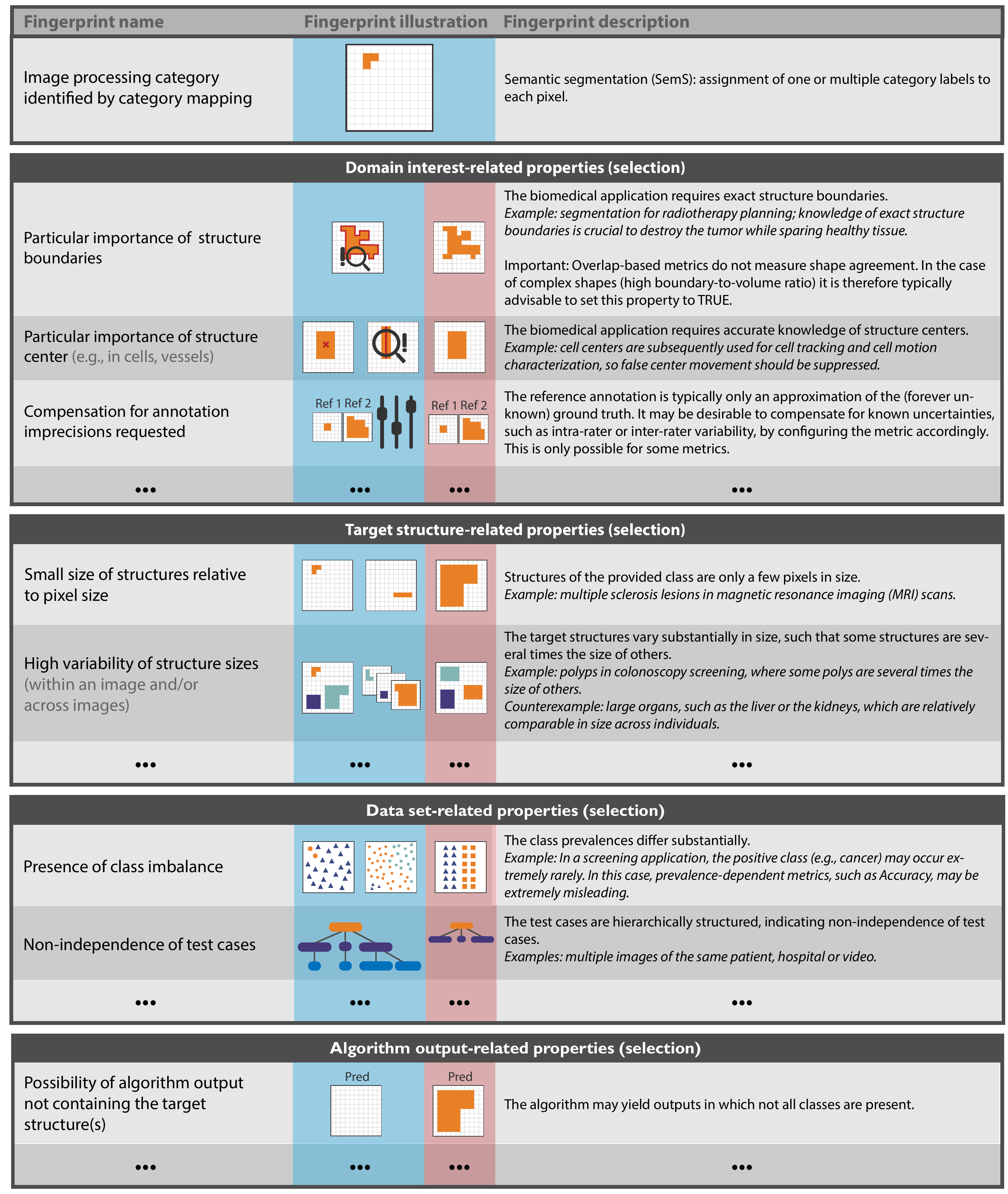}  \caption{\textbf{Relevant properties of a driving biomedical image analysis problem are captured by the \textit{problem fingerprint}} (selection for semantic segmentation shown here). The fingerprint comprises a set of items, each of which represents a specific property of the problem, is either binary or categorical, and must be instantiated by the user. Besides the problem category, the fingerprint comprises \textit{domain interest-related}, \textit{target structure-related}, \textit{data set-related} and  \textit{algorithm output-related} properties. A comprehensive version of the fingerprints for all problem categories can be found in Figs.~\ref{fig:fingerprint-ilc}-\ref{fig:fingerprint-ilc-3} (image-level classification), Figs.~\ref{fig:fingerprint-ss-1} /\ref{fig:fingerprint-ss-2}  (semantic segmentation), Figs.~\ref{fig:fingerprint-od-1}-\ref{fig:fingerprint-od-3} (object detection) and Figs.~\ref{fig:fingerprint-is-1}-\ref{fig:fingerprint-is-3} (instance segmentation). Used abbreviations: Prediction (Pred), Reference (Ref).}
    \label{fig:fingerprint}
\end{figure}

Based on a kickoff workshop held in December 2020, the \textit{Metrics Reloaded} framework (Fig.~\ref{fig:metrics_matter_overview}b and Fig.~\ref{fig:meta-mapping}) was developed using a multi-stage Delphi process \citep{brown1968delphi} for consensus building. Its primary purpose is to enable users to make educated decisions on which metrics to choose for a driving biomedical problem. The foundation of the metric selection process is the new concept of \textit{problem fingerprinting} (Fig.~\ref{fig:fingerprint}). Abstracting from a specific domain, problem fingerprinting is the generation of a structured representation of the given biomedical problem that captures all properties relevant for metric selection. As depicted in Fig.~\ref{fig:fingerprint}, the properties captured by the fingerprint comprise \textit{domain interest-related} properties, such as the particular importance of structure boundary, volume or center, \textit{target structure-related} properties, such as the shape complexity or the size of structures relative to the image grid size, \textit{data set-related} properties, such as class imbalance, as well as \textit{algorithm output-related} properties, such as the theoretical possibility of the algorithm output not containing any target structure. 

Based on the problem fingerprint, the user is then, in a transparent and understandable manner, guided through the process of selecting an appropriate set of metrics while being made aware of
potential pitfalls related to the specific characteristics of the underlying biomedical problem. The \textit{Metrics Reloaded} framework currently supports problems in which categorical target variables are to be predicted based on a given \textit{n}-dimensional input image (possibly enhanced with context information) at pixel, object or image level, as illustrated in Fig.~\ref{fig:problem_categories}. It thus supports problems that can be assigned to one of the following four \textit{problem categories}: \textit{image-level classification} (image level), \textit{object detection} (object level), \textit{semantic segmentation} (pixel level), or \textit{instance segmentation} (pixel level). Designed to be imaging modality-independent, \textit{Metrics Reloaded} can be suited for application in various image analysis domains even beyond the field of biomedicine.

Here, we present the key contributions of our work in detail, namely (1) the \textit{Metrics Reloaded} framework for problem-aware metric selection along with the key findings and design decisions that guided its development (Fig.~\ref{fig:meta-mapping}), (2) the application of the framework to common biomedical use cases, showcasing its broad applicability (selection shown in Fig.~\ref{fig:scenario-table}) and (3) the open online tool that has been implemented to improve the user experience with our framework.

%The following section~\ref{sec:framework} presents the general concept of the recommendation framework, while sections~\ref{sec:rec-classification}-\ref{sec:rec-is} present the specific recommendations for the four problem categories. Cross-topic recommendations are then summarized in section~\ref{sec:cross-topic}. We end with an introduction of the \textit{Metrics Reloaded tool} (section~\ref{sec:tool}) and a discussion of our work in the context of the state of the art (section~\ref{sec:discussion}). %before we end with a short summary of our work in section~\ref{sec:summary}. 

\newpage
%------------------------------------
\section*{Results}
\addcontentsline{toc}{chapter}{\protect\numberline{}Results}
\label{sec:results}
\subsection*{\textit{Metrics Reloaded} framework}
\begin{figure}[H]
    \centering
    \includegraphics[width=0.9\linewidth]{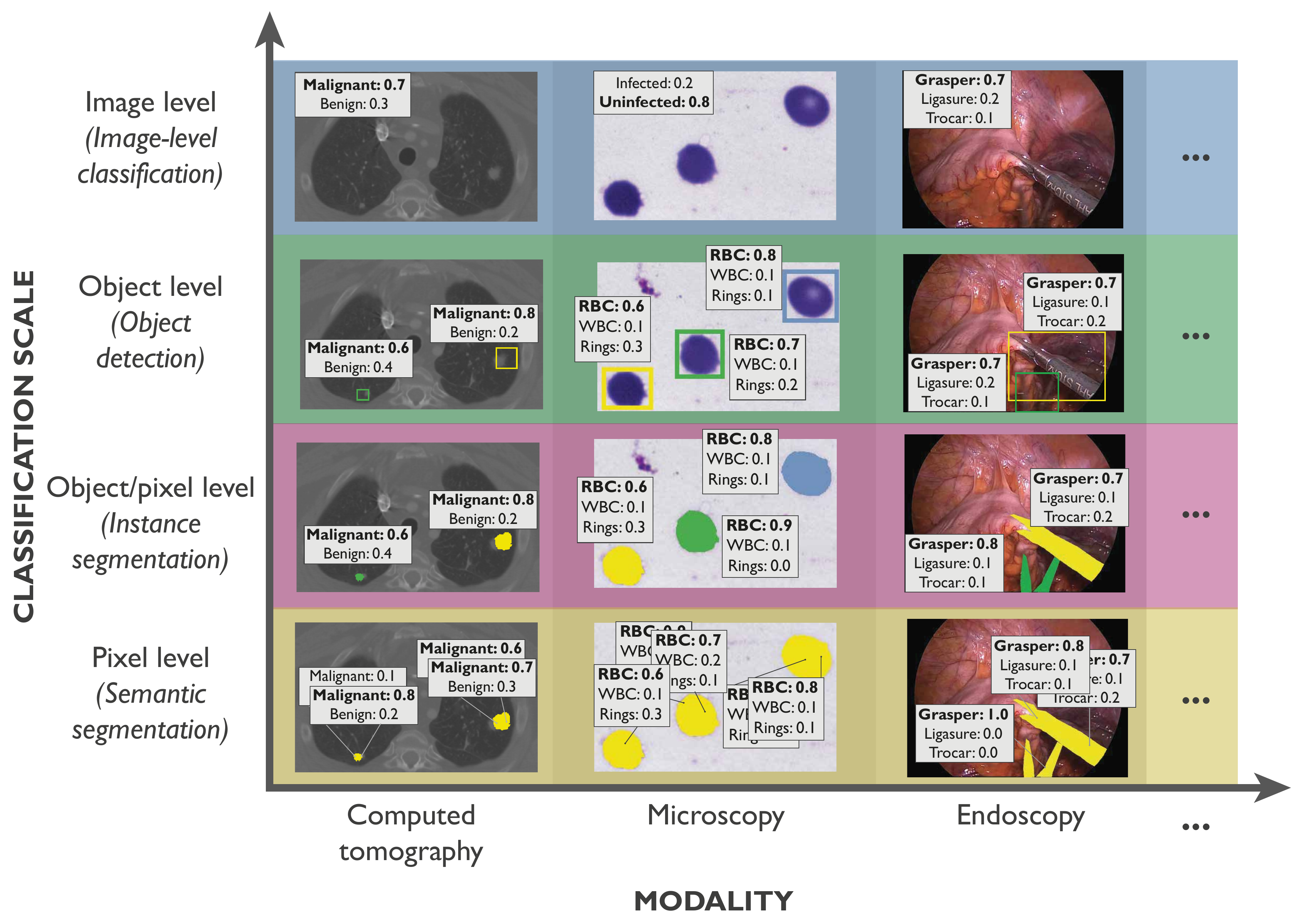}
    \caption{\textbf{\textit{Metrics Reloaded} fosters the convergence of validation methodology across modalities, application domains and classification scales.} The framework considers problems in which categorical target variables are to be predicted at image, object and/or pixel level, resulting (from top to bottom) in \textit{image-level classification}, \textit{object detection}, \textit{instance segmentation} or \textit{semantic segmentation}  problems. These problem categories are relevant across modalities (here \ac{CT}, microscopy and endoscopy) and application domains. From left to right: annotation of (left) benign and malignant lesions in \ac{CT} images \cite{armato2011lung}, (middle) different cell types in microscopy images \cite{ljosa2012annotated}, and (right) medical instruments in laparoscopy images \cite{maier2021heidelberg}.}
    \label{fig:problem_categories}
\end{figure}

\addcontentsline{toc}{section}{\protect\numberline{}Metrics Reloaded framework}
\label{sec:framework}
%------------------------------------
 \textit{Metrics Reloaded} is the result of a multi-stage Delphi process, comprising five international workshops, nine surveys, numerous expert group meetings, and crowdsourced feedback processes, all conducted between 2020 and 2022. 
 As a foundation of the recommendation framework, we identified common and rare pitfalls related to metrics in the field of biomedical image analysis using a community-powered process, detailed in this work's sister publication~\cite{reinke2021common}. We found that common practice is often not well-justified, and poor practices may even be propagated from one generation of scientists to the next. Importantly, many pitfalls generalize not only across the four problem categories that our framework addresses but also across domains (Fig.~\ref{fig:problem_categories}). This is because the source of the pitfall, such as class imbalance, uncertainties in the reference, or poor image resolution, can occur irrespective of a specific modality or application. 

  \textit{Following the convergence of \ac{AI} methodology across domains and problem categories, we therefore argue for the analogous convergence of validation methodology}.
%------------------------------------
%\subsection*{Image analysis pitfalls generalize across domains}
%\addcontentsline{toc}{section}{\protect\numberline{}Image analysis pitfalls generalize across domains}
%\label{ssec:generalize}
%------------------------------------

  %------------------------------------
%\subsection*{Historically grown practices are not always justified}
%\addcontentsline{toc}{section}{\protect\numberline{}Historically grown practices are not always justified}
%\label{ssec:historic}
%------------------------------------
 
\newpage
%------------------------------------
\subsection*{Cross-domain approach enables integration of distributed knowledge}
\addcontentsline{toc}{section}{\protect\numberline{}Cross-domain approach enables integration of distributed knowledge}
\label{ssec:results-delphi}
%------------------------------------
To break historically grown poor practices, we followed a multidisciplinary cross-domain approach that enabled us to critically question common practice in different communities and integrate distributed knowledge in one common framework. 
%To leverage distributed knowledge on metric pitfalls, strengths and weaknesses, and thus arrive at a "best of all worlds" solution, 
To this end, we formed an international multidisciplinary consortium of 73 experts from various biomedical image analysis-related fields. Furthermore, we crowdsourced metric pitfalls and feedback on our approach in a social media campaign. Ultimately, a total of 156 researchers contributed to this work, including 84 mentioned in the acknowledgements. Consideration of the different knowledge and perspectives on metrics led to the following key design decisions for \textit{Metrics Reloaded}:

\begin{description}
    \item[Encapsulating domain knowledge:] The questions asked to select a suitable metric are mostly similar regardless of image modality or application: Are the classes balanced? Is there a specific preference for the positive or negative class? What is the accuracy of the reference annotation? Is the structure boundary or volume of relevance for the target application? Importantly, \textit{while answering these questions requires domain expertise, the consequences in terms of metric selection can largely be regarded as domain-independent}. Our approach is thus to abstract from the specific image modality and domain of a given problem by capturing the properties relevant for metric selection in a \textit{problem fingerprint} (Fig.~\ref{fig:fingerprint}).
      \item[Exploiting synergies across classification scales:] Similar considerations apply with
      regard to metric choice for classification, detection and segmentation tasks, as they can all be regarded as classification tasks at different scales (Fig.~\ref{fig:problem_categories}). The similarities between the categories, however, can also lead to problems when the wrong category is chosen (see Fig.~\ref{fig:metrics_matter_overview}a, top left). Therefore, we (1) address all four problem categories in one common framework (Fig.~\ref{fig:meta-mapping}) and (2) cover the selection of the problem category itself in our framework (Fig.~\ref{fig:category_mapping}).
    \item[Setting new standards:] As the development and implementation of recommendations that go beyond the state of the art often requires critical mass, we involved stakeholders of various communities and societies in our consortium. Notably, out crowdsourcing-based approach led to a pool of metric candidates (Tab.~\ref{app:metric-pool}) that not only includes commonly applied metrics, but also metrics that have to date received little attention in biomedical image analysis. 
    \item[Abstracting from inference methodology:] Metrics should be chosen based solely on the driving biomedical problem and not be affected by algorithm design choices. For example, the error functions applied in common neural network architectures do not justify the use of corresponding metrics (e.g. validating with \ac{DSC} to match the Dice loss used for training a neural network). Instead, the domain interest should guide the choice of metric, which, in turn, can guide the choice of the loss term.
    \item[Exploiting complementary metric strengths:] A single metric typically cannot cover the complex requirements of the driving biomedical problem~\cite{reinke2018exploit}. To account for the complementary strengths and weakness of metrics, we generally recommend the usage of multiple complementary metrics to validate image analysis problems. As detailed in our recommendations (Suppl. Note~2), we specifically recommend the selection of metrics from different families.
         \item[Validation by consensus building and community feedback:] A major challenge for research on metrics is its validation, due to the lack of methods capable of quantitatively assessing the superiority of a given metric set over another. Following the spirit of large consortia formed to develop reporting guidelines (e.g., CONSORT ~\cite{schulz2010consort}, TRIPOD ~\cite{moons2015transparent}, STARD ~\cite{bossuyt2003towards}), we built the validation of our framework on three main pillars: (1) Delphi processes to challenge and refine the proposals of the expert groups that worked on individual components of the framework, (2) community feedback obtained by broadcasting the framework via society mailing lists and social media platforms and (3) and instantiation of the framework to a range of different biological and medical use cases. 
         \item[Involving and educating users:] Choosing adequate validation metrics is a complex process. Rather than providing a black box recommendation, \textit{Metrics Reloaded}  guides the user through the process of metric selection while raising awareness on pitfalls that may occur. In cases in which the tradeoffs between different choices must be considered, \textit{decision guides} (Suppl. Note~\ref{app:decision_guides}) assist in deciding between competing metrics while respecting individual preferences.
      \end{description} 
%------------------------------------
\subsection*{Problem fingerprints encapsulate relevant domain knowledge}
\addcontentsline{toc}{section}{\protect\numberline{}Problem fingerprints encapsulate relevant domain knowledge}
%Problem fingerprinting can capture relevant domain knowledge in a domain-independent manner
\label{ssec:fingerprinting}
%------------------------------------
To encapsulate relevant domain knowledge in a common format and then enable a modality-agnostic metric recommendation approach that generalizes over domains, we developed the concept of \textit{problem fingerprinting}, illustrated in Fig.~\ref{fig:fingerprint}. As a foundation, we crowdsourced all properties of a driving biomedical problem that are potentially relevant for metric selection via surveys issued to the consortium (see Suppl. Methods). This process resulted in a list of binary and categorical variables (\textit{fingerprint items}) that must be instantiated by a user to trigger the \textit{Metrics Reloaded} recommendation process. 
Common issues often relate to selecting metrics from the wrong problem category, as illustrated in Fig.~\ref{fig:metrics_matter_overview}a (top left). To avoid such issues, problem fingerprinting begins with mapping a given problem with all its intrinsic and data set-related properties to the corresponding problem category via the \textit{category mapping} shown in Fig.~\ref{fig:category_mapping}. The problem category is a fingerprint item itself.

In the following, we will refer to all fingerprint items with the notation FP\textit{X.Y}, where \textit{Y} is a numerical identifier and the index \textit{X} represents one of the following families:

\begin{description}
    \item [FP1 - Problem category] refers to the problem category generated by S1 (Extended Data Fig.~\ref{fig:category_mapping}).
    \item [FP2 - Domain interest-related properties] reflect user preferences and are highly dependent on the target application. A semantic image segmentation that serves as the foundation for radiotherapy planning, for example, would require exact contours (FP2.1 \textit{Particular importance of structure boundaries} = TRUE). On the other hand, for a cell segmentation problem that serves as prerequisite for cell tracking, the object centers may be much more important (FP2.3 = TRUE). Both problems could be tackled with identical network architectures, but the validation metrics should be different.
     \item [FP3 - Target structure-related properties] represent inherent properties of target structure(s) (if any), such as the size, size variability and the shape. Here, the term target structures can refer to any object/structure of interest, such as cells, vessels, medical instruments or tumors.
     \item [FP4 - Data set-related properties] capture properties inherent to the provided data to which the metric is applied. They primarily relate to class prevalences, uncertainties of the reference annotations, and whether the data structure is hierarchical.
     \item [FP5 - Algorithm output-related properties] encode properties of the output, such as the availability of predicted class scores.
\end{description} 

Note that not all properties are relevant for all problem categories. For example, the shape and size of target structures is highly relevant for segmentation problems but irrelevant for image classification problems. The complete problem category-specific fingerprints are provided in Suppl. Note~\ref{app:fingerprints}.
%Figs.~SN~2.7-\ref{fig:fingerprint-ilc-3} (image-level classification), Figs.~\ref{fig:fingerprint-ss-1} /\ref{fig:fingerprint-ss-2}  (semantic segmentation), Figs.~\ref{fig:fingerprint-od-1}-\ref{fig:fingerprint-od-3} (object detection) and Figs.~\ref{fig:fingerprint-is-1}-\ref{fig:fingerprint-is-3} (instance segmentation), respectively. Fingerprints are referred to in various parts of the manuscript including the decision trees~(Fig.~\ref{fig:meta-mapping} and Suppl. Note~2), the decision guides (Suppl. Note~2.7) and in the cross-category recommendations (Tab.~\ref{tab:cross-topic}~1).

%------------------------------------
\subsection*{\textit{Metrics Reloaded} addresses all three types of metric pitfalls}
\addcontentsline{toc}{section}{\protect\numberline{}Metrics Reloaded addresses all three types of metric pitfalls}
\label{ssec:metrics-reloaded-pitfalls}
%------------------------------------
\textit{Metrics Reloaded} was designed to address all three types of metric pitfalls identified in~\cite{reinke2021common} and illustrated in Fig.~\ref{fig:metrics_matter_overview}a. More specifically, each of the three steps shown in  Fig.~\ref{fig:meta-mapping} addresses one type of pitfall: 

 %------------------------------------
\subsubsection*{Step 1 - Fingerprinting}
\addcontentsline{toc}{subsection}{\protect\numberline{}Step 1 - Fingerprinting}
%------------------------------------
A user should begin by reading the general instructions of the recommendation framework, provided in Suppl. Note~\ref{app:general-instructions}. Next, the user should convert the driving biomedical problem to a problem fingerprint. This step is not only a prerequisite for applying the framework across application domains and classification scales, but also specifically addresses the \textit{inappropriate choice of the problem category} via the integrated category mapping. Once the user's domain knowledge has been encapsulated in the problem fingerprint, the actual metric selection is conducted according to a domain- and modality-agnostic process. 
 
 %------------------------------------
\subsubsection*{Step 2 - Metric Selection}
\addcontentsline{toc}{subsection}{\protect\numberline{}Step 2 - Metric Selection}
\label{sec:metric-selection}
%------------------------------------
  A Delphi process yielded the \textit{Metrics Reloaded} pool of reference-based validation metrics shown in Tab.~\ref{app:metric-pool}. Notable, this pool contains metrics that are currently not widely known in some biomedical image analysis communities. A prominent example is the \ac{NB}~\cite{vickers2016net} metric, popular in clinical prediction tasks and designed to determine whether basing decisions on a method would do more good than harm. A diagnostic test, for example, may lead to early identification and treatment of a disease, but typically will also cause a number of patients without disease being subjected to unnecessary further interventions.
 \ac{NB} allows to consider such tradeoffs by putting benefits and harms on the same scale so that they can be directly compared. Another example is the \ac{EC} metric~\cite{leeuwen2007introduction}, which can be seen as a generalization of Accuracy with many desirable added features, but is not well-known in the biomedical image analysis communities \citep{ferrer2022analysis}. Based on the \textit{Metrics Reloaded}  pool, the metric recommendation is performed with a \ac{BPMN}-inspired flowchart (see Fig.~\ref{app:symbols}), in which conditional operations are based on one or multiple fingerprint properties (Fig.~\ref{fig:meta-mapping}). The main flowchart has three substeps, each addressing the complementary strengths and weaknesses of common metrics. First, common \textit{reference-based metrics}, which are based on the comparison of the algorithm output to a reference annotation, are selected. Next, the pool of standard metrics can be complemented with custom metrics to address application-specific complementary properties.
Finally, non-reference-based metrics assessing speed, memory consumption or carbon footprint, for example, can be added to the metric pool(s). In this paper, we focus on the step of selecting reference-based metrics, because this is where synergies across modalities and scales can be exploited.

These synergies are showcased by the substantial overlap between the different paths that, depending on the problem category, are taken through the mapping during metric selection. All paths comprise several subprocesses $S$ (indicated by the $\boxplus$-symbol), each of which holds a subsidiary decision tree representing one specific step of the selection process. Traversal of a subprocess typically leads to the addition of a metric to the problem-specific metric pool. In multi-class prediction problems, dedicated metric pools for each class may need to be generated as relevant properties may differ from class to class. A three-dimensional semantic segmentation problem, for example, could require the simultaneous segmentation of both tubular and non-tubular structures (e.g., liver vessels and tissue). These require different metrics for validation. Although this is a corner case, our framework addresses this issue in principle. In ambiguous cases, i.e., when the user can choose between two options in one step of the decision tree, a corresponding \textit{decision guide} details the tradeoffs that need to be considered (Suppl. Note~\ref{app:decision_guides}). For example, the \ac{IoU} and the \ac{DSC} are mathematically closely related. The concrete choice typically boils down to a simple user or community preference. 

Fig.~\ref{fig:meta-mapping} along with the corresponding Subprocesses S1-S9 (Extended Data Figs.~\ref{fig:category_mapping}~-~\ref{fig:metric-selection:s9-assignment-strategy}) captures the core contribution of this paper, namely the consensus recommendation of the \textit{Metrics Reloaded} consortium according to the final Delphi process. For all ten components, the required Delphi consensus threshold (>75\% agreement) was met. In all cases of disagreement, which ranged from 0\% to 7\% for Fig.~\ref{fig:meta-mapping} and S1-S9, each remaining point of criticism was respectively only raised by a single person. The following paragraphs present a summary of the four different colored paths through Step 2 - Metric Selection of the recommendation tree~(Fig.~\ref{fig:meta-mapping}) for the task of selecting reference-based metrics from the \textit{Metrics Reloaded} pool of common metrics. More comprehensive textual descriptions can be found in \ref{app:metric-selection}.
%------------------------------------
\subsubsection*{\textcolor{ILCblue}{\textbf{Image-level Classification (ImLC)}}}
\label{sec:rec-classification}
%------------------------------------
Image-level classification is conceptually the most straightforward problem category, as the task is simply to assign one of multiple possible labels to an entire image (see Suppl. Note~\ref{app:ilc}). The validation metrics are designed to measure two key properties: \textit{discrimination} and \textit{calibration}.

\textit{Discrimination} refers to the ability of a classifier to discriminate between two or more classes. This can be achieved with \textit{counting metrics} that operate on the cardinalities of a fixed confusion matrix (i.e., the true/false positives/negatives in the binary classification case). Prominent examples are Sensitivity, Specificity or F$_1$ Score for binary settings and \ac{MCC} for multi-class settings. Converting predicted class scores to a fixed confusion matrix (in the binary case by setting a potentially arbitrary cutoff) can, however, be regarded as problematic in the context of performance assessment~\cite{reinke2021common}. \textit{Multi-threshold metrics}, such as \ac{AUROC}, are therefore based on varying the cutoff, which enables the explicit analysis of the tradeoff between competing properties such as Sensitivity and Specificity. 

While most research in biomedical image analysis focuses on the discrimination capabilities of classifiers, a complementary important property is the \textit{calibration} of a model. An uncertainty-aware model should yield predicted class scores that represent the true likelihood of events~\cite{gruber2022better}, as detailed in Suppl. Note~\ref{app:calibration}.
Overoptimistic or underoptimistic classifiers can be especially problematic in prediction tasks where a clinical decision may be made based on the risk of the patient of developing a certain condition. \textit{Metrics Reloaded} hence provides recommendations for validating the algorithm performance both in terms of discrimination and calibration. We recommend the following process for classification problems (\textcolor{ILCblue}{blue path} in Fig.~\ref{fig:meta-mapping}; detailed description in Suppl. Note~\ref{app:ilc}): 
\begin{description}
    \item[1: Select multi-class metric (if any):] Multi-class metrics have the unique advantage of capturing the performance of an algorithm for all classes in a single value. With the ability of taking into account all entries of the multi-class confusion matrix, they provide a holistic measure of performance without the need for customized class-aggregation schemes. We recommend using a multi-class metric if a decision rule applied to the predicted class scores is available (FP2.6). In certain use cases, especially in the presence of ordinal data, there is an unequal severity of class confusions (FP2.5.2), meaning that different costs should be applied to different misclassifications reflected by the confusion matrix. In such cases, we generally recommend \ac{EC} as metric. Otherwise, depending on the specific scenario, Accuracy, \ac{BA} and \ac{MCC} may be viable alternatives. The concrete choice of metric depends primarily on the prevalences (e.g. frequencies) of classes in the provided validation set and the target population (FP4.1/2), as detailed in Subprocess S2 (Extended Data Fig.~\ref{fig:metric-selection:s2-multi-class}) and the corresponding textual description in Suppl. Note~\ref{app:ilc}.  \\
    
    As class-specific analyses are not possible with multi-class metrics, which can potentially hide poor performance on individual classes, we recommend an additional validation with per-class counting metrics (optional) and multi-threshold metrics (always recommended).
    \item[2: Select per-class counting metric (if any):]  If a decision rule applied to the predicted class scores is available (FP2.6), a per-class counting metric, such as the $F_{\beta}$ Score, should be selected. Each class of interest is separately assessed, preferably in a "one-versus-rest" fashion. The choice depends primarily on the decision rule and the distribution of classes (FP4.2). Details can be found in Subprocess S3 for selecting per-class counting metrics (Extended Data Fig.~\ref{fig:metric-selection:s3-single-threshold}).
    \item[3: Select multi-threshold metric (if any):] 
    Counting metrics reduce the potentially complex output of a classifier (the continuous class scores) to a single value (the predicted class), such that they can work with a fixed confusion matrix. To compensate for this loss of information and obtain a more comprehensive picture of a classifier's discriminatory performance, multi-threshold metrics work with a dynamic confusion matrix reflecting a range of possible thresholds applied to the predicted class scores. While we recommend the popular, well-interpretable and prevalence-independent \ac{AUROC} as the default multi-threshold metric for classification, \ac{AP} can be more suitable in the case of high class balance because it incorporates predicted values, as detailed in Subprocess S4 for selecting multi-threshold metrics (Extended Data Fig.~\ref{fig:metric-selection:s4-multi-threshold}). 
     \item[4: Select calibration metric (if any):] If calibration assessment is requested (FP2.7), one or multiple calibration metrics should be added to the metric pool as detailed in Subprocess S5 for selecting calibration metrics (Extended Data Fig.~\ref{fig:metric-selection:s5-calibration}). 
     
\end{description}

%------------------------------------
\subsubsection*{\textcolor{SSyellow}{\textbf{Semantic segmentation (SemS)}}}
\label{sec:rec-ss}
%------------------------------------
In semantic segmentation, classification occurs at pixel level. However, it is not advisable to simply apply the standard classification metrics to the entire collection of pixels in a data set for two reasons. Firstly, pixels of the same image are highly correlated. Hence, to respect the hierarchical data structure, metric values should first be computed per image and then be aggregated over the set of images. Note in this context that the commonly used \ac{DSC} is mathematically identical to the popular F$_1$ Score applied at pixel level. Secondly, in segmentation problems, the user typically has an inherent interest in structure boundaries, centers or volumes of structures (FP2.1, FP2.2, FP2.3). The family of \textit{boundary-based metrics} (subset of \textit{distance-based metrics}) therefore requires the extraction of structure boundaries from the binary segmentation masks as a foundation for segmentation assessment. Based on these considerations and given all the complementary strengths and weaknesses of common segmentation metrics~\cite{reinke2021common}, we recommend the following process for segmentation problems (\textcolor{SSyellow}{yellow path} in Fig.~\ref{fig:meta-mapping}; detailed description in Suppl. Note~\ref{app:ss}): 

\begin{description}
    \item[1: Select overlap-based metric (if any):] In segmentation problems, counting metrics such as the \ac{DSC} or \ac{IoU} measure the overlap between the reference annotation and the algorithm prediction. As they can be considered the de facto standard for assessing segmentation quality and are well-interpretable, we recommend using them by default unless the target structures are consistently small, relative to the grid size (FP3.1), \textit{and} the reference may be noisy~(FP4.3.1). Depending on the specific properties of the problems, we recommend the \ac{DSC} or \ac{IoU} (default recommendation), the F$_\beta$ Score (preferred when there is a preference for either \ac{FP} or \ac{FN}) or the \ac{clDice} (for tubular structures). Details can be found in Subprocess S6 for selecting overlap-based metrics (Extended Data Fig.~\ref{fig:metric-selection:s6-overlap}).
     \item[2: Select boundary-based metric (if any):]
    Key weaknesses of overlap-based metrics include shape unawareness and limitations when dealing with small structures or high size variability~\cite{reinke2021common}. Our general recommendation is therefore to complement an overlap-based metric with a boundary-based metric. If annotation imprecisions should be compensated for (FP2.5.7), our default recommendation is the \ac{NSD}. Otherwise, the fundamental user preference guiding metric selection is whether errors should be penalized by existence or distance (FP2.5.6), as detailed in Subprocess S7 for selecting boundary-based metrics (Extended Data Fig.~\ref{fig:metric-selection:s7-boundary}). 
\end{description}

%------------------------------------
\subsubsection*{\textcolor{ODgreen}{\textbf{Object detection (ObD)}}}
\label{sec:rec-od}
%------------------------------------

Object detection problems differ from segmentation problems in several key features with respect to metric selection. Firstly, they involve distinguishing different instances of the same class and thus require the step of locating objects and assigning them to the corresponding reference object. Secondly, the granularity of localization is comparatively rough, which is why no boundary-based metrics are required (otherwise the problem would be phrased as an instance segmentation problem). Finally, and crucially important from a mathematical perspective, the absence of \acp{TN} in object detection problems renders many popular classification metrics (e.g. Accuracy, Specificity, \ac{AUROC}) invalid. In binary problems, for example, suitable counting metrics can only be based on three of the four entries of the confusion matrix. Based on these considerations and taking into account all the complementary strengths and weaknesses of existing metrics~\cite{reinke2021common}, we propose the following steps for object detection problems (\textcolor{ODgreen}{green path} in Fig.~\ref{fig:meta-mapping}; detailed description in Suppl. Note~\ref{app:od}):  

\begin{description}
    \item[1: Select localization criterion:] An essential part of the validation is to decide whether a prediction matches a reference object. To this end, (1) the location of both the reference objects and the predicted objects must be adequately represented (e.g., by masks, bounding boxes or center points), and (2) a metric for deciding on a match (e.g. Mask \ac{IoU}) must be chosen. As detailed in Subprocess S8 for selecting the localization criterion (Extended Data Fig.~\ref{fig:metric-selection:s8-localization-criterion}), our recommendation considers both the granularity of the provided reference (FP4.4) and the required granularity of the localization (FP2.4).
    
       \item[2: Select assignment strategy:] As the localization does not necessarily lead to unambiguous matchings, an assignment strategy needs to be chosen to potentially resolve ambiguities that occurred during localization. As detailed in Subprocess S9 for selecting the assignment strategy (Extended Data Fig.~\ref{fig:metric-selection:s9-assignment-strategy}), the recommended strategy depends on the availability of continuous class scores (FP5.1) as well as on whether double assignments should be punished (FP2.5.8).

    \item[3: Select classification metric(s) (if any):] Once objects have been located and assigned to reference objects, generation of a confusion matrix (without \ac{TN}) is possible. The final step therefore simply comprises choosing suitable classification metrics for validation.
   Several subfields of biomedical image analysis have converged to choosing solely a counting metric, such as the F$_\beta$ Score, as primary metric in object detection problems. We follow this recommendation when no continuous class scores are available for the detected objects (FP5.1). Otherwise, we disagree with the practice of basing performance assessment solely on a single, potentially suboptimal cutoff on the continuous class scores. Instead, we follow the recommendations for image-level classification and propose complementing a counting metric (Subprocess S3, Extended Data Fig.~\ref{fig:metric-selection:s3-single-threshold}) with a multi-threshold metric (Subprocess S4, Extended Data Fig.~\ref{fig:metric-selection:s4-multi-threshold}) to obtain a more holistic picture of performance. As multi-threshold metric, we recommend \ac{AP} or \ac{FROC} Score, depending on whether an easy interpretation (\ac{FROC} Score) or a standardized metric (\ac{AP}) is preferred. The choice of per-class counting metric depends primarily on the decision rule (FP2.6).

\end{description}

Note that the previous description implicitly assumed single-class problems, but generalization to multi-class problems is straightforward by applying the validation per-class. It is further worth mentioning that metric \textit{application} is not straightforward in object detection problems as the number of objects in an image may be extremely small, or even zero, compared to the number of pixels in an image. Special considerations with respect to aggregation must therefore be made, as detailed in Suppl. Note~\ref{app:od}. 

%------------------------------------
\subsubsection*{\textcolor{ISpink}{\textbf{Instance segmentation (InS)}}}
\label{sec:rec-is}
%------------------------------------

Instance segmentation delivers the tasks of object detection and semantic segmentation at the same time. Thus, the pitfalls and recommendations for instance segmentation problems are closely related to those for segmentation and object detection~\cite{reinke2021common}. This is directly reflected in our metric selection process (\textcolor{ISpink}{purple path} in Fig.~\ref{fig:meta-mapping}; detailed description in Suppl. Note~\ref{app:is}):

\begin{description}
    \item[1: Select object detection metric(s):] To overcome problems related to instance unawareness (Fig.~\ref{fig:metrics_matter_overview}a, top left), we recommend selection of a set of detection metrics to explicitly measure detection performance. To this end, we recommend almost the exact process as for object detection with two exceptions. Firstly, given the fine granularity of both the output and the reference annotation, our recommendation for the localization strategy differs, as detailed in Subprocess S8 (Extended Data Fig.~\ref{fig:metric-selection:s8-localization-criterion}). Secondly, as depicted in S3 (Extended Data Fig.~\ref{fig:metric-selection:s3-single-threshold}), we recommend the \ac{PQ} \cite{kirillov2019panoptic} as an alternative to the F$_\beta$ Score. This metric is especially suited for instance segmentation, as it combines the assessment of overall detection performance and segmentation quality of successfully matched (\ac{TP}) instances in a single score. 
    \item[2: Select segmentation metric(s) (if any):] In a second step, metrics to explicitly assess the segmentation quality for the \ac{TP} instances may be selected.  Here, we follow the exact same process as in semantic segmentation (Subprocesses S6, Extended Data Fig.~\ref{fig:metric-selection:s6-overlap} and S7, Extended Data Fig.~\ref{fig:metric-selection:s7-boundary}). The primary difference is that the segmentation metrics are applied per-instance.
\end{description} 

Importantly, the development process of the \textit{Metrics Reloaded} framework was designed such that the pitfalls identified in the sister publication of this work \cite{reinketizabi2022metrics} are comprehensively addressed. Tab.~\ref{tab:recommendations-pitfalls} makes the recommendations and design decisions corresponding to specific pitfalls explicit.

{\scriptsize
\begin{longtable}{p{3.0cm}p{10cm}}
%{p{2.0cm}p{1.1cm}p{1.8cm}cccp{1.4cm}cccc}
\caption{\textbf{\textit{Metrics Reloaded} addresses common and rare pitfalls in metric selection, as compiled in \cite{reinketizabi2022metrics}.} The first column lists all pitfall sources captured by the published taxonomy that relate to either the \textit{inadequate choice of the problem category} or \textit{poor metric selection}. The second column summarizes how \textit{Metrics Reloaded} addresses these pitfalls. The notation FP\textit{X.Y} refers to a fingerprint item (Suppl. Note~\ref{app:fingerprints}).}
\label{tab:recommendations-pitfalls}\\
\toprule
\textbf{Source of Pitfall} & \textbf{Addressed in Metrics Reloaded by} \\
\midrule 
\endfirsthead
%  Recommendations on metric implementation
\rowcolor[HTML]{FFFFFF}
\midrule
\multicolumn{2}{c}{\textbf{Inadequate choice of the problem category}} \\
\midrule
\rowcolor[HTML]{C0C0C0} 
Wrong choice of problem category & Problem category mapping (Subprocess S1, Extended Data Fig.~\ref{fig:problem_categories}) as a prerequisite for metric selection. \\
\rowcolor[HTML]{FFFFFF}
\midrule
\multicolumn{2}{c}{\textbf{Disregard of the domain interest}} \\
\midrule
\rowcolor[HTML]{C0C0C0} 
Importance of structure boundaries & FP2.1 -- Particular importance of structure boundaries; recommendation to complement common overlap-based segmentation metrics with boundary-based metrics (Fig.~\ref{fig:meta-mapping}, Suppl. Note~\ref{app:ss}) if the property holds.\\
 \rowcolor[HTML]{EFEFEF} 
Importance of structure volume & FP2.2 -- Particular importance of structure volume; recommendation to complement common overlap-based and boundary-based segmentation metrics with volume-based metrics (see Suppl. Note~\ref{app:ss}) if the property holds. \\
\rowcolor[HTML]{C0C0C0} 
Importance of structure center(line) & FP2.3 -- Particular importance of structure center(line); recommendation of the \acf{clDice} as alternative to the common \acf{DSC} or \acf{IoU} in segmentation problems (Subprocess S6, Extended Data Fig.~\ref{fig:metric-selection:s6-overlap}) and recommendation of center point-based localization criterion in object detection (Subprocess S8, Extended Data Fig.~\ref{fig:metric-selection:s8-localization-criterion}) if the property holds.     \\
 \rowcolor[HTML]{EFEFEF} 
Importance of confidence awareness & FP2.7.1 -- Calibration assessment requested; dedicated recommendations on calibration (Suppl. Note~\ref{app:calibration}). \\
 \rowcolor[HTML]{C0C0C0} 
Importance of comparability across data sets & FP4.2 -- Provided class prevalences reflect the population of interest; used in the Subprocesses S2-S4 (Extended Data Figs.~\ref{fig:metric-selection:s2-multi-class}-\ref{fig:metric-selection:s4-multi-threshold}); general focus on prevalence dependency of metrics in the framework.\\ % (Tab.~\ref{tab:metric-properties}).\\
\rowcolor[HTML]{EFEFEF} 
Unequal severity of class confusions & FP2.5 -- Penalization of errors; recommendation of the so far uncommon metric \ac{EC} as classification metric (Subprocess S2, Extended Data Fig.~\ref{fig:metric-selection:s2-multi-class}); setting $\beta$ in the F$_\beta$ Score according to preference for \ac{FP} (oversegmentation) and \ac{FN} (undersegmentation) (see DG3.3 in Suppl. Note~\ref{ssec:dg3}). \\
\rowcolor[HTML]{C0C0C0} 
Importance of cost-benefit-analysis & FP2.6 -- Decision rule applied to  predicted class scores: incorporation of a decision rule that is based on cost-benefit analysis; recommendation of the so far uncommon metrics \acf{NB} (Fig.~\ref{fig:cheat-sheet-nb}) and \acf{EC} (Fig.~\ref{fig:cheat-sheet-ec}).\\

\rowcolor[HTML]{FFFFFF}
\midrule
\multicolumn{2}{c}{\textbf{Disregard of target structure properties}} \\
\midrule
\rowcolor[HTML]{C0C0C0} 
Small structure sizes & FP3.1 -- Small size of structures relative to pixel size; recommendation to consider the problem an object detection problem (Suppl. Note~\ref{app:od}); complementation of overlap-based segmentation metrics with boundary-based metrics in the case of small structures with noisy reference (Subprocess S6, Extended Data Fig.~\ref{fig:metric-selection:s6-overlap}); recommendation of lower object detection localization threshold in case of small sizes (see DG8.3 in Suppl. Note~\ref{ssec:dg8}). \\
\rowcolor[HTML]{EFEFEF} 
High variability of structure sizes & FP3.2 -- High variability of structure sizes; recommendation of lower object detection localization threshold (see DG8.3 in Suppl. Note~\ref{ssec:dg8}) and size stratification (Suppl. Note~\ref{app:od}) in case of size variability. \\
\rowcolor[HTML]{C0C0C0} 
Complex structure shapes & FP3.3 -- Target structures feature tubular shape; recommendation of the \ac{clDice} as alternative to the common \ac{DSC} in segmentation problems (Subprocess S6, Extended Data Fig.~\ref{fig:metric-selection:s6-overlap}) and recommendation of Point inside Mask/Box/Approx as localization criterion in object detection if the property holds (Subprocess S8, Extended Data Fig.~\ref{fig:metric-selection:s8-localization-criterion}).\\
\rowcolor[HTML]{EFEFEF} 
Occurrence of overlapping or touching structures & FP3.5 -- Possibility of overlapping or touching target structures; explicit recommendation to phrase problem as instance segmentation rather than semantic segmentation problem (Suppl. Note~\ref{app:ss}); recommendation of higher object detection localization threshold in case of small sizes (see DG8.3 in Suppl. Note~\ref{ssec:dg8}).\\
\rowcolor[HTML]{C0C0C0} 
Occurrence of disconnected structures& FP3.6 -- Possibility of disconnected target structure(s); recommendation of appropriate localization criterion for object detection (DG8.2 in Suppl. Note~\ref{ssec:dg8}).\\

\rowcolor[HTML]{FFFFFF}
\midrule
\multicolumn{2}{c}{\textbf{Disregard of data set properties}} \\
\midrule
\rowcolor[HTML]{C0C0C0} 
High class imbalance &  FP4.1 -- High class imbalance and FP2.5.5 -- Compensation for class imbalances requested; compensation of class imbalance via prevalence-independent metrics such as \ac{EC} and \ac{BA}. \\
\rowcolor[HTML]{EFEFEF} 
Small test set size & Recommendation of confidence intervals for all metrics.\\
\rowcolor[HTML]{C0C0C0} 
Imperfect reference standard: Noisy reference standard & FP4.3.1 -- High inter-rater variability and FP2.5.7 -- Compensation for annotation imprecisions requested; default recommendation of the so far rather uncommon metric \acf{NSD} to assess the quality of boundaries.\\
\rowcolor[HTML]{EFEFEF} 
Imperfect reference standard: Spatial outliers in reference & FP4.3.2 -- Possibility of spatial outliers in reference annotation and FP2.5.6 -- Handling of spatial outliers; recommendation of outlier-robust metrics, such as \ac{NSD} in case no distance-based penalization of outliers is requested in segmentation problems.\\
\rowcolor[HTML]{C0C0C0} 
Occurrence of cases with an empty reference & FP4.6 -- Possibility of reference without target structure(s); recommendations for aggregation in the case of empty references according to Suppl. Note~\ref{app:od} and Extended Data Tab.~\ref{tab:cross-topic}.\\

\rowcolor[HTML]{FFFFFF}
\midrule
\multicolumn{2}{c}{\textbf{Disregard of algorithm output properties}} \\
\midrule
\rowcolor[HTML]{C0C0C0} 
Possibility of empty prediction &  FP5.2 -- Possibility of algorithm output not containing the target structure(s); selection of appropriate aggregation strategy in object detection (Suppl. Note~\ref{app:od}). \\
\rowcolor[HTML]{EFEFEF} 
Possibility of overlapping predictions & FP5.4 -- Possibility of overlapping predictions; recommendation of an assignment strategy based on \ac{IoU} > 0.5 if overlapping predictions are not possible and no predicted class scores are available. \\
\rowcolor[HTML]{C0C0C0} 
Lack of predicted class scores & FP5.1 -- Availability of predicted class scores; leveraging class scores for optimizing decision regions (FP2.6) and assessing calibration quality (FP2.7).\\

\bottomrule
\end{longtable}}
\normalsize
Once common reference-based metrics have been selected and, where necessary, complemented by application-specific metrics, the user proceeds with the application of the metrics to the given problem. 

\subsubsection*{Step 3 - Metric Application}
\addcontentsline{toc}{subsection}{\protect\numberline{}Step 3 - Metric Application}
\label{sec:cross-topic}

Although the application of a metric to a given data set may appear straightforward, numerous pitfalls can occur~\cite{reinke2021common}. Our recommendations for addressing them are provided in Extended Data Tab.~\ref{tab:cross-topic}~1. Following the taxonomy provided in the sister publication of this work \cite{reinketizabi2022metrics}, they are categorized in recommendations related to metric implementation, aggregation, ranking, interpretation, and reporting. While several aspects are covered in related work (e.g.~\cite{wiesenfarth2021methods}), an important contribution of the present work is the metric-specific summary of recommendations captured in the \textit{Metric Cheat Sheets} (Suppl. Note~\ref{app:metric-cheat-sheets}). A further major contribution is our implementation of all \textit{Metrics Reloaded} metrics in the open-source framework \ac{MONAI}, available at \url{https://github.com/Project-MONAI/MetricsReloaded} (see Suppl. Methods). 

%\newpage

\nocite{kang2006correlated, liang1986longitudinal, badgeley2019deep, khan2019non,summers2021nondeterminism, kottner2011guidelines, brummer2006application, steyerberg2010assessing, van2015spline, riley2016external}

% \newpage
% \input{CrossTopic_Table}

 %\newpage
%------------------------------------
\subsection*{\textit{Metrics Reloaded} is broadly applicable in biomedical image analysis}
\begin{figure}[H]
    \centering
    \includegraphics[width=1\linewidth]{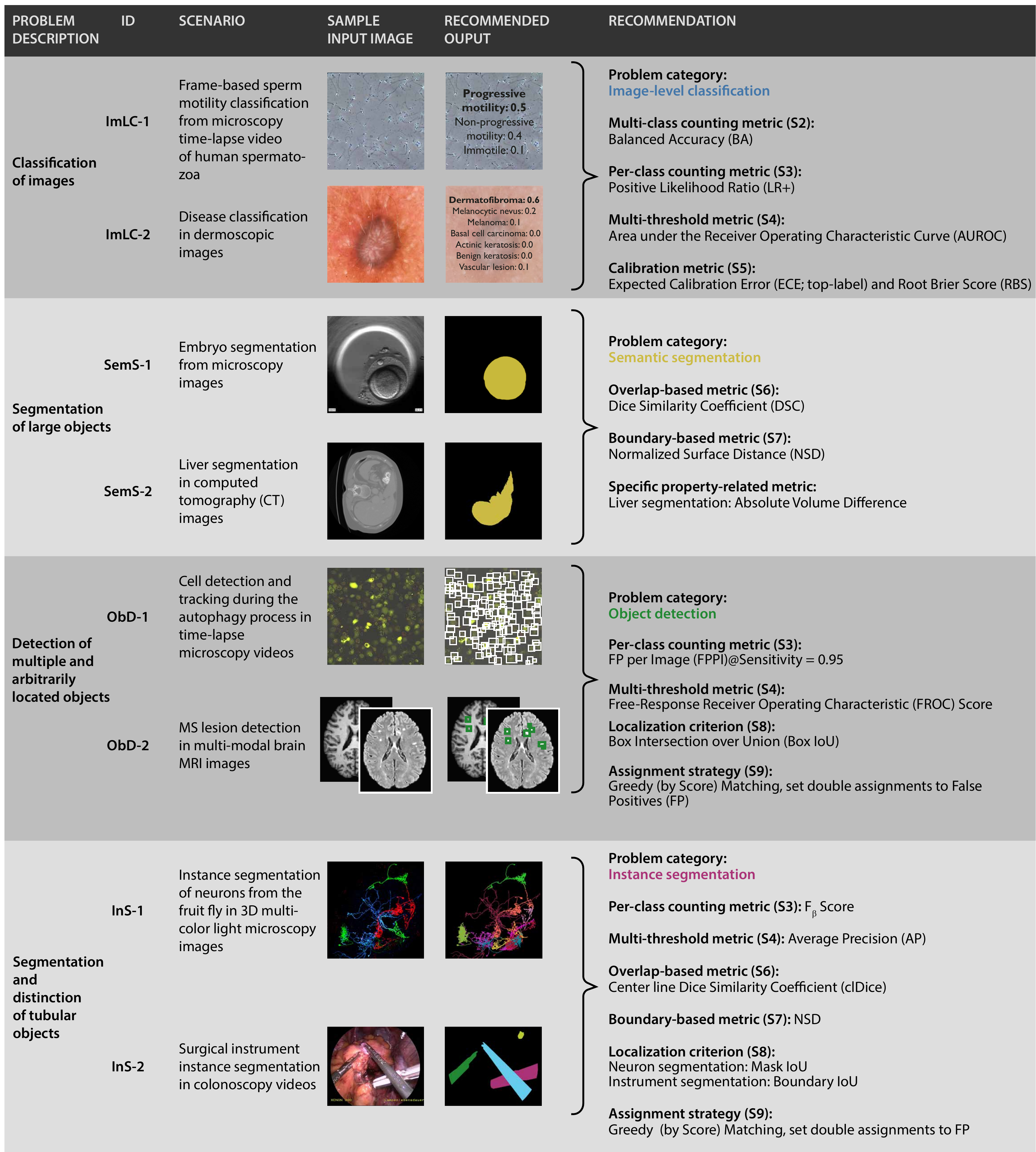}
    \caption{\textbf{Instantiation of the framework with recommendations for concrete biomedical questions.} From top to bottom: \textbf{(1)} Image classification for the examples of sperm motility classification \citep{haugen2019visem} and disease classification in dermoscopic images \citep{codella2019skin, dermoscopedia}. \textbf{(2)} Semantic segmentation of large objects for the examples of embryo segmentation from microscopy \citep{targosz2021semantic} and liver segmentation in \acf{CT} images \citep{antonelli2021medical, simpson2019large}. \textbf{(3)} Detection of multiple and arbitrarily located objects for the examples of cell detection and tracking during the autophagy process \citep{zhang2020deepphagy, nagao2020robust} and \ac{MS} lesion detection in multi-modal brain \ac{MRI} images \citep{kofler2022blob, commowick2018objective}. \textbf{(4)} Instance segmentation of tubular objects for the examples of instance segmentation of neurons from the fruit fly \citep{meissner2022searchable, tirian2017vt, mais2020patchperpix} and surgical instrument instance segmentation \citep{maier2021heidelberg}. The corresponding traversals through the decision trees are shown in\ref{app:scenarios}. An overview of the recommended metrics can be found in Suppl. Note~\ref{app:metric-cheat-sheets}, including relevant information for each metric.}
    \label{fig:scenario-table}
\end{figure}

\addcontentsline{toc}{section}{\protect\numberline{}Metrics Reloaded is broadly applicable in biomedical image analysis}
%------------------------------------
To validate the \textit{Metrics Reloaded} framework, we used it to generate recommendations for common use cases in biomedical image processing (see \ref{app:scenarios}). The traversal through the decision tree of our framework is detailed for eight selected use cases corresponding to the four different problem categories (Fig.~\ref{fig:scenario-table}):
\begin{description}
    \item[Image-level classification (Figs.~\ref{fig:scenario-s2}~-~\ref{fig:scenario-s5}):] frame-based sperm motility classification from time-lapse microscopy video of human spermatozoa (ImLC-1) and disease classification in dermoscopic images (ImLC-2).
    \item [Semantic segmentation (Figs.~\ref{fig:scenario-s6}~-~\ref{fig:scenario-s7}):] embryo segmentation in microscopy images (SemS-1) and liver segmentation in Computed Tomography (\ac{CT}) images (SemS-2).
    \item [Object detection (Figs.~\ref{fig:scenario-s3}~-~\ref{fig:scenario-s4}, \ref{fig:scenario-s8}~-~\ref{fig:scenario-s9}):] cell detection and tracking during the autophagy process in time-lapse microscopy (ObD-1) and \ac{MS} lesion detection in multi-modal brain \ac{MRI} images (ObD-2).
    \item[Instance segmentation (Figs.~\ref{fig:scenario-s3}~-~\ref{fig:scenario-s4}, \ref{fig:scenario-s6}~-~\ref{fig:scenario-s9}):] instance segmentation of neurons from the fruit fly in 3D multi-color light microscopy images (InS-1) and surgical instrument instance segmentation in colonoscopy videos (InS-2). 
\end{description}

The resulting metric recommendations (Fig.~\ref{fig:scenario-table}) demonstrate that a common framework across domains is sensible. In the showcased examples, shared properties of problems from different domains result in almost identical recommendations. In the semantic segmentation use cases, for example, the specific image modality is irrelevant for metric selection. What matters is the fact that a single object with a large size relative to the grid size should be segmented -- properties that are captured by the proposed fingerprint. In \ref{app:scenarios}, we present recommendations for several other biomedical use cases.

%------------------------------------
\subsection*{The \textit{Metrics Reloaded} online tool allows user-friendly metric selection}
\addcontentsline{toc}{section}{\protect\numberline{}The Metrics Reloaded online tool allows user-friendly metric selection}
%------------------------------------
Selecting appropriate validation metrics while considering all potential pitfalls that may occur is a highly complex process, as demonstrated by the large number of figures in this paper. Some of the complexity, however, also results from the fact that the figures need to capture all possibilities at once. For example, many of the figures could be simplified substantially for problems based on only two classes. To leverage this potential and to improve the general user experience with our framework, 
we developed the \textit{Metrics Reloaded} online tool, which is currently available as a beta version with restricted access (see Suppl. Methods). The tool captures our framework in a user-centric manner and can serve as a trustworthy common access point for image analysis validation.

\newpage

%------------------------------------
%\section*{Discussion}
%------------------------------------
%------------------------------------
\section*{Discussion}
\addcontentsline{toc}{chapter}{\protect\numberline{}Discussion}
%------------------------------------
Conventional scientific practice often grows through historical accretion, leading to standards that are not always well-justified. This holds particularly true for the validation standards in biomedical image analysis.

The present work represents the first comprehensive investigation and, importantly,  constructive set of recommendations challenging the state of the art in biomedical image analysis algorithm validation with a specific focus on metrics. With the intention of revisiting -- literally "re-searching" -- common validation practices and developing better standards, we brought together experts from traditionally disjunct fields to leverage distributed knowledge. Our international consortium of more than 70 experts from the fields of biomedical image analysis, machine learning, statistics, epidemiology, biology, and medicine, representing a large number of relevant biomedical imaging initiatives and societies, developed the \textit{Metrics Reloaded} framework that offers guidelines and tools to choose performance metrics in a problem-aware manner. The expert consortium was primarily compiled in a way to cover the required expertise from various fields but also consisted of researchers of different countries, (academic) ages, roles, and backgrounds (details can be found in the Methods). Importantly, \textit{Metrics Reloaded} comprehensively addresses all pitfalls related to metric selection (Tab.~\ref{tab:recommendations-pitfalls}) and application (Extended Data Tab.~\ref{tab:cross-topic}) that were identified in this work's sister publication~\cite{reinketizabi2022metrics}.

%------------------------------------
%\paragraph{Impact}
%------------------------------------
Today, the acceptance of a paper often relies on demonstrating a method's superior performance with respect to related work according to common validation metrics ~\cite{Birhane2021-zh}. Using appropriate validation metrics is not only fundamental for identifying new promising solutions and catalyzing scientific progress, but also key to the successful translation of \ac{ML} methods into practice. In fact, the importance of using metrics relevant to the end user has recently been prominently highlighted in the context of structured assessment of the (real-world) "readiness" of \ac{AI} technology at various points in its development cycle~\cite{lavin2022technology} as well as regulatory science enabling the clinical deployment of \ac{AI} ~\cite{lennerz2022unifying}. By significantly easing the choice of appropriate validation metrics, we expect the \textit{Metrics Reloaded} framework to increase the quality of research in the field of biomedical image analysis and catalyze faster initial translation of AI methods into (clinical) practice, as well as open up entirely new avenues of quality control and continuous performance monitoring of AI-based applications.

%------------------------------------
%\paragraph{Methodology}
%------------------------------------
%Inclusion criteria 
\textit{Metrics Reloaded} is the result of a 2.5-year long process involving numerous workshops, surveys, and expert group meetings. Many controversial debates were conducted during this time.
Even deciding on the exact scope of the paper was anything but trivial. Our consortium eventually agreed on focusing on biomedical classification problems with categorical reference data and thus exploiting synergies across classification scales. Generating and handling fuzzy reference data (e.g., from multiple observers) is a topic of its own~\cite{taha2015metrics, liu2021baseline} and was decided to be out of scope for this work. Furthermore, the inclusion of calibration metrics in addition to discrimination metrics was originally not intended because calibration is a complex topic in itself, and the corresponding field is relatively young and currently highly dynamic. This decision was reversed due to high demand from the community, expressed through crowdsourced feedback on the framework. 

Extensive discussions also evolved around the inclusion criteria for metrics, considering the tradeoff between established (potentially flawed) and new (not yet stress-tested) metrics. Our strategy for arriving at the \textit{Metrics Reloaded} recommendations balanced this tradeoff by using common metrics as a starting point and making adaptations where needed. For example, \ac{WCK}, originally designed for assessing inter-rater agreement, is the state-of-the-art metric used in the medical imaging community when handling ordinal data. Unlike other common multi-class metrics, such as (Balanced) Accuracy or \ac{MCC}, it allows the user to specify different costs for different class confusions, thereby addressing the ordinal rating. However, our consortium deemed the (not widely known) metric \ac{EC} generally more appropriate due to its favorable mathematical properties. Importantly, our framework does not intend to impose recommendations or act as a "black box" ; instead, it enables users to make educated decisions while considering ambiguities and tradeoffs that may occur. This is reflected by our use of \textit{decision guides} (Suppl. Note~\ref{app:decision_guides}), which actively involve users in the decision-making process (for the example above, for instance, see DG2.1). 

An important further challenge that our consortium faced was how to best provide recommendations in case multiple questions are asked for a single given data set. For example, a clinician's ultimate interest may lie in assessing whether tumor progress has occurred in a patient. While this would be phrased as an image-level classification task (given two images as input), an interesting \textit{surrogate task} could be seen in a segmentation task assessing the quality of tumor delineation and providing explainability for the results. \textit{Metrics Reloaded} addresses the general challenge of multiple different driving biomedical questions corresponding to one data set pragmatically by generating a recommendation separately for each question. The same holds true for multi-label problems, for example, when multiple different types of abnormalities potentially co-occur in the same image/patient.

%Validation
Another key challenge we faced was the validation of our framework due to the lack of ground truth "best metrics" to be applied for a given use case. Our solution builds upon three pillars. Firstly, we adopted established consensus building approaches utilized for developing widely used guidelines such as CONSORT ~\cite{schulz2010consort}, TRIPOD ~\cite{moons2015transparent}, or STARD ~\cite{bossuyt2003towards}). Secondly, we challenged our initial recommendation framework by acquiring feedback via a social media campaign. Finally, we instantiated the final framework to a range of different biological and medical use cases. Our approach showcases the benefit of crowdsourcing as a means of expanding the horizon beyond the knowledge peculiar to specific scientific communities. The most prominent change effected in response to the social media feedback was the inclusion of the aforementioned \ac{EC}, a powerful metric from the speech recognition community. Furthermore, upon popular demand, we added recommendations on assessing the interpretability of model outputs, now captured by Subprocess S5 (Extended Data Fig.~\ref{fig:metric-selection:s5-calibration}). 

After many highly controversial debates, the consortium ultimately converged on a consensus recommendation, as indicated by the high agreement in the final Delphi process (median agreement with the Subprocesses: 93\%). While some subsprocesses (S1, S7, S8) were unanimously agreed on without a single negative vote, several issues were raised by individual researchers. While most of them were minor (e.g., concerning wording), a major debate revolved around calibration metrics. Some members, for example, questioned the value of stand-alone calibration metrics altogether. The reason for this view is the critically important misconception that the predicted class scores of a well-calibrated model express the true posterior probability of an input belonging to a certain class~\cite{perez2022beyond} -- e.g., a patient's risk for a certain condition based on an image. As this is not the case, several researchers argued for basing calibration assessment solely on proper scoring rules (such as the \ac{BS}), which assess the quality of the posteriors better than the stand-alone calibration metrics. We have addressed all these considerations in our recommendation framework including a detailed rationale for our recommendations, provided in Suppl. Note~\ref{app:calibration}. 

While we believe our framework to cover the vast majority of biomedical image analysis use cases, suggesting a comprehensive set of metrics for every possible biomedical problem may be out of its scope. The focus of our framework lies in correcting poor practices related to the selection of common metrics. However, in some use cases, common reference-based metrics -- as a matter of principle -- be unsuitable. In fact, the use of application-specific metrics may be required in some cases. A prominent example are instance segmentation problems in which the matching of reference and predicted instances is infeasible, causing overlap-based localization criteria to fail. Metrics such as \ac{RI} \cite{rand1971objective} and \ac{VoI} \cite{meilua2003comparing} address this issue by avoiding one-to-one correspondences between predicted and reference instances. To make our framework applicable to such specific use cases, we integrated the step of choosing application-specific metrics in the main workflow (Fig.~\ref{fig:meta-mapping}). Examples of such application-specific metrics can be found in related work~\cite{ellis2021qualitative,cote2013tractometer}.

\textit{Metrics Reloaded} primarily provides guidance for the selection of metrics that measure some notion of the “correctness” of an algorithm's predictions on a set of test cases. It should be noted that holistic algorithm performance assessment also includes other aspects. One of them is robustness. For example, the accuracy of an algorithm for detecting disease in medical scans should ideally be the same across different hospitals that may use different acquisition protocols or scanners from different manufacturers. Recent work, however, shows that even the exact same models with nearly identical test set performance in terms of predictive accuracy may behave very differently on data from different distributions~\cite{DAmour2020-kh}. 
%This is highly relevant in medical imaging, where continuous changes can be expected in the data distribution due to population, acquisition, and prevalence shifts, all directly affecting the characteristics of the imaging data. 
%One way to assess robustness is to use stress testing, where the performance is monitored under simulated yet realistic perturbations of the image characteristics~\cite{Eche2021-wi}. 

Reliability is another important algorithmic property to be taken into account during validation. A reliable algorithm should have the ability to communicate its confidence and raise a flag when the uncertainty is high and the prediction should be discarded~\cite{Schulam2019-xo}. For calibrated models, this can be achieved via the predicted class scores, although other methods based on dedicated model outputs trained to express the confidence or on density estimation techniques are similarly popular. Importantly, an algorithm with reliable uncertainty estimates or increased robustness to distribution shift might not always be the best performing in terms of predictive performance~\cite{jaeger2022call}. For safe use of classification systems in practice, careful balancing of the tradeoff between robustness and reliability over accuracy might be necessary. 
%For example, an algorithm with well calibrated predictions could enable automatic failure detection and referral to human assessment for uncertain cases, providing an important safety net for clinical deployment. However, it may not be obvious how such ‘reject’ options should be quantified as part of the algorithmic performance assessment, as the algorithm may only be able to provide automated predictions for a subset of test samples. 

%Choosing the baseline methods for comparison and methods for evaluating the comparison (for example statistical significance) can also present an algorithm in a more favorable light, see Varoquaux and Cheplygina~\cite{Varoquaux2022-he} for a more detailed overview.

So far, \textit{Metrics Reloaded} focuses on common reference-based methods that compare model outputs to corresponding reference annotations. We made this design choice due to our hypothesis that reference-based metrics can be chosen in a modality- and application-agnostic manner using the concept of problem fingerprinting. As indicated by the step of choosing potential \textit{non-reference-based} metrics (Fig.~\ref{fig:meta-mapping}), however, it should be noted that validation and evaluation of algorithms should go far beyond purely technical performance~\cite{The_Institute_for_Ethical_Ai_undated-ed,declaration_montreal_ai}. In this context, Jannin introduced a global concept of “responsible research” to encompass all possible high-level assessment aspects of a digital technology~\cite{Jannin2021-uf}, including environmental, ethical, economical, social and societal aspects. For example, there are increasing efforts specifically devoted to the estimation of energy consumption and greenhouse gas emission of machine learning algorithms~\cite{Strubell2019-pw,Patterson2021-sw,lacoste2019quantifying}. For these considerations, we would like to point the reader to available tools such as the Green Algorithms calculator~\cite{lannelongue2021green} or Carbontracker~\cite{Wolff_Anthony2020-zl}.

It must further be noted that while \textit{Metrics Reloaded} places a focus on the \textit{selection} of metrics, adequate \textit{application} is also important. Detailed failure case analysis ~\cite{ross2021can} and performance assessment on relevant subgroups, for example, have been highlighted as critical components for better understanding when and where an algorithm may fail~\cite{Char2018-hs,Oakden-Rayner2020-co}. Given that learning-based algorithms rely on the availability of historical data sets for training, there is a real risk that any existing biases in the data may be picked up and replicated or even exacerbated when an algorithm makes predictions~\cite{Adamson2018-lr, Geirhos2020-rd}. This is of particular concern in the context of systemic biases in healthcare, such as the scarcity of representative data from underserved populations and often higher error rates in diagnostic labels in particular subgroups~\cite{Obermeyer2019-pp,Ibrahim2021-id}. Relevant meta information such as patient demographics, including biological sex and ethnicity, needs to be accessible for the test sets such that potentially disparate performance across subgroups can be detected~\cite{McCradden2022-kw}. Here, it is important to make use of adequate aggregations over the validation metrics as disparities in minority groups might otherwise be missed.

Finally, it must be noted that our framework addresses metric choice in the context of technical validation of biomedical algorithms. For translation of an algorithm into, for example, clinical routine, this validation may be followed by a (clinical) validation step assessing its performance compared to conventional, non-algorithm-based care according to patient-related outcome measures, such as overall survival \cite{park2023methods}. 

%However, we should note that adequate subgroup analysis in itself is far from trivial~\cite{McCradden2020-dw}. First, it may not be obvious or even possible to assign subgroup membership to individual samples due to ethical, statistical, and clinical reasons. In many cases, we may not even know what constitutes a meaningful subgroup. Second, any analysis of algorithmic bias cannot be decoupled from the issue that the biases already exist in the data sets. Performance analysis, however, relies on having access to well curated, high quality test data where the annotations are assumed to be largely correct and fair. This may not be the case when using historical data, and annotation bias, label noise, or the use of silver standards for obtaining a “pseudo ground truth” will directly affect the observed performance and impact any conclusions one may draw from the analysis.

%------------------------------------
%\paragraph{Evolution of Metrics Reloaded}
%------------------------------------
A key remaining challenge for \textit{Metric Reloaded}  is its dissemination such that it will substantially contribute to raising the quality of biomedical imaging research. To encourage widespread adherence to new standards, entry barriers should be as low as possible. 
While the framework with its vast number of subprocesses may seem very complex at first, it is important to note that from a user perspective only a fraction of the framework is relevant for a  given task, making the framework more tangible. This is notably illustrated by the \textit{Metric Reloaded} online tool, which substantially simplifies the metric selection procedure. 
%To further enhance usability, we are now working with a design team to implement the tool in the spirit of user-centric design principles.
As is common in scientific guideline and recommendation development, we intend to regularly update our framework to reflect current developments in the field, such as the inclusion of new metrics or biomedical use cases. This is intended to include an expansion of the framework's scope to further problem categories, such as regression and reconstruction. In order to accommodate future developments in a fast and efficient manner, we envision our consortium building consensus through accelerated Delphi rounds organized by the \textit{Metric Reloaded} core team. Once consensus is obtained, changes will be implemented in both the framework and online tool and highlighted so that users can easily identify changes to the previous version, which will ensure full transparency and comparability of results. In this way, we envision the \textit{Metrics Reloaded} framework and online tool as a dynamic resource reliably reflecting the current state of the art at any given time point in the future, for years to come. 

%Further/more broader future work
Of note, while the provided recommendations originate from the biomedical image analysis community, many aspects generalize to imaging research as a whole. Particularly the recommendations derived for individual fingerprints (e.g., implications of class imbalance) hold across domains, although it is possible that for different domains the existing fingerprints would need to be complemented by further features that this community is not aware of.

    %How to measure added benefit to clinicians

%------------------------------------
%\subsection*{Conclusion}
%------------------------------------
In conclusion, the \textit{Metrics Reloaded} framework provides biomedical image analysis researchers with the first systematic guidance on choosing validation metrics across different imaging tasks in a problem-aware manner. Through its reliance on methodology that can be generalized, we envision the \textit{Metrics Reloaded} framework to spark a scientific debate and hopefully lead to similar efforts being undertaken in other areas of imaging research, thereby raising research quality on a much larger scale than originally anticipated. In this context, our framework and the process by which it was developed could serve as a blueprint for broader efforts aimed at providing reliable recommendations and enforcing adherence to good practices in imaging research.

%In applying methodology that can be generalized to different \ac{AI} and imaging research domains, and showcasing the value of critically revisiting common validation practice in \ac{AI}, we envision our work to both raise the immediate quality of algorithm validation in biomedical imaging as well as inspire similar efforts of systematically improving validation practice in other domains.

%\input{OnlineMethods}

\section*{Code Availability Statement}
\addcontentsline{toc}{chapter}{\protect\numberline{}Code Availability Statement}
We provide reference implementations for all \textit{Metrics Reloaded} metrics within the \ac{MONAI} open-source framework. They are accessible at \url{https://github.com/Project-MONAI/MetricsReloaded}.

\section*{Acknowledgements}
This work was initiated by the Helmholtz Association of German Research Centers in the scope of the Helmholtz Imaging Incubator (HI), the MICCAI Special Interest Group on biomedical image analysis challenges and the benchmarking working group of the MONAI initiative. It received funding from the European Research Council (ERC) under the European Union’s Horizon 2020 research and innovation programme (grant agreement No. [101002198], NEURAL SPICING). It was further supported in part by the Intramural Research Program of the National Institutes of Health (NIH) Clinical Center as well as by the National Cancer Institute (NCI) and the National Institute of Neurological Disorders and Stroke (NINDS) of the NIH, under award numbers NCI:U01CA242871, NCI:U24CA279629, and NINDS:R01NS042645. The content of this publication is solely the responsibility of the authors and does not represent the official views of the NIH. T.A. acknowledges the Canada Institute for Advanced Research (CIFAR) AI Chairs program, the Natural Sciences and Engineering Research Council of Canada. F.B. was co-funded by the European Union (ERC, TAIPO, 101088594). Views and opinions expressed are however those of the authors only and do not necessarily reflect those of the European Union or the European Research Council. Neither the European Union nor the granting authority can be held responsible for them. V.C. acknowledges funding from NovoNordisk Foundation (NNF21OC0068816) and Independent Research Council Denmark (1134-00017B). B.A.C. was supported by NIH grant P41 GM135019 and grant 2020-225720 from the Chan Zuckerberg Initiative DAF, an advised fund of the Silicon Valley Community Foundation. G.S.C. was supported by Cancer Research UK (programme grant: C49297/A27294). M.M.H. is supported by the Natural Sciences and Engineering Research Council of Canada (RGPIN-2022-05134). A.Kara. is supported by French State Funds managed by the “Agence Nationale de la Recherche (ANR)” - “Investissements d’Avenir” (Investments for the Future), Grant ANR-10-IAHU-02 (IHU Strasbourg). M.K. was supported by the Ministry of Education, Youth and Sports of the Czech Republic (Project LM2018129). T.K. was supported in part by 4UH3-CA225021-03, 1U24CA180924-01A1, 3U24CA215109-02, and 1UG3-CA225-021-01 grants from the National Institutes of Health. G.L. receives research funding from the Dutch Research Council, the Dutch Cancer Association, HealthHolland, the European Research Council, the European Union, and the Innovative Medicine Initiative. C.H.S. is supported by an Alzheimer's Society Junior Fellowship (AS-JF-17-011). M.R is supported by Innosuisse grant number 31274.1 and Swiss National Science Foundation Grant Number 205320\_212939. R.M.S. is supported by the Intramural Research Program of the NIH Clinical Center. A.T. acknowledges support from Academy of Finland (Profi6 336449 funding program), University of Oulu strategic funding, Finnish Foundation for Cardiovascular Research, Wellbeing Services County of North Ostrobothnia (VTR project K62716), and Terttu foundation. S.A.T. acknowledges the support of Canon Medical and the Royal Academy of Engineering and the Research Chairs and Senior Research Fellowships scheme (grant RCSRF1819\textbackslash 8\textbackslash 25). B.V.C. was supported by Research Foundation Flanders (FWO grant G097322N) and Internal Funds KU Leuven (grant C24M/20/064).\

We thank Nina Sautter, Patricia Vieten and Tim Adler for proposing the name for the project. \

We would like to thank Peter Bankhead, Fred Hamprecht, Hannes Kenngott, David Moher, and Bram Stieltjes for fruitful discussions on the framework. \

We thank Susanne Steger for the data protection supervision and Anke Trotter for the hosting of the surveys. \

We would like to thank Lisa Mais for instantiating the use case for instance segmentation of neurons from the fruit fly in 3D multicolor light microscopy images. We would further like to thank the Janelia FlyLight Project Team for providing us with example images for this use case. \

We would like to thank the following people for testing the metric mappings, reviewing the recommendations and performing metric-centric testing: Tim Adler, Christoph Bender, Ahmad Bin Qasim, Kris Dreher, Niklas Holzwarth, Marco Hübner, Dominik Michael, Lucas-Raphael Müller, Maike Rees, Tom Rix, Melanie Schellenberg, Silvia Seidlitz, Jan Sellner, Akriti Srivastava, Fabian Wolf, Amine El Yamlahi, Silvia D. Almeida, Michael Baumgartner, Dimitrios Bounias, Till Bungert, Maximilian Fischer, Lukas Klein, Gregor Köhler, Bálint Kovács, Carsten Lueth, Tobias Norajitra, Constantin Ulrich, Tassilo Wald, Iuliia Alekseenko, Xiao Liu, Andrea Marheim Storås, Vajira Thambawita. \

We would like to thank the following people for taking our social media community survey and providing helpful feedback for improving the framework: Yamashita Akemi, Roi Anteby, Callum Arthurs, Pieter De Backer, Henry Badgery, Matthew Baugh, Jose Bernal, Matthew Blaschko, Dimitrios Bounias, Felipe Campos Kitamura, Jacob Carse, Chen Chen, Ivo Flipse, Nicolas Gaggion, Camila González, Pedro M. Gordaliza, Tim Horeman, Leo Joskowicz, Abin Jose, Amith Kamath, Brendan Kelly, Yannick Kirchhoff, Levin Arne Kobelke, Lars Krämer, Mira Krendel, John LaMaster, Thomas de Lange, Joël L. Lavanchy, Jianning Li, Carsten Lüth, Lisa Mais, Andrea Marheim Storås, Vishwesh Nath, Cian Scannell, Constantin Pape, M.P. Schijven, Alberto Selvanetti, Bella Specktor Fadida, Roger Staff, Jeremy Tan, Eric Tkaczyk, Rodrigo Tripodi Calumby, Athanasios Vlontzos, Weitong Zhang, Can Zhao, Jiayi Zhu. \

% We would further like to thank Niklas Holzwarth, Thijs Kooi, Gorkem Polat, Hassan Rivaz, Julien Schroeter, Anindo Saha and Ziv R. Yaniv for proposing metric pitfalls.

\addcontentsline{toc}{chapter}{\protect\numberline{}Acknowledgements}
% \input{Author-contributions}
% \addcontentsline{toc}{chapter}{\protect\numberline{}Author Contributions}
\section*{Competing Interests}
The authors declare the following competing interests: Under his terms of employment, M.B.B. is entitled to stock options in Mona.health, a KU Leuven spinoff. F.B. is an employee of Siemens AG (Munich, Germany). F.B. reports funding from Merck (Darmstadt, Germany). B.v.G. is a shareholder of Thirona (Nijmegen, NL). B.G. was an employee of HeartFlow Inc (California, USA) and Kheiron Medical Technologies Ltd (London, UK). M.M.H. received an Nvidia GPU Grant. B.K. is a consultant for ThinkSono Ldt (London, UK). G.L. is on the advisory board of Canon Healthcare IT (Minnetonka, USA) and is a shareholder of Aiosyn BV (Nijmegen, NL). N.R. is an employee of Nvidia GmbH (Munich, Germany). J.S.-R. reports funding from GSK (Heidelberg, Germany), Pfizer (New York, USA) and Sanofi (Paris, France) and fees from Travere Therapeutics (California, USA), Stadapharm (Bad Vilbel, Germany), Astex Therapeutics (Cambridge, UK), Pfizer (New York, USA), and Grunenthal (Aachen, Germany). R.M.S. receives patent royalties from iCAD (New Hampshire, USA), ScanMed (Nebraska, USA), Philips (Amsterdam, NL), Translation Holdings (Alabama, USA) and PingAn (Shenzhen, China); his lab received research support from PingAn through a Cooperative Research and Development Agreement. S.A.T. receives financial support from Canon Medical Research Europe (Edinburgh, Scotland).\\

\addcontentsline{toc}{chapter}{\protect\numberline{}Competing Interests}

%\bibliographystyle{ACM-Reference-Format}
\newpage
\renewcommand{\bibsection}{\section*{References}}
\bibliography{sample-base}

\begin{thebibliography}{175}
\providecommand{\natexlab}[1]{#1}
\providecommand{\url}[1]{\texttt{#1}}
\expandafter\ifx\csname urlstyle\endcsname\relax
  \providecommand{\doi}[1]{doi: #1}\else
  \providecommand{\doi}{doi: \begingroup \urlstyle{rm}\Url}\fi

\bibitem[Adamson and Smith(2018)]{Adamson2018-lr}
Adewole~S Adamson and Avery Smith.
\newblock Machine learning and health care disparities in dermatology, 2018.

\bibitem[Antonelli et~al.(2022)Antonelli, Reinke, Bakas, Farahani, Kopp-Schneider, Landman, Litjens, Menze, Ronneberger, Summers, et~al.]{antonelli2021medical}
Michela Antonelli, Annika Reinke, Spyridon Bakas, Keyvan Farahani, Annette Kopp-Schneider, Bennett~A Landman, Geert Litjens, Bjoern Menze, Olaf Ronneberger, Ronald~M Summers, et~al.
\newblock The medical segmentation decathlon.
\newblock \emph{Nature Communications}, 13\penalty0 (1):\penalty0 1--13, 2022.

\bibitem[Aresta et~al.(2019)Aresta, Ara{\'u}jo, Kwok, Chennamsetty, Safwan, Alex, Marami, Prastawa, Chan, Donovan, et~al.]{aresta2019bach}
Guilherme Aresta, Teresa Ara{\'u}jo, Scotty Kwok, Sai~Saketh Chennamsetty, Mohammed Safwan, Varghese Alex, Bahram Marami, Marcel Prastawa, Monica Chan, Michael Donovan, et~al.
\newblock Bach: Grand challenge on breast cancer histology images.
\newblock \emph{Medical image analysis}, 56:\penalty0 122--139, 2019.

\bibitem[Armato~III et~al.(2015)Armato~III, McLennan, Bidaut, McNitt-Gray, Meyer, Reeves, Zhao, Aberle, Henschke, Hoffman, et~al.]{armato2011lungdata}
S.~G. Armato~III, G.~McLennan, L.~Bidaut, M.~F. McNitt-Gray, C.~R. Meyer, A.~P. Reeves, B.~Zhao, D.~R. Aberle, C.~I. Henschke, E.~A. Hoffman, et~al.
\newblock Data from lidc-idri [data set].
\newblock \emph{The Cancer Imaging Archive}, 2015.

\bibitem[Armato~III et~al.(2011)Armato~III, McLennan, Bidaut, McNitt-Gray, Meyer, Reeves, Zhao, Aberle, Henschke, Hoffman, et~al.]{armato2011lung}
Samuel~G Armato~III, Geoffrey McLennan, Luc Bidaut, Michael~F McNitt-Gray, Charles~R Meyer, Anthony~P Reeves, Binsheng Zhao, Denise~R Aberle, Claudia~I Henschke, Eric~A Hoffman, et~al.
\newblock The lung image database consortium (lidc) and image database resource initiative (idri): a completed reference database of lung nodules on ct scans.
\newblock \emph{Medical physics}, 38\penalty0 (2):\penalty0 915--931, 2011.

\bibitem[Asgari~Taghanaki et~al.(2021)Asgari~Taghanaki, Abhishek, Cohen, Cohen-Adad, and Hamarneh]{asgari2021deep}
Saeid Asgari~Taghanaki, Kumar Abhishek, Joseph~Paul Cohen, Julien Cohen-Adad, and Ghassan Hamarneh.
\newblock Deep semantic segmentation of natural and medical images: a review.
\newblock \emph{Artificial Intelligence Review}, 54\penalty0 (1):\penalty0 137--178, 2021.

\bibitem[Attia(2003)]{wales2003moving}
John Attia.
\newblock Moving beyond sensitivity and specificity: using likelihood ratios to help interpret diagnostic tests.
\newblock \emph{Australian prescriber}, 26\penalty0 (5):\penalty0 111--113, 2003.

\bibitem[Aubreville et~al.(2022)Aubreville, Stathonikos, Bertram, Klopleisch, ter Hoeve, Ciompi, Wilm, Marzahl, Donovan, Maier, et~al.]{aubreville2022mitosis}
Marc Aubreville, Nikolas Stathonikos, Christof~A Bertram, Robert Klopleisch, Natalie ter Hoeve, Francesco Ciompi, Frauke Wilm, Christian Marzahl, Taryn~A Donovan, Andreas Maier, et~al.
\newblock Mitosis domain generalization in histopathology images--the midog challenge.
\newblock \emph{arXiv preprint arXiv:2204.03742}, 2022.

\bibitem[Badgeley et~al.(2019)Badgeley, Zech, Oakden-Rayner, Glicksberg, Liu, Gale, McConnell, Percha, Snyder, and Dudley]{badgeley2019deep}
MA~Badgeley, JR~Zech, L~Oakden-Rayner, BS~Glicksberg, M~Liu, W~Gale, MV~McConnell, B~Percha, TM~Snyder, and JT~Dudley.
\newblock Deep learning predicts hip fracture using confounding patient and healthcare variables. npj digit med. 2019; 2: 31, 2019.

\bibitem[Bandos et~al.(2009)Bandos, Rockette, Song, and Gur]{bandos2009area}
Andriy~I Bandos, Howard~E Rockette, Tao Song, and David Gur.
\newblock Area under the free-response roc curve (froc) and a related summary index.
\newblock \emph{Biometrics}, 65\penalty0 (1):\penalty0 247--256, 2009.

\bibitem[Baumgartner et~al.(2017)Baumgartner, Kamnitsas, Matthew, Fletcher, Smith, Koch, Kainz, and Rueckert]{baumgartner2017sononet}
Christian~F Baumgartner, Konstantinos Kamnitsas, Jacqueline Matthew, Tara~P Fletcher, Sandra Smith, Lisa~M Koch, Bernhard Kainz, and Daniel Rueckert.
\newblock Sononet: real-time detection and localisation of fetal standard scan planes in freehand ultrasound.
\newblock \emph{IEEE transactions on medical imaging}, 36\penalty0 (11):\penalty0 2204--2215, 2017.

\bibitem[Bejnordi et~al.(2017)Bejnordi, Veta, Van~Diest, Van~Ginneken, Karssemeijer, Litjens, Van Der~Laak, Hermsen, Manson, Balkenhol, et~al.]{bejnordi2017diagnostic}
Babak~Ehteshami Bejnordi, Mitko Veta, Paul~Johannes Van~Diest, Bram Van~Ginneken, Nico Karssemeijer, Geert Litjens, Jeroen~AWM Van Der~Laak, Meyke Hermsen, Quirine~F Manson, Maschenka Balkenhol, et~al.
\newblock Diagnostic assessment of deep learning algorithms for detection of lymph node metastases in women with breast cancer.
\newblock \emph{Jama}, 318\penalty0 (22):\penalty0 2199--2210, 2017.

\bibitem[Bene{\v{s}} and Zitov{\'a}(2015)]{benevs2015performance}
Miroslav Bene{\v{s}} and Barbara Zitov{\'a}.
\newblock Performance evaluation of image segmentation algorithms on microscopic image data.
\newblock \emph{Journal of microscopy}, 257\penalty0 (1):\penalty0 65--85, 2015.

\bibitem[Bernal et~al.(2019)Bernal, Histace, Masana, Angermann, S{\'a}nchez-Montes, Rodr{\'\i}guez~de Miguel, Hammami, Garc{\'\i}a-Rodr{\'\i}guez, C{\'o}rdova, Romain, et~al.]{bernal2019gtcreator}
Jorge Bernal, Aymeric Histace, Marc Masana, Quentin Angermann, Cristina S{\'a}nchez-Montes, Cristina Rodr{\'\i}guez~de Miguel, Maroua Hammami, Ana Garc{\'\i}a-Rodr{\'\i}guez, Henry C{\'o}rdova, Olivier Romain, et~al.
\newblock Gtcreator: a flexible annotation tool for image-based datasets.
\newblock \emph{International journal of computer assisted radiology and surgery}, 14\penalty0 (2):\penalty0 191--201, 2019.

\bibitem[Bernard et~al.(2018)Bernard, Lalande, Zotti, Cervenansky, Yang, Heng, Cetin, Lekadir, Camara, Ballester, et~al.]{bernard2018deep}
Olivier Bernard, Alain Lalande, Clement Zotti, Frederick Cervenansky, Xin Yang, Pheng-Ann Heng, Irem Cetin, Karim Lekadir, Oscar Camara, Miguel Angel~Gonzalez Ballester, et~al.
\newblock Deep learning techniques for automatic mri cardiac multi-structures segmentation and diagnosis: is the problem solved?
\newblock \emph{IEEE transactions on medical imaging}, 37\penalty0 (11):\penalty0 2514--2525, 2018.

\bibitem[Bickelhaupt et~al.(2018)Bickelhaupt, Jaeger, Laun, Lederer, Daniel, Kuder, Wuesthof, Paech, Bonekamp, Radbruch, et~al.]{bickelhaupt2018radiomics}
Sebastian Bickelhaupt, Paul~Ferdinand Jaeger, Frederik~Bernd Laun, Wolfgang Lederer, Heidi Daniel, Tristan~Anselm Kuder, Lorenz Wuesthof, Daniel Paech, David Bonekamp, Alexander Radbruch, et~al.
\newblock Radiomics based on adapted diffusion kurtosis imaging helps to clarify most mammographic findings suspicious for cancer.
\newblock \emph{Radiology}, 287\penalty0 (3):\penalty0 761--770, 2018.

\bibitem[Birhane et~al.(2021)Birhane, Kalluri, Card, Agnew, Dotan, and Bao]{Birhane2021-zh}
Abeba Birhane, Pratyusha Kalluri, Dallas Card, William Agnew, Ravit Dotan, and Michelle Bao.
\newblock The values encoded in machine learning research.
\newblock \emph{arXiv}, June 2021.

\bibitem[Bishop and Nasrabadi(2006)]{bishop2006pattern}
Christopher~M Bishop and Nasser~M Nasrabadi.
\newblock \emph{Pattern recognition and machine learning}, volume~4.
\newblock Springer, 2006.

\bibitem[Bossuyt et~al.(2003)Bossuyt, Reitsma, Bruns, Gatsonis, Glasziou, Irwig, Lijmer, Moher, Rennie, De~Vet, et~al.]{bossuyt2003towards}
Patrick~M Bossuyt, Johannes~B Reitsma, David~E Bruns, Constantine~A Gatsonis, Paul~P Glasziou, Les~M Irwig, Jeroen~G Lijmer, David Moher, Drummond Rennie, Henrica~CW De~Vet, et~al.
\newblock Towards complete and accurate reporting of studies of diagnostic accuracy: the stard initiative.
\newblock \emph{Annals of internal medicine}, 138\penalty0 (1):\penalty0 40--44, 2003.

\bibitem[Brier et~al.(1950)]{brier1950verification}
Glenn~W Brier et~al.
\newblock Verification of forecasts expressed in terms of probability.
\newblock \emph{Monthly weather review}, 78\penalty0 (1):\penalty0 1--3, 1950.

\bibitem[Brown(1968)]{brown1968delphi}
Bernice~B Brown.
\newblock Delphi process: a methodology used for the elicitation of opinions of experts.
\newblock Technical report, Rand Corp Santa Monica CA, 1968.

\bibitem[Br{\"u}mmer and Du~Preez(2006)]{brummer2006application}
Niko Br{\"u}mmer and Johan Du~Preez.
\newblock Application-independent evaluation of speaker detection.
\newblock \emph{Computer Speech \& Language}, 20\penalty0 (2-3):\penalty0 230--275, 2006.

\bibitem[Budd et~al.(2020)Budd, Patkee, Baburamani, Rutherford, Robinson, and Kainz]{budd2020surface}
Samuel Budd, Prachi Patkee, Ana Baburamani, Mary Rutherford, Emma~C Robinson, and Bernhard Kainz.
\newblock Surface agnostic metrics for cortical volume segmentation and regression.
\newblock In \emph{Machine Learning in Clinical Neuroimaging and Radiogenomics in Neuro-Oncology}, pages 3--12. Springer, 2020.

\bibitem[Caicedo et~al.(2019{\natexlab{a}})Caicedo, Goodman, Karhohs, Cimini, Ackerman, Haghighi, Heng, Becker, Doan, McQuin, et~al.]{caicedo2019nucleus}
Juan~C Caicedo, Allen Goodman, Kyle~W Karhohs, Beth~A Cimini, Jeanelle Ackerman, Marzieh Haghighi, CherKeng Heng, Tim Becker, Minh Doan, Claire McQuin, et~al.
\newblock Nucleus segmentation across imaging experiments: the 2018 data science bowl.
\newblock \emph{Nature methods}, 16\penalty0 (12):\penalty0 1247--1253, 2019{\natexlab{a}}.

\bibitem[Caicedo et~al.(2019{\natexlab{b}})Caicedo, Roth, Goodman, Becker, Karhohs, Broisin, Molnar, McQuin, Singh, Theis, et~al.]{caicedo2019evaluation}
Juan~C Caicedo, Jonathan Roth, Allen Goodman, Tim Becker, Kyle~W Karhohs, Matthieu Broisin, Csaba Molnar, Claire McQuin, Shantanu Singh, Fabian~J Theis, et~al.
\newblock Evaluation of deep learning strategies for nucleus segmentation in fluorescence images.
\newblock \emph{Cytometry Part A}, 95\penalty0 (9):\penalty0 952--965, 2019{\natexlab{b}}.

\bibitem[Carass et~al.(2020)Carass, Roy, Gherman, Reinhold, Jesson, Arbel, Maier, Handels, Ghafoorian, Platel, et~al.]{carass2020evaluating}
Aaron Carass, Snehashis Roy, Adrian Gherman, Jacob~C Reinhold, Andrew Jesson, Tal Arbel, Oskar Maier, Heinz Handels, Mohsen Ghafoorian, Bram Platel, et~al.
\newblock Evaluating white matter lesion segmentations with refined s{\o}rensen-dice analysis.
\newblock \emph{Scientific reports}, 10\penalty0 (1):\penalty0 1--19, 2020.

\bibitem[Char et~al.(2018)Char, Shah, and Magnus]{Char2018-hs}
Danton~S Char, Nigam~H Shah, and David Magnus.
\newblock Implementing machine learning in health care - addressing ethical challenges.
\newblock \emph{N. Engl. J. Med.}, 378\penalty0 (11):\penalty0 981--983, March 2018.

\bibitem[Cheng et~al.(2021)Cheng, Girshick, Doll{\'a}r, Berg, and Kirillov]{cheng2021boundary}
Bowen Cheng, Ross Girshick, Piotr Doll{\'a}r, Alexander~C Berg, and Alexander Kirillov.
\newblock Boundary iou: Improving object-centric image segmentation evaluation.
\newblock In \emph{Proceedings of the IEEE/CVF Conference on Computer Vision and Pattern Recognition}, pages 15334--15342, 2021.

\bibitem[Chenouard et~al.(2014)Chenouard, Smal, De~Chaumont, Ma{\v{s}}ka, Sbalzarini, Gong, Cardinale, Carthel, Coraluppi, Winter, et~al.]{chenouard2014objective}
Nicolas Chenouard, Ihor Smal, Fabrice De~Chaumont, Martin Ma{\v{s}}ka, Ivo~F Sbalzarini, Yuanhao Gong, Janick Cardinale, Craig Carthel, Stefano Coraluppi, Mark Winter, et~al.
\newblock Objective comparison of particle tracking methods.
\newblock \emph{Nature methods}, 11\penalty0 (3):\penalty0 281--289, 2014.

\bibitem[Chinchor(1992)]{Chinchor1992}
Nancy Chinchor.
\newblock Muc-4 evaluation metrics.
\newblock In \emph{Proceedings of the 4th Conference on Message Understanding}, MUC4 '92, page 22–29, USA, 1992. Association for Computational Linguistics.
\newblock ISBN 1558602739.
\newblock \doi{10.3115/1072064.1072067}.
\newblock URL \url{https://doi.org/10.3115/1072064.1072067}.

\bibitem[Clancy et~al.(2020)Clancy, Jones, Maier-Hein, Elson, and Stoyanov]{clancy2020surgical}
Neil~T Clancy, Geoffrey Jones, Lena Maier-Hein, Daniel~S Elson, and Danail Stoyanov.
\newblock Surgical spectral imaging.
\newblock \emph{Medical image analysis}, 63:\penalty0 101699, 2020.

\bibitem[Clark et~al.(2013)Clark, Vendt, Smith, Freymann, Kirby, Koppel, Moore, Phillips, Maffitt, Pringle, et~al.]{clark2013cancer}
Kenneth Clark, Bruce Vendt, Kirk Smith, John Freymann, Justin Kirby, Paul Koppel, Stephen Moore, Stanley Phillips, David Maffitt, Michael Pringle, et~al.
\newblock The cancer imaging archive (tcia): maintaining and operating a public information repository.
\newblock \emph{Journal of digital imaging}, 26\penalty0 (6):\penalty0 1045--1057, 2013.

\bibitem[Codella et~al.(2019)Codella, Rotemberg, Tschandl, Celebi, Dusza, Gutman, Helba, Kalloo, Liopyris, Marchetti, et~al.]{codella2019skin}
Noel Codella, Veronica Rotemberg, Philipp Tschandl, M~Emre Celebi, Stephen Dusza, David Gutman, Brian Helba, Aadi Kalloo, Konstantinos Liopyris, Michael Marchetti, et~al.
\newblock Skin lesion analysis toward melanoma detection 2018: A challenge hosted by the international skin imaging collaboration (isic).
\newblock \emph{arXiv preprint arXiv:1902.03368}, 2019.

\bibitem[Cohen(1960)]{cohen1960coefficient}
Jacob Cohen.
\newblock A coefficient of agreement for nominal scales.
\newblock \emph{Educational and psychological measurement}, 20\penalty0 (1):\penalty0 37--46, 1960.

\bibitem[Collins et~al.(2021)Collins, Dhiman, Navarro, Ma, Hooft, Reitsma, Logullo, Beam, Peng, Van~Calster, et~al.]{collins2021protocol}
Gary~S Collins, Paula Dhiman, Constanza L~Andaur Navarro, Jie Ma, Lotty Hooft, Johannes~B Reitsma, Patricia Logullo, Andrew~L Beam, Lily Peng, Ben Van~Calster, et~al.
\newblock Protocol for development of a reporting guideline (tripod-ai) and risk of bias tool (probast-ai) for diagnostic and prognostic prediction model studies based on artificial intelligence.
\newblock \emph{BMJ open}, 11\penalty0 (7):\penalty0 e048008, 2021.

\bibitem[Commowick et~al.(2018)Commowick, Istace, Kain, Laurent, Leray, Simon, Pop, Girard, Ameli, Ferr{\'e}, et~al.]{commowick2018objective}
Olivier Commowick, Audrey Istace, Michael Kain, Baptiste Laurent, Florent Leray, Mathieu Simon, Sorina~Camarasu Pop, Pascal Girard, Roxana Ameli, Jean-Christophe Ferr{\'e}, et~al.
\newblock Objective evaluation of multiple sclerosis lesion segmentation using a data management and processing infrastructure.
\newblock \emph{Scientific reports}, 8\penalty0 (1):\penalty0 1--17, 2018.

\bibitem[{CONSORT-AI and SPIRIT-AI Steering Group}(2019)]{CONSORT-AI_and_SPIRIT-AI_Steering_Group2019-om}
{CONSORT-AI and SPIRIT-AI Steering Group}.
\newblock Reporting guidelines for clinical trials evaluating artificial intelligence interventions are needed.
\newblock \emph{Nat. Med.}, 25\penalty0 (10):\penalty0 1467--1468, October 2019.

\bibitem[Correia and Pereira(2006)]{correia2006video}
Paulo Correia and Fernando Pereira.
\newblock Video object relevance metrics for overall segmentation quality evaluation.
\newblock \emph{EURASIP Journal on Advances in Signal Processing}, 2006:\penalty0 1--11, 2006.

\bibitem[Cybenko et~al.(1998)Cybenko, O'Leary, and Rissanen]{cybenko1998mathematics}
George Cybenko, Dianne~P O'Leary, and Jorma Rissanen.
\newblock \emph{The Mathematics of Information Coding, Extraction and Distribution}, volume 107.
\newblock Springer Science \& Business Media, 1998.

\bibitem[Côté et~al.(2013)Côté, Girard, Boré, Garyfallidis, Houde, and Descoteaux]{cote2013tractometer}
Marc-Alexandre Côté, Gabriel Girard, Arnaud Boré, Eleftherios Garyfallidis, Jean-Christophe Houde, and Maxime Descoteaux.
\newblock Tractometer: towards validation of tractography pipelines.
\newblock \emph{Medical Image Analysis}, 17\penalty0 (7):\penalty0 844--857, October 2013.
\newblock ISSN 1361-8423.
\newblock \doi{10.1016/j.media.2013.03.009}.

\bibitem[D'Amour et~al.(2020)D'Amour, Heller, Moldovan, Adlam, and {others}]{DAmour2020-kh}
A~D'Amour, K~Heller, D~Moldovan, B~Adlam, and {others}.
\newblock Underspecification presents challenges for credibility in modern machine learning.
\newblock \emph{arXiv preprint arXiv}, 2020.

\bibitem[Davis and Goadrich(2006)]{davis2006relationship}
Jesse Davis and Mark Goadrich.
\newblock The relationship between precision-recall and roc curves.
\newblock In \emph{Proceedings of the 23rd international conference on Machine learning}, pages 233--240, 2006.

\bibitem[de~Montréal(2017)]{declaration_montreal_ai}
Université de~Montréal.
\newblock The {Declaration} - {Montreal} {Responsible} {AI}, 2017.
\newblock URL \url{https://www.montrealdeclaration-responsibleai.com/the-declaration}.

\bibitem[DeGroot and Fienberg(1983)]{degroot1983comparison}
Morris~H DeGroot and Stephen~E Fienberg.
\newblock The comparison and evaluation of forecasters.
\newblock \emph{Journal of the Royal Statistical Society: Series D (The Statistician)}, 32\penalty0 (1-2):\penalty0 12--22, 1983.

\bibitem[Dice(1945)]{dice1945measures}
Lee~R Dice.
\newblock Measures of the amount of ecologic association between species.
\newblock \emph{Ecology}, 26\penalty0 (3):\penalty0 297--302, 1945.

\bibitem[Dolezal et~al.(2022)Dolezal, Srisuwananukorn, Karpeyev, Ramesh, Kochanny, Cody, Mansfield, Rakshit, Bansal, Bois, et~al.]{dolezal2022uncertainty}
James~M Dolezal, Andrew Srisuwananukorn, Dmitry Karpeyev, Siddhi Ramesh, Sara Kochanny, Brittany Cody, Aaron~S Mansfield, Sagar Rakshit, Radhika Bansal, Melanie~C Bois, et~al.
\newblock Uncertainty-informed deep learning models enable high-confidence predictions for digital histopathology.
\newblock \emph{Nature communications}, 13\penalty0 (1):\penalty0 6572, 2022.

\bibitem[Ellis et~al.(2021)Ellis, Alvarez, and Aizenberg]{ellis2021qualitative}
David~G Ellis, Carlos~M Alvarez, and Michele~R Aizenberg.
\newblock Qualitative criteria for feasible cranial implant designs.
\newblock In \emph{Cranial Implant Design Challenge}, pages 8--18. Springer, 2021.

\bibitem[Everingham et~al.(2006)Everingham, Zisserman, Williams, Van~Gool, Allan, Bishop, Chapelle, Dalal, Deselaers, Dork{\'o}, et~al.]{everingham20062005}
Mark Everingham, Andrew Zisserman, Christopher~KI Williams, Luc Van~Gool, Moray Allan, Christopher~M Bishop, Olivier Chapelle, Navneet Dalal, Thomas Deselaers, Gyuri Dork{\'o}, et~al.
\newblock The 2005 pascal visual object classes challenge.
\newblock In \emph{Machine Learning Challenges. Evaluating Predictive Uncertainty, Visual Object Classification, and Recognising Tectual Entailment: First PASCAL Machine Learning Challenges Workshop, MLCW 2005, Southampton, UK, April 11-13, 2005, Revised Selected Papers}, pages 117--176. Springer, 2006.

\bibitem[Everingham et~al.(2010)Everingham, Van~Gool, Williams, Winn, and Zisserman]{everingham2010pascal}
Mark Everingham, Luc Van~Gool, Christopher~KI Williams, John Winn, and Andrew Zisserman.
\newblock The pascal visual object classes (voc) challenge.
\newblock \emph{International journal of computer vision}, 88\penalty0 (2):\penalty0 303--338, 2010.

\bibitem[Everingham et~al.(2015)Everingham, Eslami, Van~Gool, Williams, Winn, and Zisserman]{everingham2015pascal}
Mark Everingham, SM~Ali Eslami, Luc Van~Gool, Christopher~KI Williams, John Winn, and Andrew Zisserman.
\newblock The pascal visual object classes challenge: A retrospective.
\newblock \emph{International journal of computer vision}, 111\penalty0 (1):\penalty0 98--136, 2015.

\bibitem[Ferrer(2022)]{ferrer2022analysis}
Luciana Ferrer.
\newblock Analysis and comparison of classification metrics.
\newblock \emph{arXiv preprint arXiv:2209.05355}, 2022.
\newblock The document discusses common performance metrics used in machine learning classification, and introduces the expected cost (EC) metric. It compares these metrics and argues that EC is superior due to its generality, simplicity, and intuitive nature. Additionally, it highlights the potential of EC in measuring calibration and optimal decision-making using class posteriors.

\bibitem[Geirhos et~al.(2020)Geirhos, Jacobsen, Michaelis, Zemel, Brendel, Bethge, and Wichmann]{Geirhos2020-rd}
Robert Geirhos, J{\"o}rn-Henrik Jacobsen, Claudio Michaelis, Richard Zemel, Wieland Brendel, Matthias Bethge, and Felix~A Wichmann.
\newblock Shortcut learning in deep neural networks.
\newblock \emph{Nature Machine Intelligence}, 2\penalty0 (11):\penalty0 665--673, November 2020.

\bibitem[Gneiting and Raftery(2007)]{gneiting2007strictly}
Tilmann Gneiting and Adrian~E Raftery.
\newblock Strictly proper scoring rules, prediction, and estimation.
\newblock \emph{Journal of the American statistical Association}, 102\penalty0 (477):\penalty0 359--378, 2007.

\bibitem[Gooding et~al.(2018)Gooding, Smith, Tariq, Aljabar, Peressutti, van~der Stoep, Reymen, Emans, Hattu, van Loon, et~al.]{gooding2018comparative}
Mark~J Gooding, Annamarie~J Smith, Maira Tariq, Paul Aljabar, Devis Peressutti, Judith van~der Stoep, Bart Reymen, Daisy Emans, Djoya Hattu, Judith van Loon, et~al.
\newblock Comparative evaluation of autocontouring in clinical practice: a practical method using the turing test.
\newblock \emph{Medical physics}, 45\penalty0 (11):\penalty0 5105--5115, 2018.

\bibitem[Grandini et~al.(2020)Grandini, Bagli, and Visani]{grandini2020metrics}
Margherita Grandini, Enrico Bagli, and Giorgio Visani.
\newblock Metrics for multi-class classification: an overview.
\newblock \emph{arXiv preprint arXiv:2008.05756}, 2020.

\bibitem[Gruber and Buettner(2022)]{gruber2022better}
Sebastian~Gregor Gruber and Florian Buettner.
\newblock Better uncertainty calibration via proper scores for classification and beyond.
\newblock In \emph{Advances in Neural Information Processing Systems}, 2022.

\bibitem[Guo et~al.(2017)Guo, Pleiss, Sun, and Weinberger]{guoCalibrationModernNeural2017}
Chuan Guo, Geoff Pleiss, Yu~Sun, and Kilian~Q Weinberger.
\newblock On {{Calibration}} of {{Modern Neural Networks}}.
\newblock \emph{ICML}, page~10, 2017.

\bibitem[Gupta et~al.(2020)Gupta, Rahimi, Ajanthan, Mensink, Sminchisescu, and Hartley]{gupta2020calibration}
Kartik Gupta, Amir Rahimi, Thalaiyasingam Ajanthan, Thomas Mensink, Cristian Sminchisescu, and Richard Hartley.
\newblock Calibration of neural networks using splines.
\newblock \emph{arXiv preprint arXiv:2006.12800}, 2020.

\bibitem[Gurcan et~al.(2010)Gurcan, Madabhushi, and Rajpoot]{gurcan2010pattern}
Metin~N Gurcan, Anant Madabhushi, and Nasir Rajpoot.
\newblock Pattern recognition in histopathological images: An icpr 2010 contest.
\newblock In \emph{International Conference on Pattern Recognition}, pages 226--234. Springer, 2010.

\bibitem[Hanley and McNeil(1982)]{hanley1982meaning}
James~A Hanley and Barbara~J McNeil.
\newblock The meaning and use of the area under a receiver operating characteristic (roc) curve.
\newblock \emph{Radiology}, 143\penalty0 (1):\penalty0 29--36, 1982.

\bibitem[Hastie et~al.(2009)Hastie, Tibshirani, Friedman, and Friedman]{hastie2009elements}
Trevor Hastie, Robert Tibshirani, Jerome~H Friedman, and Jerome~H Friedman.
\newblock \emph{The elements of statistical learning: data mining, inference, and prediction}, volume~2.
\newblock Springer, 2009.

\bibitem[Haugen et~al.(2019)Haugen, Hicks, Andersen, Witczak, Hammer, Borgli, Halvorsen, and Riegler]{haugen2019visem}
Trine~B Haugen, Steven~A Hicks, Jorunn~M Andersen, Oliwia Witczak, Hugo~L Hammer, Rune Borgli, P{\aa}l Halvorsen, and Michael Riegler.
\newblock Visem: A multimodal video dataset of human spermatozoa.
\newblock In \emph{Proceedings of the 10th ACM Multimedia Systems Conference}, pages 261--266, 2019.

\bibitem[Honauer et~al.(2015)Honauer, Maier-Hein, and Kondermann]{honauer2015hci}
Katrin Honauer, Lena Maier-Hein, and Daniel Kondermann.
\newblock The hci stereo metrics: Geometry-aware performance analysis of stereo algorithms.
\newblock In \emph{Proceedings of the IEEE International Conference on Computer Vision}, pages 2120--2128, 2015.

\bibitem[Huttenlocher et~al.(1993)Huttenlocher, Klanderman, and Rucklidge]{huttenlocher1993comparing}
Daniel~P Huttenlocher, Gregory~A. Klanderman, and William~J Rucklidge.
\newblock Comparing images using the hausdorff distance.
\newblock \emph{IEEE Transactions on pattern analysis and machine intelligence}, 15\penalty0 (9):\penalty0 850--863, 1993.

\bibitem[Ibrahim et~al.(2021)Ibrahim, Liu, Zariffa, Morris, and Denniston]{Ibrahim2021-id}
Hussein Ibrahim, Xiaoxuan Liu, Nevine Zariffa, Andrew~D Morris, and Alastair~K Denniston.
\newblock Health data poverty: an assailable barrier to equitable digital health care.
\newblock \emph{Lancet Digit Health}, 3\penalty0 (4):\penalty0 e260--e265, April 2021.

\bibitem[ISO~9000(2000)]{iso2000quality}
BSEN ISO~9000.
\newblock Quality management systems: Fundamentals and vocabulary.
\newblock \emph{London: British Standards Institution}, 2000.

\bibitem[Jaccard(1912)]{jaccard1912distribution}
Paul Jaccard.
\newblock The distribution of the flora in the alpine zone. 1.
\newblock \emph{New phytologist}, 11\penalty0 (2):\penalty0 37--50, 1912.

\bibitem[Jaeger et~al.(2023)Jaeger, L{\"u}th, Klein, and Bungert]{jaeger2022call}
Paul~F Jaeger, Carsten~T L{\"u}th, Lukas Klein, and Till~J Bungert.
\newblock A call to reflect on evaluation practices for failure detection in image classification.
\newblock \emph{International Conference on Learning Representations}, 2023.

\bibitem[J{\"a}ger(2020)]{jager2020challenges}
Paul~Ferdinand J{\"a}ger.
\newblock Challenges and opportunities of end-to-end learning in medical image classification.
\newblock \emph{Karlsruher Institut für Technologie}, 2020.

\bibitem[Jannin(2021)]{Jannin2021-uf}
Pierre Jannin.
\newblock Towards responsible research in digital technology for health care.
\newblock \emph{arXiv}, September 2021.

\bibitem[Jannin et~al.(2006)Jannin, Grova, and Maurer]{jannin2006model}
Pierre Jannin, Christophe Grova, and Calvin~R Maurer.
\newblock Model for defining and reporting reference-based validation protocols in medical image processing.
\newblock \emph{International Journal of Computer Assisted Radiology and Surgery}, 1\penalty0 (2):\penalty0 63--73, 2006.

\bibitem[Jin et~al.(2020)Jin, Yang, Kuang, Ni, Gao, Sun, Gao, Ma, Tan, Kang, et~al.]{jin2020deep}
Liang Jin, Jiancheng Yang, Kaiming Kuang, Bingbing Ni, Yiyi Gao, Yingli Sun, Pan Gao, Weiling Ma, Mingyu Tan, Hui Kang, et~al.
\newblock Deep-learning-assisted detection and segmentation of rib fractures from ct scans: Development and validation of fracnet.
\newblock \emph{EBioMedicine}, 62:\penalty0 103106, 2020.

\bibitem[Judge et~al.(2022)Judge, Bernard, Porumb, Chartsias, Beqiri, and Jodoin]{judge2022reliable}
Thierry Judge, Olivier Bernard, Mihaela Porumb, Agisilaos Chartsias, Arian Beqiri, and Pierre-Marc Jodoin.
\newblock -reliable uncertainty estimation for medical image segmentation.
\newblock In \emph{International Conference on Medical Image Computing and Computer-Assisted Intervention}, pages 492--502. Springer, 2022.

\bibitem[Kang et~al.(2006)Kang, Jin, and Sukthankar]{kang2006correlated}
Feng Kang, Rong Jin, and Rahul Sukthankar.
\newblock Correlated label propagation with application to multi-label learning.
\newblock In \emph{2006 IEEE Computer Society Conference on Computer Vision and Pattern Recognition (CVPR'06)}, volume~2, pages 1719--1726. IEEE, 2006.

\bibitem[Kelly et~al.(2019)Kelly, Karthikesalingam, Suleyman, Corrado, and King]{kelly2019key}
Christopher~J Kelly, Alan Karthikesalingam, Mustafa Suleyman, Greg Corrado, and Dominic King.
\newblock Key challenges for delivering clinical impact with artificial intelligence.
\newblock \emph{BMC medicine}, 17:\penalty0 1--9, 2019.

\bibitem[Khan et~al.(2019)Khan, Li, Sha, Liu, Jimenez, Raj, and Singh]{khan2019non}
Daanish~Ali Khan, Linhong Li, Ninghao Sha, Zhuoran Liu, Abelino Jimenez, Bhiksha Raj, and Rita Singh.
\newblock Non-determinism in neural networks for adversarial robustness.
\newblock \emph{arXiv preprint arXiv:1905.10906}, 2019.

\bibitem[Kirillov et~al.(2019)Kirillov, He, Girshick, Rother, and Doll{\'a}r]{kirillov2019panoptic}
Alexander Kirillov, Kaiming He, Ross Girshick, Carsten Rother, and Piotr Doll{\'a}r.
\newblock Panoptic segmentation.
\newblock In \emph{Proceedings of the IEEE/CVF Conference on Computer Vision and Pattern Recognition}, pages 9404--9413, 2019.

\bibitem[Kofler et~al.(2021)Kofler, Ezhov, Isensee, Berger, Korner, Paetzold, Li, Shit, McKinley, Bakas, et~al.]{kofler2021DICE}
Florian Kofler, Ivan Ezhov, Fabian Isensee, Christoph Berger, Maximilian Korner, Johannes Paetzold, Hongwei Li, Suprosanna Shit, Richard McKinley, Spyridon Bakas, et~al.
\newblock Are we using appropriate segmentation metrics? {I}dentifying correlates of human expert perception for {CNN} training beyond rolling the {DICE} coefficient.
\newblock \emph{arXiv preprint arXiv:2103.06205v1}, 2021.

\bibitem[Kofler et~al.(2022)Kofler, Shit, Ezhov, Fidon, Al-Maskari, Li, Bhatia, Loehr, Piraud, Erturk, et~al.]{kofler2022blob}
Florian Kofler, Suprosanna Shit, Ivan Ezhov, Lucas Fidon, Rami Al-Maskari, Hongwei Li, Harsharan Bhatia, Timo Loehr, Marie Piraud, Ali Erturk, et~al.
\newblock blob loss: instance imbalance aware loss functions for semantic segmentation.
\newblock \emph{arXiv preprint arXiv:2205.08209}, 2022.

\bibitem[Konukoglu et~al.(2012)Konukoglu, Glocker, Ye, Criminisi, and Pohl]{konukoglu2012discriminative}
Ender Konukoglu, Ben Glocker, Dong~Hye Ye, Antonio Criminisi, and Kilian~M Pohl.
\newblock Discriminative segmentation-based evaluation through shape dissimilarity.
\newblock \emph{IEEE transactions on medical imaging}, 31\penalty0 (12):\penalty0 2278--2289, 2012.

\bibitem[Kottner et~al.(2011)Kottner, Audig{\'e}, Brorson, Donner, Gajewski, Hr{\'o}bjartsson, Roberts, Shoukri, and Streiner]{kottner2011guidelines}
Jan Kottner, Laurent Audig{\'e}, Stig Brorson, Allan Donner, Byron~J Gajewski, Asbj{\o}rn Hr{\'o}bjartsson, Chris Roberts, Mohamed Shoukri, and David~L Streiner.
\newblock Guidelines for reporting reliability and agreement studies (grras) were proposed.
\newblock \emph{International journal of nursing studies}, 48\penalty0 (6):\penalty0 661--671, 2011.

\bibitem[Kuhn(1955)]{kuhn1955hungarian}
Harold~W Kuhn.
\newblock The hungarian method for the assignment problem.
\newblock \emph{Naval research logistics quarterly}, 2\penalty0 (1-2):\penalty0 83--97, 1955.

\bibitem[Kull et~al.(2019)Kull, Perello~Nieto, K{\"a}ngsepp, Silva~Filho, Song, and Flach]{kull2019beyond}
Meelis Kull, Miquel Perello~Nieto, Markus K{\"a}ngsepp, Telmo Silva~Filho, Hao Song, and Peter Flach.
\newblock Beyond temperature scaling: Obtaining well-calibrated multi-class probabilities with dirichlet calibration.
\newblock \emph{Advances in neural information processing systems}, 32, 2019.

\bibitem[Kumar et~al.(2019)Kumar, Liang, and Ma]{kumar2019verified}
Ananya Kumar, Percy~S Liang, and Tengyu Ma.
\newblock Verified uncertainty calibration.
\newblock \emph{Advances in Neural Information Processing Systems}, 32, 2019.

\bibitem[Kuppers et~al.(2020)Kuppers, Kronenberger, Shantia, and Haselhoff]{kuppers2020multivariate}
Fabian Kuppers, Jan Kronenberger, Amirhossein Shantia, and Anselm Haselhoff.
\newblock Multivariate confidence calibration for object detection.
\newblock In \emph{Proceedings of the IEEE/CVF Conference on Computer Vision and Pattern Recognition Workshops}, pages 326--327, 2020.

\bibitem[Lacoste et~al.(2019)Lacoste, Luccioni, Schmidt, and Dandres]{lacoste2019quantifying}
Alexandre Lacoste, Alexandra Luccioni, Victor Schmidt, and Thomas Dandres.
\newblock Quantifying the carbon emissions of machine learning.
\newblock \emph{arXiv preprint arXiv:1910.09700}, 2019.

\bibitem[Lannelongue et~al.(2021)Lannelongue, Grealey, and Inouye]{lannelongue2021green}
Lo{\"\i}c Lannelongue, Jason Grealey, and Michael Inouye.
\newblock Green algorithms: quantifying the carbon footprint of computation.
\newblock \emph{Advanced science}, 8\penalty0 (12):\penalty0 2100707, 2021.

\bibitem[Lavin et~al.(2022)Lavin, Gilligan-Lee, Visnjic, Ganju, Newman, Ganguly, Lange, Baydin, Sharma, Gibson, et~al.]{lavin2022technology}
Alexander Lavin, Ciar{\'a}n~M Gilligan-Lee, Alessya Visnjic, Siddha Ganju, Dava Newman, Sujoy Ganguly, Danny Lange, At{\'\i}l{\'\i}m~G{\"u}ne{\c{s}} Baydin, Amit Sharma, Adam Gibson, et~al.
\newblock Technology readiness levels for machine learning systems.
\newblock \emph{Nature Communications}, 13\penalty0 (1):\penalty0 1--19, 2022.

\bibitem[Le et~al.(2019)Le, Wang, Huang, Hickman, and Gilbert]{le2019artificial}
EPV Le, Y~Wang, Yuan Huang, Sarah Hickman, and FJ~Gilbert.
\newblock Artificial intelligence in breast imaging.
\newblock \emph{Clinical radiology}, 74\penalty0 (5):\penalty0 357--366, 2019.

\bibitem[Leeuwen and Br{\"u}mmer(2007)]{leeuwen2007introduction}
David A~van Leeuwen and Niko Br{\"u}mmer.
\newblock An introduction to application-independent evaluation of speaker recognition systems.
\newblock In \emph{Speaker classification I}, pages 330--353. Springer, 2007.

\bibitem[Lennerz et~al.(2022)Lennerz, Green, Williamson, and Mahmood]{lennerz2022unifying}
Jochen~K Lennerz, Ursula Green, Drew~FK Williamson, and Faisal Mahmood.
\newblock A unifying force for the realization of medical ai.
\newblock \emph{npj Digital Medicine}, 5\penalty0 (1):\penalty0 1--3, 2022.

\bibitem[Li et~al.(2022)Li, Kamnitsas, Islam, Chen, and Glocker]{li2022estimating}
Zeju Li, Konstantinos Kamnitsas, Mobarakol Islam, Chen Chen, and Ben Glocker.
\newblock Estimating model performance under domain shifts with class-specific confidence scores.
\newblock In \emph{International Conference on Medical Image Computing and Computer-Assisted Intervention}, pages 693--703. Springer, 2022.

\bibitem[Liang and Zeger(1986)]{liang1986longitudinal}
Kung-Yee Liang and Scott~L Zeger.
\newblock Longitudinal data analysis using generalized linear models.
\newblock \emph{Biometrika}, 73\penalty0 (1):\penalty0 13--22, 1986.

\bibitem[Lin et~al.(2014)Lin, Maire, Belongie, Hays, Perona, Ramanan, Doll{\'a}r, and Zitnick]{lin2014microsoft}
Tsung-Yi Lin, Michael Maire, Serge Belongie, James Hays, Pietro Perona, Deva Ramanan, Piotr Doll{\'a}r, and C~Lawrence Zitnick.
\newblock Microsoft coco: Common objects in context.
\newblock In \emph{European conference on computer vision}, pages 740--755. Springer, 2014.

\bibitem[Liu et~al.(2021)Liu, Parks, Saknite, Reasat, Cronin, Wheless, Dawant, and Tkaczyk]{liu2021baseline}
Xiaoqi Liu, Kelsey Parks, Inga Saknite, Tahsin Reasat, Austin~D Cronin, Lee~E Wheless, Benoit~M Dawant, and Eric~R Tkaczyk.
\newblock Baseline photos and confident annotation improve automated detection of cutaneous graft-versus-host disease.
\newblock \emph{Clinical hematology international}, 3\penalty0 (3):\penalty0 108, 2021.

\bibitem[Ljosa et~al.(2012)Ljosa, Sokolnicki, and Carpenter]{ljosa2012annotated}
Vebjorn Ljosa, Katherine~L Sokolnicki, and Anne~E Carpenter.
\newblock Annotated high-throughput microscopy image sets for validation.
\newblock \emph{Nature methods}, 9\penalty0 (7):\penalty0 637--637, 2012.

\bibitem[Maier-Hein et~al.(2018)Maier-Hein, Eisenmann, Reinke, Onogur, Stankovic, Scholz, Arbel, Bogunovic, Bradley, Carass, et~al.]{maier2018rankings}
Lena Maier-Hein, Matthias Eisenmann, Annika Reinke, Sinan Onogur, Marko Stankovic, Patrick Scholz, Tal Arbel, Hrvoje Bogunovic, Andrew~P Bradley, Aaron Carass, et~al.
\newblock Why rankings of biomedical image analysis competitions should be interpreted with care.
\newblock \emph{Nature communications}, 9\penalty0 (1):\penalty0 1--13, 2018.
\newblock With this comprehensive analysis of biomedical image analysis competitions (challenges), the authors initiated a shift in how such challenges are designed, performed, and reported in the biomedical domain. Its concepts and guidelines have been adopted by reputed organizations such as MICCAI.

\bibitem[Maier-Hein et~al.(2021)Maier-Hein, Wagner, Ross, Reinke, Bodenstedt, Full, Hempe, Mindroc-Filimon, Scholz, Tran, et~al.]{maier2021heidelberg}
Lena Maier-Hein, Martin Wagner, Tobias Ross, Annika Reinke, Sebastian Bodenstedt, Peter~M Full, Hellena Hempe, Diana Mindroc-Filimon, Patrick Scholz, Thuy~Nuong Tran, et~al.
\newblock Heidelberg colorectal data set for surgical data science in the sensor operating room.
\newblock \emph{Scientific data}, 8\penalty0 (1):\penalty0 1--11, 2021.

\bibitem[Maier-Hein et~al.(2022)Maier-Hein, Reinke, Christodoulou, Glocker, Godau, Isensee, Kleesiek, Kozubek, Reyes, Riegler, et~al.]{maier2022metrics}
Lena Maier-Hein, Annika Reinke, Evangelia Christodoulou, Ben Glocker, Patrick Godau, Fabian Isensee, Jens Kleesiek, Michal Kozubek, Mauricio Reyes, Michael~A Riegler, et~al.
\newblock Metrics reloaded: Pitfalls and recommendations for image analysis validation.
\newblock \emph{arXiv preprint arXiv:2206.01653}, 2022.

\bibitem[Mais et~al.(2020)Mais, Hirsch, and Kainmueller]{mais2020patchperpix}
Lisa Mais, Peter Hirsch, and Dagmar Kainmueller.
\newblock Patchperpix for instance segmentation.
\newblock In \emph{European Conference on Computer Vision}, pages 288--304. Springer, 2020.

\bibitem[Margolin et~al.(2014)Margolin, Zelnik-Manor, and Tal]{margolin2014evaluate}
Ran Margolin, Lihi Zelnik-Manor, and Ayellet Tal.
\newblock How to evaluate foreground maps?
\newblock In \emph{Proceedings of the IEEE conference on computer vision and pattern recognition}, pages 248--255, 2014.

\bibitem[Ma{\v{s}}ka et~al.(2014)Ma{\v{s}}ka, Ulman, Svoboda, Matula, Matula, Ederra, Urbiola, Espa{\~n}a, Venkatesan, Balak, et~al.]{mavska2014benchmark}
Martin Ma{\v{s}}ka, Vladim{\'\i}r Ulman, David Svoboda, Pavel Matula, Petr Matula, Cristina Ederra, Ainhoa Urbiola, Tom{\'a}s Espa{\~n}a, Subramanian Venkatesan, Deepak~MW Balak, et~al.
\newblock A benchmark for comparison of cell tracking algorithms.
\newblock \emph{Bioinformatics}, 30\penalty0 (11):\penalty0 1609--1617, 2014.

\bibitem[Matthews(1975)]{matthews1975comparison}
Brian~W Matthews.
\newblock Comparison of the predicted and observed secondary structure of t4 phage lysozyme.
\newblock \emph{Biochimica et Biophysica Acta (BBA)-Protein Structure}, 405\penalty0 (2):\penalty0 442--451, 1975.

\bibitem[Matula et~al.(2015)Matula, Ma{\v{s}}ka, Sorokin, Matula, Ortiz-de Sol{\'o}rzano, and Kozubek]{matula2015cell}
Pavel Matula, Martin Ma{\v{s}}ka, Dmitry~V Sorokin, Petr Matula, Carlos Ortiz-de Sol{\'o}rzano, and Michal Kozubek.
\newblock Cell tracking accuracy measurement based on comparison of acyclic oriented graphs.
\newblock \emph{PloS one}, 10\penalty0 (12):\penalty0 e0144959, 2015.

\bibitem[McCradden et~al.(2022)McCradden, Anderson, A~Stephenson, Drysdale, Erdman, Goldenberg, and Zlotnik~Shaul]{McCradden2022-kw}
Melissa~D McCradden, James~A Anderson, Elizabeth A~Stephenson, Erik Drysdale, Lauren Erdman, Anna Goldenberg, and Randi Zlotnik~Shaul.
\newblock A research ethics framework for the clinical translation of healthcare machine learning.
\newblock \emph{Am. J. Bioeth.}, pages 1--15, January 2022.

\bibitem[Mehrtash et~al.(2020)Mehrtash, Wells, Tempany, Abolmaesumi, and Kapur]{mehrtash2020confidence}
Alireza Mehrtash, William~M Wells, Clare~M Tempany, Purang Abolmaesumi, and Tina Kapur.
\newblock Confidence calibration and predictive uncertainty estimation for deep medical image segmentation.
\newblock \emph{IEEE transactions on medical imaging}, 39\penalty0 (12):\penalty0 3868--3878, 2020.

\bibitem[Meil{\u{a}}(2003)]{meilua2003comparing}
Marina Meil{\u{a}}.
\newblock Comparing clusterings by the variation of information.
\newblock In \emph{Learning theory and kernel machines}, pages 173--187. Springer, 2003.

\bibitem[Meissner et~al.(2022)Meissner, Nern, Dorman, G.M., Forster, Gibney, J.H., He, Iyer, Jeter, et~al.]{meissner2022searchable}
G.~Meissner, A.~Nern, Z.~Dorman, DePasquale G.M., K.~Forster, T.~Gibney, Hausenfluck J.H., Y.~He, N.~Iyer, J.~Jeter, et~al.
\newblock A searchable image resource of drosophila gal4-driver expression patterns with single neuron resolution.
\newblock \emph{BioRxiv}, page 2020.05.29.080473, 2022.

\bibitem[Moons et~al.(2015)Moons, Altman, Reitsma, Ioannidis, Macaskill, Steyerberg, Vickers, Ransohoff, and Collins]{moons2015transparent}
Karel~GM Moons, Douglas~G Altman, Johannes~B Reitsma, John~PA Ioannidis, Petra Macaskill, Ewout~W Steyerberg, Andrew~J Vickers, David~F Ransohoff, and Gary~S Collins.
\newblock Transparent reporting of a multivariable prediction model for individual prognosis or diagnosis (tripod): explanation and elaboration.
\newblock \emph{Annals of internal medicine}, 162\penalty0 (1):\penalty0 W1--W73, 2015.

\bibitem[Naeini et~al.(2015)Naeini, Cooper, and Hauskrecht]{naeini2015obtaining}
Mahdi~Pakdaman Naeini, Gregory Cooper, and Milos Hauskrecht.
\newblock Obtaining well calibrated probabilities using bayesian binning.
\newblock In \emph{Twenty-Ninth AAAI Conference on Artificial Intelligence}, 2015.

\bibitem[Nagao et~al.(2020)Nagao, Sakamoto, Chinen, Okada, and Takao]{nagao2020robust}
Yukiko Nagao, Mika Sakamoto, Takumi Chinen, Yasushi Okada, and Daisuke Takao.
\newblock Robust classification of cell cycle phase and biological feature extraction by image-based deep learning.
\newblock \emph{Molecular biology of the cell}, 31\penalty0 (13):\penalty0 1346--1354, 2020.

\bibitem[Nai et~al.(2021)Nai, Teo, Tan, O'Doherty, Stephenson, Thian, Chiong, and Reilhac]{nai2021comparison}
Ying-Hwey Nai, Bernice~W Teo, Nadya~L Tan, Sophie O'Doherty, Mary~C Stephenson, Yee~Liang Thian, Edmund Chiong, and Anthonin Reilhac.
\newblock Comparison of metrics for the evaluation of medical segmentations using prostate mri dataset.
\newblock \emph{Computers in Biology and Medicine}, 134:\penalty0 104497, 2021.

\bibitem[Nasa et~al.(2021)Nasa, Jain, and Juneja]{nasa2021delphi}
Prashant Nasa, Ravi Jain, and Deven Juneja.
\newblock Delphi methodology in healthcare research: how to decide its appropriateness.
\newblock \emph{World Journal of Methodology}, 11\penalty0 (4):\penalty0 116, 2021.

\bibitem[Nath et~al.(2020)Nath, Schilling, Parvathaneni, Huo, Blaber, Hainline, Barakovic, Romascano, Rafael-Patino, Frigo, et~al.]{nath2020tractography}
Vishwesh Nath, Kurt~G Schilling, Prasanna Parvathaneni, Yuankai Huo, Justin~A Blaber, Allison~E Hainline, Muhamed Barakovic, David Romascano, Jonathan Rafael-Patino, Matteo Frigo, et~al.
\newblock Tractography reproducibility challenge with empirical data (traced): the 2017 ismrm diffusion study group challenge.
\newblock \emph{Journal of Magnetic Resonance Imaging}, 51\penalty0 (1):\penalty0 234--249, 2020.

\bibitem[Neumann et~al.(2018)Neumann, Zisserman, and Vedaldi]{neumann2018relaxed}
Lukas Neumann, Andrew Zisserman, and Andrea Vedaldi.
\newblock Relaxed softmax: Efficient confidence auto-calibration for safe pedestrian detection.
\newblock \emph{2018 NIPS Workshop on Machine Learning for Intelligent Transportation Systems}, 2018.

\bibitem[Nikolov et~al.(2021)Nikolov, Blackwell, Zverovitch, Mendes, Livne, De~Fauw, Patel, Meyer, Askham, Romera-Paredes, et~al.]{nikolov2021clinically}
Stanislav Nikolov, Sam Blackwell, Alexei Zverovitch, Ruheena Mendes, Michelle Livne, Jeffrey De~Fauw, Yojan Patel, Clemens Meyer, Harry Askham, Bernadino Romera-Paredes, et~al.
\newblock Clinically applicable segmentation of head and neck anatomy for radiotherapy: deep learning algorithm development and validation study.
\newblock \emph{Journal of Medical Internet Research}, 23\penalty0 (7):\penalty0 e26151, 2021.

\bibitem[Oakden-Rayner et~al.(2020)Oakden-Rayner, Dunnmon, Carneiro, and R{\'e}]{Oakden-Rayner2020-co}
Luke Oakden-Rayner, Jared Dunnmon, Gustavo Carneiro, and Christopher R{\'e}.
\newblock Hidden stratification causes clinically meaningful failures in machine learning for medical imaging.
\newblock \emph{Proc ACM Conf Health Inference Learn (2020)}, 2020:\penalty0 151--159, April 2020.

\bibitem[Obermeyer et~al.(2019)Obermeyer, Powers, Vogeli, and Mullainathan]{Obermeyer2019-pp}
Ziad Obermeyer, Brian Powers, Christine Vogeli, and Sendhil Mullainathan.
\newblock Dissecting racial bias in an algorithm used to manage the health of populations.
\newblock \emph{Science}, 366\penalty0 (6464):\penalty0 447--453, October 2019.

\bibitem[Park et~al.(2023)Park, Han, Jang, Park, Lee, Kim, and Choi]{park2023methods}
Seong~Ho Park, Kyunghwa Han, Hye~Young Jang, Ji~Eun Park, June-Goo Lee, Dong~Wook Kim, and Jaesoon Choi.
\newblock Methods for {Clinical} {Evaluation} of {Artificial} {Intelligence} {Algorithms} for {Medical} {Diagnosis}.
\newblock \emph{Radiology}, 306\penalty0 (1):\penalty0 20--31, January 2023.
\newblock ISSN 0033-8419.
\newblock \doi{10.1148/radiol.220182}.
\newblock URL \url{https://pubs.rsna.org/doi/10.1148/radiol.220182}.
\newblock Publisher: Radiological Society of North America.

\bibitem[Patterson et~al.(2021)Patterson, Gonzalez, Le, Liang, Munguia, Rothchild, So, Texier, and Dean]{Patterson2021-sw}
David Patterson, Joseph Gonzalez, Quoc Le, Chen Liang, Lluis-Miquel Munguia, Daniel Rothchild, David So, Maud Texier, and Jeff Dean.
\newblock Carbon emissions and large neural network training.
\newblock \emph{arXiv}, April 2021.

\bibitem[Pauker and Kassirer(1975)]{pauker1975therapeutic}
Stephen~G Pauker and Jerome~P Kassirer.
\newblock Therapeutic decision making: a cost-benefit analysis.
\newblock \emph{New England Journal of Medicine}, 293\penalty0 (5):\penalty0 229--234, 1975.

\bibitem[Perez-Lebel et~al.(2023)Perez-Lebel, Morvan, and Varoquaux]{perez2022beyond}
Alexandre Perez-Lebel, Marine~Le Morvan, and Ga{\"e}l Varoquaux.
\newblock Beyond calibration: estimating the grouping loss of modern neural networks.
\newblock \emph{International Conference on Learning Representations}, 2023.

\bibitem[Popordanoska et~al.(2022)Popordanoska, Sayer, and Blaschko]{popordanoska2022consistent}
Teodora Popordanoska, Raphael Sayer, and Matthew~B Blaschko.
\newblock A consistent and differentiable lp canonical calibration error estimator.
\newblock In \emph{Advances in Neural Information Processing Systems}, 2022.

\bibitem[Quinonero-Candela et~al.(2005)Quinonero-Candela, Rasmussen, Sinz, Bousquet, and Sch{\"o}lkopf]{quinonero2005evaluating}
Joaquin Quinonero-Candela, Carl~Edward Rasmussen, Fabian Sinz, Olivier Bousquet, and Bernhard Sch{\"o}lkopf.
\newblock Evaluating predictive uncertainty challenge.
\newblock In \emph{Machine Learning Challenges Workshop}, pages 1--27. Springer, 2005.

\bibitem[Rand(1971)]{rand1971objective}
William~M Rand.
\newblock Objective criteria for the evaluation of clustering methods.
\newblock \emph{Journal of the American Statistical association}, 66\penalty0 (336):\penalty0 846--850, 1971.

\bibitem[Reinke et~al.(2018)Reinke, Eisenmann, Onogur, Stankovic, Scholz, Full, Bogunovic, Landman, Maier, Menze, et~al.]{reinke2018exploit}
Annika Reinke, Matthias Eisenmann, Sinan Onogur, Marko Stankovic, Patrick Scholz, Peter~M Full, Hrvoje Bogunovic, Bennett~A Landman, Oskar Maier, Bjoern Menze, et~al.
\newblock How to exploit weaknesses in biomedical challenge design and organization.
\newblock In \emph{International Conference on Medical Image Computing and Computer-Assisted Intervention}, pages 388--395. Springer, 2018.

\bibitem[Reinke et~al.(2021)Reinke, Eisenmann, Tizabi, Sudre, R{\"a}dsch, Antonelli, Arbel, Bakas, Cardoso, Cheplygina, et~al.]{reinke2021common}
Annika Reinke, Matthias Eisenmann, Minu~D Tizabi, Carole~H Sudre, Tim R{\"a}dsch, Michela Antonelli, Tal Arbel, Spyridon Bakas, M~Jorge Cardoso, Veronika Cheplygina, et~al.
\newblock Common limitations of image processing metrics: A picture story.
\newblock \emph{arXiv preprint arXiv:2104.05642}, 2021.

\bibitem[Reinke et~al.(2024)Reinke, Tizabi, Baumgartner, Eisenmann, Heckmann-N{\"o}tzel, Kavur, R{\"a}dsch, Sudre, Acion, Antonelli, et~al.]{reinketizabi2022metrics}
Annika Reinke, Minu~D Tizabi, Michael Baumgartner, Matthias Eisenmann, Doreen Heckmann-N{\"o}tzel, A~Emre Kavur, Tim R{\"a}dsch, Carole~H Sudre, Laura Acion, Michela Antonelli, et~al.
\newblock Understanding metric-related pitfalls in image analysis validation.
\newblock \emph{Nature methods}, pages 1--13, 2024.

\bibitem[Ren et~al.(2015)Ren, He, Girshick, and Sun]{ren2015faster}
Shaoqing Ren, Kaiming He, Ross Girshick, and Jian Sun.
\newblock Faster r-cnn: Towards real-time object detection with region proposal networks.
\newblock \emph{Advances in neural information processing systems}, 28, 2015.

\bibitem[Riley et~al.(2016)Riley, Ensor, Snell, Debray, Altman, Moons, and Collins]{riley2016external}
Richard~D Riley, Joie Ensor, Kym~IE Snell, Thomas~PA Debray, Doug~G Altman, Karel~GM Moons, and Gary~S Collins.
\newblock External validation of clinical prediction models using big datasets from e-health records or ipd meta-analysis: opportunities and challenges.
\newblock \emph{bmj}, 353, 2016.

\bibitem[Ro{\ss} et~al.(2021)Ro{\ss}, Bruno, Reinke, Wiesenfarth, Koeppel, Full, Pekdemir, Godau, Trofimova, Isensee, et~al.]{ross2021can}
Tobias Ro{\ss}, Pierangela Bruno, Annika Reinke, Manuel Wiesenfarth, Lisa Koeppel, Peter~M Full, B{\"u}nyamin Pekdemir, Patrick Godau, Darya Trofimova, Fabian Isensee, et~al.
\newblock How can we learn (more) from challenges? a statistical approach to driving future algorithm development.
\newblock \emph{arXiv preprint arXiv:2106.09302}, 2021.

\bibitem[Sage et~al.(2015)Sage, Kirshner, Pengo, Stuurman, Min, Manley, and Unser]{sage2015quantitative}
Daniel Sage, Hagai Kirshner, Thomas Pengo, Nico Stuurman, Junhong Min, Suliana Manley, and Michael Unser.
\newblock Quantitative evaluation of software packages for single-molecule localization microscopy.
\newblock \emph{Nature methods}, 12\penalty0 (8):\penalty0 717--724, 2015.

\bibitem[Samuelson and Petrick(2006)]{samuelson2006comparing}
Frank~W Samuelson and Nicholas Petrick.
\newblock Comparing image detection algorithms using resampling.
\newblock In \emph{3rd IEEE International Symposium on Biomedical Imaging: Nano to Macro, 2006.}, pages 1312--1315. IEEE, 2006.

\bibitem[S{\'a}nchez-Montes et~al.(2019)S{\'a}nchez-Montes, S{\'a}nchez, Bernal, C{\'o}rdova, L{\'o}pez-Cer{\'o}n, Cuatrecasas, De~Miguel, Garc{\'\i}a-Rodr{\'\i}guez, Garc{\'e}s-Dur{\'a}n, Pellis{\'e}, et~al.]{sanchez2019computer}
Cristina S{\'a}nchez-Montes, Francisco~Javier S{\'a}nchez, Jorge Bernal, Henry C{\'o}rdova, Mar{\'\i}a L{\'o}pez-Cer{\'o}n, Miriam Cuatrecasas, Cristina~Rodr{\'\i}guez De~Miguel, Ana Garc{\'\i}a-Rodr{\'\i}guez, Rodrigo Garc{\'e}s-Dur{\'a}n, Mar{\'\i}a Pellis{\'e}, et~al.
\newblock Computer-aided prediction of polyp histology on white light colonoscopy using surface pattern analysis.
\newblock \emph{Endoscopy}, 51\penalty0 (03):\penalty0 261--265, 2019.

\bibitem[Sander et~al.(2019)Sander, de~Vos, Wolterink, and I{\v{s}}gum]{sander2019towards}
J{\"o}rg Sander, Bob~D de~Vos, Jelmer~M Wolterink, and Ivana I{\v{s}}gum.
\newblock Towards increased trustworthiness of deep learning segmentation methods on cardiac mri.
\newblock In \emph{Medical imaging 2019: image Processing}, volume 10949, pages 324--330. SPIE, 2019.

\bibitem[Schulam and Saria(2019)]{Schulam2019-xo}
Peter Schulam and Suchi Saria.
\newblock Can you trust this prediction? auditing pointwise reliability after learning.
\newblock In Kamalika Chaudhuri and Masashi Sugiyama, editors, \emph{Proceedings of the {Twenty-Second} International Conference on Artificial Intelligence and Statistics}, volume~89 of \emph{Proceedings of Machine Learning Research}, pages 1022--1031. PMLR, 2019.

\bibitem[Schulz et~al.(2010)Schulz, Altman, Moher, and Group*]{schulz2010consort}
Kenneth~F Schulz, Douglas~G Altman, David Moher, and CONSORT Group*.
\newblock Consort 2010 statement: updated guidelines for reporting parallel group randomized trials.
\newblock \emph{Annals of internal medicine}, 152\penalty0 (11):\penalty0 726--732, 2010.

\bibitem[Setio et~al.(2017)Setio, Traverso, De~Bel, Berens, Van Den~Bogaard, Cerello, Chen, Dou, Fantacci, Geurts, et~al.]{setio2017validation}
Arnaud Arindra~Adiyoso Setio, Alberto Traverso, Thomas De~Bel, Moira~SN Berens, Cas Van Den~Bogaard, Piergiorgio Cerello, Hao Chen, Qi~Dou, Maria~Evelina Fantacci, Bram Geurts, et~al.
\newblock Validation, comparison, and combination of algorithms for automatic detection of pulmonary nodules in computed tomography images: the luna16 challenge.
\newblock \emph{Medical image analysis}, 42:\penalty0 1--13, 2017.

\bibitem[Shah et~al.(2019)Shah, Milstein, and Bagley]{shah2019making}
Nigam~H Shah, Arnold Milstein, and Steven~C Bagley.
\newblock Making machine learning models clinically useful.
\newblock \emph{Jama}, 322\penalty0 (14):\penalty0 1351--1352, 2019.

\bibitem[Shit et~al.(2021)Shit, Paetzold, Sekuboyina, Ezhov, Unger, Zhylka, Pluim, Bauer, and Menze]{shit2021cldice}
Suprosanna Shit, Johannes~C Paetzold, Anjany Sekuboyina, Ivan Ezhov, Alexander Unger, Andrey Zhylka, Josien~PW Pluim, Ulrich Bauer, and Bjoern~H Menze.
\newblock cldice-a novel topology-preserving loss function for tubular structure segmentation.
\newblock In \emph{Proceedings of the IEEE/CVF Conference on Computer Vision and Pattern Recognition}, pages 16560--16569, 2021.

\bibitem[Simpson et~al.(2019)Simpson, Antonelli, Bakas, Bilello, Farahani, Van~Ginneken, Kopp-Schneider, Landman, Litjens, Menze, et~al.]{simpson2019large}
Amber~L Simpson, Michela Antonelli, Spyridon Bakas, Michel Bilello, Keyvan Farahani, Bram Van~Ginneken, Annette Kopp-Schneider, Bennett~A Landman, Geert Litjens, Bjoern Menze, et~al.
\newblock A large annotated medical image dataset for the development and evaluation of segmentation algorithms.
\newblock \emph{arXiv preprint arXiv:1902.09063}, 2019.

\bibitem[Sounderajah et~al.(2020)Sounderajah, Ashrafian, Aggarwal, De~Fauw, Denniston, Greaves, Karthikesalingam, King, Liu, Markar, McInnes, Panch, Pearson-Stuttard, Ting, Golub, Moher, Bossuyt, and Darzi]{Sounderajah2020-ou}
Viknesh Sounderajah, Hutan Ashrafian, Ravi Aggarwal, Jeffrey De~Fauw, Alastair~K Denniston, Felix Greaves, Alan Karthikesalingam, Dominic King, Xiaoxuan Liu, Sheraz~R Markar, Matthew D~F McInnes, Trishan Panch, Jonathan Pearson-Stuttard, Daniel S~W Ting, Robert~M Golub, David Moher, Patrick~M Bossuyt, and Ara Darzi.
\newblock Developing specific reporting guidelines for diagnostic accuracy studies assessing {AI} interventions: The {STARD-AI} steering group.
\newblock \emph{Nat. Med.}, 26\penalty0 (6):\penalty0 807--808, June 2020.

\bibitem[Steyerberg et~al.(2010)Steyerberg, Vickers, Cook, Gerds, Gonen, Obuchowski, Pencina, and Kattan]{steyerberg2010assessing}
Ewout~W Steyerberg, Andrew~J Vickers, Nancy~R Cook, Thomas Gerds, Mithat Gonen, Nancy Obuchowski, Michael~J Pencina, and Michael~W Kattan.
\newblock Assessing the performance of prediction models: a framework for some traditional and novel measures.
\newblock \emph{Epidemiology (Cambridge, Mass.)}, 21\penalty0 (1):\penalty0 128, 2010.

\bibitem[Strubell et~al.(2019)Strubell, Ganesh, and McCallum]{Strubell2019-pw}
Emma Strubell, Ananya Ganesh, and Andrew McCallum.
\newblock Energy and policy considerations for deep learning in {NLP}.
\newblock \emph{arXiv}, June 2019.

\bibitem[Summers and Dinneen(2021)]{summers2021nondeterminism}
Cecilia Summers and Michael~J Dinneen.
\newblock Nondeterminism and instability in neural network optimization.
\newblock In \emph{International Conference on Machine Learning}, pages 9913--9922. PMLR, 2021.

\bibitem[Taha and Hanbury(2015)]{taha2015metrics}
Abdel~Aziz Taha and Allan Hanbury.
\newblock Metrics for evaluating 3d medical image segmentation: analysis, selection, and tool.
\newblock \emph{BMC medical imaging}, 15\penalty0 (1):\penalty0 1--28, 2015.
\newblock The paper discusses the importance of effective metrics for evaluating the accuracy of 3D medical image segmentation algorithms. The authors analyze existing metrics, propose a selection methodology, and develop a tool to aid researchers in choosing appropriate evaluation metrics based on the specific characteristics of the segmentation task.

\bibitem[Targosz et~al.(2021)Targosz, Przysta{\l}ka, Wiaderkiewicz, and Mrugacz]{targosz2021semantic}
Anna Targosz, Piotr Przysta{\l}ka, Ryszard Wiaderkiewicz, and Grzegorz Mrugacz.
\newblock Semantic segmentation of human oocyte images using deep neural networks.
\newblock \emph{BioMedical Engineering OnLine}, 20\penalty0 (1):\penalty0 40, 2021.

\bibitem[Tharwat(2020)]{tharwat2020classification}
Alaa Tharwat.
\newblock Classification assessment methods.
\newblock \emph{Applied Computing and Informatics}, 2020.

\bibitem[{The Institute for Ethical Ai} and {Machine Learning}(2018)]{The_Institute_for_Ethical_Ai_undated-ed}
{The Institute for Ethical Ai} and {Machine Learning}.
\newblock The institute for ethical {AI} \& machine learning.
\newblock \url{https://ethical.institute/principles.html}, 2018.
\newblock Accessed: 2022-5-21.

\bibitem[Timmins et~al.(2021)Timmins, van~der Schaaf, Bennink, Ruigrok, An, Baumgartner, Bourdon, De~Feo, Di~Noto, Dubost, et~al.]{timmins2021comparing}
Kimberley~M Timmins, Irene~C van~der Schaaf, Edwin Bennink, Ynte~M Ruigrok, Xingle An, Michael Baumgartner, Pascal Bourdon, Riccardo De~Feo, Tommaso Di~Noto, Florian Dubost, et~al.
\newblock Comparing methods of detecting and segmenting unruptured intracranial aneurysms on tof-mras: The adam challenge.
\newblock \emph{Neuroimage}, 238:\penalty0 118216, 2021.

\bibitem[Tirian and Dickson(2017)]{tirian2017vt}
Laszlo Tirian and Barry~J Dickson.
\newblock The vt gal4, lexa, and split-gal4 driver line collections for targeted expression in the drosophila nervous system.
\newblock \emph{BioRxiv}, page 198648, 2017.

\bibitem[Tran et~al.(2022)Tran, Adler, Yamlahi, Christodoulou, Godau, Reinke, Tizabi, Sauer, Persicke, Albert, and Maier-Hein]{Tran2022}
Thuy~N Tran, Tim Adler, Amine Yamlahi, Evangelia Christodoulou, Patrick Godau, Annika Reinke, Minu~D Tizabi, Peter Sauer, Tillmann Persicke, Jörg~G. Albert, and Lena Maier-Hein.
\newblock Sources of performance variability in deep learning-based polyp detection.
\newblock \emph{arXiv preprint arXiv:2211.09708}, 2022.

\bibitem[Ulman et~al.(2017)Ulman, Ma{\v{s}}ka, Magnusson, Ronneberger, Haubold, Harder, Matula, Matula, Svoboda, Radojevic, et~al.]{ulman2017objective}
Vladim{\'\i}r Ulman, Martin Ma{\v{s}}ka, Klas~EG Magnusson, Olaf Ronneberger, Carsten Haubold, Nathalie Harder, Pavel Matula, Petr Matula, David Svoboda, Miroslav Radojevic, et~al.
\newblock An objective comparison of cell-tracking algorithms.
\newblock \emph{Nature methods}, 14\penalty0 (12):\penalty0 1141--1152, 2017.

\bibitem[Usatine and Manci(2021)]{dermoscopedia}
Richard Usatine and Rachel Manci.
\newblock Dermoscopedia, 2021.
\newblock \url{https://dermoscopedia.org/File:DF_chinese_dms.JPG}.

\bibitem[Vaassen et~al.(2020)Vaassen, Hazelaar, Vaniqui, Gooding, van~der Heyden, Canters, and van Elmpt]{vaassen2020evaluation}
Femke Vaassen, Colien Hazelaar, Ana Vaniqui, Mark Gooding, Brent van~der Heyden, Richard Canters, and Wouter van Elmpt.
\newblock Evaluation of measures for assessing time-saving of automatic organ-at-risk segmentation in radiotherapy.
\newblock \emph{Physics and Imaging in Radiation Oncology}, 13:\penalty0 1--6, 2020.

\bibitem[Vaicenavicius et~al.(2019)Vaicenavicius, Widmann, Andersson, Lindsten, Roll, and Sch{\"o}n]{vaicenavicius2019evaluating}
Juozas Vaicenavicius, David Widmann, Carl Andersson, Fredrik Lindsten, Jacob Roll, and Thomas Sch{\"o}n.
\newblock Evaluating model calibration in classification.
\newblock In \emph{The 22nd International Conference on Artificial Intelligence and Statistics}, pages 3459--3467. PMLR, 2019.

\bibitem[Van~Calster et~al.(2016)Van~Calster, Nieboer, Vergouwe, De~Cock, Pencina, and Steyerberg]{van2016calibration}
Ben Van~Calster, Daan Nieboer, Yvonne Vergouwe, Bavo De~Cock, Michael~J Pencina, and Ewout~W Steyerberg.
\newblock A calibration hierarchy for risk models was defined: from utopia to empirical data.
\newblock \emph{Journal of clinical epidemiology}, 74:\penalty0 167--176, 2016.

\bibitem[Van~Calster et~al.(2019)Van~Calster, McLernon, Van~Smeden, Wynants, and Steyerberg]{van2019calibration}
Ben Van~Calster, David~J McLernon, Maarten Van~Smeden, Laure Wynants, and Ewout~W Steyerberg.
\newblock Calibration: the achilles heel of predictive analytics.
\newblock \emph{BMC medicine}, 17\penalty0 (1):\penalty0 1--7, 2019.

\bibitem[Van~Ginneken et~al.(2010)Van~Ginneken, Armato~III, de~Hoop, van Amelsvoort-van~de Vorst, Duindam, Niemeijer, Murphy, Schilham, Retico, Fantacci, et~al.]{van2010comparing}
Bram Van~Ginneken, Samuel~G Armato~III, Bartjan de~Hoop, Saskia van Amelsvoort-van~de Vorst, Thomas Duindam, Meindert Niemeijer, Keelin Murphy, Arnold Schilham, Alessandra Retico, Maria~Evelina Fantacci, et~al.
\newblock Comparing and combining algorithms for computer-aided detection of pulmonary nodules in computed tomography scans: the anode09 study.
\newblock \emph{Medical image analysis}, 14\penalty0 (6):\penalty0 707--722, 2010.

\bibitem[Van~Hoorde et~al.(2015)Van~Hoorde, Van~Huffel, Timmerman, Bourne, and Van~Calster]{van2015spline}
Kirsten Van~Hoorde, Sabine Van~Huffel, Dirk Timmerman, Tom Bourne, and Ben Van~Calster.
\newblock A spline-based tool to assess and visualize the calibration of multiclass risk predictions.
\newblock \emph{Journal of biomedical informatics}, 54:\penalty0 283--293, 2015.

\bibitem[Van~Rijsbergen(1979)]{van1979information}
C~Van~Rijsbergen.
\newblock Information retrieval: theory and practice.
\newblock In \emph{Proceedings of the Joint IBM/University of Newcastle upon Tyne Seminar on Data Base Systems}, volume~79, 1979.

\bibitem[Vickers and Elkin(2006)]{vickers2006decision}
Andrew~J Vickers and Elena~B Elkin.
\newblock Decision curve analysis: a novel method for evaluating prediction models.
\newblock \emph{Medical Decision Making}, 26\penalty0 (6):\penalty0 565--574, 2006.

\bibitem[Vickers et~al.(2016)Vickers, Van~Calster, and Steyerberg]{vickers2016net}
Andrew~J Vickers, Ben Van~Calster, and Ewout~W Steyerberg.
\newblock Net benefit approaches to the evaluation of prediction models, molecular markers, and diagnostic tests.
\newblock \emph{bmj}, 352, 2016.

\bibitem[Wack et~al.(2012)Wack, Dwyer, Bergsland, Di~Perri, Ranza, Hussein, Ramasamy, Poloni, and Zivadinov]{wack2012improved}
David~S Wack, Michael~G Dwyer, Niels Bergsland, Carol Di~Perri, Laura Ranza, Sara Hussein, Deepa Ramasamy, Guy Poloni, and Robert Zivadinov.
\newblock Improved assessment of multiple sclerosis lesion segmentation agreement via detection and outline error estimates.
\newblock \emph{BMC medical imaging}, 12\penalty0 (1):\penalty0 1--10, 2012.

\bibitem[Warrens(2012)]{warrens2012some}
Matthijs~J Warrens.
\newblock Some paradoxical results for the quadratically weighted kappa.
\newblock \emph{Psychometrika}, 77\penalty0 (2):\penalty0 315--323, 2012.

\bibitem[Wenger et~al.(2020)Wenger, Kjellstr{\"o}m, and Triebel]{wenger2020non}
Jonathan Wenger, Hedvig Kjellstr{\"o}m, and Rudolph Triebel.
\newblock Non-parametric calibration for classification.
\newblock In \emph{International Conference on Artificial Intelligence and Statistics}, pages 178--190. PMLR, 2020.

\bibitem[Widmann et~al.(2019)Widmann, Lindsten, and Zachariah]{widmann2019calibration}
David Widmann, Fredrik Lindsten, and Dave Zachariah.
\newblock Calibration tests in multi-class classification: A unifying framework.
\newblock \emph{Advances in Neural Information Processing Systems}, 32, 2019.

\bibitem[Wiesenfarth et~al.(2021)Wiesenfarth, Reinke, Landman, Eisenmann, Saiz, Cardoso, Maier-Hein, and Kopp-Schneider]{wiesenfarth2021methods}
Manuel Wiesenfarth, Annika Reinke, Bennett~A Landman, Matthias Eisenmann, Laura~Aguilera Saiz, M~Jorge Cardoso, Lena Maier-Hein, and Annette Kopp-Schneider.
\newblock Methods and open-source toolkit for analyzing and visualizing challenge results.
\newblock \emph{Scientific reports}, 11\penalty0 (1):\penalty0 1--15, 2021.

\bibitem[Wolff~Anthony et~al.(2020)Wolff~Anthony, Kanding, and Selvan]{Wolff_Anthony2020-zl}
Lasse~F Wolff~Anthony, Benjamin Kanding, and Raghavendra Selvan.
\newblock Carbontracker: Tracking and predicting the carbon footprint of training deep learning models.
\newblock \emph{arXiv}, July 2020.

\bibitem[Wynants et~al.(2019)Wynants, Van~Smeden, McLernon, Timmerman, Steyerberg, and Van~Calster]{wynants2019three}
Laure Wynants, Maarten Van~Smeden, David~J McLernon, Dirk Timmerman, Ewout~W Steyerberg, and Ben Van~Calster.
\newblock Three myths about risk thresholds for prediction models.
\newblock \emph{BMC medicine}, 17\penalty0 (1):\penalty0 1--7, 2019.

\bibitem[Yang and Buettner(2021)]{yang2021multi}
Yinchong Yang and Florian Buettner.
\newblock Multi-output gaussian processes for uncertainty-aware recommender systems.
\newblock In \emph{Uncertainty in Artificial Intelligence}, pages 1505--1514. PMLR, 2021.

\bibitem[Yeghiazaryan and Voiculescu(2015)]{yeghiazaryan2015overview}
Varduhi Yeghiazaryan and Irina Voiculescu.
\newblock An overview of current evaluation methods used in medical image segmentation.
\newblock \emph{Department of Computer Science, University of Oxford}, 2015.

\bibitem[Yeghiazaryan and Voiculescu(2018)]{yeghiazaryan2018family}
Varduhi Yeghiazaryan and Irina~D Voiculescu.
\newblock Family of boundary overlap metrics for the evaluation of medical image segmentation.
\newblock \emph{Journal of Medical Imaging}, 5\penalty0 (1):\penalty0 015006, 2018.

\bibitem[Zhang et~al.(2020)Zhang, Xie, Liu, Deng, Peng, Wang, Xu, Ruan, Deng, Guo, et~al.]{zhang2020deepphagy}
Ying Zhang, Yubin Xie, Wenzhong Liu, Wankun Deng, Di~Peng, Chenwei Wang, Haodong Xu, Chen Ruan, Yongjie Deng, Yaping Guo, et~al.
\newblock Deepphagy: a deep learning framework for quantitatively measuring autophagy activity in saccharomyces cerevisiae.
\newblock \emph{Autophagy}, 16\penalty0 (4):\penalty0 626--640, 2020.

\bibitem[Zhu(2020)]{zhu2020OnMatthews}
Qiuming Zhu.
\newblock On the performance of matthews correlation coefficient (mcc) for imbalanced dataset.
\newblock \emph{Pattern Recognit. Lett.}, 136:\penalty0 71--80, 2020.

\end{thebibliography}
\addcontentsline{toc}{chapter}{\protect\numberline{}References}

\newpage
\section*{Acronyms}
\addcontentsline{toc}{chapter}{\protect\numberline{}Acronyms}
\begin{acronym}[]
\hypersetup{hidelinks}
\acro{Acc}{Accuracy}
\acro{AI}{artificial intelligence}
\acro{AP}{Average Precision}
\acro{ASSD}{Average Symmetric Surface Distance}
\acro{AUC}{Area under the curve}
\acro{AUROC}{Area under the Receiver Operating Characteristic Curve}
\acro{BA}{Balanced Accuracy}
\acro{BM}{Bookmaker Informedness}
\acro{BPMN}{Business Process Model and Notation}
\acro{BS}{Brier Score}
\acro{BSS}{Brier Skill Score}
\acro{CE}{Calibration Error}
\acro{CI}{Confidence Interval}
\acro{CK}{Cohen's Kappa}
\acro{clDice}{centerline Dice Similarity Coefficient}
\acro{CT}{computed tomography}
\acro{CWCE}{Class-wise Calibration Error}
\acro{DSC}{Dice Similarity Coefficient}
\acro{EC}{Expected Cost}
\acro{ECE}{Expected Calibration Error}
%\acro{ECE$^{KDE}$}{Expected Calibration Error Kernel Density Estimate}
\acro{ECEKDE}[ECE\textsuperscript{KDE}]{Expected Calibration Error Kernel Density Estimate}
\acro{ECI}{Expected Calibration Index}
\acro{ECN}{normalized EC}
\acro{EQUATOR}{Enhancing the QUAlity and Transparency Of health Research}
\acro{ER}{Error Rate}
\acro{FN}{False Negative}
\acro{FNR}{False Negative Rate}
\acro{FP}{False Positive}
\acro{FPPI}{False Positives per Image}
\acro{FDR}{False Discovery Rate}
\acro{FOR}{False Omission Rate}
\acro{FPR}{False Positive Rate}
\acro{FROC}{Free-Response Receiver Operating Characteristic}
\acro{HD}{Hausdorff Distance}
\acro{HD95}{Hausdorff Distance 95th Percentile}
\acro{HR}{Hit Rate}
\acro{ICI}{Integrated Calibration Index}
\acro{ImLC}{Image-level Classification}
\acro{InS}{Instance Segmentation}
\acro{IoU}{Intersection over Union}
\acro{IoR}{Intersection over Reference}
\acro{J}{Youden Index}
\acro{JAC}{Jaccard Index}
\acro{KCE}{Kernel Calibration Error}
\acro{LR+}{Positive Likelihood Ratio}
\acro{Log Score}{Logarithmic Score}
\acro{mAP}{Mean Average Precision}
\acro{MASD}{Mean Average Surface Distance}
\acro{MCC}{Matthews Correlation Coefficient}
\acro{MCE}{Maximum Calibration Error}
\acro{MCP}{Maximum Class Probability}
\acro{MICCAI}{Medical Image Computing and Computer Assisted Interventions}
\acro{MK}{Markedness}
\acro{ML}{machine learning}
\acro{MONAI}{Medical Open Network
for Artificial Intelligence}
\acro{MRI}{magnetic resonance imaging}
\acro{MS}{multiple sclerosis}
\acro{NaN}{'Not a Number'}
\acro{NB}{Net Benefit}
\acro{NLL}{Negative Log Likelihood}
\acro{NPV}{Negative Predictive Value}
\acro{NSD}{Normalized Surface Distance}
\acro{ObD}{Object Detection}
\acro{O:E ratio}{Observed:Expected ratio}
\acro{PPV}{Positive Predictive Value}
\acro{PQ}{Panoptic Quality}
\acro{PHI}{Protected Health Information}
\acro{PR}{Precision-Recall}
\acro{PREC}{Precision}
\acro{PSR}{Proper Scoring Rule}
\acro{RBS}{Root Brier Score}
\acro{RI}{Rand Index}
\acro{ROC}{Receiver Operating Characteristic}
\acro{ROI}{Region of Interest}
\acro{SEL}{Selectivity}
\acro{Sens}{Sensitivity}
\acro{Spec}{Specificity}
\acro{SemS}{Semantic Segmentation}
\acro{TN}{True Negative}
\acro{TNR}{True Negative Rate}
\acro{TOF-MRA}{time-of-flight magnetic resonance angiography}
\acro{TP}{True Positive}
\acro{TPR}{True Positive Rate}
\acro{VoI}{Variation of Information}
\acro{WCK}{Weighted Cohen's Kappa}
\acro{WSI}{whole slide imaging}
\acro{X$^{th}$ Percentile HD}{X$^{th}$ Percentile Hausdorff Distance}

\end{acronym}

% \section{Full Author Affiliations}
% \label{app:authors}
% \printauthors

\newpage
\section*{Extended Data}
\addcontentsline{toc}{chapter}{\protect\numberline{}Extended Data}

\setcounter{figure}{0}
\setcounter{table}{0}

\makeatletter
\renewcommand{\fnum@figure}{Extended Data Fig. \thefigure}
\renewcommand{\fnum@table}{Extended Data Tab. \thetable}
\makeatother

\begin{figure}[H]
    \centering
    \includegraphics[width=1\linewidth]{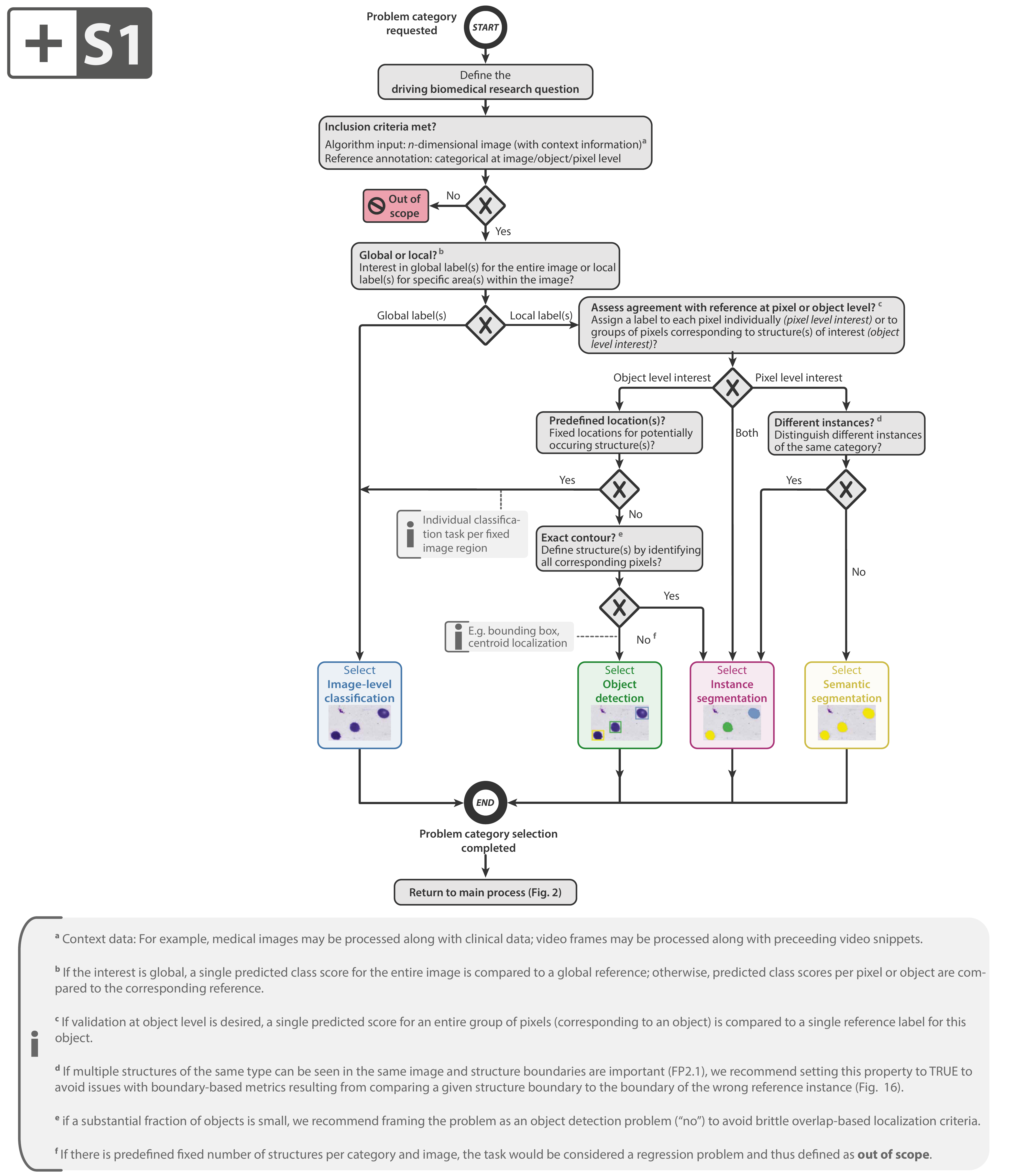}
    \caption{\textbf{Subprocess S1 for selecting a problem category.} The \textit{Category Mapping} maps a given research problem to the appropriate problem category with the goal of \textbf{grouping problems by similarity of validation.} The leaf nodes represent the categories: image-level classification, object detection, instance segmentation, or semantic segmentation. FP2.1 refers to fingerprint 2.1 (see Fig.~\ref{fig:fingerprint-ss-1}). An overview of the symbols used in the process diagram is provided in Fig.~\ref{fig:process-diagram-symbols}.}
    \label{fig:category_mapping}
\end{figure}

\begin{figure}[H]
    \centering
    \includegraphics[width=1\linewidth]{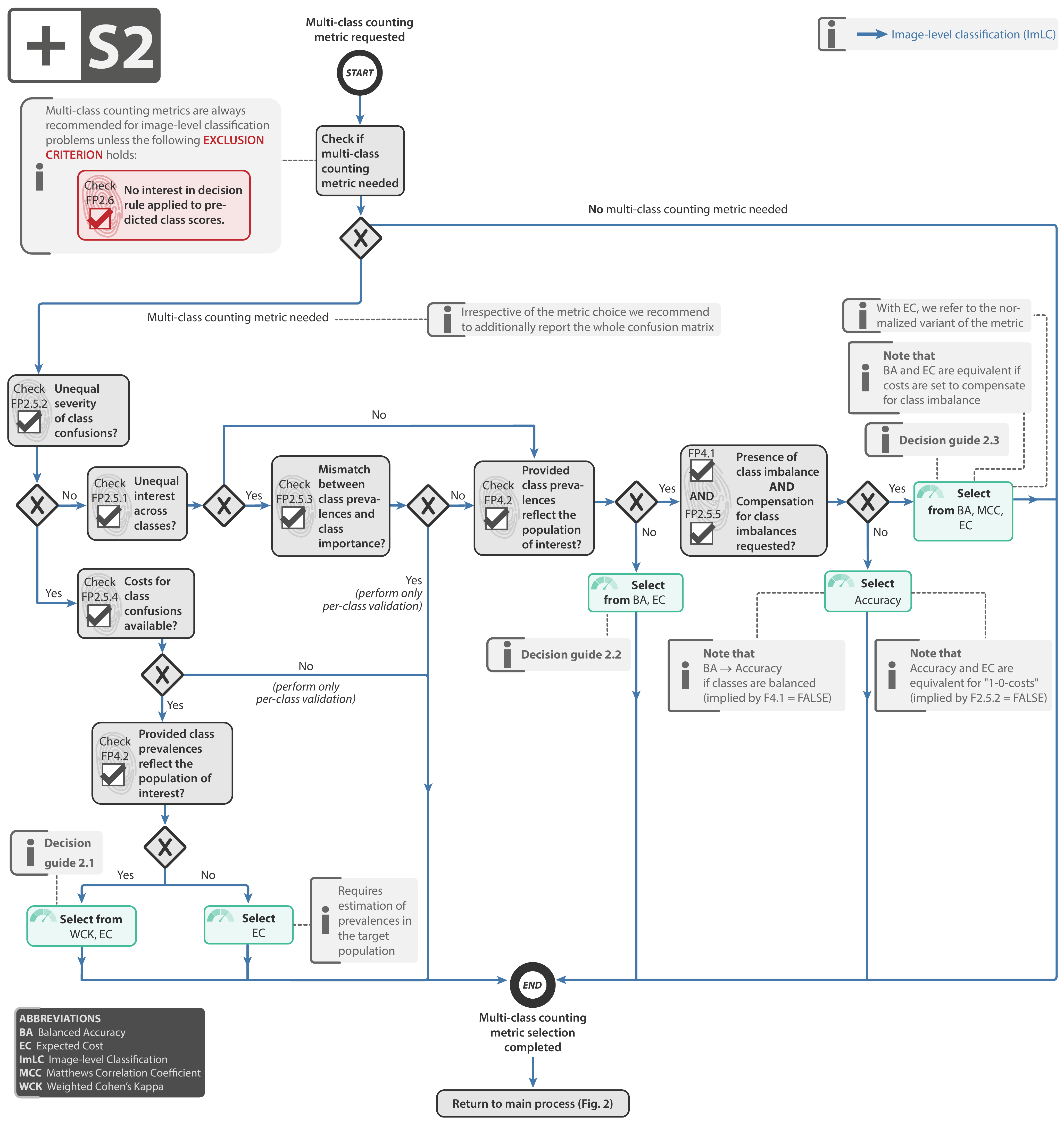}
    \caption{\textbf{Subprocess S2 for selecting multi-class metrics (if any)}. Applies to: image-level classification (ImLC). In the case of presence of class imbalance and no compensation of class imbalance being requested, one should follow the "No" branch. Decision guides are provided in Suppl. Note~\ref{ssec:dg2}. A detailed description of the subprocess is given in Suppl. Note~\ref{app:ilc}.} 
    \label{fig:metric-selection:s2-multi-class}
\end{figure}

\begin{figure}[H]
    \centering
    \includegraphics[width=1\linewidth]{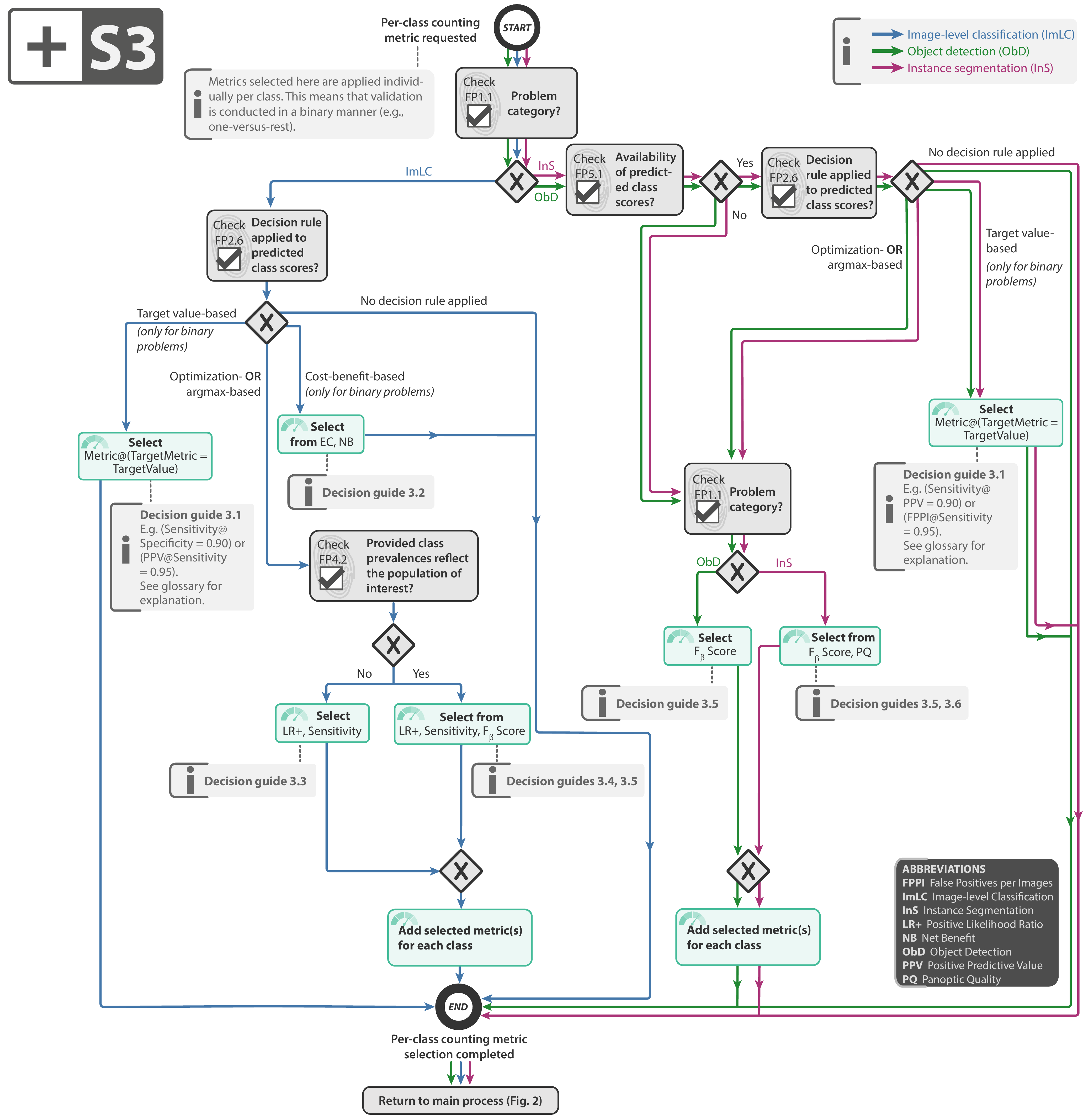}
    \caption{\textbf{Subprocess S3 for selecting a per-class counting metric (if any)}. Applies to: image-level classification (ImLC), object detection (ObD), and instance segmentation (InS). Decision guides are provided in Suppl. Note~\ref{ssec:dg3}. A detailed description of the subprocess is given in Suppl. Notes~\ref{app:ilc}, \ref{app:od} and \ref{app:is}.}
    \label{fig:metric-selection:s3-single-threshold}
\end{figure}

\begin{figure}[H]
    \centering
    \includegraphics[width=1\linewidth]{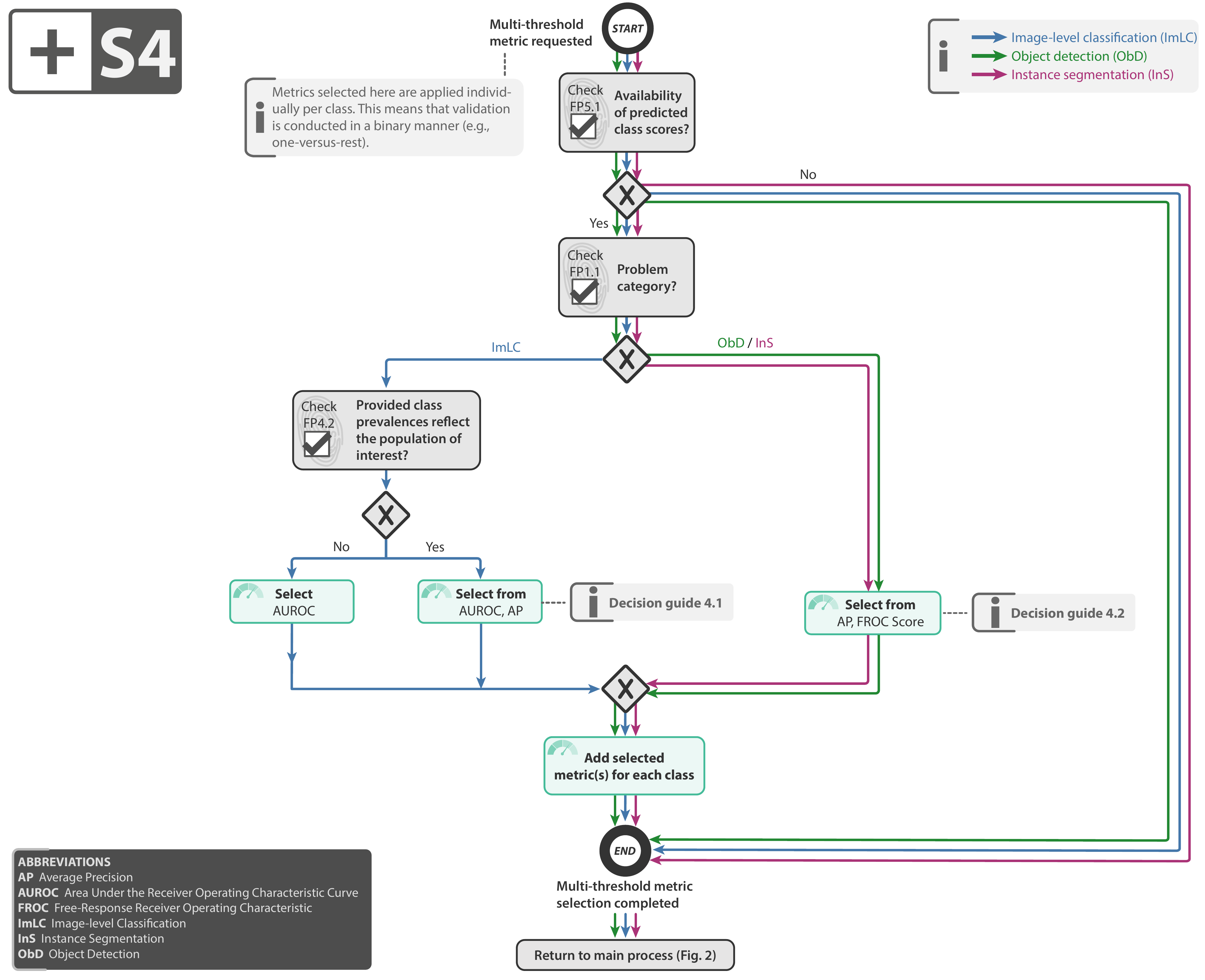}
    \caption{\textbf{Subprocess S4 for selecting a multi-threshold metric (if any)}. Applies to: image-level classification (ImLC), object detection (ObD), and instance segmentation (InS). Decision guides are provided in Suppl. Note~\ref{ssec:dg4}. A detailed description of the subprocess is given in Suppl. Notes~\ref{app:ilc}, \ref{app:od} and \ref{app:is}.}
    \label{fig:metric-selection:s4-multi-threshold}
\end{figure}

\begin{figure}[H]
    \centering
    \includegraphics[width=1\linewidth]{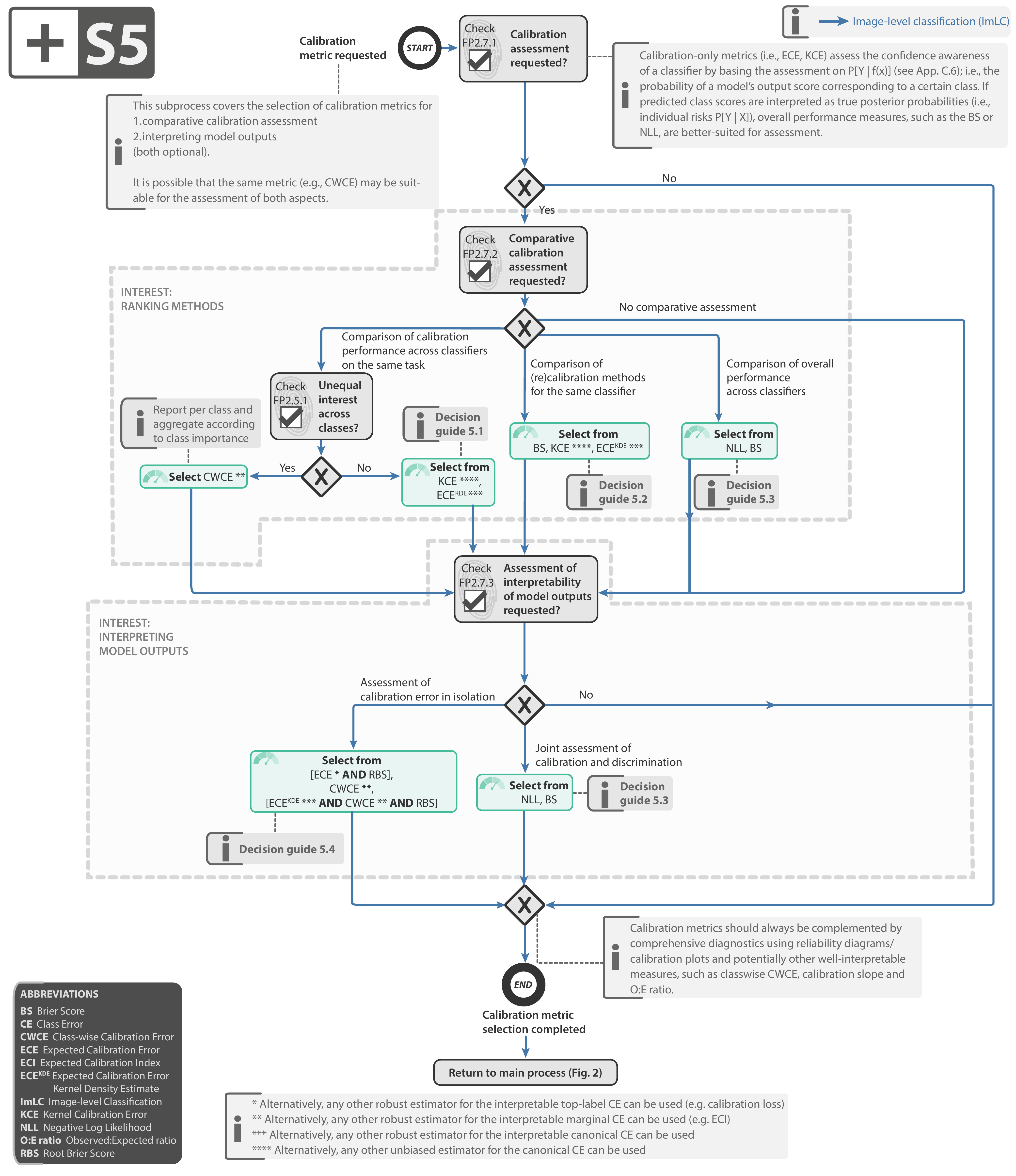}
    \caption{\textbf{Subprocess S5 for selecting a calibration metric (if any)}. Applies to: image-level classification (ImLC). Decision guides are provided in Suppl. Note~\ref{ssec:dg5}. A detailed description of the subprocess is given in Suppl. Note~\ref{app:calibration}. Further suggested calibration metrics include the calibration loss \cite{brummer2006application}, calibration slope \cite{steyerberg2010assessing}, \acf{ECI} \cite{van2015spline} and \acf{O:E ratio} \cite{riley2016external}.}
    \label{fig:metric-selection:s5-calibration}
\end{figure}

\begin{figure}[H]
    \centering
    \includegraphics[width=1\linewidth]{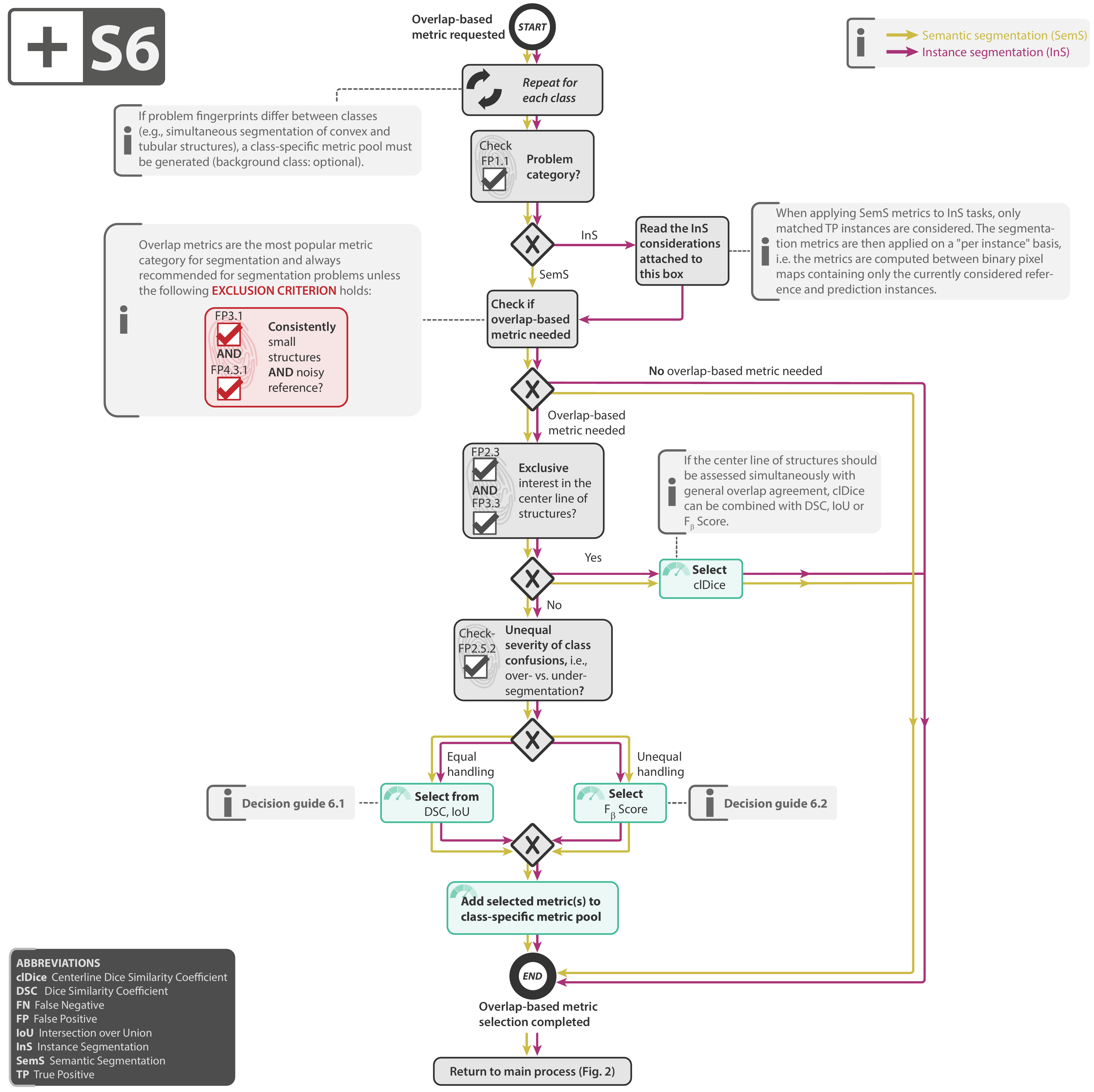}
    \caption{\textbf{Subprocess S6 for selecting overlap-based segmentation metrics (if any)}. Applies to: semantic segmentation (SemS) and instance segmentation (InS). Decision guides are provided in Suppl. Note~\ref{ssec:dg6}. A detailed description of the subprocess is given in Suppl. Notes~\ref{app:ss} and \ref{app:is}.}
    \label{fig:metric-selection:s6-overlap}
\end{figure}

\begin{figure}[H]
    \centering
    \includegraphics[width=1\linewidth]{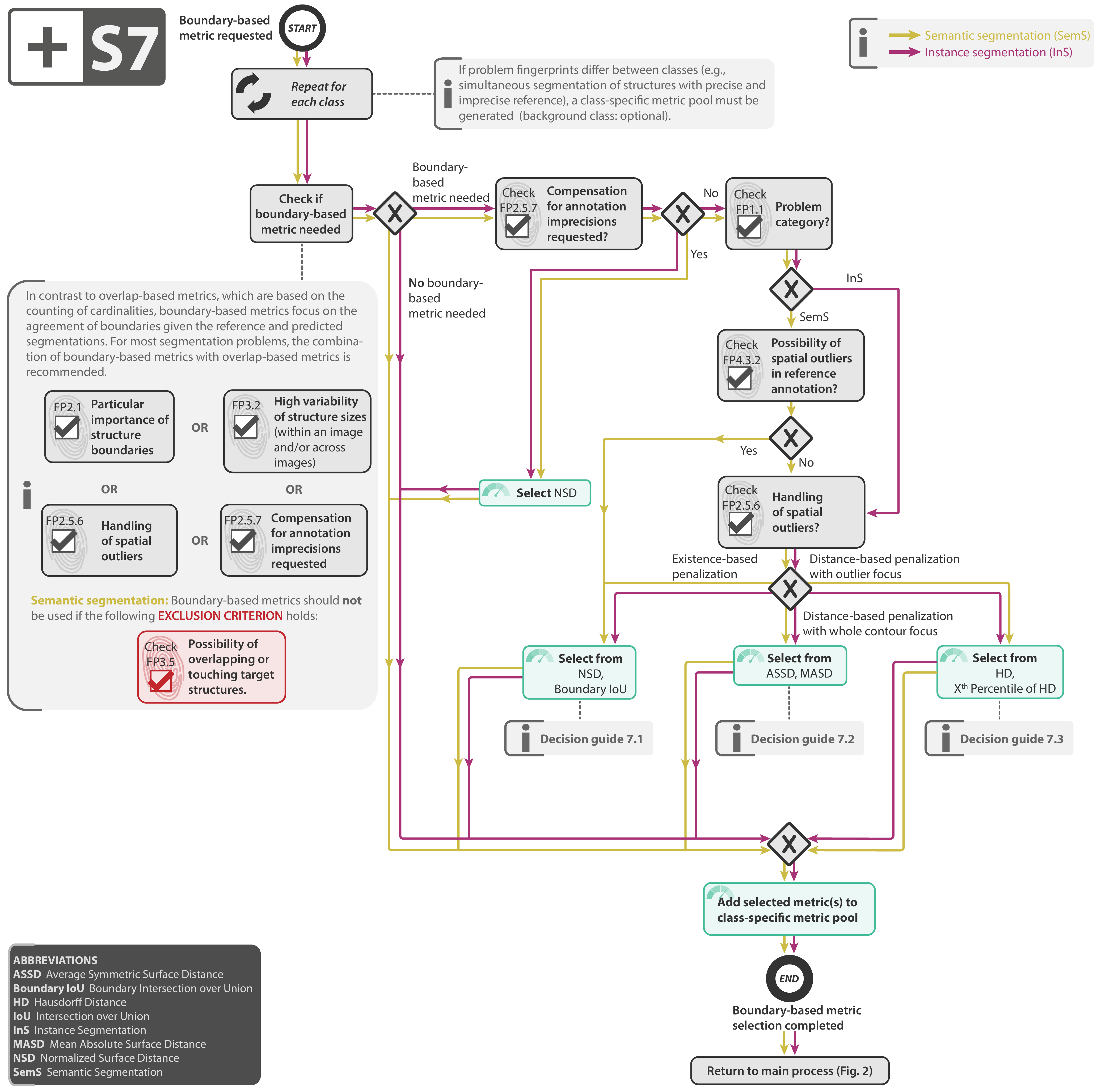}
    \caption{\textbf{Subprocess S7 for selecting a boundary-based segmentation metric (if any)}. Applies to: semantic segmentation (SemS) and instance segmentation (InS). Decision guides are provided in Suppl. Note~\ref{ssec:dg7}. A detailed description of the subprocess is given in Suppl. Notes~\ref{app:ss} and \ref{app:is}.}
    \label{fig:metric-selection:s7-boundary}
\end{figure}

\begin{figure}[H]
    \centering
    \includegraphics[width=1\linewidth]{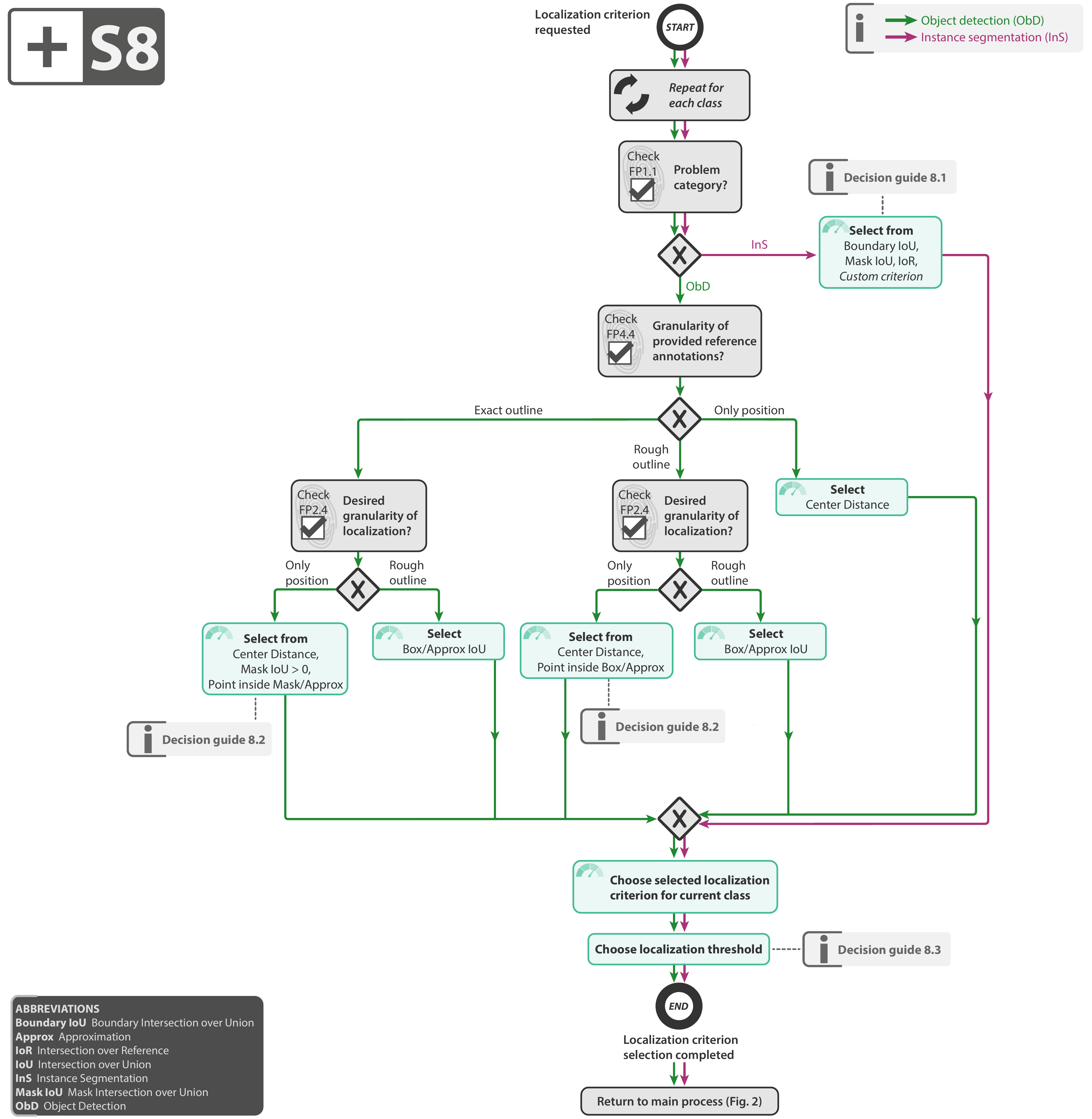}
    \caption{\textbf{Subprocess S8 for selecting the localization criterion}. Applies to: object detection (ObD) and instance segmentation (InS). Definitions of the localization criteria can be found in \cite{reinke2021common}. Decision guides are provided in Suppl. Note~\ref{ssec:dg8}. A detailed description of the subprocess is given in Suppl. Notes~\ref{app:od} and \ref{app:is}.}
    \label{fig:metric-selection:s8-localization-criterion}
\end{figure}

\begin{figure}[H]
    \centering
    \includegraphics[width=1\linewidth]{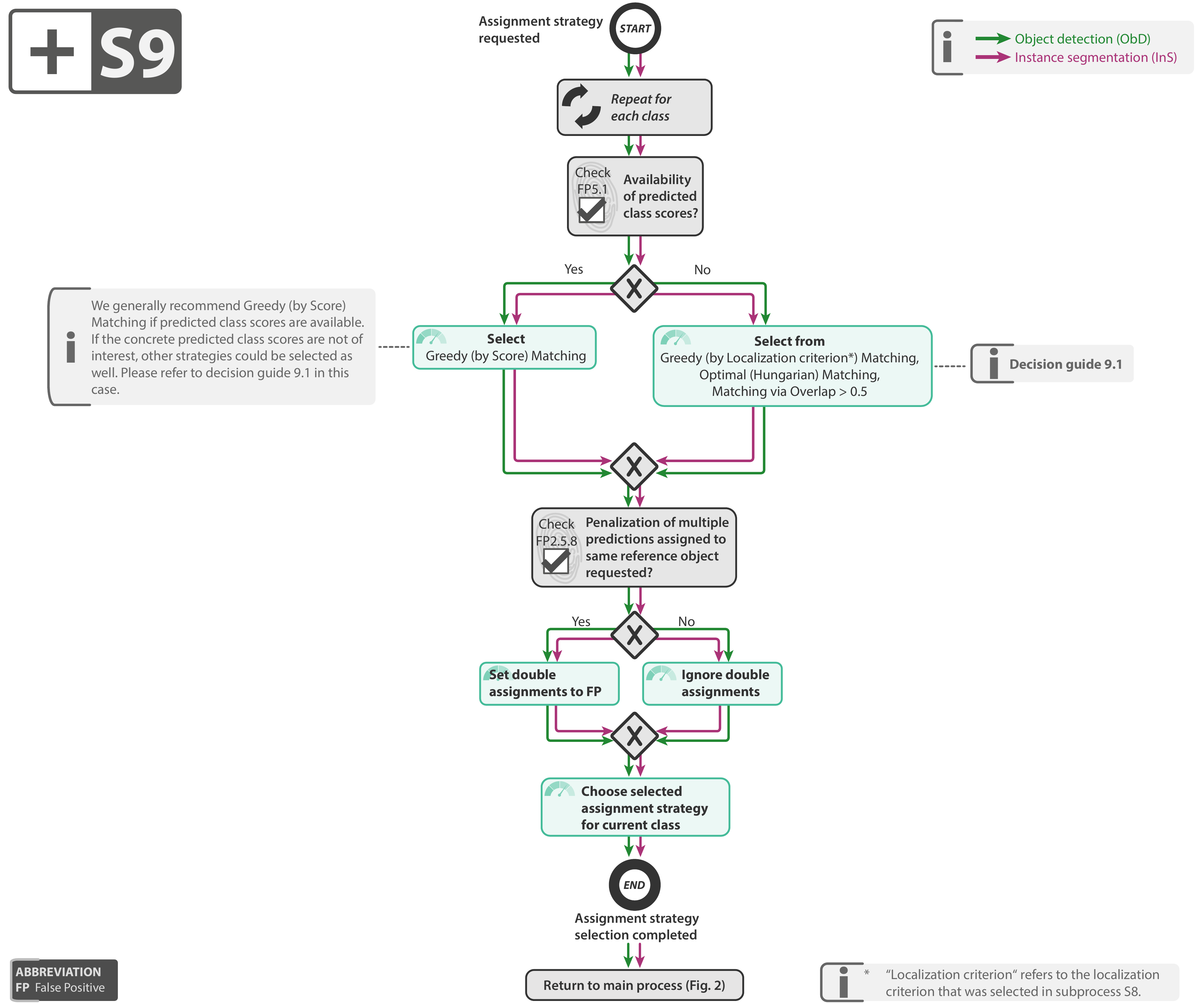}
    \caption{\textbf{Subprocess S9 for selecting the assignment strategy.} Applies to: object detection (ObD) and instance segmentation (InS). Assignment strategies are defined in \cite{reinke2021common}. Decision guides are provided in Suppl. Note~\ref{ssec:dg9}. A detailed description of the subprocess is given in Suppl. Notes~\ref{app:od} and \ref{app:is}.}
    \label{fig:metric-selection:s9-assignment-strategy}
\end{figure}

\newpage

{\small
\begin{longtable}{p{4.0cm}p{10cm}}
%{p{2.0cm}p{1.1cm}p{1.8cm}cccp{1.4cm}cccc}
\caption{\textbf{Recommendations for metric application addressing the pitfalls collected in \cite{reinketizabi2022metrics}.}  The first column comprises \textit{all} sources of pitfalls captured by the published taxonomy that relate to the application of (already selected) metrics. The second column provides the \textit{Metrics Reloaded} recommendation. The notation FPX.Y refers to a fingerprint item (Suppl. Note~\ref{app:fingerprints}).}
\label{tab:cross-topic}\\
\toprule
\textbf{Source of Pitfall} & \textbf{Recommendation} \\
\midrule 
\endfirsthead
%  Recommendations on metric implementation
\rowcolor[HTML]{FFFFFF}
\midrule
\multicolumn{2}{c}{\textbf{Metric implementation}} \\
\midrule
\rowcolor[HTML]{C0C0C0} 
Non-standardized metric definition and undefined corner cases & Use reference implementations provided at \url{https://github.com/Project-MONAI/MetricsReloaded}. \\
\rowcolor[HTML]{EFEFEF}
Discretization issues & Use unbiased estimates of properties of interest if possible (Suppl. Note~\ref{app:calibration}).\\
\rowcolor[HTML]{C0C0C0} 
Metric-specific issues including sensitivity to hyperparameters & Read metric-specific recommendations in the cheat sheets  (Suppl. Note~\ref{app:metric-cheat-sheets}).\\
%  Recommendations on metric aggregation
\rowcolor[HTML]{FFFFFF}
\midrule
\multicolumn{2}{c}{\textbf{Aggregation}} \\
\midrule
 \rowcolor[HTML]{EFEFEF} 
Hierarchical label/class structure & Address the potential correlation between classes when aggregating~\cite{kang2006correlated}.     \\
\rowcolor[HTML]{C0C0C0} 
Multi-class problem & Complement validation with multi-class metrics such as Expected Cost (EC) or Matthews Correlation Coefficient (MCC) with per-class validation (Fig.~\ref{fig:meta-mapping}); perform weighted class aggregation if \textit{FP2.5.1 Unequal interest across classes} holds.  \\
\rowcolor[HTML]{EFEFEF} 
Non-independence of test cases (FP4.5) & Respect the hierarchical data structure when aggregating metrics~\cite{liang1986longitudinal}. \\
\rowcolor[HTML]{C0C0C0} 
Risk of bias & Leverage metadata (e.g. on imaging device/protocol/center) to reveal potential algorithmic bias~\cite{badgeley2019deep}.  \\
 \rowcolor[HTML]{EFEFEF} 
Possibility of invalid prediction (FP5.3) & Follow category-specific aggregation strategy detailed in Suppl. Note~\ref{app:metric-selection}.\\

% Recommendations on metric combination
\rowcolor[HTML]{FFFFFF}
\midrule
\multicolumn{2}{c}{\textbf{Ranking}} \\
\midrule
\rowcolor[HTML]{C0C0C0} 
Metric relationships & Avoid combining closely related metrics (see Fig.~\ref{fig:metric-relations}) when choosing  metrics to be used in algorithm ranking. \\

\rowcolor[HTML]{EFEFEF} 
 Ranking uncertainties & Provide information beyond plain tables that make possible uncertainties in rankings explicit as detailed in \cite{wiesenfarth2021methods}.  \\

\rowcolor[HTML]{FFFFFF}
\midrule
\multicolumn{2}{c}{\textbf{Reporting}} \\
\midrule
\rowcolor[HTML]{C0C0C0} 
Non-determinism of algorithms & Consider multiple test set runs to address the variability of results resulting from non-determinism \cite{khan2019non,summers2021nondeterminism}. 
\\
\rowcolor[HTML]{EFEFEF} 
Uninformative visualization
& Include a visualization of the raw metric values \cite{wiesenfarth2021methods} and report the full confusion matrix unless \textit{FP2.6 = no decision rule applied} holds. \\

%  Recommendations on interpretation
\rowcolor[HTML]{FFFFFF}
\midrule
\multicolumn{2}{c}{\textbf{Interpretation}} \\
\midrule
\rowcolor[HTML]{C0C0C0} 
Low resolution & Read metric-related recommendations to obtain awareness of the pitfall (Suppl. Note~\ref{app:metric-cheat-sheets}). \\
\rowcolor[HTML]{EFEFEF} 
Lack of lower/upper bounds & Read metric-related recommendations to obtain awareness of the pitfall (Suppl. Note~\ref{app:metric-cheat-sheets}).\\
\rowcolor[HTML]{C0C0C0} 
Insufficient domain relevance of metric score differences & Report on the quality of the reference (e.g. intra-rater and inter-rater variability) \cite{kottner2011guidelines}. Choose the number of decimal places such that they reflect both relevance and uncertainties of the reference. More than one decimal number is often not useful given the typically high inter-rater variability. \\
\bottomrule
\end{longtable}}
\normalsize

\section*{Supplementary Information}
\addcontentsline{toc}{chapter}{\protect\numberline{}Supplementary Information}
\setlength{\parskip}{0.5em}

\setcounter{section}{0}

\renewcommand{\thefigure}{SN~\arabic{section}.\arabic{figure}}
\renewcommand{\thetable}{SN~\arabic{section}.\arabic{table}}

\renewcommand*{\thesection}{SUPPL. NOTE \arabic{section}}
\renewcommand*{\thesubsection}{\arabic{section}.\arabic{subsection}}

\begin{tcolorbox}[title=Content of Supplementary Notes]
\small

Our metric recommendations are detailed in the following sections.
\begin{table}[H]
    \centering
    \begin{tabular}{c l}
        % \begin{minipage}{2cm}\includegraphics[width=1.5cm]{images/Pictograms/subprocesses.png}\end{minipage}& \begin{minipage}{10cm}\textbf{Subprocesses (Suppl. Note~\ref{app:subprocesses})} \\ Subprocesses (S1-S9) referred to by the recommendation overview (Fig.~2)\end{minipage} \\
        %  & \\
        \begin{minipage}{2cm}\includegraphics[width=1.5cm]{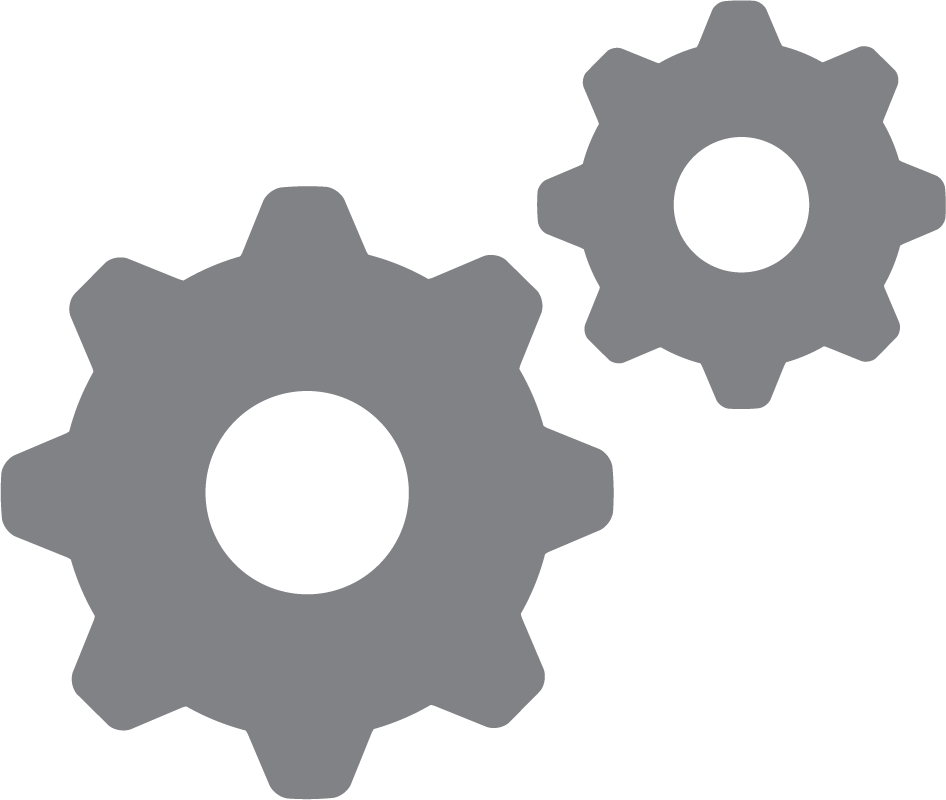}\end{minipage}& \begin{minipage}{10cm}\textbf{SUPPL. METHODS} \\ Delphi process, expert consortium, reference implementation, web-based tool\end{minipage} \\
         & \\
         \begin{minipage}{2cm}\includegraphics[width=0.9cm]{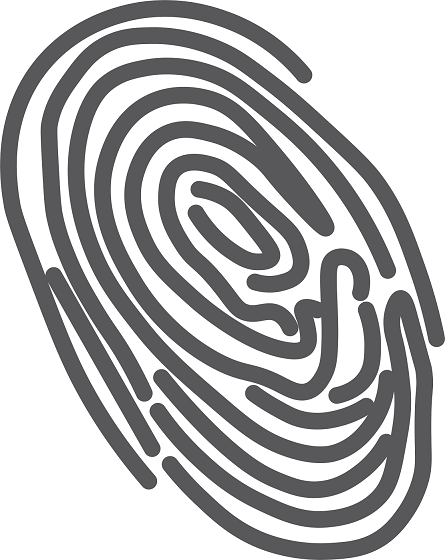}\end{minipage}& \begin{minipage}{10cm}\textbf{Step 1 - Problem Fingerprinting (\ref{app:problem-fingerprinting})} \\ General instructions (Suppl. Note~\ref{app:general-instructions}), problem category mapping (Suppl. Note~\ref{app:fingerprint-category}), generation of the problem fingerprint (Suppl. Note~\ref{app:fingerprints})\end{minipage} \\
         & \\
         \begin{minipage}{2cm}\includegraphics[width=1.3cm]{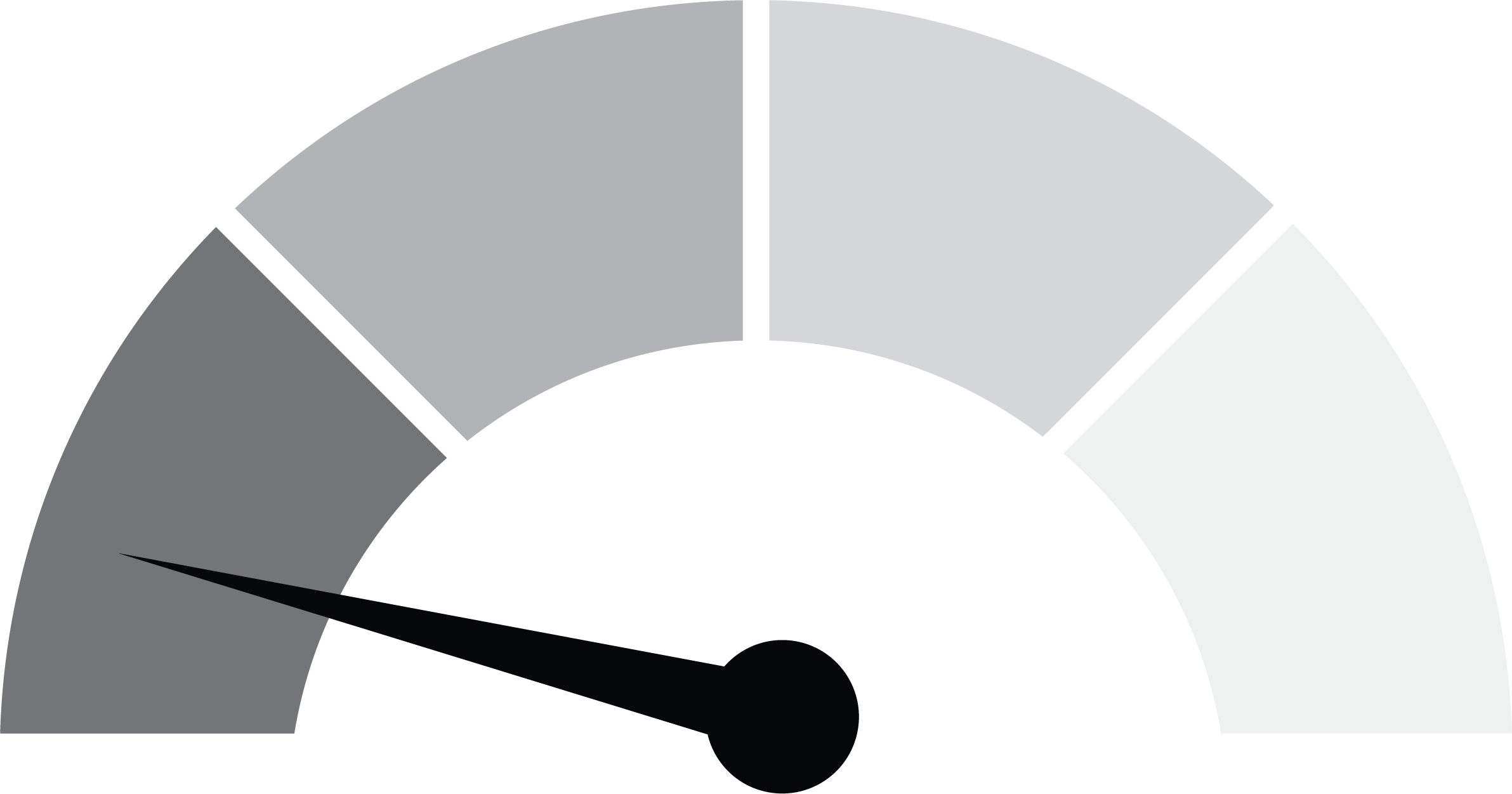}\end{minipage}& \begin{minipage}{10cm}\textbf{Step 2 - Metric Selection (\ref{app:metric-selection})} \\ \textit{Metrics Reloaded} pool of reference-based metrics (Suppl. Note~\ref{app:metric-pool}), recommendations for metric selection (Suppl. Notes~\ref{app:ilc}-\ref{app:calibration}), decision guides (Suppl. Note~\ref{app:decision_guides})\end{minipage}\\ & \\
         \begin{minipage}{2cm}\includegraphics[width=1cm]{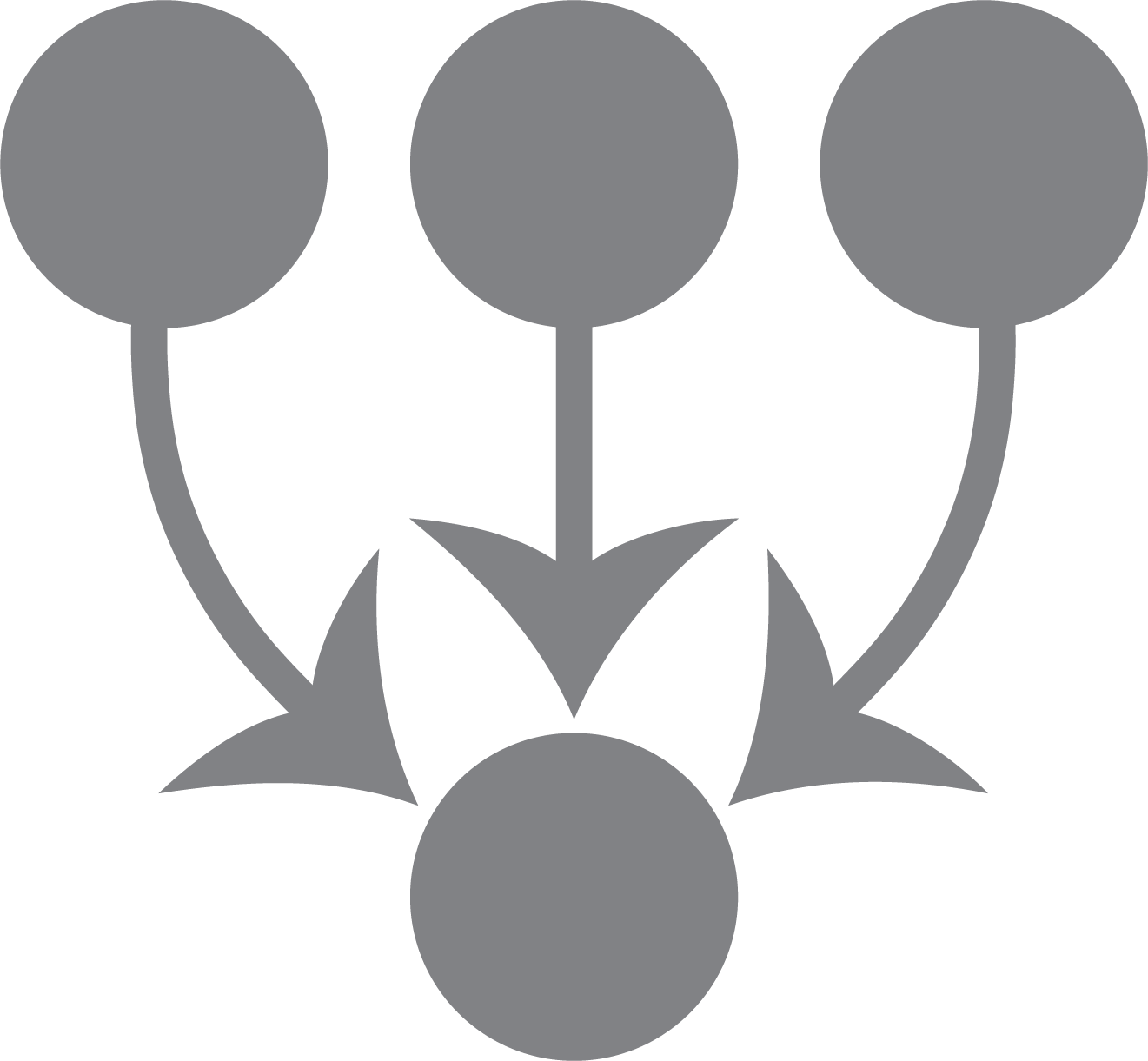}\end{minipage}& \begin{minipage}{10cm}\textbf{Step 3 - Metric Application (\ref{app:metric-application})} \\ Metric cheat sheets (Suppl. Note~\ref{app:metric-cheat-sheets})\end{minipage}\\ & \\
         \begin{minipage}{2cm}\includegraphics[width=1.4cm]{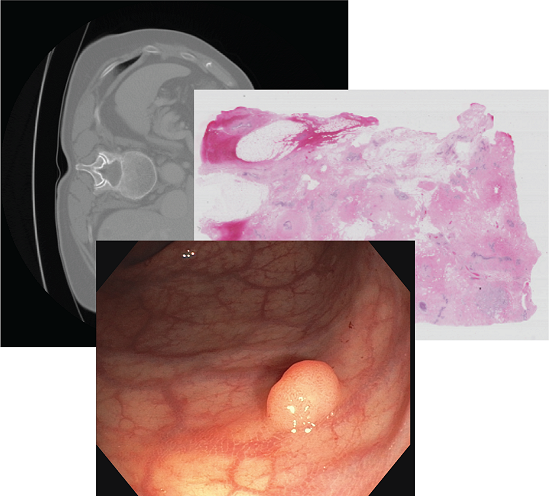}\end{minipage}& \begin{minipage}{10cm}\textbf{Recommendations for selected use cases (\ref{app:scenarios})}\end{minipage} \\
         & \\
         \begin{minipage}{2cm}\includegraphics[width=0.9cm]{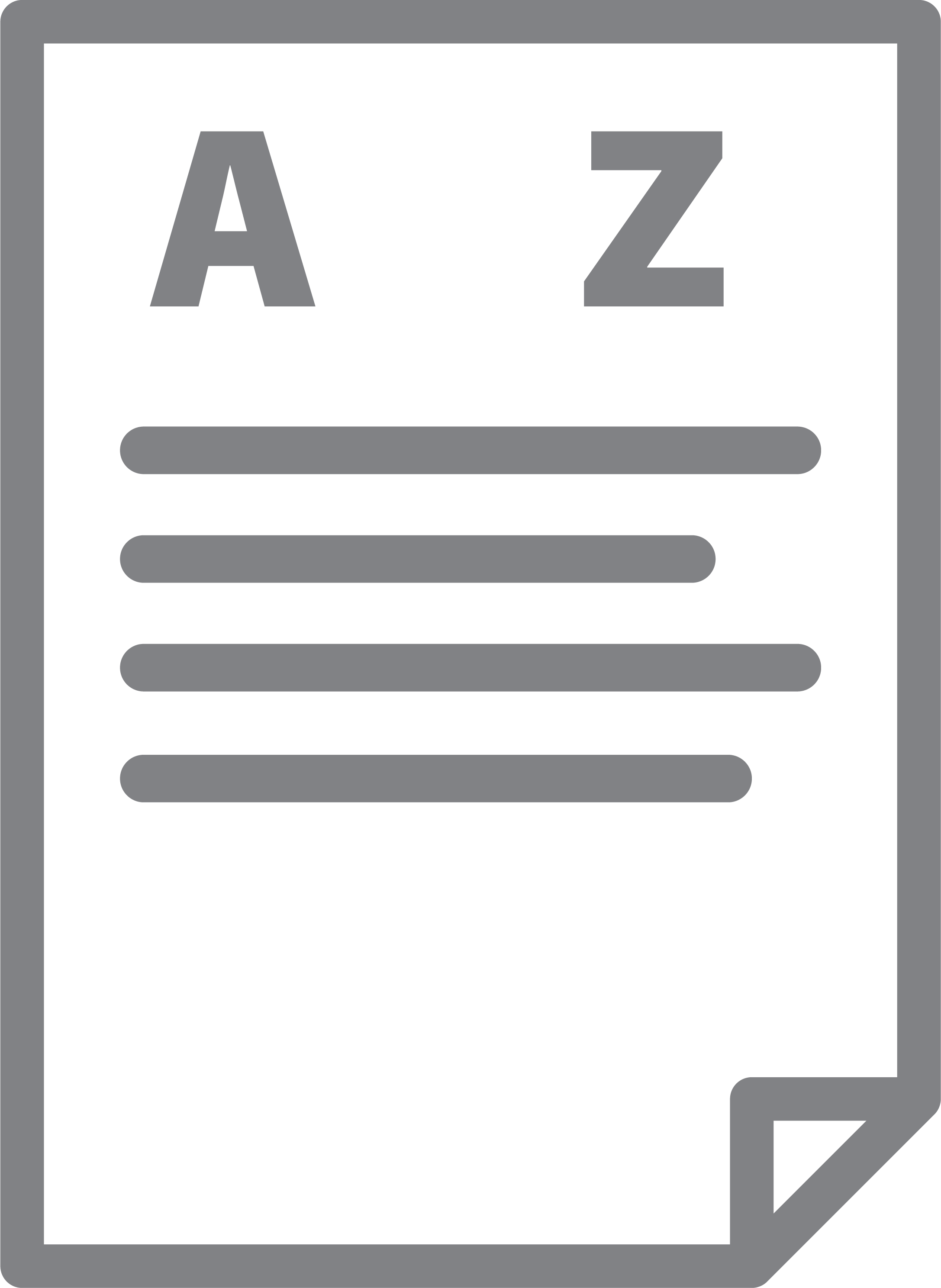}\end{minipage}& \begin{minipage}{10cm}\textbf{Terminology and Notation (\ref{app:terminology-notation})} \\ Symbols (Suppl. Note~\ref{app:symbols}), acronyms (Suppl. Note~\ref{app:acronyms}), glossary (Suppl. Note~\ref{app:terminology})\end{minipage}\\ & \\
         % \begin{minipage}{2cm}\includegraphics[width=1.4cm]{images/Pictograms/authors.png}\end{minipage}& \begin{minipage}{10cm}\textbf{Full Author Affiliations (Suppl. Note~\ref{app:authors})}\end{minipage} \\
    \end{tabular}
    %\caption{Caption}
    %\label{tab:my_label}
\end{table}

\begin{table}[H]
    \centering
    \scriptsize
    \captionsetup{labelformat=empty}
    \caption*{\textit{Overview of relevant content for \acf{ImLC}, \acf{SemS}, \acf{ObD} and \acf{InS} problems.}}
    \label{tab:overview-table}
    \renewcommand{\arraystretch}{2}
    \begin{tabular}{P{2.4cm} P{2.1cm} P{2.1cm} P{2.1cm} P{2.1cm}}
        \toprule
        & \textbf{\acs{ImLC}} &\textbf{ \acs{SemS}} &\textbf{ \acs{ObD}} & \textbf{\acs{InS}} \\ \midrule
        \textbf{Problem fingerprint (\ref{app:problem-fingerprinting})} & Figs.~\ref{fig:fingerprint-ilc}~-~\ref{fig:fingerprint-ilc-3} & Figs.~\ref{fig:fingerprint-ss-1}~-~\ref{fig:fingerprint-ss-2}& Figs.~\ref{fig:fingerprint-od-1}~-~\ref{fig:fingerprint-od-3} & Figs.~\ref{fig:fingerprint-is-1}~-~\ref{fig:fingerprint-is-3}\\
        \textbf{Recommendations for metric selection (\ref{app:metric-selection})} & Suppl. Note~\ref{app:ilc}; \linebreak Calibration: Suppl. Note~\ref{app:calibration} & Suppl. Note~\ref{app:ss} & Suppl. Note~\ref{app:od} & Suppl. Note~\ref{app:is} \\
        % \textbf{Subprocesses (Suppl. Note~\ref{app:subprocesses})} & S2~-~S5 (Figs.~\ref{fig:metric-selection:s2-multi-class}~-~\ref{fig:metric-selection:s5-calibration}) & S6~-~S7 (Figs.~\ref{fig:metric-selection:s6-overlap}~-~\ref{fig:metric-selection:s7-boundary})& S3~-~S5, S8~-~S9 (Figs.~\ref{fig:metric-selection:s3-single-threshold}~-~\ref{fig:metric-selection:s5-calibration}, \ref{fig:metric-selection:s8-localization-criterion}~-~\ref{fig:metric-selection:s9-assignment-strategy})& S3~-~S9 (Figs.~\ref{fig:metric-selection:s3-single-threshold}~-~\ref{fig:metric-selection:s9-assignment-strategy})\\
        % \textbf{Decision guides (denoted as DGX.Y; Suppl. Note~\ref{app:decision_guides})} & 2.1~-~2.3, 3.1~-~3.5, 4.1, 5.1~-~5.3& 6.1~-~6.2, 7.1~-~7.3 & 3.1, 3.5, 4.2, 5.1~-~5.3, 8.2~-~8.3, 9.1 & 3.1, 3.5, 3.6, 4.2, 5.1~-~5.3, 6.1~-~6.2, 7.1~-~7.3, 8.1, 8.3, 9.1\\
        \textbf{Metric cheat sheets (\ref{app:metric-application})} & Suppl. Note~\ref{app:metric-cheat-sheets} & Suppl. Note~\ref{app:metric-cheat-sheets} & Suppl. Note~\ref{app:metric-cheat-sheets} & Suppl. Note~\ref{app:metric-cheat-sheets} \\
        \textbf{Instantiation for common use cases (\ref{app:scenarios})} & Fig.~\ref{fig:scenario-ImLC-overview} & Fig.~\ref{fig:scenario-SemS-overview} & Fig.~\ref{fig:scenario-ObD-overview} & Fig.~\ref{fig:scenario-InS-overview} \\
         \bottomrule
    \end{tabular}
\end{table}

\end{tcolorbox}

\newpage
%------------------------------------
\section*{Supplementary Methods}
\addcontentsline{toc}{section}{\protect\numberline{}Supplementary Methods}
%------------------------------------
\acresetall
\subsection*{Delphi process}
\label{ssec:delphi}

We compiled the recommendations provided by the \textit{Metrics Reloaded} framework by assembling an international expert consortium which then underwent a multi-stage Delphi process. A Delphi process is a structured group communication process that serves to gather opinions from an expert panel via a series of individual interrogations, usually in the form of questionnaires, interspersed with feedback from the respondents \cite{brown1968delphi}. The technique is widely used for establishing consensus among experts in medicine, particularly in the development of best practices in areas where evidence may be limited, conflicting, or absent \cite{nasa2021delphi}. The initial panel participating in our Delphi process comprised 30 international biomedical image analysis experts representing 25 institutions. Member selection was initially based on membership in one of the three initiatives that triggered this research, namely the Biomedical Image Analysis Challenges (BIAS) initiative, the \ac{MONAI} Working Group for Evaluation, Reproducibility and Benchmarks, and the \ac{MICCAI} Special Interest Group for Challenges (previously \ac{MICCAI} board working group). To reflect as broad a range of imaging domains as possible and expand the available expertise, the number of consortium members was gradually increased from the initial 30 to a final number of 73. The members provided a wide range of expertise ranging from biology, medicine, epidemiology and biomedical image analysis all the way to statistics, mathematics and computer science. Furthermore, leading members of major standardization initiatives were included, such as the \ac{EQUATOR} network, 
%TODO complement
from which imaging and clinical guidelines have originated, including CONSORT/CONSORT-AI~\cite{schulz2010consort, CONSORT-AI_and_SPIRIT-AI_Steering_Group2019-om}, TRIPOD/TRIPOD-AI ~\cite{moons2015transparent,collins2021protocol}, STARD/STARD-AI ~\cite{bossuyt2003towards,Sounderajah2020-ou}, and others.

% If we want to include it, we could add the following information:
% Technical: 66,67\%	
% Technical/Clinical: 13,89\%	
% Technical/Biological: 5,56\%	
% Technical/Clinical/Biological: 4,17\%	
% Clinical: 6,94\%
% Biological: 2,78\%

Overall, the process comprised six distinct stages and encompassed five workshops and nine surveys before a final Delphi consensus voting was performed. Each survey was developed by the Metrics Reloaded core team and taken by the remaining members of the consortium; in other words, the researchers that designed the surveys did not take part in them. Upon completion, the core team then analyzed the results, discussed them with team members where necessary, and integrated the feedback, thus iteratively refining the framework.
The main stages of the compilation and consensus building process are detailed in the following:

\paragraph{1. Initialization} A kickoff workshop was held in December 2020 with the primary goal of deciding on the concrete scope of the recommendation framework. 
Prior to the workshop, an initial survey had been conducted with a focus on gathering relevant literature as well as theoretical and practical failure cases of metrics in the broader scope of classification, segmentation and detection.
Based on the discussions at the workshop, a series of three surveys was issued whose responses resulted in (1) a joint terminology (see Suppl. Note~\ref{app:terminology}), (2) inclusion criteria for the paper, namely the decision to cover classification tasks at image/object and pixel level (Fig.~\ref{fig:problem_categories}), (3) a shortlist of relevant metrics for each category (whose subsequent refinement resulted in Tab.~\ref{app:metric-pool}), and (4) an initial set of fingerprint items, whose refinement resulted in the final fingerprints presented in Suppl. Note~\ref{app:fingerprints}. It was further decided by the consortium that the choice of problem category should be part of the recommendation itself (covered in Subprocess S1, Extended Data Fig.~\ref{fig:category_mapping}).

\paragraph{2. Compilation of first draft of recommendations in expert groups} The primary purpose of the second Delphi workshop in June 2021 was the formation of expert groups that should coordinate individual task forces. Five expert groups were initially formed; three dedicated to the problem categories addressed in the framework (one for image-level classification, one for semantic segmentation and one for object detection and instance segmentation), plus a biomedical expert group and a cross-topic expert group. The task of the expert groups corresponding to the problem categories was to develop recommendations for their respective  categories that address the pitfalls compiled in the sister publication \cite{reinketizabi2022metrics} (now captured in Fig.~\ref{fig:meta-mapping} and the nine subprocesses in Extended Data Figs.~\ref{fig:category_mapping}~-~\ref{fig:metric-selection:s9-assignment-strategy}). The task of the cross-topic group was to identify and tackle metric-related issues going beyond pure metric selection, such as metric aggregation, reporting and implementation, statistical considerations, rankings, and biases (now captured in Extended Data Tab.~\ref{tab:cross-topic}). The task of the biomedical expert group was to ensure that the recommendation framework would satisfy the needs of domain experts, such as clinicians, and to identify relevant biomedical scenarios (now captured in \ref{app:scenarios}). Group-specific surveys were issued to support the work of the task forces. To give the experts enough freedom, no specific restrictions were imposed with respect to how the individual groups would arrive at their recommendations. In the following third Delphi workshop, held in October 2021, the expert group leaders discussed their preliminary results with the entire core team. 

\paragraph{3. Consolidation of recommendations by Metrics Reloaded core team} Once the expert groups had finalized their initial recommendation drafts for the problem categories, the \textit{Metrics Reloaded} core team consolidated and harmonized the recommendations in close collaboration with the groups. In the fourth Delphi workshop in January 2022, the resulting decision trees capturing the core recommendations (Fig.~\ref{fig:meta-mapping}; S1-S4, S6-S9) were presented and discussed.

\paragraph{4. Revision by Metrics Reloaded consortium} The decision trees were then subjected to internal tests by members of the consortium and their teams. The Metrics Reloaded core team incorporated the survey-based feedback in close collaboration with the expert groups. The first draft of the entire framework was then presented and discussed at the fifth Delphi workshop in March 2022. 

\paragraph{5. Crowdsourcing of  feedback} Finally, community feedback was obtained via a social media campaign. The recommendation framework was released on arXiv~\cite{maier2022metrics}, and a survey link was sent by the \textit{Metrics Reloaded} core team to various mailing lists, as well as posted on the social media platforms LinkedIn, with the hashtag \#imageanalysis, and Twitter, where we tagged various relevant accounts, e.g., @MICCAIStudents, @WomenInMICCAI, @midl\_conference, @ELLISforEurope, @ProjectMONAI, and @naturemethods. The original tweet received more than 42,000 impressions. All co-authors were asked to distribute the survey among their colleagues and societies. Furthermore, the survey link was added to the released arXiv version. The survey was initially opened at the beginning of June 2022 and closed at the end of July 2022. The community had the choice between submitting one-click feedback or detailed feedback by answering questions on the comprehensiveness and usefulness of our approach, the specific mappings, as well as voicing concerns or questions. In addition, we specifically asked for which biomedical use cases the framework should be instantiated. All contributors were given the choice to be included in the acknowledgements.  total of 186 researchers participated in the survey. Of those, 82 provided feedback in the form of free text answers. 58 participants chose to give detailed feedback rather than one-click feedback. A total of 46 researchers wished to be mentioned in the acknowledgements and provided their names. Contributors who provided substantial feedback were invited into the consortium (seven in total). The social media survey was used as a basis to select biomedical use cases for which the framework was instantiated. Based on the feedback, we designed the metric cheat sheets (see \ref{app:scenarios}). The implementation of the web toolkit was highly encouraged by several survey participants. Moreover, in response to the feedback, an additional expert group on the topic of calibration was established with newly recruited consortium members, which led to the generation of the calibration recommendations captured in S5 (Extended Data Fig.~\ref{fig:metric-selection:s5-calibration}). A revised framework, with the community feedback integrated (e.g., including new classification metrics, such as the EC), was presented to the consortium in another survey, based on which the Metrics Reloaded core team compiled the final recommendations (captured in Fig.~\ref{fig:meta-mapping} and S1-S9) that served as basis for the final Delphi-based consensus building.

\paragraph{6. Final Delphi consensus building} In the final stage, an accelerated final Delphi process was initiated to vote for the ten core components of the recommendation framework (Fig.~\ref{fig:meta-mapping} and Subprocesses S1-S9). In response to the consortium's comments, final modifications to the calibration recommendations were made. After two rounds of revisions to S5, the final recommendation received strong support (only one member disagreed). For all other nine components, the first round had already resulted in a very strong consensus (disagreement 0\%-7\%). Minor modifications, primarily concerning formatting and style, were communicated to the entire consortium whose members were then given the opportunity to veto any of the changes, which none of the consortium made use of. 

\subsection*{Expert consortium}
\label{ssec:experts}

he expert consortium consisted of a total of 73 researchers (73\% male, 27\% female) from a total of 65 institutions. The majority of experts (52\%) were professors, followed by postdoctoral researchers (37\%). The median h-index of the consortium was 34 (mean: 27; minimum: 6; maximum: 113) and the median academic age was 18 years (mean: 19; minimum: 3; max: 42). Experts were from 18 countries and 5 continents. 66\% of experts had a technical, 7\% a clinical, 3\% a biological, and 24\% a mixed background. From the 65 institutions, we could identify the number of employees for 88\% of institutions. From those, the majority of institutions had a size between 1,000 and 10,000 employees (58\%), followed by even larger institutions between 10,000 and 100,000 employees (25\%), and smaller institutions below 1,000 employees (16\%). Only a small portion of institutions were above 100,000 employees (2\%).

\subsection*{Reference implementations}
\label{ssec:implementation}

To overcome pitfalls related to metric implementation~\cite{reinketizabi2022metrics}, we provide reference implementations for all \textit{Metrics Reloaded} metrics within the \ac{MONAI} open-source framework. They are accessible at \url{https://github.com/Project-MONAI/MetricsReloaded}. %Most metrics are included in the \texttt{metrics/pairwise\_measures.py} file, which also include all necessary hyperparameters for each metric. Metrics that are measured for a specific target value can be found in the \texttt{metrics/prob\_pairwise\_measures.py} file (e.g. Sensitivity@Specificity). Calibration metrics are included in the \texttt{metrics/calibration\_measures.py} file. Generally, warnings are included to point researchers to potential design designs. For example, if a metric would turn \texttt{\ac{NaN}} due to division by zero errors, the package will give a warning such as "\texttt{prediction empty, PPV not defined but set to 0}".

\subsection*{Web-based tool}
\label{ssec:toolkit}

The recommendation framework was implemented as a web-based tool, which guides the users through the entire recommendation processes of Fig.~\ref{fig:meta-mapping}. The core advantage of the tool compared to the decision trees depicted in S1-S9 is the fact that the tool automatically restricts the visualization only to the relevant information that is required in each specific step and for the specific use case. It further provides comprehensive profiles of all metrics contained in the \textit{Metric Reloaded} pool.

The \textit{Metric Reloaded} tool is available at \url{https://metrics-reloaded.dkfz.de}.%(for reviewers, please use: username: mrtest; password: mr\_imsy\_69120).

%

%mr\_imsy\_69120. 
%One of the most important aspects of the web-based tool is its user-friendly interface. We recruited a professional UI/UX team to develop the tool's interface. The interface uses clear and concise language to explain each step of the process, and includes helpful visual aids, such as flowcharts and diagrams, to make the information easy to access. 
%As a result, users can simply access the functions they want without losing time browsing to locate them or going through confusing menus.

% \newpage
% %------------------------------------
% \section{SUBPROCESSES}
% \label{app:subprocesses}
% %------------------------------------
% This section contains the subprocesses (S1-S9) referred to by the recommendation overview (Fig.~\ref{fig:meta-mapping}) and represented by the $\boxplus$-symbol. They capture the \textit{Metrics Reloaded} core recommendations in static flowcharts, while the online tool (see Online Methods) serves as a user-centric alternative. 
% %We first present the problem category mapping in Suppl. Note~\ref{app:fingerprint-category}. The formal decision trees (subprocesses) are shown in Suppl. Note~\ref{app:subprocesses-metric-selection}.
% \input{Subprocesses.tex}

%------------------------------------
\section{STEP 1~-~PROBLEM FINGERPRINTING}
\label{app:problem-fingerprinting}
\setcounter{figure}{0}
%------------------------------------
\acresetall
The first step in the framework \textit{Step 1: Problem fingerprinting} (Fig.~\ref{fig:meta-mapping}) requires the user to read the general instructions provided in Suppl. Note~\ref{app:general-instructions}, perform the problem category mapping according to Suppl. Note~\ref{app:fingerprint-category}, and generate the corresponding problem fingerprint as detailed in Suppl. Note~\ref{app:fingerprints}. 

%\input{EssentialsBox.tex}

%------------------------------------
\subsection{General Instructions}
\label{app:general-instructions}
%------------------------------------
\acresetall
Users of the \textit{Metrics Reloaded} framework should read the following instructions prior to metric selection.  
 
\paragraph{Inclusion criteria}
The \textit{Metrics Reloaded} framework currently considers problems in which categorical target variables are to be predicted based on a given \textit{n}-dimensional input image. Hence, it covers a broad range of imaging modalities from classical 2D/3D modalities, such as fluorescence, \ac{CT} or X-ray imaging, to novel, for example spectral, imaging modalities that yield high-dimensional output per pixel~\cite{clancy2020surgical}. Classification can occur at pixel, object or image level, resulting in the four problem problem categories covered by the framework and depicted in Fig.~\ref{fig:problem_categories}: 
\begin{description}
    \item[Image-level classification] refers to the assignment of one or multiple category labels to the entire image or fixed regions/predefined locations within an image. 
    \item[Semantic segmentation] refers to the assignment of one or multiple category labels to each pixel. For many segmentation problems, object boundaries are generated in addition to the pixel-wise classification images, which enables the computation of distance-based metrics, such as the \ac{NSD}. 
    \item[Object detection] refers to the localization and categorization of an unknown number of structures.  
    \item[Instance segmentation] refers to the localization and delineation of each distinct structure of a particular class. It can be regarded as delivering the tasks of object detection and semantic segmentation at the same time. In contrast to object detection, instance segmentation also involves the accurate marking of the structure boundary. In contrast to semantic segmentation, it distinguishes different structures of the same class. 
\end{description}
Notably, the four different categories are mathematically closely related (Fig.~\ref{fig:problem_categories}) as they typically rely on the generation of confusion matrices as a foundation of metric computation. Application examples for all categories can be found in Fig.~\ref{fig:scenario-table}.
Importantly, \textit{Metrics Reloaded} does not require an entire image to be provided as input for the validation. For example, the classification of a \ac{ROI} within a medical image may be required. In this example, the framework would proceed with the \ac{ROI} as input as if it was an entire image. Furthermore, the shape of the image/input does not need to be rectangular. Finally, context information may be provided along with the input. For example, medical images may be processed along with clinical data to arrive at a diagnosis; video frames may be processed along with preceeding video snippets. Ultimately, only the algorithm \textit{output} must correspond to an \textit{n}-dimensional image. 

\paragraph{Phrasing of the biomedical task}
The recommendation framework has been designed in a way to support the metric selection and application process for one specific driving biomedical question. In practice, multiple questions are often addressed with one given data set. For example, a clinician may have the ultimate interest of diagnosing brain cancer in a patient based on a given \ac{MRI} data set. While this would be phrased as an image-level classification task, an interesting \textit{surrogate task} could be that of segmentation to assess the quality of tumor delineation. In the case of multiple different driving biomedical questions, a recommendation is generated separately for each question. This specifically holds true for multi-label problems, in which multiple labels can simultaneously be assigned to the same image/object/pixel (e.g., multiple sclerosis and brain tumor both assigned to the same magnetic resonance image). In such a case, the problem should be converted to multiple binary problems, for which the framework is traversed individually.

\paragraph{Matching reference annotations}
The metric selection process begins with the step of mapping a given problem with all its intrinsic and data set-related properties to the corresponding problem category via the \textit{category mapping} shown in Fig.~\ref{fig:category_mapping}. Our framework assumes that the reference annotations of the given data set meet the requirements of the identified problem. Expected formats for both the reference annotations and the algorithm output are provided in Suppl. Note~\ref{app:formats}. 

\paragraph{Model-agnostic metric recommendation}
Metrics should be chosen based solely on the driving biomedical problem and not be affected by algorithm design choices. For example, the error functions applied in common neural network architectures do not justify the use of corresponding metrics (e.g., validating with \ac{DSC} to match the Dice loss used for training a neural network). Instead, the domain interest should guide the choice of metric, which, in turn, can guide the choice of the loss term. 

\paragraph{Dealing with multiple classes} 
 Multi-class metrics, such as Accuracy or \ac{MCC}, have the unique advantage that they capture the performance of an algorithm for all classes in a single score without the need for customized class-aggregation schemes. On the other hand, they do now allow for detailed class-specific analyses. \textit{Metrics Reloaded} therefore generally recommends performing a per-class validation for all classes (in addition to potential multi-class validation). Specifically in segmentation problems, problem properties may differ from class to class (e.g. the size or size variability of target structures). In these rare cases, the problem fingerprint needs to be generated separately for each class and several subprocesses (denoted by the $\boxplus$-symbol in the framework overview shown in Fig.~\ref{fig:meta-mapping}) need to be traversed separately for each class. Although not common in current validation practice, this may - in theory - lead to different validation metrics for different classes. We speak of \textit{class-specific metric pools} in this case, which are generated in addition to the multi-class metric pool.

\paragraph{Primary and secondary metrics} 
In general, biomedical interest cannot be captured with a single metric. The framework has therefore been designed to recommend multiple complementary metrics for a given task. We assume two main use cases for our framework. In \textbf{comparative benchmarking studies} (e.g., competitive challenges), multiple algorithms or algorithm variants are compared on identical data sets. This requires the ranking of the competing algorithms according to performance. Typically, multiple complementary validation metrics are applied in this use case, resulting in either multiple rankings or a merged ranking that takes all or several metric values into account. We refer to the metrics that contribute to the (primary) ranking(s) as \textit{primary metrics}. While our framework focuses on the recommendation of primary metrics, users are invited to complement them with  \textit{secondary metrics} according to their specific needs. Secondary metrics can additionally be applied for comprehensive reporting, for example because they reflect complementary properties of interest (e.g., compute time, carbon footprint), or for providing performance measures that are comparable across publications. The computer vision community, for instance, typically reports the \ac{IoU} rather than the \ac{DSC}. The second use case of metrics addressed by our framework are \textbf{validation studies centered around a single algorithm} that focus on comprehensive diagnostics rather than comparative assessment. In this case, it is often desired to report as many complementary metrics as possible in order to comprehensively analyze the properties of an algorithm. Users interested in this second use case can ignore the discrimination between primary and secondary metrics. 

%\newpage
\paragraph{Decision rule applied to predicted class scores} 
A classification system in practice operates by making decisions. Converting the raw continuous model outputs -- the predicted class scores -- into discrete decisions is achieved by determining an appropriate decision rule. Common options are detailed in Suppl. Note~\ref{app:fingerprints} ($\rightarrow$ FP2.6: Decision rule applied to predicted class scores). While identifying the optimal decision rule for a classification system is beyond the scope of this work, it is important to know that the choice affects the selection of adequate validation metrics. 

\begin{figure}[H]
   \centering
   \includegraphics[width=0.8\textwidth]{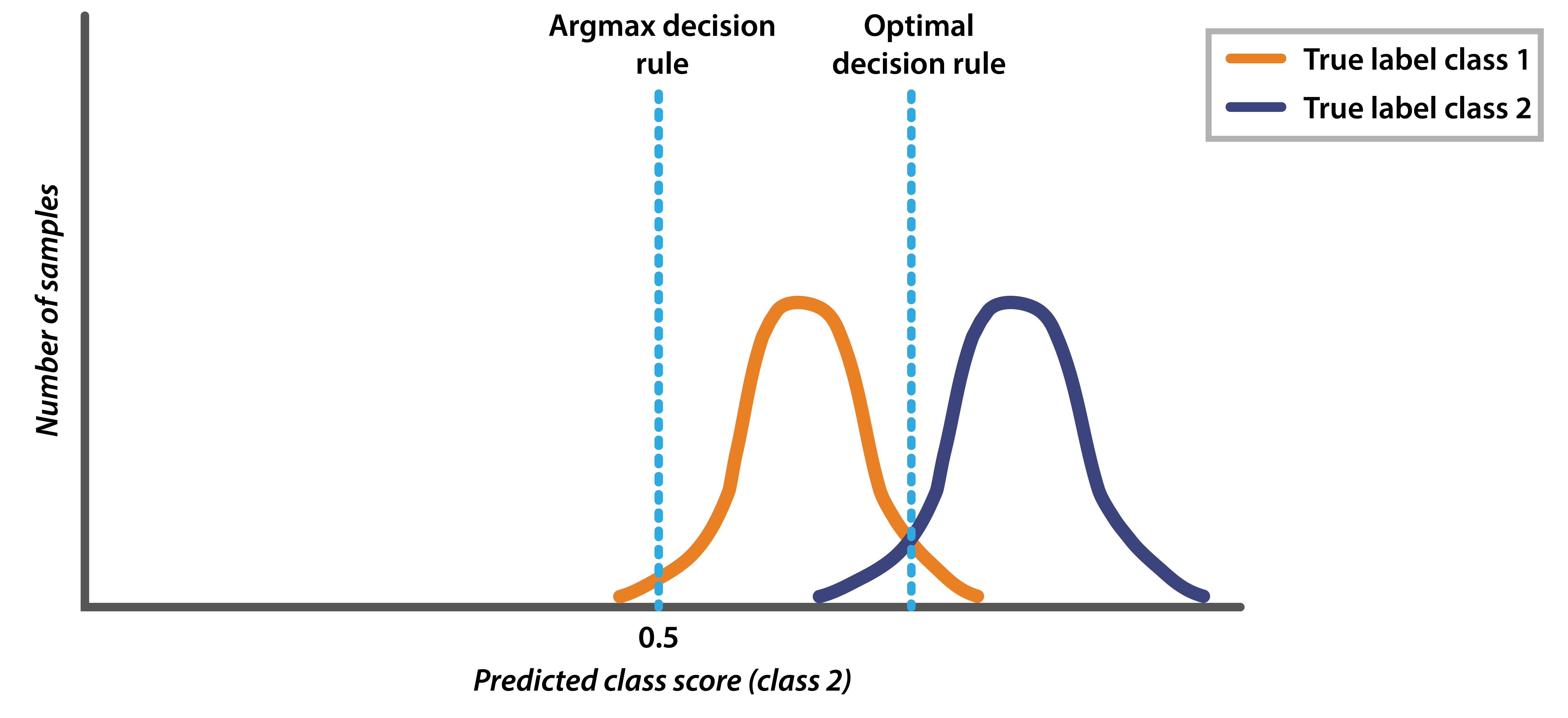}
   \caption{Argmax decision rule for converting predicted class scores to a categorical label. Choosing the class with the highest predicted class score does not necessarily result in the best decision-making, as for example measured by Accuracy. 
   \label{fig:top-label-pitfall}}
\end{figure}

In binary tasks, defining an optimal decision rule boils down to determining a suitable cutoff (i.e., threshold) on predicted class scores (Fig.~\ref{fig:top-label-pitfall}). In contrast, identifying an optimal decision rule for multi-class problems is generally more complex. A common, intuitive workaround for this challenge is to determine an individual decision rule for each predicted class score. However, this strategy implies that multiple decisions are made for the same input, thus fundamentally changing the task to multi-label classification (in this framework, multi-label classification is handled as separate binary tasks, as detailed in Suppl. Note~\ref{app:general-instructions} - inclusion criteria).
Instead, in practice, a multi-class system requires a single global decision rule for all classes, which amounts to identifying optimal global 'decision regions'. The most common global decision rule is to simply select the class associated with the highest predicted class score, which is typically implemented as an 'argmax' operation and is also referred to as a 'maximum a posteriori' decision. Bayesian decision theory, however, shows that this argmax rule is only the optimal choice in case of equal severity of class confusions (FP2.5.2=False) and no compensation for class imbalances being requested (FP2.5.5=False). If one of these requirements is not fulfilled, a cost-dependent variation of the argmax-rule should be employed (see equation 44 in~\cite{ferrer2022analysis}). Further, the argmax decision rule assumes that predicted class scores are calibrated (see Section~\ref{app:calibration} for details on calibration). Fig.~\ref{fig:top-label-pitfall} showcases a hypothetical example of how argmax can be a suboptimal decision rule in combination with miscalibrated model outputs. While a variety of calibration metrics is discussed in Section~\ref{app:calibration}, it should be noted that \ac{EC} features a framework to directly validate the effect of a decision rule on the quality of associated decisions. Moreover, any measured negative effect can be associated with the miscalibration of scores, thus guiding users to enhance their decision making. 

A further potential pitfall associated with the global decision rule of a classifier can occur when the validation of a multi-class problem is primarily based on multi-threshold metrics. This is because multi-threshold metrics, which do not rely on a decision rule, may conceal the fact that in practice, the optimal global decision rule will not be identified. Thus, the resulting metric scores may overestimate the decision-making performance of a model in practice. 

Finally, an important consideration for identifying a decision rule of a classifier is that any data-based optimization or search must be performed on a separate data split different from the validation data. This consideration includes any configuration of re-calibration methods.

\newpage

\paragraph{Notation} The notation for our recommendations has been based on \ac{BPMN}\footnote{\url{https://www.omg.org/spec/BPMN/}}. The individual components used in the recommendation diagrams are explained in Fig.~\ref{fig:process-diagram-symbols}. Please note that we do not strictly follow \ac{BPMN} to improve clarity of presentation.

%\newpage
\paragraph{Terminology} Terminology may differ substantially across communities. For example, the statistics community prefers the term \ac{PPV} over \textit{Precision}, as the latter can be confused with the mathematical precision (repeatability) term. In the medical domain, the term \textit{validation} is used for an independent assessment (untouched test set) of an algorithm, while the machine learning community commonly uses a \textit{validation set} for hyperparameter tuning. To avoid confusion resulting from unclear terminology, we follow the general terminology of \citep{reinke2021common} and have included a glossary in the Suppl. Note~\ref{app:terminology}. 

% \newpage
% \begin{figure}[H]
%     \centering
%     \includegraphics[width=0.8\linewidth]{pitfall-paper/images/Classification/AUROC_classes_threshold.png}
%     \caption{Reporting the \acf{AUROC} for multiple classes may convey overly optimistic classifier performance. This is because any classifier upon application requires a global decision rule, which can not be optimized w.r.t to each class individually (see paragraph \textit{Global decision rule}). Thus, in practice, \ac{ROC} curves for most classes will be subject to a suboptimal cutoff. In the depicted example, all class-specific \ac{AUROC} scores would be 1.0. In contrast, the (realistic) application of a global cutoff does not lead to good performance for the classes it has not been optimized on. Used abbreviations: \acf{PPV}, \acf{NPV}, \acf{MCC}, Cohen's Kappa $\kappa$ and \acf{BA}.}
%     \label{fig:auroc-classes-threshold}
% \end{figure}

\newpage
%------------------------------------
\subsection{Problem Category Mapping}
\label{app:fingerprint-category}
\begin{figure}[H]
    \centering
    \includegraphics[width=1\linewidth]{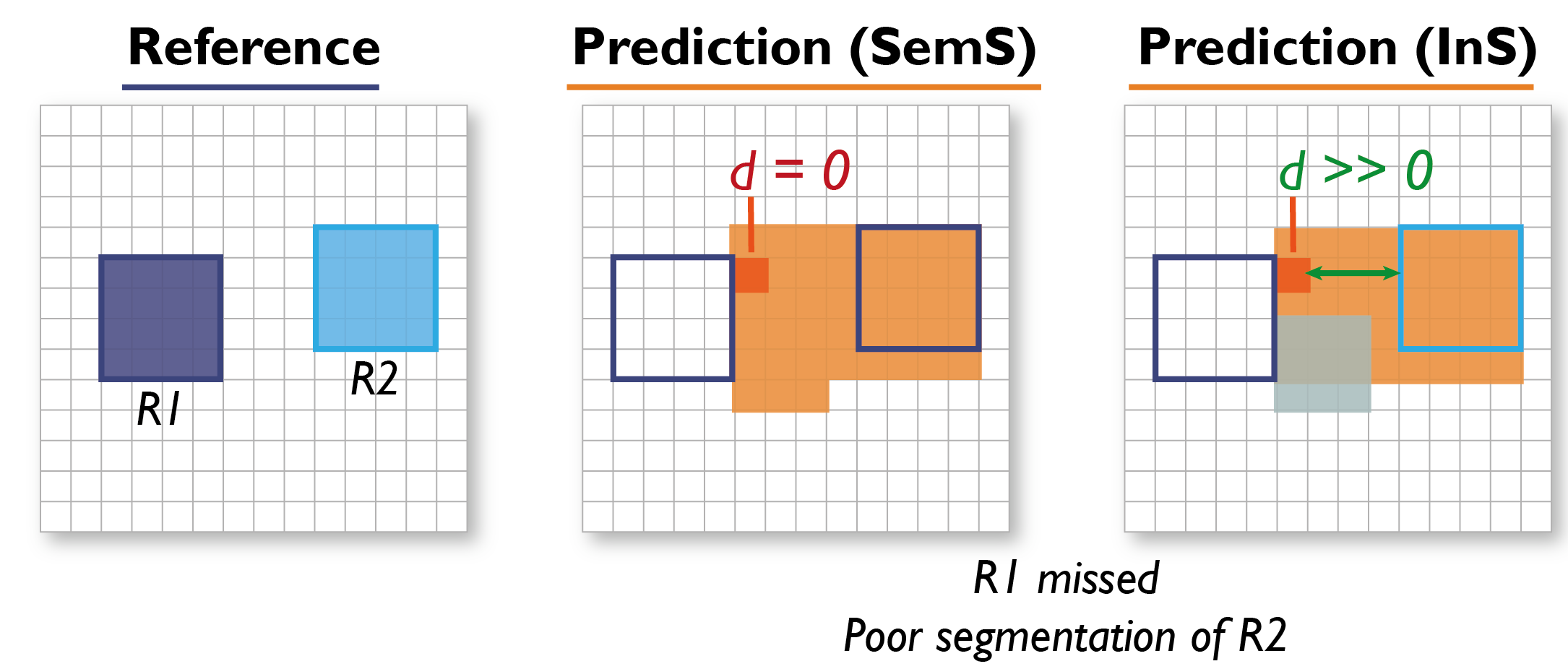}
    \caption{\textbf{Boundary-based metrics in semantic/instance segmentation problems.} If multiple structures of the same type can be seen within the same image (here: reference objects \textit{R1} and \textit{R2}), it is generally advisable to phrase the problem as  instance segmentation (InS; right) rather than semantic segmentation (SemS; left). This way, issues with boundary-based metrics resulting from comparing a given structure boundary to the boundary of the wrong instance in the reference can be avoided. In the provided example, the distance of the red boundary pixel to the reference, as measured by a boundary-based metric in SemS problems, would be zero, because different instances of the same structure cannot be distinguished. This problem is overcome by phrasing the problem as InS. In this case, (only) the boundary of the matched instance (here: R2) is considered for distance computation.}
     \label{fig:boundary-is-ss}
\end{figure}

%------------------------------------
\acresetall
The problem fingerprinting (Step 1 in Fig.~\ref{fig:meta-mapping}; see Sec.~\ref{app:fingerprints}) begins with the step of mapping a given problem with all its intrinsic and data set-related properties to the corresponding problem category via the \textit{category mapping} shown in Fig.~\ref{fig:category_mapping}. This step is crucial for avoiding pitfalls related to the inappropriate choice of the problem category, as detailed in the sister publication of this work \cite{reinketizabi2022metrics}. Specifically, when multiple instances of the same structure type can occur in an image, it is typically advisable to phrase the underlying problem as an object detection or instance segmentation problem rather than a semantic segmentation problem (Figs.1 and ~\ref{fig:boundary-is-ss}).

\newpage
%------------------------------------
%\subsection{Problem Fingerprinting}
%\label{app:fingerprint-category}
%------------------------------------
%------------------------------------
\subsection{Generation of the problem fingerprint}
\label{app:fingerprints}
%------------------------------------
\acresetall
\textit{Metrics Reloaded} is based on the novel concept of \textit{problem fingerprinting} -- the generation of a structured representation of the given problem that captures all aspects that are relevant for metric selection. 
Specifically, the fingerprint comprises a set of items, each of which represents a specific property of the problem, is either binary or categorical, and must be instantiated by the user. In the following, we will refer to all fingerprint items with the notation \textit{FPX.Y}, where Y is a numerical identifier and the index \textit{X} represents one of the following families: general, domain interest-related, target structure-related, data set-related, algorithm output-related.

Fingerprint generation begins with the aforementioned mapping of the underlying problem with its intrinsic and data set-related properties to the corresponding problem category via the \textit{category mapping} (Subprocess S1) shown in Extended Data \ref{fig:category_mapping}. Next, the user needs to instantiate the category-specific fingerprint items provided in Figs.~\ref{fig:fingerprint-ilc}-\ref{fig:fingerprint-ilc-3} (image-level classification), Figs.~\ref{fig:fingerprint-ss-1}/\ref{fig:fingerprint-ss-2} (semantic segmentation), Figs.~\ref{fig:fingerprint-od-1}-\ref{fig:fingerprint-od-3} (object detection), and Figs.~\ref{fig:fingerprint-is-1}-\ref{fig:fingerprint-is-3} (instance segmentation). 

 Instantiating fingerprint items may not always be straightforward due to their binary/categorical nature. Therefore, the \textit{Metrics Reloaded} tool comprises a "Why are we asking this question?" button in each branch based on a fingerprint that may not be straightforward to instantiate. In case of ongoing doubt, the user may traverse all appropriate branches originating from the questions. 

\begin{figure}[H]
    \centering
    \includegraphics[width=\textwidth]{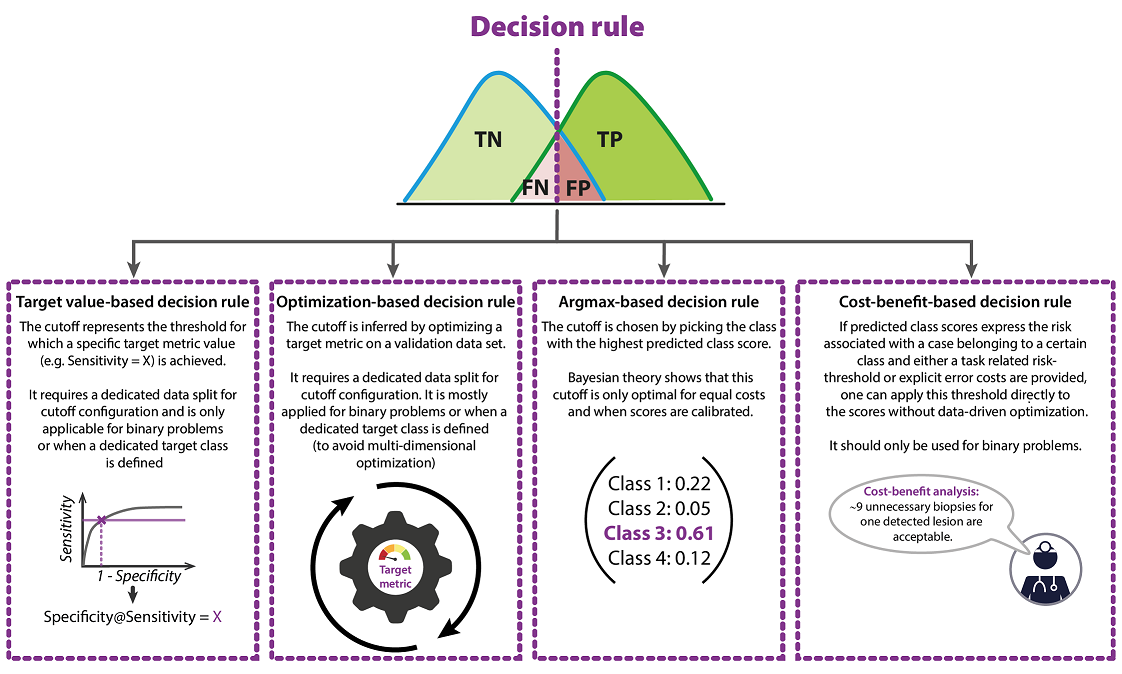}
    \caption{Illustration of strategies for identifying a decision rule applied to predicted class scores.}
    \label{fig:cutoff-strategies}
\end{figure}

Importantly, some fingerprint items require particularly careful consideration and/or are not sufficiently self-explanatory. These are the following: 

\paragraph{FP2.6: Decision rule applied to predicted class scores} 
Modern algorithms output (continuous) predicted class scores. To classify cases in an actual biomedical application (i.e., to make actual decisions), however, applying a decision rule to the scores is required; this amounts to setting a cutoff value in the binary classification case. The deciding factor for whether or not to apply a decision rule during validation should be how much focus is to be put on the quality of the actual decisions of a classification system versus the general quality of its continuous predictions. While some communities have converged to decision rule-based validation (e.g., cell instance segmentation \cite{caicedo2019nucleus}), recent clinical initiatives advocate for decision rule-agnostic validation, arguing that decision rules are often over-optimized on a specific data set, associated results are not transferable across study cohorts (e.g., with differing disease prevalence) and clinical applications (e.g., with differing cost-benefit trade-offs for patients), and continuous "risk scores" might be beneficial for communicating results with patients \cite{vickers2016net,moons2015transparent,bejnordi2017diagnostic}. One study goes so far as to call out the common practice of imposing decision rules on continuous predictions as  ‘dichotomania’~\cite{wynants2019three}. We handle this controversy in current practices by making validation with specific decision rules applied optional (for all tasks except semantic segmentation) and encoding user preferences in this fingerprint. The fingerprint offers the following decision rule strategies (Fig.~\ref{fig:cutoff-strategies}):

\begin{description}
\item [Target-value based (for binary image-level classification problems)] Sometimes, the underlying problem provides a specific target metric value to be reached (e.g., Sensitivity of 0.95), requiring a corresponding cutoff value. In this case, we use the notation Metric@(TargetMetric = TargetValue), for example, Specificity@(Sensitivity = 0.95), denoting the Specificity for a Sensitivity matching the target value (here: 0.95). Importantly, this cutoff needs to be configured on a separate and dedicated data split.

\item [Optimization-based] If no specific target value is provided, a data-based decision rule can also be identified by optimizing a primary metric (e.g., F$_1$ Score) using a dedicated data set for decision rule configuration. Notably, simple (one-dimensional) cutoff scans are only possible in binary tasks, while identifying decision rules in multiple classes represents a computationally and technically complex process.

\item [Argmax-based] An alternative widely used strategy is to simply apply a decision rule based on the 'argmax' operation, which boils down to a cutoff of 0.5 in binary classification problems. The underlying hypothesis for this strategy is that the highest class score resembles the highest probability for the associated class being the true class. In Bayesian theory, this decision rule defines a Bayes classifier, and the theory further shows that the underlying hypothesis only holds for equal severity of class confusions (FP2.5.2=False) and when the class scores are calibrated. Detailed considerations for this decision rule strategy are provided in Sec.~\ref{app:general-instructions}.

\item [Cost-benefit-based] If the predicted class scores are calibrated (see FP2.7), and task-related error costs or a risk cutoff (the latter only for binary classification tasks, e.g., "only treat patients with cancer risk >10\%") are provided, one can apply this decision rule directly to the scores without data-driven optimization. Notably, in binary classification tasks, cost-benefit-based cutoffs often correspond to a cost ratio of \ac{TP} versus \ac{FP} (e.g., not more than 10\ac{FP} per 1 \ac{TP} should be treated), while for cost-based cutoffs the explicit costs for both errors \ac{FP} and \ac{FN} are defined (see DG3.2, Suppl. Note~\ref{ssec:dg3}). Cost-based decision rules are further extendable to multi-class problems~\cite{ferrer2022analysis}. %However, in the context of this fingerprint, per-class counting metrics are selected, and thus multi-class problems are validated in a one-versus-rest scheme, where cost-benefit analysis is not intuitively applicable.

\item[No decision rule applied] A complementary strategy is to abstain from validating algorithms under a certain decision rule and exclusively report results on multi-threshold metrics (averaging over various cutoffs) instead. 
\end{description}

%\paragraph{FP2.7: Interpretability of predicted class scores requested} When validating classification methods, it is often crucial for the predicted class scores themselves to be interpretable. This holds especially true for problems that involve direct human read-out. Interpretability is commonly achieved by requesting an algorithm to be well-calibrated. This is the case if $p$ percent of all predictions reported at probability $p$ are true \cite{seidenfeld1985calibration}. In a clinical setting, for example, this would imply that out of all patients that are assigned a score of 0.9, 90\% should actually belong to the target class (e.g. have a certain disease). In practice, many methods are overconfident, meaning that less than $p$ percent of all predictions reported at probability $p$ are true. Calibration is commonly measured by the \ac{ECE}. An alternative to achieve interpretability are \ac{PSR} \cite{gneiting2007strictly}, such as the \ac{BS} and the \ac{Log Score}, which validate discrimination and calibration in a single score.  %, which uses the entire data set .
% is typically measured either w.r.t the entire data set (calibration), or via matching of posterior probabilities on the individual level (Proper Scoring Rules). In application scenarios that involve interpreting the raw algorithm output (specifically the  predicted class scores), a\textit{calibrated output} may be highly desirable.

\newpage

\begin{figure}[H]
    \centering
    \includegraphics[width=0.7\linewidth]{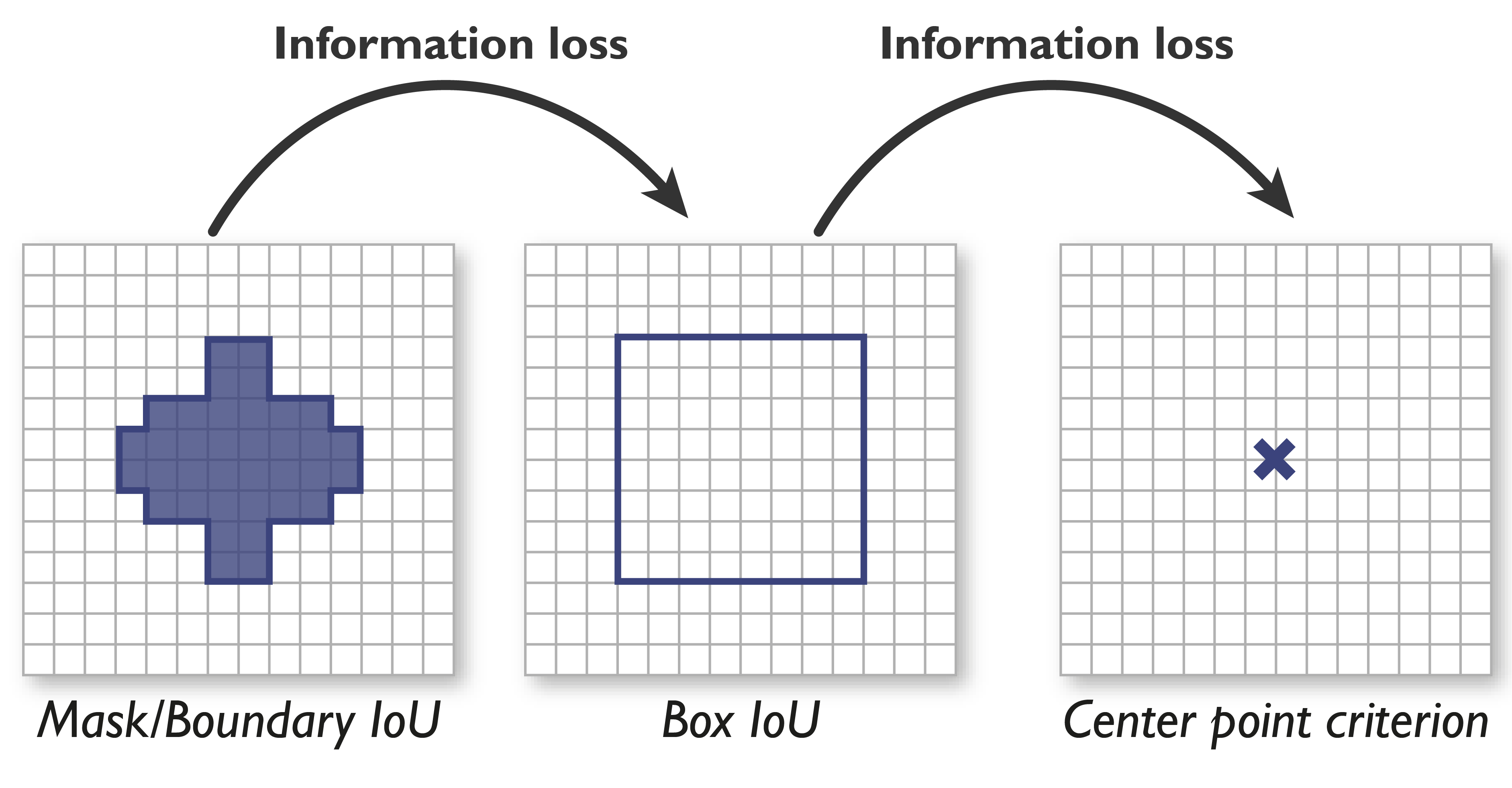}
    \caption{Selection of a localization criterion that discards spatial information should be well motivated by the given task.}
    \label{fig:od-information-loss}
\end{figure}

\begin{figure}[H]
    \centering
    \includegraphics[width=1\linewidth]{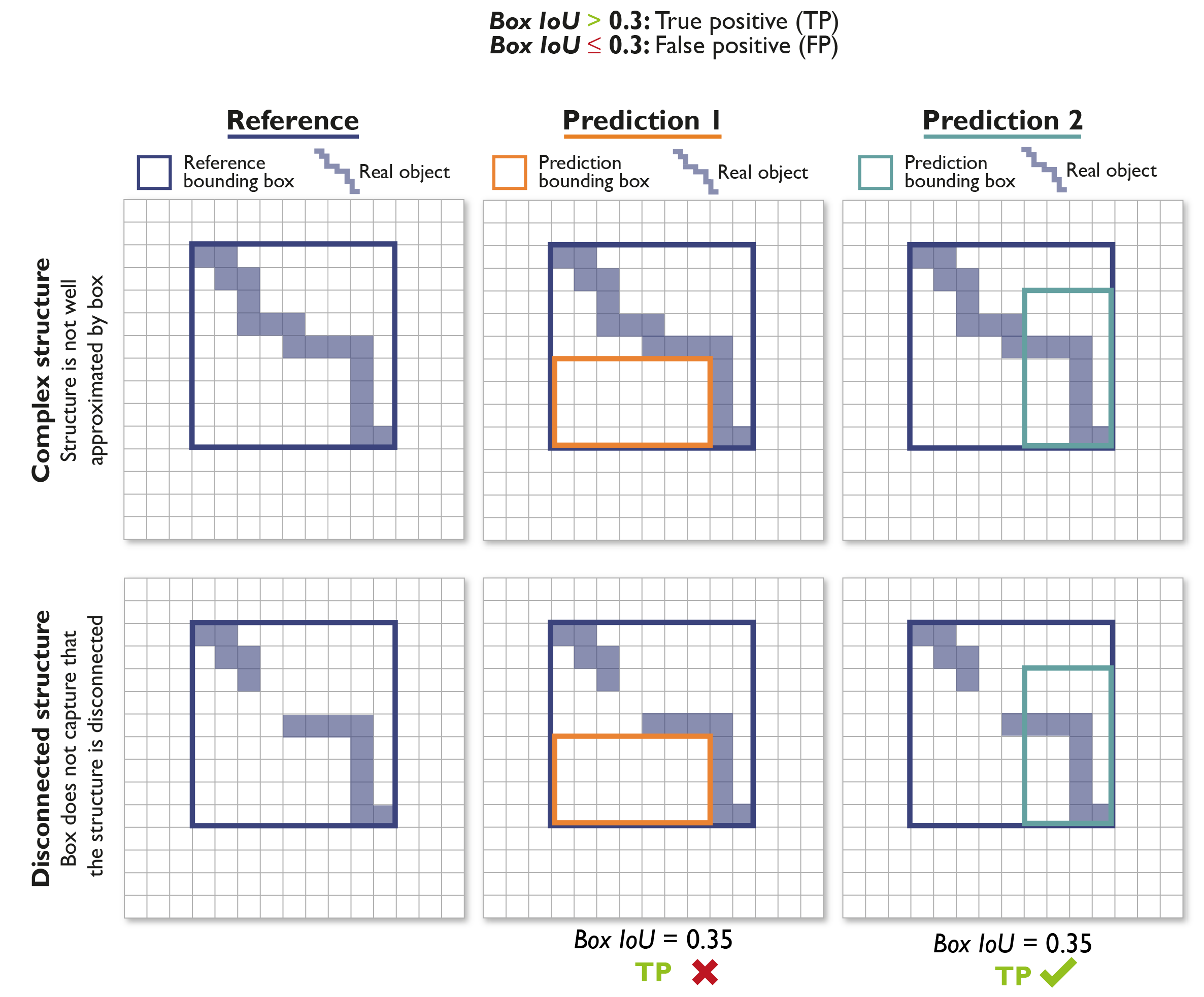}
    \caption{Bounding boxes are not well-suited for representing complex (top) and disconnected (bottom) shapes. Specifically, they are not well-suited for capturing multi-component structures. \textit{Predictions 1} and \textit{2}  both end up in a \acf{TP} detection, as the Box \acf{IoU} is larger than the cutoff 0.3. However, \textit{Prediction 1} does not hit the real objects at all.}
    \label{fig:disconnected}
\end{figure}

\paragraph{FP2.4 Desired granularity of localization}
Selecting a localization criterion operating on a lower/ coarser resolution with regard to provided reference annotations effectively discards spatial information and should be well motivated by the given task (see Fig.~\ref{fig:od-information-loss}). For instance, Box \ac{IoU} is sometimes employed despite access to pixel-mask annotations (FP4.4) because associated models (object detectors) are considered simpler approaches compared to instance segmentation models. Such simplification may cause problems if structures are not well-approximated by a box shape -- especially for 3D shapes, boxes usually constitute poor approximations -- or if structures can overlap (FP3.5), causing multi-component masks (see Fig.~\ref{fig:disconnected}).

\newpage

\paragraph{FP2.5.5: Compensation for class imbalances} 

While Accuracy is the de facto standard metric in multi-class settings with balanced class frequencies and error costs, this metric is prone to several pitfalls when class imbalances are present. To give an example, consider the following confusion matrix for a binary classification task: $TP=0$, $FP=1$, $FN=1$, $TN=$ 10,000, which leads to an accuracy of $\approx 1$. Three pitfalls pertain to this metric score, which at the same time represent the three reasons why users may want to compensate for the underlying effects caused by the class imbalance:\\\\

\newpage
 
 \textbf{Misleading metric values due to missing reference value for naive classifier:} In the provided example, the near-perfect score hides the fact that the same performance could have been achieved by a naive system always predicting the dominant class. Generally, in balanced scenarios, the Accuracy of a naive classifier is known to be “1/number of classes”, which serves as an important anchor when interpreting the metric scores. However, when class imbalances are present, no such interpretation can be made and the naive reference depends on the class prevalences. 

\textbf{Misleading metric values due to unequal contribution of classes to the metric score:} In the provided example, the near-perfect score hides the fact that all samples of the positive class (here: one sample) were misclassified. While all classes contribute similarly to the Accuracy metric in balanced scenarios, frequent classes dominate the performance value in imbalanced settings. While 0\% (0/1) of the rare cases have been classified correctly, the metric achieves an almost perfect score due to the very good performance on the dominant class. Other prevalence-independent metrics, such as \ac{BA}, are based on the equal contribution of each class irrespective of prevalence.

\textbf{Misleading metric values due to missing consideration of predictive values:} In the provided example, the near-perfect score hides the fact that the positive predictive value of this system is 0. Generally, in balanced scenarios, high accuracy scores imply high predictive values (\ac{PPV} and \ac{NPV}), which are important indicators of the utility of a classification system in practice. This is not necessarily the case in imbalanced scenarios, as seen in the provided example, where the \ac{PPV} is 0 despite a high Accuracy. To compensate for this effect, alternative metrics such as \ac{MCC} can be considered, which explicitly assess the predictive performance of a classifier.

\begin{figure}[H]
    \centering
    \includegraphics[width=0.9\linewidth]{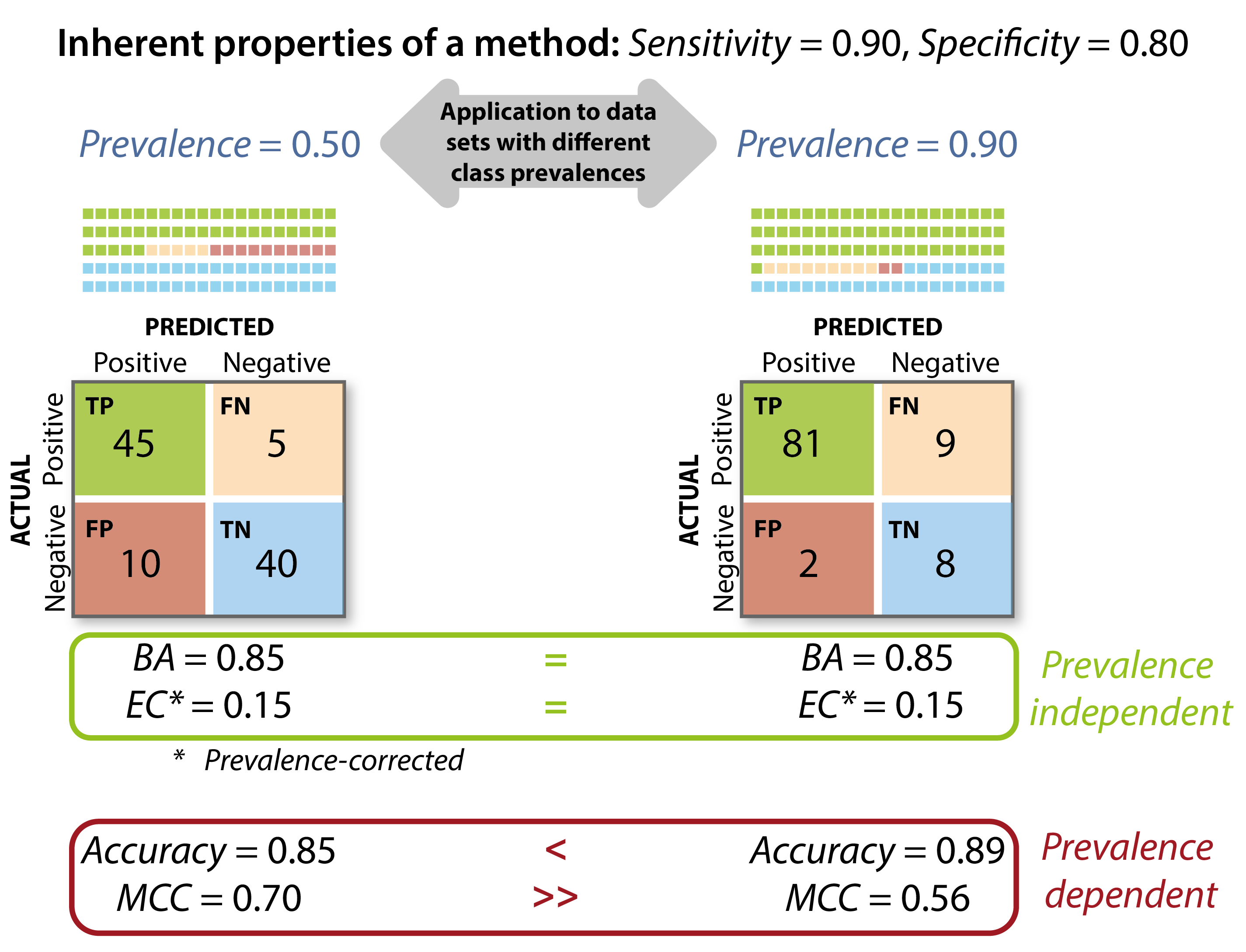}
    \caption{Effect of prevalence dependency. An algorithm with specific inherent properties (here: Sensitivity of 0.9 and Specificity of 0.8) may perform completely differently on different data sets if the prevalences differ (here: 50\% (left) and 90\% (right)) and prevalence-dependent metrics are used for validation (here: Accuracy and \acf{MCC}). In contrast, prevalence-independent metrics (here: \acf{BA} and the prevalence-corrected \acf{EC}) can be used to compare validation results across different data sets. Used abbreviations: \acf{TP}, \acf{FN}, \acf{FP} and \acf{TN}.}
    \label{fig:prevalence-dependency}
\end{figure}

\begin{figure}[H]
    \centering
    \includegraphics[width=1\linewidth]{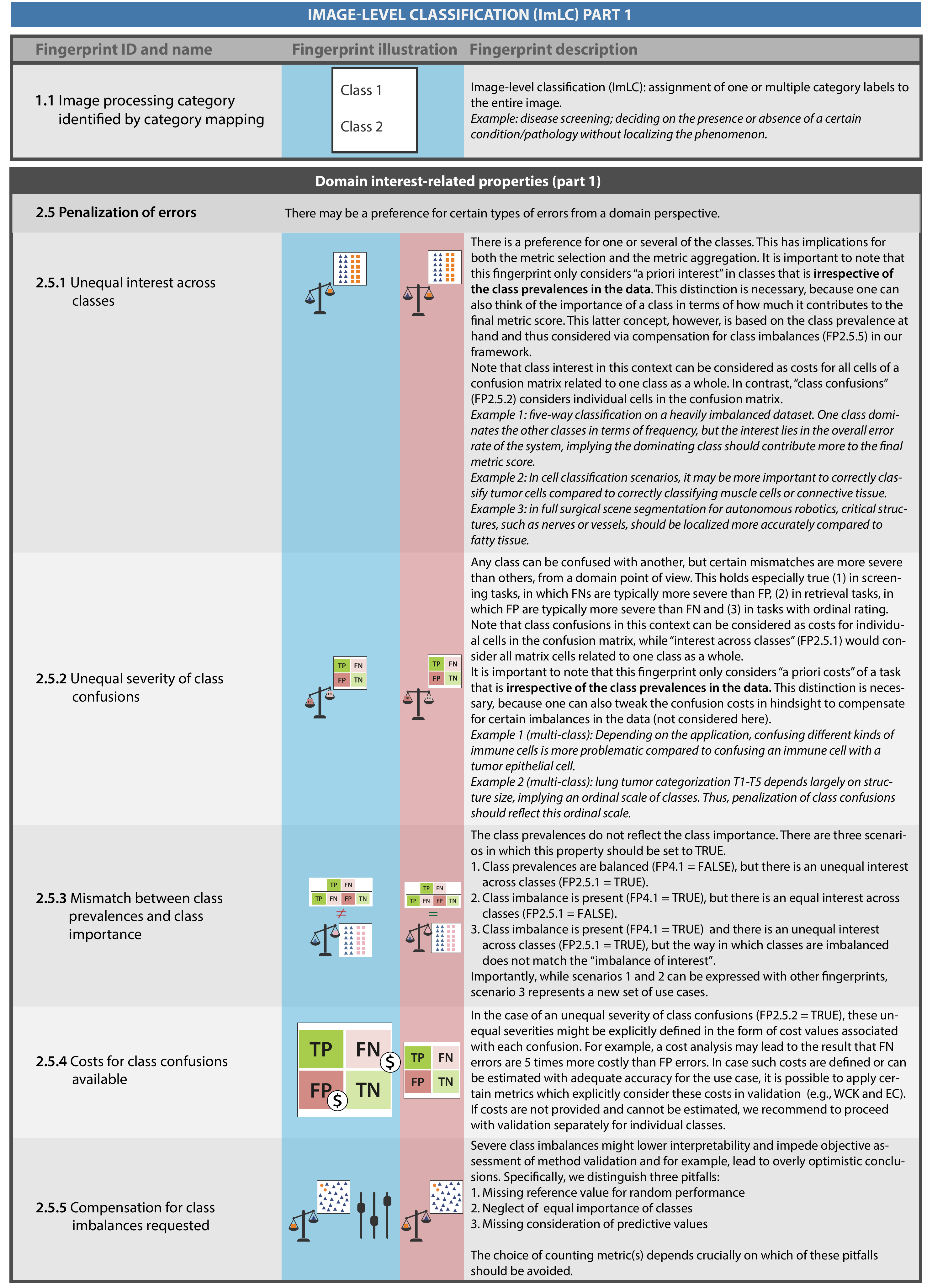}
    \caption{\textbf{Fingerprint for image-level classification (Part 1).} In the case of binary fingerprint items, the blue column shows examples for which the property is true while the red column shows counterexamples. Categorical fingerprint items are only shown in blue. Suppl. Note~\ref{app:fingerprints} provides more detailed explanations of selected fingerprint items. Used abbreviations: \acf{TP}, \acf{FP}, \acf{FN}, \acf{TN}.}
    \label{fig:fingerprint-ilc}
\end{figure}

\begin{figure}[H]
    \centering
    \includegraphics[width=1\linewidth]{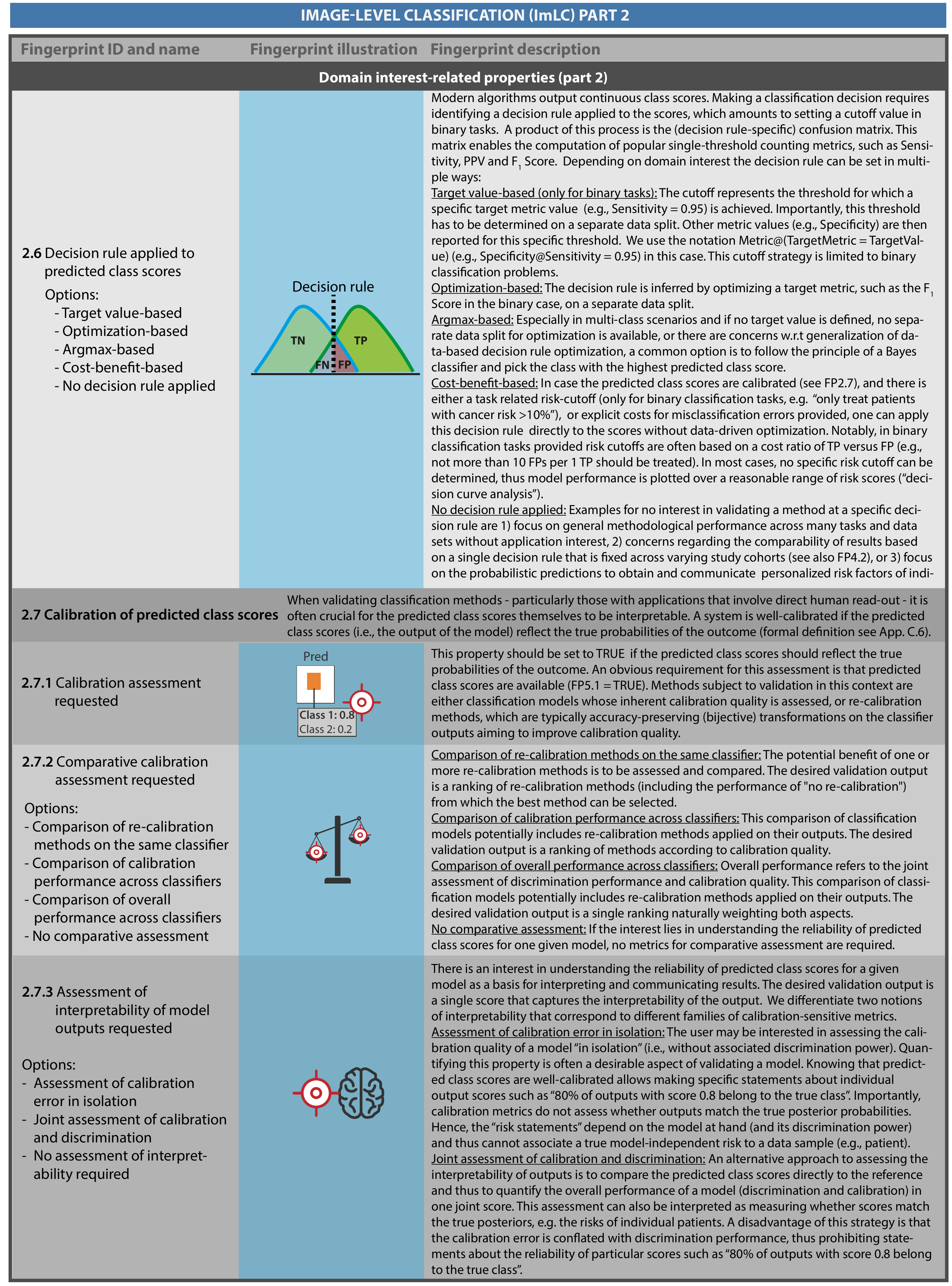}
    \caption{\textbf{Fingerprint for image-level classification (Part 2).} In the case of binary fingerprint items, the blue column shows examples for which the property is true while the red column shows counterexamples. Categorical fingerprint items are only shown in blue. Suppl. Note~\ref{app:fingerprints} provides more detailed explanations of selected fingerprint items. Used abbreviations: \acf{TP}, \acf{FP}, \acf{FN}, \acf{TN}.}
    \label{fig:fingerprint-ilc-2}
\end{figure}

\begin{figure}[H]
    \centering
    \includegraphics[width=1\linewidth]{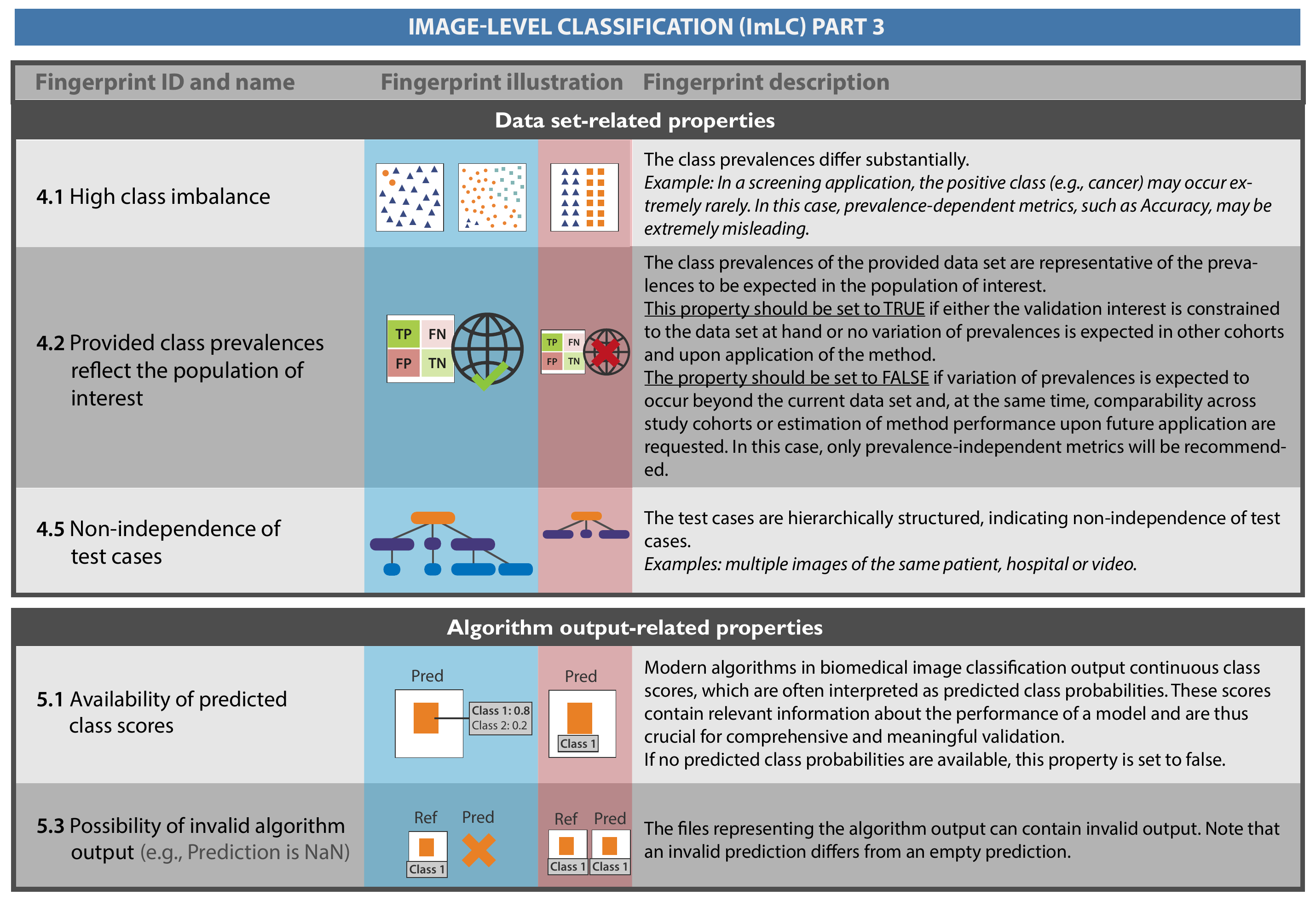}
    \caption{\textbf{Fingerprint for image-level classification (Part 3).} In the case of binary fingerprint items, the blue column shows examples for which the property is true while the red column shows counterexamples. Categorical fingerprint items are only shown in blue. Suppl. Note~\ref{app:fingerprints} provides more detailed explanations of selected fingerprint items. Used abbreviations: Reference (Ref), Prediction (Pred), \acf{TP}, \acf{FP}, \acf{FN}, \acf{TN}.}
    \label{fig:fingerprint-ilc-3}
\end{figure}

\begin{figure}[H]
    \centering
    \includegraphics[width=1\linewidth]{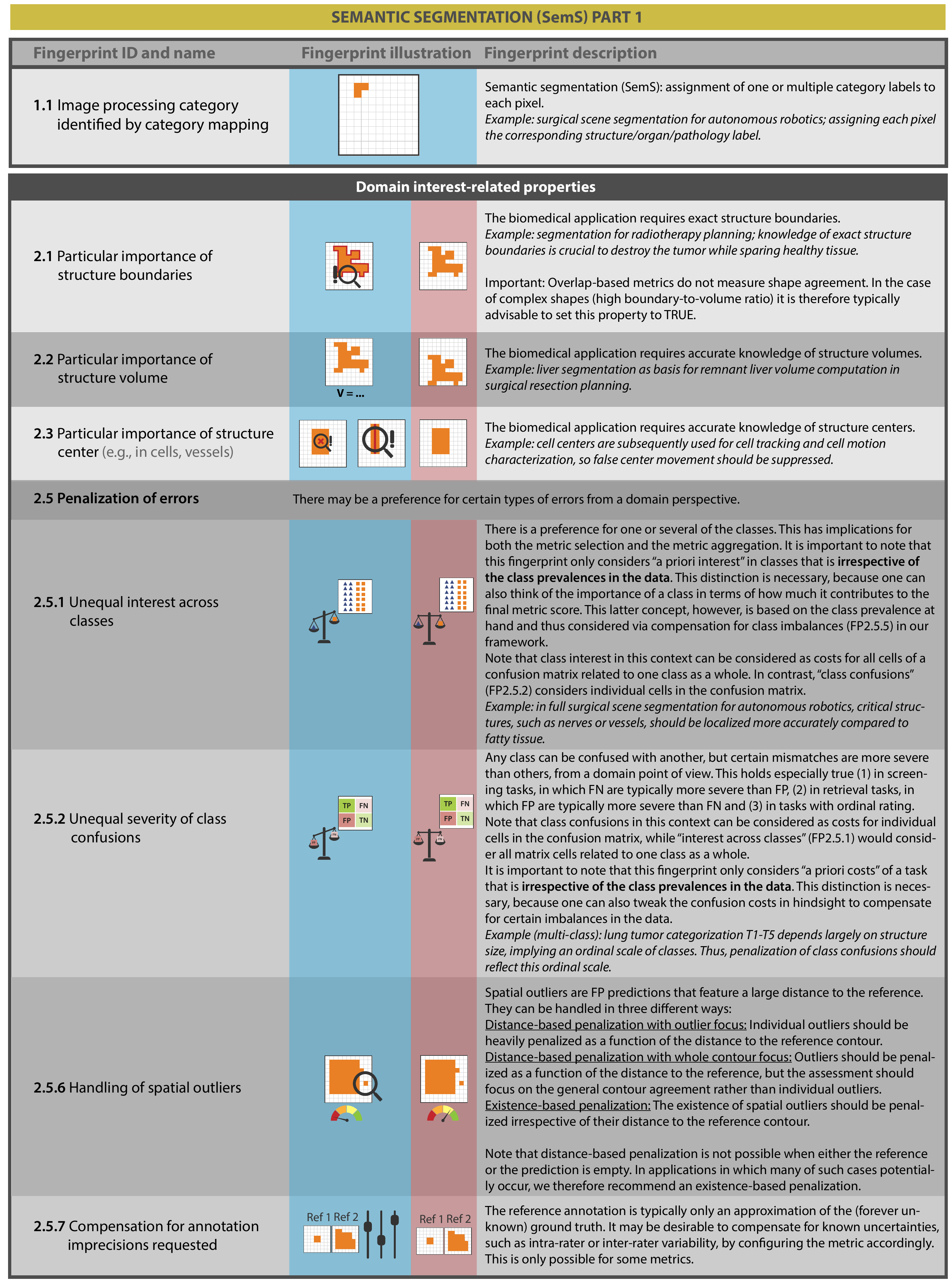}
    \caption{\textbf{Fingerprint for semantic segmentation (Part 1).} In the case of binary fingerprint items, the blue column shows examples for which the property is true while the red column shows counterexamples. Categorical fingerprint items are only shown in blue. Suppl. Note~\ref{app:fingerprints} provides more detailed explanations of selected fingerprint items. Used abbreviations: Reference (Ref), Prediction (Pred), \acf{TP}, \acf{FP}, \acf{FN}, \acf{TN}.}
    \label{fig:fingerprint-ss-1}
\end{figure}

\begin{figure}[H]
    \centering
    \includegraphics[width=1\linewidth]{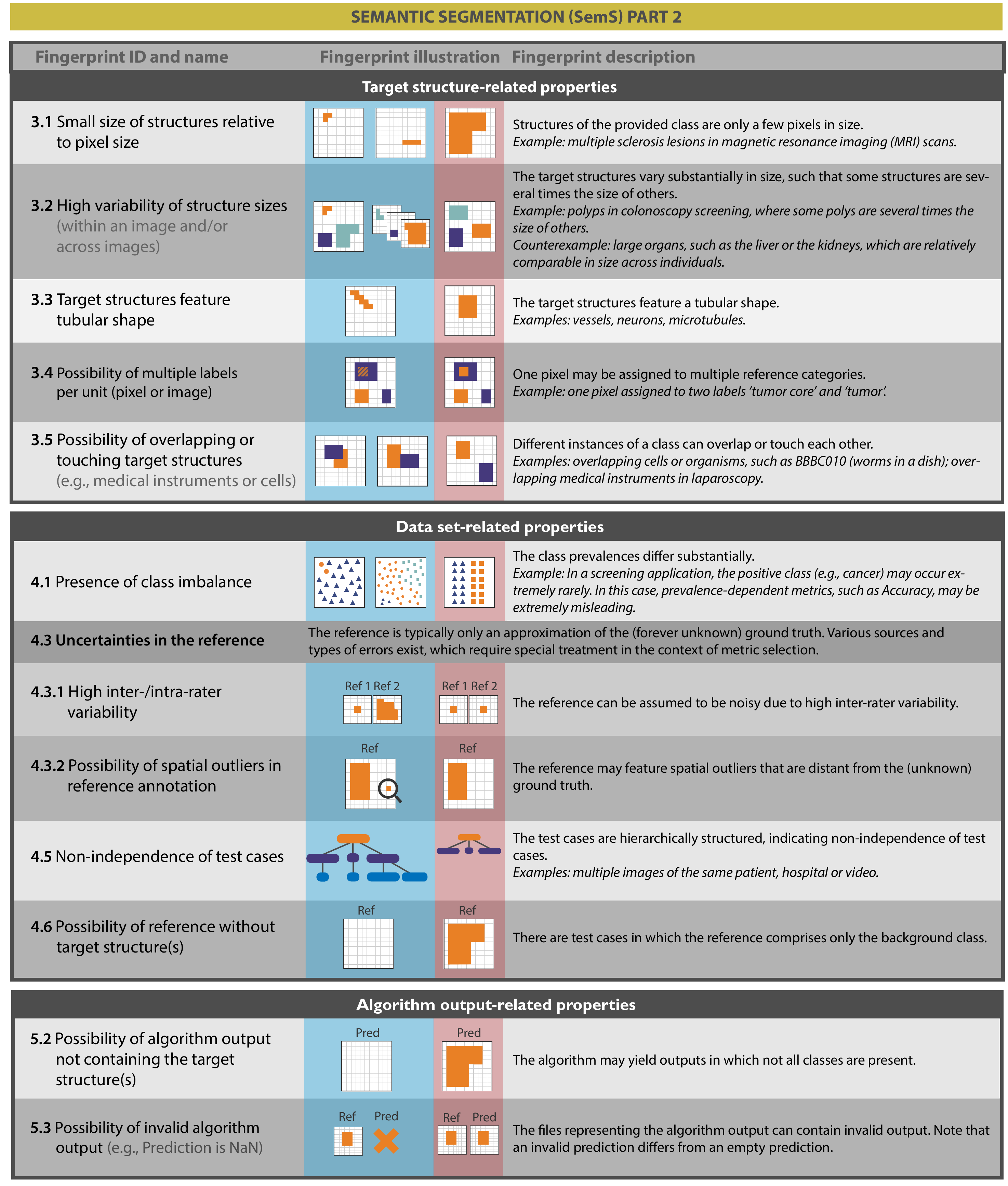}
    \caption{\textbf{Fingerprint for semantic segmentation (Part 2).} In the case of binary fingerprint items, the blue column shows examples for which the property is true while the red column shows counterexamples. Categorical fingerprint items are only shown in blue. Suppl. Note~\ref{app:fingerprints} provides more detailed explanations of selected fingerprint items.}
    \label{fig:fingerprint-ss-2}
\end{figure}

\begin{figure}[H]
    \centering
    \includegraphics[width=0.9\linewidth]{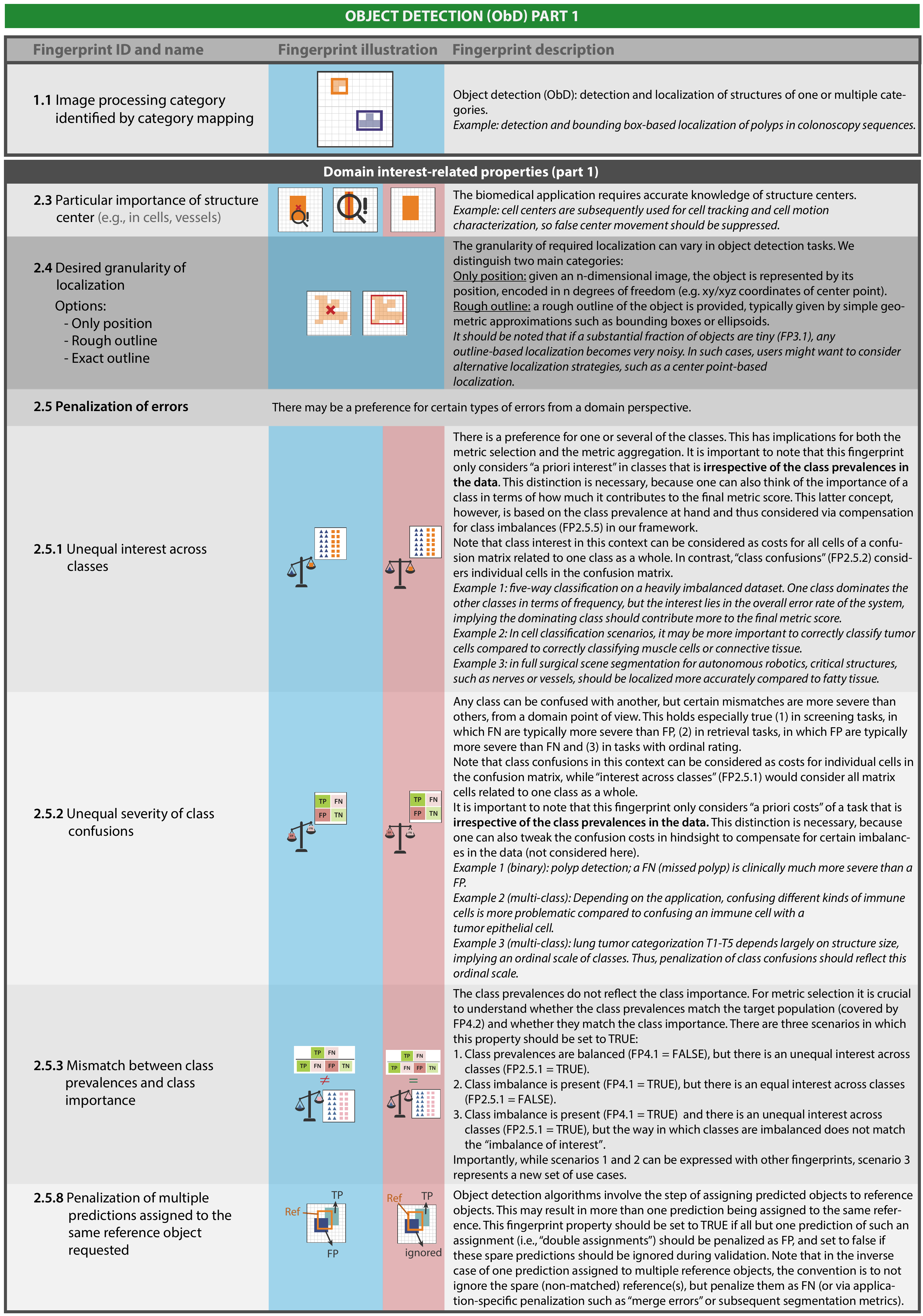}
    \caption{\textbf{Fingerprint for object detection (Part 1).} In the case of binary fingerprint items, the blue column shows examples for which the property is true while the red column shows counterexamples. Categorical fingerprint items are only shown in blue. Suppl. Note~\ref{app:fingerprints} provides more detailed explanations of selected fingerprint items. Used abbreviations: \acf{TP}, \acf{FP}, \acf{FN}, \acf{TN}.}
    \label{fig:fingerprint-od-1}
\end{figure}

\begin{figure}[H]
    \centering
    \includegraphics[width=1\linewidth]{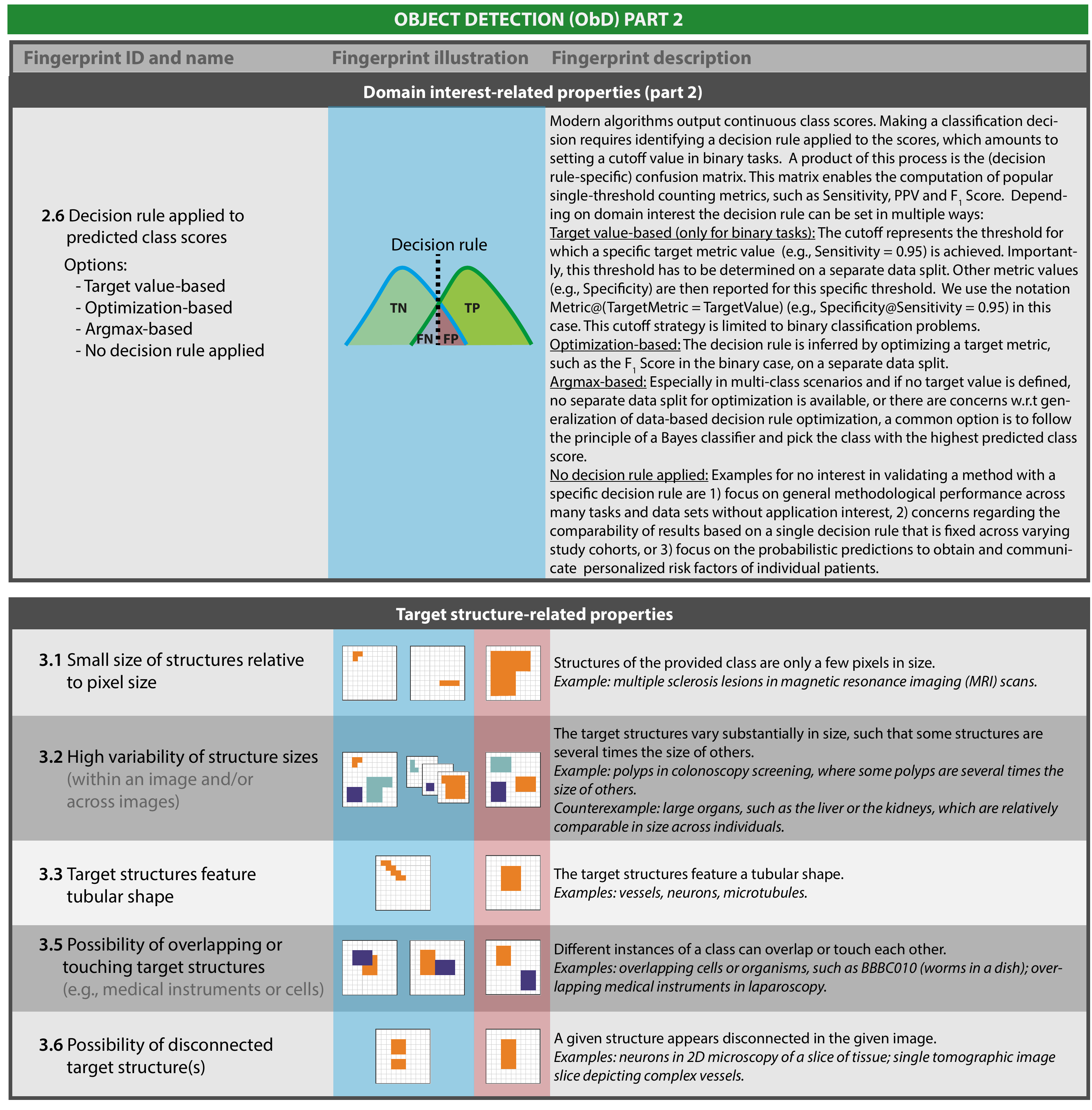}
    \caption{\textbf{Fingerprint for object detection (Part 2).} In the case of binary fingerprint items, the blue column shows examples for which the property is true while the red column shows counterexamples. Categorical fingerprint items are only shown in blue. Suppl. Note~\ref{app:fingerprints} provides more detailed explanations of selected fingerprint items. Used abbreviations: \acf{TP}, \acf{FP}, \acf{FN}, \acf{TN}.}
    \label{fig:fingerprint-od-2}
\end{figure}

\begin{figure}[H]
    \centering
    \includegraphics[width=1\linewidth]{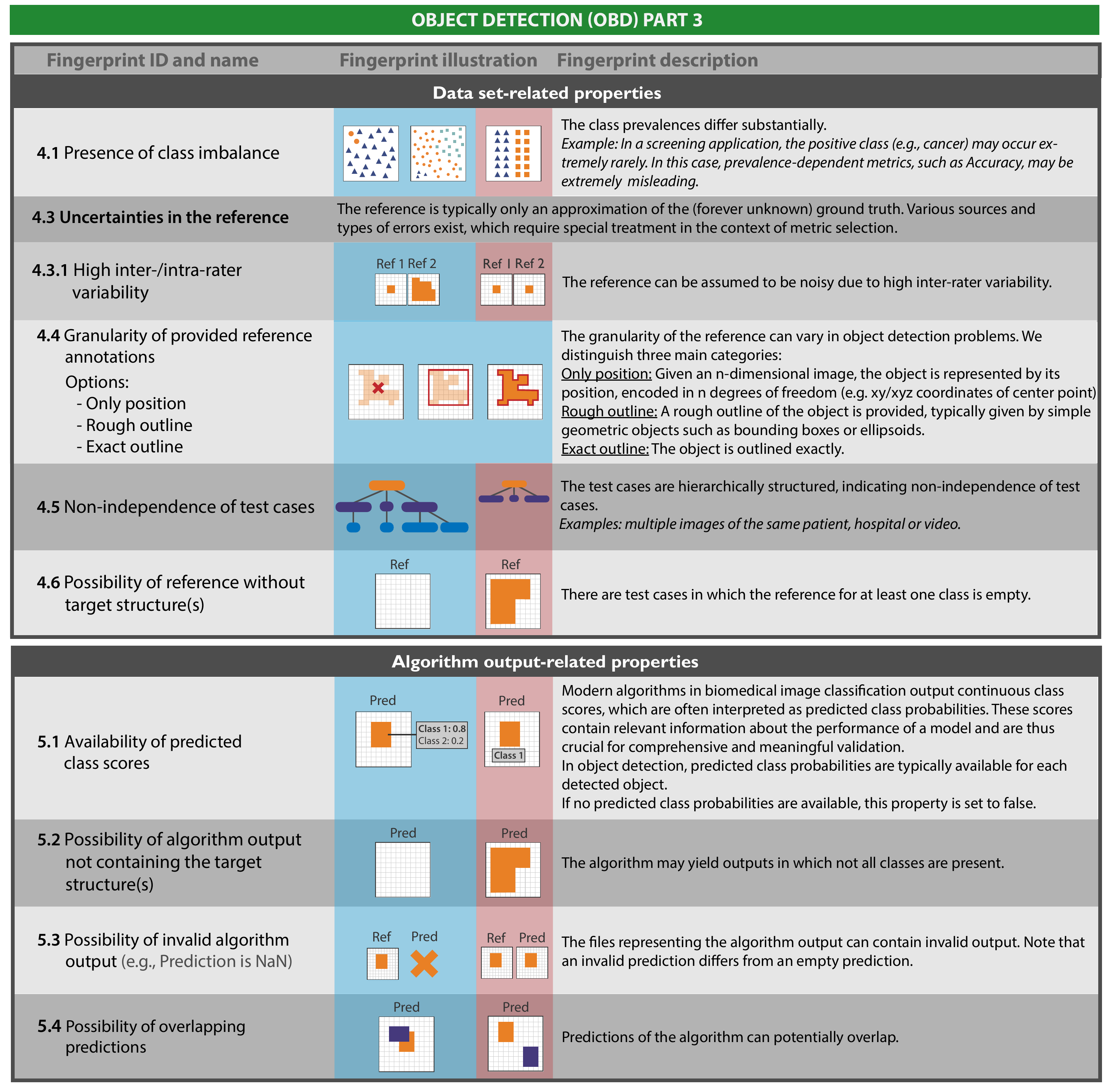}
    \caption{\textbf{Fingerprint for object detection (Part 3).} In the case of binary fingerprint items, the blue column shows examples for which the property is true while the red column shows counterexamples. Categorical fingerprint items are only shown in blue. Suppl. Note~\ref{app:fingerprints} provides more detailed explanations of selected fingerprint items. Used abbreviations: Reference (Ref), Prediction (Pred).}
    \label{fig:fingerprint-od-3}
\end{figure}

\begin{figure}[H]
    \centering
    \includegraphics[width=0.9\linewidth]{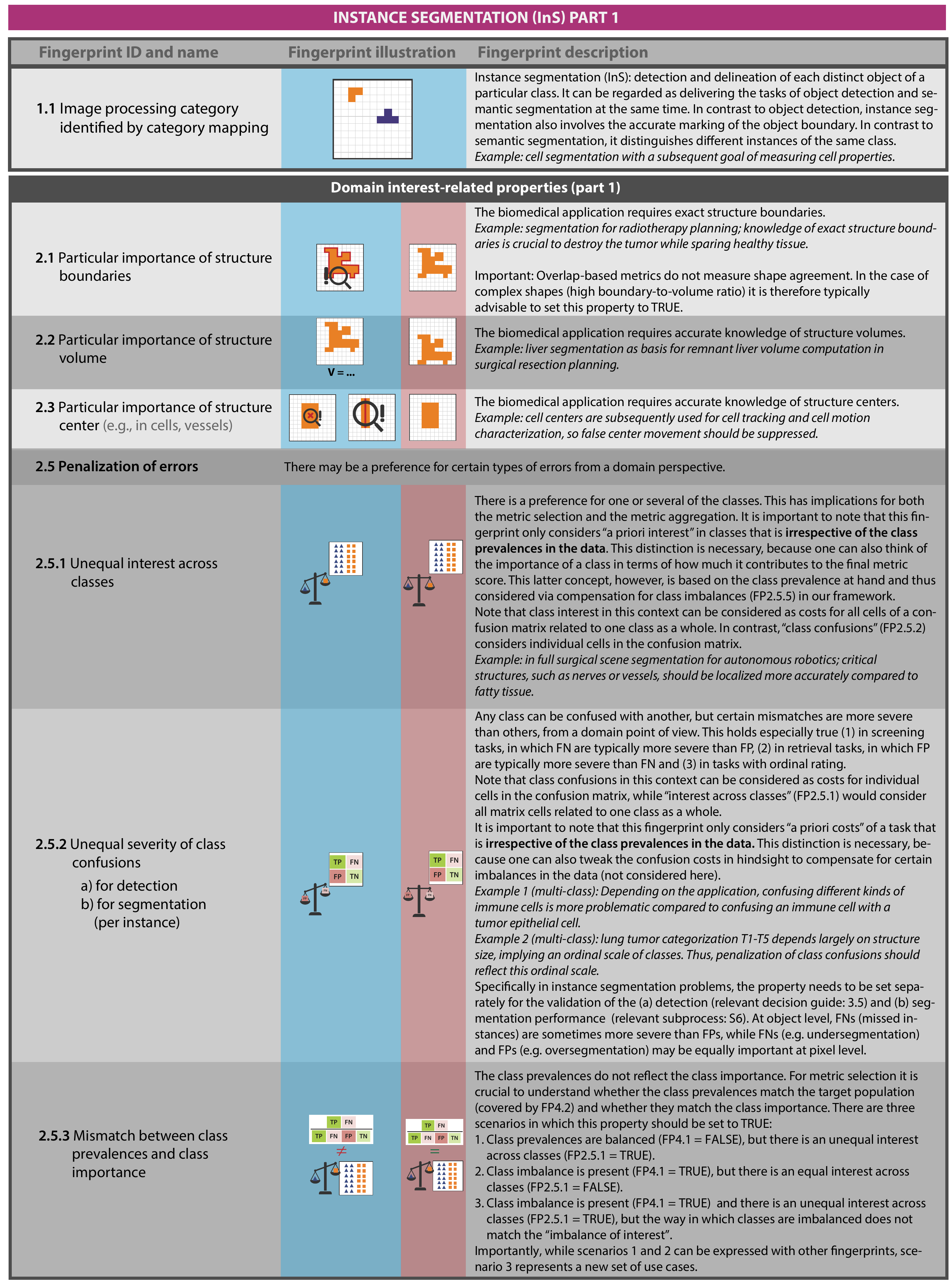}
    \caption{\textbf{Fingerprint for instance segmentation (Part 1).} In the case of binary fingerprint items, the blue column shows examples for which the property is true while the red column shows counterexamples. Categorical fingerprint items are only shown in blue. Suppl. Note~\ref{app:fingerprints} provides more detailed explanations of selected fingerprint items. Used abbreviations: \acf{TP}, \acf{FP}, \acf{FN}, \acf{TN}.}
    \label{fig:fingerprint-is-1}
\end{figure}

\begin{figure}[H]
    \centering
    \includegraphics[width=1\linewidth]{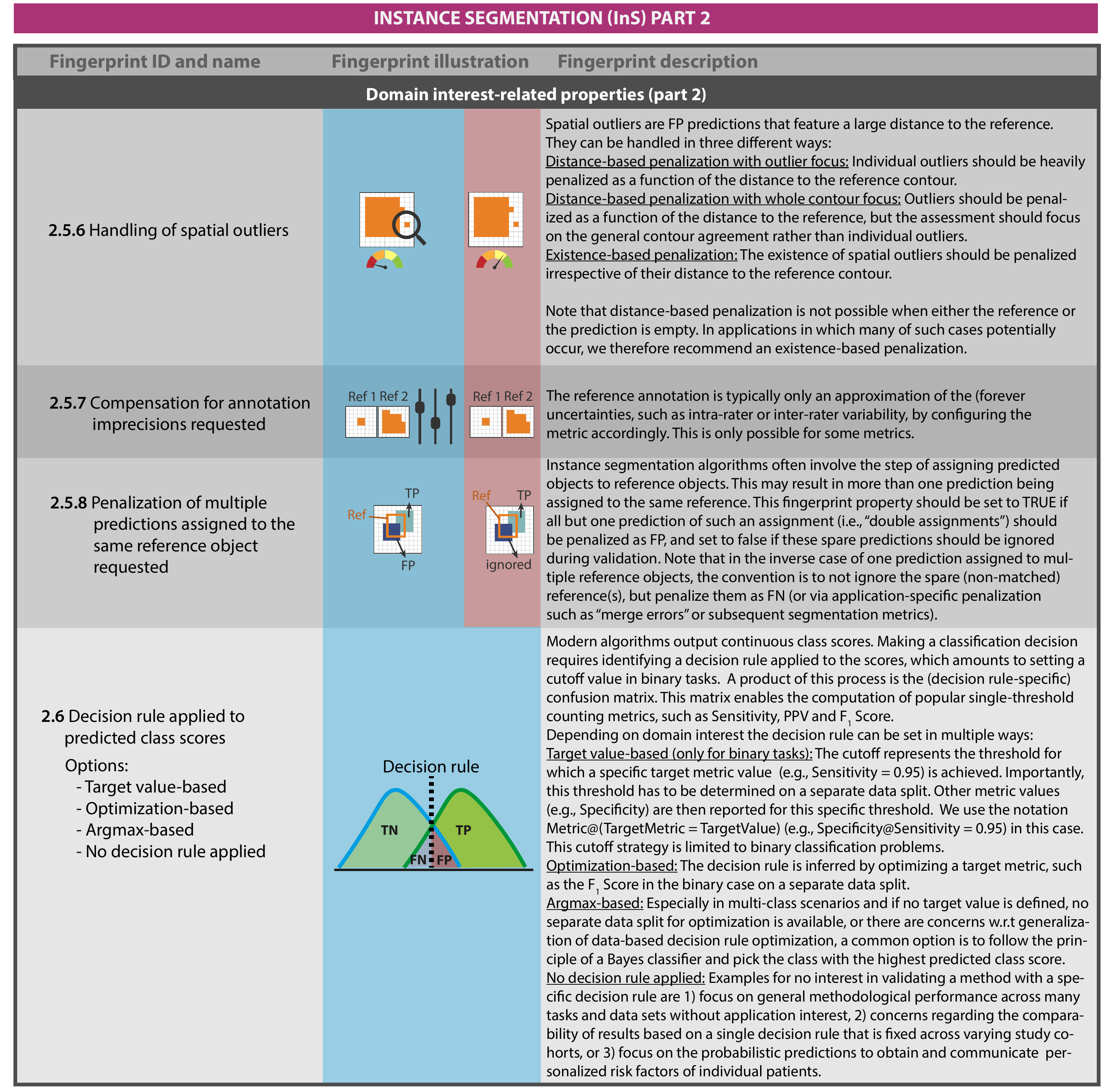}
    \caption{\textbf{Fingerprint for instance segmentation (Part 2).} In the case of binary fingerprint items, the blue column shows examples for which the property is true while the red column shows counterexamples. Categorical fingerprint items are only shown in blue. Suppl. Note~\ref{app:fingerprints} provides more detailed explanations of selected fingerprint items. Used abbreviations: Reference (Ref), \acf{TP}, \acf{FP}, \acf{FN}, \acf{TN}.}
    \label{fig:fingerprint-is-2}
\end{figure}

\begin{figure}[H]
    \centering
    \includegraphics[width=1\linewidth]{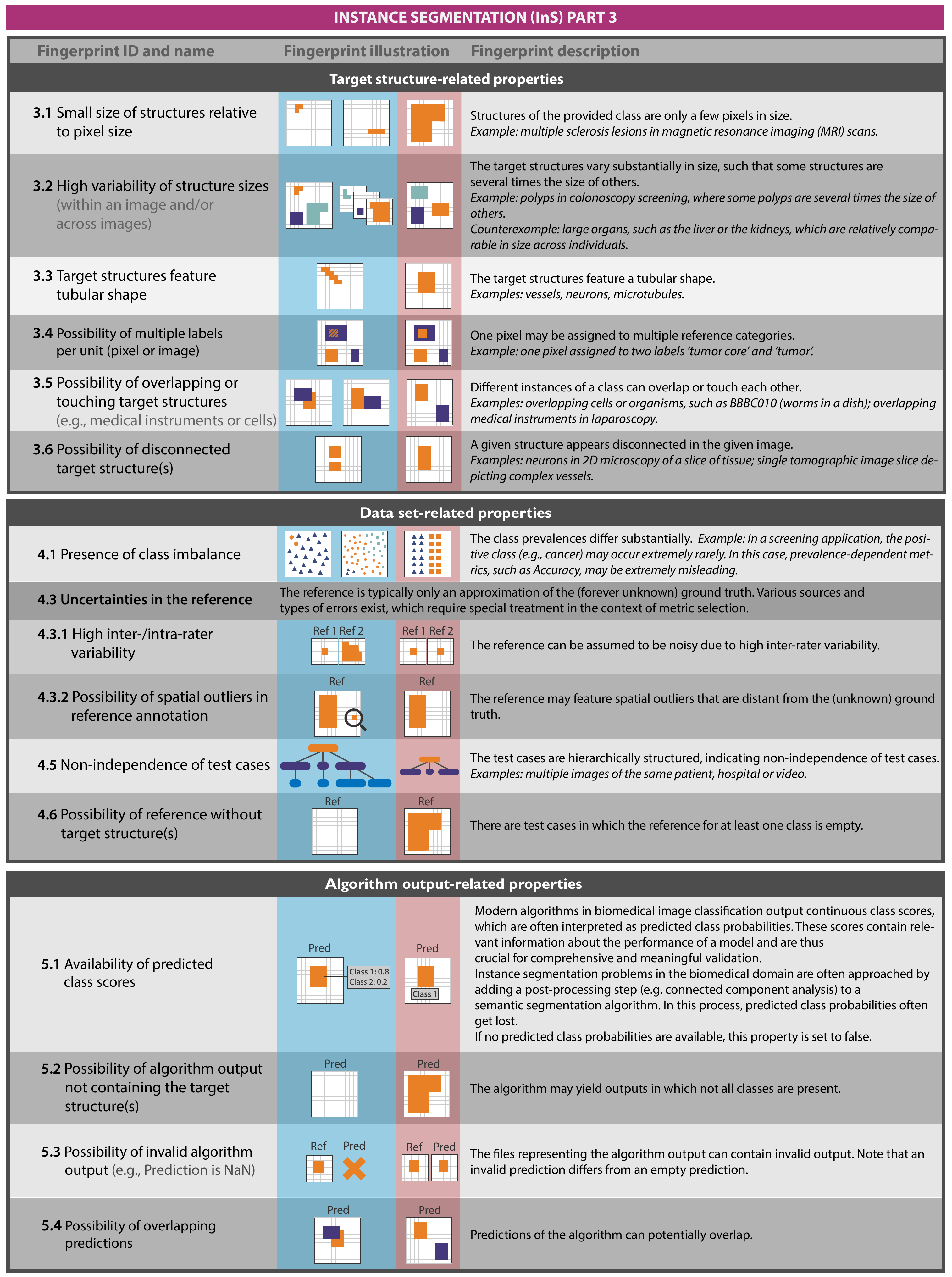}
    \caption{\textbf{Fingerprint for instance segmentation (Part 3).} In the case of binary fingerprint items, the blue column shows examples for which the property is true while the red column shows counterexamples. Categorical fingerprint items are only shown in blue. Suppl. Note~\ref{app:fingerprints} provides more detailed explanations of selected fingerprint items. Used abbreviations: Reference (Ref), Prediction (Pred).}
    \label{fig:fingerprint-is-3}
\end{figure}

\paragraph{FP4.2 Class prevalences reflect the population of interest}
Class prevalences and their differences across data sets are highly important, although this aspect is often ignored in common validation practice. This can best be explained with the example of diagnostic tests, for example image-based disease classification. While several metrics, such as Sensitivity and Specificity, are independent of class frequencies and measure the inherent properties of the test, other metrics, such as Accuracy, measure the test performance for the specific prevalence of the test set. This is not problematic if the class prevalences of the provided test set reflect the population of interest, but can lead to problems otherwise (see Fig.~\ref{fig:prevalence-dependency}). This fingerprint should hence be set to true if either the validation interest is constrained to the data set at hand (no future comparison to data sets with different class prevalences is desired) or no variation of prevalences is expected in other cohorts and upon application of the method. 

\newpage

%TODO 2.5.5 Penalization of errors:  Compensation for class imbalances requested

\newpage

%------------------------------------
%\subsection{\textit{Metrics Reloaded} Pool of reference-based metrics}
%\label{app:metric-pool}
%------------------------------------

\newpage
%------------------------------------
\section{STEP 2~-~METRIC SELECTION}
\label{app:metric-selection}
\setcounter{figure}{0}
%------------------------------------
\acresetall
As a foundation for the metric selection process, the \textit{Metrics Reloaded} consortium compiled a set of common reference-based validation metrics (Suppl. Note~\ref{app:metric-pool}). The framework leverages the problem fingerprints to guide the user through the process of selecting an appropriate set of category-specific reference-based validation metrics while being made aware of potential pitfalls related to individual choices. A bird's eye perspective of the process is shown in Fig.~\ref{fig:meta-mapping}. A detailed explanation for the selection of reference-based metrics is provided separately for all four problem categories in Suppl. Notes~\ref{app:ilc}-\ref{app:is}. Details on selecting appropriate calibration metrics, if desired, are given in Suppl. Note~\ref{app:calibration}. The corresponding formal decision trees (subprocesses) along with corresponding decision guides are shown in Extended Data Figs.~\ref{fig:category_mapping}~-~\ref{fig:metric-selection:s9-assignment-strategy} and Suppl. Note~\ref{app:decision_guides}, respectively. % To reduce the complexity for the user, cheat sheets for common settings (e.g. binary classification) are provided in Suppl. Note~\ref{app:mapping-cheat-sheets}. 
%Note that this entire section focuses on common reference-based common metrics can then be complemented by application-specific reference-based metrics as well non-reference-based metrics, measuring run time or carbon footprint, for example. While a recommendation of these is beyond the scope of the present paper, we provide a discussion on this important topic.

%\input{EssentialsBox.tex}

%------------------------------------
\subsection{\textit{Metrics Reloaded} pool of reference-based metrics}
\label{app:metric-pool}
%------------------------------------
\acresetall
The \textit{Metrics Reloaded} pool of common reference-based validation metrics is shown in Tab.~\ref{tab:metrics}.
Most of these metrics are directly or indirectly based on the cardinalities of the \textit{confusion matrix} (i.e., the true (T)/false (F) positives (P)/negatives (N) in binary problems). For the purpose of metric recommendation, we follow the terminology in the sister publication of this work~\cite{reinke2021common} and classify the metrics into \textbf{counting metrics} that operate directly on a single fixed confusion matrix and express the metric value as a function of the cardinalities, \textbf{multi-threshold metrics} that operate on a dynamic confusion matrix, and \textbf{distance-based metrics}, designed to measure differences between boundaries, volumes, center (line)s or shapes \cite{reinke2021common}. In addition to these, our framework considers metrics designed to measure calibration capabilities of models.

\begin{figure}[H]
    \centering
    \includegraphics[width=0.9\linewidth]{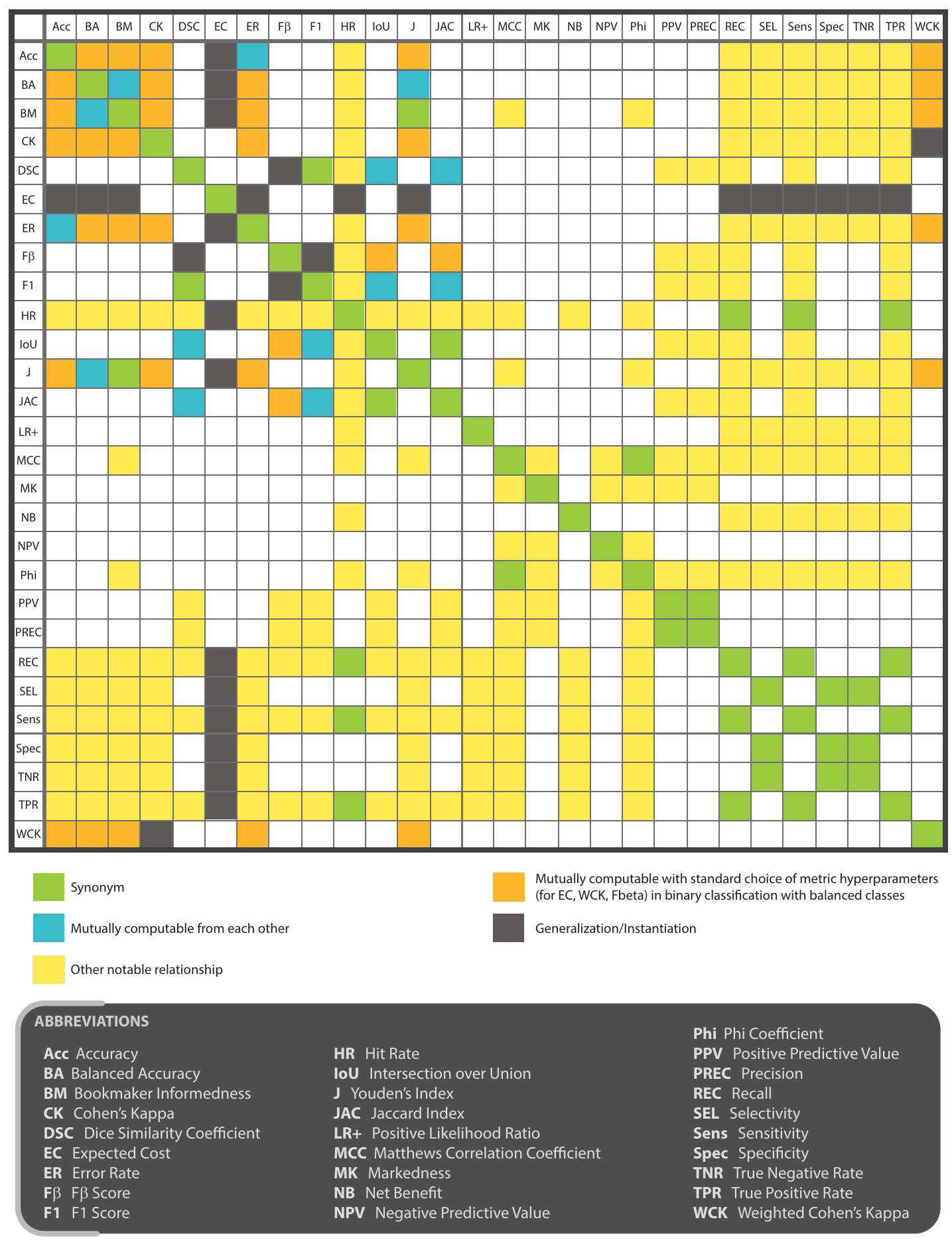}
    \caption{Common counting metrics and their relation to each other. The depicted metrics comprise the counting metrics of the \textit{Metrics Reloaded} pool (Tab.~\ref{tab:metrics}) as well as closely related metrics, namely \acf{ER}, \acf{BM}, \acf{MK}, and \acf{CK} with synonyms. \acf{PQ}, \acf{clDice}, and \acf{FPPI} have been excluded as they rely on more information than solely the confusion matrix.}
    \label{fig:metric-relations}
\end{figure}

Importantly, many popular counting metrics are closely related. In Fig.~\ref{fig:metric-relations}, we categorize the relationship as follows.
\begin{enumerate}
    \item \textcolor{RelGreen}{Synonyms}: For some metrics, various terms exist. Popular examples are:
            \begin{itemize}
                \item Recall = Sensitivity = \ac{TPR} = Hit rate
                \item \ac{PPV} = Precision
             \item \ac{DSC} = F$_\text{1}$ Score (at pixel level)
            \end{itemize}
    \item \textcolor{RelBlue}{Mutually computable from each other}: Some metrics are directly computable from each other without further information.
    Popular examples are:
            \begin{itemize}
                \item Accuracy = 1 - \ac{ER}
                \item \ac{DSC} = (2 * \ac{IoU})/(1 + \ac{IoU})
                \item Balanced Accuracy = (\ac{BM} + 1) / 2
            \end{itemize}
    \item \textcolor{RelGray}{Generalization/Instantiation}: Some metrics are an instantiation of others. 
    Popular examples are:
            \begin{itemize}
                \item \ac{DSC} is an instantiation of the F$_\beta$ Score for $\beta = 1$
                \item Accuracy is a specialization of \ac{EC}, where costs are chosen as "0-1-costs", meaning $c_{ii} = 0$ and $c_{ij} = 1$ otherwise. 
                \item \ac{BA} is a specialization of \ac{EC}, where costs are chosen such that $c_{ii} = 0$ and $c_{ij} = \frac{1}{C P_i}$ with $P_i$ reflecting the class prevalence of class $i$ and $C$ denoting the number of classes.  
            \end{itemize}
    \item \textcolor{RelOrange}{Mutually computable under certain conditions}: Assuming a simple problem setup, additional metrics coincide. Popular examples are:
    \begin{itemize}
        \item Assuming perfect class balance in a binary problem, \ac{BA} = Accuracy = \ac{CK}
        \item Assuming $\beta = 1$ allows to compute \ac{IoU} and Jaccard index from F$_\beta$ Score
    \end{itemize}
    \item \textcolor{RelYellow}{Other notable relationship}: Some metrics share another notable relationship. These are detailed in the metric cheat sheets (Suppl. Note~\ref{app:metric-cheat-sheets})
    \end{enumerate}

 Cheat sheets for all metrics, comprising basic information such as definition and links to reference implementations, relationships to other metrics, and \textit{Metrics Reloaded} recommendations for their usage, can be found in Suppl. Note~\ref{app:metric-cheat-sheets}. 

\newpage
\vspace*{-\baselineskip}
{\tiny
\begin{table}[H]
\caption{\textbf{Overview of recommended reference-based metrics.}
For each metric, name, acronym, synonyms, reference to the definition and illustration, range and corresponding problem categories are provided. The direction of the arrow in the 'range' column indicates whether higher (up) or lower scores (down) are better. A detailed introduction and discussion of all metrics can be found in the sister publication of this work~\cite{reinke2021common}. ImLC: image-level classification; SemS: semantic segmentation; ObD: object detection; InS: instance segmentation.}
\label{tab:metrics}

\begin{tabular}{p{3cm}p{1.2cm}p{3.2cm}cccccc}

\toprule
\textbf{Metric} & \textbf{Acronym} & \textbf{Synonyms} & \textbf{Definition} &  \textbf{Range} & \multicolumn{4}{c}{\textbf{Recommended for}} \\
 & &  &   & & \textbf{ImLC} & \textbf{SemS} & \textbf{ObD} & \textbf{InS} \\

%  Counting Metrics
\midrule
\multicolumn{9}{c}{\textbf{Counting Metrics}} \\
\midrule
\rowcolor[HTML]{C0C0C0} 
Accuracy &  &  & \cite{tharwat2020classification, grandini2020metrics} &  [0, 1] $\uparrow$ & x &  &  &  \\
\rowcolor[HTML]{EFEFEF} 
\acl{BA} & \acs{BA} &  & \cite{tharwat2020classification, grandini2020metrics} &  [0, 1] $\uparrow$  &  x &  &  &  \\
\rowcolor[HTML]{C0C0C0} 
\acl{WCK} & \acs{WCK}  & Cohen's Kappa Coefficient, Kappa Statistic, Kappa Score & \cite{cohen1960coefficient} &  [-1, 1] $\uparrow$  & x &  &  &  \\
\rowcolor[HTML]{EFEFEF} 
\acl{clDice} & \acs{clDice} &  & \cite{shit2021cldice} &  [0, 1] $\uparrow$  &  & x &  & x \\
\rowcolor[HTML]{C0C0C0} 
\acl{DSC} & \acs{DSC} & Sørensen–Dice Coefficient, F$_1$ Score,  Balanced F Score & \cite{dice1945measures} &  [0, 1] $\uparrow$ &  & x &  & x \\
\rowcolor[HTML]{EFEFEF} 
\acl{EC} & \acs{EC} &  & \cite{bishop2006pattern, ferrer2022analysis} &  (-$\infty$, $\infty$) $\downarrow$ & x &  &  &  \\
\rowcolor[HTML]{C0C0C0} 
F$_\beta$ Score &  &  & \cite{Chinchor1992} &  [0, 1] $\uparrow$  & x & x & x & x \\
\rowcolor[HTML]{EFEFEF} 
\acl{FPPI}* & \acs{FPPI} & & \cite{bandos2009area,van2010comparing} &   [0, $\infty$) $\downarrow$  & &  & x & x \\
\rowcolor[HTML]{C0C0C0} 
\acl{IoU} & \acs{IoU} & Jaccard Index, Tanimoto Coefficient & \cite{jaccard1912distribution} &  [0, 1] $\uparrow$  &  & x &  & x \\
\rowcolor[HTML]{EFEFEF} 
\acl{MCC} & \acs{MCC} & Phi Coefficient & \cite{matthews1975comparison} &  [-1, 1] $\uparrow$  & x &  &  &  \\
\rowcolor[HTML]{C0C0C0} 
\acl{PQ} & \acs{PQ} &  & \cite{kirillov2019panoptic} &   [0, 1] $\uparrow$  &  &  &  & x \\
\rowcolor[HTML]{EFEFEF} 
\acl{NB} & \acs{NB} &  & \cite{vickers2016net} &  (-$\infty$,$\infty$) $\uparrow$  & x &  &  & \\
\rowcolor[HTML]{C0C0C0} 
\acl{NPV}* & &  & \citep{tharwat2020classification, bickelhaupt2018radiomics} &  [0, 1] $\uparrow$  &   x &  &  &  \\
\rowcolor[HTML]{EFEFEF} 
\acl{LR+} & \acs{LR+} &  Likelihood Ratio Positive, Likelihood Ratio for Positive Results & \cite{wales2003moving} &  [0, $\infty$) $\uparrow$  & x &  &  &  \\
\rowcolor[HTML]{C0C0C0} 
\acl{PPV}* & \acs{PPV} & Precision & \citep{tharwat2020classification, grandini2020metrics, bickelhaupt2018radiomics} &  [0, 1] $\uparrow$  &  x &  & x & x \\
\rowcolor[HTML]{EFEFEF} 
Sensitivity* & &  Recall, Hit Rate, \acf{TPR} & \citep{tharwat2020classification, grandini2020metrics, bickelhaupt2018radiomics} &  [0, 1] $\uparrow$  &  x &  & x & x \\
\rowcolor[HTML]{C0C0C0} 
Specificity* &  & Selectivity, \acf{TNR} & \citep{tharwat2020classification, grandini2020metrics, bickelhaupt2018radiomics}  &  [0, 1] $\uparrow$ &  x &  &  &  \\
% \rowcolor[HTML]{EFEFEF} 
% Sensitivity@\acl{FPPI} & Sensitivity @\acs{FPPI} &  Sensitivity: Recall, Hit Rate, True Positive Rate & \citep{tharwat2020classification, grandini2020metrics, bickelhaupt2018radiomics} &   [0, 1] $\uparrow$ &    &  & x & x \\
% \rowcolor[HTML]{C0C0C0} 
% Sensitivity@Specificity &  & Sensitivity: Recall, Hit Rate, True Positive Rate; Specificity: Selectivity, True Negative Rate & \citep{tharwat2020classification, grandini2020metrics, bickelhaupt2018radiomics}  &  [0, 1] $\uparrow$ &   x &  &  &  \\
% \rowcolor[HTML]{EFEFEF} 
% Youden's Index & J & Youden's~J Statistic, Youden's~J, \acf{BM} & \citep{youden1950index} &  Fig. \ref{fig:def-classification-2} & [-1, 1] $\uparrow$ & Figs. \ref{fig:lr+ba}, \ref{fig:misleading-ba-j} & x &  &  & \\ 

% Multi-threshold Metrics
\rowcolor[HTML]{FFFFFF}
\midrule
\multicolumn{9}{c}{\textbf{Multi-threshold Metrics}} \\
\midrule
\rowcolor[HTML]{C0C0C0} 
\acl{AUROC} & \acs{AUROC} &  Area under the curve (AUC), AUC Receiver Operating Characteristic (ROC), C-Index, C-Statistics &  \cite{hanley1982meaning} &   [0, 1] $\uparrow$  & x &  &  &  \\
\rowcolor[HTML]{EFEFEF} 
\acl{AP} & \acs{AP} &  & \cite{lin2014microsoft} &  [0, 1] $\uparrow$ & x &  & x & x \\
\rowcolor[HTML]{C0C0C0} 
\acl{FROC} Score & \acs{FROC} Score &  & \cite{bandos2009area,van2010comparing} &   [0, 1] $\uparrow$ &  &  & x & x \\
\rowcolor[HTML]{FFFFFF}
% Distance-based Metrics
\midrule
\multicolumn{9}{c}{\textbf{Distance-based Metrics}} \\
\midrule
\rowcolor[HTML]{C0C0C0} 
\acl{ASSD} & \acs{ASSD} &  & \cite{yeghiazaryan2015overview} &  [0, $\infty$) $\downarrow$  &  & x &  & x \\
\rowcolor[HTML]{EFEFEF} 
Boundary \acl{IoU} & Boundary IoU &  & \cite{cheng2021boundary} &  [0, 1] $\uparrow$  & & x &  & x \\
\rowcolor[HTML]{C0C0C0} 
\acl{HD} & \acs{HD} & Hausdorff Metric, Pompeiu–Hausdorff Distance, Maximum Symmetric Surface Distance & \cite{huttenlocher1993comparing} &  [0, $\infty$) $\downarrow$  &  & x &  & x \\
\rowcolor[HTML]{EFEFEF} 
\acl{MASD} & \acs{MASD} &  & \cite{benevs2015performance} & [0, $\infty$) $\downarrow$  &  & x &  & x \\
\rowcolor[HTML]{C0C0C0} 
\acl{NSD} & \acs{NSD} & Normalized Surface Dice, Surface Distance, Surface Dice & \cite{nikolov2021clinically} &  [0, 1] $\uparrow$  &  & x &  & x \\
\rowcolor[HTML]{EFEFEF} 
\acl{X$^{th}$ Percentile HD} & X$^{th}$ Percentile~HD &  & \cite{huttenlocher1993comparing} &  [0, $\infty$) $\downarrow$  &  & x &  & x \\

% Calibration Metrics
\rowcolor[HTML]{FFFFFF}
\midrule
\multicolumn{9}{c}{\textbf{Calibration Metrics}}\\
\midrule
\rowcolor[HTML]{C0C0C0} 
\acl{BS} & \acs{BS} & &  \cite{brier1950verification} & [0, 1] $\downarrow$  & x &  & x  &  x\\
\rowcolor[HTML]{EFEFEF} 
\acl{CWCE} & \acs{CWCE} & &  \cite{kull2019beyond, kumar2019verified} & [0, 1] $\downarrow$  & x &  & x  &  x\\
\rowcolor[HTML]{C0C0C0} 
\acl{ECE} & \acs{ECE} & &  \cite{naeini2015obtaining,guoCalibrationModernNeural2017} & [0, 1] $\downarrow$  & x &  & x  &  x\\
\rowcolor[HTML]{EFEFEF} 
\acl{ECEKDE} & \acs{ECEKDE} & &  \cite{popordanoska2022consistent} & [0, 1] $\downarrow$  & x &  & x  &  x\\
\rowcolor[HTML]{C0C0C0} 
\acl{KCE} & \acs{KCE} & &  \cite{widmann2019calibration, gruber2022better} & [0, 1] $\downarrow$ & x &  & x  &  x\\
\rowcolor[HTML]{EFEFEF} 
\acl{NLL} & \acs{NLL} & Cross Entropy Loss &  \cite{cybenko1998mathematics} & [0, $\infty$) $\downarrow$ & x &  &   &  \\
\rowcolor[HTML]{C0C0C0} 
\acl{RBS} & \acs{RBS} & &  \cite{gruber2022better} & [0, 1] $\downarrow$ & x &  & x  &  x\\
\bottomrule
\multicolumn{9}{p{0.9\linewidth}}{ *: This metric is best used in combination with another metric using a predefined target value (see "Target value-based cutoff" in the definition of \textit{FP2.6: Cutoff on predicted class scores} (Suppl. Note~\ref{app:fingerprint-category}).} \\
\end{tabular}
\end{table}
\normalsize
}

\newpage
 \hfill \\

\newpage

%------------------------------------
\newpage
\subsection{\textcolor{ILCblue}{\textbf{Recommendations for Image-level Classification}}}
\label{app:ilc}
%------------------------------------
\acresetall

\begin{tcolorbox}[title=Essentials]
    \textbf{FPX.Y} refers to a fingerprint item detailed in Figs.~\ref{fig:fingerprint-ilc}-\ref{fig:fingerprint-ilc-3}. \\

    \textbf{SX} refers to a subprocess in Extended Data Figs.~\ref{fig:metric-selection:s2-multi-class}-\ref{fig:metric-selection:s5-calibration}. \\

    \textbf{DGX.Y} refers to a decision guide in Suppl. Notes~\ref{ssec:dg2}-\ref{ssec:dg5}.
\end{tcolorbox}

This section provides recommendations for selecting \textit{common reference-based metrics} for image-level classification problems. As depicted in Fig.~\ref{fig:meta-mapping}, these common metrics can then be complemented by application-specific metrics as well as non-reference-based metrics (assessing run time or carbon footprint, for example).

Image-level classification refers to the process of assigning one or multiple labels (\textit{classes}) to an image. Modern algorithms usually output \textbf{predicted class scores} between 0 and 1 for every image and class, which are often interpreted as the probability of the image belonging to a specific class. In binary classification, a threshold can be applied to convert the continuous scores to a classification decision (e.g. cancer = true for values above 0.5). In multi-class classification, the class associated with the highest predicted score is often selected as the final prediction ('argmax' operation). The most common strategies for converting predicted class scores into discrete decisions are captured in the fingerprint \textit{FP2.6 Decision rule applied to  predicted class scores} and are detailed in Suppl. Note~\ref{app:fingerprints}. 

Comparing the algorithm predictions with the reference labels enables the generation of a confusion matrix, which captures the number of correct class assignments on the diagonal for each class and the numbers for all possible class confusions in the remaining cells. In the binary case, these numbers, here referred to as the \textit{cardinalities}, are simply the true/false positives/negatives arranged in a 2$\times$2 matrix. \textbf{Counting metrics} operate on this matrix by relating the cardinalities of different matrix entries~\cite{reinke2021common}. They can be classified into \textbf{multi-class counting metrics} that operate on the full, potentially multi-class confusion matrix, such as Accuracy, \ac{MCC} and \ac{EC}, and \textbf{per-class counting metrics} that validate the performance of a particular class of interest defined as the \textit{positive class} (e.g. with a one-vs-rest comparison for multi-class scenarios), such as the F$_\beta$ Score. Per-class validation is typically recommended (see below) to obtain an in-depth understanding of the performance of each individual class, as multi-class metrics may potentially hide poor performance of individual classes. All counting metrics differ exclusively in which cardinalities of the confusion matrix they use and how they are combined. 
%REVISION: The metric properties that the \textit{Metrics Reloaded} consortium deemed particularly relevant for assessing the discrimination capabilities of a classifier via counting metrics are provided in Tab.~\ref{tab:metric-properties}. 
%TODO PAUL SUMMARIZE MOST IMPORTANT FINGERPRINTS
\newpage
\begin{figure}[H]
    \centering
    \includegraphics[width=\textwidth]{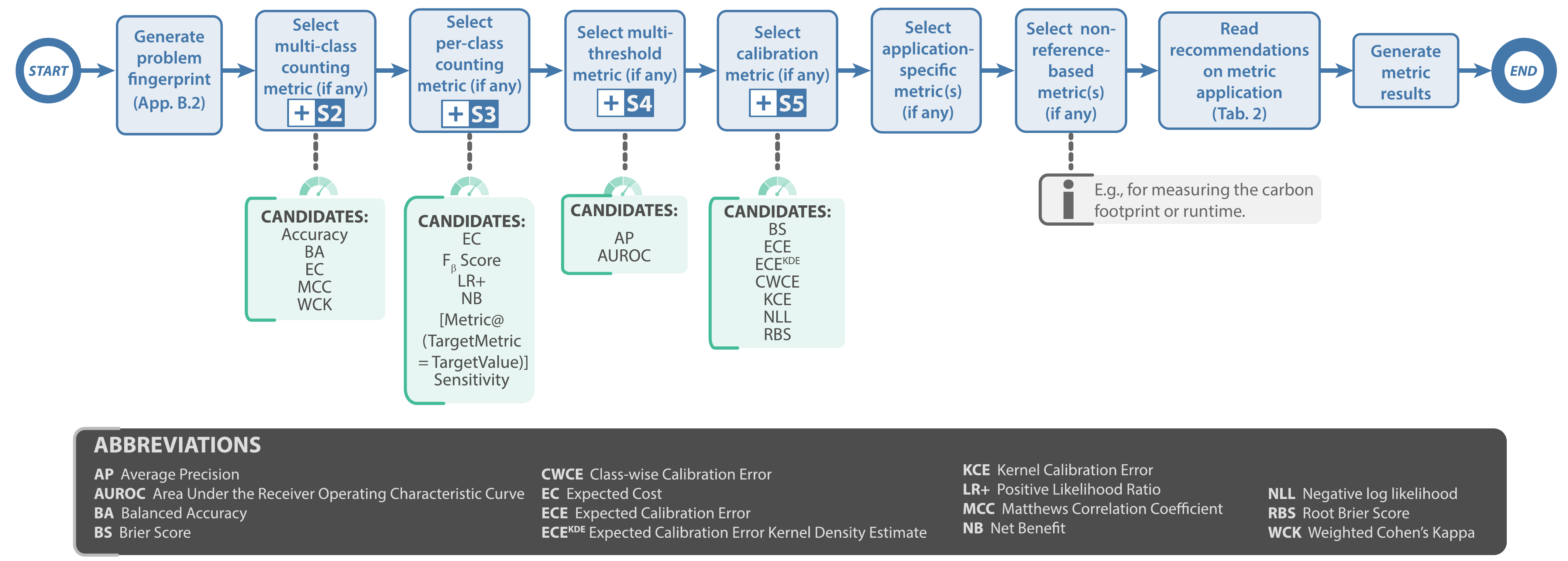}
    \caption{\textit{Metrics Reloaded} recommendation framework for image-level classification at a glance.}
    \label{fig:reduced-metamapping-ilc}
\end{figure}

Counting metrics in general reflect the fact that systems in practice need to define a strategy for converting the predicted class scores (if available) into actual decisions. Choosing a decision rule for the generation of a confusion matrix, however, is not necessarily straightforward, and counting metrics may fail to capture the full capacities of a classifier by restricting performance analysis to a single working point on the decision curve \cite{reinketizabi2022metrics} (Fig.~\ref{fig:cutoff-strategies}). \textbf{Multi-threshold metrics} (Fig.~\ref{fig:def-auc} such as \ac{AUROC} overcome the limitation of a potentially arbitrary threshold by calculating metric scores based on a range of thresholds. They are commonly only defined for binary classification (again, one-versus-rest validation can be performed) and relate basic complementary properties, such as Sensitivity and Specificity in the case of \ac{AUROC}, or Sensitivity and \ac{PPV} in the case of \ac{AP}, to each other. The metric value is then obtained by computing the area under the resulting curve \cite{reinketizabi2022metrics}.

While both counting metrics and multi-threshold metrics measure the \textit{discrimination} capabilities of a classifier, they do not assess whether the predicted class scores reflect the true probability of cases belonging to the predicted class. An orthogonal class of metrics has therefore been designed to assess the interpretability of classifier outputs. As detailed in Suppl. Note\ref{app:calibration}, these \textbf{calibration metrics} can roughly be categorized in metrics that assess discrimination and calibration quality together, such as the \ac{BS}, and those that assess only calibration, such as the \ac{ECE}).

\begin{figure}[H]
    \centering
    \includegraphics[width=0.6\linewidth]{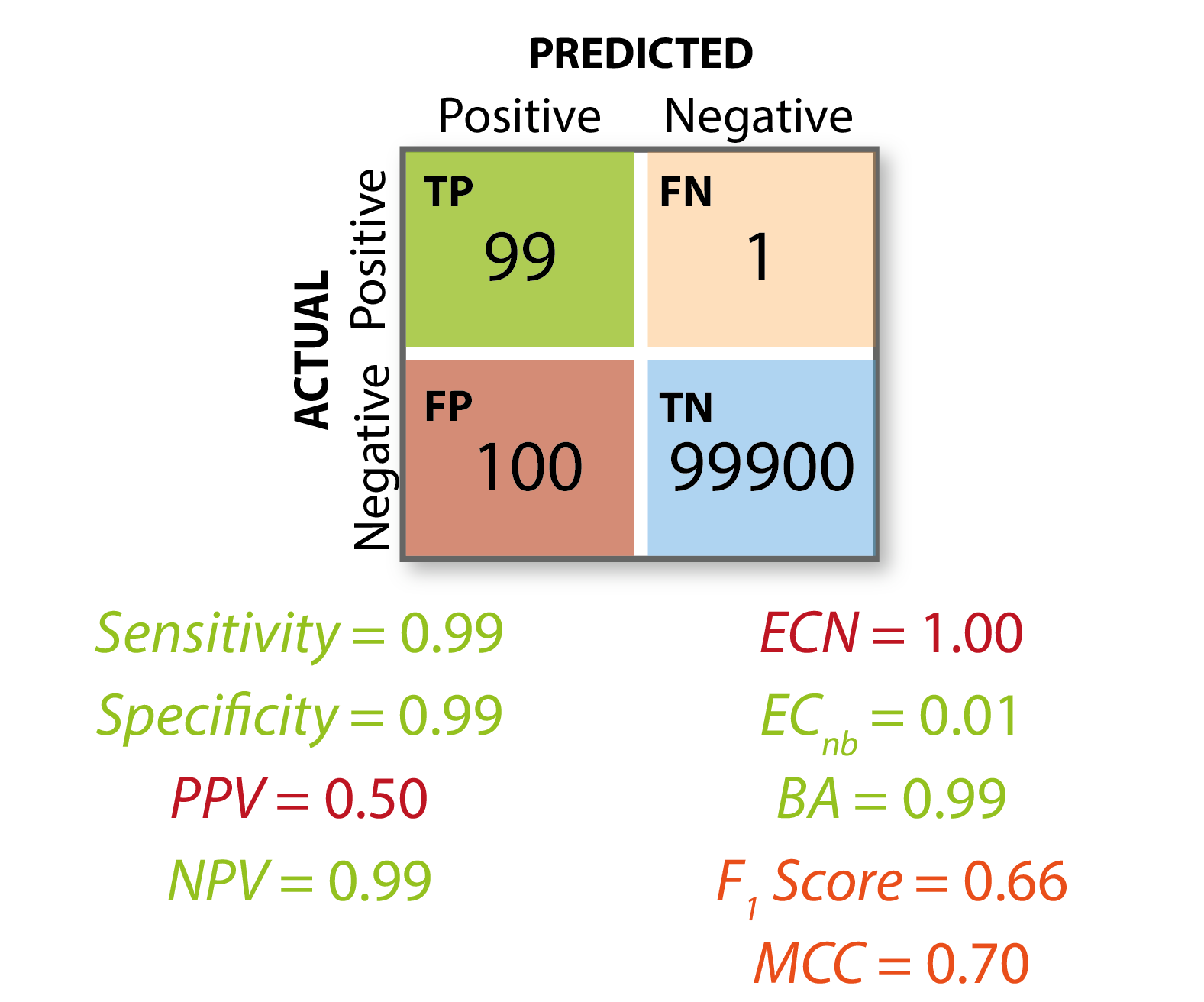}
    \caption{Failure of prevalence-independent metrics in a screening scenario with high class imbalance. Intuitively, the system is substantially better than random guessing because almost all positive cases have been retrieved out of a large database. At the same time, it is not perfect because only about half of the retrieved cases are correctly classified. However, common popular metrics either indicate near-perfect performance (Sensitivity and Specificity close to 1) or random performance (\acf{ECN} about 1). Only the F$_\text{1}$ Score and \acf{MCC} reflect the intuitive scoring of “quite good, but not perfect” because they incorporate predictive values.}
    \label{fig:cm-retrieval}
\end{figure}

Taking into account these considerations as well as the complementary strengths and weaknesses of classification metrics, we recommend the following process for selecting reference-based classification metrics (blue path in Fig.~\ref{fig:meta-mapping} and Fig.~\ref{fig:reduced-metamapping-ilc}): 
\begin{description}
    \item[1: Select multi-class counting metric (if any):] Multi-class counting metrics have the unique advantage that they capture the performance of an algorithm for all classes in a single value. With the ability to take into account all entries of the multi-class confusion matrix, they provide a holistic measure of performance without the need for customized class-aggregation schemes. We therefore recommend the selection of a multi-class counting metric based on Subprocess S2 (Extended Data Fig.~\ref{fig:metric-selection:s2-multi-class}) if a decision rule should be applied to the predicted class scores (FP2.6). In some use cases and especially in the presence of ordinal data, there may be an unequal severity of class confusions (FP2.5.2 = TRUE), implying that different costs to be applied to different errors reflected by the confusion matrix must be available (FP2.5.4 = TRUE). In this case, the only viable options are \ac{WCK} (Fig.~\ref{fig:cheat-sheet-wck}) and \ac{EC} (Fig.~\ref{fig:cheat-sheet-ec}). While \ac{WCK} is widely used, it comes with severe drawbacks (see Suppl. Note~\ref{ssec:dg2} for details), such as high prevalence dependency and 'paradoxical results'~\citep{warrens2012some} for the most common variant based on quadratic weights. For this reason, the consortium recommends \ac{EC} as the default choice for the described scenario. In the case of equal costs, Accuracy (Fig.~\ref{fig:cheat-sheet-accuracy}) is the most widely used multi-class metric, but we recommend it in only one specific scenario: when the class prevalences in the data set reflect those in the target population (FP4.2) and potential class imbalances should not be compensated for. In the more general case, the decision boils down to either picking one of the prevalence-independent metrics \ac{EC} or \ac{BA} (Fig.~\ref{fig:cheat-sheet-ba}), which is specifically recommended if the class prevalences do \textit{not} reflect the target population, or \ac{MCC} (Fig.~\ref{fig:cheat-sheet-mcc}), which has the important property that it requires not only the class-specific Sensitivities (i.e. Sensitivity and Specificity in the binary case) but also the corresponding predictive values (\ac{PPV} and \ac{NPV}) to be high (see Fig.~\ref{fig:cm-retrieval}). Irrespective of the metric choice, we recommend additionally reporting the whole confusion matrix in the case of a reasonable number of classes.
    \item[2: Select per-class counting metric (if any):] As detailed class-specific analyses are not possible with multi-class counting metrics, which may potentially hide the poor performance of individual classes, we recommend an additional per-class validation with metrics selected according to Subprocess S3 (Extended Data Fig.~\ref{fig:metric-selection:s3-single-threshold}). To this end, class-specific metric pools are generated. The choice of metric depends primarily on the decision rule applied to the predicted class scores (FP2.6; see Suppl. Note~\ref{app:fingerprints} for a detailed explanation). If a \textbf{target value-based} strategy is preferred, the decision rule applied to the predicted class scores is optimized such that a specific target value (e.g. Sensitivity = 0.95; see Fig.~\ref{fig:cheat-sheet-sensitivity}) is achieved (see Fig.~\ref{fig:cutoff-strategies}). Complementary metrics, such as Specificity (Fig.~\ref{fig:cheat-sheet-specificity}), can then be reported for this fixed value of the target metric (see decision guide 3.1 in Suppl. Note~\ref{ssec:dg3}). In this case, the target metric is only reported for the specified target class. If a \textbf{cost-benefit-based} strategy is chosen (only recommended for binary classification tasks), we recommend selecting either \ac{NB} (explicit risk-centric view; Fig.~\ref{fig:cheat-sheet-nb}) or \ac{EC} (cost-centric view; Fig.~\ref{fig:cheat-sheet-ec}) (see decision guide 3.2 in Suppl. Note~\ref{ssec:dg3}). In contrast, in the case of \textbf{optimization-based} or \textbf{argmax-based} decision rules, the  metric choice  should be made between Sensitivity, \ac{LR+} (Fig.~\ref{fig:cheat-sheet-lr+}), and F-$_\beta$ Score (Fig.~\ref{fig:cheat-sheet-fbeta}) (see decision guide 3.3 and 3.4 in Suppl. Note~\ref{ssec:dg3}).
    \item[3: Select multi-threshold metric:] 
   To obtain a more comprehensive picture of the discrimination performance of a classifier, we always recommend the selection of a multi-threshold metric according to Subprocess S4 (Extended Data Fig.~\ref{fig:metric-selection:s4-multi-threshold}), irrespective of the decision rule. Multi-threshold metrics are again applied per class. A particular strength of \ac{AUROC} (Fig.~\ref{fig:cheat-sheet-auroc}) is the fact that it is well-interpretable, as the value simply reflects the probability of a sample from the positive class being assigned a higher predicted class score compared to a sample from the negative class. Furthermore, it is prevalence-independent and therefore well-suited for comparison of performance across different data sets. \ac{AP} (Fig.~\ref{fig:cheat-sheet-ap}), on the other hand, is a prevalence-dependent metric, which comes with the advantage that predictive values are considered. This may be a crucial property in class-imbalanced scenarios where the focus is to be put on the rare class while \ac{AUROC} scores are dominated by the frequent class and may lead to overly optimistic interpretation.
     \item[4: Select calibration metric (if any):] If the calibration of a method should be assessed in addition to its discrimination capabilities (FP2.7.1), one or multiple calibration metrics should be chosen based on Subprocess S5  (Extended Data Fig.~\ref{fig:metric-selection:s5-calibration}). Details on this process are provided in Suppl. Note~\ref{app:calibration}.
\end{description}

\newpage

\newpage

%------------------------------------
\subsection{\textcolor{SSyellow}{\textbf{Recommendations for Semantic segmentation}}}
\label{app:ss}
%------------------------------------
\acresetall
\begin{tcolorbox}[title=Essentials]
    \textbf{FPX.Y} refers to a fingerprint item detailed in Figs.~\ref{fig:fingerprint-ss-1}-\ref{fig:fingerprint-ss-2}. \\

    \textbf{SX} refers to a subprocess in Extended Data Figs.~\ref{fig:metric-selection:s6-overlap}-\ref{fig:metric-selection:s7-boundary}. \\

    \textbf{DGX.Y} refers to a decision guide in Apps.~\ref{ssec:dg6}-\ref{ssec:dg7}.
\end{tcolorbox}

This section provides recommendations for selecting \textit{common reference-based metrics} for semantic segmentation problems. As depicted in Fig.~\ref{fig:meta-mapping}, these common metrics can then be complemented by application-specific metrics as well as non-reference-based metrics (assessing run time or carbon footprint, for example).

Semantic segmentation is commonly defined as the process of partitioning an image into multiple segments/regions. To this end, one or multiple labels are assigned to each pixel such that pixels with the same label share certain characteristics. Semantic segmentation can therefore also be regarded as pixel-level classification. As in image-classification problems, predicted class probabilities are typically calculated for each pixel deciding on the class affiliation based on a threshold over the class scores \cite{asgari2021deep}. In semantic segmentation problems, the pixel-level classification is typically followed by a post-processing step, in which boundary pixels are identified. 

The most common semantic segmentation metrics (e.g. \ac{DSC} and \ac{IoU}) are per-class counting metrics -- here referred to as \textbf{overlap-based metrics}, which measure the overlap between the reference annotation and the prediction of the algorithm. They can be considered the \textit{de facto} standard for assessing segmentation quality and are well-interpretable. 

A key weakness of overlap-based metrics is their shape and contour unawareness~\cite{reinke2021common}. A second class of metrics, the \textbf{distance-based metrics}, therefore explicitly assess certain spatial characteristics such as the quality of structure centers or boundaries. 
Note that in scenarios in which multiple structures of the same type are present within the same image (e.g., in \ac{MS} lesion segmentation), a potential pitfall is related to comparing a given structure boundary to the boundary of the wrong instance in the reference (Fig.~\ref{fig:boundary-is-ss}). Similar issues arise in the case of completely missed instances. In such scenarios, we explicitly recommend reconsideration to phrase the problem as an instance segmentation problem. If semantic segmentation remains the chosen category, we advise against the use of distance-based metrics, as these are not designed for cases where confusion of boundaries across different instances can occur.
\newpage
\begin{figure}[H]
    \centering
    \includegraphics[width=1\textwidth]{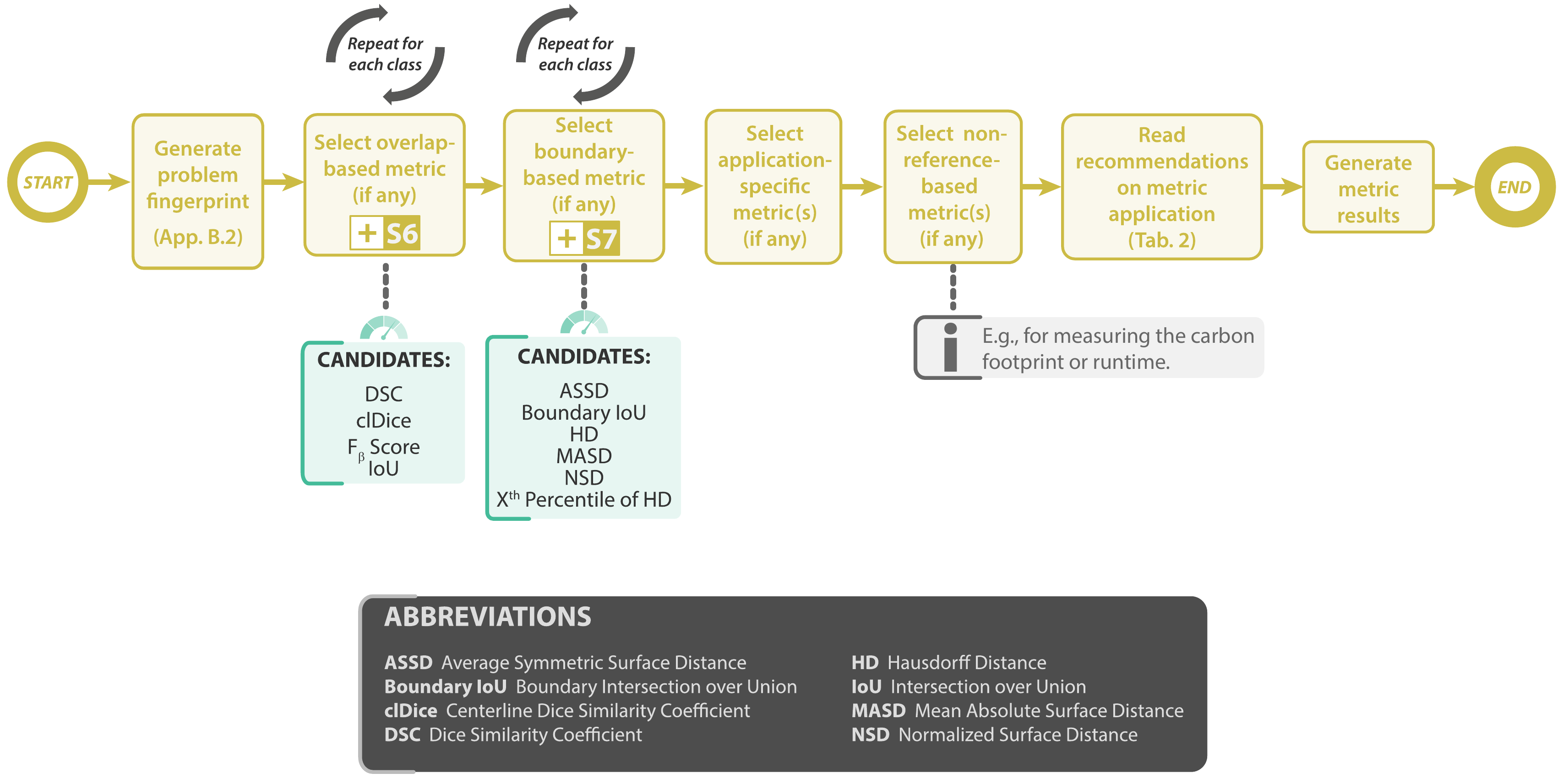}
    \caption{\textit{Metrics Reloaded} recommendation framework for semantic segmentation at a glance.}
    \label{fig:reduced-metamapping-ss}
\end{figure}

Based on the complementary strengths and weaknesses of common segmentation metrics~\cite{reinke2021common}, we recommend the following process for segmentation problems (orange path in Fig.~\ref{fig:meta-mapping} and Fig.~\ref{fig:reduced-metamapping-ss}): 

\begin{description}
    \item[1: Select overlap-based metric (if any):] We recommend selecting an overlap-based metric by default unless the target structures are consistently small (FP3.1) \textit{and} the reference may be noisy (FP4.3.1). As detailed in Subprocess S6 for selecting overlap-based metrics (Extended Data Fig.~\ref{fig:metric-selection:s6-overlap}), our default recommendation is the \ac{DSC} (Fig.~\ref{fig:cheat-sheet-dsc}), which is almost identical to the \ac{IoU} (Fig.~\ref{fig:cheat-sheet-iou}). The F$_\beta$ Score (Fig.~\ref{fig:cheat-sheet-fbeta}), which can be seen as a generalization of the \ac{DSC} = F$_\text{1}$ Score, is an alternative if there is a preference for either \ac{FP} or \ac{FN}. In the specific case of tubular structures (FP3.3), the \ac{clDice} (Fig.~\ref{fig:cheat-sheet-cldice}) as an increasingly used variant of the \ac{DSC} can also be applied (optionally in addition to the \ac{DSC}). 
     \item[2: Select boundary-based metric (if any):]
      To compensate for the weakness of overlap-based metrics, specifically their shape unawareness and limitations when dealing with small structures or high size variability~\cite{reinke2021common}, our general recommendation is to complement an overlap-based metric with a boundary-based metric according to Subprocess S7 (Extended Data Fig.~\ref{fig:metric-selection:s7-boundary}). If annotation imprecisions should be compensated for, our default recommendation is the \ac{NSD} (Fig.~\ref{fig:cheat-sheet-nsd}). Otherwise, the fundamental user preference guiding metric selection is whether errors should be penalized by existence or distance (FP2.5.6). In the case of existence-based penalization, Boundary \ac{IoU} (Fig.~\ref{fig:cheat-sheet-boundary-iou}) should be preferred over \ac{NSD} if even slight contour errors can be seen as crucial inconsistencies that should be assessed. In the case of distance-based penalization, \ac{MASD} (Fig.~\ref{fig:cheat-sheet-masd}) is our default recommendation, as it has mathematical advantages over \ac{ASSD} (Fig.~\ref{fig:cheat-sheet-assd}; see decision guide 7.2 in Suppl. Note~\ref{ssec:dg7}) and is not as sensitive to annotation outliers as \ac{HD} and its variants (Fig.~\ref{fig:cheat-sheet-hd}).
\end{description}

While overlap- and distance-based metrics are the standard metrics used by the general computer vision community, biomedical applications sometimes have special domain-specific requirements. To accommodate specific domain needs, the standard metrics can therefore be complemented by further 'application-specific' metrics as shown in Fig.~\ref{fig:meta-mapping}. In medical imaging, for example, the actual volume of an object, for example a tumor, may be of particular interest (FP2.2). In this case, \textbf{volume metrics} such as the \textit{Absolute} or \textit{Relative Volume Error} and the \textit{Symmetric Relative Volume Difference} can be computed \cite{nai2021comparison}. Also, the clDice can be complemented by application-specific connectivity metrics, for instance in the case of tubular structures \cite{cote2013tractometer, nath2020tractography}.  Similarly, the explicit agreement of object centers (e.g., in cells) or shapes may be of interest. Note that the latter can often be addressed by choosing a boundary metric with a high tolerance. In other cases, shape agreement may be measured by comparing specific object properties, such as curvatures or principal components. 
 Finally, compliance with prior knowledge, such as hierarchical label structure (FP3.4), can be measured with additional application-specific metrics. 

\begin{figure}[H]
    \centering
    \includegraphics[width=0.7\linewidth]{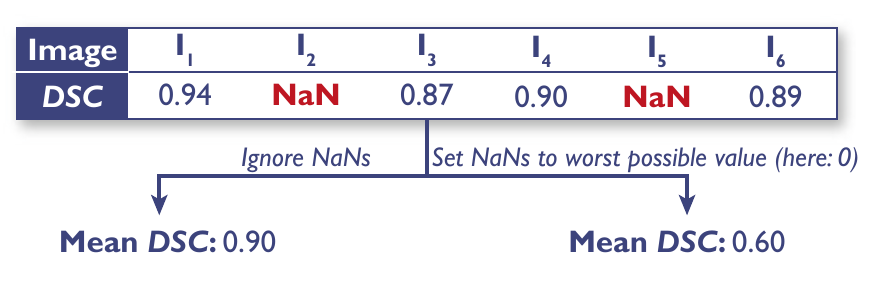}
    \caption{Effect of missing values when aggregating metric values. In this example, ignoring missing values leads to a substantially higher \acf{DSC} compared to setting missing values to the worst possible value (here: 0).}
    \label{fig:missings}
\end{figure}

\begin{figure}[H]
    \centering
    \includegraphics[width=0.6\linewidth]{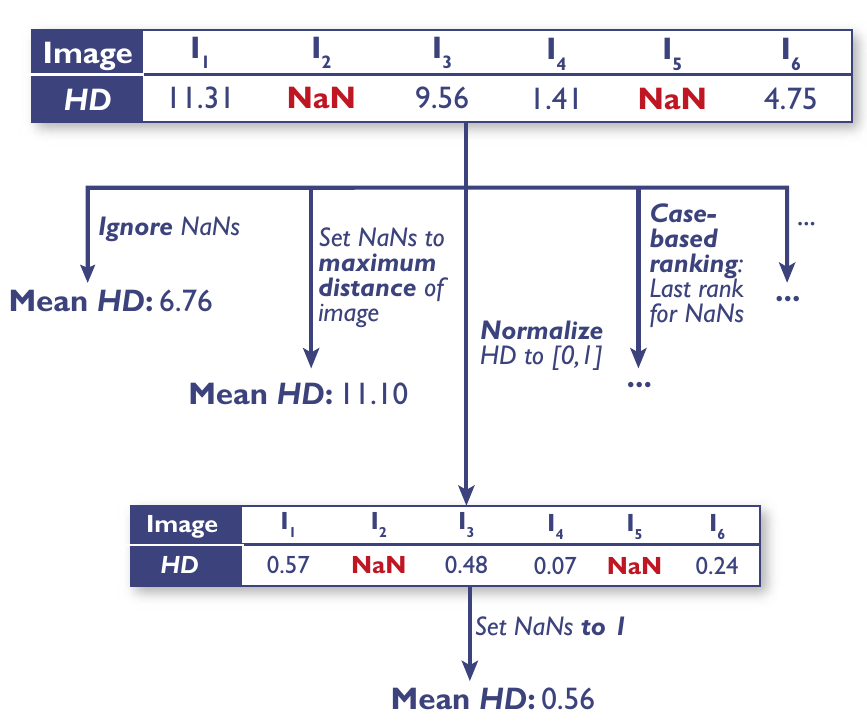}
    \caption{Effect of missing values when aggregating metric values for metrics without fixed boundaries (here: \acf{HD}). In this example, ignoring or treating missing values in different ways leads to substantially different \ac{HD} values.}
    \label{fig:missings-hd}
\end{figure}

Once a set of metrics has been selected, an appropriate \textbf{aggregation strategy} should be chosen. We recommend handling of \texttt{\acp{NaN}} by setting the corresponding metric value to the worst possible value (see Fig.~\ref{fig:missings}). In the case of distance-based metrics such as the \ac{HD}, the image diagonal can be chosen, for example (see Fig.~\ref{fig:missings-hd}). In a benchmarking setting, an alternative lies in using a "rank-then-aggregate" strategy~\cite{wiesenfarth2021methods}. A test case with a \texttt{\ac{NaN}} value can then be assigned the worst rank for the given image.
%\newpage

\newpage
\begin{figure}[H]
    \centering
    \includegraphics[width=0.7\linewidth]{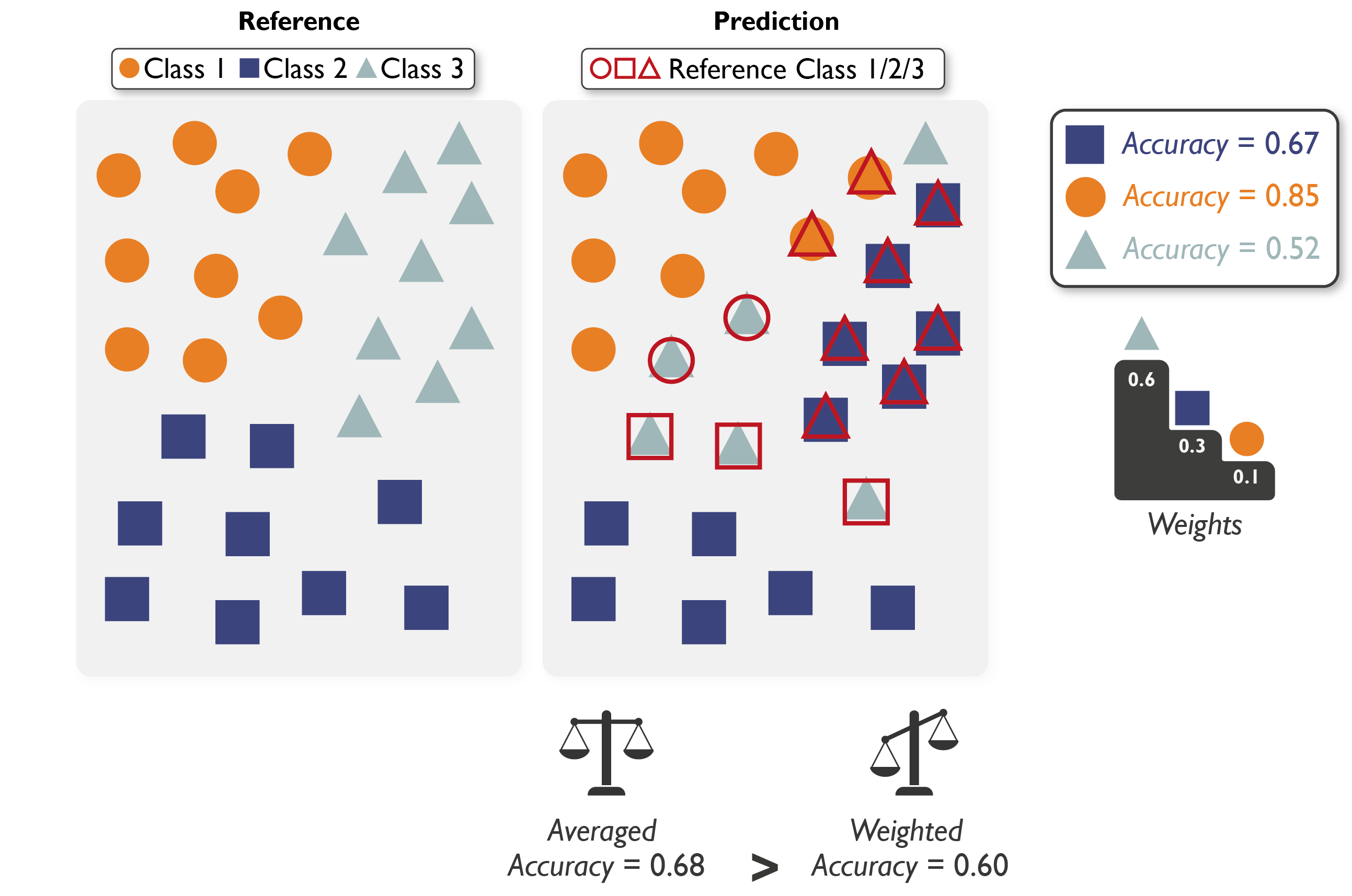}
    \caption{Effect of unequal handling of classes. Simple averaging (macro-averaging) of the Accuracy ignores the unequal importance of classes, given by pre-defined weights of classes. Incorrect predictions are indicated by a red square.}
    \label{fig:unequal-class}
\end{figure}

In multi-class settings, the metric values for the individual classes can be combined in a single score. This can be done via \textit{macro averaging} over class-specific scores, indicating equal importance for each class (FP2.5.1 = FALSE) and an interest to compensate for potential class imbalance (FP2.5.5 = TRUE). Alternatively, weighted averaging, which takes the unequal interest across classes and/or different class prevalences into account, may be performed (Fig.~\ref{fig:unequal-class}).

\newpage

%------------------------------------
\subsection{\textcolor{ODgreen}{\textbf{Recommendations for Object detection}}}
\label{app:od}
%------------------------------------
\acresetall
\begin{tcolorbox}[title=Essentials]
    \textbf{FPX.Y} refers to a fingerprint item detailed in Figs.~\ref{fig:fingerprint-od-1}-\ref{fig:fingerprint-od-3}. \\

    \textbf{SX} refers to a subprocess in Extended Data Figs.~\ref{fig:metric-selection:s3-single-threshold}-\ref{fig:metric-selection:s4-multi-threshold} and Extended Data Figs.~\ref{fig:metric-selection:s8-localization-criterion}-\ref{fig:metric-selection:s9-assignment-strategy}. \\

    \textbf{DGX.Y} refers to a decision guide in Suppl. Note~\ref{ssec:dg3}~-~\ref{ssec:dg5}, \ref{ssec:dg8}-\ref{ssec:dg9}.
\end{tcolorbox}

This section provides recommendations for selecting \textit{common reference-based metrics} for object detection problems. As depicted in Fig.~\ref{fig:meta-mapping}, these common metrics can then be complemented by application-specific metrics as well as non-reference-based metrics (assessing run time or carbon footprint, for example).

Object detection refers to the detection and localization of structures of one or multiple categories. A key feature of object detection algorithms is their ability to distinguish different instances of the same class, which may be of crucial domain interest (see \ref{fig:metrics_matter_overview}). The confusion matrix is generated by comparing reference objects to predicted objects. Based on their matching (see below), \acp{TP} (prediction matched to reference object), \acp{FP} (prediction without assigned reference object) and \acp{FN} (references without assigned predictions) can be computed. While the design choices in image-level classification are primarily related to the selection of discrimination and calibration metrics (see Suppl. Note~\ref{app:ilc}), additional design decisions must be made in object detection due to the object-centric validation. Specifically, a \textbf{localization criterion} must be chosen to determine whether a predicted object spatially corresponds to one of the reference objects and vice versa. To this end, an appropriate representation of objects must be chosen. Typical choices are bounding boxes or other approximating shapes. As the object localization step might lead to ambiguous matchings, such as two predictions being assigned to the same reference object, an \textbf{assignment strategy} needs to be picked as well. Overall, predictions in object detection have to fulfill the following requirements to be labeled \ac{TP}: Firstly, the localization criterion must be fulfilled (spatial correspondence). Secondly, the predicted class must match the class of the reference object (always given in the binary case). Finally, the assignment strategy must yield a matching reference object for the given prediction. The recommended localization criteria are provided in Suppl. Note~\ref{ssec:cheat-sheets-localization}. 
\newpage
\begin{figure}[H]
    \centering
    \includegraphics[width=\textwidth]{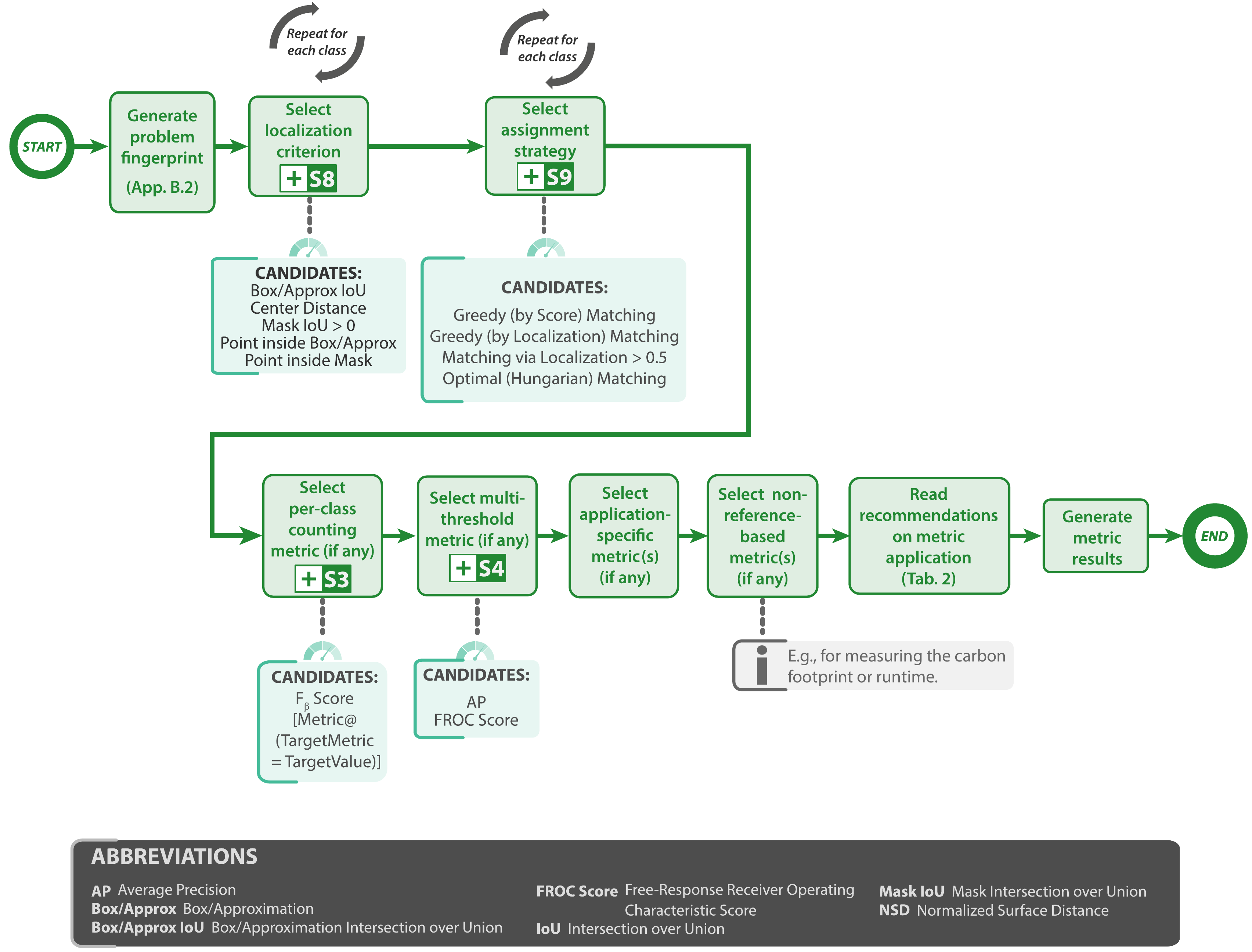}
    \caption{\textit{Metrics Reloaded} recommendation framework for object detection at a glance.}
    \label{fig:reduced-metamapping-od}
\end{figure}

\begin{figure}[H]
    \centering
    \includegraphics[width=0.9\linewidth]{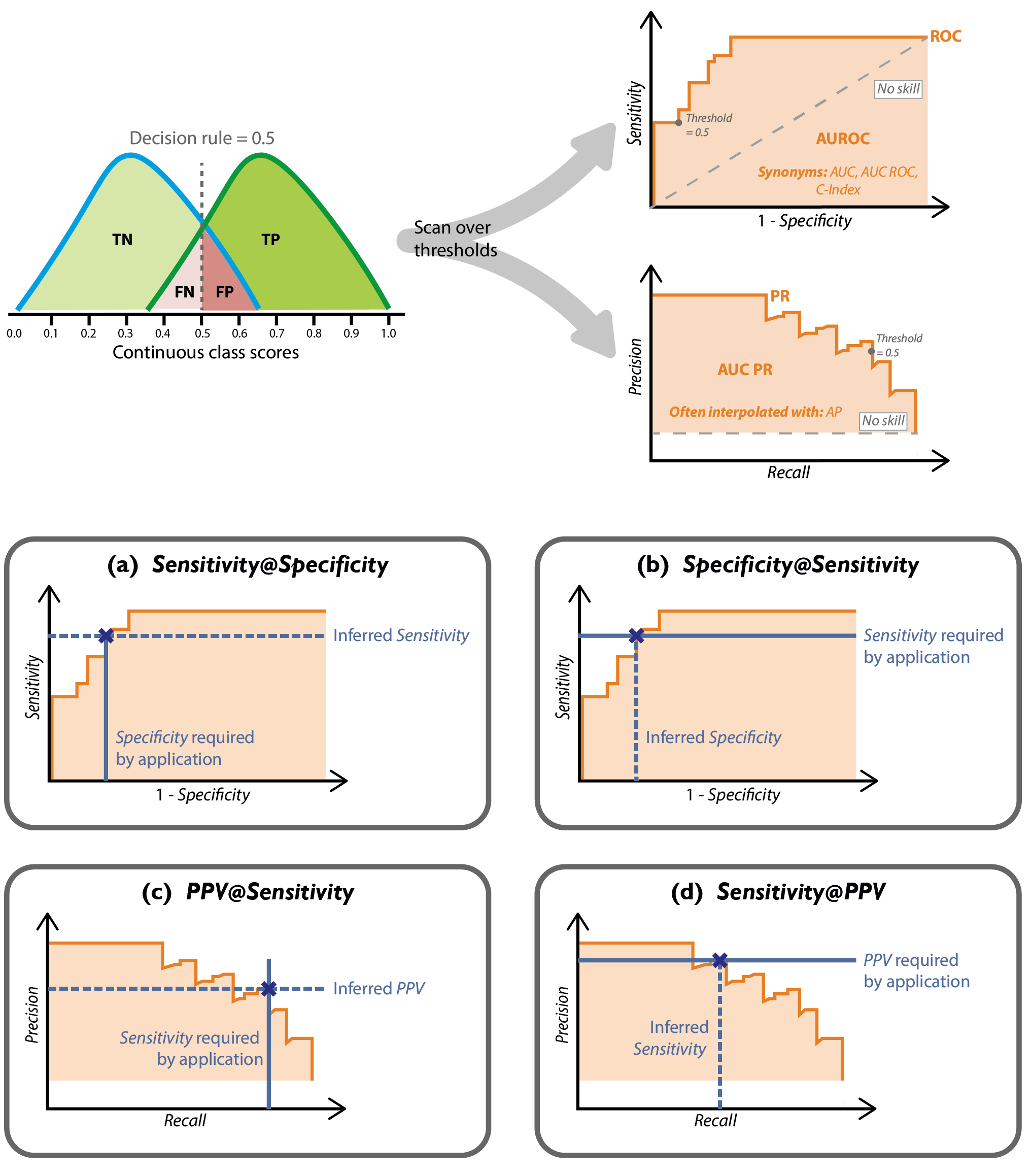}
    \caption{Principle of multi-threshold metrics (top) and per-class counting metrics with application-driven thresholds (bottom). Rather than being based on a static threshold (e.g., for generating the confusion matrix), multi-threshold-based metrics integrate over a range of thresholds. Prominent examples are the \acf{AUROC} (also known as \acf{AUC} or AUC \acf{ROC}) and the Area under the \acf{PR} curve (AUC PR). Cardinalities, i.e., the true (T)/false (F) positives (P)/negatives (N), are computed based on a threshold (e.g., 0.5) of predicted class probabilities (left). Based on those values, Sensitivity (also known as Recall) and 1 - Specificity/\acf{PPV} are calculated and plotted against each other (right). The procedure is repeated for several thresholds, resulting in the \ac{ROC}/\ac{PR} curve. The area under the \ac{ROC}/\ac{PR} curve is referred to as \ac{AUROC}/AUC PR. The latter is often interpolated by the \acf{AP} metric. The dashed gray lines refer to a classifier with no skill level (random guessing). In the case of an application-driven threshold (e.g., required Sensitivity of 0.95), the metrics Sensitivity@Specificity, Specificity@Sensitivity, \ac{PPV}@Sensitivity and Sensitivity@\ac{PPV} can be calculated on the basis of the \ac{ROC}/\ac{PR} curves. Please note that we use the synonyms Precision instead of \ac{PPV} and Recall instead of Sensitivity for the \ac{PR} curve, given their common use.}
    \label{fig:def-auc}
\end{figure}

Based on the choice of localization criterion and assignment strategy, standard classification metrics can be computed on object level. Importantly from a mathematical perspective in this context, the absence of \acp{TN} in object detection problems renders many popular classification metrics (e.g., Accuracy, Specificity, \ac{AUROC}) invalid. Based on these considerations and taking into account all the complementary strengths and weaknesses of existing metrics~\cite{reinke2021common}, we propose the following steps for object detection problems (green path in Fig.~\ref{fig:meta-mapping} and ~\ref{fig:reduced-metamapping-od}):  

\begin{description}
    \item[1: Select localization criterion:] The selection of the localization criterion should be performed according to Subprocess S8 (Extended Data Fig.~\ref{fig:metric-selection:s8-localization-criterion}). If a rough outline of objects is desired, rather than just obtaining the object position (FP2.4 Desired granularity of localization, see Suppl. Note~\ref{app:fingerprints} for details), our recommendation is the \textit{Box/Approximation \ac{IoU}} (Fig.~\ref{fig:cheat-sheet-iou-localization-crit}). If only the position of objects is relevant from a domain interest (e.g. for determining the location of cells), the \textit{Center Distance} (Fig.~\ref{fig:center-dist}) is often an attractive option, although \textit{Mask \ac{IoU} > 0} (Fig.~\ref{fig:cheat-sheet-iou-localization-crit-0}) or \textit{Point inside Mask/Box/Approximation} (Fig.~\ref{fig:cheat-sheet-point-inside}) are viable alternatives in case of fine-granular reference annotations (FP4.4 Granularity of provided references). 
   \item[2: Select assignment strategy:] The recommendations for the assignment strategy are provided in Subprocess S9 (Extended Data Fig.~\ref{fig:metric-selection:s9-assignment-strategy}). In case of the availability of predicted class scores (FP5.1 = TRUE) \textit{Greedy (by Score) Matching} (Fig.~\ref{fig:cheat-sheet-greedy-score}) is our default recommendation. Otherwise, \textit{Greedy (by Localization criterion) Matching} (Fig.~\ref{fig:cheat-sheet-greedy-localization}), \textit{Optimal (Hungarian) Matching} (Fig.~\ref{fig:cheat-sheet-hungarian}) or \textit{Matching via Overlap > 0.5} (Fig.~\ref{fig:cheat-sheet-matching-greater}) are viable options, as detailed in decision guide 9.1 in Suppl. Note~\ref{ssec:dg9}. The user must also decide whether double assignments should be punished (FP2.5.8). 
    \item[3: Select classification metric(s) (if any):] Once objects have been located and assigned to reference objects, generation of a confusion matrix (without \ac{TN}) is possible. The final step therefore simply comprises choosing suitable classification metrics.  
    \begin{description}
    \item[a: Select counting metric (if any):] The selection of a per-class counting metric according to Subprocess S3 (Extended Data Fig.~\ref{fig:metric-selection:s3-single-threshold}) is governed by the decision rule (FP2.6). If a target value for a specific target metric is provided (e.g. Sensitivity = 0.95; Fig.~\ref{fig:cheat-sheet-sensitivity}), complementary metrics such as \ac{PPV} (Fig.~\ref{fig:cheat-sheet-ppv}) can be assessed at the provided point on the decision curve (Fig.~\ref{fig:def-auc}). Otherwise, we recommend the F$_\beta$ Score (Fig.~\ref{fig:cheat-sheet-fbeta}) as a counting metric.  
    \item[b: Select multi-threshold metric:]
   Several subfields of biomedical image analysis have converged to choosing solely a counting metric as the primary metric. This choice seems to be a historical artifact from when algorithms did not provide predicted class scores. We generally recommend not discarding the scores typicalls provided by current algorithms and disagree with the practice of basing performance assessment solely on a single, potentially suboptimal, decision rule applied to the predicted class scores. Instead, we primarily propose selecting a multi-threshold metric (Subprocess S4, Extended Data Fig.~\ref{fig:metric-selection:s4-multi-threshold}) to present a more holistic picture of performance. As multi-threshold metric, we recommend \ac{AP} (Fig.~\ref{fig:cheat-sheet-ap}) or \ac{FROC} Score (Fig.~\ref{fig:cheat-sheet-froc}), depending on whether an easy interpretation (\ac{FROC} Score) or a standardized metric (\ac{AP}) is preferred (see decision guide 4.2 in Suppl. Note~\ref{ssec:dg4}). 
   \end{description}
\end{description}

Note that the previous description implicitly assumed single-class problems, but generalization to multi-class problems is straightforward by applying the validation per class. 
\newpage

\begin{figure}[H]
    \centering
    \includegraphics[width=1\linewidth]{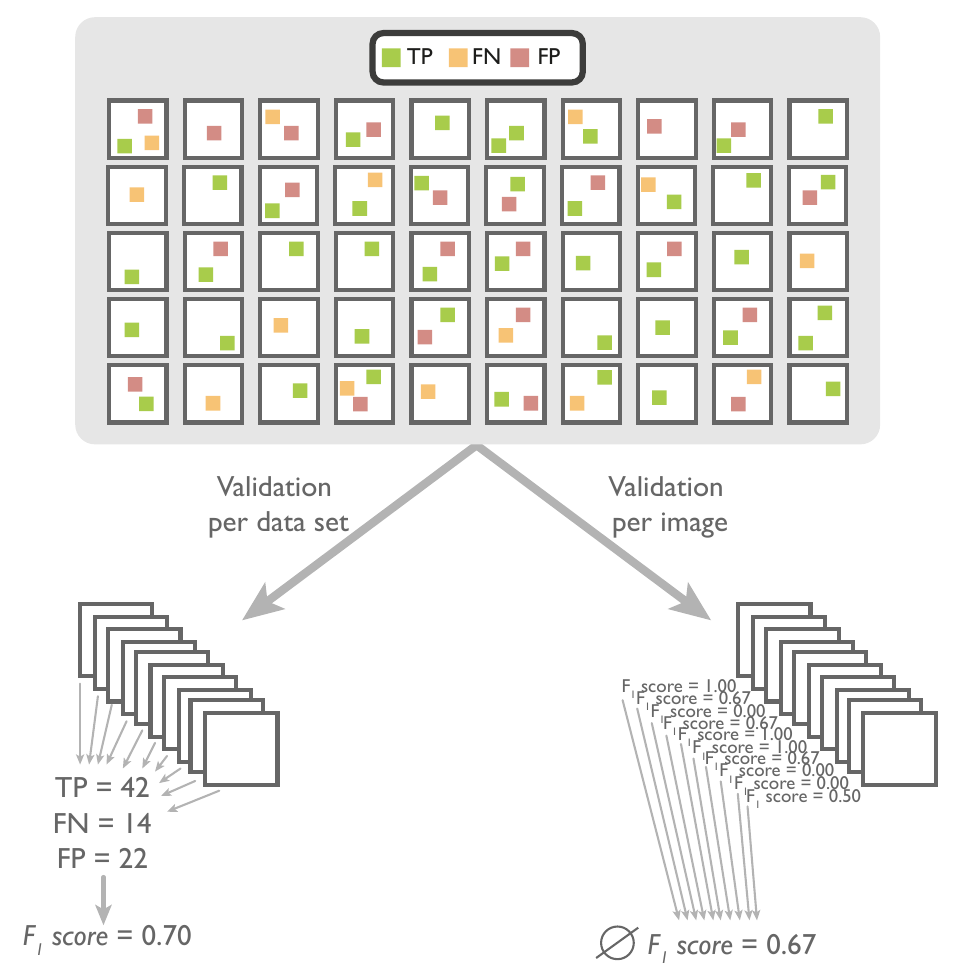}
    \caption{Validation on object level can be performed per data set (left) or per image (right). For the per-data set validation of objects, the cardinalities are calculated over the whole data set. For the per-image validation of objects, metric scores are computed per image and aggregated afterward. $\varnothing$ refers to the average F$_1$ Score.}
    \label{fig:perimage-perdataset} 
\end{figure}
\newpage

\begin{figure}[H]
    \centering
    \includegraphics[width=0.7\linewidth]{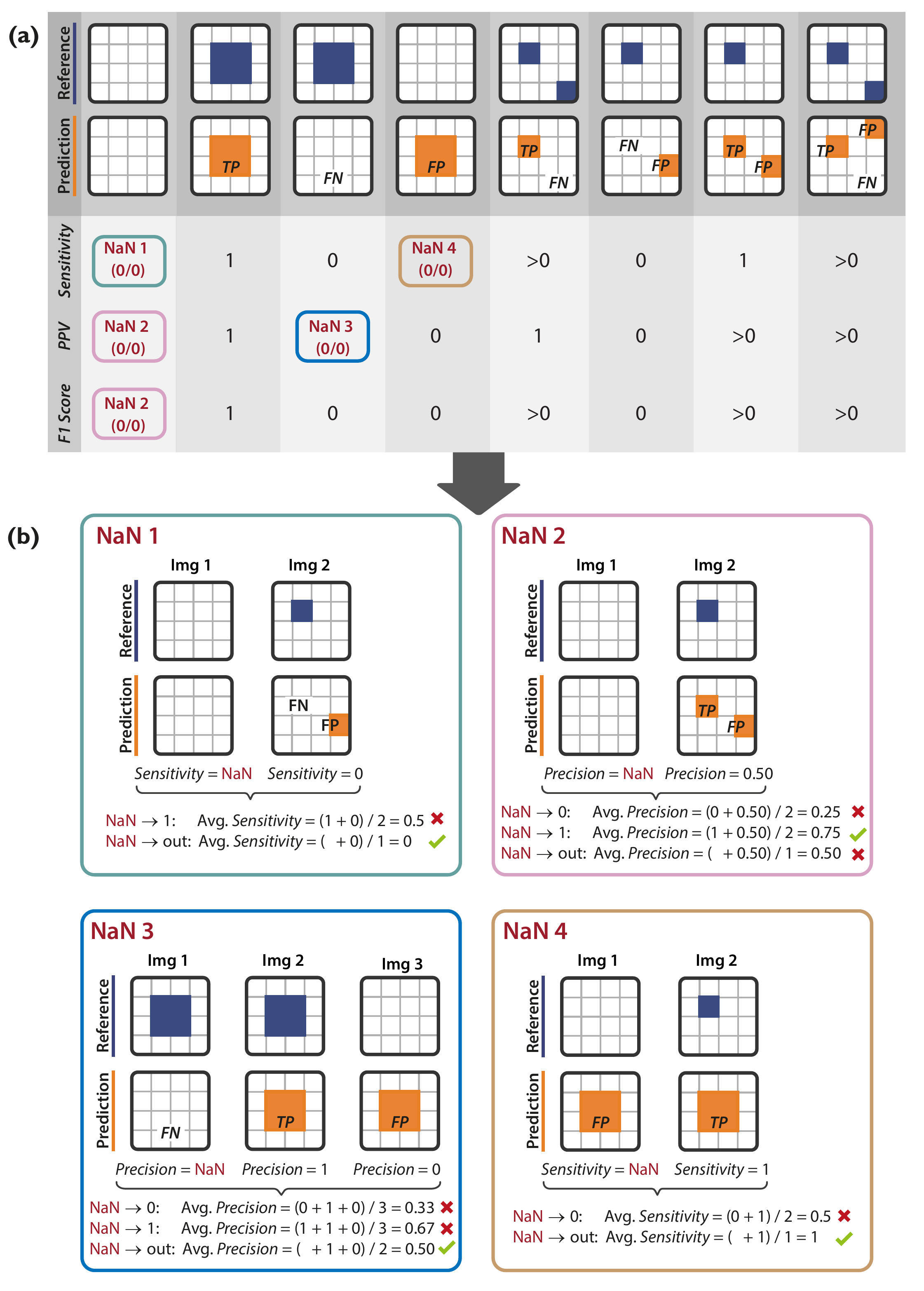}
    \caption{Effect of handling a \acf{NaN} occurring during metric computation, when object detection/instance segmentation tasks are validated per image. Specifically, \ac{NaN} cases occur when an image features no target structures and/or no object predictions by the model, which causes division by zero errors in prevalent metrics. \textbf{(a)} Demonstration of how and when \ac{NaN} can occur. Each column represents a potential scenario for per-image validation of objects, categorized by whether \acp{TP}, \acp{FN}, and \acp{FP} are present (n > 0) or not present (n = 0) after matching/assignment. The sketches on the top showcase each scenario when setting "n > 0" to "n = 1". For each scenario, Sensitivity, \acf{PPV}, and F$_\text{1}$ Score are calculated. \textbf{(b)} Effect of different \ac{NaN} handling strategies based on different conventions for the aggregation across multiple images. Four examples are shown for the \ac{NaN} scenarios from (a) (\textbf{\ac{NaN} 1-4}). \textbf{\ac{NaN} 1} and \textbf{4}: The intuitive penalization for \acp{FP} in "empty" images is already established by means of \ac{PPV} scores (see \textbf{\ac{NaN} 4}) and further penalization by means of Sensitivity is neither required nor appropriate. Instead, images without reference objects should be ignored when averaging Sensitivity scores over images. \textbf{\ac{NaN} 2}: The intuitive penalization for \ac{FP} in "empty" images is established when assigning a \ac{PPV} (and F$_\text{1}$ Score) of 1. \textbf{\ac{NaN} 3}: The intuitive penalization for \ac{FP} is established when removing images with \ac{FN} and no \ac{FP} from the aggregation of \ac{PPV} (and F$_\text{1}$) scores.}
     \label{fig:empty-ref-pred-od}
\end{figure}

It is further worth mentioning that metric \textit{application} is not straightforward in object detection problems. One example is the fact that the number of objects in an image may be extremely small (even zero) compared to the number of pixels in an image. Special considerations with respect to \textbf{aggregation strategy} must therefore be made (Fig.~\ref{fig:perimage-perdataset}). 
In fact, in the machine learning community, object detection tasks are typically validated by pooling all matched objects (i.e, \ac{TP}, \ac{FP}, and \ac{FN}) over the entire data set and computing global metrics on the entire pool ('per-data set aggregation'). An alternative strategy is the 'per-image aggregation', where matched objects are aggregated per individual image to compute corresponding metrics (e.g., F$_\beta$ Score). The per-image metrics are subsequently averaged over the data set. This alternative aggregation may be desirable for two reasons. Firstly, due to the hierarchical data structure (potentially multiple objects per image and/or multiple images per patient), a hierarchical aggregation of metric values, which compensates for the non-independence of images, is generally recommended. Secondly, from a domain interest, the expected metric value per image (rather than per entire data set) may be desirable.
%because individual errors (\ac{FP}, and \ac{FN}) are penalized depending on how much they affect the outcome per image (or diagnosis per patient in a clinical scenario). 
Importantly, the per-image aggregation strategy also changes the way multi-threshold metrics such as \ac{AP} are computed: While the thresholds are still scanned over the scale of predicted class scores simultaneously for the entire data set, the precision and recall used to generate the \ac{PR}-curve are now per-image scores averaged over the data set rather than the global per-data set scores. It should further be noted that validating an object-level problem per image rather than per data set comes with the problem that images containing no reference or prediction objects lead to division by zero for some metrics and thus to \ac{NaN} as metric output. We therefore propose strategies for \ac{NaN} handling in Fig.~\ref{fig:empty-ref-pred-od}a. In summary, we recommend to exclude \ac{NaN} cases from metric computation except when an empty prediction corresponds to an empty reference, in which case \ac{PPV}, and in extension F$_\beta$ Score, should be set to 1.

A further critical consideration for metric application in object detection is the fact that structure sizes may have a large effect on performance metrics~\cite{Tran2022}. We therefore recommend size stratification, i.e., the separate validation for different size ranges, if size variability is high (FP3.2 = TRUE).

\newpage

%------------------------------------
\subsection{\textcolor{ISpink}{\textbf{Recommendations for Instance segmentation}}}
\label{app:is}
%------------------------------------
\acresetall
\begin{tcolorbox}[title=Essentials]
    \textbf{FPX.Y} refers to a fingerprint item detailed in Figs.~\ref{fig:fingerprint-is-1}-\ref{fig:fingerprint-is-3}. \\

    \textbf{SX} refers to a subprocess in Extended Data Figs.~\ref{fig:metric-selection:s3-single-threshold}-\ref{fig:metric-selection:s4-multi-threshold} and Extended Data Figs.~\ref{fig:metric-selection:s6-overlap}-\ref{fig:metric-selection:s9-assignment-strategy}. \\

    \textbf{DGX.Y} refers to a decision guide in Suppl. Notes~\ref{ssec:dg3}-\ref{ssec:dg9}. \\

    Segmentation metrics are to be applied per instance.
\end{tcolorbox}

This section provides recommendations for selecting \textit{common reference-based metrics} for instance segmentation problems. As depicted in Fig.~\ref{fig:meta-mapping}, these common metrics can then be complemented by application-specific metrics as well as non-reference-based metrics (assessing run time or carbon footprint, for example).

Instance segmentation can be regarded as delivering the tasks of object detection and semantic segmentation at the same time. In contrast to object detection, instance segmentation also involves the accurate marking of the object boundary. In contrast to semantic segmentation, it distinguishes different instances of the same class. The pitfalls and recommendations for instance segmentation problems are closely related to those for segmentation and object detection~\cite{reinke2021common} and we recommend reading Suppl. Note~\ref{app:od} and Suppl. Note~\ref{app:ss} as a foundation for this section. 

Our recommendations for assessing instance segmentation quality can be summarized as follows (purple path in Fig.~\ref{fig:meta-mapping} and Fig.~\ref{fig:reduced-metamapping-is}):
\newpage
\begin{figure}[H]
    \centering
    \includegraphics[width=1\textwidth]{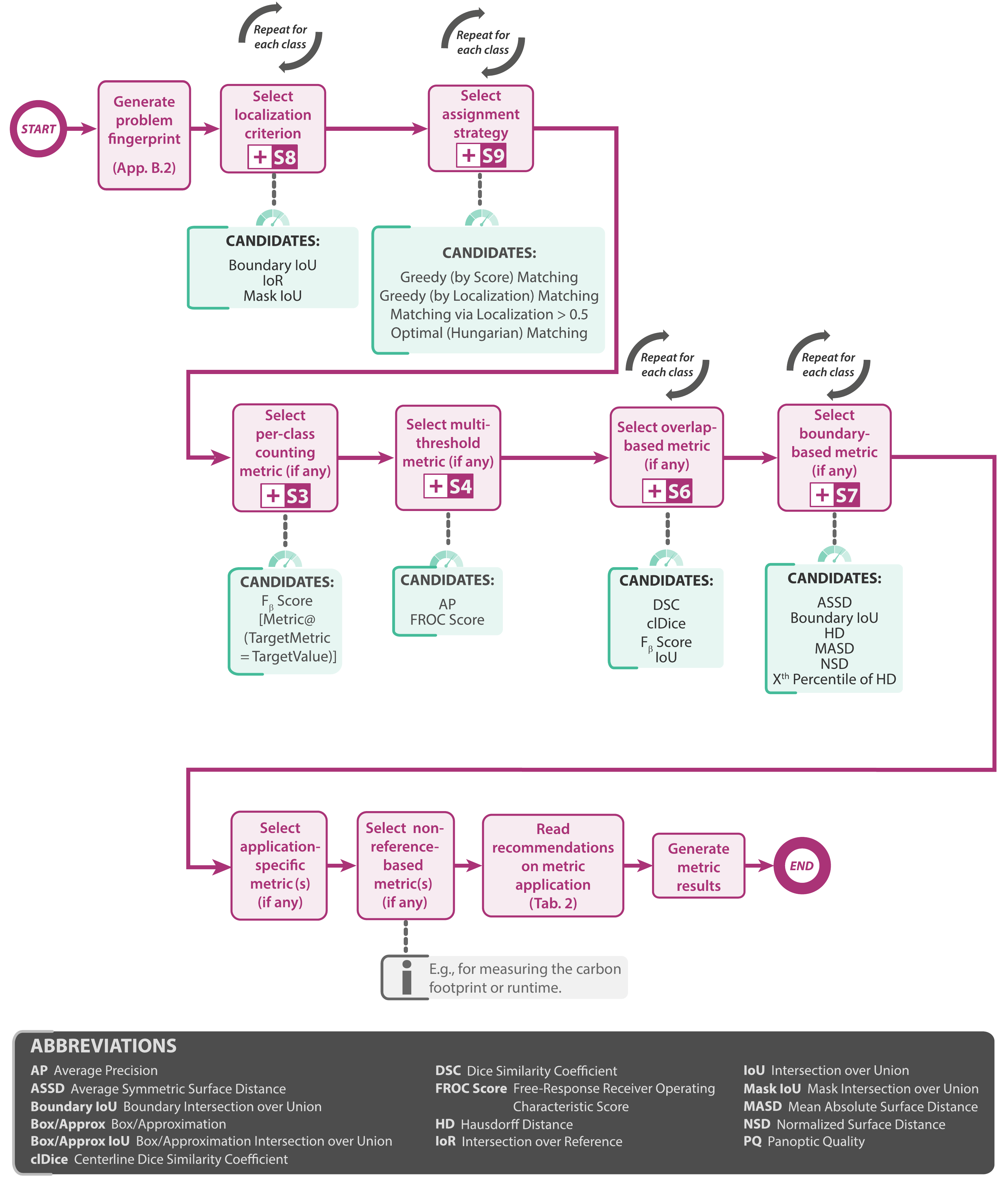}
    \caption{\textit{Metrics Reloaded} recommendation framework for instance segmentation at a glance.}
    \label{fig:reduced-metamapping-is}
\end{figure}

\begin{figure}[H]
    \centering
    \hspace{-0.5cm}
    \includegraphics[width=0.9\linewidth]{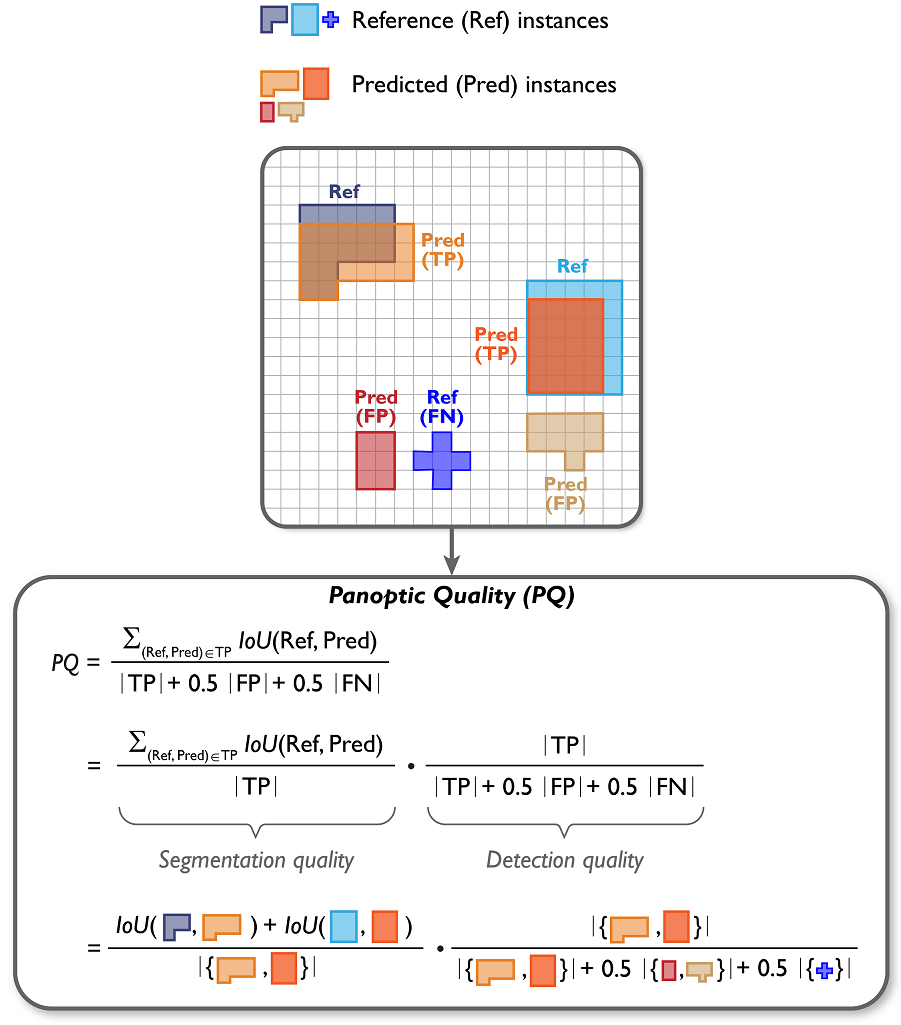}
    \caption{The \acf{PQ} measures the \textbf{segmentation and detection quality} of a prediction in one score. The metric simply averages the \ac{IoU} scores for all \acf{TP} instances and multiplies the result with the F$_1$ score. For perfect segmentation results, i.e., an average \ac{IoU} of 1, the \ac{PQ} would equal the F$_1$ Score.}
    \label{fig:definition-pq}
\end{figure}

\begin{description}
    \item[1: Select object detection metric(s):] From a semantic segmentation perspective, overcoming problems related to instance unawareness (\ref{fig:metrics_matter_overview}a (top left)) requires the selection of a set of detection metrics to explicitly measure detection performance. To this end, we follow the same general process as in the object detection recommendation by selecting a localization criterion, an assignment strategy, and suitable classification metrics. The specific recommendations for instance segmentation are: 
     \begin{description}
    \item[a: Select localization criterion:] Although not common in practice, we argue that for consistency it might be appropriate to base the localization criterion on the corresponding target segmentation metric (see step 2: "Select segmentation metric(s) (if any)" below). For example, if the target metric is \ac{NSD}, the localization criterion could be defined accordingly. This may not always be possible, for instance because the target metric has no fixed upper bound (e.g., \ac{HD}), rendering the setting of adequate cutoffs challenging. As an alternative strategy, we therefore recommend choosing the localization criterion according to common practice (see Subprocess S8, Extended Data Fig.~\ref{fig:metric-selection:s8-localization-criterion}). For this strategy, given the fine granularity of both the output and the reference annotation, we recommend selecting between Boundary \ac{IoU} (Fig.~\ref{fig:cheat-sheet-boundary-iou-localization-crit}), Mask \ac{IoU} (Fig.~\ref{fig:cheat-sheet-iou-localization-crit}), and \ac{IoR}, (Fig.~\ref{fig:cheat-sheet-ior}) using decision guide 8.1 in Suppl. Note~\ref{ssec:dg8}.
    \item[b: Select assignment strategy:] The recommendations for the assignment strategy are identical to those for object detection (Extended Data Fig.~\ref{fig:metric-selection:s9-assignment-strategy}). In case of the availability of predicted class scores (FP5.1 = TRUE) Greedy (by Score) Matching is our default recommendation. Otherwise, Greedy (by Localization criterion) Matching, Optimal (Hungarian) Matching. or Matching via Overlap > 0.5 are viable options, as detailed in decision guide 9.1 in Suppl. Note~\ref{ssec:dg9}. The user must also decide whether double assignments should be punished (FP2.5.8). 
    \item [c: Select classification metrics:] 
    Our recommendations with respect to classification metrics are identical to those for object detection (Suppl. Note~\ref{app:od}) with a single exception. As depicted in S3, Extended Data Fig.~\ref{fig:metric-selection:s3-single-threshold}, we recommend the \ac{PQ} (Fig.~\ref{fig:cheat-sheet-pq}) \cite{kirillov2019panoptic} as an alternative to the F$_\beta$ Score (Fig.~\ref{fig:cheat-sheet-fbeta}). As illustrated in Fig.~\ref{fig:definition-pq}, this metric is especially suited for instance segmentation, as it combines the assessment of overall detection performance and segmentation quality of successfully matched (\ac{TP}) instances in a single score. 
    \end{description}
    \item[2: Select segmentation metric(s) (if any):] In a second step, metrics for explicit assessment of the segmentation quality for the \ac{TP} instances, i.e., successfully matched instances, may be selected. Here, we follow the exact same process as in semantic segmentation (Subprocesses S6, Extended Data Fig.~\ref{fig:metric-selection:s6-overlap} and S7, Extended Data Fig.~\ref{fig:metric-selection:s7-boundary}). The primary difference is that the segmentation metrics are computed per-instance and subsequently averaged resulting, for example, in a '\ac{DSC} per instance' score.
\end{description}

While we have found our recommendations for instance segmentation to match the majority of biomedical problems,  standard reference-based metrics are not well-suited for some applications. Specifically, standard metrics struggle in images with structures of extreme density and complex shapes, because overlap often fails as a criterion to establish unique correspondences between predicted and reference instances. In such cases, specialized metrics not relying on one-to-one correspondences may be required, such as pair-counting metrics or information theoretic-based metrics~\cite{taha2015metrics}. Another example that calls for application-specific metrics is cell nucleus segmentation, where splitting a reference object by two separate predictions is assessed by a dedicated 'split error', and the converse by a dedicated "merge error'~\cite{caicedo2019evaluation}. These application-specific errors can either be used as stand-alone metrics or integrated into compound metrics such as F$_1$ Score. 

Recommendations for aggregating object detection and instance segmentation metrics are provided in the respective appendices Suppl. Note~\ref{app:od} and Suppl. Note~\ref{app:is}, respectively.
\newpage

\newpage

%\newpage
\subsection{Recommendations for Calibration of Predicted Class Scores}
\label{app:calibration}
\acresetall

While most research in biomedical image analysis focuses on the discrimination capabilities of classifiers, a complementary property of relevance is the \textit{calibration} of predicted class scores.
Importantly, a large portion of the research in this field is comparatively young, and a variety of new calibration metrics are proposed every year. As it might be premature to call for rigid standardization in such a vibrant environment, the following recommendations are to be seen as general guidance through the current landscape of calibration metrics, which might be subject to updates in the following years.

Intuitively speaking, a system is well-calibrated if the predicted class scores (i.e., the output of the model) reflect the true probabilities of the outcome. In practice, this means that calibrated scores match the empirical success rate of associated predictions. For example, in a binary classification task, calibration implies that of all the data samples assigned a predicted score of $0.8$ for the positive class, empirically, $80\%$ belong to this class.

One common but critically important misconception about calibration is that the predicted class scores of a well-calibrated model express the true posterior probability $\mathbb{P}_{Y \mid X}$ of an input belonging to a certain class~\cite{perez2022beyond}, e.g., that they express a patient's risk for a certain condition based on an image. While this probability is commonly of interest in classification problems, common calibration metrics instead typically consider $\mathbb{P}_{Y \mid f(X)}$, i.e., the probability of a model's output score belonging to a certain class. Conditioning on the output entirely ignores the mapping $f \colon \mathcal{X} \to \mathcal{P}$. Thus, while calibration allows making statements about the empirical class membership of predicted scores, such as in the example above, these statements are conditioned on the discrimination power of a model. This means that different models may predict different probabilities for the same input even though all of them are perfectly calibrated~\cite{perez2022beyond}. Going back to the clinical example, this implies that a classifier that always predicts the score $0.5$ is considered perfectly calibrated on a balanced binary task, although another perfectly calibrated model with better discrimination ability could output completely different, practically more meaningful scores. Again, this discrepancy occurs because calibrated scores reflect the empirical success rate of predictions and not a patient-specific (model-agnostic) inherent risk. The clinical prediction modelling community therefore traditionally distinguishes different levels of calibration \cite{van2016calibration}, where \textit{level 4 strong calibration} implies correct posteriors ($\mathbb{P}_{Y \mid X}$). As level 4 is practically unfeasible to measure (the true individual posteriors are unknown), common research focuses on  \textit{level 3 moderate calibration}, which implies that the predicted scores match the empirical success rate.

For a more formal definition of (level 3) calibration, let the random variables $X$ and $Y$ correspond to the feature (e.g., an image) and target variables (encoding the outcome), respectively, with feature and target spaces $\mathcal{X}$ and $\mathcal{Y}$. Let $f \colon \mathcal{X} \to \mathcal{P}$ denote a classifier with predicted class scores $f \left( X \right)$ and $\mathcal{P}$ a set of distributions on $\mathcal Y$.
We further use the notation $\mathbb{P}_Y, \mathbb{P}_{Y \mid f(X)} \in \mathcal{P}$, where $\mathbb{P}_Y$ refers to the distribution of $Y$, and $\mathbb{P}_{Y \mid f(X)}$ to the conditional distribution of $Y$ given $f(X)$. 

\begin{figure}[H]
    \centering
    \includegraphics[width=0.7\linewidth]{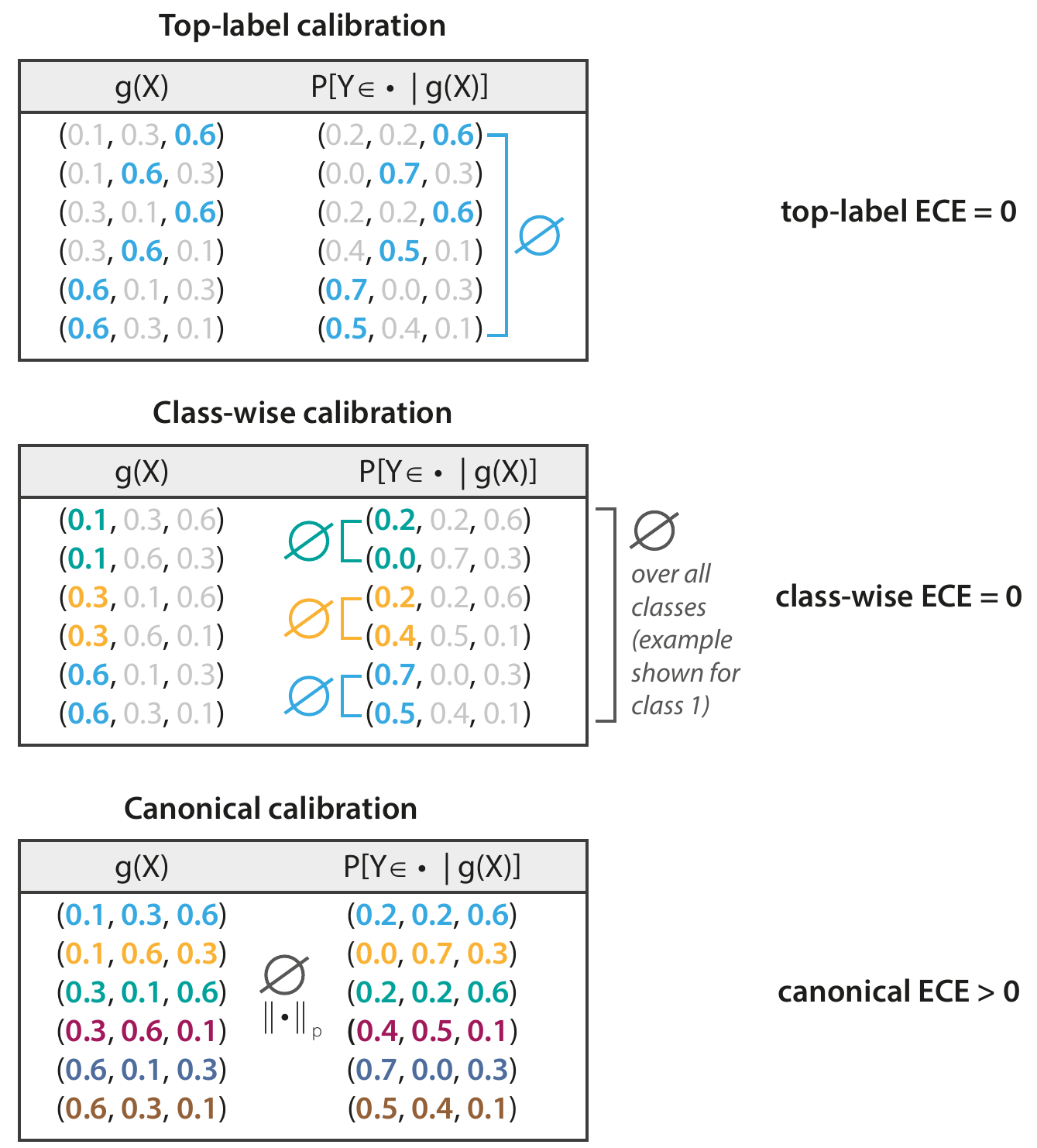}
    \caption{Estimating the \acf{CE} according to the three different conditions in multi-class settings yields inconsistent results. For the top-label calibration, only the maximum values of the predicted class scores $f \left( X \right)$ are considered, while all other values are neglected. For the computation of the \ac{CE}, for each distinct output value of $f \left( X \right)$ (only 0.6 in this case), $\mathbb{P}_{Y \mid f \left( X \right)}$ is determined as the average over the empirical rates of this output (0.6, 0.7, 0.6, 0.5, 0.7, 0.5 in this case). The top-label calibration condition (i.e., matching the two scores) results in a perfect \ac{CE} $ = 0$ in this scenario. Similarly, for the class-wise calibration, the predicted class scores are compared per class, a requirement that is also fulfilled by the depicted system. Only the canonical calibration, which comes with the strict requirement that the model output must match the full probability distribution (implying the comparison of entire vectors rather than single values) indicates a miscalibrated system ($CE > 0$). This figure is inspired by ~\cite{vaicenavicius2019evaluating}.} 
    \label{fig:ece-levels}
\end{figure}

In practice, three different variations of calibration conditions can be distinguished~\cite{vaicenavicius2019evaluating}:
\begin{itemize}
    
\item \textbf{Canonical calibration:} $f \left( X \right) = \mathbb{P}_{Y \mid f \left( X \right)}$. This condition implies pairwise matching of all entries across the two distributions (see also the top panel in Fig.~\ref{fig:ece-levels}). Although not the most commonly applied condition in practice, a common perception is that this condition is the appropriate perspective on calibration in many application scenarios as the weaker conditions (see below) are prone to underestimating miscalibration~\cite{gruber2022better,popordanoska2022consistent,ferrer2022analysis}.

\item \textbf{Class-wise calibration:} $f_k \left( X \right) = \mathbb{P} \left(Y = k \mid f_k \left( X \right) \right)$ for all classes $k \in \mathcal{Y}$. This is a weaker condition, where not the joint, but the marginal distributions for each class are required to match (see also the middle panel in Fig.~\ref{fig:ece-levels}). Assessing the calibration quality for individual classes provides crucial information, for example in scenarios where there is a mismatch between class prevalences and class importance (FP2.5.3=TRUE).

\item \textbf{Top-label calibration:} $f_K \left( X \right) = \mathbb{P} \left(Y = K \mid f_K \left( X \right) \right)$, where $K = \arg\max_k f_k \left( X \right)$ of a model $f \colon \mathcal{X} \to \mathcal{P}$. This is the weakest of the three conditions, where only the maximum entry (top label) of each predicted class vector is considered (see also the bottom panel in Fig.~\ref{fig:ece-levels}). This condition assesses only the highest class score, which is often used to determine the predicted class, and thus implies a strong focus on validating the reliability of a model’s decisions.

\end{itemize}

While these three conditions are equivalent for binary classification problems, they may differ substantially in the broader multi-class setting, as illustrated in Fig.~\ref{fig:ece-levels}.

In practice, no model is perfectly calibrated. Calibration quality  is captured by the \acf{CE}, which can be computed via a divergence, i.e., a distance function, between predictions $f \left( X \right)$ and either of the three conditions (canonical, class-wise, top-label). For instance, typical choices for quantifying the canonical \ac{CE} are expected $L_1$ or $L_2$ errors~\cite{gruber2022better,kumar2019verified,naeini2015obtaining}. These can be further generalized to the $L_p$ \ac{CE}: For $1 \leq p \in \mathbb{R}$, the canonical \textbf{$\ell_p$ \ac{CE}} (CE$_p$) of model $f \colon \mathcal{X} \to \mathcal{P}$ is defined as:
\begin{equation}
    \text{CE}_p \left( f \right) = \left( \mathbb{E} \left[ \left\| f \left( X \right) - \mathbb{P}_{Y \mid f \left( X \right)} \right\|^p_p \right] \right)^{\frac{1}{p}}.
\label{def:lp_ce}
\end{equation}

The relations of \ac{CE} variants associated with the three conditions above intuitively translate to $CE_{canonical} \geq CE_{class-wise} \geq CE_{top-label}$. In the example provided in Fig.~\ref{fig:ece-levels}, the weaker conditions of top-label calibration and class-wise calibration are fulfilled (associated errors are zero), while the broader canonical condition for calibration is not met. The fact that the calibration quality of a classifier varies when assessed through the lens of different conditions causes common calibration measures to be characterized as \textit{inconsistent} in multi-class settings~\cite{gruber2022better}. 

The canonical $\ell_p$ \ac{CE} can be generalized by replacing the $\ell_p$ norm as a distance measure between $f \left( X \right)$ and $\mathbb{P}_{Y \mid f \left( X \right)}$  with alternative distance functions. For example, \cite{widmann2019calibration} introduced a canonical \ac{CE} based on matrix-valued kernels. 

\newpage

Generally, measuring the \ac{CE} is challenging, because $\mathbb{P}_{Y \mid f \left( X \right)}$ is unknown and needs to be estimated as the expected value on the available data. Returning to the simple example above, as we only have access to a small subset of all potential cases for which the model would predict a score of $0.8$, we do not know whether the corresponding success rate of these cases is $80\%$ in general; instead, our assessment relies on the estimated success rate based on the available samples. The fact that classifier outputs are generally continuous often reduces the number of available samples per prediction to one. Strategies for alleviating the sparse sampling problem include binning the continuous scale of $f \left( X \right)$ and estimating the \ac{CE} per bin (such as done for \ac{ECE} (Fig.~\ref{fig:cheat-sheet-ece}) and \ac{CWCE} (Fig.~\ref{fig:cheat-sheet-cwce}), as illustrated in Fig.~\ref{fig:ece-illustration}), or using kernel density estimation methods (such as done for \ac{ECEKDE}, see Fig.~\ref{fig:cheat-sheet-ecekde}). Despite these efforts, the most popular calibration measures are generally biased estimators of the true error, which means their estimates depend on the number of samples (i.e., size of the validation data set). Gruber et al.~\cite{gruber2022better} recently described this bias and how it leads to substantial under- and over-estimations of the true error. Popordanoska et al.~\cite{popordanoska2022consistent} showed that straightforward estimators of $\ell_p$ calibration based on density estimation (such as done for \ac{ECEKDE}) have a generally lower bias compared to statistical estimators (such as binning) and presented means to additionally de-bias estimators. There are also ongoing efforts investigating canonical \acp{CE} that are not based on the $\ell_p$ norm, such as the \ac{KCE}, where 'maximum mean discrepancy' is used as a distance function instead (see Fig.~\ref{fig:cheat-sheet-ecekde}). These efforts have resulted in fully unbiased estimators, which, however, do not allow for interpretable calibration assessment and further require nontrivial configuration of the kernels and associated hyperparameters. 

An attractive alternative to estimate \acp{CE} are so-called \acfp{PSR} (also referred to as \textit{overall performance measures}~\cite{steyerberg2010assessing}), which measure discrimination and calibration in a single score (e.g. \ac{NLL}, \ac{BS}; Figs.~\ref{fig:cheat-sheet-bs}, \ref{fig:cheat-sheet-nll}). An intuitive example for this metric category is the \ac{BS}: For a model $f \colon \mathcal{X} \to \mathcal{P}$ the \ac{BS} is defined as the expected value of the squared error between predictions and reference values as determined on the validation data:
\begin{equation}
    \text{BS} \left( f \right) = \mathbb{E} \left[ \left\| f \left( X \right) - Y^\prime \right\|^2_2 \right],
\label{def:bs}
\end{equation}
where $Y^\prime$ is the one-hot-encoded version of the reference vector $Y$ \textit{for each individual data sample}. This equation illustrates the difference between overall performance measures and calibration metrics measuring the \ac{CE} in Equation~\ref{def:lp_ce}. While the \ac{CE} measures whether the predicted class scores match the empirical success rate (see also~\ref{fig:ece-levels}), \ac{BS} is defined as an expected value over \textit{every single prediction}, thus posing a stronger requirement on the scores which can be interpreted as assessing the true posterior probabilities or individual risks. In theory, \ac{BS} can be decomposed into explicit terms for discrimination and calibration assessment \cite{degroot1983comparison}. In practice, however, although overall performance measures do not suffer from the sampling problem, they conflate the true \ac{CE} with the discrimination error and can thus not make calibration quality explicit. However, proper scores are still useful for comparative studies. Furthermore, it has been shown that the square root of the \ac{BS}, referred to as the \ac{RBS} (Fig.~\ref{fig:cheat-sheet-rbs}), represents a robust estimator and upper bound of the canonical \ac{CE}~\cite{gruber2022better}. Such a guarantee can be particularly relevant in safety-critical scenarios where the error must not be underestimated.

\begin{figure}[H]
    \centering
    \includegraphics[width=1\linewidth]{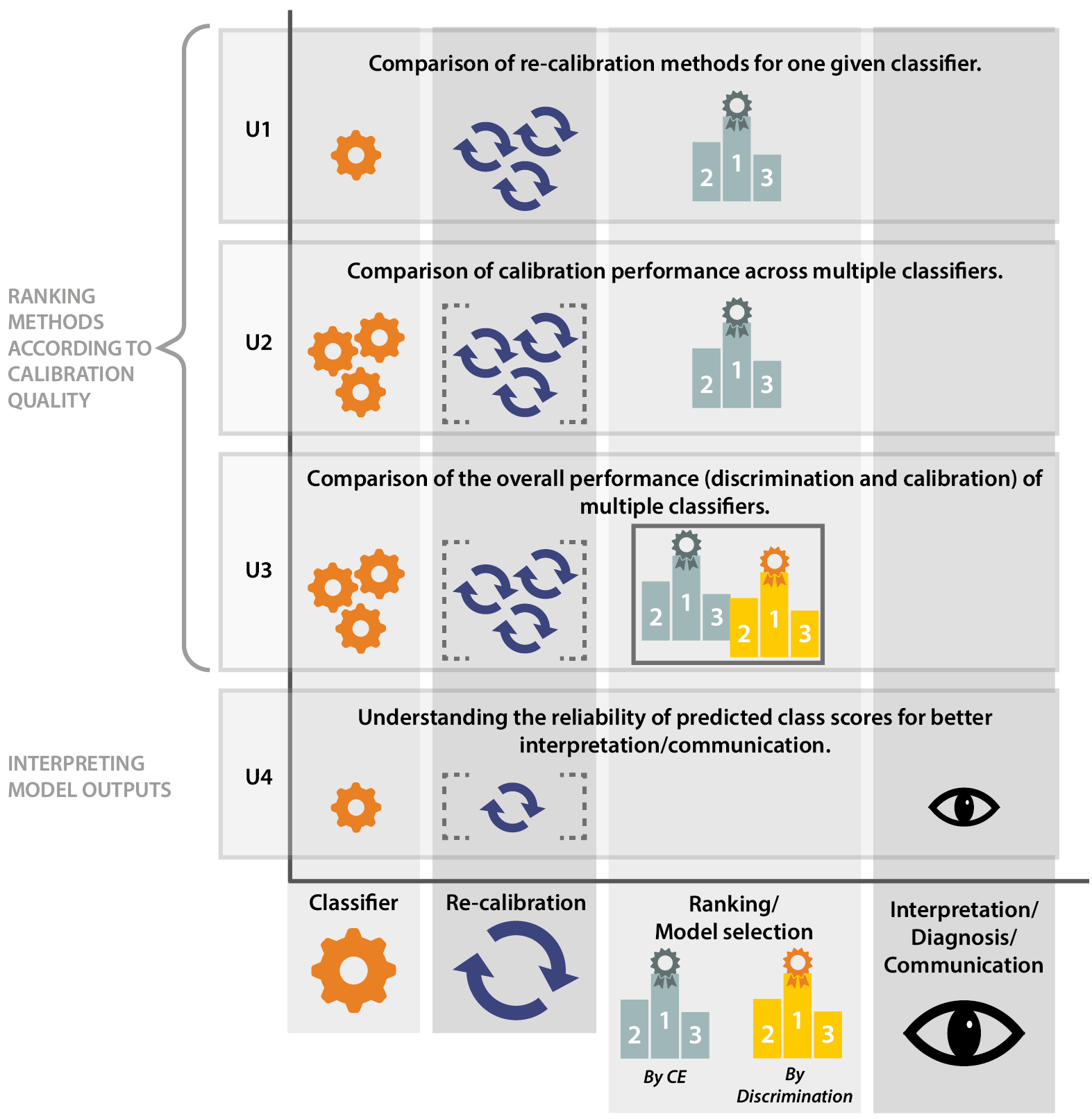}
    \caption{Underlying interest related to the assessment of calibration quality. The user is interested in the comparative calibration assessment (U1-U3) and/or obtaining a reliable estimate of the \acf{CE} for interpreting and communicating the algorithm output (U4). The use cases are detailed in Suppl. Note~\ref{app:calibration}. The brackets around re-calibration methods denote that their application is optional in the corresponding use case.} 
    \label{fig:calibration-comparison-vs-interpretation}
\end{figure}

\begin{figure}
    \centering
    \includegraphics[width=\textwidth]{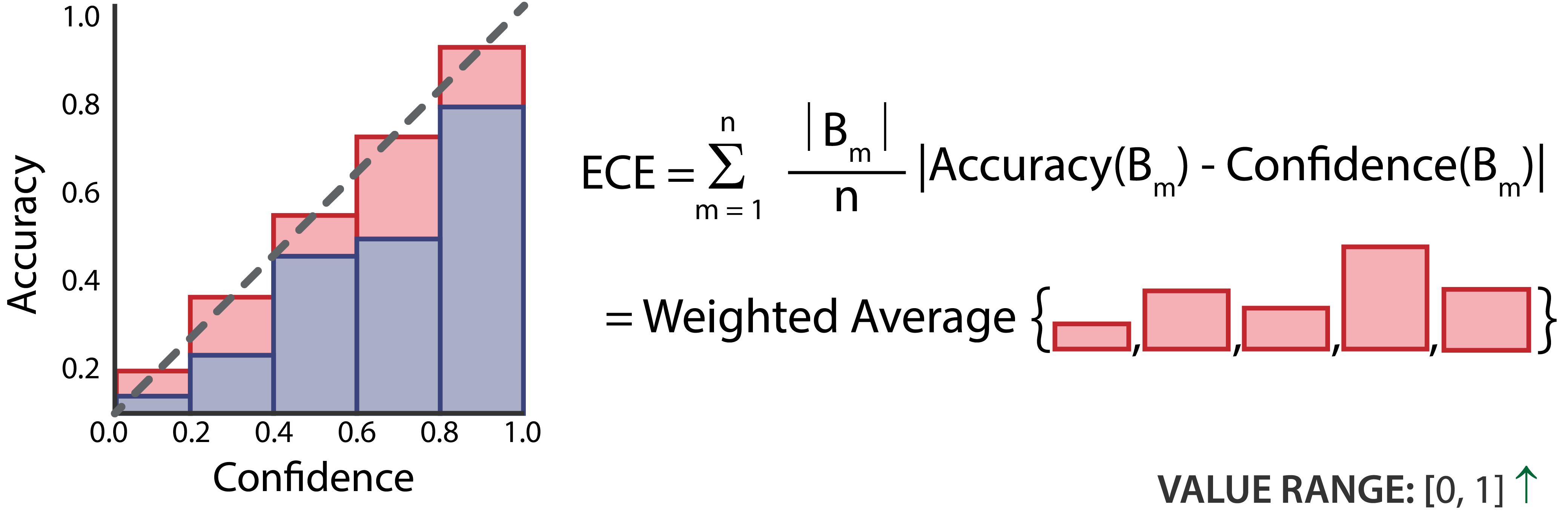}
    \caption{Computation of the \acf{ECE} based on the binning of predicted class scores. The error is based on the discrepancy between the Accuracy per bin $Accuracy(B_m) = 1/\vert B_m \vert \sum_{i \in B_m} \mathbf{1}(\hat{y}_i = y_i)$ and the average over predicted class scores per bin $Confidence(B_m) = 1/\vert B_m \vert \sum_{i \in B_m} \hat{p}_i$. The final \ac{ECE} score is obtained as the average over bin discrepancies weighted by the number of samples $|B_m|$ per bin. Here, $n$ denotes the total number of samples, $\hat{y}_i$denotes the predicted class labels and $y_i$ the true class labels, $\mathbf{1}$ is the indicator function (1 if $\hat{y}_i = y_i$, 0 otherwise), and $\hat{p}_i$ refers to the predicted class scores. The dashed diagonal line acts as a visual reference for a perfectly calibrated system, where discrepancies between per-bin confidences and accuracies are zero.}
    \label{fig:ece-illustration}
\end{figure}

The choice of which calibration condition to validate as well as which metric to use depends on the task interest. Methods subject to validation in this context are either classification models whose inherent calibration quality shall be assessed, or so-called 're-calibration methods', i.e., transformations on the classifier outputs aiming to improve the calibration quality. In the most common scenarios, the driving interest may either be a comparative performance assessment, in which methods are ranked according to calibration quality, or an absolute performance assessment, in which an interpretable and communicable measure of calibration quality is desired. We identified four main use cases (U1-U4) which our framework addresses (Fig.~\ref{fig:calibration-comparison-vs-interpretation}).
\newpage
\newpage
\begin{enumerate}
     \item  \textbf{Ranking methods to determine calibration quality:} The following use cases focus on the comparative assessment of the calibration quality of one or multiple classifiers. 
        \begin{enumerate}
        \item  \textbf{Use case 1 (U1):} comparing the effect of one or more re-calibration methods on the same (fixed) classifier. The desired validation output is a ranking of re-calibration methods (possibly including the performance of 'no re-calibration') from which the best method can be selected.
        \item  \textbf{Use case 2 (U2):} comparing the calibration quality across multiple classifiers on the same task. The desired validation output is a ranking of classifiers according to calibration quality. In practice, such a ranking should be accompanied by a ranking according to discrimination performance, as it is not recommended to base model selection purely on calibration performance.
        \item  \textbf{Use case 3 (U3):} comparing the 'overall performance' of classifiers (optionally including potential re-calibration methods), i.e., a joint assessment of discrimination performance and calibration quality. The desired validation output is a single ranking naturally weighing both aspects.
        \end{enumerate}
     \item  \textbf{Interpreting model outputs:} Complementary interest may lie in the analysis of the \ac{CE} to the end of assessing the reliability of the predicted class scores of one or multiple classifiers.
        \begin{enumerate}
        \item  \textbf{Use case 4 (U4):} interest in understanding the reliability of predicted class scores for a given model as a basis for interpreting and communicating results. The desired validation output is a single score which provides insight into how well the model is calibrated. The reliability of model outputs is often considered crucial upon application, such as for clinical prediction models~\cite{van2019calibration, dolezal2022uncertainty, yang2021multi}. Importantly, U4 can be used in addition to U1, U2 or U3 as it is based on an orthogonal interest. 
        \end{enumerate}
\end{enumerate}

 Because some decision rules assume calibrated model outputs, a further potential interest in calibration validation may lie in determining the quality of a decision rule applied to predicted class scores (see FP2.6), i.e., answering the question: "How much better could the classifier's decisions have been under this rule if predicted class scores were calibrated?".  While such ablations of classifier design decisions are generally out of the scope of our framework, decision rule-related pitfalls and countermeasures are discussed in Sec.~\ref{app:general-instructions}.
 
Based on all of the above considerations, we recommend selecting calibration metrics using Subprocess S5 (Extended Data Fig.~\ref{fig:metric-selection:s5-calibration}) in case the assessment of calibration quality is desired (FP2.7.1~=~TRUE): 

\begin{description}
    \item[1: Select metric for comparative calibration assessment (if any):]  
    This step selects an adequate metric in case a comparative assessment of calibration methods is desired (FP2.7.2). The fingerprint FP2.7.2 covers the presented use cases U1-U3 (Fig.~\ref{fig:calibration-comparison-vs-interpretation}). For \textbf{U1} "Comparison of re-calibration methods for the same fixed classifier", one option is to select a metric that assesses the canonical \ac{CE}, such as \ac{KCE} as an unbiased estimator of a canonical \ac{CE} based on an alternative distance function, or \ac{ECEKDE} as a well-interpretable estimator of canonical $\ell_p$ calibration. Alternatively, an overall performance measure such as the \ac{BS} can be used (see DG5.2), because the classifier is fixed in this scenario, the conflation of the \ac{CE} with discrimination errors is no disturbing factor, and the true \ac{CE} is exposed for relative comparison of scores. For \textbf{U2} "Comparison of calibration quality across classifiers on the same task", we recommend reporting the \ac{CE} per class by using an estimator of marginal \ac{CE}, such as \ac{CWCE}, if there is an unequal interest across classes (FP2.5.1). Otherwise the canonical \ac{CE} should be assessed, e.g. using \ac{KCE} or \ac{ECEKDE} (see DG5.1). For \textbf{U3} "Comparison of overall performance across classifiers", we recommend reporting a \ac{PSR} (i.e., \ac{BS} or \ac{NLL}, see DG5.3) as the joint assessment of calibration and discrimination is exactly what this category of metrics is designed for. 
    
    \item[2: Select metric for assessing output interpretability (if any):]  
    This step selects an adequate metric for assessing the interpretability of the model output (FP2.7.3), which corresponds to \textbf{U4}. The first decision to be made in FP2.7.3 is whether to assess the calibration quality in isolation, as measured by \ac{CE} estimates, or jointly with discrimination as measured by overall performance measures. 
    %The latter amounts to assessing an "overall reliability" of predicted class scores, i.e. how well they match the true posterior probabilities.
    When deciding for calibration-only assessment, the core decision to be made is whether to measure top-label, marginal or canonical \ac{CE}, as detailed in DG 5.4. If there is an unequal interest across classes (FP2.5.3), a well-interpretable estimator of the marginal \ac{CE}, such as \ac{CWCE}, is recommended. Otherwise, the default option is to select a well-interpretable estimator of the canonical \ac{CE} (e.g., \ac{ECEKDE}) and a corresponding guaranteed upper bound (e.g., \ac{RBS}), together with the a per-class estimator of marginal \ac{CE} (e.g., \ac{CWCE}).
    Top-label calibration (as measured by \ac{ECE}) is only recommended in rare cases, as detailed in DG5.4. %The latter might be preferable when the user interest focuses on the decisions made by the classifier, and only the  probabilities of the resulting decisions are of interest. The latter strategy, in turn, implies an interest in a thorough model diagnosis or a broader task interest in all predictions including those with low scores. For example, an aggressive tumor may not be the most likely class, but a probability of 30\% would still have substantial implications for patient care. For both options, we recommend to additionally report the \ac{RBS} as a guaranteed upper bound of the canonical \ac{CE}, as both are known to potentially underestimate the true error.
\end{description}
Note that the selection of the same metric (e.g., \ac{CWCE}) in both steps is a potential outcome of the mapping. Crucially, metrics involving calibration assessment are generally prevalence-dependent. Thus, comparative studies as described in U2 and U3 are generally restricted to one data set and, if the prevalence of the data set does not represent the population of interest (see FP4.2), the calibration quality of a classifier needs to be re-validated on each new study cohort (see Fig.~\ref{fig:prevalence-dependency}). 

%Extra considerations:
%- in ILC metrics very different because no TN, no problem here
%- What is included in validatioN?
%- Background class: Ignore even though n+1 vector available (background discarded also in ILC because XY). Here can be ignored as well with the following arguments.
%- One exception: NLL

Calibration is most commonly assessed for image-level classification tasks. Due to the comparatively sparse research basis in the other problem categories, no specific recommendations are provided in our framework at this time. There are however, a few recent studies employing calibration metrics in object detection~\cite{kuppers2020multivariate,neumann2018relaxed} and slightly more studies in semantic segmentation, especially in the biomedical domain~\cite{judge2022reliable,mehrtash2020confidence,li2022estimating,sander2019towards}. 

Nevertheless, in theory, Subprocess S5 may also be traversed for object detection, instance segmentation, and semantic segmentation. When traversing S5 for object-level tasks, the following considerations should be noted:
\begin{itemize}
 \item \textbf{Calibration recommendations only apply to the classification part of object detectors:} As described in Suppl. Note~\ref{app:formats}, object detection and instance segmentation methods commonly provide outputs beyond the predicted class score vector such as bounding box coordinates or, depending on the method, 'intermediate objectness scores'~\cite{ren2015faster}. Thus, it is important to note that when utilizing calibration recommendations in this framework for object-level methods, the recommendations only apply to the classification output.
 \item \textbf{Why considerations in image-level classification translate to object detection:} When validating \textit{discrimination} performance, a fundamental conceptual difference between image level and object level is the fact that \acp{TN} are not defined in the latter case. This difference does not translate to calibration, where only predictions of the model $f \left( X \right)$ are validated. As the background class is discarded from validation (see below), this means that only \ac{TP} and \ac{FP} predictions are relevant for calibration, i.e., non-matched predictions are considered while non-matched reference objects (\acp{FN}) are not. A further conceptual difference between object-level classification and image-level classification is the former's additional requirement of localization to distinguish \acp{TP} and \acp{FP}. This aspect is inherently covered by calibration validation because non-matched predictions are simply considered as additional \ac{FP} errors (mistaking the 'true' background class for one of the foreground classes), equivalently to the standard \ac{FP} error case (mistaking two classes).
  \item \textbf{Dealing with the background class:} When validating classification performance in object-level tasks, the model output predicting class scores for the background class is commonly discarded (see Suppl. Note~\ref{app:formats})~\cite{everingham2015pascal}, because rewarding correct background predictions contradicts the task interest (there are no 'background objects') and would be easily exploitable (predicting high numbers of background objects). Further, penalization of background predictions is already ensured implicitly by considering them as \acp{FP} with respect to the true foreground class. Discarding the background class leads to class prediction vectors that do not sum to one, which is of no concern for validation as metrics do not rely on the probabilistic interpretation of scores. These considerations translate directly to calibration validation, which is equivalently exploitable by predicting high numbers of background objects. Here, the background class is discarded from outputs, and calibration of outputs only refers to the foreground class predictions (the actual 'object detection' outputs). Moreover, the calibration conditions introduced above describe the matching of single entries across two distributions and do not rely on their scores summing to one.
    \item \textbf{Non-applicability of the \ac{NLL} on object level:} There is one exception for object-level tasks where metric recommendations differ when traversing S5: The \ac{NLL} is not applicable. This is because the \ac{NLL} considers the predicted class score for the correct reference class ('true class probability'). For non-matched predictions, this 'true class' is the background class, which is discarded from validation as described above. In contrast, \ac{BS} remains applicable and a meaningful measure of \ac{CE} under the recommended protocol (i.e., when only considering foreground classes).
\end{itemize}

%\newpage
%\input{Subprocesses}

\newpage

\newpage
%------------------------------------
\subsection{Decision Guides}
\label{app:decision_guides}
%------------------------------------
\acresetall
While the problem fingerprint helps exclude common metrics that are not suitable for the driving problem, the final choice in each subprocess may not be unambiguous. In these cases, decision guides support the users in making an educated decision that best matches their preferences.

%---------------------------------------
\subsubsection{Decision guide S2}
%---------------------------------------
\label{ssec:dg2}
\hfill\\\noindent\textbf{DG2.1: \acf{WCK} versus \acf{EC}} 
\begin{tcolorbox}[title= Summary of DG2.1: \acs{WCK} versus \acs{EC}, colback=white]
    \begin{table}[H]
        \centering
        \begin{tabular}{p{5.5cm} p{0.3cm} p{5cm}}
            \textbf{\ac{WCK}} && \textbf{\ac{EC}}\\ 
                \textcolor{red}{\faMinusCircle } Designed for symmetric situations (guesses of two raters) && \textcolor{gray}{\faArrowCircleRight } Designed for asymmetric situations \\
               \textcolor{red}{\faMinusCircle } Limited interpretability && \textcolor{ODgreen}{\faPlusCircle} Good interpretability with normalized variant\\
                 \textcolor{ODgreen}{\faPlusCircle } Widely used && \textcolor{gray}{\faArrowCircleRight } Not widely used in biomedical image analysis \\
                             \textcolor{red}{\faMinusCircle} Lack of framework to identify and validate the decision rule applied to class scores && \textcolor{ODgreen}{\faPlusCircle } Availability of framework to identify and validate the decision rule applied to class scores\\
                              \textcolor{red}{\faMinusCircle} Possibility of paradoxical results &&   \\
 
        \end{tabular}
        \caption{Comparison of \acf{WCK} to \acf{EC} in the context of the decision guide DG2.1 for Subprocess S2. Context: unequal severity of class confusions (FP2.5.2 = TRUE), costs for class confusions available (FP2.5.3 = TRUE), and provided class prevalences reflect the population of interest (FP4.2 = TRUE).}
        \label{tab:d2-1}
    \end{table}
\end{tcolorbox}

Both \ac{WCK} (Fig.~\ref{fig:cheat-sheet-wck}) and \ac{EC} (Fig.~\ref{fig:cheat-sheet-ec}) are metrics that allow for incorporating task-specific penalties for confusions between individual pairs of classes. Common use cases for this property are tasks with ordinal classes or diagnostic decisions with errors of varying clinical severity. Importantly, however, Kappa statistics in general and \ac{WCK} in particular were originally proposed to compare annotations/guesses of two raters, which is a symmetric problem by nature. Validation studies, on the other hand, involve the comparison of a prediction to a reference that approximates the truth (asymmetric setting). Hence, unlike \ac{EC}, \ac{WCK} does not conceptually match the intended comparison. For this reason and due to further favorable properties, we generally recommend the usage of \ac{EC} rather than \ac{WCK}. When deciding between the two metrics, the following further properties are of relevance:
\begin{itemize}
    \item \textbf{Interpretability}: While both metrics can be interpreted as 'measures of (dis)agreement', the main difference is the fact that \ac{WCK} provides this measure in reference to 'agreement by chance'. The equivalent concept for \ac{EC} is its normalized variant \ac{ECN}, where the disagreement measure is divided by a 'random performance' measure. Due to the conceptual similarity, it is more sensible to compare \ac{WCK} to \ac{ECN}. Both metrics are prevalence-dependent due to relating model performance to a random performance reference. Their main difference is the definition of the 'random reference': In \ac{ECN} this reference is straightforward to interpret as the 'best possible naive classification system' which always predicts the most dominant class. The definition in \ac{WCK} stems from its symmetric concept to compare the predictions of two raters. The random reference in this case is the probability of both raters agreeing by chance. Using this definition in classification tasks results in random reference systems that can be weaker than the naive system of \ac{ECN}. Thus, the random reference in \ac{WCK} is less intuitive and arguably not useful in classification tasks (i.e., asymmetric settings). % Another important downside of interpretability in \ac{WCK} is the fact that the maximum score of the metric depends on the task, while for \ac{ECN} this is always 0.
    \item \textbf{Undesired behaviour in practice}:
    Using \ac{WCK} with quadratic weights, often done for ordinal tasks, has been found to lead to 'paradoxical results', as detailed in~\cite{warrens2012some}.
    \item \textbf{Popularity}: \ac{WCK} is widely used in the biomedical domain, whenever customized penalties for class confusion are required. \ac{EC}, on the other hand, is currently mostly found either in statistical textbooks or in non-related domains such as speech recognition.
    \item \textbf{Theoretical foundation}: 
    \ac{EC} comes with a comprehensive theoretical foundation based on Bayesian decision theory~\cite{ferrer2022analysis}. As a consequence, it is possible to analytically derive the optimal decision rule applied to the predicted class scores (more generally: decision region for more than two classes) and empirically validate the quality of this decision rule by means of calibration assessment. This is an important property in this context because the standard argmax-based decision rule (i.e., predicting the class with the highest class score) is by definition not optimal in scenarios with unequal costs of misclassifications.
\end{itemize}

\noindent\textbf{DG2.2: \acf{BA} versus \acf{EC}}
\begin{tcolorbox}[title= Summary of DG2.2: \acs{BA} versus \acs{EC}, colback=white]
    \begin{table}[H]
        \centering
        \begin{tabular}{p{5.5cm} p{0.3cm} p{5cm}}
            \textbf{\ac{BA}} & & \textbf{\ac{EC}} \\
                 \textcolor{gray}{\faArrowCircleRight } Prevalence independence
                   && \textcolor{ODgreen}{\faPlusCircle } Possibility of reflecting expected prevalences in the target population \\
                                  \textcolor{ODgreen}{\faPlusCircle } Widely used  && \textcolor{gray}{\faArrowCircleRight } Not commonly known in biomedical image analysis \\
        \end{tabular}
        \caption{Comparison of \acf{BA} to \acf{EC} in the context of the decision guide DG2.2 for Subprocess S2. Context: Equal severity of class confusions (FP2.5.2 = FALSE), either (1) unequal interest across classes (FP2.5.1 = TRUE) and no mismatch between class prevalences and class importance (FP2.5.3 = FALSE), or (2) equal interest across classes (FP2.5.1 = FALSE) and  provided class prevalences do not reflect the population of interest (FP4.2 = FALSE).}
        \label{tab:d2-2}
    \end{table}
\end{tcolorbox}
% \textcolor{red}{\faMinusCircle }
% \textcolor{ODgreen}{\faPlusCircle }
% \textcolor{gray}{\faArrowCircleRight }

When deciding between \ac{BA} and \ac{EC} in the provided context, two primary scenarios should be distinguished:

\textbf{Classes should contribute according to prevalence in the \textit{target} application:} Although the user may have an inherently equal interest in all classes (FP2.5.1 = FALSE), reporting a metric score to which all classes contribute equally may \textit{not} necessarily be desired. Instead, the user may simply be interested in the overall performance on a given data set and thus want classes to contribute according to their prevalence in the target application. This is not straightforward in the provided scenario because the data set at hand does not match the prevalences of the target population (FP4.2 = FALSE). In this case, we recommend \ac{EC}, because it offers a mechanism to explicit set (expected) class prevalences directly in the formula. This strategy allows getting a glimpse of a model's performance on the target application while validating on the data at hand. Application of \ac{EC} in this way, however, is only possible if the prevalences can be specified upfront.

%This can be useful for identifying the optimal cutoff on a new data set, as described in~\cite{ferrer2022analysis}, essentially rendering the re-calibration of class scores unnecessary. %(for whether calibration is of interest see FP2.7). 
\textbf{Each class should contribute equally to the metric:} In this case,  compensation for potential class imbalance is required in order to ensure equal contribution from each class. Here, we recommend \ac{BA} as metric because it was designed for exactly this purpose. Although \ac{EC} can be configured to be identical to \ac{BA} (Suppl. Note~\ref{app:metric-pool}), we favor \ac{BA} due to its widespread use. 

\ac{EC} also offers a complementary way to address class imbalance as it allows for the exchange of the class priors directly in the formula: When the class priors upon application on a new data set are known, they can be incorporated in \ac{EC}. This can be useful for identifying the optimal decision rule applied to predicted class scores on a new data set, as described in~\cite{ferrer2022analysis}, essentially rendering the re-calibration of class scores unnecessary. %(for whether calibration is of interest see FP2.7). 
However, one might argue that class priors being known upfront is quite a strong assumption for a new application.

\newpage
\noindent\textbf{DG2.3: \acf{BA} versus \acf{MCC} versus \acf{ECN}}
\begin{tcolorbox}[title= Summary of DG2.3: \acs{BA} versus \acs{MCC} versus \acs{ECN}, colback=white]
    \begin{table}[H]
        \centering
        \begin{tabular}{p{3cm} p{0.3cm} p{3cm} p{0.3cm} p{3cm}}
            \textbf{\ac{BA}} & & \textbf{\ac{MCC}} & & \textbf{\ac{ECN}}\\ 
            % \textcolor{red}{\faMinusCircle } Prevalence-dependent && \textcolor{ODgreen}{\faPlusCircle } Prevalence-independent && \textcolor{red}{\faMinusCircle } Prevalence-dependent\\
            \textcolor{ODgreen}{\faPlusCircle } Inherent interpretability with respect to naive classifier && \textcolor{ODgreen}{\faPlusCircle } Inherent interpretability with respect to naive classifier && \textcolor{ODgreen}{\faPlusCircle } Inherent interpretability with respect to naive classifier\\
            \textcolor{gray}{\faArrowCircleRight } Implication of equal class contribution && \textcolor{gray}{\faArrowCircleRight } Implication of equal class contribution && \textcolor{gray}{\faArrowCircleRight } No establishment of equal class contribution \\
            \textcolor{gray}{\faArrowCircleRight } Insensitive to predictive values (\ac{PPV} and \ac{NPV}) && \textcolor{gray}{\faArrowCircleRight } High scores ensure high predictive values (\ac{PPV} and \ac{NPV}) && \textcolor{gray}{\faArrowCircleRight } Limited sensitivity to predictive values (\ac{PPV} and \ac{NPV})\\
      \textcolor{ODgreen}{\faPlusCircle } Availability of framework to identify and validate decision rule applied to class scores && \textcolor{gray}{\faArrowCircleRight } Lack of framework to identify and validate the decision rule applied to class scores && \textcolor{ODgreen}{\faPlusCircle } Availability of framework to identify and validate the decision rule applied to class scores\\            \textcolor{ODgreen}{\faPlusCircle } Good interpretability && \textcolor{gray}{\faArrowCircleRight } Limited interpretability && \textcolor{ODgreen}{\faPlusCircle } Good interpretability\\
          \textcolor{ODgreen}{\faPlusCircle } Widely used && \textcolor{gray}{\faArrowCircleRight } Fairly well-known but not much used && \textcolor{gray}{\faArrowCircleRight } Not known or used in the biomedical imaging domain although based on well-studied statistical concepts\\
            % \textcolor{gray}{\faArrowCircleRight } Easy interpretation && \textcolor{gray}{\faArrowCircleRight } Hard to interpret && TODO\\

            % && \textcolor{gray}{\faArrowCircleRight } A low value indicates low values of related metrics && TODO
        \end{tabular}
        \caption{Comparison of \acf{BA} to \acf{MCC} to \acf{ECN} in the context of the decision guide DG2.3 for Subprocess S2. Context: Equal severity of class confusions (FP2.5.2 = FALSE), either (1) unequal interest across classes (FP2.5.1 = TRUE) and no mismatch between class prevalences and class importance (FP2.5.3 = FALSE) or (2) equal interest across classes (FP2.5.1 = FALSE),  provided class prevalences reflect the population of interest (FP4.2 = TRUE), presence of class imbalance (FP4.1 = TRUE) and compensation for class imbalances requested (FP2.5.5 = TRUE).}
        \label{tab:d2-3}
    \end{table}
\end{tcolorbox}

Three metrics are particularly attractive when class prevalences reflect the population of interest but compensation for class imbalance is desired (FP4.1 = TRUE and FP2.5.5 = TRUE). These are \ac{MCC}, \ac{BA}, and the normalized variant of \ac{EC}, \ac{ECN}. As described in Suppl. Note~\ref{app:fingerprints} (FP2.5.5 \textit{Compensation for class imbalance requested}), there are three effects of class imbalance that can be compensated for. 
\begin{itemize}
    \item \textbf{Effect 1:} Misleading metric values due to missing reference value for naive classifier
    \item \textbf{Effect 2:} Misleading metric values due to unequal contribution of classes
    \item \textbf{Effect 3:} Misleading metric values due to missing consideration of predictive values
\end{itemize}
While the most common multi-class metric, Accuracy, is subject to all three pitfalls when used in imbalanced settings, this decision guide discusses the three aforementioned alternatives (\ac{BA}, \ac{MCC}, and \ac{ECN}) that compensate for one or more of these effects. The following aspects are relevant when deciding between the three:

\textbf{Compensating for Effect 1:} All three metrics establish a fixed score for the performance of a naive classifier, i.e., one that always predicts the most frequent class -- which is a more realistic baseline in class imbalanced scenarios -- compared to an entirely random system. The corresponding scores are 0 for \ac{MCC}, 1 for \ac{ECN}, and 1$/$C for \ac{BA}, where C is the number of classes. However, the nature of the different compensation methods is fundamentally different. Consider the following confusion matrix of a binary classification system, as shown in Tab.~\ref{tab:cf-dg-2-3-eff-1}: 

\begin{table}[H]
    \centering
        \caption{Confusion matrix illustrating Effect 1.}
    \label{tab:cf-dg-2-3-eff-1}
    \begin{tabular}{l|l|c|c|}
        \multicolumn{2}{c}{}&\multicolumn{2}{c}{\textbf{Prediction}}\\
        \cline{3-4}
        \multicolumn{2}{c|}{}&\textit{Positive}&\textit{Negative}\\
        \cline{2-4}
        \multirow{ 2}{*}{\textbf{Actual}}& \textit{Positive} & TP = 100 & FN = 1\\
        \cline{2-4}
        & \textit{Negative} & FP = 100 & TN = 10,000 \\
        \cline{2-4}
        \end{tabular}
\end{table}

Respective metric values are \ac{BA}: 0.99, \ac{MCC}: 0.7, \ac{ECN}: 1. Although all metrics feature fixed values for a random classifier, the same system can be assessed differently, as it is being considered 'near-perfect' by \ac{BA} (0.99), 'fairly good' by \ac{MCC} (0.7), and 'random'/'naive' by \ac{ECN} (1). Intuitively, the \ac{BA} assessment seems overly optimistic, which can be attributed to the fact that \ac{BA} does not compensate for Effect 3, as described in more detail below. On the other hand, the \ac{ECN} assessment appears overly strict, which can be attributed to the fact that \ac{ECN} does not compensate for Effect 2 as described in more detail below. 

%A formal assessment of the interpretability of these metrics in a certain use case, however, requires knowledge about the specific task interests.

\textbf{Compensating for Effect 2:} In balanced scenarios, all classes are weighted equally by common discrimination metrics. In contrast, in imbalanced scenarios, common metrics such as Accuracy are dominated by the frequent classes. Equal contribution of classes in this context would imply that each class can contribute equally to the final metric score, irrespective of prevalence. This is exactly what \ac{BA} does by computing the average of individual class Sensitivities. An alternative way of thinking about this compensation is tweaking the costs of misclassification errors by assigning higher costs for errors in rare classes and vice versa. Hence, \ac{BA} can be thought of as a cost instantiation of \ac{EC} if the costs are set proportional to the inverse of class prevalences (see Suppl. Note~\ref{app:metric-pool}). Importantly, the normalized variant of \ac{EC}, \ac{ECN}, does not generally compensate for Effect 2, but merely rescales metric scores in a way that the value of 1 corresponds to a naive classifier always predicting the most frequent class (see Effect 1). In other words, the rankings obtained for a set of test cases would be the same for \ac{EC} and \ac{ECN}. Analogously to \ac{EC}, it is also possible to tweak the costs to compensate for Effect 2 in \ac{ECN}, but the resulting metric would yield no advantages over \ac{BA}. Importantly, the fact that \ac{ECN} does not compensate for Effect 2 implies that if there is an unequal interest across classes (FP2.5.1 = TRUE), then \ac{ECN} is the only correct choice. Analogously to \ac{BA}, \ac{MCC} establishes equal contribution of classes by assessing individual class sensitivities. 

\textbf{Compensating for Effect 3:} The predictive values (\ac{PPV} and \ac{NPV}) are an important aspect of assessing the quality of a classification system. To showcase this importance, consider the following confusion matrix of a binary classification task, as shown in Tab.~\ref{tab:cf-dg-2-3-eff-3}: 

\begin{table}[H]
    \centering
        \caption{Confusion matrix illustrating Effect 3.}
    \label{tab:cf-dg-2-3-eff-3}
    \begin{tabular}{l|l|c|c|}
        \multicolumn{2}{c}{}&\multicolumn{2}{c}{\textbf{Prediction}}\\
        \cline{3-4}
        \multicolumn{2}{c|}{}&\textit{Positive}&\textit{Negative}\\
        \cline{2-4}
        \multirow{ 2}{*}{\textbf{Actual}}& \textit{Positive} & TP = 10 & FN = 1\\
        \cline{2-4}
        & \textit{Negative} & FP = 100 & TN = 10,000 \\
        \cline{2-4}
        \end{tabular}
\end{table}

This system is assessed as 'near-perfect' by \ac{BA} (0.95), 'better than random, but not really useful' by \ac{MCC} (0.29), and 'much worse than random' by \ac{ECN} (9.2).

This example shows that \ac{BA} does not consider predictive values, thus yielding a near-perfect score despite a low \ac{PPV} of 0.09. This assessment could be considered a pitfall in many scenarios, where the classification system would be fairly useless. Consider, for instance, a breast cancer screening program where, based on the provided system, $>90\%$ of all biopsies (\acp{TP} + \acp{FP}) would be unnecessary (\acp{FP}).

In contrast, the \ac{MCC} score could be considered intuitive for many scenarios such as the screening example. This is due to \ac{MCC} explicitly considering all four basic rates \ac{TPR}, \ac{TNR}, \ac{PPV}, and \ac{NPV}. Thus, \ac{MCC} poses further requirements compared to \ac{BA}, which focuses only on Sensitivities. \ac{ECN} also ensures high predictive values by design. In practice, however, it is not always a good indicator for predictive values because of the sometimes overly strict penalization of errors, as seen in the above example. In theory the weights in \ac{ECN} could be adjusted to simulate the behavior of predictive value-sensitive metrics like \ac{MCC}, but this implies a trial-and-error tuning process on each new task.\\ 

\textbf{Identifying the optimal decision rule applied to predicted class scores:} The different strategies for identifying a decision rule applied to predicted class scores are described in FP2.6 (see Suppl. Note~\ref{app:fingerprints}). In the multi-class setting, argmax-based decision rules (i.e., predicting the class with the highest class score) are very common, but make arguably strong assumptions such as calibrated scores and equal penalization of all misclassifications. %Associated pitfalls are described in Suppl. Note~\textcolor{red}{TODO - WIP}. 

It should be noted here that metrics that can be viewed as instantiations of EC (in this case \ac{BA} and \ac{ECN}) come with a theoretical framework on how to validate the decision rule, i.e., answering the question “how much better could the classification performance have been with an optimal decision rule?”~\cite{ferrer2022analysis}.  \ac{MCC}, on the other hand, lacks such a framework.

\textbf{Interpretability:} Arguably, \ac{BA} features the most straightforward interpretation as the average over individual class Sensitivities, with bounded scores $[0, 1]$ and a fixed random reference at 1$/$C. \ac{ECN} scores are also fairly interpretable (“the EC of the system in relation to the EC of a naive system”), but scores are not bounded $[0, \infty)$. Furthermore, the random reference could be interpreted as 'too strict' for many scenarios such as the provided example. As for \ac{MCC}, a random reference value is provided at 0 and the scores are bounded [-1,1], but all intermediate scores are arguably less intuitive. The general interpretation of \ac{MCC} would be that it is a metric that depends on individual class Sensitivities and predictive values, i.e., a high \ac{MCC} score guarantees all of these being high and a low \ac{MCC} score indicates that at least one of them is low.

\textbf{Popularity:} \ac{BA} is a widely used metric. \ac{MCC} is fairly well-known but arguably not used as much. \ac{ECN} is used prominently in the field of speaker verification but has not been introduced to the biomedical imaging or clinical community yet, although the statistical concepts it is based upon are long-standing in Bayesian decision theory.

%\newpage
%---------------------------------------
\subsubsection{Decision guide S3}
%---------------------------------------
\label{ssec:dg3}

\hfill\\\noindent\textbf{DG3.1: Metric@(TargetMetric = TargetValue)} \\ 
If a target value for a specific metric (typically Sensitivity; Fig.~\ref{fig:cheat-sheet-sensitivity}) is provided, the decision rule applied to the predicted class scores is optimized such that the specific target value is reached on a validation data set (Suppl. Note~\ref{app:terminology}). Other metrics, depending on the target application, can then be reported for that specific threshold. Example Specificity@(Sensitivity = 0.95): As illustrated in Fig.~\ref{fig:def-auc}, the decision rule is set such that a Sensitivity of 0.95 is achieved. Other metric values (here Specificity; Fig.~\ref{fig:cheat-sheet-specificity}) can then be obtained from the corresponding fixed confusion matrix. In the example, this yields the Specificity at the predefined Sensitivity level. Possible candidates include Sensitivity (Fig.~\ref{fig:cheat-sheet-sensitivity}), Specificity (Fig.~\ref{fig:cheat-sheet-specificity}), \ac{PPV} (Fig.~\ref{fig:cheat-sheet-ppv}), \ac{NPV}, (Fig.~\ref{fig:cheat-sheet-npv}) and \ac{FPPI} (Fig.~\ref{fig:cheat-sheet-fppi}).

%\newpage
\hfill\\\noindent\textbf{DG3.2: \acf{NB} versus \acf{EC}} 
\begin{tcolorbox}[title= Summary of DG3.2: \acs{NB} versus \acs{EC}, colback=white]
    \begin{table}[H]
        \centering
        \begin{tabular}{p{5.5cm} p{0.3cm} p{5cm}}
            \textbf{\ac{NB}} & & \textbf{\ac{EC}} \\ 
                \textcolor{gray}{\faArrowCircleRight } Decisions can be defined directly based on predicted class scores, interpreted as risks && \textcolor{gray}{\faArrowCircleRight } Decisions based on explicit definition of misclassification costs \\
                 \textcolor{gray}{\faArrowCircleRight } Weighting of \acf{TP} against \acf{FP} in risk perspective && \textcolor{gray}{\faArrowCircleRight } Weighting of \acf{FP} against \acf{FN} in cost perspective \\
              \textcolor{gray}{\faArrowCircleRight } Lack of framework to validate the decision rule applied to class scores && \textcolor{ODgreen}{\faPlusCircle } Availability of framework to validate the decision rule applied to class scores \\
               \textcolor{ODgreen}{\faPlusCircle } Focus on reflectance of the (e.g., clinical) interest in the scores && \textcolor{ODgreen}{\faPlusCircle } Inherent interpretability with respect to naive classifier \\
                \textcolor{gray}{\faArrowCircleRight } Popular metric in clinical studies but not common in image analysis && \textcolor{gray}{\faArrowCircleRight }  Not known or used in the biomedical imaging domain although based on well-studied statistical concepts \\
               % \textcolor{gray}{\faArrowCircleRight } Assesses clinical utility of a prediction && \textcolor{gray}{\faArrowCircleRight } Requires the a priori definition of a range of clinically relevant cutoffs\\
        \end{tabular}
        \caption{Comparison of \acf{NB} and \acf{EC} in the context of the decision guide DG3.2 for Subprocess S3. Context: FP2.6 = cost-benefit-based decision rule applied to predicted class scores requested.}
        \label{tab:d3-2}
    \end{table}
\end{tcolorbox}

This decision guide is embedded in the framework in Subprocess S3, which guides the selection of metrics that are reported separately for each class. In multi-class tasks (i.e., more than two classes present) this reporting amounts to a one-versus-rest validation scheme. However, this scheme is not intuitively applicable to a cost-benefit analysis (what are the costs and benefits of the 'rest' class?), which is the concept behind decision rules of both metrics in this decision guide. Thus, for multi-class tasks we recommend to only proceed with the metrics selected in Subprocess S2 (e.g., \ac{EC} or \ac{WCK}) and not select any further metrics here to be reported in a one-versus-rest fashion, i.e.. we recommend to skip the guide. 

Both \ac{NB} (Fig.~\ref{fig:cheat-sheet-nb}) and \ac{EC} (Fig.~\ref{fig:cheat-sheet-ec}) are linked to cost-benefit analysis~\cite{pauker1975therapeutic} and are well-suited when a cost-benefit-based approach for determining an appropriate decision rule applied to the predicted class scores is desired (FP2.6 = cost-benefit-based). To this end, both require the knowledge of task-dependent tradeoffs between benefits and costs, as detailed below.
%to determine a corresponding decision rule (i.e. cutoff on predicted class scores). 
%The two metrics are calculated on the resulting confusion matrix and technically very similar in the sense that both multiply costs/risk with (variations of) error rates.\\
The following aspects are relevant when deciding between \ac{EC} and \ac{NB} (note that cost-based decision rule applied to predicted class scores is only considered for binary classification tasks in the scope of this work, thus referred to as a cutoff in this context):

\textbf{Cost versus risk perspective:} \textit{Cost perspective:} For \ac{EC}, explicit costs for both basic misclassifications (\ac{FP}, \ac{FN}) need to be defined or estimated. The optimal decision rule (i.e., cutoff on predicted class scores) that minimizes these costs can be analytically determined without data-based optimization. \textit{Risk perspective:} In contrast, \ac{NB} does not require the costs to be defined explicitly. Instead, predicted class scores are interpreted as probabilities or 'risks' of certain model output scores belonging to the positive class and the cutoff on the scores is defined directly on this scale based on task interest (e.g., “only treat patients with cancer risk >10\%”). This can be interpreted as an implicit cost-benefit analysis resulting in a single intuitive risk score. However, it is also common for \ac{NB} to make this cost-benefit analysis more explicit and define the risk as a relation of the benefit of \acp{TP} to the harms caused by \acp{FP}. A diagnostic test, for example, may lead to early identification and treatment of a disease, but typically the process will also cause some patients without disease being subjected to unnecessary further interventions. \ac{NB} allows to consider such tradeoffs by putting the benefits and harms of the test on the same scale so that they can be directly compared. A physician may, for example, state that 10 \acp{FP}, resulting in unnecessary biopsies, are acceptable to find one more cancer case (\ac{TP}).

\textbf{Decision curves:} In most scenarios it is not possible to precisely define the costs or risks associated with the task. For example, it is not straightforward to make an exact decision on how many \acp{FP} would be acceptable to obtain one more \ac{TP}. To compensate for this uncertainty, it is common practice to plot \ac{NB} over a “reasonable range of risk thresholds” resulting in so-called decision curves~\cite{vickers2016net}. This analysis allows assessing and comparing methods according to their \ac{NB} scores without relying on a single cutoff.  Although not common practice, one could also generate such curves for \ac{EC} when expressing cost ratios as a risk score (i.e., switching from the cost to the risk perspective).

\textbf{Cutoff on predicted class scores:}
In \ac{NB}, the cost-benefit-based cutoff, which is determined directly from provided knowledge about the task and does not require data-based optimization, is an explicit part of the metric computation. In contrast, \ac{EC} allows to alternatively determine a data-based cutoff by taking into account the provided costs in the metric calculation and minimizing \ac{EC} on a dedicated data split, if available. A further difference between the two metrics is the way prevalence dependency is handled: \ac{EC} isolates the class priors from the predicted class scores and defines them as a parameter of the cutoff itself, such that all application dependent parameters (costs and class priors) are part of the cutoff~\cite{ferrer2022analysis}. Upon deployment of a model on a new data set, the threshold can simply be updated analytically. Note that this process only works under the arguably strong assumption that the class priors of the new data set are known. In contrast, \ac{NB} considers risk scores that incorporate the class priors, implying that the threshold depends solely on the cost-benefit tradeoff. As a consequence, when the class priors shift on a new data set, the risk-cutoff in \ac{NB} requires class scores to be re-calibrated. The latter might be a harder requirement because it requires a labeled validation set for re-calibration as opposed to requiring merely the class priors of the new data set for a threshold update.

\textbf{Interpretability:} The following tradeoff exists between the two metrics regarding interpretability: \ac{EC} allows reporting a normalized version (\ac{ECN}), which makes the metric scores interpretable with regard to the performance of a random classifier. In contrast, in \ac{NB}, the reference to a random classifier is typically done manually (by comparing the two scores), because NB itself allows for an interpretation as the 'proportion of net-TP', which would get lost by normalization.

\textbf{Calibration:} Both metrics rely on the fact that predicted class scores are well-calibrated with regard to a chosen cutoff. \ac{EC} allows assessing this requirement by calculating the extra cost entailed by miscalibration (or the potential for reducing cost by calibrating scores)~\cite{ferrer2022analysis}. The calibration error here is measured as the increase of EC with the analytical, i.e., task interest-based, cutoff compared to an empirical cutoff optimized on the data. Compared to related calibration errors (see Suppl. Note~\ref{app:calibration}), this technique assesses a weaker calibration condition, which is directly targeted to the decision process at hand. For instance, even when assessing the relatively weak top-label calibration condition by means of \ac{ECE} with two bins and the border at the determined cutoff value, the distribution inside the bins would be considered, while \ac{EC} only focuses on how many more cases would have been on the 'correct side of the cutoff' if scores were calibrated, without considering score distributions on either side of the cutoff.

\textbf{Popularity:} Neither \ac{NB} nor \ac{EC} are widely used in the biomedical image analysis community. \ac{NB} is a popular metric in clinical studies, while \ac{EC} is currently not used but is part of a coherent framework of intuitive metric formulations (linked to Accuracy, \ac{BA}, and extends to multi-class scenarios).

%\newpage
\hfill\\\noindent\textbf{DG3.3: \acf{LR+} versus Sensitivity} 
\begin{tcolorbox}[title= Summary of DG3.3: \acs{LR+} versus Sensitivity, colback=white]
    \begin{table}[H]
        \centering
        \begin{tabular}{p{5.5cm} p{0.3cm} p{5cm}}
            \textbf{\ac{LR+}} & & \textbf{Sensitivity} \\ 
                 \textcolor{ODgreen}{\faPlusCircle } Straightforward application in the case of an optimization-based decision rule (FP2.6) &&  \textcolor{red}{\faMinusCircle } Challenging application in the case of an optimization-based decision rule (FP2.6)\\
                 \textcolor{ODgreen}{\faPlusCircle } Interpretation often reflecting interest in binary tasks && \textcolor{ODgreen}{\faPlusCircle } Good interpretability\\
        \end{tabular}
        \caption{Comparison of \ac{LR+} and Sensitivity in the context of the decision guide DG3.3 for Subprocess S3. Context: FP2.6 = optimization- or argmax-based decision rule applied to predicted class scores requested and provided class prevalences do not reflect the population of interest (FP4.2 = FALSE).}
        \label{tab:d3-3}
    \end{table}
\end{tcolorbox}
% \textcolor{red}{\faMinusCircle }
% \textcolor{ODgreen}{\faPlusCircle }
% \textcolor{gray}{\faArrowCircleRight }
This decision guide helps deciding between \ac{LR+} and Sensitivity in the context of per-class validation (Subprocess S3) with an optimization- or argmax-based decision rule applied to predicted class scores (FP2.6).

\ac{LR+} (Fig.~\ref{fig:cheat-sheet-lr+}) is the likelihood ratio of the positive class. In a clinical example where the quality of a diagnostic test is to be assessed,  this could be interpreted as the ratio of the likelihood of a diseased patient receiving a positive test result versus a healthy patient receiving a positive test result ($P(t+ | d+) / P(t+|d-)$, where t/d denotes a positive(+)/negative(-) test/disease status). In other words: How much more likely is the occurrence of a positive test result for a diseased person compared to a healthy person? The formal calculation for this metric  boils down to the following formula: \ac{LR+} = \ac{TPR} / (1-\ac{TNR}), where \ac{TPR}/\ac{TNR} are the Sensitivities of the positive/negative class.

In the provided context of this decision guide, where metrics are reported individually per class, Sensitivity (Fig.~\ref{fig:cheat-sheet-sensitivity}) and \ac{LR+} convey similar information and there is no 'incorrect' choice. Thus, the choice between the two can generally be made as the metric that is easier to interpret in the given task: In binary classification tasks (e.g., the provided example), \ac{LR+} conveys Sensitivities of both classes in a single score. Due to its intuitive and meaningful interpretation, it is often reported in clinical studies. In multi-class settings (which, in this context, amount to a one-versus-rest validation scheme), Sensitivities are generally easier to interpret, while the direct interpretation of \ac{LR+} as a property of a (clinical) test does not apply.

In case the decision rule applied to predicted class scores is to be determined on the basis of optimization on the target class, one additional consideration is of importance  (FP2.6 = optimization-based decision rule). When reporting Sensitivity per class, the decision rule can not be optimized based solely on the single Sensitivity at hand because this would always yield a cutoff value of 1. \ac{LR+} naturally overcomes this problem. Other possible workarounds include choosing a different decision rule (FP2.6) or optimizing a weighted average over Sensitivity for all classes instead. The latter option should only be considered if meaningful weights across classes can be defined (e.g., based on class importance).

%\newpage
\hfill\\\noindent\textbf{DG3.4: \acf{LR+} versus Sensitivity versus F$_\beta$ Score} 
\begin{tcolorbox}[title= Summary of DG3.4: \acs{LR+} versus Sensitivity versus F$_\beta$ Score, colback=white]
    \begin{table}[H]
        \centering
        \begin{tabular}{p{3cm} p{0.3cm} p{3cm} p{0.3cm} p{3cm}}
            \textbf{\ac{LR+}} & & \textbf{Sensitivity} & & \textbf{F$_\beta$ Score}\\ 
                \textcolor{ODgreen}{\faPlusCircle } Meaningful interpretation in binary tasks && \textcolor{ODgreen}{\faPlusCircle } Generally good interpretability && \textcolor{gray}{\faArrowCircleRight } Limited interpretability\\
               \textcolor{ODgreen}{\faPlusCircle } Inherent interpretability with respect to naive classifier && \textcolor{ODgreen}{\faPlusCircle } Inherent interpretability with respect to naive classifier only when averaging over classes &&  \textcolor{red}{\faMinusCircle } No interpretability with respect to naive classifier\\
                \textcolor{red}{\faMinusCircle } Insensitive to \ac{PPV} && \textcolor{red}{\faMinusCircle } Insensitive to \ac{PPV} && \textcolor{ODgreen}{\faPlusCircle } High scores ensures high \ac{PPV}\\
        \end{tabular}
        \caption{Comparison of \ac{LR+}, Sensitivity and F$_\beta$ Score in the context of the decision guide DG3.4 for Subprocess S3. Context: FP2.6 = optimization- or argmax-based decision rule applied to predicted class scores requested and provided class prevalences reflecting the population of interest (FP4.2 = TRUE).}
        \label{tab:d3-4}
    \end{table}
\end{tcolorbox}
% \textcolor{red}{\faMinusCircle }
% \textcolor{ODgreen}{\faPlusCircle }
% \textcolor{gray}{\faArrowCircleRight }
 
In the context of this decision guide, prevalence dependency is not an exclusion criterion (see FP4.2) and thus F$_\beta$ Score (Fig.~\ref{fig:cheat-sheet-fbeta}) can be considered as an alternative to Sensitivity-based metrics (Sensitivity and \ac{LR+}, Figs.~\ref{fig:cheat-sheet-sensitivity} and~\ref{fig:cheat-sheet-lr+}). Details for the decision between the latter are provided in DG3.3; the present guide focuses on the pros and cons of opting for F$_\beta$ Score.

Per-class validation is commonly performed in a one-versus-rest fashion, naturally introducing class imbalance into the validation. Exceptions are binary scenarios with two balanced classes. 
For this exception, no compensation for class imbalance is needed (FP2.5.5 = FALSE) and the choice between F$_\beta$ Score and Sensitivity-based metrics becomes less relevant, i.e., there are no obvious incorrect choices. Thus, the decision can be made on the basis of which metric is easier to interpret in a given task. For all other cases, the decision should be based on whether compensation for class imbalance is required  (FP2.5.5 = TRUE).

\textbf{Compensation for class imbalance:} As described in FP2.5.5 (and explained in more detail in Suppl. Note~\ref{app:fingerprints}, "Compensation for class imbalance"), there are three aspects of compensation for class imbalance:
\begin{enumerate}
\item \textbf{Establishing a reference value for random performance:} \ac{LR+} provides a fixed random reference value at \ac{LR+} $=1$, while for Sensitivity the scores of individual classes can vary and only their average is fixed at “1/number of classes” (equivalent to \ac{BA}). F$_\beta$ Score does not provide a reference value for random performance.
\item \textbf{Establishing equal class contribution:} In the provided context (S3), the validation is performed per class, such that this aspect is irrelevant. 
\item \textbf{Establishing consideration of predictive values:} This aspect is the main reason to opt for F$_\beta$ Score in this decision guide, because it is the only metric of the three where high scores ensure a high \ac{PPV}. In contrast, \ac{LR+} and Sensitivity are insensitive to \ac{PPV}, which, depending on the task interest, can substantially diminish their utility. An exemplary pitfall related to this choice is the confusion matrix of a binary classification task, as shown in Tab.~\ref{tab:cf-dg-2-3-eff-3}. This classification system yields (for the two individual classes) Sensitivities of ($90\%$, $99\%$) and \ac{LR+} of (90, 90), respectively.  Resulting F$_1$ Scores are (0.165, 0.995), indicating a low \ac{PPV} and thus unsatisfying performance for the rare positive class.
This pitfall may be of practical relevance in class-imbalanced tasks where \acp{FP} shall not be neglected. For example, in breast cancer screening, the provided classifier would not be useful, because $>90\%$ of all biopsies (\ac{TP}+\ac{FP}) would be unnecessary (\ac{FP}).
\end{enumerate}

\textbf{Interpretability:} Out of the three, Sensitivity is arguably the easiest-to-interpret metric (exceptions are binary tasks, where \ac{LR+} might be preferable as detailed in DG3.3, see Tab.~\ref{tab:d3-3}). F$_\beta$ Score can be interpreted as the harmonic mean of Sensitivity and \ac{PPV}, which adds a layer of complexity to the interpretation compared to Sensitivity. Thus, if the aspects discussed in “compensation for class imbalance” are not relevant, F$_\beta$ Score might not be the metric of choice.

\begin{figure}[H]
    \centering
    \includegraphics[width=1\linewidth]{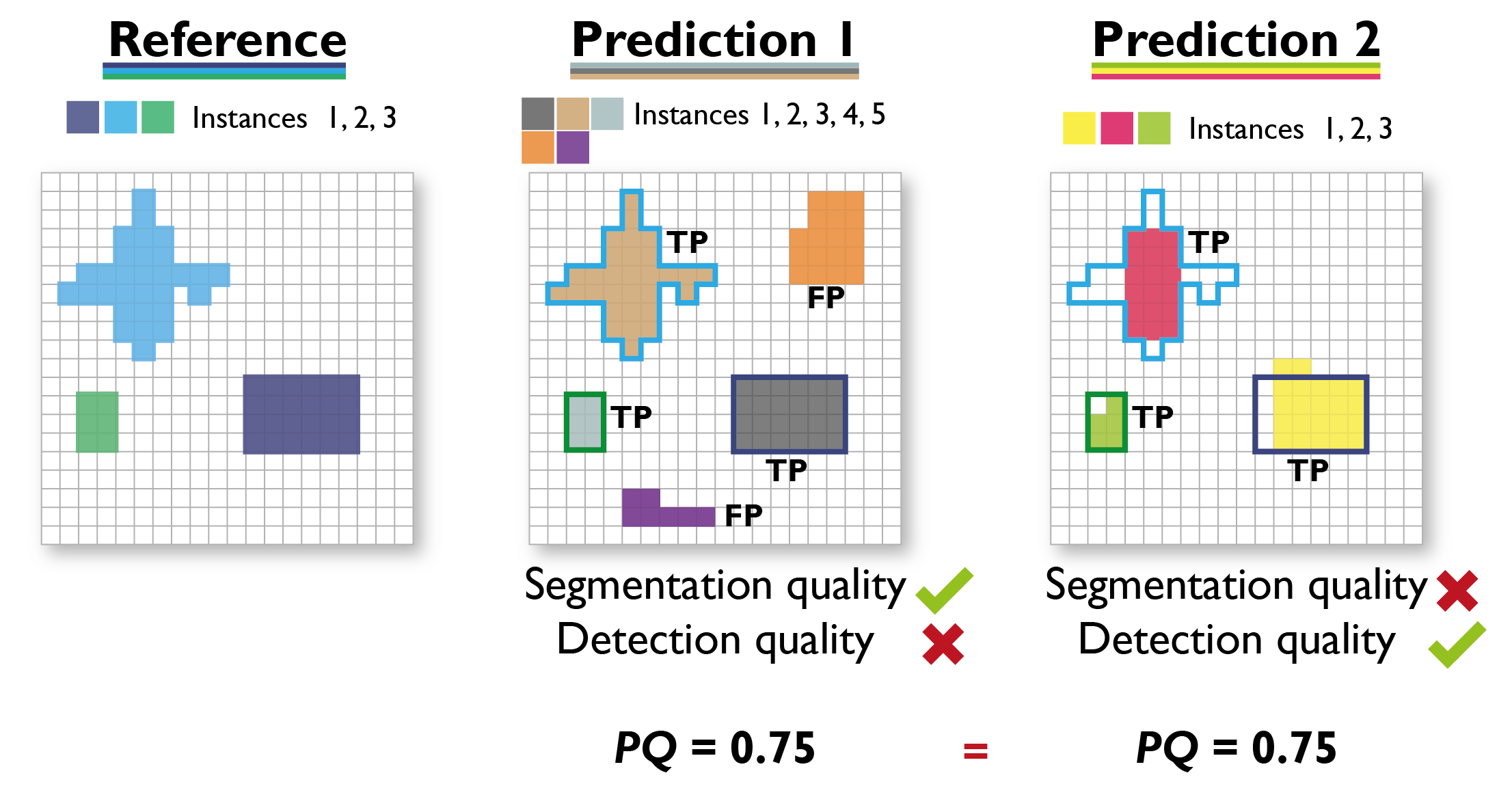}
    \caption{Effect of assessing segmentation and detection quality in a single score. \textit{Prediction 1} achieves a high segmentation but low detection quality (with several \acf{FP} predictions); vice versa for \textit{Prediction 2} (only predicting \acf{TP} instances; no \ac{FP} but low segmentation quality). However, both yield the same \acf{PQ} score.}
     \label{fig:pq-sq-vs-dq}
\end{figure}

\newpage
\noindent\textbf{DG3.5: How to determine $\beta$ in F$_\beta$ Score} 
\begin{tcolorbox}[title= Summary of DG3.5: $\beta$ in F$_\beta$ Score, colback=white]
    \begin{table}[H]
        \centering
        \begin{tabular}{p{3cm} p{0.3cm} p{3cm} p{0.3cm} p{3cm}}
            \textbf{$\beta$ < 1} & & \textbf{$\beta$ = 1} & & \textbf{$\beta$ > 1} \\ 
            \textcolor{gray}{\faArrowCircleRight } Higher weighting of \acf{FP} penalties (\acf{PPV}) & & \textcolor{gray}{\faArrowCircleRight } Harmonic mean of \ac{PPV} and Sensitivity & & \textcolor{gray}{\faArrowCircleRight } Higher weighting of \acf{FN} penalties (Sensitivity)\\
        \end{tabular}
        \caption{Determining the hyperparameter of the F$_\beta$ Score in the context of the decision guide DG3.5 for Subprocess S3. Context: [Image-level classification (ImLC)]: FP2.6 = optimization- or argmax-based decision rule applied to predicted class scores requested and provided class prevalences reflecting the population of interest (FP4.2 = TRUE). [Object detection (ObD) or instance segmentation (InS)]: Either no predicted class scores available (FP5.1 = FALSE) or FP2.6 = optimization- or argmax-based decision rule applied to predicted class scores requested.}
        \label{tab:d3-5}
    \end{table}
\end{tcolorbox}
The F$_\beta$ Score (Fig.~\ref{fig:cheat-sheet-fbeta}) is defined as:
\begin{equation}
F_\beta = (1 + \beta^2) \cdot \frac{\text{PPV} \cdot \text{Sensitivity}}{(\beta^2 \cdot \text{PPV}) + \text{Sensitivity}} = \frac{(1 + \beta^2) \cdot \text{TP}}{(1 + \beta^2) \cdot \text{TP} + \beta^2 \cdot \text{FN} + \text{FP}} 
\end{equation}
The most common choice is to set $\beta$ to $1$, resulting in equal weighting of \ac{FP} and \ac{FN} penalties. If unequal penalization of class confusions is desired (see FP2.5.2), higher values of $\beta$ result in higher weights on \ac{FN} penalties compared to \ac{FP} penalties and thus imply a focus on Sensitivity compared to \ac{PPV}. \\ \\
%\newpage
\noindent\textbf{DG3.6: F$_\beta$ Score versus \acf{PQ}} 
\begin{tcolorbox}[title= Summary of DG3.6: F$_\beta$ Score versus \acs{PQ}, colback=white]
    \begin{table}[H]
        \centering
        \begin{tabular}{p{5.5cm} p{0.3cm} p{5cm}}
            \textbf{F$_\beta$ Score} & & \textbf{\ac{PQ}} \\ 
             \textcolor{gray}{\faArrowCircleRight } Pure detection metric &&  \textcolor{gray}{\faArrowCircleRight } Hybrid metric for assessing detection and segmentation quality\\
        \end{tabular}
        \caption{Comparison of F$_\beta$ Score and \acf{PQ} in the context of the decision guide DG3.6 for Subprocess S3. Context: FP2.6 = optimization- or argmax-based decision rule applied to predicted class scores and FP1.1 = instance segmentation (InS).}
        \label{tab:d3-6}
    \end{table}
\end{tcolorbox}
% \textcolor{red}{\faMinusCircle }
% \textcolor{ODgreen}{\faPlusCircle }
% \textcolor{gray}{\faArrowCircleRight }
The F$_\beta$ Score (Fig.~\ref{fig:cheat-sheet-fbeta}) is a pure detection metric counting \ac{TP}, \ac{FP}, and \ac{FN} detections on instance level (specifically, it represents the harmonic mean of \ac{PPV} (Fig.~\ref{fig:cheat-sheet-ppv}) and Sensitivity (Fig.~\ref{fig:cheat-sheet-sensitivity}; see also \citep{reinke2021common}). The “segmentation aspect” of instance segmentation is here only incorporated via a prior cutoff on the localization criterion operating on pixel level (e.g., “\ac{IoU} > 0.5”). In case shifting the focus of validation more towards the segmentation quality of successfully matched (\ac{TP}) instances is desired, there are two options: 
\begin{enumerate}
    \item \textbf{Complementary segmentation metric:} One option is to select separate segmentation metrics in addition to object detection metrics such as F$_\beta$ Score on a per-instance basis (e.g. “\ac{DSC} per \ac{TP}-instance”). This selection is naturally incorporated in the instance segmentation recommendation (Fig.~\ref{fig:meta-mapping}) by the subroutines S6 (Extended Data Fig.~6) and S7 (Extended Data Fig.~7).
    \item \textbf{Hybrid metric:} An alternative is to select \ac{PQ} (Fig.~\ref{fig:cheat-sheet-pq}) instead of F$_\beta$ Score, which allows expressing both interests (detection performance and segmentation quality) in a single score. Essentially, \ac{PQ} is a modified F$_1$ Score, where \ac{TP} instances do not count as “1” in the calculation, but the “1” is replaced with the associated \ac{DSC} score (range [0,1]) of the instance. While combining the two aspects in a single score might be desirable, e.g., for method benchmarking or ranking, on the downside, such combined metrics make it harder to trace back performance to individual aspects (in this case: object detection versus segmentation; see Fig.~\ref{fig:pq-sq-vs-dq}). 
\end{enumerate}

%\newpage

\newpage
%---------------------------------------
\subsubsection{Decision guide S4}
%---------------------------------------
\label{ssec:dg4}

\begin{figure}[H]
    \centering
    \includegraphics[width=0.8\textwidth]{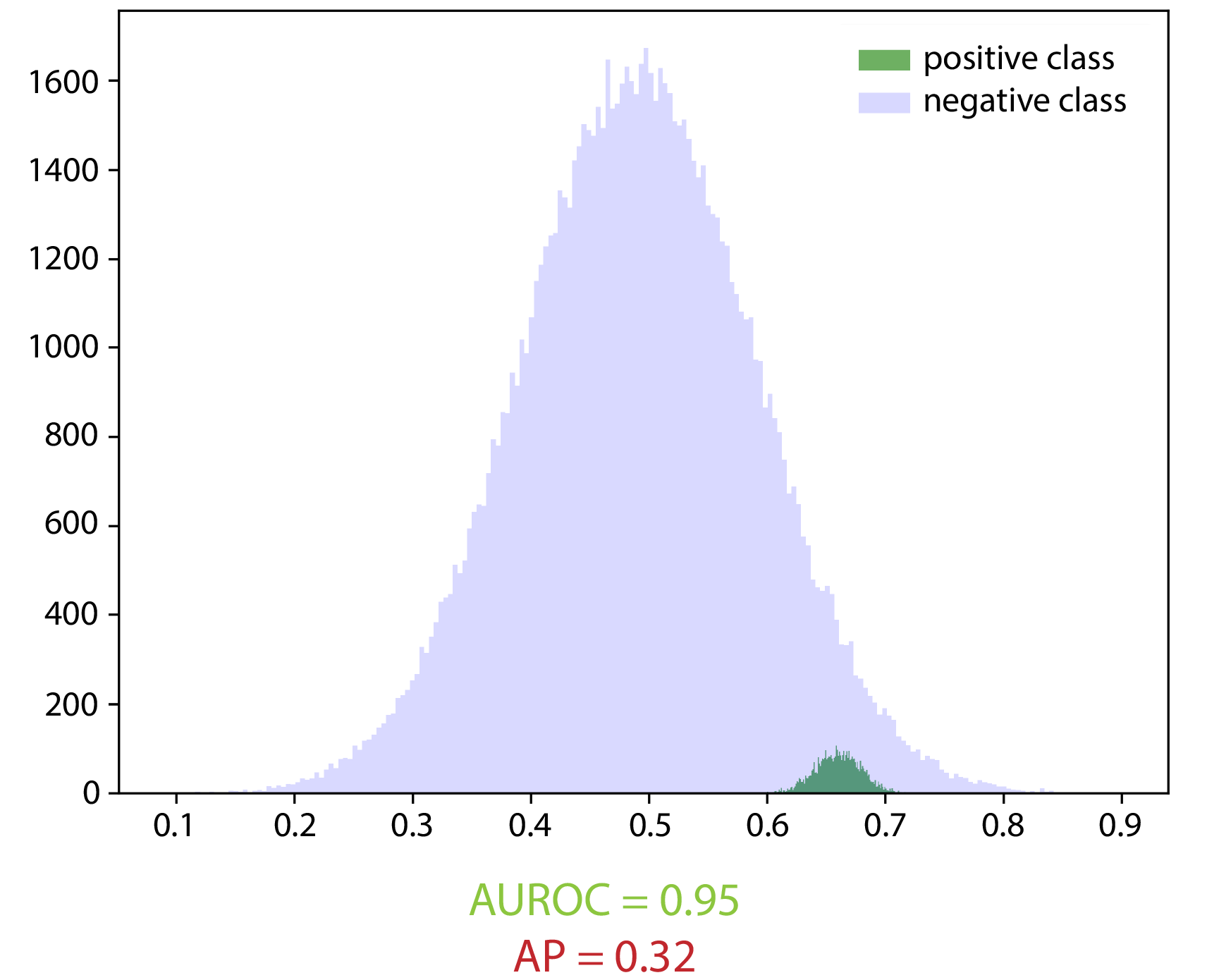}
    \caption{\acf{AUROC} scores neglect the \acf{PPV} in class-imbalanced settings and might lead to misinterpretation of a model's discrimination quality. The figure shows the simulation outcome for a binary classification problem with a low prevalence for the positive class. A clinical example of this scenario are cancer screening programs, where most of the subjects are healthy. While \ac{AUROC} is agnostic to the class prevalence and thus implies near-perfect discrimination with a score of 0.95, the prevalence-dependent \acf{AP} allows focusing on discrimination of the rare positive class by explicitly considering the \ac{PPV} and yields an intuitive score of 0.32.}
    \label{fig:hist-pitfall}
\end{figure}

\hfill\\\noindent\textbf{DG4.1: \acf{AUROC} versus \acf{AP}}
\begin{tcolorbox}[title= Summary of DG4.1: \acs{AUROC} versus \acs{AP}, colback=white]
    \begin{table}[H]
        \centering
        \begin{tabular}{p{5.5cm} p{0.3cm} p{5cm}}
            \textbf{\ac{AUROC}} & & \textbf{\ac{AP}} \\ 
             \textcolor{gray}{\faArrowCircleRight } Insensitive to \acf{PPV} under class imbalance  &&   \textcolor{ODgreen}{\faPlusCircle } High scores ensure high \ac{PPV} including under class imbalance\\
            \textcolor{ODgreen}{\faPlusCircle } Inherent interpretability with respect to naive classifier  &&  \textcolor{gray}{\faArrowCircleRight } Prevalence-dependent reference value for naive classifier\\
             \textcolor{ODgreen}{\faPlusCircle } Straightforward interpretability &&  \textcolor{gray}{\faArrowCircleRight } Limited interpretability\\
        \end{tabular}
        \caption{Comparison of \acf{AUROC} and \acf{AP} in the context of the decision guide DG4.1 for Subprocess S4. Context: availability of predicted class scores (FP5.1 = TRUE), FP1.1 = image-level classification (ImLC) and provided class  prevalences reflecting the population of interest.}
        \label{tab:d4-1}
    \end{table}
\end{tcolorbox}
The comparison between the two concepts behind \ac{AUROC} and \ac{AP}, i.e., the comparison between \ac{ROC} curves and \ac{PR} curves has been extensively studied \cite{davis2006relationship}. In practice, the choice between the two metrics boils down to the following aspects (if no clear choice can be made, we recommend reporting both metrics):

\textbf{Compensation for class imbalance effects:} Of relevance in the context of this decision guide is pitfall 3 from FP2.5.5: "Missing consideration of predictive values" (more on this topic can be found in Suppl. Note~\ref{app:fingerprints}, "Compensation for class imbalance"). \ac{AUROC} is based on the Sensitivities of the two classes and does not consider predictive values. In class-imbalanced scenarios, this may lead to pitfalls such as depicted in Fig.~\ref{fig:hist-pitfall}, where near-perfect \ac{AUROC} scores hide the fact that a system might have limited to no predictive utility. \ac{AP} assesses the predictive value of the positive class (\ac{PPV}) and thus compensates for the undesired effects caused by class imbalance: In the provided example, \ac{AP} yields an intuitive score of 0.32, reflecting the low \ac{PPV} and thus low utility of the system in the context of the task. A technical explanation is given by the fact that the high number of \acp{TN} dominate and suppress the \acp{FP} in the calculation of the \ac{TNR}, thus yielding high scores for \ac{AUROC}. A practical example for this pitfall might be a breast screening program, where a high \ac{PPV} is of great importance to prevent unnecessary biopsies (\acp{FP}). The focus of \ac{AP} on the positive class further has the effect that the resulting scores differ depending on which of the two classes is defined as positive and negative. This is in contrast to \ac{AUROC}, which yields the same scores irrespective of this definition. The general approach for \ac{AP}-based assessment in class-imbalanced scenarios is to define the rare class as the positive class. The fact that \ac{AP} focuses on the positive class reflects the task interest of not letting rare (important) events be dominated by frequent events in the metric score.\\

\textbf{Interpretability:}
\ac{AUROC} is easy to interpret as it simply represents the probability of a randomly sampled positive case having a higher predicted class score than a randomly sampled negative case. It further comes with a fixed reference value for the performance of a random classifier at 0.5. \ac{AP}, on the other hand, is harder to interpret and features no fixed random reference value. Instead, the \ac{AP} score of a random classifier is the prevalence of the positive class which varies on each data set.

\textbf{Implementations:} For reasons described in~\cite{davis2006relationship}, the \ac{PR} curve is more complex to interpolate compared to the \ac{ROC} curve. This results in the existence of various implementations of \ac{AP}, whereas no such heterogeneities exist for \ac{AUROC}. 

\textbf{Popularity:}
Although \ac{AUROC} is the common choice for multi-threshold metrics, \ac{AP} is also widely known and used.

\begin{figure}[H]
    \centering
    \includegraphics[width=0.8\linewidth]{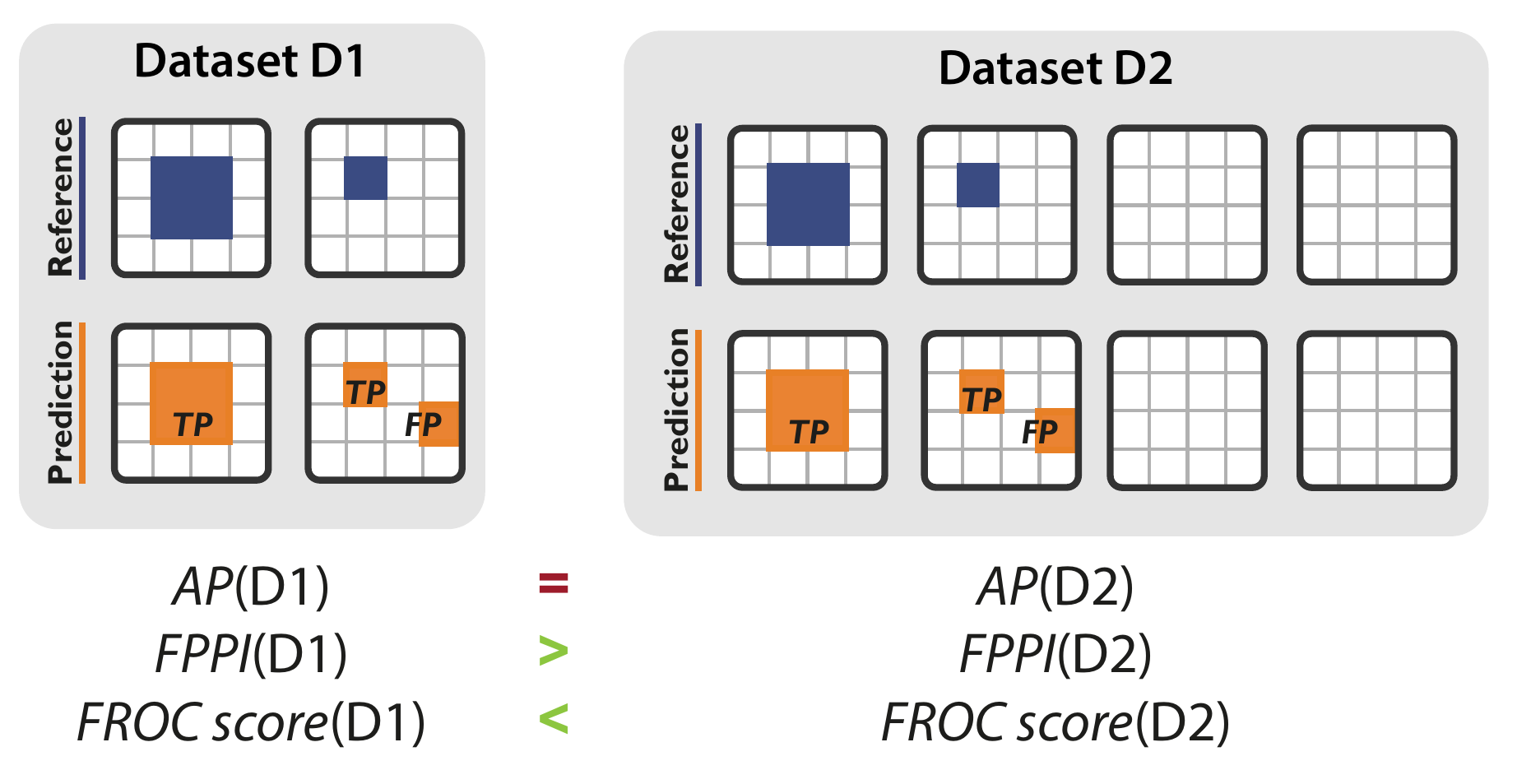}
    \caption{Effect of the number of images per data set on the metric scores. The \acf{AP} metric does not take into account the total number of images, yielding the same score for data sets D1 and D2. The \acf{FROC} curve plots the average number of \acf{FPPI} against the Sensitivity, therefore accounting for the number of images. The \ac{FPPI} is lower for D2, yielding a higher \ac{FROC} score.}
     \label{fig:ap-froc} 
\end{figure}

\begin{figure}[H]
    \centering
    \includegraphics[width=1\linewidth]{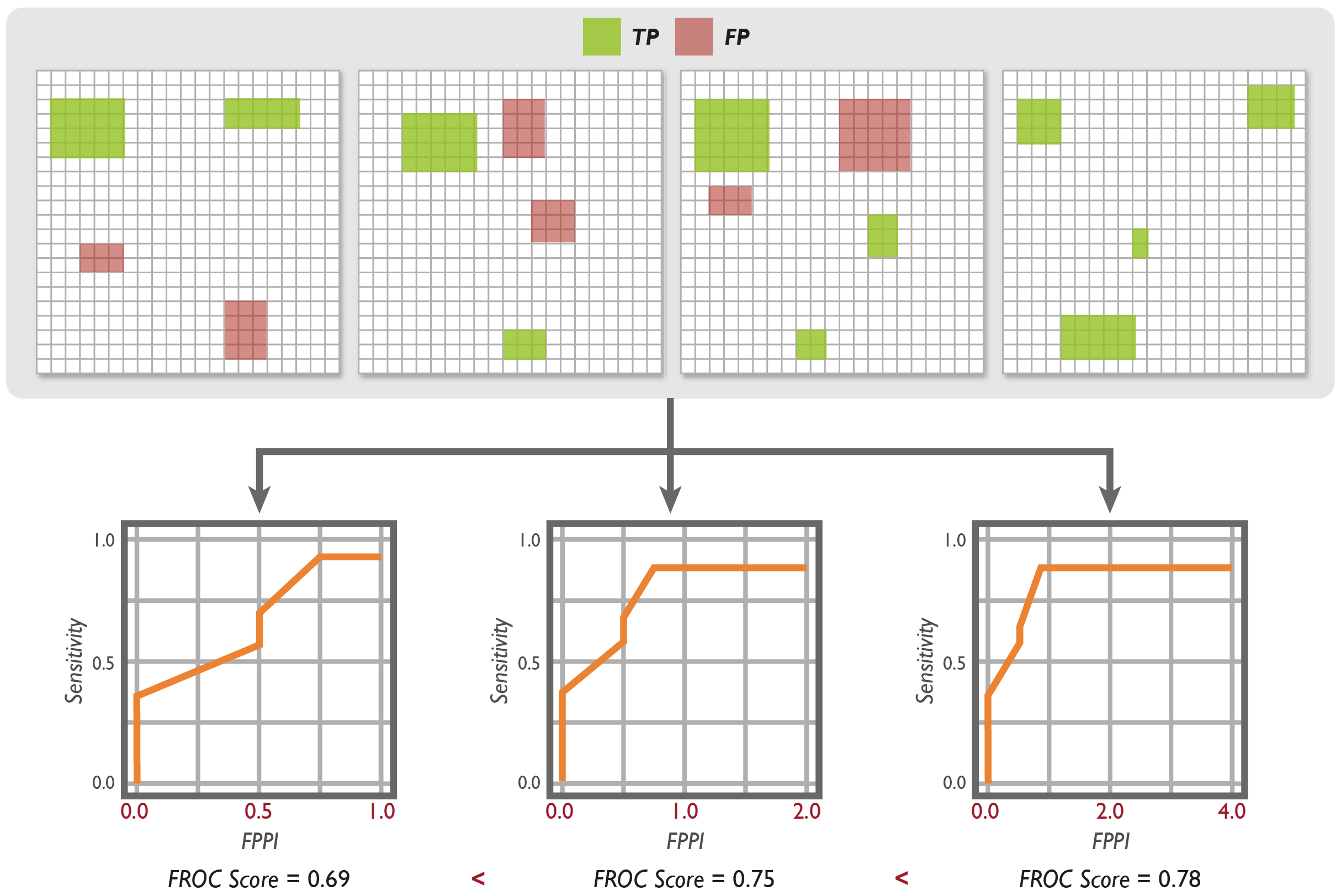}
    \caption{Effect of defining different ranges for the \acf{FPPI} used to draw the \acf{FROC} curve for the same prediction (top). The resulting \ac{FROC} Scores change for different boundaries of the x-axis.}
    \label{fig:froc-no-standard}
\end{figure}

% "Use ranged version if required" WHERE?
\newpage
\hfill\\\noindent\textbf{DG4.2: \acf{AP} versus \acf{FROC} Score}
\begin{tcolorbox}[title= Summary of DG4.2: \acs{AP} versus \acs{FROC} Score, colback=white]
    \begin{table}[H]
        \centering
        \begin{tabular}{p{5.5cm} p{0.3cm} p{5cm}}
            \textbf{\ac{AP}} & & \textbf{\ac{FROC} Score} \\ 
            \textcolor{gray}{\faArrowCircleRight }  Standard metric in computer vision community && \textcolor{gray}{\faArrowCircleRight }  Preference in clinical context due to its domain-centered approach\\
            \textcolor{gray}{\faArrowCircleRight } Unawareness of data set sizes && \textcolor{gray}{\faArrowCircleRight } Consideration of data set sizes \\
            \textcolor{red}{\faMinusCircle } For filtering low confidence predictions, a cutoff on confidence scores is required && \textcolor{ODgreen}{\faPlusCircle } No consideration of low-confidence predictions \\
            \textcolor{ODgreen}{\faPlusCircle } Relatively good standardization of hyperparameters && \textcolor{red}{\faMinusCircle } Lack of standardization\\
        \end{tabular}
        \caption{Comparison of \acf{AP} and \acf{FROC} Score in the context of the decision guide DG4.2 for Subprocess S4. Context: availability of predicted class scores (FP5.1 = TRUE) and FP1.1 = object detection (ObD) or instance segmentation (InS).}
        \label{tab:d4-2}
    \end{table}
\end{tcolorbox}
% \textcolor{red}{\faMinusCircle }
% \textcolor{ODgreen}{\faPlusCircle }
% \textcolor{gray}{\faArrowCircleRight }
The following aspects should be taken into account when deciding between \ac{AP} (Fig.~\ref{fig:cheat-sheet-ap}) and \ac{FROC} Score (Fig.~\ref{fig:cheat-sheet-froc}):
    \begin{itemize}
    \item \textbf{Community preferences:} While \ac{AP} constitutes the undisputed standard metric for object detection and instance segmentation in the computer vision community, the \ac{FROC} Score is often favoured in the clinical context due to its easier interpretability despite its lack of standardization (employed \ac{FPPI} Scores vary across studies \cite{jin2020deep,bejnordi2017diagnostic,setio2017validation}). Thus, the decision between the two metrics often boils down to a decision between a standardized and technical validation versus an interpretable and application-focused validation. 
    \item \textbf{Data set size awareness:} In its default configuration, where \ac{AP} is computed globally over the entire data set, the metric is insensitive to performance per individual images. In contrast, the \ac{FROC} Score takes into account the total number of images in the data set (see also Fig.~\ref{fig:ap-froc}). For example, given a data set and an \ac{AP} as well as \ac{FROC} score computed for this data set, adding more 'empty' images (i.e., images with no reference objects and no model predictions) would lead to an improved \ac{FROC} score, because \ac{FROC} rewards the model for correctly not predicting structures on these images. In contrast, the \ac{AP} score would not be affected, because globally no new \acp{TP}, \acp{FP} or \acp{FN} are added that would alter the metric score. This property does not affect relative method comparison and can be related to the underlying question "at which scale are matched objects (cardinalities) aggregated/counted?". As described in Suppl. Note~\ref{app:od}, \ac{AP} can alternatively be configured to aggregate scores per individual image, in which case the total number of images in the data set is considered equally to \ac{FROC}. \citep{reinke2021common} demonstrates how to apply \ac{AP} and other object detection metrics to, for example, clinical scenarios requiring per-image aggregation. \ac{FROC} score is a hybrid metric in this context, where \textit{Sensitivity} is computed per data set while \acp{FP} are averaged over single images (\ac{FPPI}).  
    \item \textbf{Dealing with low-confidence predictions:} It is often desired to filter low confidence predictions (e.g., objects with high confidence of being background) prior to metric computation. For \ac{AP} computation, this requires a cutoff on the confidence score or upper limits of considered predictions per image or per data set. For \ac{FROC}, however, with typical values of \ac{FPPI}, such low-confidence predictions naturally go unconsidered, thus allowing to avoid additional filtering measures.
    \item \textbf{\ac{FPPI}:} Different \ac{FPPI} values are used in the field for computing the FROC Score, yielding non-standardized results (see Fig.~\ref{fig:froc-no-standard}). A potential default are the values 1/8, 1/4, 1/2, 1, 2, 4, 8, as used for multiple popular benchmarks \cite{van2010comparing,setio2017validation}. Here, lower \ac{FPPI} values (smaller than one) are weighted equally to higher \ac{FPPI} values (greater than 1; four values each). Deviation from this weighting might be appropriate depending on the application, but should be explained. In the biostatistics community, areas under the curve are sometimes computed constraining the \ac{FPPI} range to [0,1] \cite{samuelson2006comparing}.
\end{itemize}

%\newpage

\newpage

%-----------------------------------

\newpage
\subsubsection{Decision guide S5}

\label{ssec:dg5}
\hfill\\\noindent\textbf{DG5.1: \acf{KCE} versus \acf{ECEKDE}}
\begin{tcolorbox}[title= Summary of DG5.1: \acs{KCE} versus \acs{ECEKDE}, colback=white]
    \begin{table}[H]
        \centering
        \begin{tabular}{p{5.5cm} p{0.3cm} p{5cm}}
            \textbf{\ac{KCE}} & & \textbf{\ac{ECEKDE}} \\ 
\textcolor{ODgreen}{\faPlusCircle } Capture of isolated calibration quality && \textcolor{ODgreen}{\faPlusCircle } Capture of isolated calibration quality\\
    \textcolor{gray}{\faArrowCircleRight } Unbiased estimator of canonical calibration error based on an alternative distance function && \textcolor{gray}{\faArrowCircleRight } Potentially biased estimator of an $\ell_p$ canonical calibration error (bias might be rendered neglectable by future de-biasing schemes) \\
         \textcolor{red}{\faMinusCircle } Bad interpretability, also due to negative output values && \textcolor{ODgreen}{\faPlusCircle } Straightforward interpretability of relative improvement\\
           \textcolor{gray}{\faArrowCircleRight } Recent proposition, not widely used && \textcolor{gray}{\faArrowCircleRight } Recent proposition, not widely used\\      
          \textcolor{red}{\faMinusCircle } Depends on nontrivial configuration choices of kernels and associated hyperparameters  && \\
        \end{tabular}
        \caption{Comparison of \acf{KCE} and \acf{ECEKDE} in the context of the decision guide DG5.1 for Subprocess S5. Context: FP2.7.2 = U2 - comparison of calibration performance across classifiers on the same task requested and no mismatch between class prevalences and class importance (F2.5.3 = FALSE).}
        \label{tab:d5-1}
    \end{table}
\end{tcolorbox}

The context for this decision guide between \ac{KCE} (Fig.~\ref{fig:cheat-sheet-kce}) and \ac{ECEKDE} (Fig.~\ref{fig:cheat-sheet-ecekde}) is use case 2 (U2) in Fig.~\ref{fig:calibration-comparison-vs-interpretation}: "comparing the calibration quality across multiple classifiers on the same task."

\textbf{General differences:} Both \ac{KCE} and \ac{ECEKDE} are estimators of a canonical calibration error, but measure this error based on different divergences, i.e., distance functions: \ac{ECEKDE} is based on the $\ell_p$ norm and thus straightforward to interpret and configure. In contrast, \ac{KCE} is based on the “maximum mean discrepancy” and thus not interpretable (it may even take on negative values) and requires nontrivial configuration of kernels as well as associated hyperparameters. On the other hand, $\ell_p$ norm estimators such as \ac{ECEKDE} are inherently biased while \ac{KCE} is an unbiased estimator. Arguably, in the context of this decision guide (U2), interpretability of the calibration error estimate is not required, since only a comparative, or relative assessment is requested rendering the unbiased \ac{KCE} the intuitive choice. However, recent research on $\ell_p$ norm estimators presents effective de-biasing schemes~\cite{popordanoska2022consistent}, which might render the resulting bias neglectable in the near future and thus make $\ell_p$ estimators such as  \ac{ECEKDE} a viable alternative for comparative calibration assessment.

\textbf{Popularity:} Calibration error estimates \ac{KCE} and \ac{ECEKDE} are both recently proposed measures that are not widely known in the biomedical community.

\newpage
\textbf{DG5.2: \acf{BS} versus \acf{KCE} versus \acf{ECEKDE}}
\begin{tcolorbox}[title= Summary of DG5.2: \acs{BS} versus \acs{KCE} versus \acs{ECEKDE}, colback=white]
    \begin{table}[H]
        \centering
        \begin{tabular}{p{3cm} p{0.3cm} p{3cm} p{0.3cm} p{3cm}}
            \textbf{\ac{BS}} & & \textbf{\ac{KCE}} & & \textbf{\ac{ECEKDE}} \\ 
          \textcolor{gray}{\faArrowCircleRight }
 Capture of effects of (re-) calibration methods on discrimination performance in addition to calibration quality  && \textcolor{gray}{\faArrowCircleRight } Capture of isolated calibration quality && \textcolor{gray}{\faArrowCircleRight } Capture of isolated calibration quality\\
     \textcolor{ODgreen}{\faPlusCircle } Unbiased measure of an $\ell_2$ norm canonical calibration error && \textcolor{gray}{\faArrowCircleRight } Unbiased estimator of canonical calibration error based on an alternative distance function && \textcolor{gray}{\faArrowCircleRight } Potentially biased estimator of an $\ell_p$ canonical calibration error (bias might be rendered neglectable by future de-biasing schemes) \\
          \textcolor{ODgreen}{\faPlusCircle } 
 Straightforward interpretability of relative improvement && \textcolor{red}{\faMinusCircle } Bad interpretability, also due to negative output values && \textcolor{ODgreen}{\faPlusCircle } Straightforward interpretability of relative improvement\\
           \textcolor{ODgreen}{\faPlusCircle } Established statistical concept with long history of applications in many fields of research && \textcolor{gray}{\faArrowCircleRight } Recent proposition, not widely used && \textcolor{gray}{\faArrowCircleRight } Recent proposition, not widely used\\    
           && \textcolor{red}{\faMinusCircle } Depends on nontrivial configuration choices of kernels and associated hyperparameters && \\
        \end{tabular}
        \caption{Comparison of \acf{BS}, \acf{KCE} and \acf{ECEKDE} in the context of the decision guide DG5.2 for Subprocess S5. Context: U1 - FP2.7.2 = comparison of re-calibration methods for the same classifier requested.}
        \label{tab:d5-2}
    \end{table}
\end{tcolorbox}

The context for this decision guide between \ac{BS} (Fig.~\ref{fig:cheat-sheet-bs}), \ac{ECEKDE} (Fig.~\ref{fig:cheat-sheet-ecekde}), and \ac{KCE} (Fig.~\ref{fig:cheat-sheet-kce}) is use case 1 (U1) in Fig.~\ref{fig:calibration-comparison-vs-interpretation}: "comparing the effect of one or more re-calibration methods on the same (fixed) classifier."

\textbf{General differences:} 
\ac{BS} can be decomposed into discrimination and calibration terms, where the calibration term exactly resembles the canonical calibration error (as defined in Suppl. Note~\ref{app:calibration}). As the purpose of the metric in the provided context is to assess the performance of different re-calibration methods for the same classifier, a higher \ac{BS} score also implies a better calibration in the case of \textit{accuracy-preserving} calibration methods.
As a major difference to \ac{BS}, \ac{KCE} estimates the canonical calibration error directly. While this estimation is not biased (i.e., it is not dependent on the data set size), the resulting estimates are not interpretable, that is, they only allow for relative comparison on the same task (equivalently to \ac{BS}). Further, \ac{KCE} requires nontrivial configuration of kernels as well as associated hyperparameters. In contrast to \ac{KCE}, current estimators of $\ell_p$ calibration error are biased, but are highly interpretable and straightforward to configure. Moreover, recent developments in this line of research present effective de-biasing schemes~\cite{popordanoska2022consistent}, which might render the resulting bias neglectable in the near future and thus make $\ell_p$ estimators such as \ac{ECEKDE} a viable alternative also for comparative calibration assessment.

\textbf{Applicability:} Generally, \ac{BS} is attractive for ranking re-calibration methods that are guaranteed to be accuracy-preserving (such as the common temperature scaling~\cite{guoCalibrationModernNeural2017}). Otherwise, the metric must be applied with care, because altered discrimination performance will dilute the focus on calibration quality in the ranking. 
Note that it may also be desirable to capture the effect of (non-accuracy-preserving) re-calibration methods on the discrimination performance. In such cases of comprehensive assessment of re-calibration methods, it is also appropriate to apply \ac{BS}.
%is applied and these (advantageous or disadvantageous) effects on discrimination performance are to be taken into account for overall assessment of such methods, \ac{BS} is the correct choice (note that this use case can also be interpreted as U3 in Fig.~\ref{fig:calibration-comparison-vs-interpretation} and FP2.7.2). 
In contrast to \ac{BS}, calibration error estimators such as \ac{KCE} and \ac{ECEKDE} are capable of comparing the calibration error of re-calibration while being agnostic to potential changes of discrimination performance caused by the transformations. For the provided use case, this property allows the ranking of non-accuracy-preserving transformations, such as recently proposed techniques employing spline interpolations~\cite{gupta2020calibration} or Gaussian processes~\cite{wenger2020non}, purely according to their calibration error while ignoring their effects on the discrimination performance. 
%Because the classifier is changed by these transformations, the described use case might also be interpreted as U2 in Fig.~\ref{fig:calibration-comparison-vs-interpretation}.

\textbf{Interpretability:} Defined as the root mean square error between predictions and references, \ac{BS} is bounded between [0, 1] and therefore straightforward to interpret as an overall measure. However, as the calibration error is not isolated and scores are still conflated with the (same fixed) discrimination performance, only a relative comparison of calibration errors is possible. \ac{KCE} is generally hard to interpret, also because it can yield negative values. \ac{ECEKDE} as an estimator of $\ell_p$ calibration error is straightforward to interpret.

\textbf{Popularity}: \ac{BS} is a widely known metric for overall performance measures with a long history of usage. Calibration error estimates \ac{KCE} and \ac{ECEKDE} are  both recently proposed measures and not widely known in the biomedical community.

\textbf{Reasons to not recommend \ac{NLL} in this context:} \ac{NLL} essentially assesses a weighted version of the canonical calibration error as the logarithm leads to heavy penalization of tail probabilities. As the implications of this weighting on calibration assessment (as opposed to the overall performance measure) are not intuitive, we generally do not recommend \ac{NLL} in this use case.\\

\newpage
\noindent\textbf{DG5.3: \acf{BS} versus \acf{NLL}}
\begin{tcolorbox}[title= Summary of DG5.3: \acs{BS} versus \acs{NLL}, colback=white]
    \begin{table}[H]
        \centering
        \begin{tabular}{p{5cm} p{0.3cm} p{5cm}}
            \textbf{\ac{BS}} & & \textbf{\ac{NLL}} \\ 
           \textcolor{red}{\faMinusCircle } Bounded penalization of errors leads to preference of naive systems in imbalanced settings  && \textcolor{gray}{\faArrowCircleRight } Heavy penalization of extreme scores (close to 0 or 1), thus ability to capture missing rare events. General preference of conservative models \\
           \textcolor{ODgreen}{\faPlusCircle } Straightforward interpretability as the mean squared error && \textcolor{gray}{\faArrowCircleRight }
Difficult interpretability due to lack of upper bound\\
           \textcolor{ODgreen}{\faPlusCircle } Established statistical concept with long history of applications in many fields of research && \textcolor{ODgreen}{\faPlusCircle } Established statistical concept with long history of applications in many fields of research\\
        \end{tabular}
        \caption{Comparison of \acf{BS} and \acf{NLL} in the context of the decision guide DG5.3 for Subprocess S5. Context: FP2.7.2 = U3 - comparison of overall performance across classifiers requested.}
        \label{tab:d5-3}
    \end{table}
\end{tcolorbox}

The context for this decision guide between \ac{BS} and \ac{NLL} is use case 3 (U3) in Fig.~\ref{fig:calibration-comparison-vs-interpretation}: "overall performance measure requested." Both \ac{BS} (Fig.~\ref{fig:cheat-sheet-bs}) and \ac{NLL} (Fig.~\ref{fig:cheat-sheet-nll}) are overall performance measures, which capture discrimination and canonical calibration in a single score.

\textbf{Penalization of errors}: Like Accuracy, \ac{BS} penalizes errors of all events equivalently irrespective of the class prevalence. This implies that scores may drastically change when the prevalence changes and thus renders BS a highly prevalence-dependent metric. For instance, in imbalanced scenarios, a naive system that simply predicts the dominant class can receive a high \ac{BS}, similarly to a high Accuracy score. One strategy to cope with this is to divide the \ac{BS} by the \ac{BS} achieved with a naive system, resulting in the normalized variant \acf{BSS}. Equivalently to \ac{ECN}, this transformation is a rescaling of scores to establish a 'naive baseline' and enhance interpretability, but errors are still penalized equivalently irrespective of class prevalence. In other words, equal importance of classes (FP2.5.1) is not reflected in the metric, and missing a frequent event is still as heavily penalized as missing a rare event although missing a rare event has a greater effect on the respective class sensitivity. This results in a strict interpretation where the total amount of errors has to be lower than the number of events in the rare class in order for a system to be considered 'better than random'.

Compared to squared error penalization in \ac{BS}, the logarithm introduces a stronger penalization of tail probabilities~\cite{quinonero2005evaluating}. In consequence, overconfident predictions (such as a score of 1, implying scores of 0 in the other classes) lead to higher losses. For example, predicting 0.001 rather than 0.01 (when the true class is '1') increases BS by $\approx 2\%$ and \ac{NLL} by $ \approx 230\%$ (for this single entry). A practical effect of this penalty is a naturally higher penalization of naive systems in class imbalance scenarios, addressing the pitfall of \ac{BS} above. \ac{NLL} is thus of potential  interest in scenarios with high class imbalance, where missing rare events would be heavily penalized, compared to \ac{BS} which is prone to favoring naive systems. Generally, the penalization effect can also be described as \ac{NLL} favoring more conservative models that avoid predictions of extreme class scores.

\textbf{Interpretability}: \ac{BS} is relatively straightforward to interpret as the mean squared error between predictions and the reference. The resulting scores are bounded ($[0,1]$). NLL is arguably harder to interpret featuring logarithmic penalization of errors and thus no upper bound of the resulting score (bounds: $[0, \infty]$)

\textbf{Popularity}: Both metrics are common statistical concepts and come with a long history of usage in many fields of research.

%TABLE
%BS: highly prevalence-dependent VS Natural down-weighting of dominant classes and focus on detecting rare events. No heavy penalization of tail probabilities thus prone to yielding high scores for naive systemsTODO VS favors conservative models due to strong penalization of tail probabilities
%Relatively straightforward to interpret with bounds [0,1]  VS Hard to interpret without upperlower bound (-infinity, 0, infinity]
%Both widely used

\hfill\\\noindent\textbf{DG5.4: \acf{ECE}/ \acf{RBS} versus \acf{CWCE} versus \acf{ECEKDE}/ \acf{CWCE}/ \acf{RBS}} 

The decision between the sets of metrics boils down to determining whether predicted class scores should be tested for top-label calibration (as measured by \ac{ECE}, Fig.~\ref{fig:cheat-sheet-ece}), marginal calibration (as measured by \ac{CWCE}, Fig.~\ref{fig:cheat-sheet-cwce}), or canonical calibration (as measured by \ac{ECEKDE}, Fig.~\ref{fig:cheat-sheet-ecekde}. If there is an unequal interest across classes (FP2.5.1), \ac{CWCE} is the natural choice. In this case we recommend both per-class and weighted reporting (by class importance). Note that only aggregated reporting comes with the pitfall of unstable results, specifically in the case of few samples or many classes. In the case of equal interest across classes, the key question is whether the task interest is limited to the predicted scores that lead to the classification decision (top-label) or whether there are reasons to request all predicted scores to be calibrated.

Notably, in binary classification tasks, the two conditions are equivalent~\cite{vaicenavicius2019evaluating}.

\textbf{Reasons for and against focusing on top-label calibration (\ac{ECE}):} The task interest focuses on the decisions made by the classifier and only lies in the  probabilities of the resulting decisions. In case the underlying biomedical research question has a dedicated focus on the decision process, top-label error might be the right choice, because it directly reflects this focus. Conflating the calibration of decisions with other probabilities might be interpreted as washing out the task focus in this case. Although it is common practice to assess calibration quality with \ac{ECE}, this approach comes with various pitfalls. Importantly, it is often ignored that top-label calibration implies an argmax decision rule based on the predicted class scores, which is often not an optimal decision rule as discussed in Suppl. Note~\ref{app:general-instructions} ('decision rule on predicted class scores'; see Fig.~\ref{fig:top-label-pitfall}). Caution should also be exercised if there is a mismatch between class prevalences and class importance (FP2.5.3) as the top-label calibration is highly biased towards the high-prevalence classes. Furthermore, \ac{ECE} commonly relies on binning of class scores, which introduces a dependency of the resulting metric score on the specific binning scheme.
%~\cite{gruber2022better}.
The number of bins is a configuration parameter that should by no means be optimized on the final validation data.
Note in this context that binning has been shown to result in a more biased estimation compared to density estimation methods~\cite{popordanoska2022consistent}. 

\textbf{Reasons to extend the focus to all predicted scores (\ac{ECEKDE} and \ac{CWCE}):} A common perception is that the canonical calibration condition, which is the strongest condition considering all predicted class scores, is the appropriate one in many application scenarios~\cite{gruber2022better,popordanoska2022consistent,ferrer2022analysis}. One reason lies in the limitations of top-label calibration and associated binning estimators described above. Another reason could be a broad task interest in all predictions beyond the classification decision. In the clinical context, for instance, the risk for all potential outcomes might be relevant for further treatment or shall be communicated to the patient. In such scenarios, calibration of all probabilities might be of interest. Consider, for instance, a multi-way classification of tumor categories, where one category is more aggressive than others. Even though the final prediction of the system is 'benign lesion', it might be of clinical interest to know (and communicate to the patient) whether the probability for this outcome was $5\%$ or $20\%$. While the primary calibration metric for such scenarios should be \ac{ECEKDE} as an estimate of the canonical calibration, it might be of interest to additionally report marginal calibration (as measured by \ac{CWCE}) separately for each class. Notably, for these scenarios, alternatively splitting the problem into individual domain questions that result in separate traversals for each class of interest should be considered (see Suppl. Note~\ref{app:general-instructions}).

\textbf{Additional reporting of \ac{RBS} (Fig.~\ref{fig:cheat-sheet-rbs}) as a guaranteed upper bound on the calibration error:} In top-label and canonical calibration, we recommend the additional reporting of \ac{RBS} as a guaranteed upper bound on the calibration error. As popular methods to assess calibration quality such as \ac{ECE} or \ac{ECEKDE} are known to over- or underestimate the error~\cite{gruber2022better}, this guarantee provides additional information, especially in safety-critical applications where the calibration error must not be underestimated.

%\newpage
%-----------------------------------
\subsubsection{Decision guide S6}
\label{ssec:dg6}
%-----------------------------------

\hfill\\\noindent\textbf{DG6.1: \acf{DSC} versus \acf{IoU}} 
\begin{tcolorbox}[title= Summary of DG6.1: \acs{DSC} versus \acs{IoU}, colback=white]
    \begin{table}[H]
        \centering
        \begin{tabular}{p{5.3cm} p{0.3cm} p{6cm}}
            \textbf{\ac{DSC}} & & \textbf{\ac{IoU}} \\ 
            \textcolor{gray}{\faArrowCircleRight } Identical to F$_\text{1}$ Score & & \textcolor{gray}{\faArrowCircleRight } Identical to Jaccard Index \\
            \textcolor{gray}{\faArrowCircleRight } Close relation to \ac{IoU} (see Eq.~\ref{eq:dsc-iou}) & & \textcolor{gray}{\faArrowCircleRight } Close relation to \ac{DSC} (see Eq.~\ref{eq:iou-dsc}) \\
            \textcolor{gray}{\faArrowCircleRight } Preference in medical community & & \textcolor{gray}{\faArrowCircleRight } Preference in computer vision community\\
        \end{tabular}
        \caption{Comparison of \acf{DSC} and \acf{IoU} in the context of the decision guide DG6.1 for Subprocess S6. Context: no exclusive interest in the center line of structures (FP2.3 = FALSE, FP3.3 = FALSE) and equal severity of class confusions (FP2.5.2 = FALSE).}
        \label{tab:d6-1}
    \end{table}
\end{tcolorbox}
The \ac{DSC} (Fig.~\ref{fig:cheat-sheet-dsc}) is identical to the F$_1$ Score on pixel level and closely related to the \ac{IoU} (Fig.~\ref{fig:cheat-sheet-iou}), which, in turn, is identical to the Jaccard Index (see equations~\ref{eq:iou-dsc} and~\ref{eq:dsc-iou}). The two metrics will yield the same ranking of aggregated metric values in most applications (theoretically, deviations are possible), such that there is no value in combining them. Commonly, the computer vision community prefers the \ac{IoU}, while the medical image community favors the \ac{DSC}.

\noindent\hspace{-1cm}\begin{minipage}{.5\linewidth}
\begin{equation}
\label{eq:iou-dsc}
\text{\text{IoU}} = \frac{\text{\text{DSC}}}{2-\text{\text{DSC}}}
\end{equation}
\end{minipage}
\begin{minipage}{.5\linewidth}
\begin{equation}
\label{eq:dsc-iou}
\text{\text{DSC}} = \frac{2 \text{\text{IoU}}}{1+\text{\text{IoU}}}
\end{equation}
\end{minipage}

\noindent\textbf{DG6.2: How to determine $\beta$ in F$_\beta$ Score} \\ \\
% \begin{tcolorbox}[title= $\beta$ in F$_\beta$ Score, colback=white]
%     \begin{table}[H]
%         \centering
%         \begin{tabular}{p{3cm} p{0.3cm} p{3cm} p{0.3cm} p{3cm}}
%             \textbf{$\beta$ < 1} & & \textbf{$\beta$ = 1} & & \textbf{$\beta$ > 1} \\ 
%             \textcolor{gray}{\faArrowCircleRight } Higher weights on \ac{FP} penalties (oversegmentation) & & \textcolor{gray}{\faArrowCircleRight } Higher weights on \ac{FP} penalties (oversegmentation) & & \textcolor{gray}{\faArrowCircleRight } Higher weights on \ac{FN} penalties (undersegmentation)\\
%         \end{tabular}
%         \caption{Caption}
%         \label{tab:d6-2}
%     \end{table}
% \end{tcolorbox}
The F$_\beta$ Score (Fig.~\ref{fig:cheat-sheet-fbeta}) is defined as:
\begin{equation}
F_\beta = (1 + \beta^2) \cdot \frac{\ac{PPV} \cdot \text{Sensitivity}}{(\beta^2 \cdot \ac{PPV}) + \text{Sensitivity}} = \frac{(1 + \beta^2) \cdot \ac{TP}}{(1 + \beta^2) \cdot \ac{TP} + \beta^2 \cdot \ac{FN} + \ac{FP}} 
\end{equation}
The most common choice is to set $\beta$ to $1$, resulting in equal weighting of Sensitivity (Fig.~\ref{fig:cheat-sheet-sensitivity}) and \ac{PPV} (Fig.~\ref{fig:cheat-sheet-ppv}). Higher values of $\beta$ result in higher weights on \ac{FN} penalties (undersegmentation in segmentation problems) compared to \ac{FP} penalties (oversegmentation) and thus imply a focus on Sensitivity compared to \ac{PPV}.

\newpage
%-----------------------------------
\subsubsection{Decision guide S7}
\label{ssec:dg7}
%-----------------------------------

\hfill\\\noindent\textbf{DG7.1: \acf{NSD} versus Boundary \acf{IoU}}
\begin{tcolorbox}[title= Summary of DG7.1: \acs{NSD} versus Boundary \acs{IoU}, colback=white]
    \begin{table}[H]
        \centering
        \begin{tabular}{p{5.5cm} p{0.5cm} p{5.5cm}}
            \textbf{\ac{NSD}} & & \textbf{Boundary \ac{IoU}} \\ 
            \textcolor{gray}{\faArrowCircleRight } Accounting for noisy images, limited resolution or imprecise reference annotations & & \textcolor{gray}{\faArrowCircleRight } Measurement of overlap between predicted and reference contours up to certain width\\
             \textcolor{gray}{\faArrowCircleRight } Influence of hyperparameter on scores: distances below tolerance threshold are considered \ac{TP} && \textcolor{gray}{\faArrowCircleRight } Influence of hyperparameter on scores: distance parameter determines thickness of the considered boundary \\ 
        \end{tabular}
        \caption{Comparison of \acf{NSD} and Boundary \acf{IoU} in the context of the decision guide DG7.1 for Subprocess S7. Context: possibility of spatial outliers in the reference annotation (FP4.3.2 = TRUE) or, if FALSE, FP2.5.6 = existence-based penalization of outliers.}
        \label{tab:d7-1}
    \end{table}
\end{tcolorbox}

The following aspects should be considered when deciding between \ac{NSD} (Fig.~\ref{fig:cheat-sheet-nsd}) and Boundary \ac{IoU} (Fig.~\ref{fig:cheat-sheet-boundary-iou}):

\begin{itemize}
    \item \textbf{Different research questions:} Both metrics set the focus on the boundary/contour of structures, but fundamentally differ in what they measure: \ac{NSD} measures the \ac{DSC} score on the surface voxels (often interpreted as the ratio of correctly predicted contour), where the strictness for what constitutes a correct boundary is controlled by a tolerance parameter. This way, noise in the image, limited resolution, or imprecise reference annotations can be accounted for. Boundary \ac{IoU} directly measures the overlap between predicted and reference contours (without tolerance) up to a certain width (which is controlled by a width parameter). Thus, \ac{NSD} is preferable if a tolerance accounting for imprecise annotations is requested. Boundary \ac{IoU}, on the other hand, is preferable if contour errors are thought of as crucial inconsistencies that should be assessed, or if a wider area around the contour line is of interest (dynamic transition to the classical \ac{IoU}).
    \item \textbf{Setting the hyperparameter:} The \ac{NSD} and Boundary \ac{IoU} both require users to manually set a hyperparameter. \textbf{\ac{NSD}:} Boundary distances below the tolerance threshold will be considered \ac{TP} (deviations do not count as errors). This parameter can be set according to the inter-rater variability or, if not available, heuristics. \textbf{Boundary \ac{IoU}:} The distance parameter determines the thickness of the considered boundary and thus also influences the sensitivity to contour errors (the smaller the distance, the higher the sensitivity). This parameter can also be set according to the inter-rater variability (here in order to capture potential inconsistencies, as opposed to disregarding noise as in \ac{NSD}) or, if not available, heuristics.
\end{itemize}
%\faArrowCircleRight

\begin{figure}[H]
    \centering
    \hspace{-0.5cm}
    \includegraphics[width=0.75\linewidth]{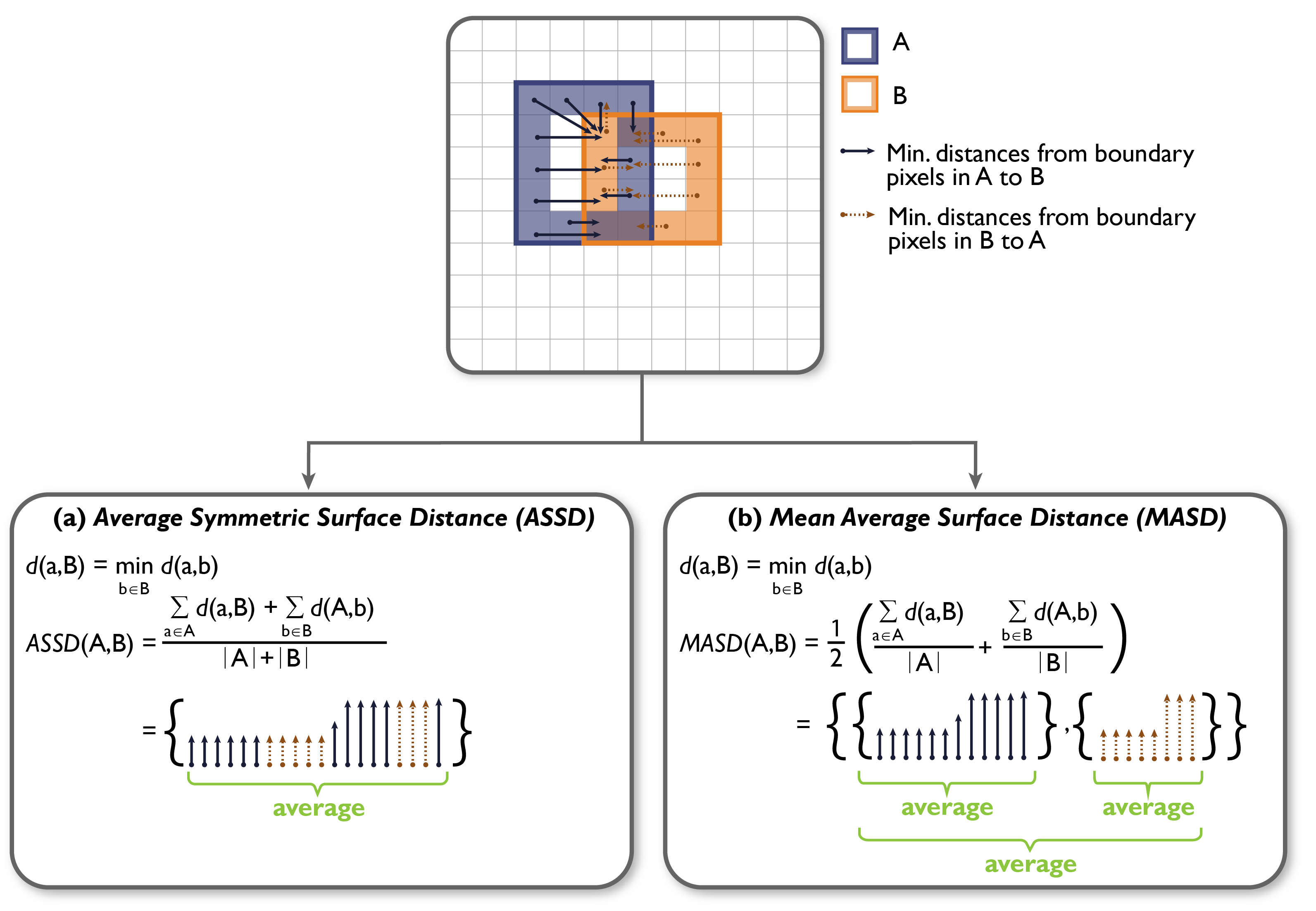}
    \caption{Most commonly used distance-based segmentation metrics: \textbf{(a)} the \acf{ASSD} and \textbf{(b)} the \acf{MASD}. The term $d(a,b)$ denotes the Euclidean distance between boundary pixels $a$ and $b$. Only the \acfp{TP} are considered.}
    \label{fig:definition-distance-assd-masd}
\end{figure}

\begin{figure}[H]
    \centering
    \includegraphics[width=0.6\linewidth]{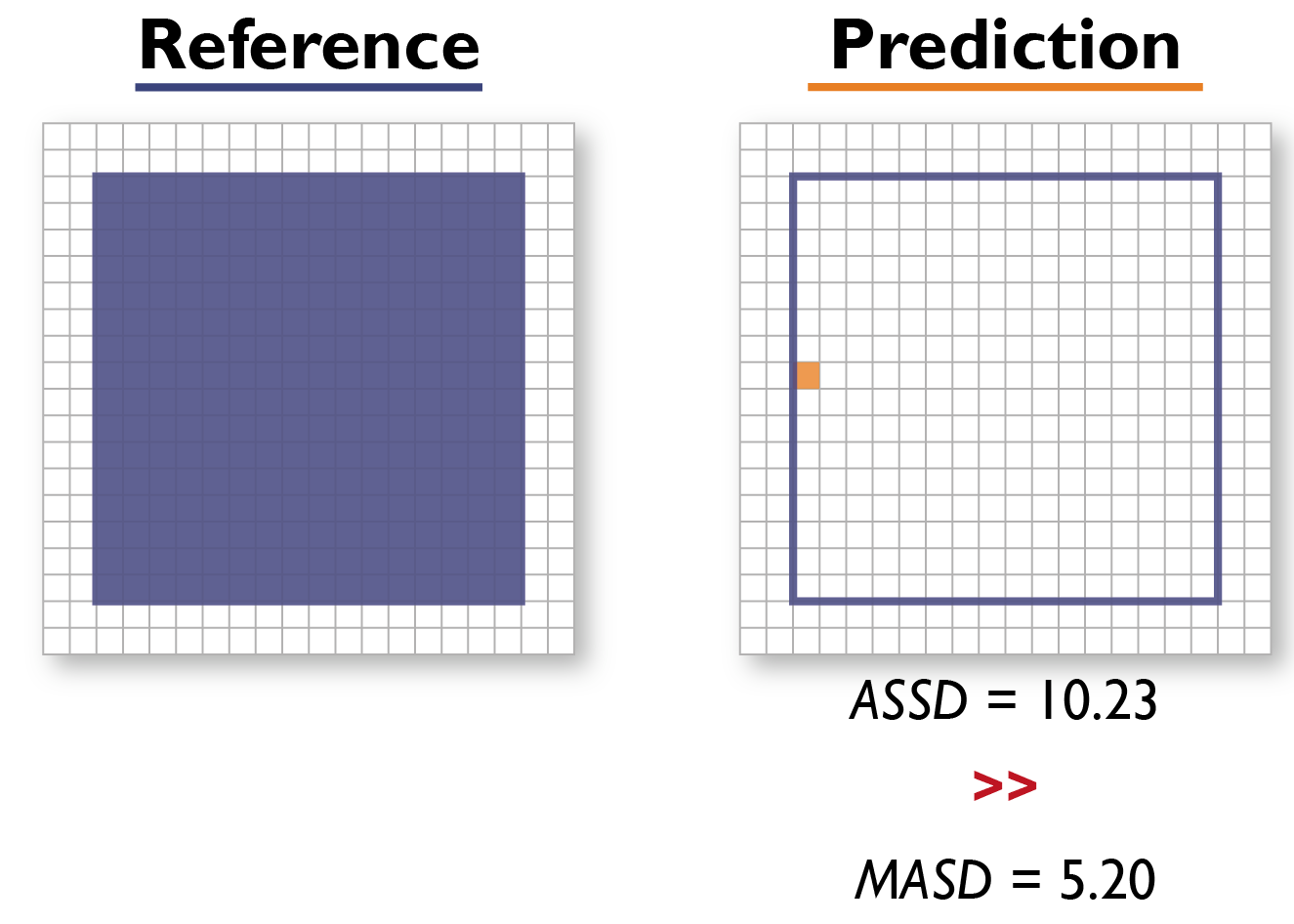}
    \caption{Corner case in which \acf{MASD} yields an undesired result. If the \textit{Prediction} is very small (here: one pixel) and located close to the reference boundary, the \acf{MASD} will be much lower compared to the \acf{ASSD}.}
     \label{fig:masd-assd}
\end{figure}

\newpage
\noindent\textbf{DG7.2: \acf{MASD} versus \acf{ASSD}}
\begin{tcolorbox}[title= Summary of DG7.2: \acs{MASD} versus \acs{ASSD}, colback=white]
    \begin{table}[H]
        \centering
        \begin{tabular}{p{5.5cm} p{0.5cm} p{5.5cm}}
            \textbf{\ac{MASD}} & & \textbf{\ac{ASSD}} \\ 
            \textcolor{gray}{\faArrowCircleRight } Equal contribution of reference and prediction boundaries to the metric score & & \textcolor{red}{\faMinusCircle } Domination of the metric score by the larger contour\\
            \textcolor{red}{\faMinusCircle } Possibility of misleading results in corner cases (e.g., tiny prediction closely located to the reference)& & \\
        \end{tabular}
        \caption{Comparison of \acf{MASD} and \acf{ASSD}  in the context of the decision guide DG7.2 for Subprocess S7. Context: FP2.5.6 = distance-based penalization of outliers with contour focus.}
        \label{tab:d7-2}
    \end{table}
\end{tcolorbox}
The \ac{ASSD} (Fig.~\ref{fig:cheat-sheet-assd}) puts all boundary distances (all distances from boundary A to boundary B and all distances from boundary B to boundary A) in a list, then takes the mean (Fig.~\ref{fig:definition-distance-assd-masd}). Thus, if one boundary is much larger than the other, this boundary will impact the mean much more. 
The \ac{MASD} (Fig.~\ref{fig:cheat-sheet-masd}) computes the sum of the mean distances from boundary A to boundary B and the mean distances from boundary B to boundary A. Therefore, the reference and prediction boundaries contribute equally (see Fig.~\ref{fig:definition-distance-assd-masd}). While there are corner cases in which \ac{MASD} features disadvantages compared to \ac{ASSD} as well (see Fig.~\ref{fig:masd-assd}), we generally recommend \ac{MASD} because of the aforementioned advantage. 

%\newpage

%\newpage

\begin{figure}[H]
    \centering
    \includegraphics[width=0.9\linewidth]{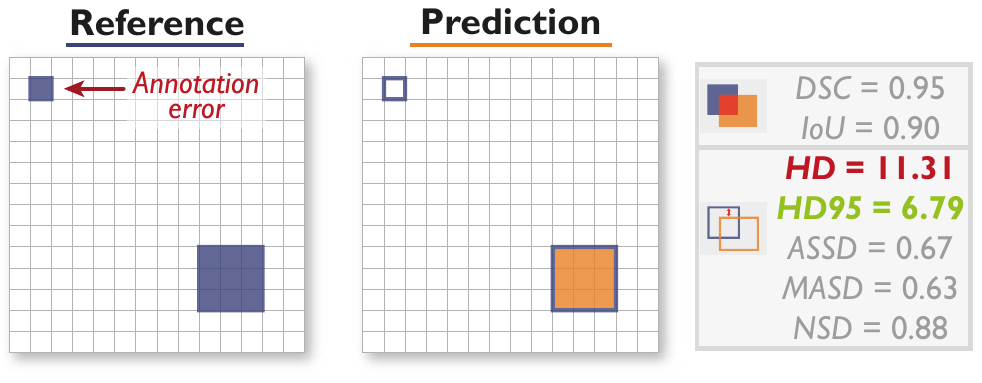}
    \caption{Effect of annotation errors/noise. A single erroneously annotated pixel may lead to a large decrease in performance, especially in the case of the \acf{HD} when applied to small structures. The \acf{HD95}, on the other hand, was designed to deal with spatial outliers. Further abbreviations: \acf{DSC}, \acf{IoU}, \acf{ASSD}, \acf{NSD}.}
    \label{fig:HD-outlier}
\end{figure}

%\newpage
\noindent\textbf{DG7.3: \acf{HD} versus \acf{X$^{th}$ Percentile HD}}
\begin{tcolorbox}[title= Summary of DG7.3: \acs{HD} versus \acs{X$^{th}$ Percentile HD}, colback=white]
    \begin{table}[H]
        \centering
        \begin{tabular}{p{5.5cm} p{0.5cm} p{5.5cm}}
            \textbf{\ac{HD}} & & \textbf{\ac{X$^{th}$ Percentile HD}} \\ 
            \textcolor{gray}{\faArrowCircleRight } Sensitivity to spatial outliers & & \textcolor{gray}{\faArrowCircleRight } Compensation for spatial outliers\\
        \end{tabular}
        \caption{Comparison of \acf{HD} and \acf{X$^{th}$ Percentile HD} in the context of the decision guide DG7.3 for Subprocess S7. Context: FP2.5.6 = distance-based penalization of outliers with outlier focus.}
        \label{tab:d7-3}
    \end{table}
\end{tcolorbox}
The \ac{HD} (Fig.~\ref{fig:cheat-sheet-hd}) calculates the maximum of all shortest distances for all points from one object boundary to the other, which is why it is also known as the Maximum Symmetric Surface Distance \cite{yeghiazaryan2018family}. The \ac{X$^{th}$ Percentile HD} calculates the X$^{th}$ percentile (e.g., 95\% percentile, the \ac{HD95}, Fig.~\ref{fig:cheat-sheet-hd95}) instead of the maximum, and should therefore be used instead if spatial outliers should be disregarded (FP2.5.6, see Fig.~\ref{fig:HD-outlier}).

\newpage

%-----------------------------------
\newpage
\subsubsection{Decision guide S8}
\label{ssec:dg8}
\hfill\\\noindent\textbf{DG8.1: Mask \acf{IoU} versus Boundary \acs{IoU} versus \acf{IoR}}
\begin{tcolorbox}[title= Summary of DG8.1: Mask \acs{IoU} versus Boundary \acs{IoU} versus \acs{IoR}, colback=white]
    \begin{table}[H]
        \centering
        \begin{tabular}{p{3cm} p{0.3cm} p{3cm} p{0.3cm} p{3cm}}
            \textbf{Mask \ac{IoU}} & & \textbf{Boundary \ac{IoU}} & & \textbf{\ac{IoR}} \\ 
          \textcolor{gray}{\faArrowCircleRight } Focus on overlap & &            
            \textcolor{gray}{\faArrowCircleRight } Focus on structure boundaries & & \textcolor{gray}{\faArrowCircleRight } Focus on overlap\\
            
            \textcolor{ODgreen}{\faPlusCircle } Widely used && \textcolor{gray}{\faArrowCircleRight } Recently proposed alternative; not well known && \textcolor{gray}{\faArrowCircleRight } Not well known \\

           \textcolor{red}{\faMinusCircle } Possible heavy penalization of predictions in the case of a high ratio of touching reference objects &&  && \textcolor{gray}{\faArrowCircleRight } Preferability in the case of a high ratio of touching reference objects\\

            \textcolor{red}{\faMinusCircle } Over-penalization of small structure sizes in tasks with high variability of sizes (FP3.2) & & && \\
            
           && \textcolor{red}{\faMinusCircle } Possibility of yielding perfect value for imperfect predictions && \\

           && \textcolor{gray}{\faArrowCircleRight } Additional hyperparameter, which can be determined based on inter-rater variability, for example && \\

 &&  &&   \textcolor{red}{\faMinusCircle }  Can be deceived  by large predictions\\
            
        \end{tabular}
        \caption{Comparison of Mask \acf{IoU}, Boundary \ac{IoU} and \acf{IoR} in the context of the decision guide DG8.1 for Subprocess S8. Context: FP1.1 = instance segmentation (InS).}
        \label{tab:d8-1}
    \end{table}
\end{tcolorbox}
% \textcolor{red}{\faMinusCircle }
% \textcolor{ODgreen}{\faPlusCircle }
% \textcolor{gray}{\faArrowCircleRight }
%\newpage

 In instance segmentation problems, it might be appropriate to base the localization criterion on the corresponding target segmentation metric (\textit{custom criterion}). For example, if the target segmentation metric chosen in Subprocess S6 (Extended Data Fig.~6) is \ac{NSD}, the localization criterion could be defined accordingly. This may not always be possible, for example because the target metric has no fixed upper bound (e.g., \ac{HD}), rendering the setting of adequate localization cutoffs challenging. An alternative strategy is to choose one of the common object detection localization criteria.

\begin{figure}[H]
    \centering
    \includegraphics[width=0.9\linewidth]{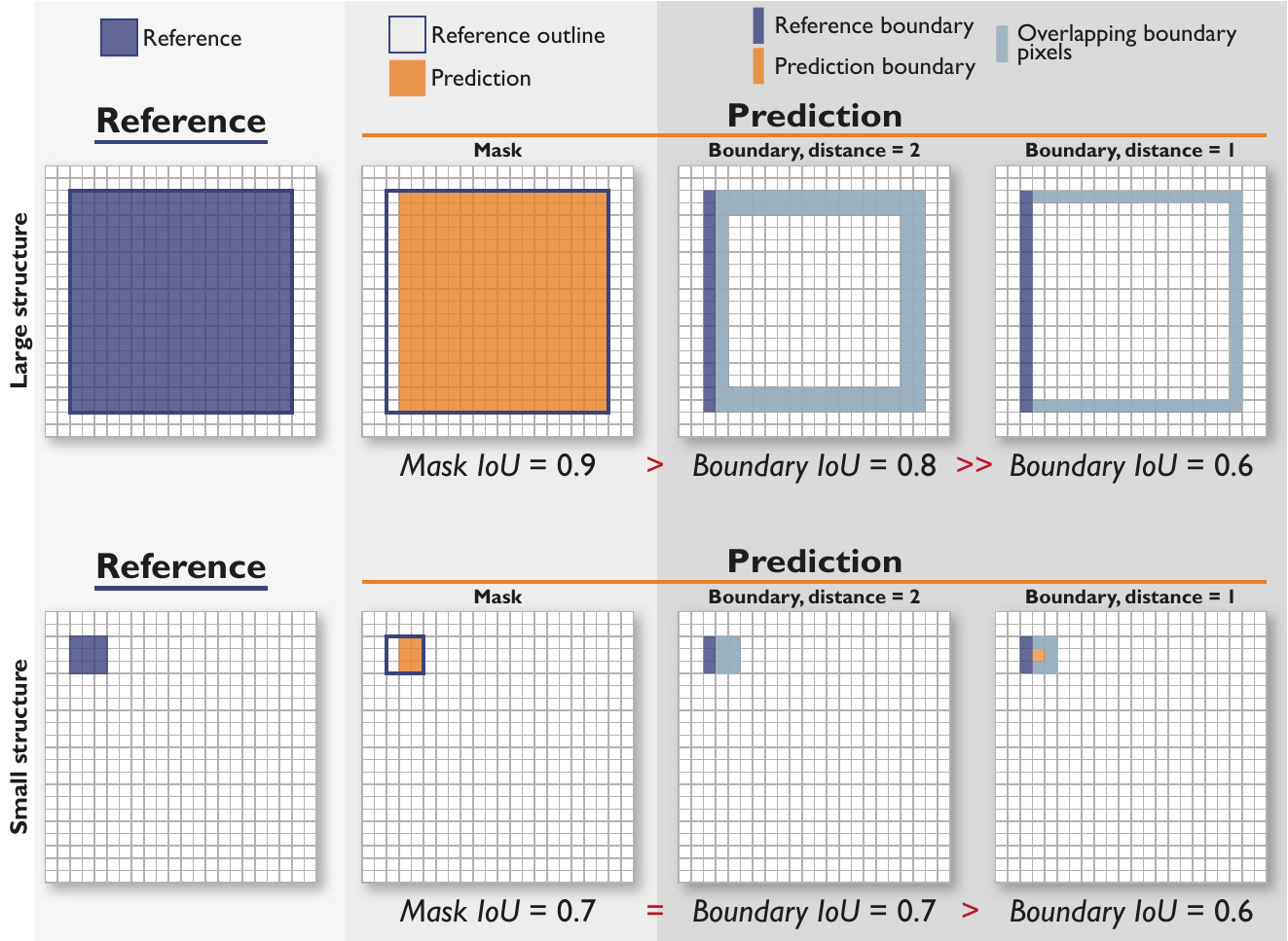}
    \caption{Compared to the Mask \acf{IoU}, the Boundary \ac{IoU} (third and fourth column, representing two different thresholds) (1) specifically penalizes errors in the boundaries and (2) is more invariant to structure sizes (top: large; bottom: small).}
     \label{fig:boundary-mask-iou} 
\end{figure}

\begin{figure}[H]
    \centering
    \includegraphics[width=0.7\linewidth]{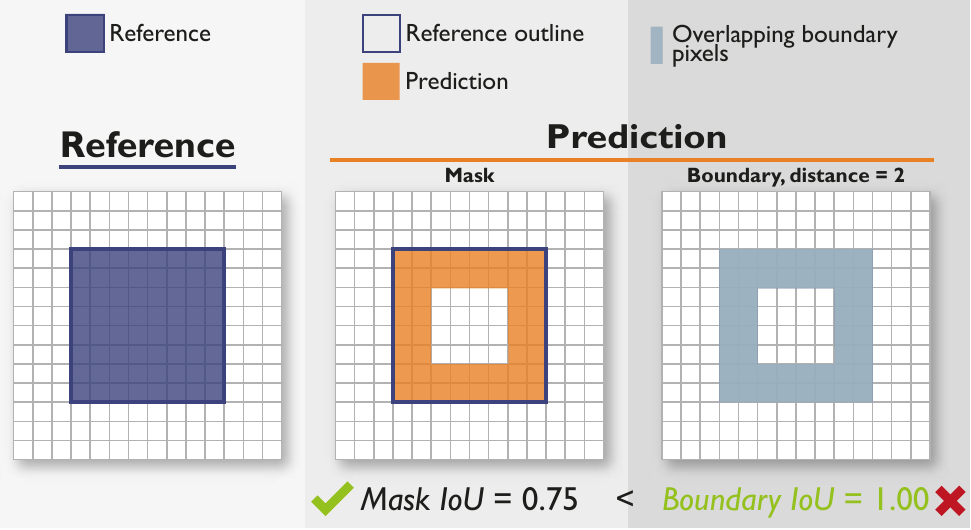}
    \caption{Example of a perfect Boundary \acf{IoU} score for an imperfect prediction. Overlapping pixels from the reference and prediction are shown in light blue. For a prediction with a hole in the middle, the Boundary \ac{IoU} may result in a score of 1.00 if the distance to border contains all mask pixels (here: distance = 2). However, the Mask \ac{IoU} spots the problem and yields a lower score.}
     \label{fig:boundary-iou} 
\end{figure}

\begin{figure}[H]
    \centering
    \includegraphics[width=0.9\linewidth]{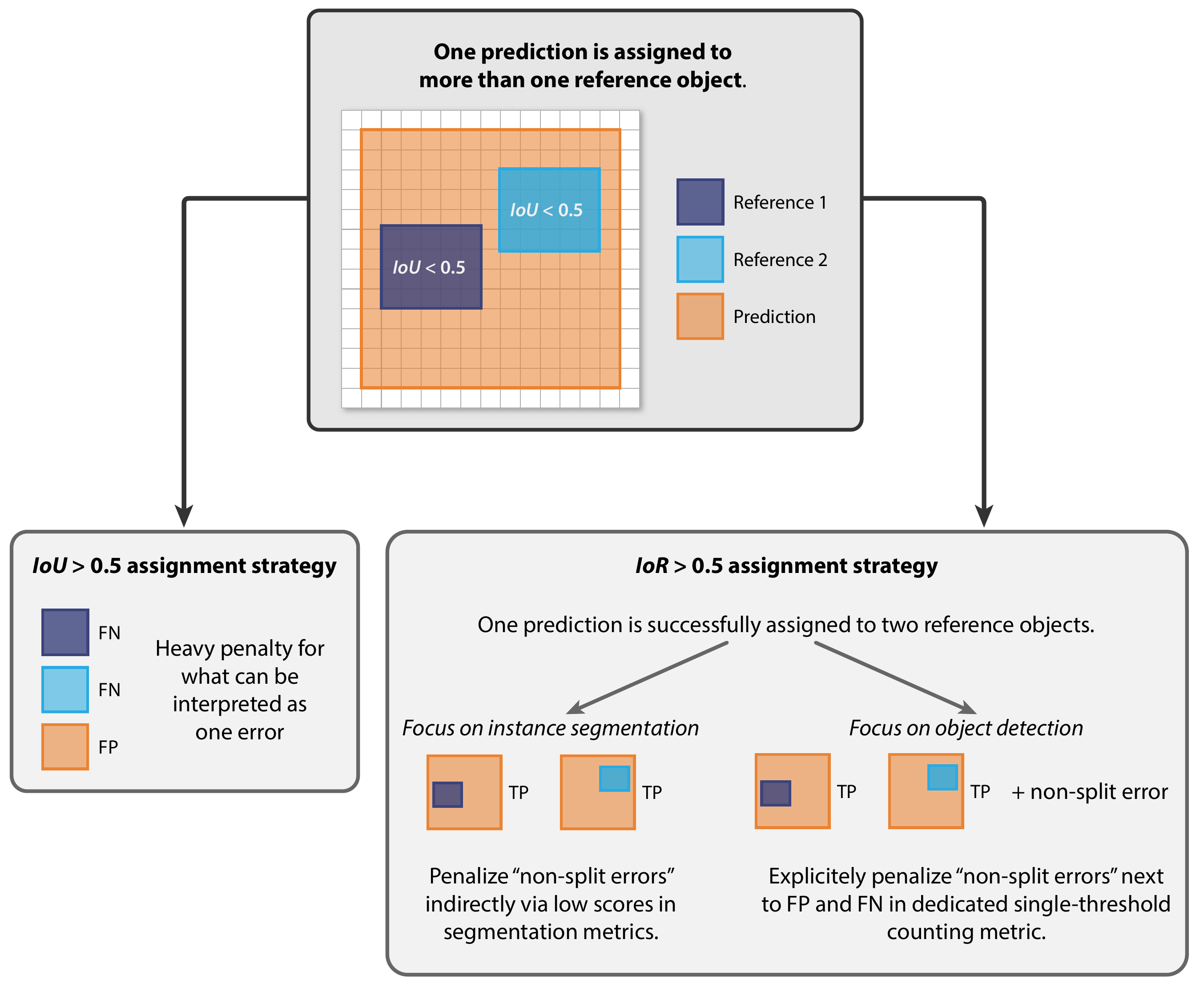}
    \caption{In case of one prediction assigned to multiple reference objects, an assignment strategy needs to be chosen. This may be based, for example, on the \textit{\acf{IoU}} > 0.5 strategy, which may result in a heavy penalty (two False Negatives (FN) and one \acf{FP}). Another option is to use the \textit{\acf{IoR}} > 0.5 strategy, which examines whether the prediction was successfully assigned to the reference objects. In an additional step, the "non-split errors" will be penalized. Used abbreviations: \acf{FN}, \acf{FP} and \acf{TP}.}
    \label{fig:od-ior}
\end{figure}

The following aspects should be taken into account when deciding between  Mask \ac{IoU} (Fig.~\ref{fig:cheat-sheet-iou}), Boundary \ac{IoU} (Fig.~\ref{fig:cheat-sheet-boundary-iou}), and \ac{IoR} (Fig.~\ref{fig:cheat-sheet-ior}) in instance segmentation problems. We will first focus on the more subtle distinction between Mask \ac{IoU} and Boundary  \ac{IoU}, and finally discuss scenarios for potential usage of \ac{IoR}: 
    \begin{description}
     \item [Boundary versus Mask \ac{IoU}] \hfill\\ 
     \begin{itemize}
     \item Boundary focus: While Mask \ac{IoU} measures the overlap of structures in general, Boundary \ac{IoU} allows to focus on the correctness of boundaries (FP2.1, see Fig.~\ref{fig:boundary-mask-iou}). Note that the focus on boundaries also comes with pitfalls. Boundary \ac{IoU} can even be deceived to result in a perfect value of $1.0$ despite an imperfect prediction (see Fig.~\ref{fig:boundary-iou}).
    \item Small structures: Mask \ac{IoU} over-penalizes small structures in tasks with high variability of structure sizes (FP3.2) because boundary pixels increase linearly (or quadratically) with size, while total pixels increase quadratically (or cubically) with size. Boundary \ac{IoU} \cite{cheng2021boundary} addresses this issue by selecting only pixels with a maximum distance of “d” with regard to the boundary for validation (see Fig.~\ref{fig:boundary-mask-iou}).
    \item Hyperparameters: For the computation of Boundary \ac{IoU}, the distance “d” constitutes an additional and sensitive hyperparameter to be determined. It can be determined based on inter-rater variability, for example.
    %or qualitative assessment of general annotation quality. 
    \item Popularity: While Mask \ac{IoU} represents an established concept that is well-known to the community, Boundary \ac{IoU} is a recently proposed modification \citep{cheng2021boundary} that might thus require specific introduction when used in validation.
    \end{itemize}
    \item [\ac{IoR}] 
    In the case of a high ratio of touching reference objects, 'non-split errors' (one prediction overlaps multiple reference objects) might occur frequently. While the \ac{IoU} criterion can potentially heavily penalize this scenario resulting in \ac{FN} and multiple \acp{FP}, a less severe penalization might be desired, e.g., in the form of the Intersection over Reference (\ac{IoR}) \cite{matula2015cell}. \ac{IoR} essentially considers the ratio of the area of a reference object that is covered by a prediction (see Fig.~\ref{fig:od-ior}), allowing for multiple TP matches of the same prediction. Appropriate penalization in these cases is then ensured either by separating such errors as 'merge errors' \cite{caicedo2019evaluation}, or by means of additional segmentation metrics. \ac{IoR} shares the behaviour of Mask \ac{IoU} regarding the above discussions on boundaries and small structures. As a major disadvantage, it can be deceived by large predictions. Widespread usage of \ac{IoR} is currently limited to the field of cell segmentation, where images with high density of structures are present \citep{matula2015cell}. 
     \end{description}

\newpage

\newpage

\begin{figure}[H]
    \centering
    \includegraphics[width=0.8\linewidth]{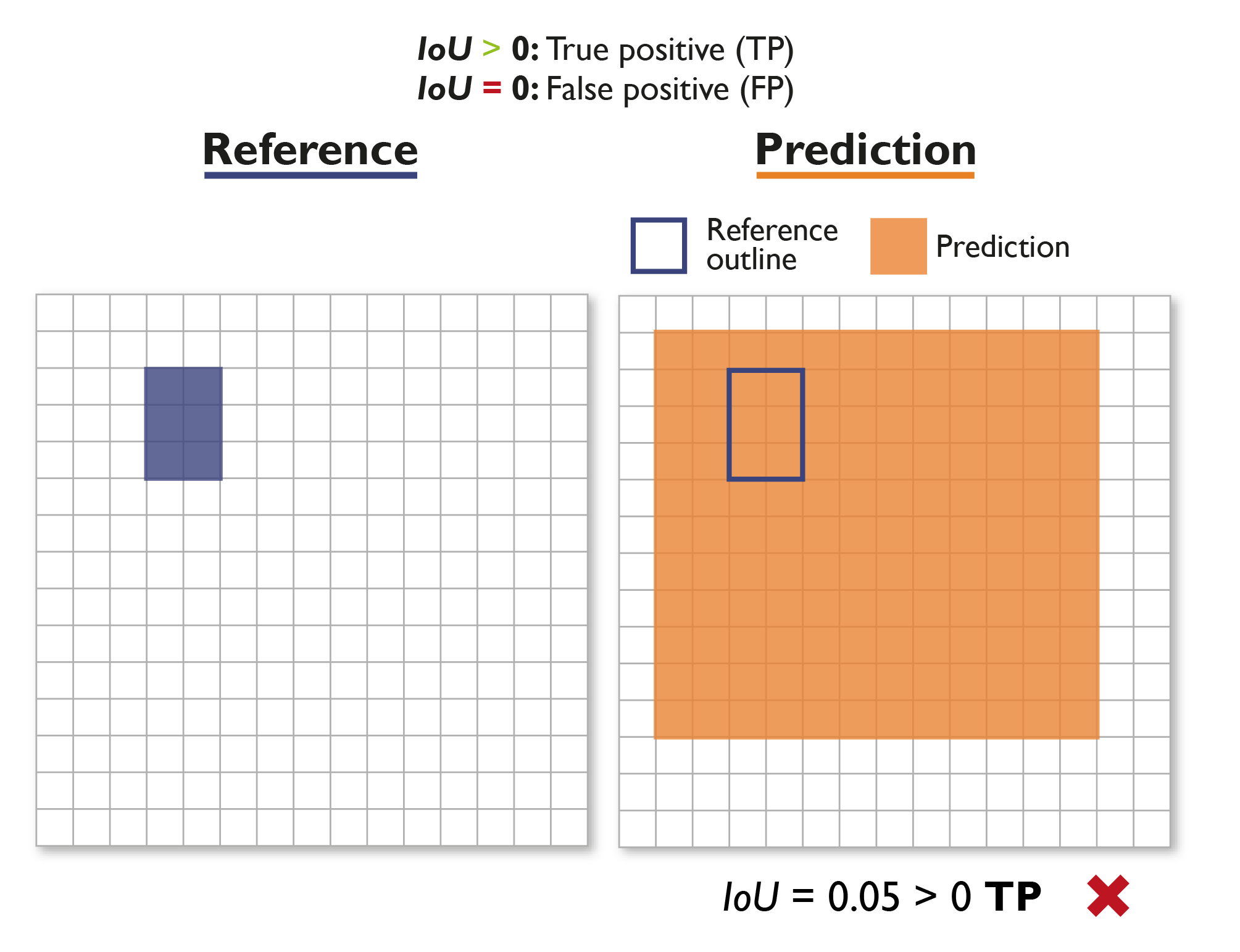}
    \caption{Effect of a loose \acf{IoU} criterion. When defining a \acf{TP} by an \ac{IoU} > 0, the resulting localizations may be deceived by very large predictions.}
    \label{fig:loose-iou}
\end{figure}

\begin{figure}[H]
    \centering
    \includegraphics[width=0.6\linewidth]{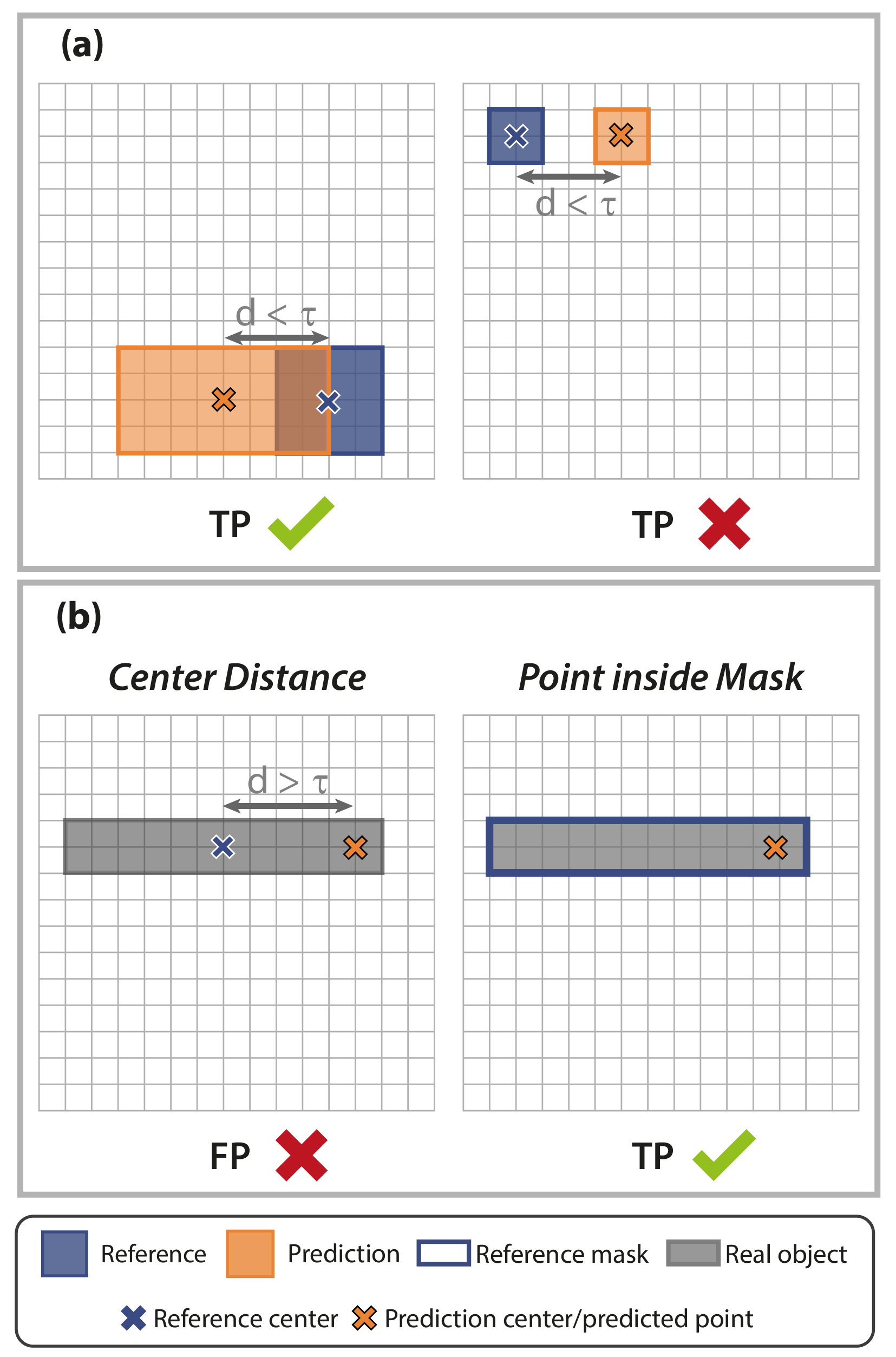}
    \caption{Pitfalls of the Center Distance. \textbf{(a) Ignoring overlap between objects.} Both predictions have the same distance to their corresponding reference center. The Center Distance, which requires a threshold distance $\tau$ between center points not be exceeded, does not take into account the overlap between objects. However, the right prediction does not overlap the reference and should thus not be considered a \acf{TP}. \textbf{(b) Tubular structures.} The Center Distance is not an ideal criterion because it implies that the prediction shown would result in a \acf{FP}, although it hits the elongated structure. This could be overcome by a Point inside Mask criterion.}
    \label{fig:center-dist}
\end{figure}

\newpage
\noindent\textbf{DG8.2: Mask \acf{IoU} > 0 versus Center Distance versus Point inside Mask/Box/Approx} 
\begin{tcolorbox}[title= Summary of DG8.2: Mask \acs{IoU} > 0 versus Center Distance versus Point inside Mask/Box/Approx, colback=white]
    \begin{table}[H]
        \centering
        \begin{tabular}{p{3.5cm} p{0.1cm} p{3.5cm} p{0.1cm} p{3.5cm}}
            \textbf{Mask \ac{IoU} > 0} & & \textbf{Center Distance} & & \textbf{Point inside Mask/Box/Approx} \\ 
                 \textcolor{ODgreen}{\faPlusCircle } No hyperparameters (standardized) & & \textcolor{gray}{\faArrowCircleRight } Distance threshold must be provided   & &  \textcolor{ODgreen}{\faPlusCircle }  No hyperparameters (standardized) \\
                
                  \textcolor{red}{\faMinusCircle } Strictness of criterion cannot be varied & & \textcolor{ODgreen}{\faPlusCircle } Strictness of criterion can be varied  & &    \textcolor{red}{\faMinusCircle } Strictness of criterion cannot be varied \\
            \textcolor{red}{\faMinusCircle } Potential large ambiguity of the predicted location as only few pixels overlap the reference & & \textcolor{ODgreen}{\faPlusCircle } Good representation of the object center (FP2.3)    & & \textcolor{ODgreen}{\faPlusCircle } Relatively good representation of tubular or disconnected structures\\
        \end{tabular}
        \caption{Comparison of Mask \acf{IoU} > 0, Center Distance and Point inside Mask/Box/Approx localization criteria in the context of the decision guide DG8.2 for Subprocess S8. Context: FP1.1 = object detection (ObD) problems in the case of either (1) FP4.4 = reference annotations provided as exact outline and FP2.4 = a desired localization as only position or (2) FP4.4 = reference annotations provided as rough outline and FP2.4 = a desired localization as only position. Note that Mask \ac{IoU} > 0 is only relevant for case (1).}
        \label{tab:d8-2}
    \end{table}
\end{tcolorbox}
When choosing a localization criterion for tasks where the mere existence of objects is of interest (as opposed to the outlining of objects), the following aspects should be considered:
    \begin{itemize}
    \item \textbf{Loose criterion:} (only recommended if the reference is provided as exact outline (FP2.4)) The intuitive choice of a very loose \ac{IoU} criterion (e.g., “\ac{IoU} > 0“ or “at least one pixel overlap”, Fig.~\ref{fig:cheat-sheet-iou}) comes with simplicity but implies the pitfall that the size of the predicted structure is in theory unbounded, i.e., the predicted location can be ambiguous (see Fig.~\ref{fig:loose-iou}). 
    \item \textbf{Point-based criteria:} A preferable alternative for the case of pure localization (without interest in outlines) is to constrain the prediction to a single coordinate. A common criterion for this scenario is the distance to the center point of the structure (Fig.~\ref{fig:cheat-sheet-center-distance}; which can also be of explicit interest, see FP2.3, Fig.~\ref{fig:center-dist}). 
    The center point\footnote{Depending on what kind of information the center point is derived from, different definitions are possible as detailed in Fig.~\ref{fig:cheat-sheet-center-distance}.}, however, might not be a good reference for tubular structures (check FP3.3) or disconnected structures (check FP3.6). In such cases (and if annotations are provided in the form of masks), a binary Point inside Mask/Box/Approx criterion (Fig.~\ref{fig:cheat-sheet-point-inside}) might be the better choice. On the other hand, the Point inside Mask/Box/Approx criterion does not allow for a variation of the criterion’s strictness (i.e., threshold). Application despite this shortcoming should be well-justified.
    \end{itemize}

\newpage
\noindent\textbf{DG8.3: Choose localization threshold} 
\begin{tcolorbox}[title= Summary DG8.3: Choose localization threshold, colback=white]
    \begin{table}[H]
        \centering
        \begin{tabular}{p{5.3cm} p{0.3cm} p{6cm}}
            \textbf{Lower thresholds} & & \textbf{Higher thresholds} \\ 
            \textcolor{gray}{\faArrowCircleRight } Interest in the existence of objects rather than their precise localization & & \textcolor{gray}{\faArrowCircleRight }  Interest in precise localization\\
            \textcolor{gray}{\faArrowCircleRight } Small size of structures (FP3.1) & & \textcolor{gray}{\faArrowCircleRight }  Dense distribution of structures in images (FP3.5)\\
            \textcolor{gray}{\faArrowCircleRight } High variability of structure sizes (FP3.2) & &  \\
            \textcolor{gray}{\faArrowCircleRight } 3D input images & &  \\
            \textcolor{gray}{\faArrowCircleRight } Uncertainties in the reference (FP4.3.1) & &  \\
        \end{tabular}
        \caption{Choosing the localization threshold in the context of the decision guide DG8.3 for Subprocess S8. This decision guide does not apply for Point inside Mask/Box/Approx criteria.}
        \label{tab:d8-3}
    \end{table}
\end{tcolorbox}
Note that most localization criteria require a threshold to be set (e.g., \ac{IoU} > 0.5 counts as detected). However, such cutoff renders the validity of results limited to the specific threshold. To increase robustness of reported metrics, it is common practice in the computer vision community to average metrics over multiple cutoff values (default for \ac{IoU} criteria: from 0.5 until 0.9 in steps of 0.05). On the other hand, certain properties of the underlying problem may limit the relevance of cutoff values to lower or higher values.

The following properties might warrant validation with lower thresholds: interest in the existence of objects rather than their precise localization, small size of structures (FP3.1), high variability of structure sizes (FP3.2), 3D input images (as volume increases cubically with size, the desired overlap ratio might require adaptation), uncertainties in the reference (FP4.3.1). Conversely, these properties typically warrant validation at higher thresholds: interest in precise localization, dense distribution of structures in images (FP3.5).

It should be noted that no threshold is needed for the Point inside Mask/Box/Approx and Mask \ac{IoU} > 0 criteria.

%-----------------------------------
\newpage
\subsubsection{Decision guide S9}
\label{ssec:dg9}
\begin{figure}
    \centering
    \includegraphics[width=\textwidth]{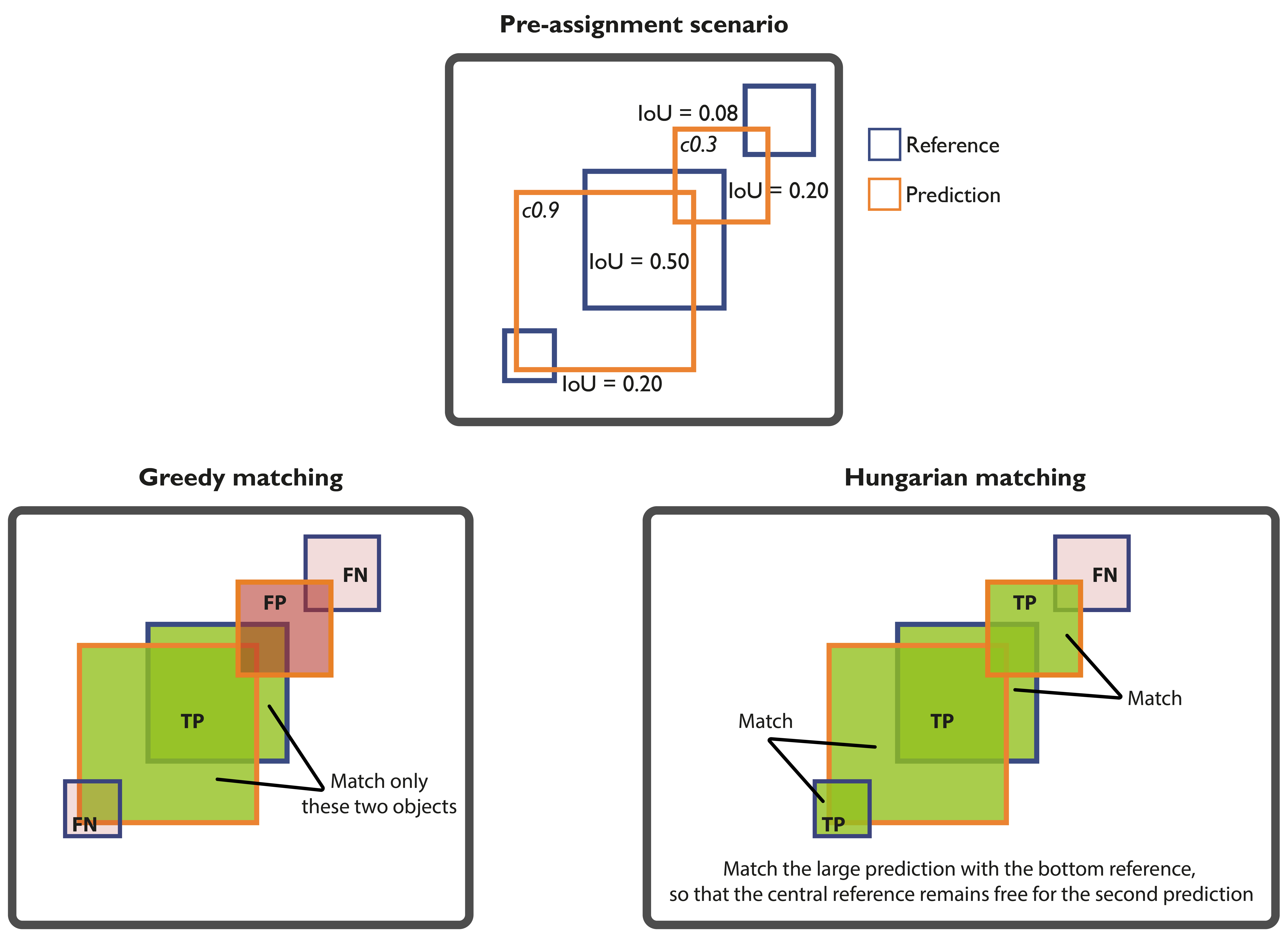}
    \caption{Comparison of Greedy versus Hungarian matching assignment strategies.}
    \label{fig:assignment-strategies}
\end{figure}

\hfill\\\noindent\textbf{DG9.1: Assignment without predicted class probabilities on instance level}
\begin{tcolorbox}[title= Summary DG9.1, colback=white]
    \begin{table}[H]
        \centering
        \begin{tabular}{p{3cm} p{0.3cm} p{3cm} p{0.3cm} p{4cm}}
            \textbf{"Localization criterion" > 0.5} & & \textbf{Greedy Matching} & & \textbf{Optimal (Hungarian) Matching} \\ 
            \textcolor{gray}{\faArrowCircleRight } Inherent avoidance of assignment ambiguities & & \textcolor{gray}{\faArrowCircleRight }  No necessity of sophisticated strategies & & \textcolor{gray}{\faArrowCircleRight } Necessity of sophisticated strategies\\
            \textcolor{red}{\faMinusCircle } Unfeasibility if overlapping predictions are possible & & & & \textcolor{gray}{\faArrowCircleRight } Optimistic interpretation/validation of ambiguous model outputs, but might not represent the most realistic approximation of model performance\\
        \end{tabular}
        \caption{Comparison of assignment strategies in the context of the decision guide DG9.1 for Subprocess S9. Context: lack of predicted class scores (FP5.1 = FALSE).}
        \label{tab:d9-1}
    \end{table}
\end{tcolorbox}
The following aspects should be considered when selecting the assignment strategy:
    \begin{itemize}
    \item \textbf{Matching via Overlap > 0.5 (Fig.~\ref{fig:cheat-sheet-matching-greater}):} If overlapping predictions are not possible (FP5.4 = FALSE), sophisticated matching strategies are often avoided in the biomedical domain by setting the threshold for the localization criterion (Mask \ac{IoU}, Boundary \ac{IoU}, or \ac{IoR}) to > 0.5. With this strategy, assignment ambiguities are inherently avoided. 
    %These two requirements inherently preclude matching conflicts). 
    However, if either overlapping predictions are possible, a non-overlap based criterion is employed, or a criterion with a threshold above 0.5 is not appropriate, one of the following strategies should be chosen.  
     \item \textbf{Greedy Matching (Figs.~\ref{fig:cheat-sheet-greedy-score}, \ref{fig:cheat-sheet-greedy-localization}):} A greedy approach can be taken, in which each reference is assigned to the best matching prediction. If predicted class scores are available (FP5.1 = TRUE) this is typically achieved based on the class score ("Greedy by Score Matching", Fig.~\ref{fig:cheat-sheet-greedy-score}). In the given scenario with FP5.1 = FALSE, an intuitive alternative is to rank predictions by the localization criterion score ("Greedy by localization criterion Matching", Fig.~\ref{fig:cheat-sheet-greedy-localization}). Assignment is then achieved by stepping through the ranked list, matching the current prediction with the most overlapping reference object, and removing the reference object from the assignment process.
    \item  \textbf{Optimal (Hungarian) Matching (Fig.~\ref{fig:cheat-sheet-hungarian}):} The Hungarian algorithm optimizes the matching between predictions and reference objects while minimizing a given cost function, such as the average overlap for all matched pairs. Notably, this optimization generally leads to optimistic interpretation/validation of ambiguous model outputs, but might not represent the most realistic approximation of model performance upon application (see Fig.~\ref{fig:assignment-strategies}).
    % What are examples for cost functions in our community? examples for applications of hungarian?
   
\end{itemize}

%-----------------------------------
\newpage

%\newpage

%\input{Mapping-Cheat-Sheets}

\newpage

%------------------------------------
\section{STEP 3~-~METRIC APPLICATION}
\label{app:metric-application}
\setcounter{figure}{0}
%------------------------------------
\acresetall
Once a suitable metric pool has been generated, the chosen metrics must be applied to the given data set. We recommend beginning with the setting of the global decision threshold in case metrics based on a fixed cutoff on the predicted class scores (FP2.6; more generally: decision region for more than two classes) have been selected, which is generally the case. In order to avoid overestimation of algorithm performance, this threshold needs to be set globally for all classes and metrics, as detailed in Suppl. Note~\ref{app:general-instructions}. 
Once raw metric values have been computed for all metrics, metric values are aggregated, potentially combined (for rankings) and reported according to the recommendations in Extended Data Tab.~\ref{tab:cross-topic}. Importantly, we support the user by providing cheat sheets for the entire pool of \textit{Metrics Reloaded} metrics that help find reference implementations and overcome metric-specific pitfalls (Suppl. Note~\ref{app:metric-cheat-sheets}).

\newpage

%------------------------------------
%\subsection{Metrics Cheat Sheets}
%\label{app:metric-cheat-sheets}
%------------------------------------
%------------------------------------
\subsection{Metrics Cheat Sheets}
\label{app:metric-cheat-sheets}
\begin{figure}[H]
    \centering
    \includegraphics[width=0.9\textwidth]{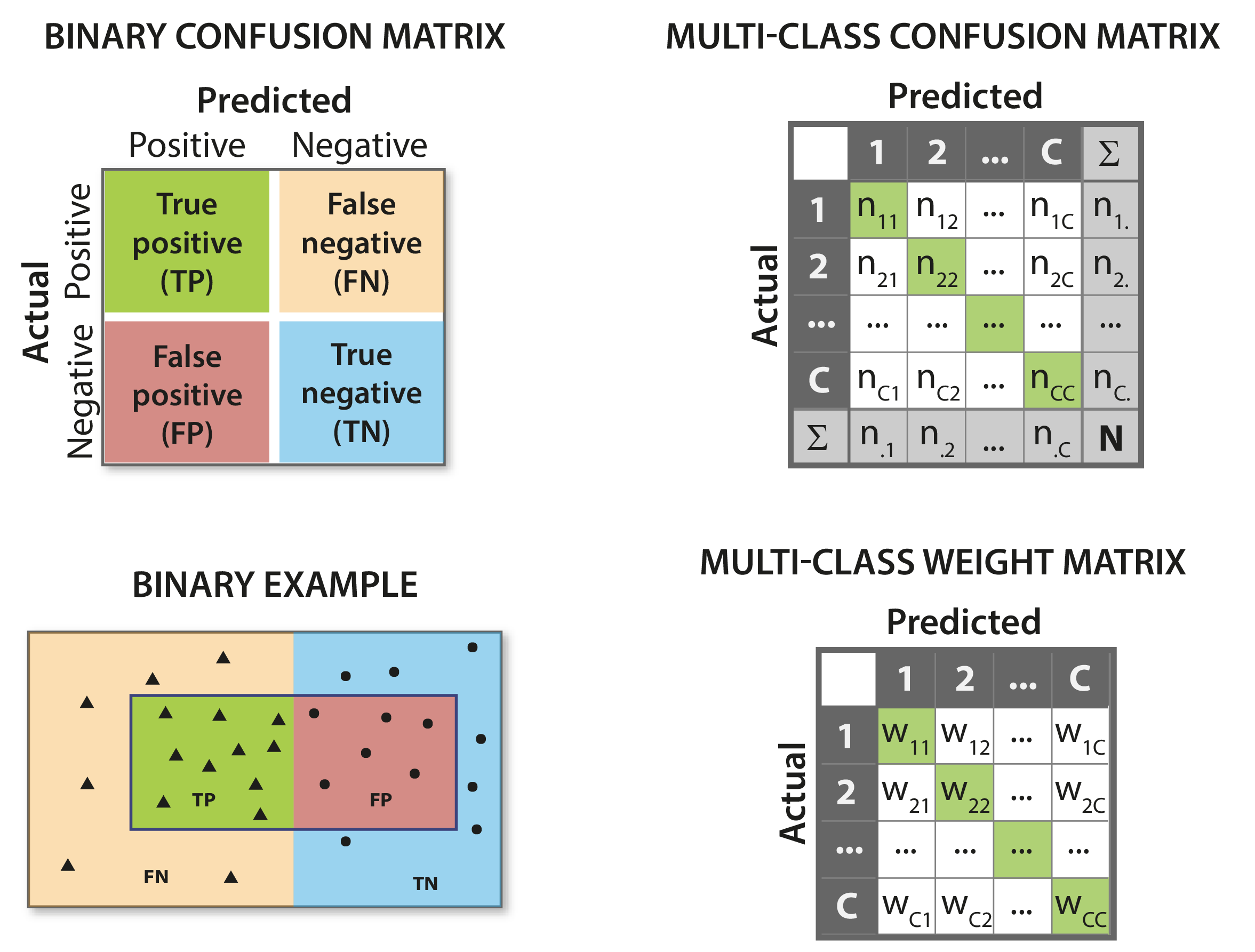}
    \caption{Schematic example of the confusion matrix for two and for $C$ classes. For the latter case, we also present a weight or cost matrix with weights $w_{ij} > 0$ without loss of generality. For the binary confusion matrix, we show an example illustrating the cardinalities for a prediction of triangles and circles.}
    \label{fig:confusion-matrix}
\end{figure}

\begin{figure}[H]
    \centering
    \includegraphics[width=\textwidth]{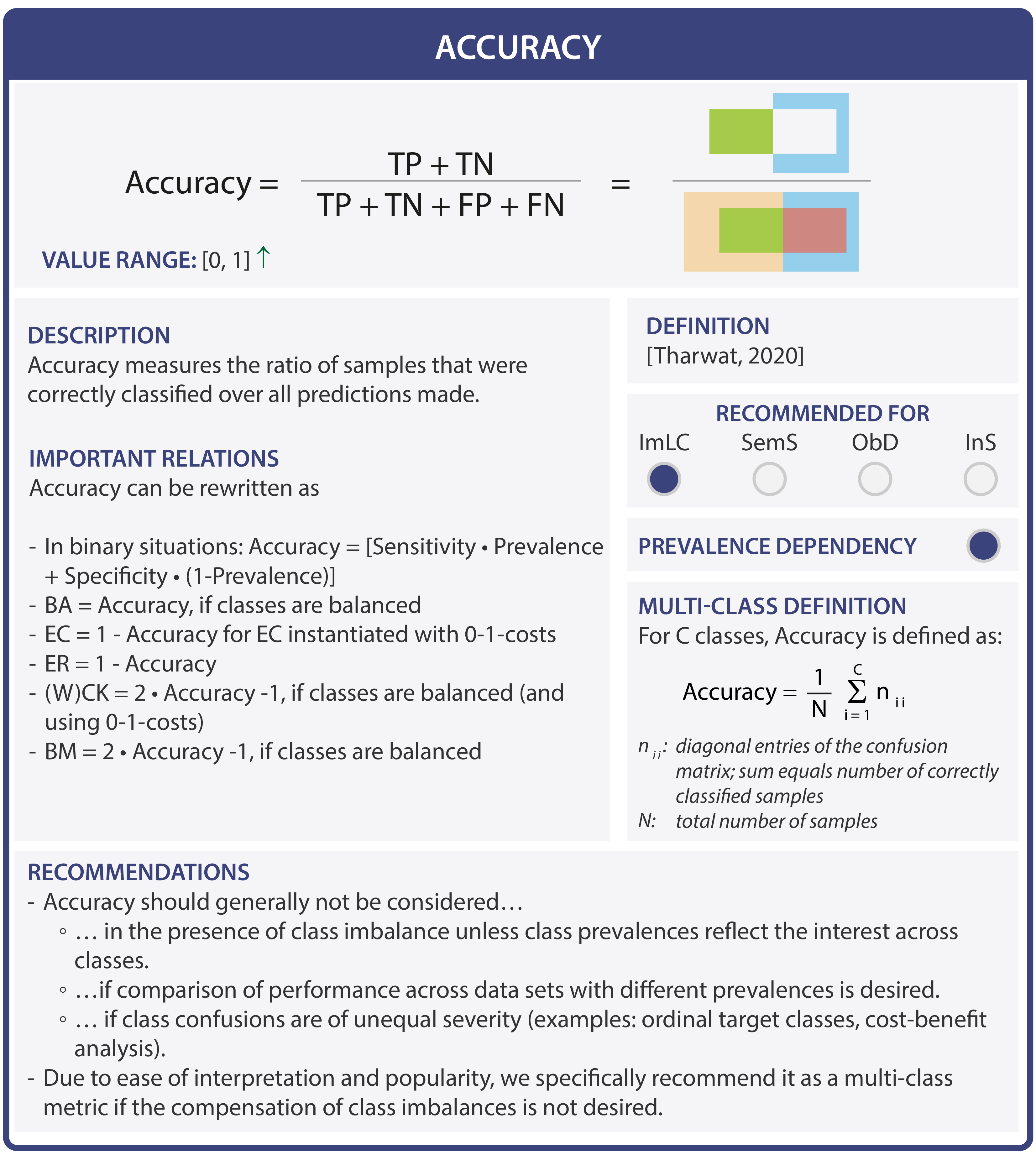}
    \caption{Cheat Sheet for the Accuracy. The upward arrow in the value range indicates that higher values are better than lower values. Abbreviations used in the figure: \acf{BA}, \acf{BM}, \acf{CK}, \acf{EC}, \acf{ER}, \acf{FN}, \acf{FP}, \acf{ImLC}, \acf{InS}, \acf{ObD}, \acf{SemS}, \acf{TN}, \acf{TP}, \acf{WCK}. Reference used in the figure: Tharwat, 2020: \cite{tharwat2020classification}. We recommend Accuracy as a multi-class counting metric in Subprocess S2 (Extended Data Fig.~\ref{fig:metric-selection:s2-multi-class}).}
    \label{fig:cheat-sheet-accuracy}
\end{figure}

\begin{figure}[H]
    \centering
    \includegraphics[width=\textwidth]{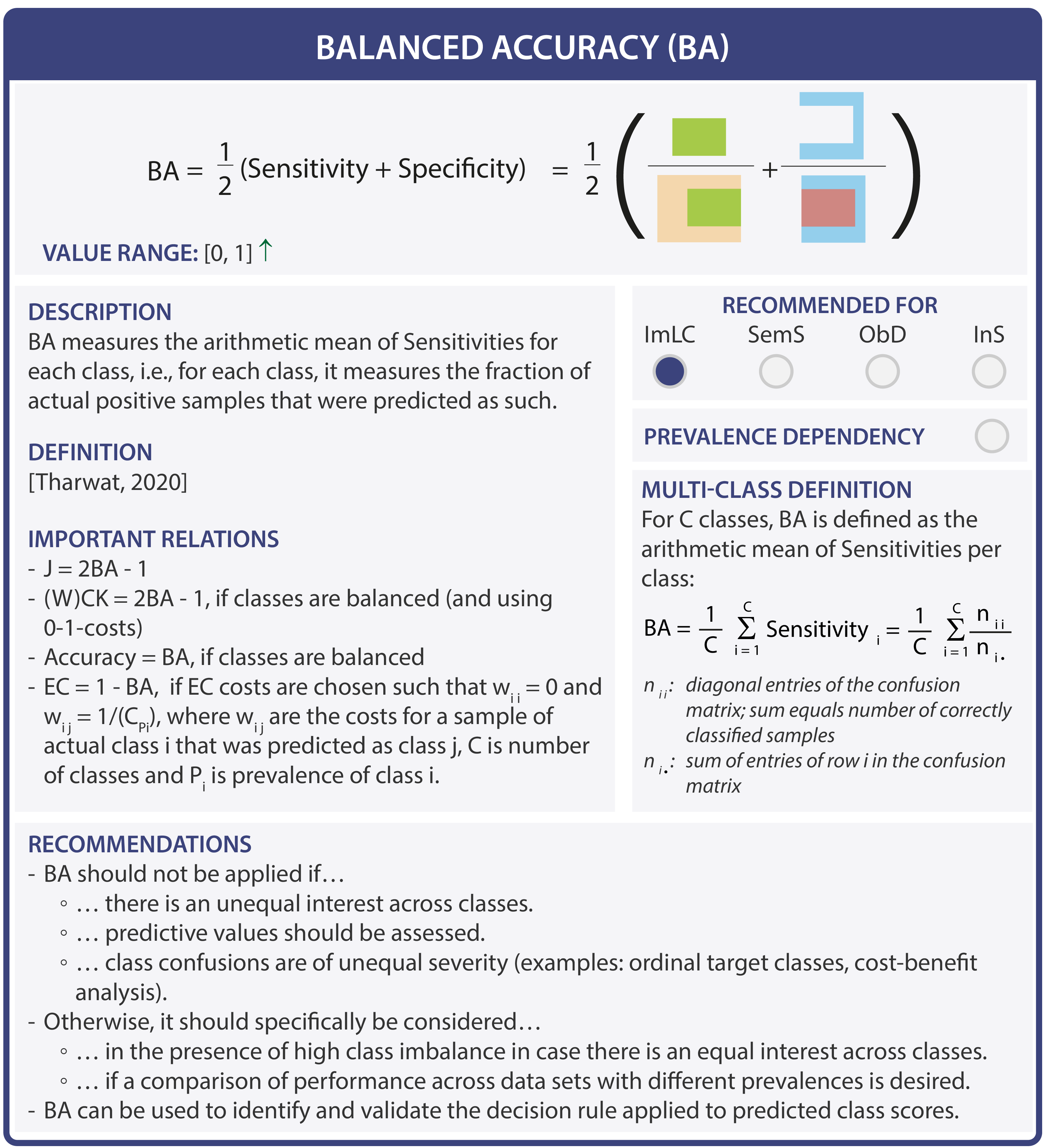}
    \caption{Cheat Sheet for the \acf{BA}. The upward arrow in the value range indicates that higher values are better than lower values. Abbreviations used in the figure: \acf{ImLC}, \acf{InS}, \acf{ObD}, \acf{SemS}, \acf{CK}, \acf{EC}, \acf{J}, \acf{WCK}. We recommend \ac{BA} as a multi-class counting metric in Subprocess S2 (Extended Data Fig.~\ref{fig:metric-selection:s2-multi-class}).}
    \label{fig:cheat-sheet-ba}
\end{figure}

\begin{figure}[H]
    \centering
    \includegraphics[width=\textwidth]{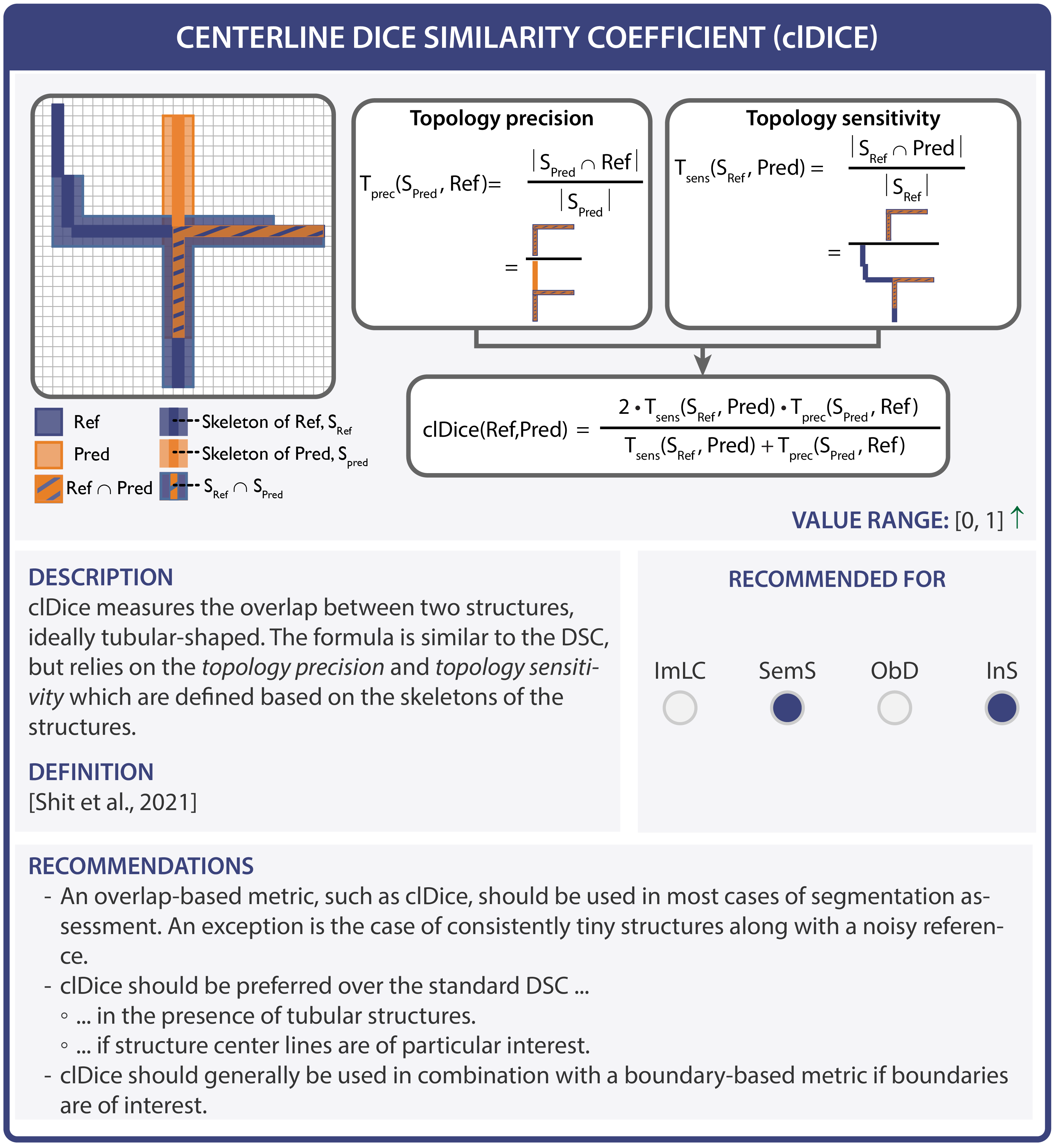}
    \caption{Cheat Sheet for the \acf{clDice}. The upward arrow in the value range indicates that higher values are better than lower values. Abbreviations used in the figure: \acf{DSC}, \acf{ImLC}, \acf{InS}, \acf{ObD}, \acf{SemS}. We recommend \ac{clDice} as an overlap-based metric in Subprocess S6 (Extended Data Fig.~\ref{fig:metric-selection:s6-overlap}).}
    \label{fig:cheat-sheet-cldice}
\end{figure}

\begin{figure}[H]
    \centering
    \includegraphics[width=\textwidth]{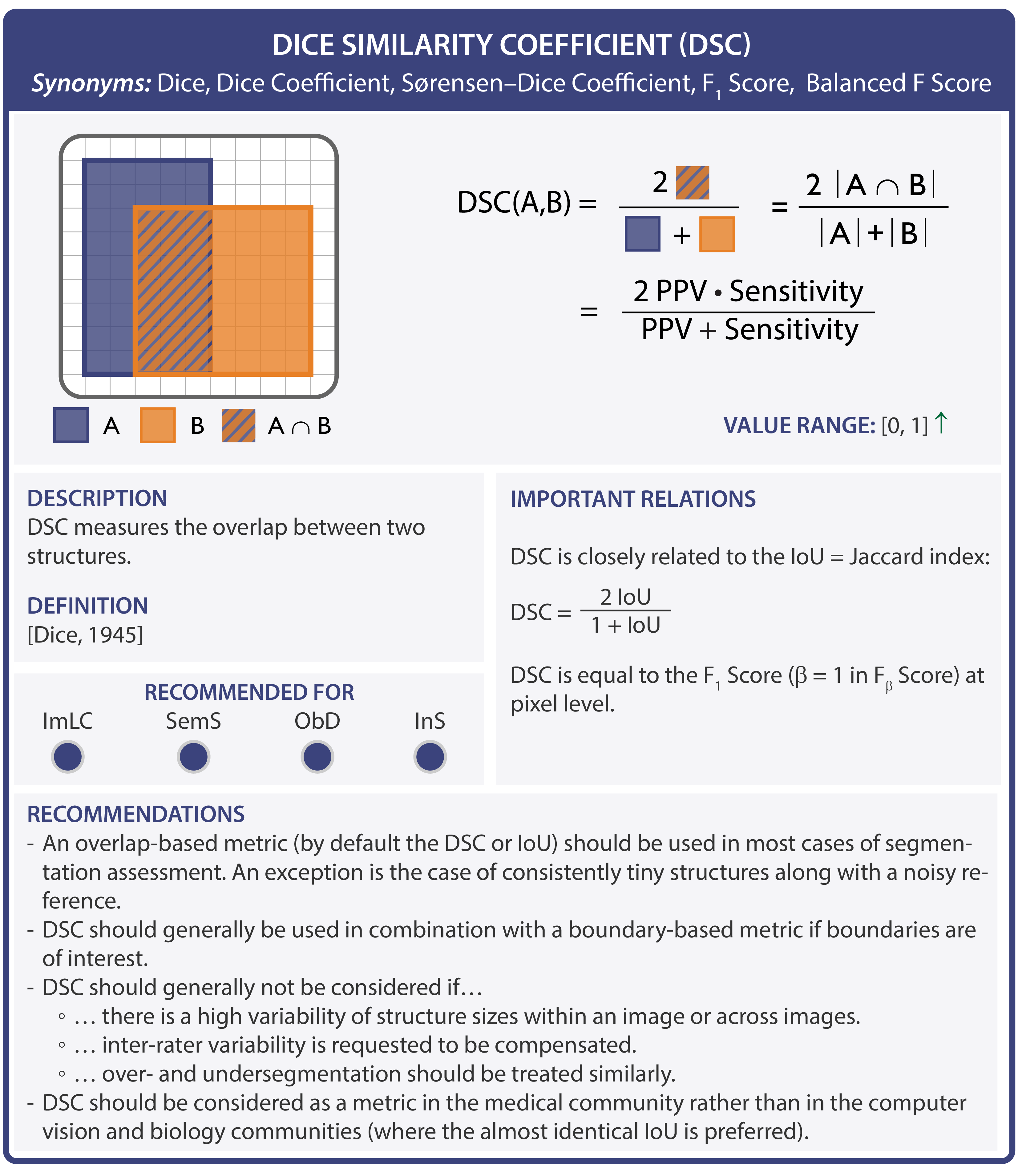}
    \caption{Cheat Sheet for the \acf{DSC}. The upward arrow in the value range indicates that higher values are better than lower values. Abbreviations used in the figure: \acf{FN}, \acf{FP}, \acf{ImLC}, \acf{InS}, \acf{IoU}, \acf{ObD}, \acf{PPV}, \acf{SemS}, \acf{TN}, \acf{TP}. Reference used in the figure: Dice, 1945: \cite{dice1945measures}. We recommend \ac{DSC} as an overlap-based  metric in Subprocess S6 (Extended Data Fig.~\ref{fig:metric-selection:s6-overlap}).}
    \label{fig:cheat-sheet-dsc}
\end{figure}

\begin{figure}[H]
    \centering
    \includegraphics[width=1\textwidth]{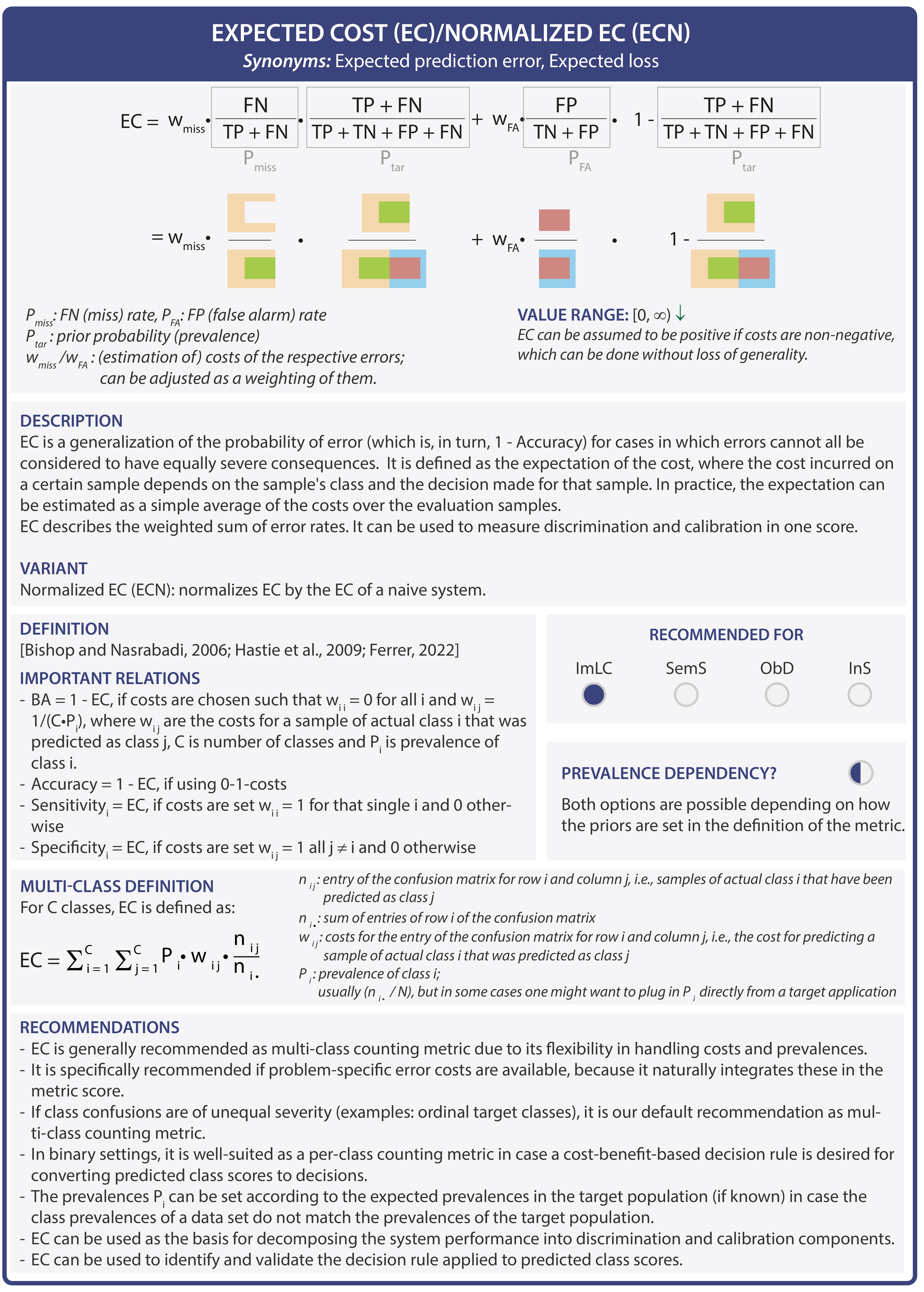}
    \caption{Cheat Sheet for the \acf{EC}/\acf{ECN}. The downward arrow in the value range indicates that lower values are better than higher values. Abbreviations used in the figure: \acf{BA}, \acf{ImLC}, \acf{InS}, \acf{ObD}, \acf{SemS}. References used in the figure: Bishop and Nasrabadi, 2006: \cite{bishop2006pattern}, Ferrer 2022: \cite{ferrer2022analysis}, Hastie et al., 2009:  \cite{hastie2009elements}. We recommend \ac{EC} as a per-class counting metric in Subprocess S2 (Extended Data Fig.~\ref{fig:metric-selection:s2-multi-class}).}
    \label{fig:cheat-sheet-ec}
\end{figure}

\begin{figure}[H]
    \centering
    \includegraphics[width=\textwidth]{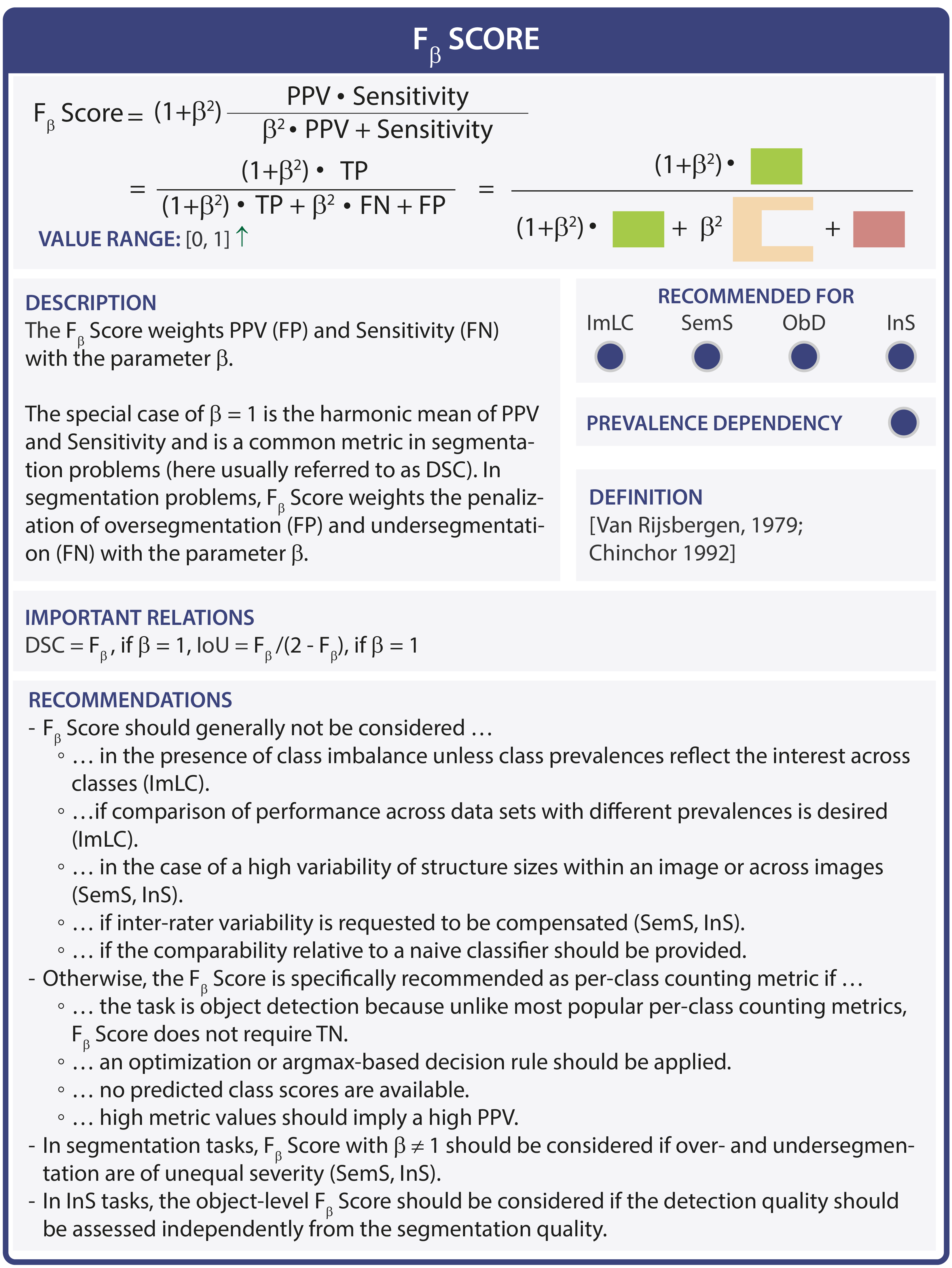}
    \caption{Cheat Sheet for the F$_\beta$ Score. The upward arrow in the value range indicates that higher values are better than lower values. Abbreviations used in the figure: \acf{DSC}, \acf{FN}, \acf{FP}, \acf{ImLC}, \acf{InS}, \acf{IoU}, \acf{ObD}, \acf{PPV}, \acf{SemS}, \acf{TN}, \acf{TP}. References used in the figure: Chinchor 1992: \cite{Chinchor1992}, Van Rijsbergen, 1979: \cite{van1979information}. We recommend F$_\beta$ Score as a per-class counting metric in Subprocess S3 (Extended Data Fig.~\ref{fig:metric-selection:s3-single-threshold}).}
    \label{fig:cheat-sheet-fbeta}
\end{figure}

\begin{figure}[H]
    \centering
    \includegraphics[width=\textwidth]{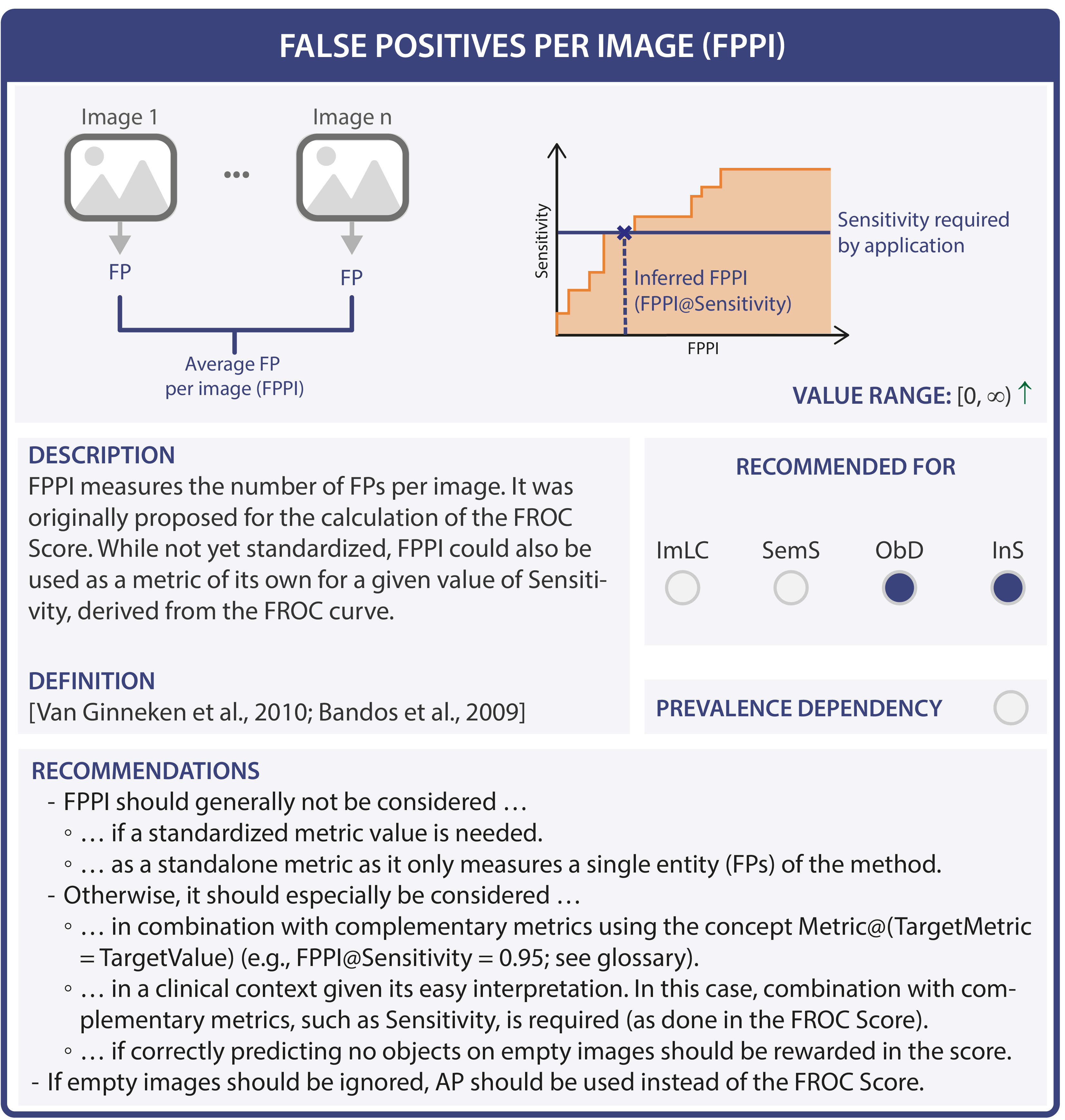}
    \caption{Cheat Sheet for the \acf{FPPI} \cite{van2010comparing, bandos2009area}. The upward arrow in the value range indicates that higher values are better than lower values. Abbreviations used in the figure: \acf{FP}, \acf{FROC}, \acf{ImLC}, \acf{InS}, \acf{ObD}, \acf{SemS}. References used in the figure: Van Ginneken et al., 2010: \cite{van2010comparing}, Bandos et al., 2009: \cite{bandos2009area}. We recommend \ac{FPPI} as a per-class counting metric in Subprocess S3 (Extended Data Fig.~\ref{fig:metric-selection:s3-single-threshold}) for target value-based optimization using the concept Metric@(TargetMetric = TargetValue) (e.g., FPPI@Sensitivity = 0.95; see glossary in Suppl. Note~\ref{app:terminology}).}
    \label{fig:cheat-sheet-fppi}
\end{figure}

\begin{figure}[H]
    \centering
    \includegraphics[width=\textwidth]{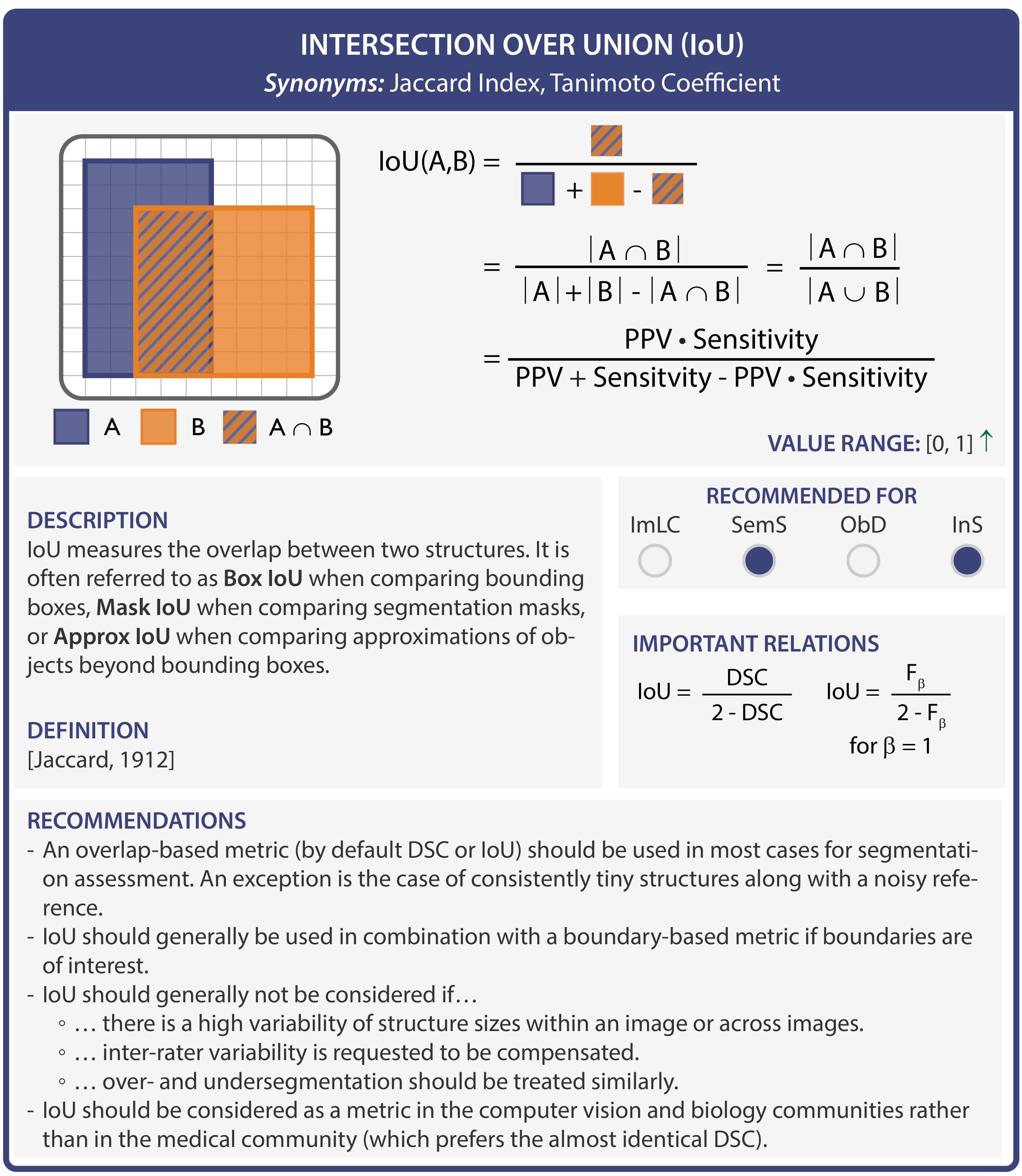}
    \caption{Cheat Sheet for the \acf{IoU}. The upward arrow in the value range indicates that higher values are better than lower values. Abbreviations used in the figure: \acf{DSC}, \acf{ImLC}, \acf{InS}, \acf{ObD}, \acf{PPV}, \acf{SemS}. Reference used in the figure: Jaccard, 1912: \cite{jaccard1912distribution}. We recommend \ac{IoU} as an overlap-based  metric in Subprocess S6 (Extended Data Fig.~\ref{fig:metric-selection:s6-overlap}).}
    \label{fig:cheat-sheet-iou}
\end{figure}

\begin{figure}[H]
    \centering
    \includegraphics[width=\textwidth]{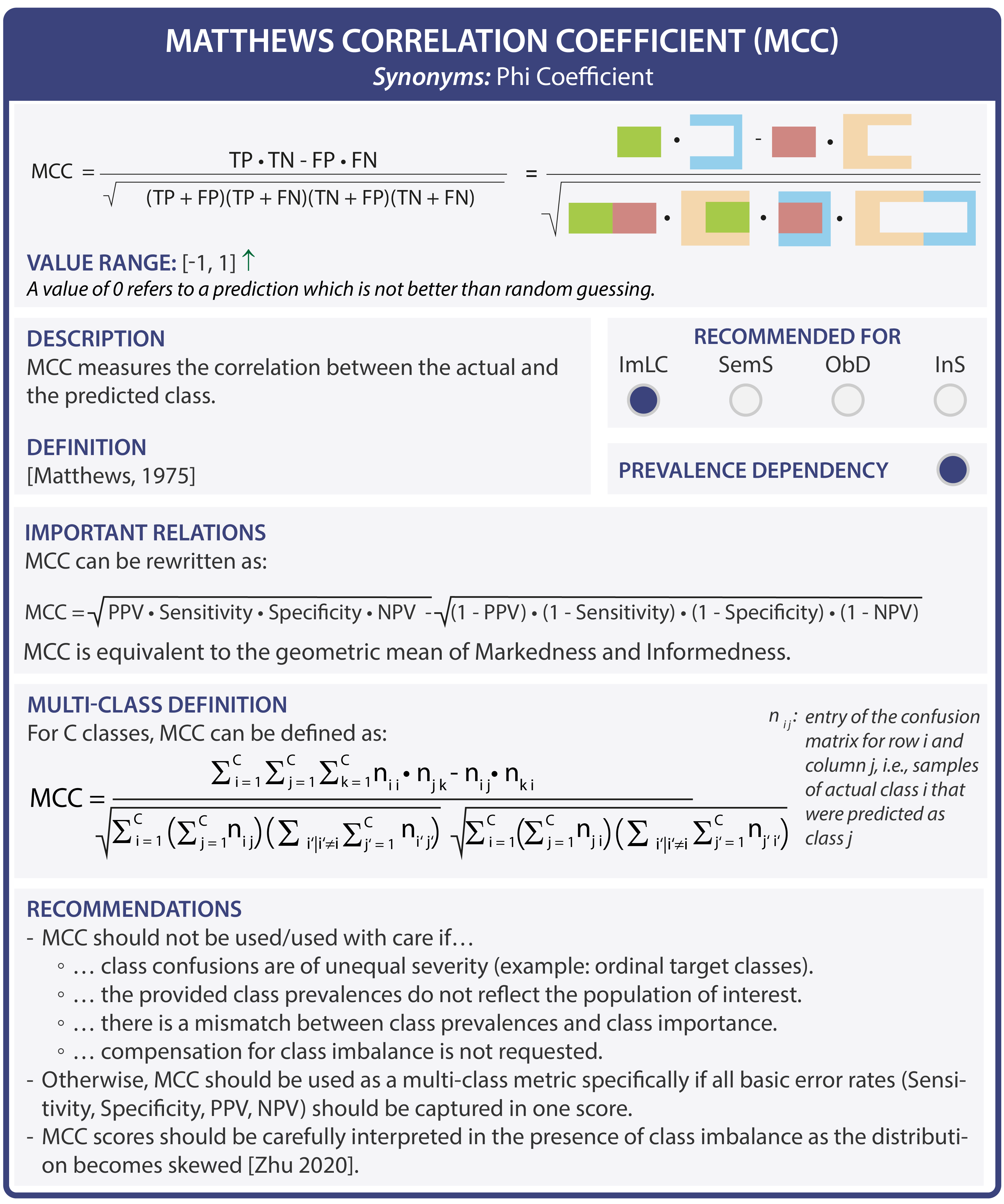}
    \caption{Cheat Sheet for the \acf{MCC}. The upward arrow in the value range indicates that higher values are better than lower values. Abbreviations used in the figure: \acf{ImLC}, \acf{InS}, \acf{NPV}, \acf{PPV}, \acf{ObD}, \acf{SemS}. References used in the figure: Matthews, 1975: \cite{matthews1975comparison}, Zhu, 2020: \cite{zhu2020OnMatthews}. We recommend \ac{MCC} as a multi-class counting metric in Subprocess S2 (Extended Data Fig.~\ref{fig:metric-selection:s2-multi-class}).}
    \label{fig:cheat-sheet-mcc}
\end{figure}

\begin{figure}[H]
    \centering
    \includegraphics[width=\textwidth]{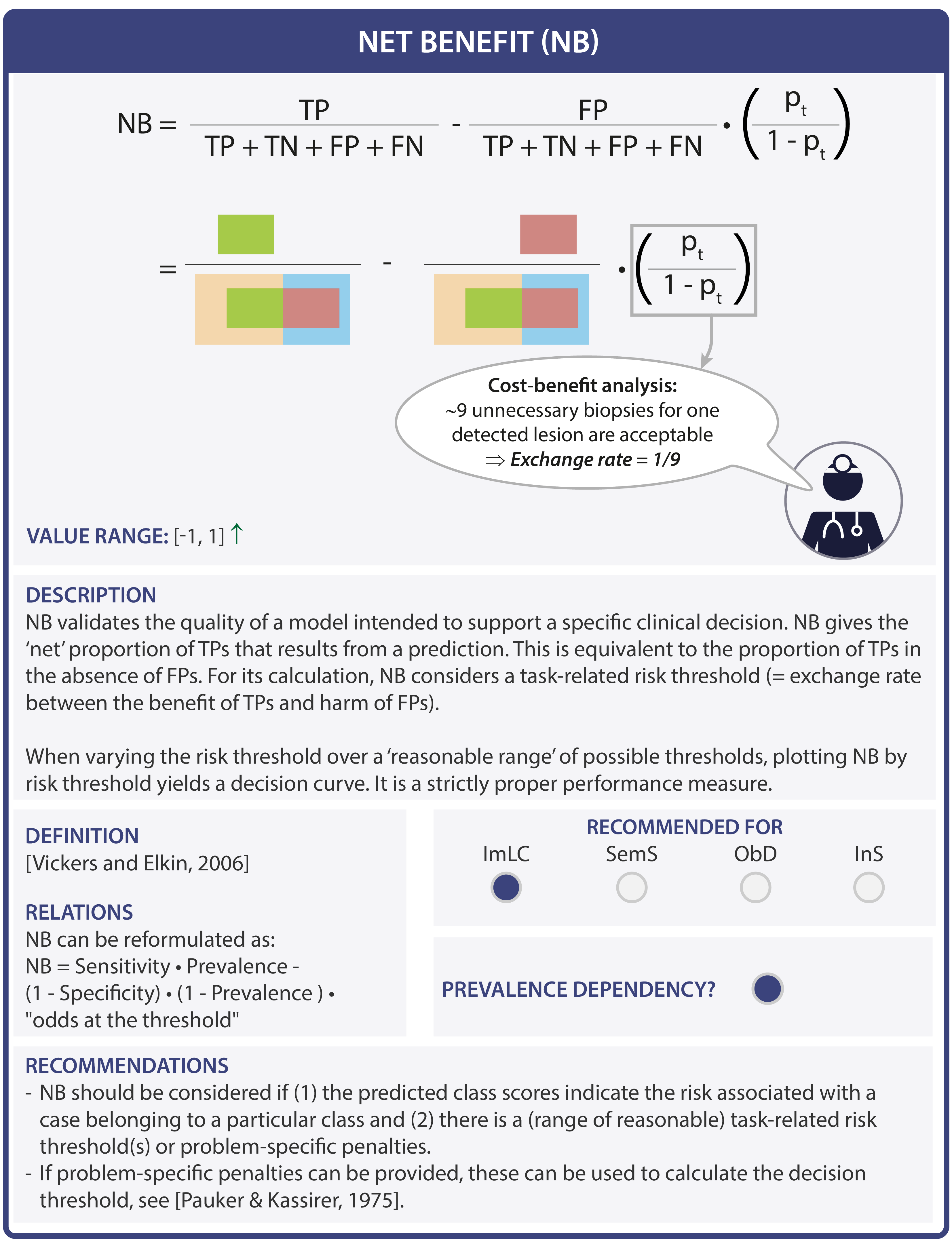}
    \caption{Cheat Sheet for the \acf{NB}. The upward arrow in the value range indicates that higher values are better than lower values. Abbreviations used in the figure: \acf{FN}, \acf{FP}, \acf{ImLC}, \acf{InS}, \acf{ObD}, \acf{SemS}, \acf{TN}. References used in the figure: Pauker and Kassirer, 1975: \cite{pauker1975therapeutic}, Vickers and Elkin, 2006: \cite{vickers2006decision}, Vickers et al., 2016: \cite{vickers2016net}. We recommend \ac{NB} as a per-class counting metric in Subprocess S3 (Extended Data Fig.~\ref{fig:metric-selection:s3-single-threshold}).}
    \label{fig:cheat-sheet-nb}
\end{figure}

\begin{figure}[H]
    \centering
    \includegraphics[width=\textwidth]{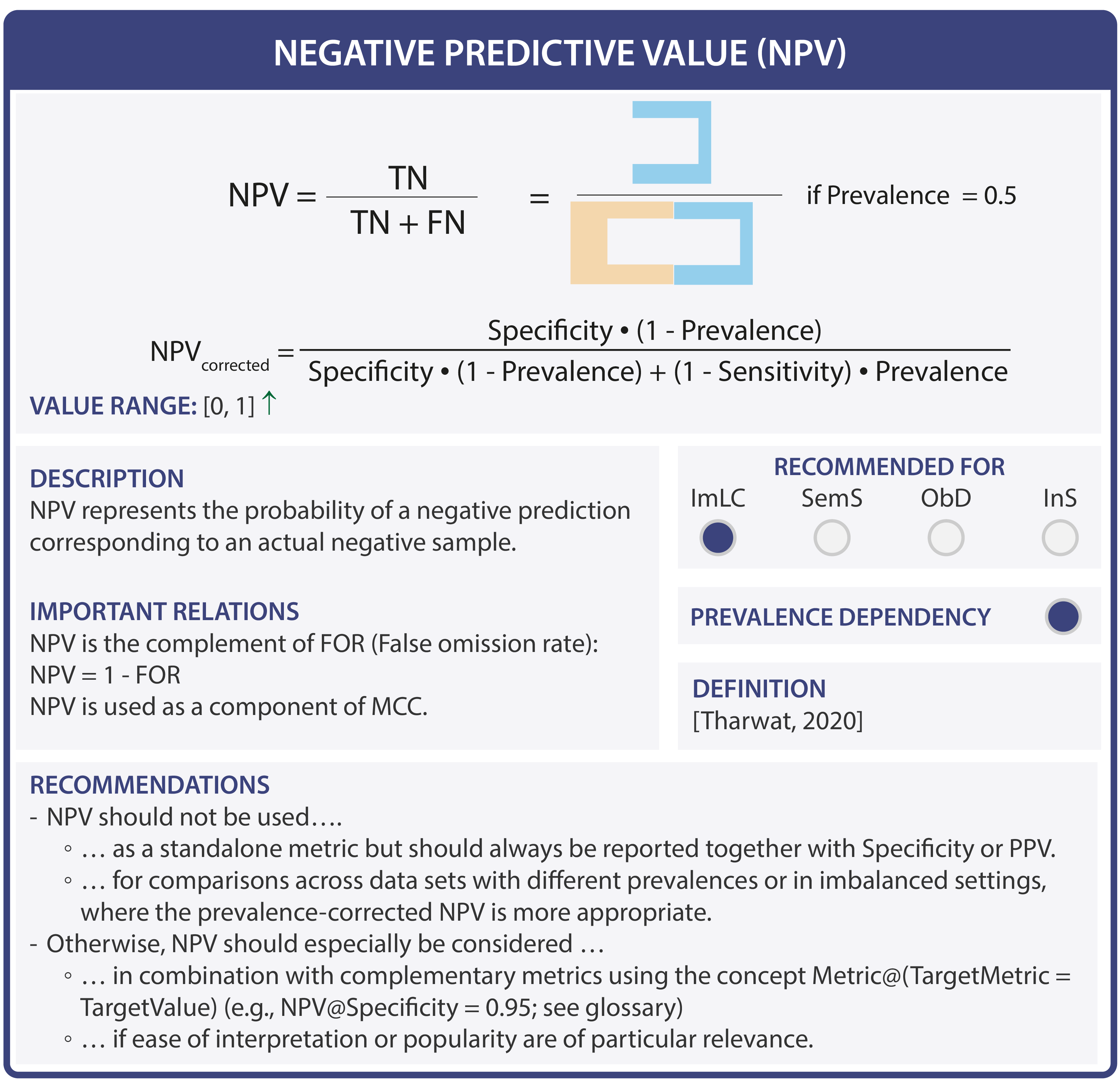}
    \caption{Cheat Sheet for the \acf{NPV}. The upward arrow in the value range indicates that higher values are better than lower values. Abbreviations used in the figure: \acf{FN}, \acf{FOR}, \acf{FP}, \acf{ImLC}, \acf{InS}, \acf{ObD}, \acf{PPV}, \acf{SemS}, \acf{TN}. Reference used in the figure: Tharwat, 2020:  \cite{tharwat2020classification}. We recommend \ac{NPV} as a per-class counting metric in Subprocess S3 (Extended Data Fig.~\ref{fig:metric-selection:s3-single-threshold}) for target value-based optimization using the concept Metric@(TargetMetric = TargetValue) (e.g., NPV@Sensitivity = 0.95; see glossary in Suppl. Note~\ref{app:terminology}).}
    \label{fig:cheat-sheet-npv}
\end{figure}

\begin{figure}[H]
    \centering
    \includegraphics[width=\textwidth]{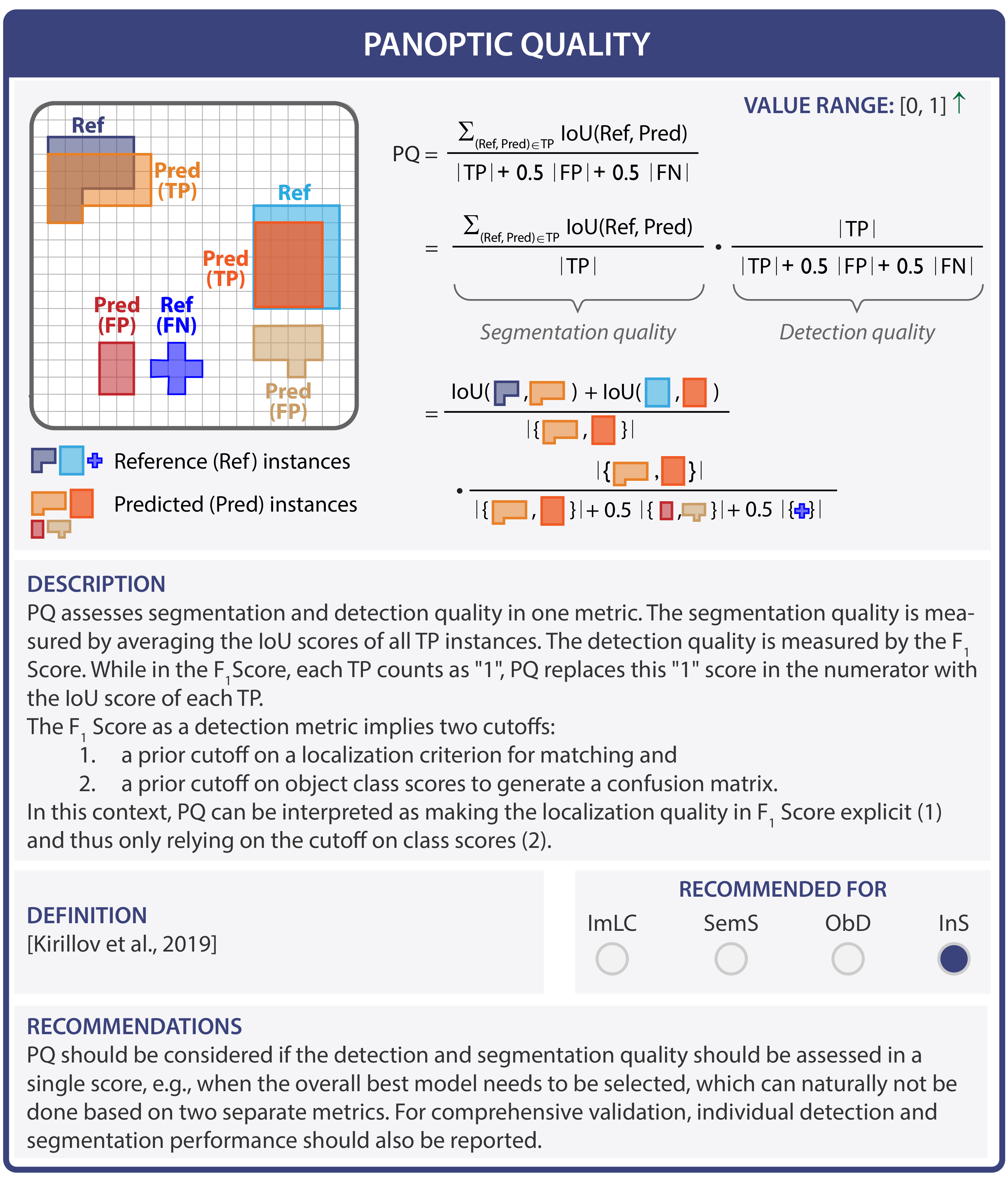}
    \caption{Cheat Sheet for the \acf{PQ}. The upward arrow in the value range indicates that higher values are better than lower values. Abbreviations used in the figure: \acf{AP}, \acf{FN}, \acf{FP}, \acf{FROC}, \acf{ImLC}, \acf{InS}, \acf{IoU}, \acf{ObD}, \acf{SemS}, \acf{TP}. Reference used in the figure: Kirillov et al., 2019: \cite{kirillov2019panoptic}. We recommend \ac{PQ} as a per-class counting metric in Subprocess S3 (Extended Data Fig.~\ref{fig:metric-selection:s3-single-threshold}).}
    \label{fig:cheat-sheet-pq}
\end{figure}

\begin{figure}[H]
    \centering
    \includegraphics[width=\textwidth]{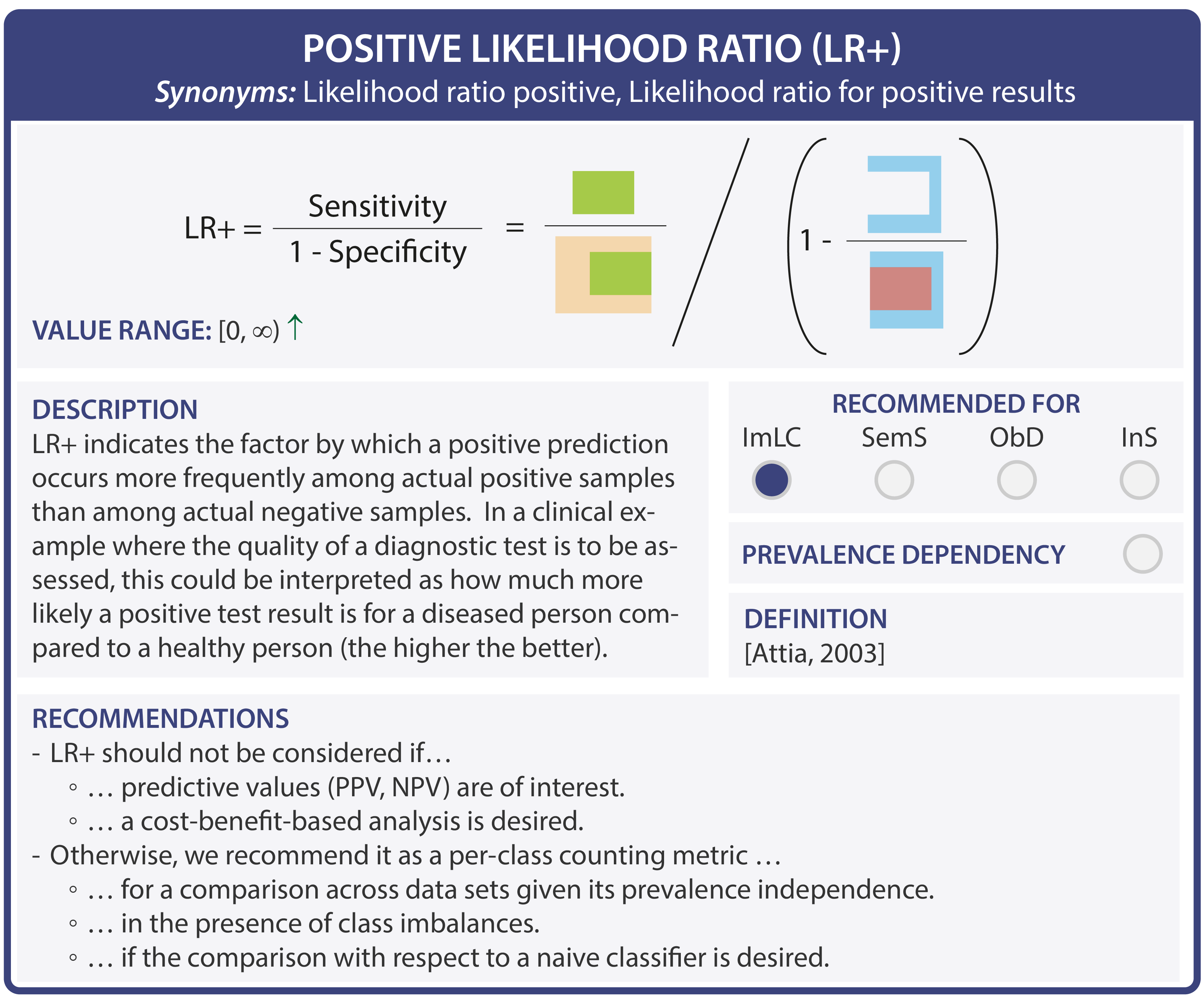}
    \caption{Cheat Sheet for the \acf{LR+}. The upward arrow in the value range indicates that higher values are better than lower values. Abbreviations used in the figure: \acf{ImLC}, \acf{InS}, \acf{ObD}, \acf{PPV}, \acf{SemS}. Reference used in the figure: Attia, 2003: \cite{wales2003moving}. We recommend \ac{LR+} as a per-class counting metric in Subprocess S3 (Extended Data Fig.~\ref{fig:metric-selection:s3-single-threshold}).}
    \label{fig:cheat-sheet-lr+}
\end{figure}

\begin{figure}[H]
    \centering
    \includegraphics[width=\textwidth]{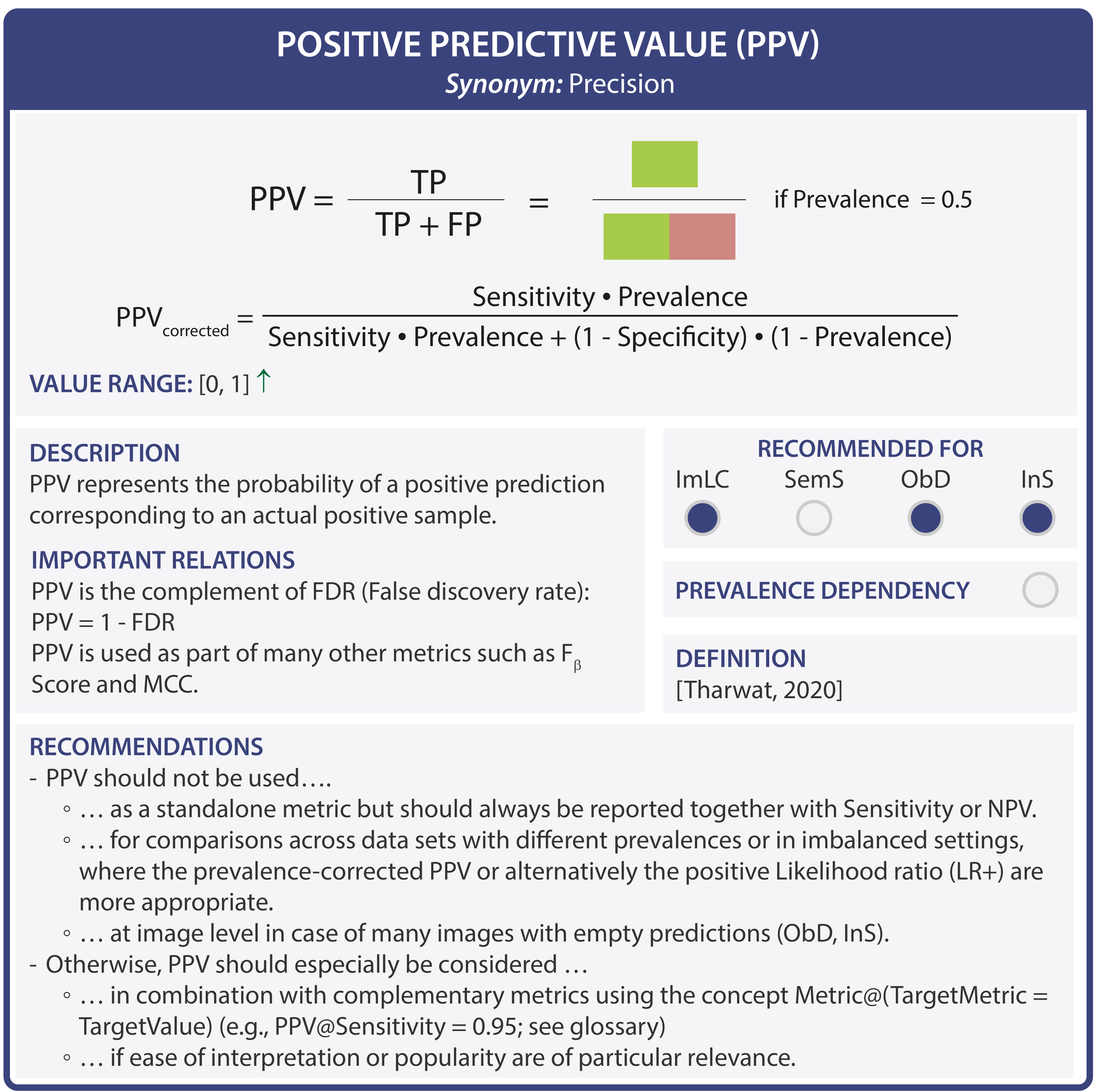}
    \caption{Cheat Sheet for the \acf{PPV}. The upward arrow in the value range indicates that higher values are better than lower values. Abbreviations used in the figure: \acf{FDR}, \acf{FP}, \acf{ImLC}, \acf{InS}, \acf{LR+}, \acf{MCC}, \acf{NPV}, \acf{ObD}, \acf{SemS}, \acf{TP}. Reference used in the figure: Tharwat, 2020:  \cite{tharwat2020classification}. We recommend \ac{PPV} as a per-class counting metric in Subprocess S3 (Extended Data Fig.~\ref{fig:metric-selection:s3-single-threshold}) for target value-based optimization using the concept Metric@(TargetMetric = TargetValue) (e.g., PPV@Sensitivity = 0.95; see glossary in Suppl. Note~\ref{app:terminology}).}
    \label{fig:cheat-sheet-ppv}
\end{figure}

\begin{figure}[H]
    \centering
    \includegraphics[width=\textwidth]{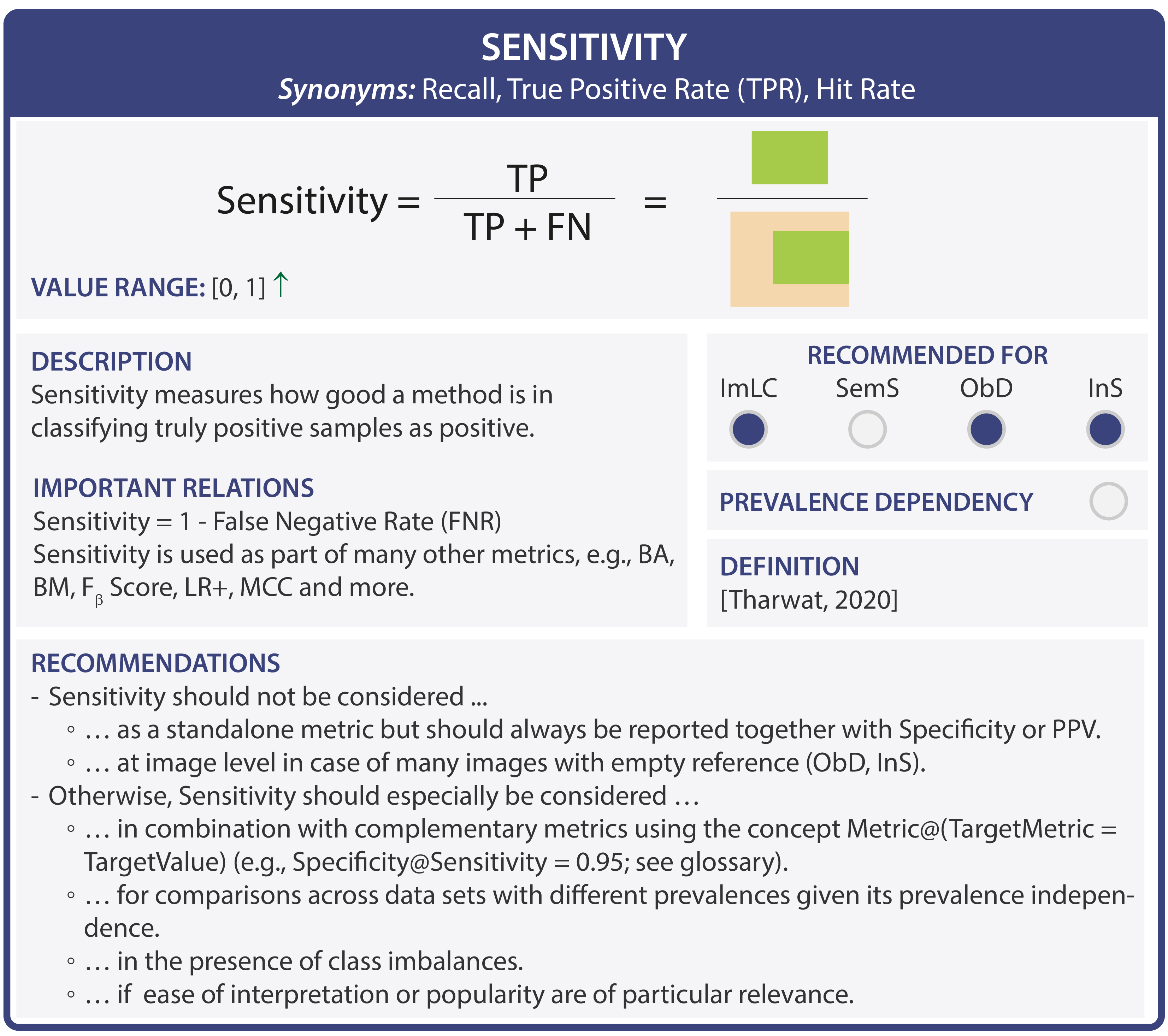}
    \caption{Cheat Sheet for the Sensitivity. The upward arrow in the value range indicates that higher values are better than lower values. Abbreviations used in the figure: \acf{BM}, \acf{FN}, \acf{ImLC}, \acf{InS}, \acf{LR+}, \acf{MCC}, \acf{ObD}, \acf{SemS}, \acf{TP}. Reference used in the figure: Tharwat, 2020:  \cite{tharwat2020classification}. We recommend Sensitivity as a per-class counting metric in Subprocess S3 (Extended Data Fig.~\ref{fig:metric-selection:s3-single-threshold}) for target value-based optimization using the concept Metric@(TargetMetric = TargetValue) (e.g., Specificity@Sensitivity = 0.95; see glossary in Suppl. Note~\ref{app:terminology}).}
    \label{fig:cheat-sheet-sensitivity}
\end{figure}

\begin{figure}[H]
    \centering
    \includegraphics[width=\textwidth]{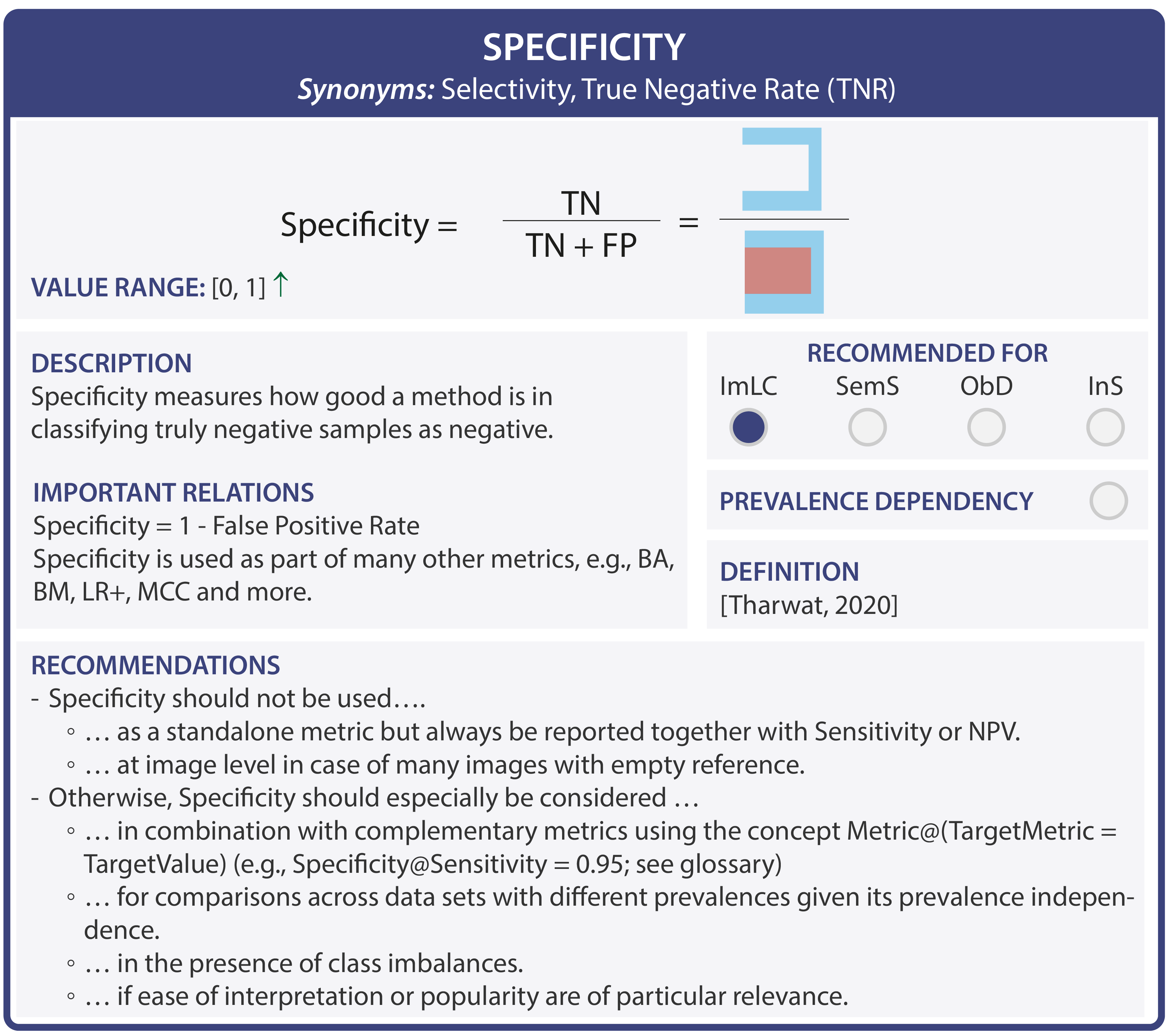}
    \caption{Cheat Sheet for the Specificity. The upward arrow in the value range indicates that higher values are better than lower values. Abbreviations used in the figure: \acf{BM}, \acf{FP}, \acf{ImLC}, \acf{InS}, \acf{LR+}, \acf{MCC}, \acf{ObD}, \acf{SemS}, \acf{TN}. Reference used in the figure: Tharwat, 2020:  \cite{tharwat2020classification}. We recommend Specificity as a per-class counting metric in Subprocess S3 (Extended Data Fig.~\ref{fig:metric-selection:s3-single-threshold}) for target value-based optimization using the concept Metric@(TargetMetric = TargetValue) (e.g., Specificity@Sensitivity = 0.95; see glossary in Suppl. Note~\ref{app:terminology}).}
    \label{fig:cheat-sheet-specificity}
\end{figure}

\begin{figure}[H]
    \centering
    \includegraphics[width=0.9\textwidth]{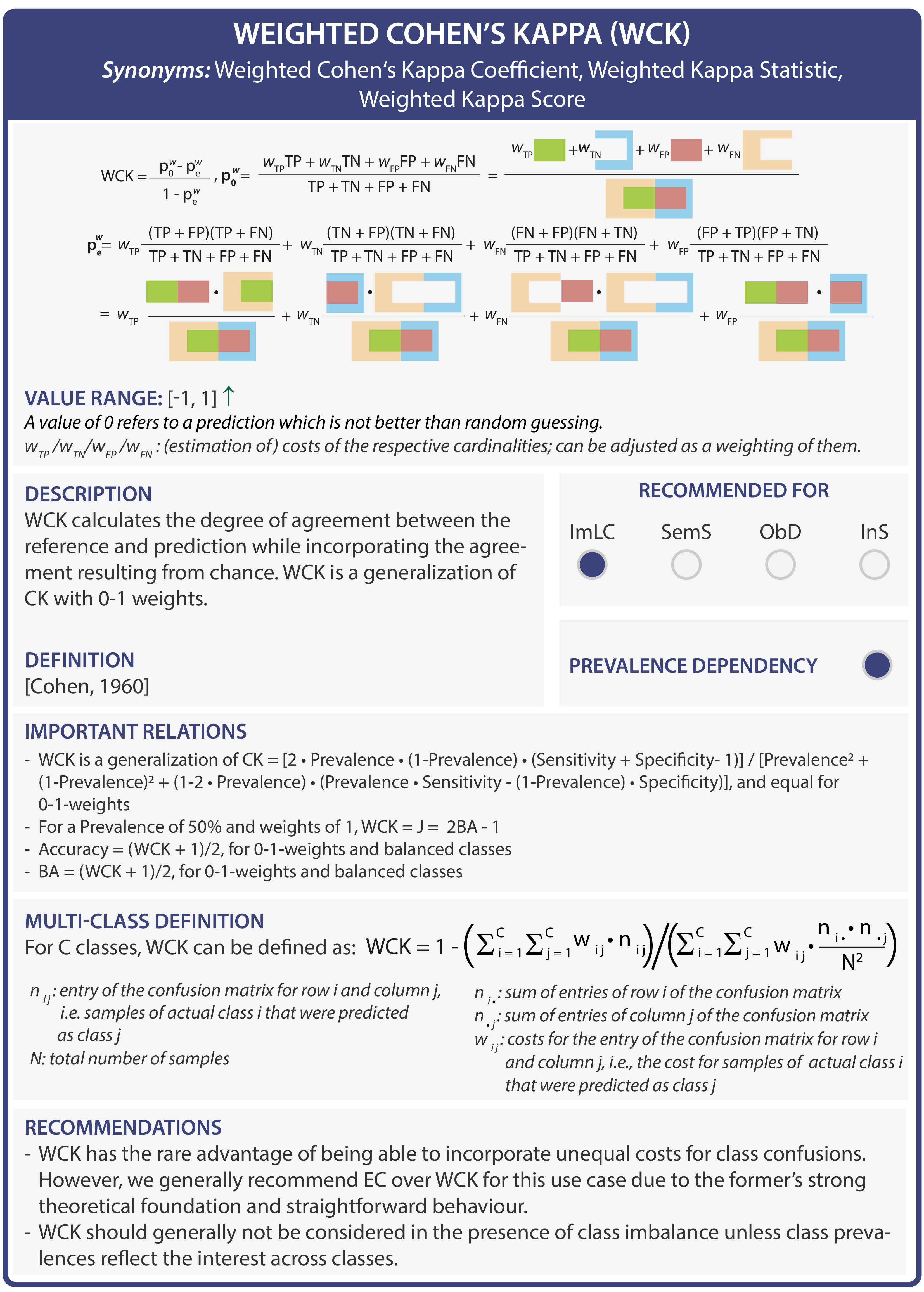}
    \caption{Cheat Sheet for the \acf{WCK}. The upward arrow in the value range indicates that higher values are better than lower values. Abbreviations used in the figure: \acf{BA}, \acf{CK}, \acf{FN}, \acf{FP}, \acf{ImLC}, \acf{InS}, \acf{ObD}, \acf{SemS}, \acf{TN}, \acf{TP}. Reference used in the figure: Cohen, 1960: \cite{cohen1960coefficient}. We recommend \ac{WCK} as a multi-class counting metric in Subprocess S2 (Extended Data Fig.~\ref{fig:metric-selection:s2-multi-class}).}
    \label{fig:cheat-sheet-wck}
\end{figure}

\begin{figure}[H]
    \centering
    \includegraphics[width=\textwidth]{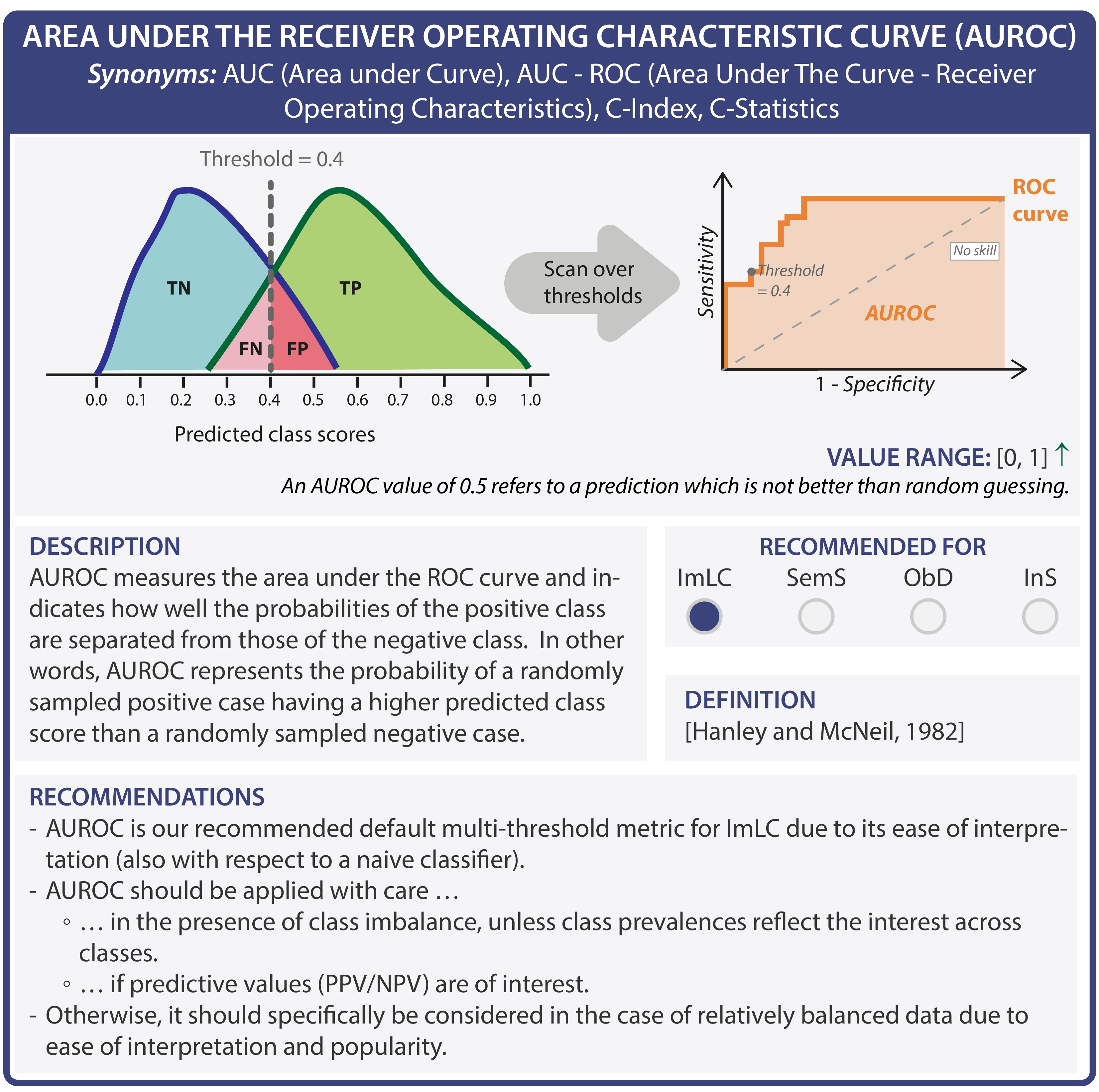}
    \caption{Cheat Sheet for the \acf{AUROC}. The upward arrow in the value range indicates that higher values are better than lower values. Abbreviations used in the figure: \acf{FN}, \acf{FP}, \acf{ImLC}, \acf{InS}, \acf{ObD}, \acf{ROC}, \acf{SemS}, \acf{TN}, \acf{TP}. Reference used in the figure: Hanley and McNeil, 1982: \cite{hanley1982meaning}. We recommend \ac{AUROC} as a multi-threshold metric in Subprocess S4 (Extended Data Fig.~\ref{fig:metric-selection:s4-multi-threshold}).}
    \label{fig:cheat-sheet-auroc}
\end{figure}

\begin{figure}[H]
    \centering
    \includegraphics[width=\textwidth]{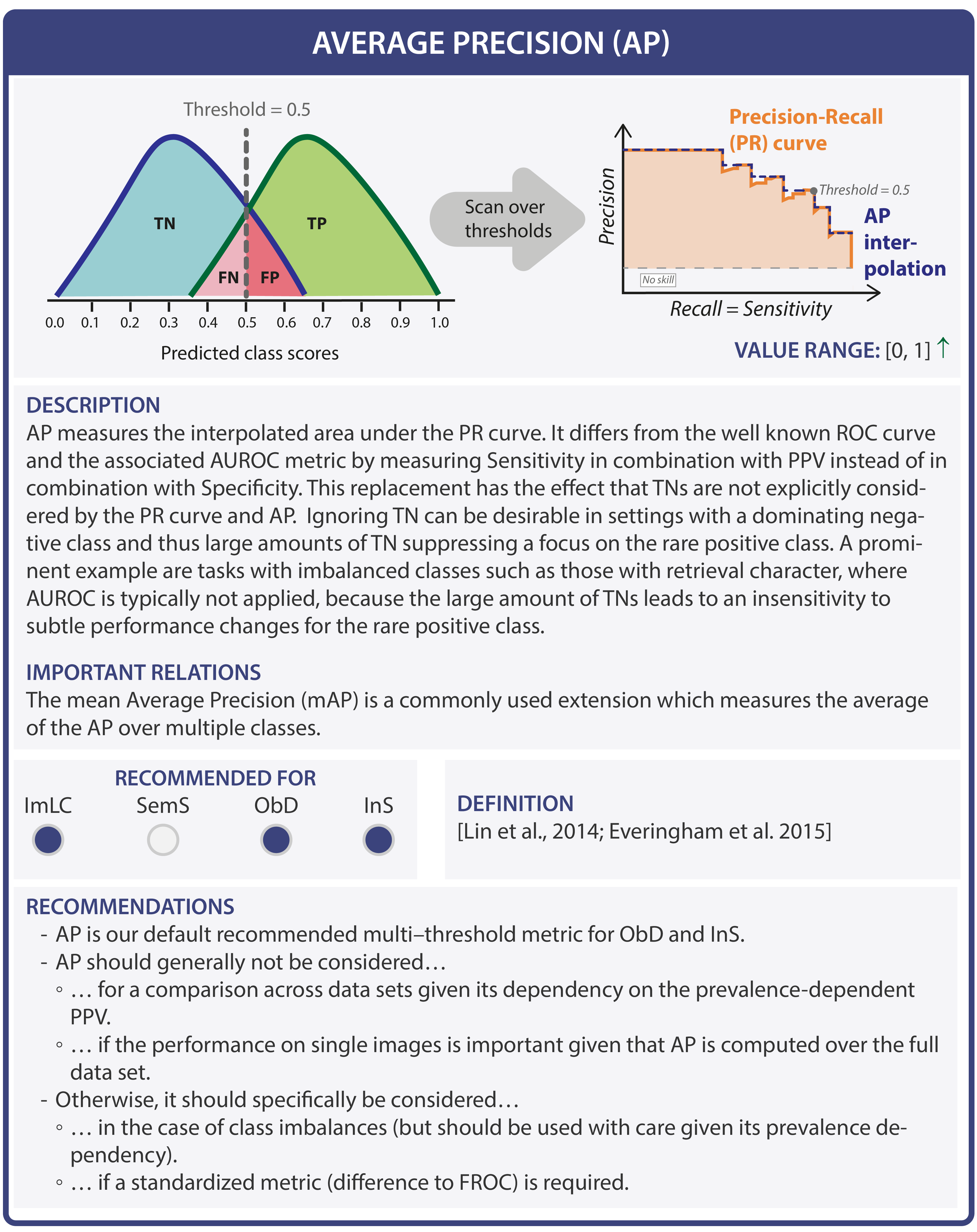}
    \caption{Cheat Sheet for the \acf{AP}. The upward arrow in the value range indicates that higher values are better than lower values. Abbreviations used in the figure: \acf{FN}, \acf{FP}, \acf{ImLC}, \acf{InS}, \acf{mAP}, \acf{ObD}, \acf{PPV}, \acf{PR}, \acf{SemS}, \acf{TN}, \acf{TP}. References used in the figure: Lin et al., 2014: \cite{lin2014microsoft}, Everingham et al., 2015: \cite{everingham2010pascal}. We recommend \ac{AP} as a multi-threshold metric in Subprocess S4 (Extended Data Fig.~\ref{fig:metric-selection:s4-multi-threshold}).}
    \label{fig:cheat-sheet-ap}
\end{figure}

\begin{figure}[H]
    \centering
    \includegraphics[width=\textwidth]{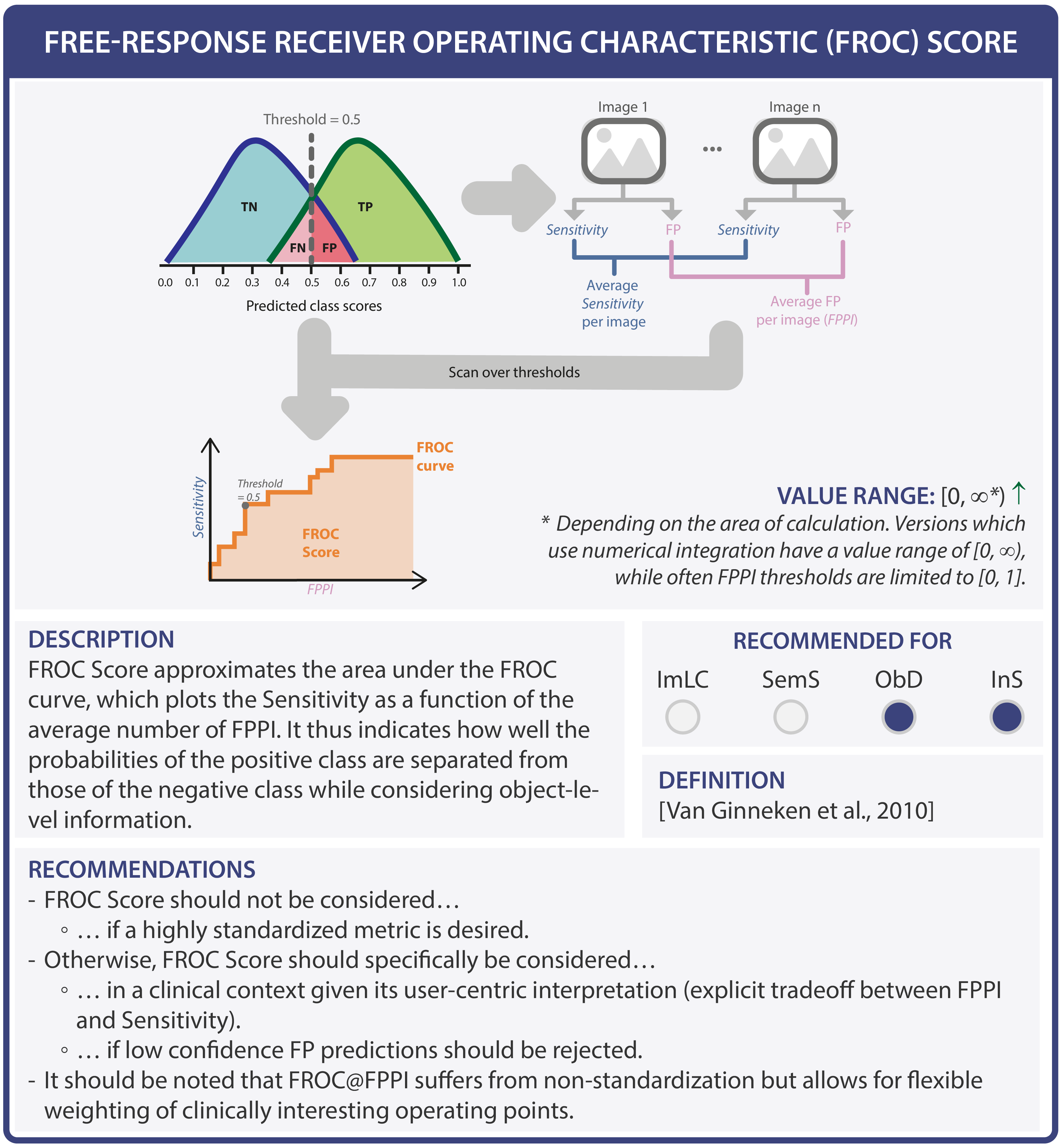}
    \caption{Cheat Sheet for the \acf{FROC}. The upward arrow in the value range indicates that higher values are better than lower values. Abbreviations used in the figure: \acf{FN}, \acf{FP}, \acf{FPPI}, \acf{ImLC}, \acf{InS}, \acf{ObD}, \acf{SemS}, \acf{TN}, \acf{TP}. Reference used in the figure: Van Ginneken et al., 2010: \cite{van2010comparing}. We recommend \ac{FROC} as a multi-threshold metric in Subprocess S4 (Extended Data Fig.~\ref{fig:metric-selection:s4-multi-threshold}).}
    \label{fig:cheat-sheet-froc}
\end{figure}

\begin{figure}[H]
    \centering
    \includegraphics[width=\textwidth]{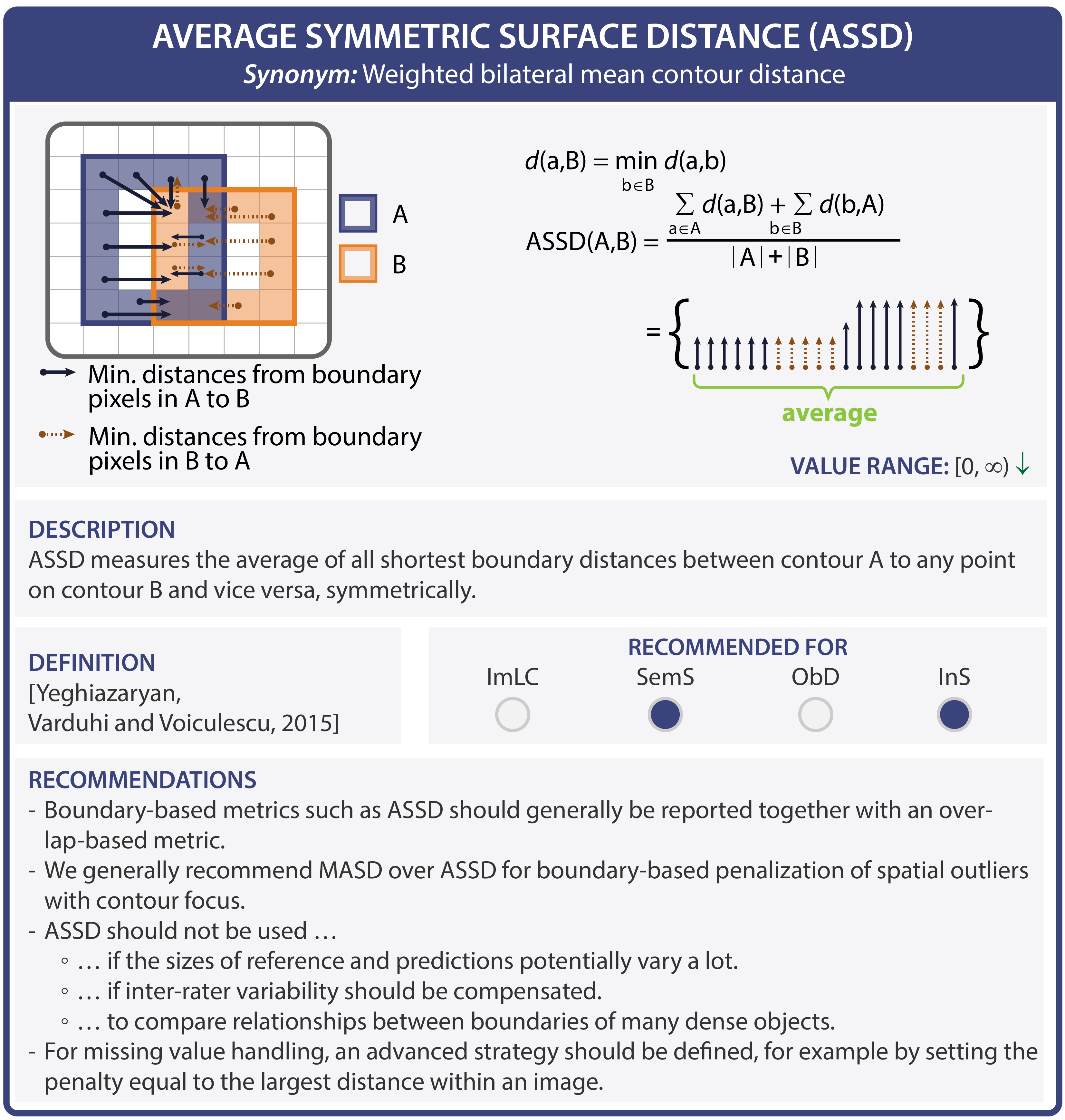}
    \caption{Cheat Sheet for the \acf{ASSD}. The downward arrow in the value range indicates that lower values are better than higher values. Abbreviations used in the figure: \acf{ImLC}, \acf{InS}, \acf{ObD}, \acf{SemS}. Reference used in the figure: Yeghiazaryan, Varduhi and Voiculescu, 2015: \cite{yeghiazaryan2015overview}. We recommend \acf{ASSD} as a boundary-based metric in Subprocess S7 (Extended Data Fig.~\ref{fig:metric-selection:s7-boundary}).}
    \label{fig:cheat-sheet-assd}
\end{figure}

\begin{figure}[H]
    \centering
    \includegraphics[width=\textwidth]{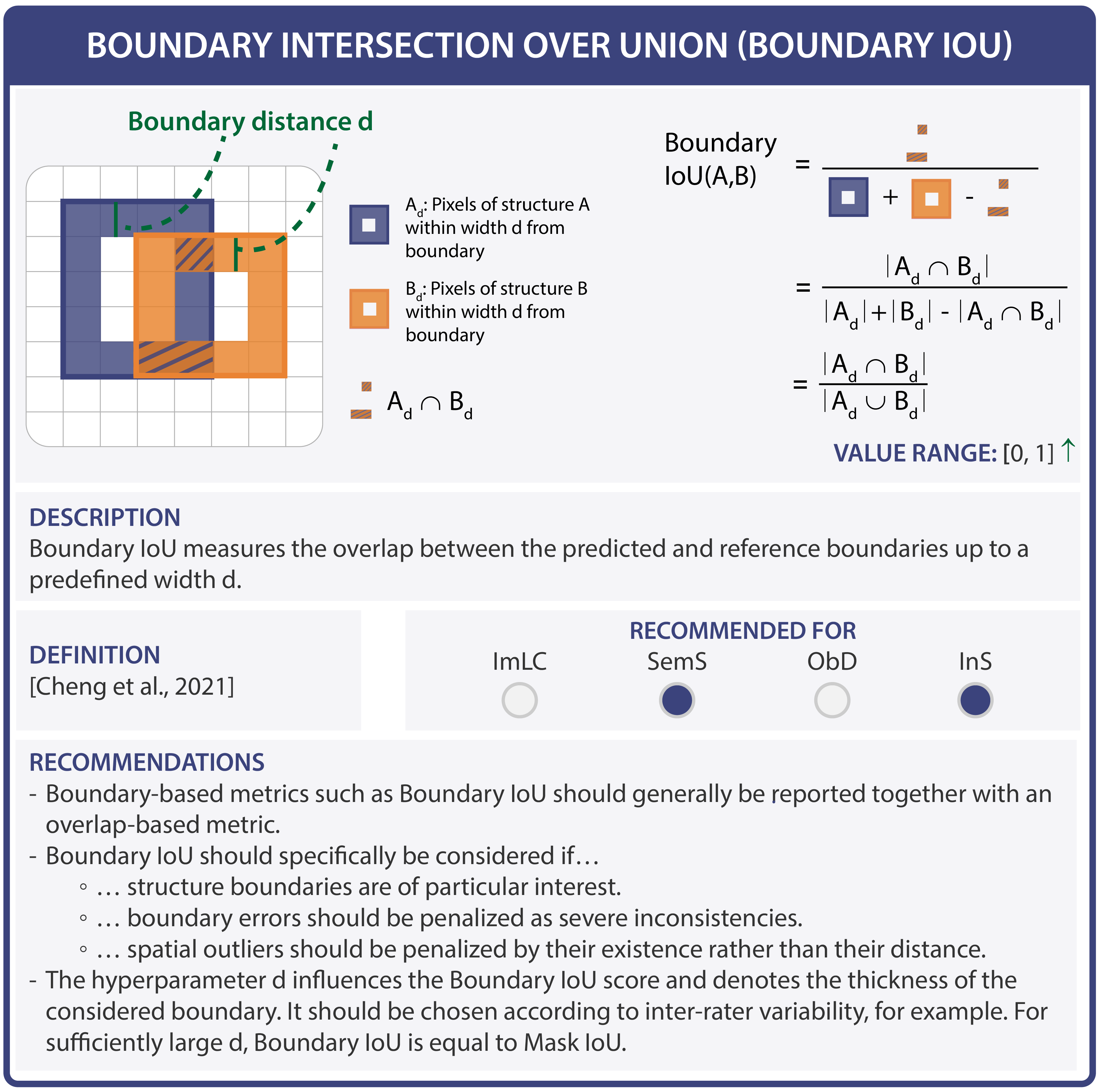}
    \caption{Cheat Sheet for the Boundary \acf{IoU}. The upward arrow in the value range indicates that higher values are better than lower values. Abbreviations used in the figure: \acf{ImLC}, \acf{InS}, \acf{MASD}, \acf{ObD}, \acf{SemS}. Reference used in the figure: Cheng et al., 2021: \cite{cheng2021boundary}. We recommend Boundary \ac{IoU} as a boundary-based metric in Subprocess S7 (Extended Data Fig.~\ref{fig:metric-selection:s7-boundary}).}
    \label{fig:cheat-sheet-boundary-iou}
\end{figure}

\begin{figure}[H]
    \centering
    \includegraphics[width=\textwidth]{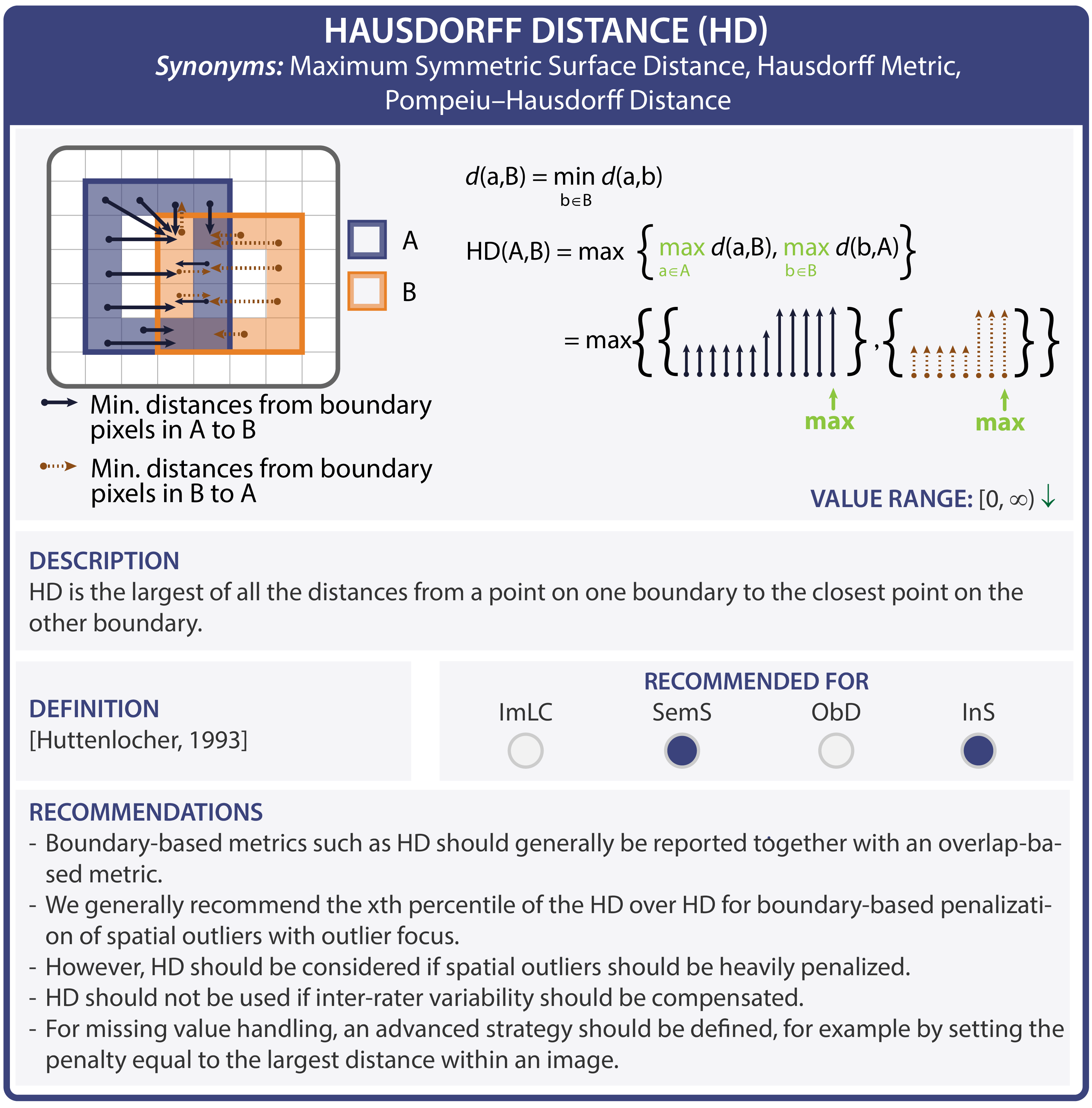}
    \caption{Cheat Sheet for the \acf{HD}. The downward arrow in the value range indicates that lower values are better than higher values. Abbreviations used in the figure: \acf{ImLC}, \acf{InS}, \acf{ObD}, \acf{SemS}. Reference used in the figure: Huttenlocher, 1993: \cite{huttenlocher1993comparing}. We recommend \ac{HD} as a boundary-based metric in Subprocess S7 (Extended Data Fig.~\ref{fig:metric-selection:s7-boundary}).}
    \label{fig:cheat-sheet-hd}
\end{figure}

\begin{figure}[H]
    \centering
    \includegraphics[width=\textwidth]{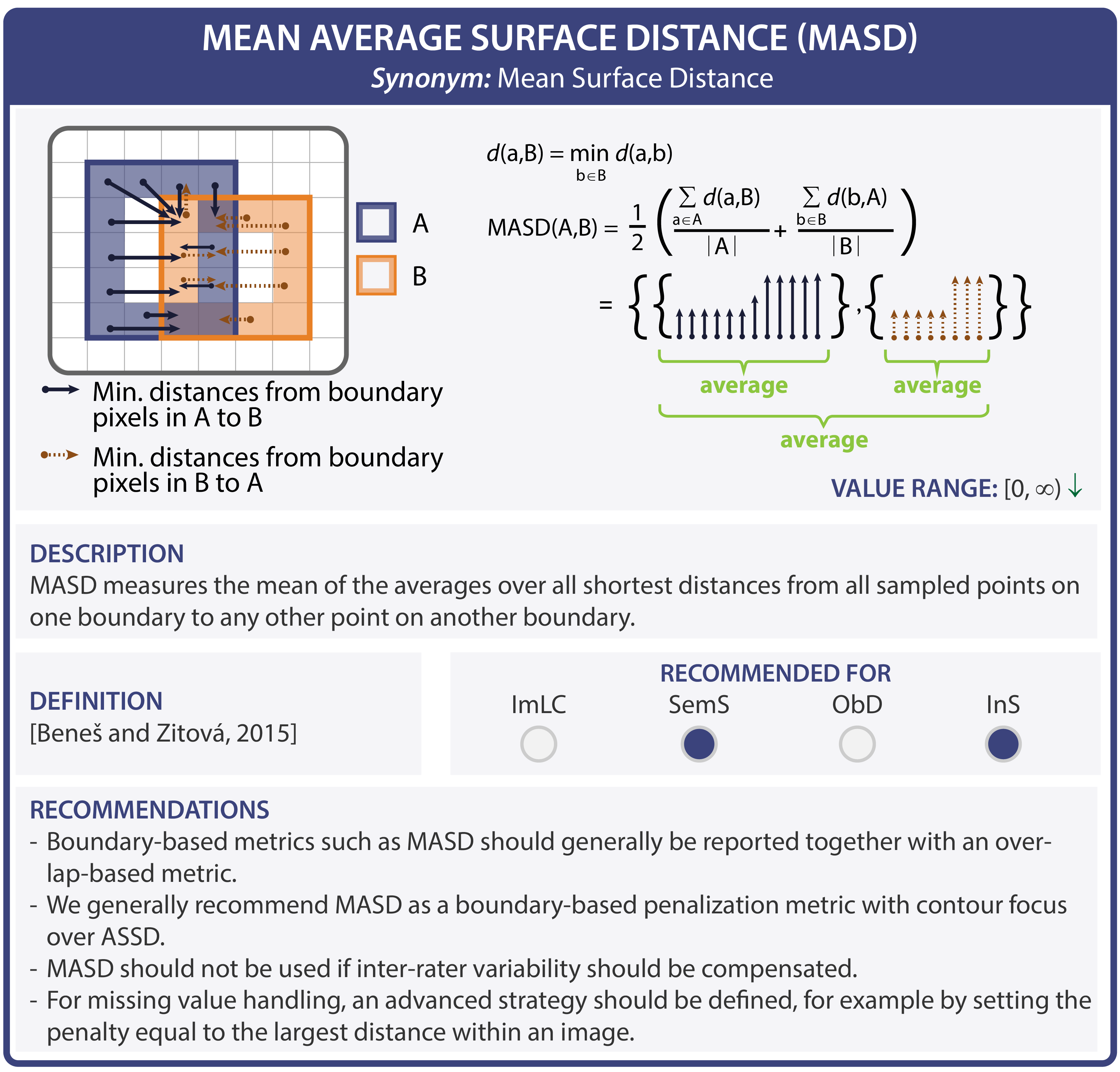}
    \caption{Cheat Sheet for the \acf{MASD}. The downward arrow in the value range indicates that lower values are better than higher values. Abbreviations used in the figure: \acf{ASSD}, \acf{ImLC}, \acf{InS}, \acf{ObD}, \acf{SemS}. Reference used in the figure: Beneš and Zitová, 2015: \cite{benevs2015performance}. We recommend \ac{MASD} as a boundary-based metric in Subprocess S7 (Extended Data Fig.~\ref{fig:metric-selection:s7-boundary}).}
    \label{fig:cheat-sheet-masd}
\end{figure}

\begin{figure}[H]
    \centering
    \includegraphics[width=\textwidth]{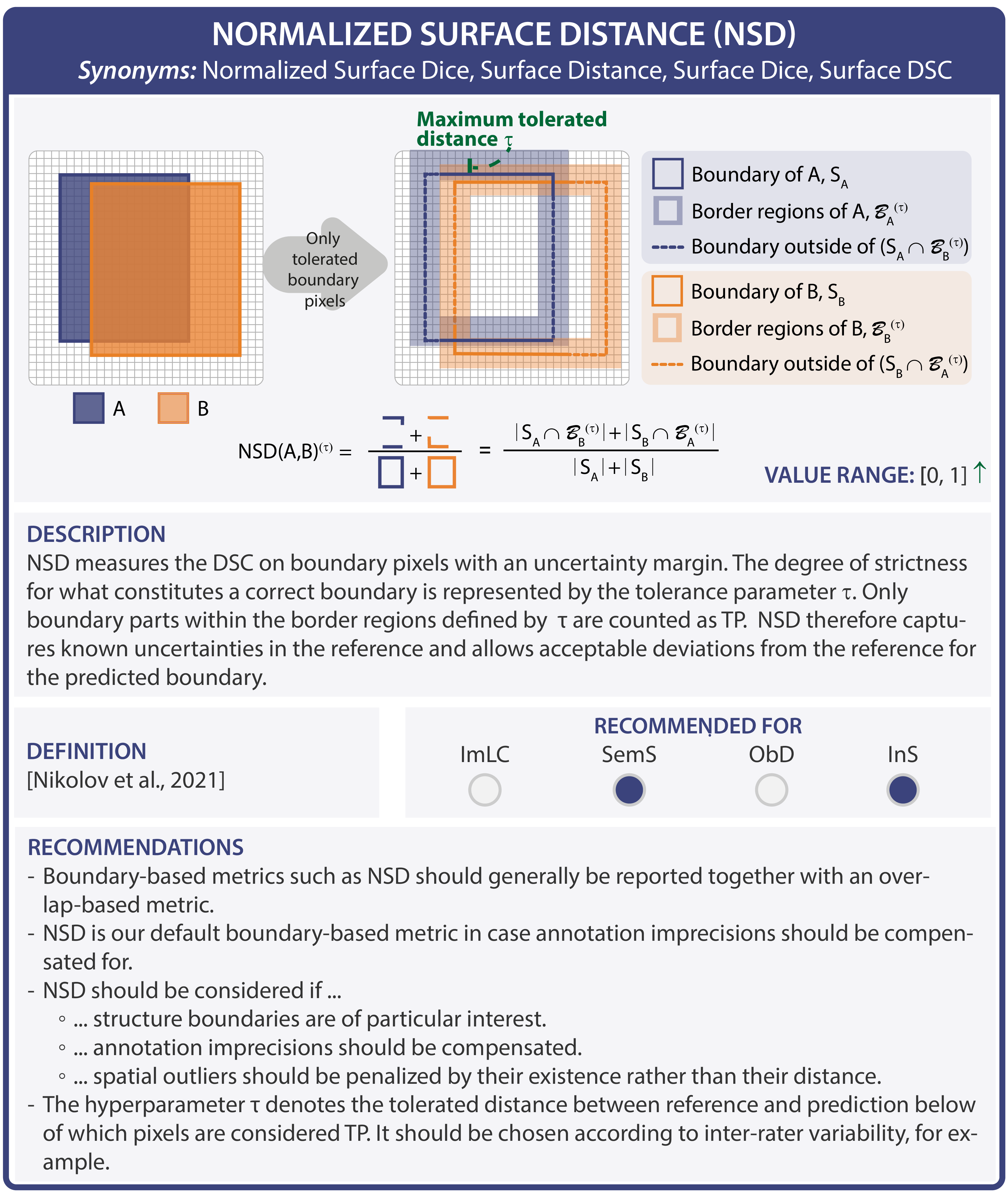}
    \caption{Cheat Sheet for the \acf{NSD}. The upward arrow in the value range indicates that higher values are better than lower values. Abbreviations used in the figure: \acf{DSC}, \acf{ImLC}, \acf{InS}, \acf{ObD}, \acf{SemS}, \acf{TP}. Reference used in the figure: Nikolov et al., 2021: \cite{nikolov2021clinically}. We recommend \ac{NSD} as a boundary-based metric in Subprocess S7 (Extended Data Fig.~\ref{fig:metric-selection:s7-boundary}).}
    \label{fig:cheat-sheet-nsd}
\end{figure}

\begin{figure}[H]
    \centering
    \includegraphics[width=\textwidth]{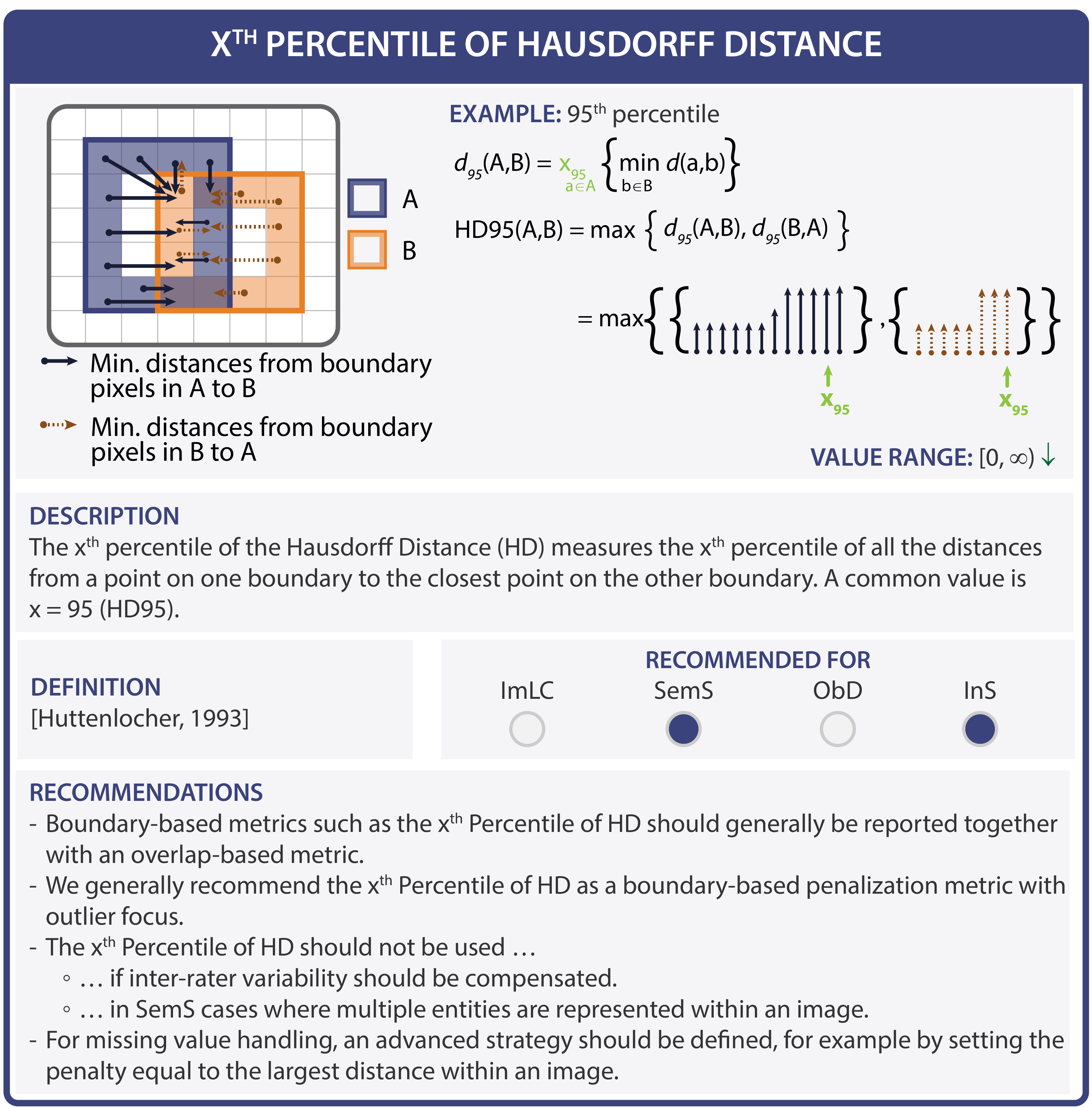}
    \caption{Cheat Sheet for the \acf{X$^{th}$ Percentile HD}. The downward arrow in the value range indicates that lower values are better than higher values. Abbreviations used in the figure: \acf{HD}, \acf{ImLC}, \acf{InS}, \acf{ObD}, \acf{SemS}. Reference used in the figure: Huttenlocher, 1993: \cite{huttenlocher1993comparing}. We recommend \ac{X$^{th}$ Percentile HD} as a boundary-based metric in Subprocess S7 (Extended Data Fig.~\ref{fig:metric-selection:s7-boundary}).}
    \label{fig:cheat-sheet-hd95}
\end{figure}

\begin{figure}[H]
    \centering
    \includegraphics[width=\textwidth]{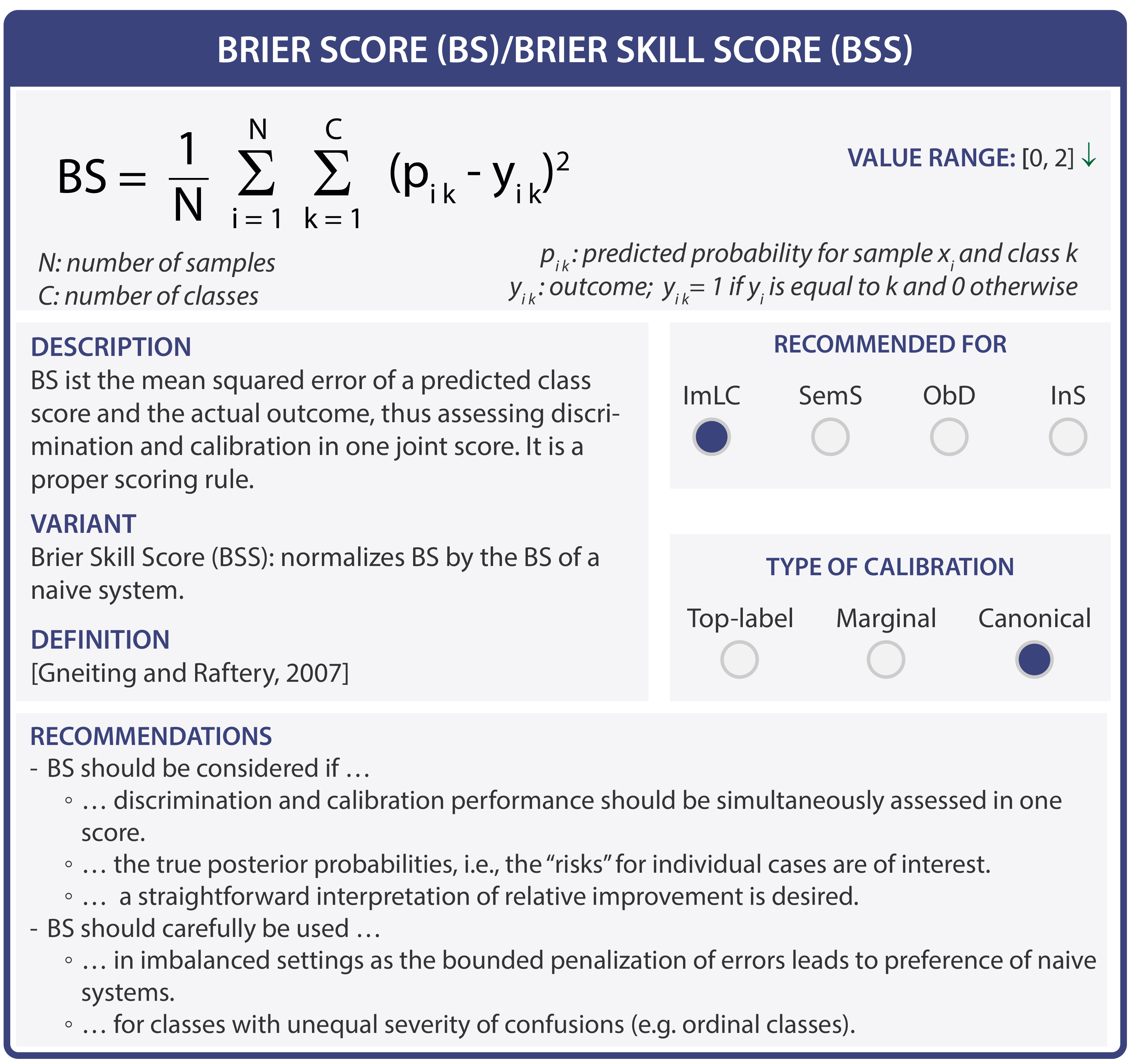}
    \caption{Cheat Sheet for the \acf{BS}/\acf{BSS}. The downward arrow in the value range indicates that lower values are better than higher values. Abbreviations used in the figure: \acf{BSS}, \acf{ImLC}, \acf{InS}, \acf{ObD}, \acf{SemS}. An introduction to calibration and corresponding terminology can be found in Suppl. Note~\ref{app:calibration}. Reference used in the figure: Gneiting and Raftery, 2007: \cite{gneiting2007strictly}. We recommend \ac{BS}/\acf{BSS} as a calibration metric in Subprocess S5 (Extended Data Fig.~\ref{fig:metric-selection:s5-calibration}).}
    \label{fig:cheat-sheet-bs}
\end{figure}

\begin{figure}[H]
    \centering
    \includegraphics[width=\textwidth]{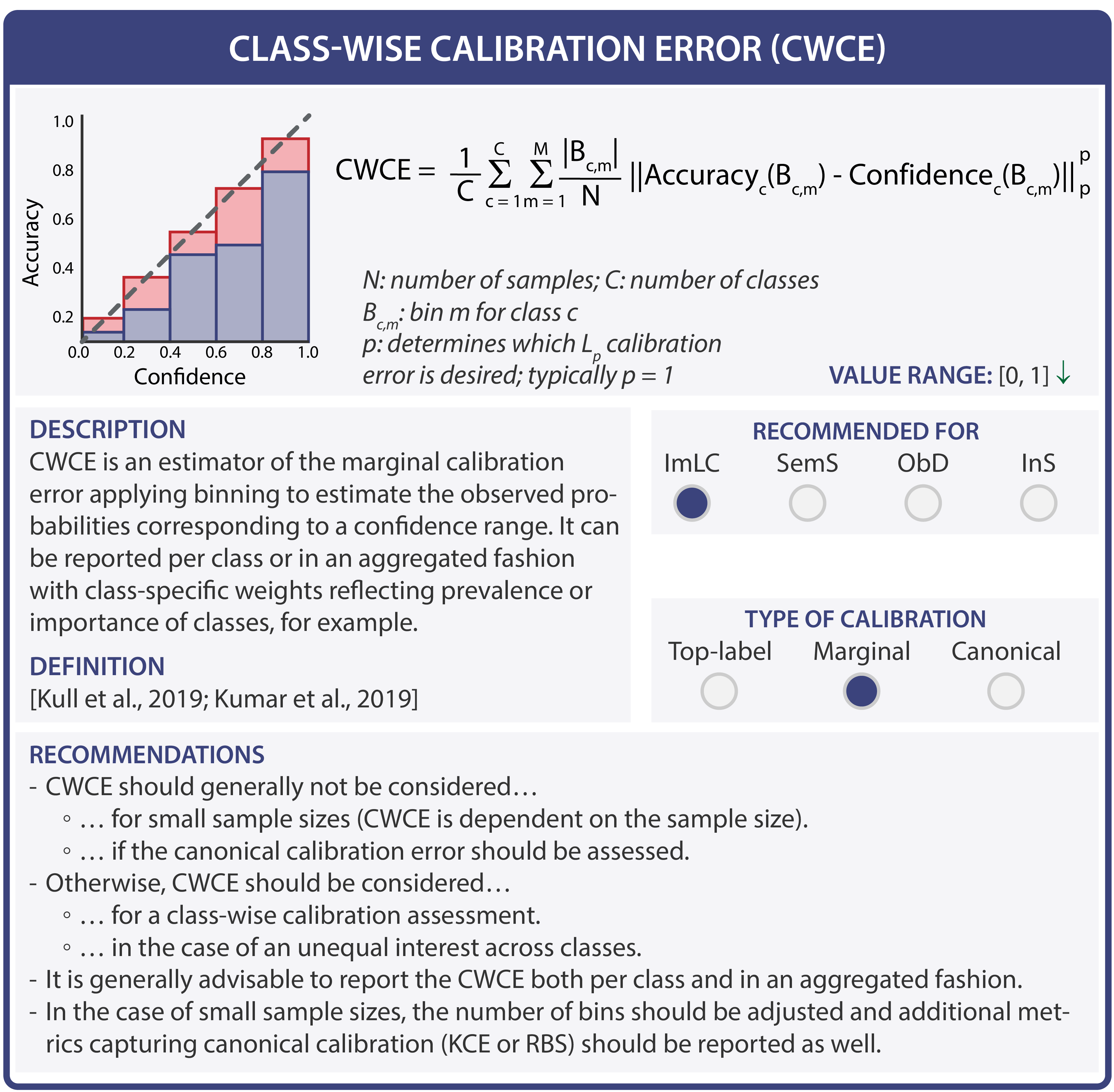}
    \caption{Cheat Sheet for the \acf{CWCE}. The downward arrow in the value range indicates that lower values are better than higher values. Abbreviations used in the figure: \acf{ImLC}, \acf{InS}, \acf{KCE}, \acf{ObD}, \acf{RBS}, \acf{SemS}. An introduction to calibration and corresponding terminology can be found in Suppl. Note~\ref{app:calibration}. References used in the figure: Kumar et al., 2019: \cite{kumar2019verified}, Kull et al., 2019: \cite{kull2019beyond}. We recommend \ac{CWCE} as a calibration metric in Subprocess S5 (Extended Data Fig.~\ref{fig:metric-selection:s5-calibration}).}
    \label{fig:cheat-sheet-cwce}
\end{figure}

\begin{figure}[H]
    \centering
    \includegraphics[width=\textwidth]{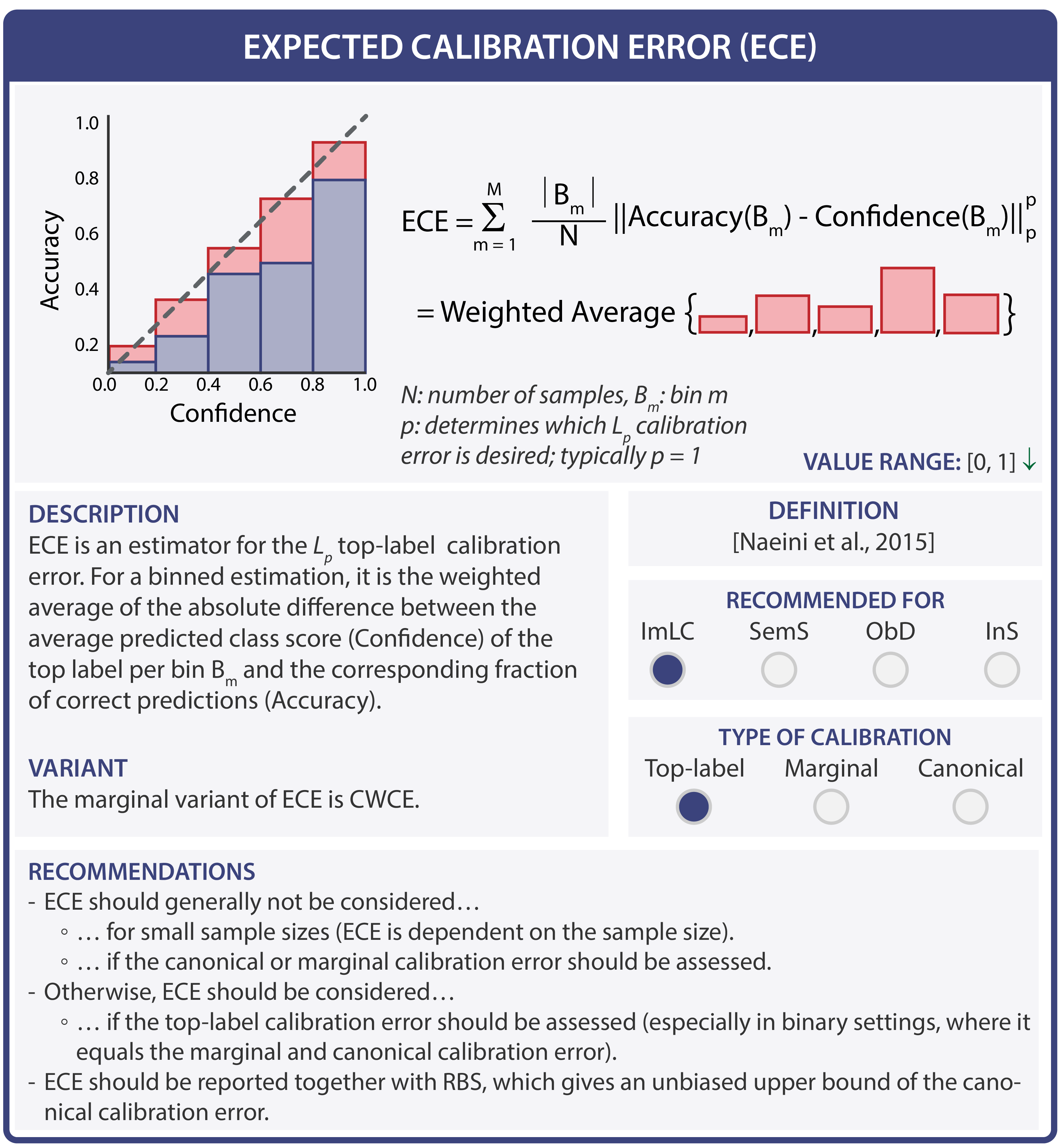}
    \caption{Cheat Sheet for the \acf{ECE}. The downward arrow in the value range indicates that lower values are better than higher values. Abbreviations used in the figure: \acf{ImLC}, \acf{InS}, \acf{ObD}, \acf{SemS}. An introduction to calibration and corresponding terminology can be found in Suppl. Note~\ref{app:calibration}. Reference used in the figure: Naeini et al., 2015: \cite{naeini2015obtaining}. We recommend \ac{ECE} as a calibration metric in Subprocess S5 (Extended Data Fig.~\ref{fig:metric-selection:s5-calibration}).}
    \label{fig:cheat-sheet-ece}
\end{figure}

\begin{figure}[H]
    \centering
    \includegraphics[width=\textwidth]{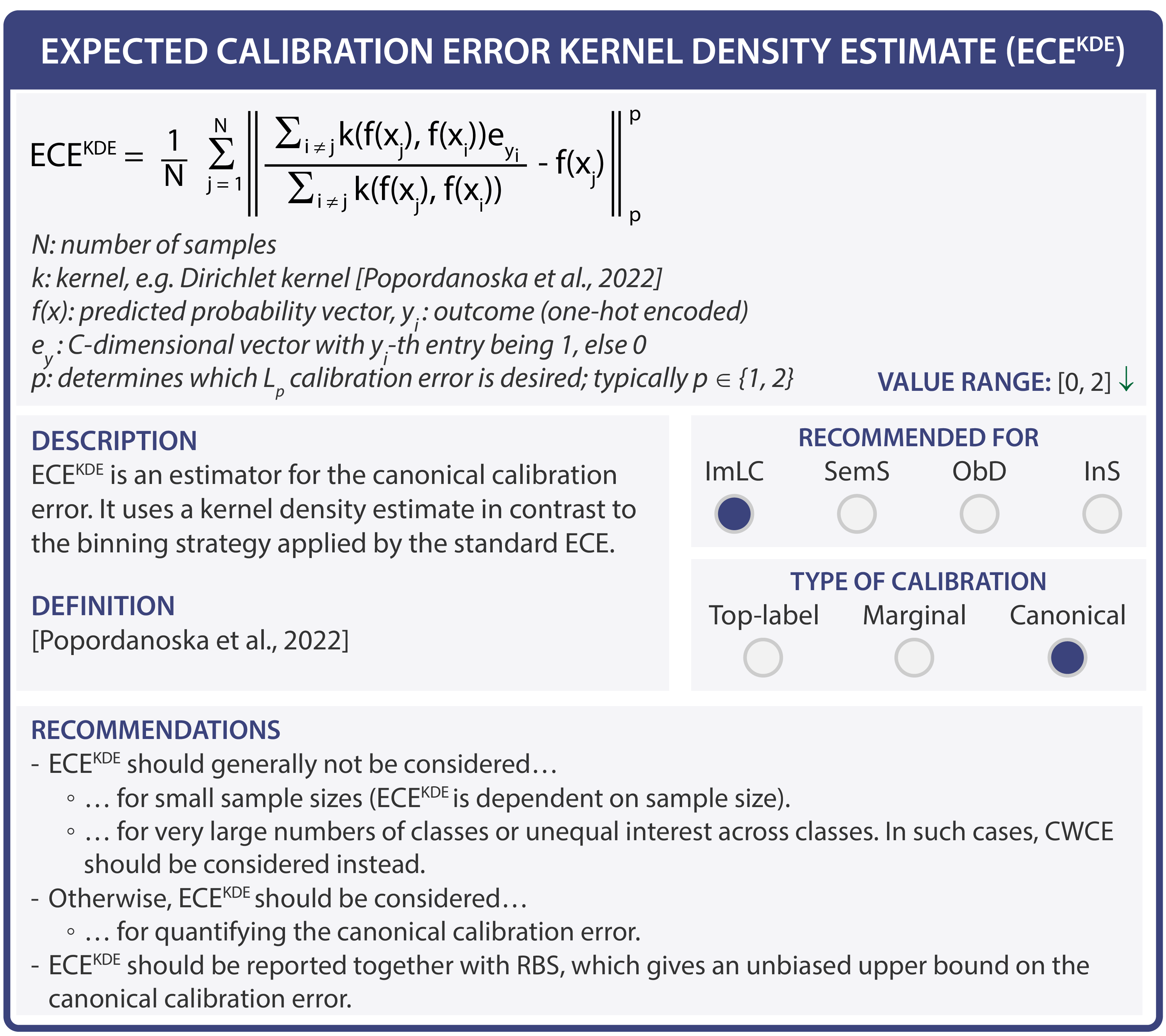}
    \caption{Cheat Sheet for the \acf{ECEKDE}. The downward arrow in the value range indicates that lower values are better than higher values. Abbreviations used in the figure: \acf{CWCE}, \acf{ImLC}, \acf{InS}, \acf{ObD}, \acf{RBS}, \acf{SemS}. An introduction to calibration and corresponding terminology can be found in Suppl. Note~\ref{app:calibration}. Reference used in the figure: Popordanoska et al., 2022: \cite{popordanoska2022consistent}. We recommend \ac{ECEKDE} as a calibration metric in Subprocess S5 (Extended Data Fig.~\ref{fig:metric-selection:s5-calibration}).}
    \label{fig:cheat-sheet-ecekde}
\end{figure}

\begin{figure}[H]
    \centering
    \includegraphics[width=\textwidth]{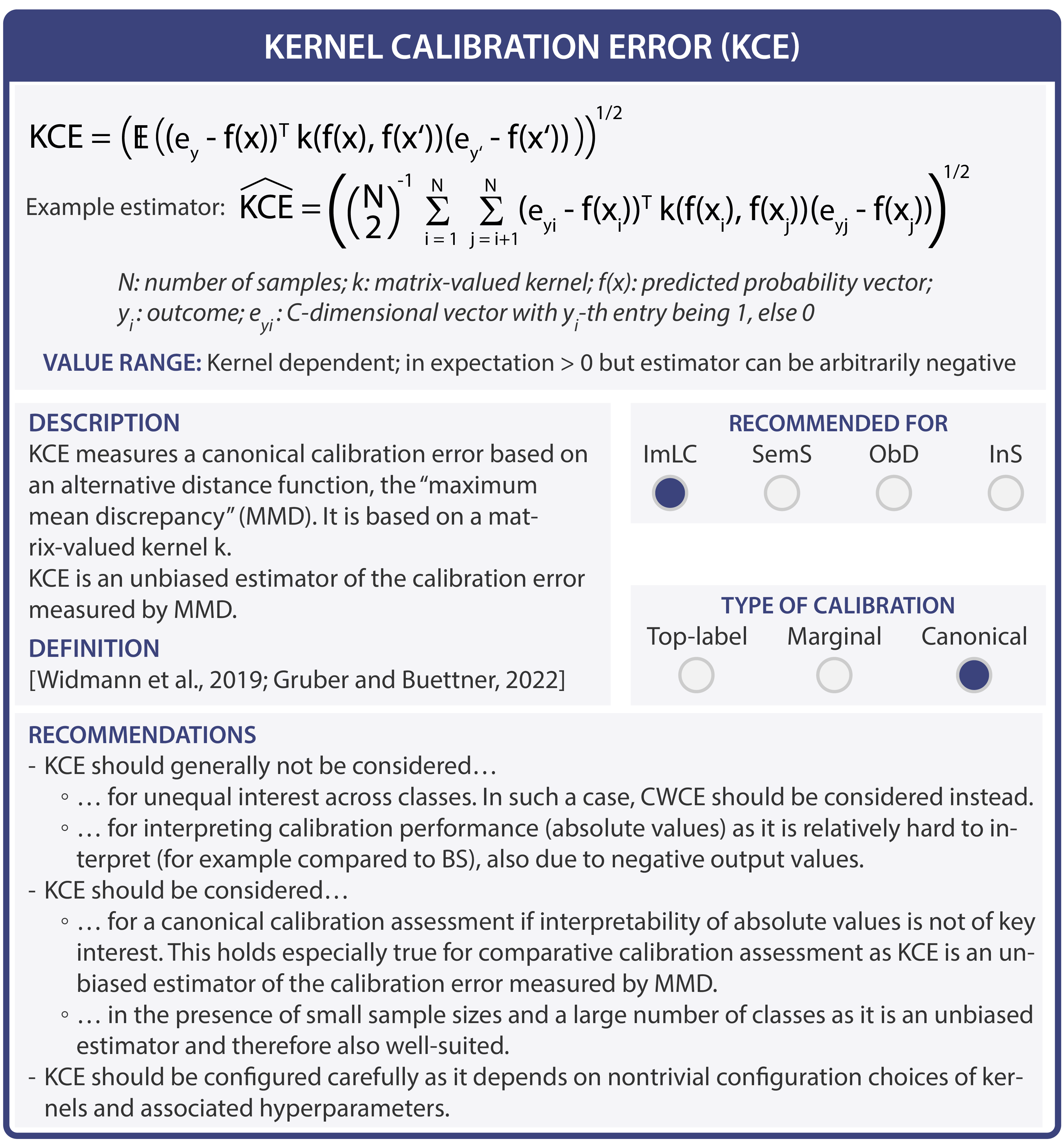}
    \caption{Cheat Sheet for the \acf{KCE}. The downward arrow in the value range indicates that lower values are better than higher values. Abbreviations used in the figure: \acf{BS}, \acf{ImLC}, \acf{InS}, \acf{ObD}, \acf{SemS}. An introduction to calibration and corresponding terminology can be found in Suppl. Note~\ref{app:calibration}. References used in the figure: Gruber and Buettner, 2022: \cite{gruber2022better}, Widmann et al., 2019: \cite{widmann2019calibration}. We recommend \ac{KCE} as a calibration metric in Subprocess S5 (Extended Data Fig.~\ref{fig:metric-selection:s5-calibration}).}
    \label{fig:cheat-sheet-kce}
\end{figure}

\begin{figure}[H]
    \centering
    \includegraphics[width=\textwidth]{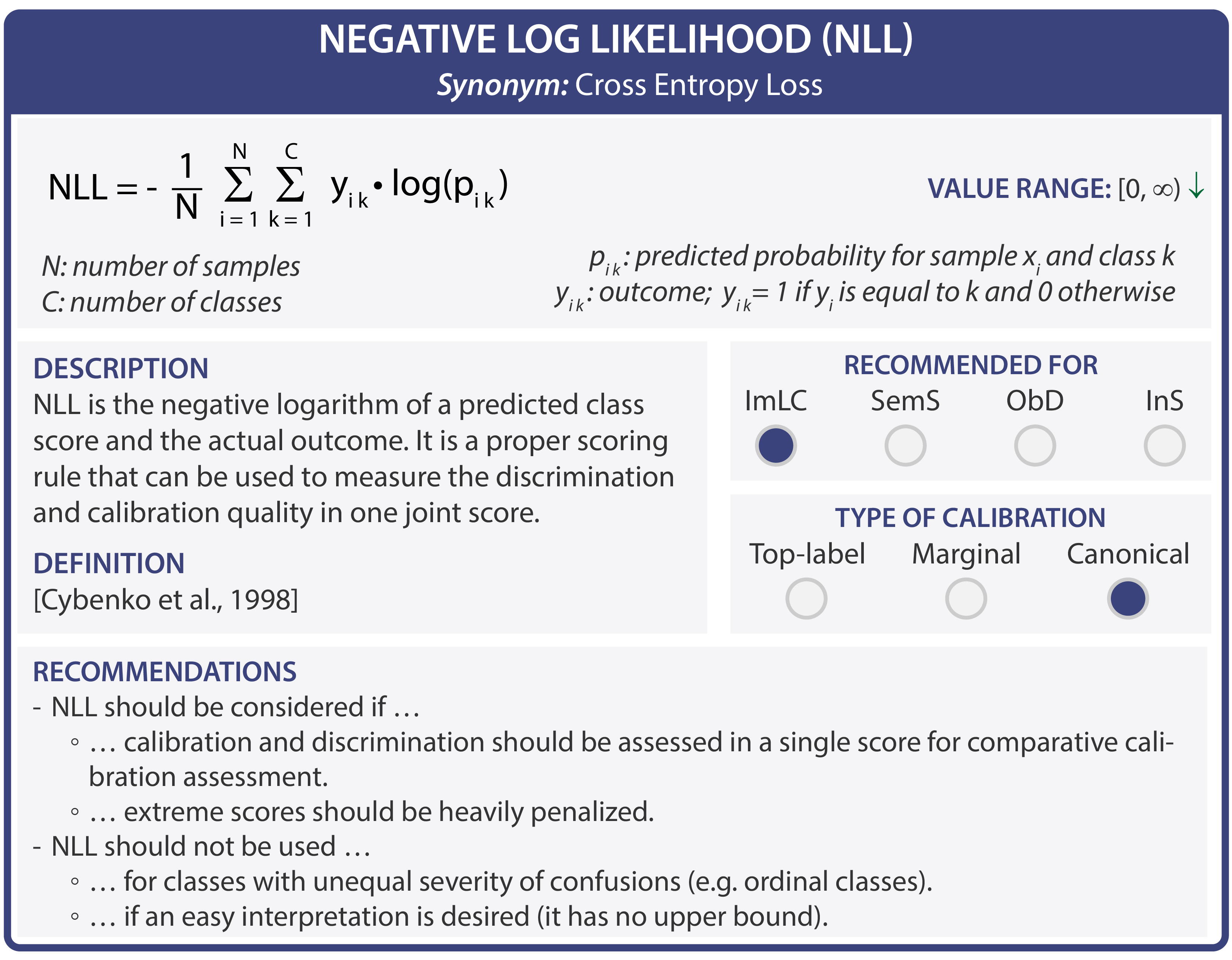}
    \caption{Cheat Sheet for the \acf{NLL}. The downward arrow in the value range indicates that lower values are better than higher values. Abbreviations used in the figure: \acf{ImLC}, \acf{InS}, \acf{ObD}, \acf{SemS}. An introduction to calibration and corresponding terminology can be found in Suppl. Note~\ref{app:calibration}. Reference used in the figure: Cybenko et al., 1998: \cite{cybenko1998mathematics}. We recommend \ac{NLL} as a calibration metric in Subprocess S5 (Extended Data Fig.~\ref{fig:metric-selection:s5-calibration}).}
    \label{fig:cheat-sheet-nll}
\end{figure}

\begin{figure}[H]
    \centering
    \includegraphics[width=\textwidth]{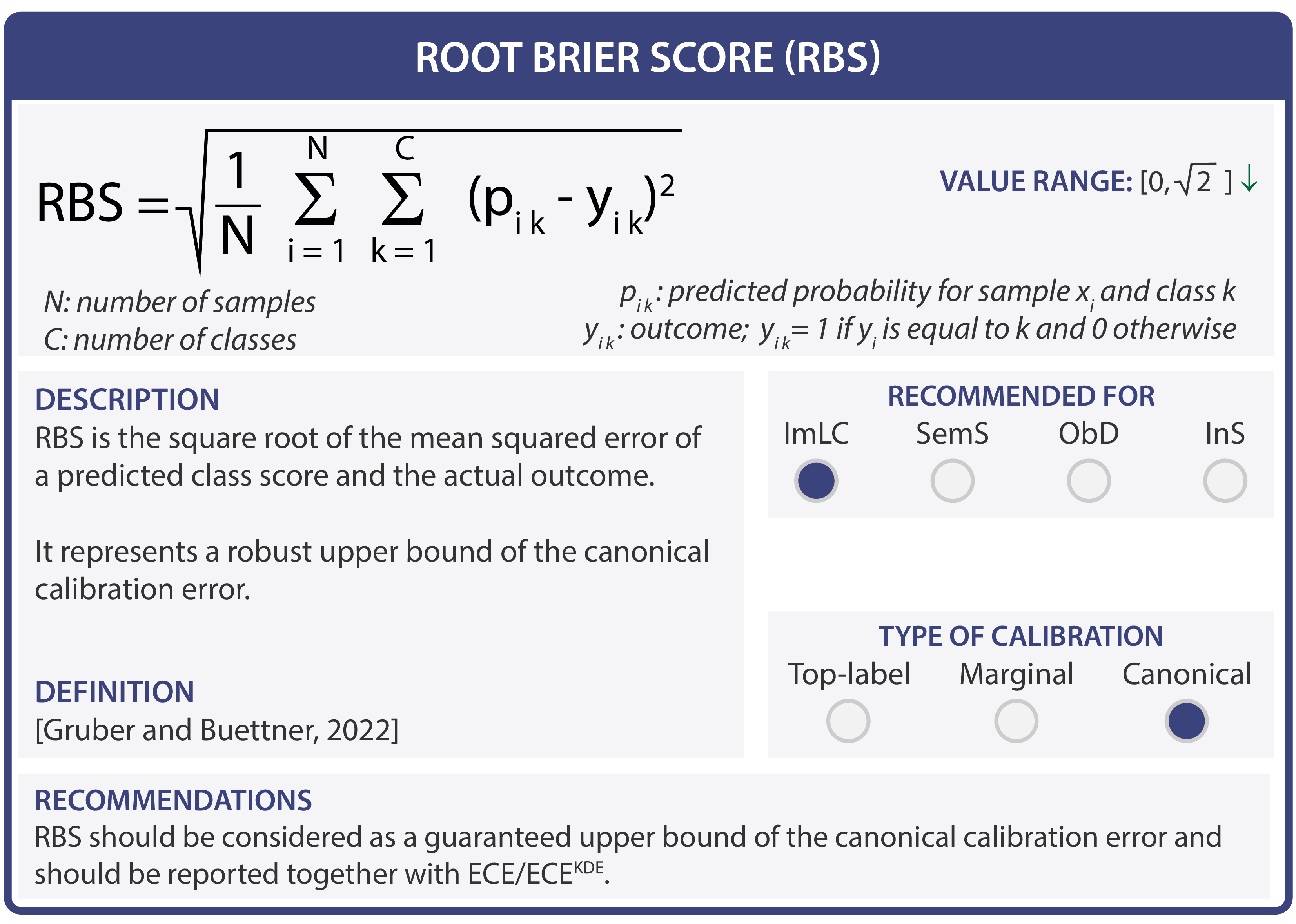}
    \caption{Cheat Sheet for the \acf{RBS}. The downward arrow in the value range indicates that lower values are better than higher values. Abbreviations used in the figure: \acf{ECE}, \acf{ECEKDE}, \acf{ImLC}, \acf{InS}, \acf{ObD}, \acf{SemS}. An introduction to calibration and corresponding terminology can be found in Suppl. Note~\ref{app:calibration}. Reference used in the figure: Gruber and Buettner, 2022: \cite{gruber2022better}. We recommend \ac{RBS} as a calibration metric in Subprocess S5 (Extended Data Fig.~\ref{fig:metric-selection:s5-calibration}).}
    \label{fig:cheat-sheet-rbs}
\end{figure}

\begin{figure}[H]
    \centering
    \includegraphics[width=\textwidth]{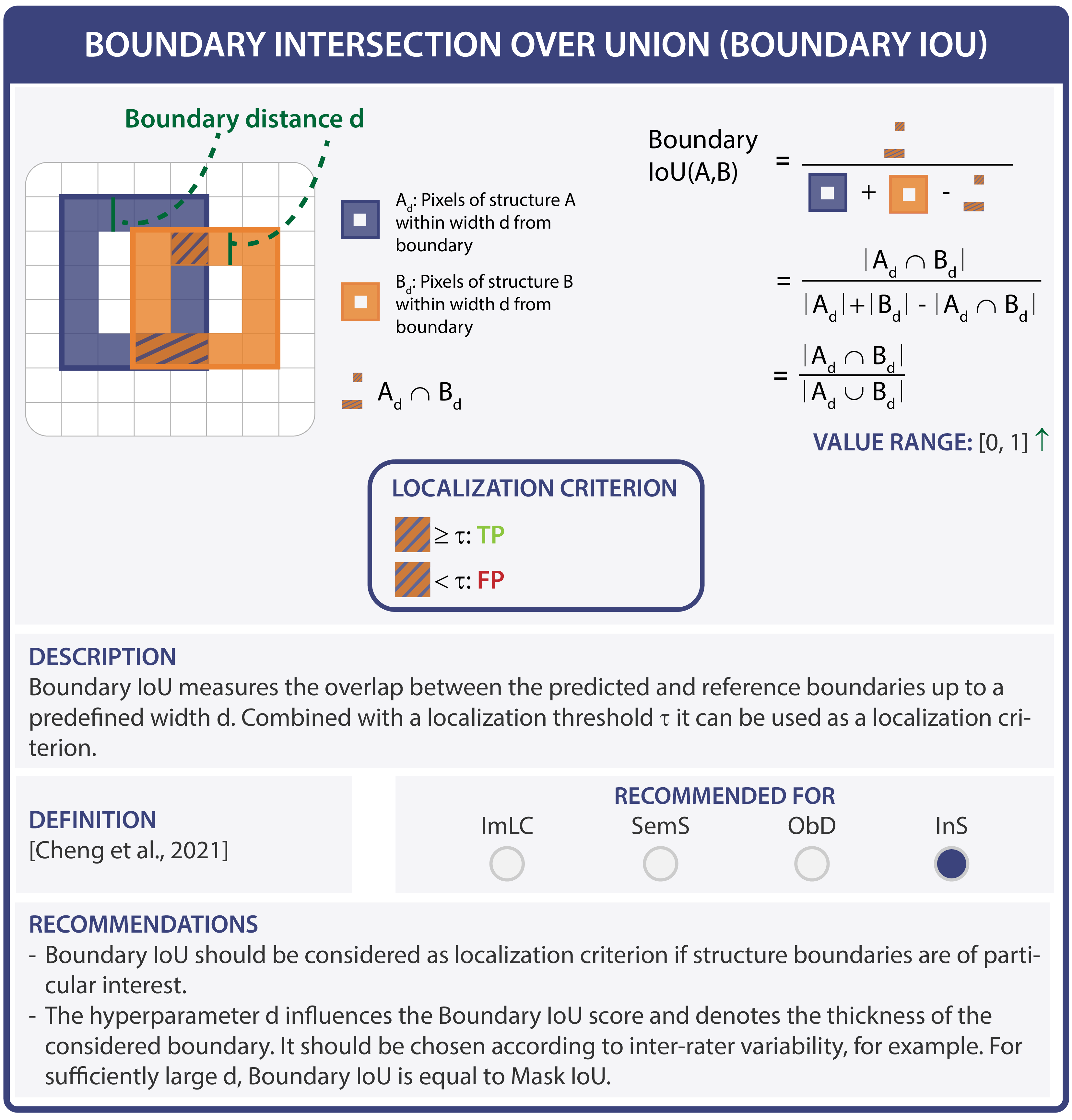}
    \caption{Metric profile of the Boundary \ac{IoU} (localization criterion). The upward arrow in the value range indicates that higher values are better than lower values. Abbreviations used in the figure: \acf{ImLC}, \acf{InS}, \acf{ObD}, \acf{SemS}. Reference used in the figure: Cheng et al., 2021: \cite{cheng2021boundary}. We recommend Boundary \acf{IoU} as a localization criterion in Subprocess S8 (Extended Data Fig.~\ref{fig:metric-selection:s8-localization-criterion}).}
    \label{fig:cheat-sheet-boundary-iou-localization-crit}
\end{figure}

\begin{figure}[H]
    \centering
    \includegraphics[width=\textwidth]{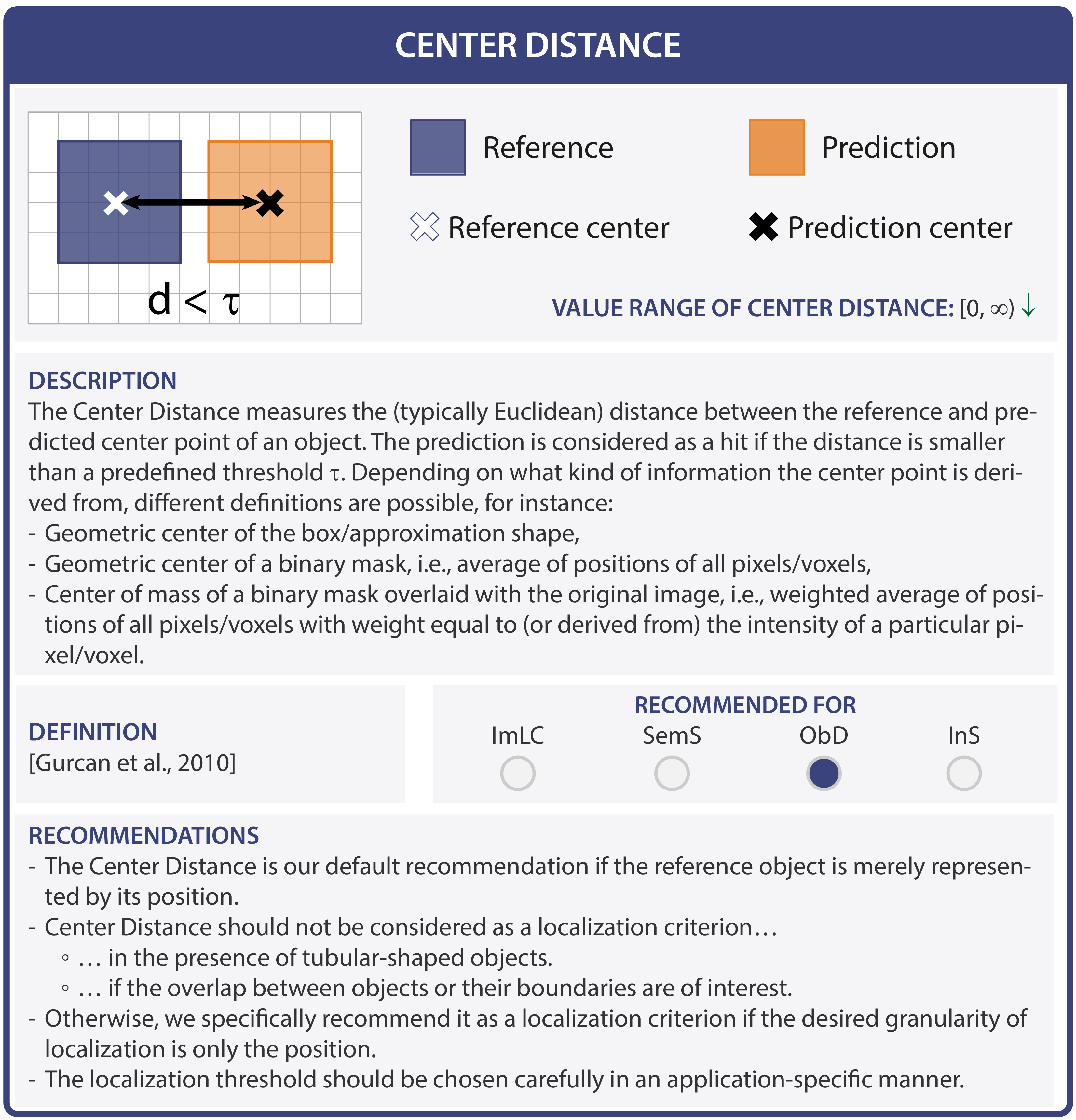}
    \caption{Cheat Sheet for the Center Distance. The downward arrow in the value range indicates that lower values are better than higher values. Abbreviations used in the figure: \acf{FN}, \acf{FP}, \acf{ImLC}, \acf{IoU}, \acf{InS}, \acf{ObD}, \acf{SemS}. Reference used in the figure: Gurcan et al., 2010: \cite{gurcan2010pattern}. We recommend Center Distance as a localization criterion in Subprocess S8 (Extended Data Fig.~\ref{fig:metric-selection:s8-localization-criterion}).}
    \label{fig:cheat-sheet-center-distance}
\end{figure}

\begin{figure}[H]
    \centering
    \includegraphics[width=\textwidth]{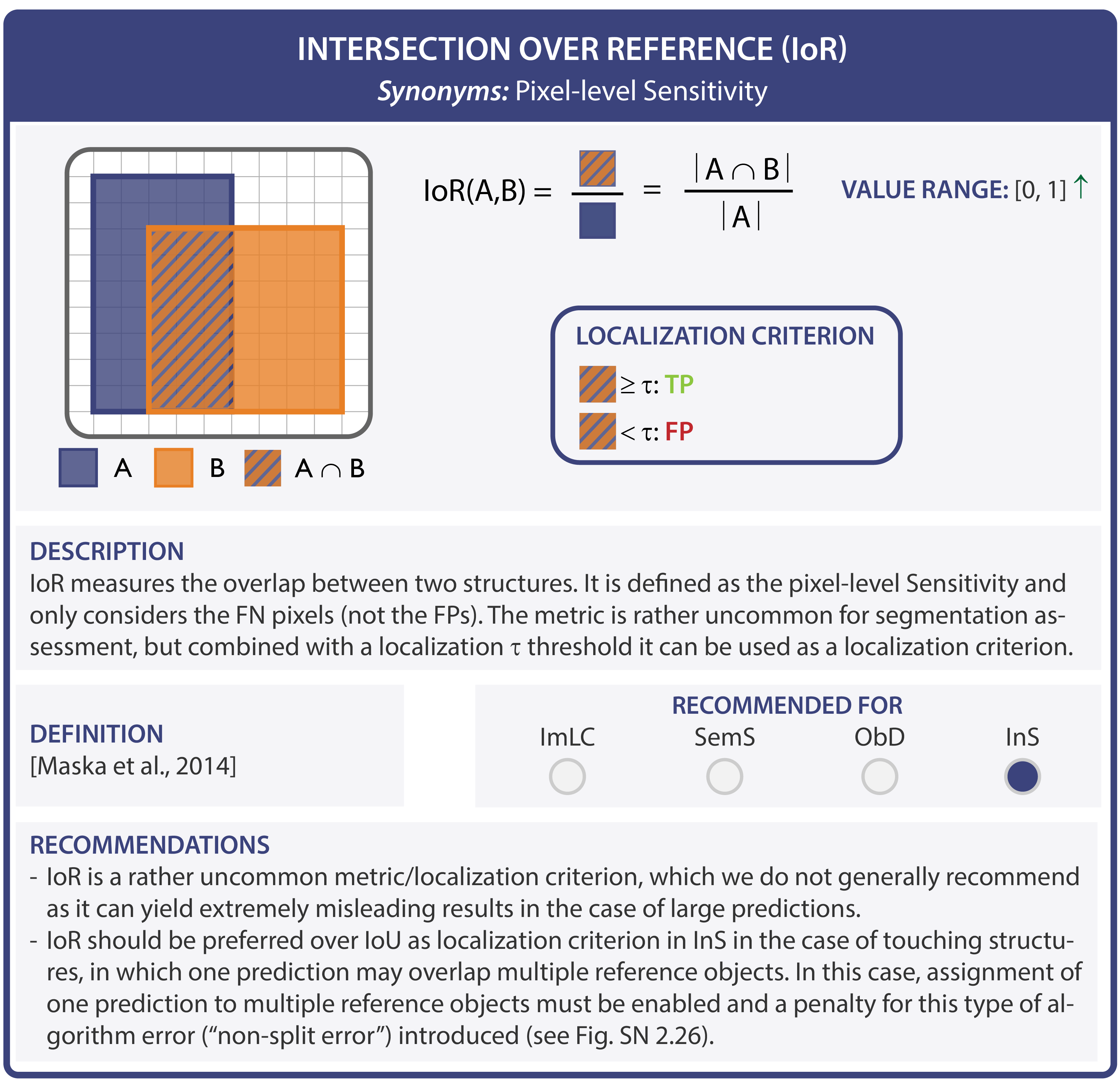}
    \caption{Cheat Sheet for the \acf{IoR}. The upward arrow in the value range indicates that higher values are better than lower values. Abbreviations used in the figure: \acf{ImLC}, \acf{InS}, \acf{ObD}, \acf{SemS}. Reference used in the figure: Maška et al., 2014: \cite{mavska2014benchmark}. We recommend \ac{IoR} as a localization criterion in Subprocess S8 (Extended Data Fig.~\ref{fig:metric-selection:s8-localization-criterion}).}
    \label{fig:cheat-sheet-ior}
\end{figure}

\begin{figure}[H]
    \centering
    \includegraphics[width=\textwidth]{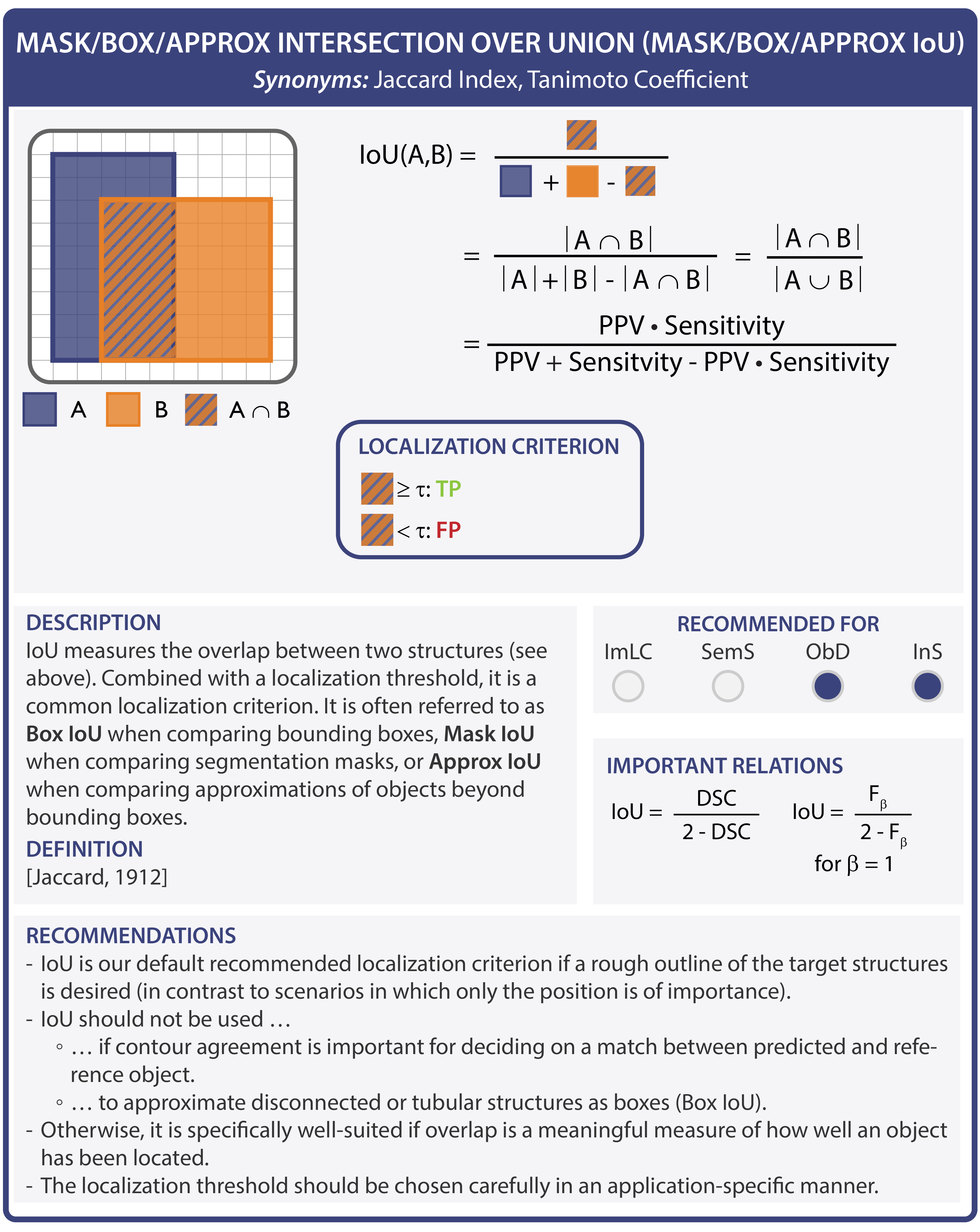}
    \caption{Metric profile of the Mask/Box/Approx \acf{IoU} (localization criterion). Abbreviations used in the figure: \acf{ImLC}, \acf{InS}, \acf{ObD}, \acf{PPV}, \acf{SemS}. Reference used in the figure: Jaccard, 1912: \cite{jaccard1912distribution}. We recommend Mask/Box/Approx \ac{IoU} as a localization criterion in Subprocess S8 (Extended Data Fig.~\ref{fig:metric-selection:s8-localization-criterion}).}
    \label{fig:cheat-sheet-iou-localization-crit}
\end{figure}

\begin{figure}[H]
    \centering
    \includegraphics[width=\textwidth]{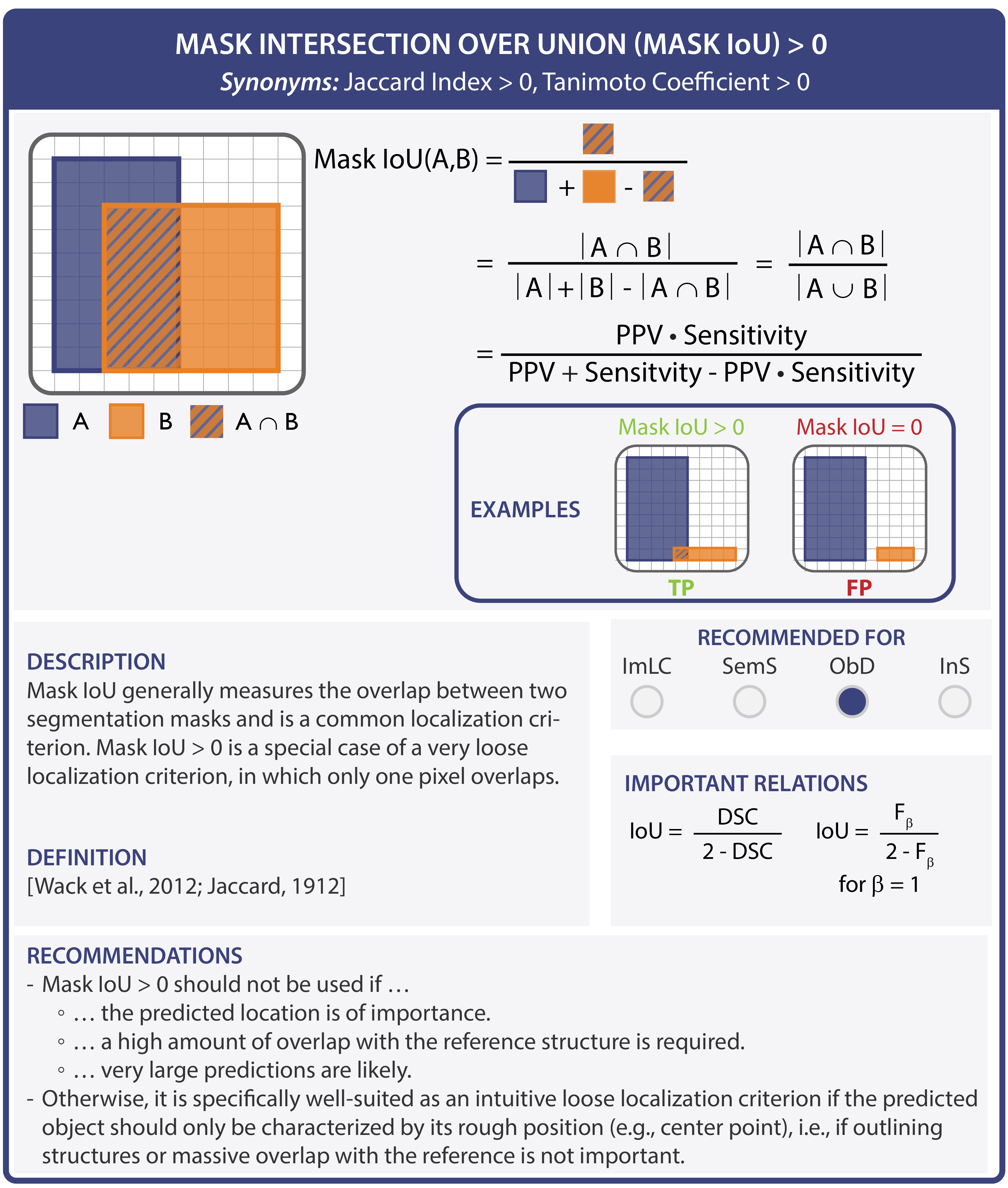}
    \caption{Metric profile of the Mask \acf{IoU} > 0. Abbreviations used in the figure: \acf{FP}, \acf{ImLC}, \acf{InS}, \acf{ObD}, \acf{PPV}, \acf{SemS}. References used in the figure: Jaccard, 1912: \cite{jaccard1912distribution}, Wack et al., 2012: \cite{wack2012improved}. We recommend Mask \ac{IoU} > 0 as a localization criterion in Subprocess S8 (Extended Data Fig.~\ref{fig:metric-selection:s8-localization-criterion}).}
    \label{fig:cheat-sheet-iou-localization-crit-0}
\end{figure}

\begin{figure}[H]
    \centering
    \includegraphics[width=\textwidth]{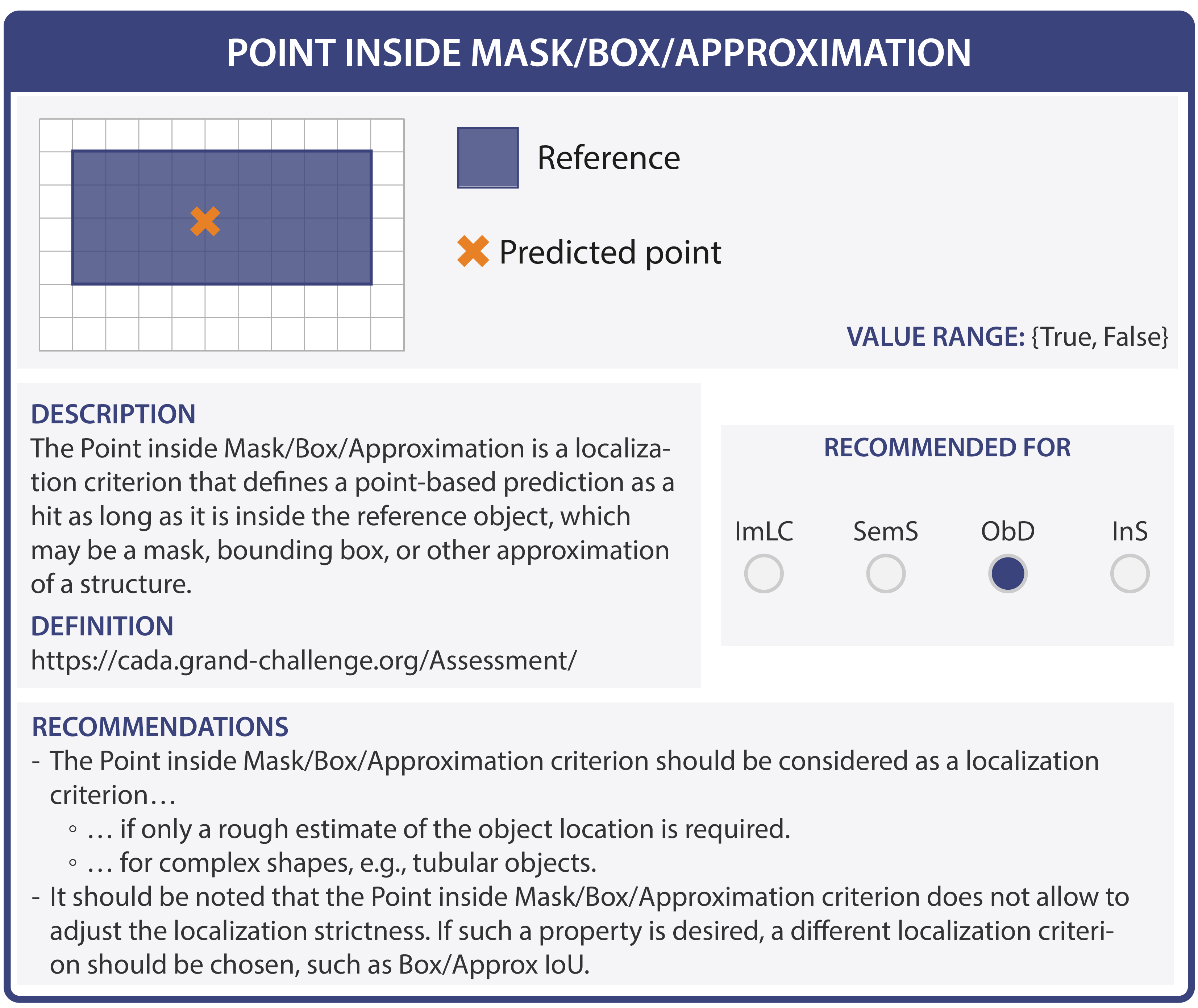}
    \caption{Cheat Sheet for the Point inside Mask/Box/Approximation. Abbreviations used in the figure: \acf{DSC}, \acf{ImLC}, \acf{InS}, \acf{ObD}, \acf{PPV}, \acf{SemS}. Reference used in the figure: \url{https://cada.grand-challenge.org/Assessment/}. We recommend Point inside Mask/Box/Approximation as a localization criterion in Subprocess S8 (Extended Data Fig.~\ref{fig:metric-selection:s8-localization-criterion}).}
    \label{fig:cheat-sheet-point-inside}
\end{figure}

\begin{figure}[H]
    \centering
    \includegraphics[width=\textwidth]{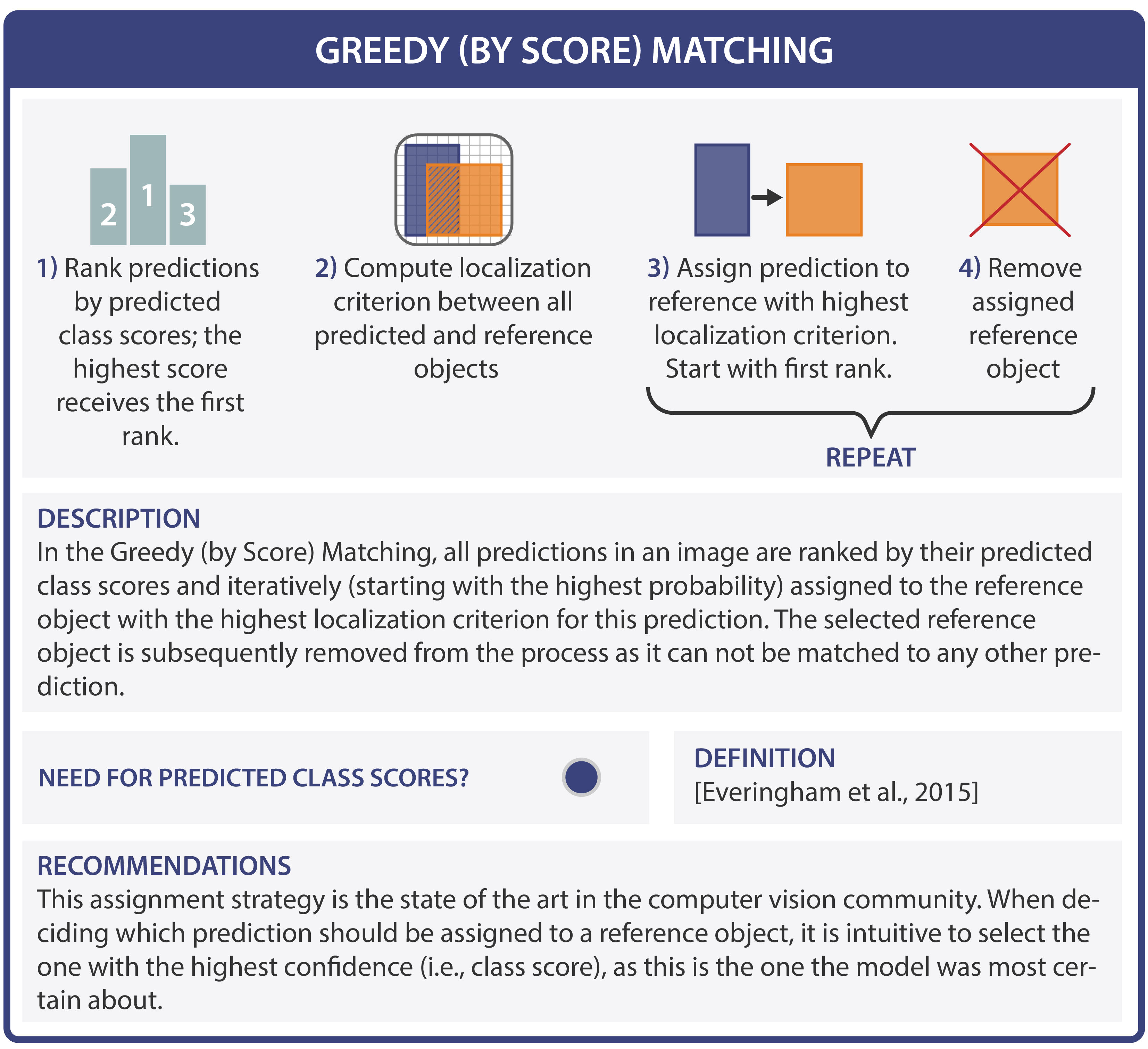}
    \caption{Cheat Sheet for the Greedy (by Score) Matching. Reference used in the figure: Everingham et al., 2015: \cite{everingham2015pascal}. We recommend Greedy (by Score) Matching as an assignment strategy in Subprocess S9 (Extended Data Fig.~\ref{fig:metric-selection:s9-assignment-strategy}).}
    \label{fig:cheat-sheet-greedy-score}
\end{figure}

\begin{figure}[H]
    \centering
    \includegraphics[width=\textwidth]{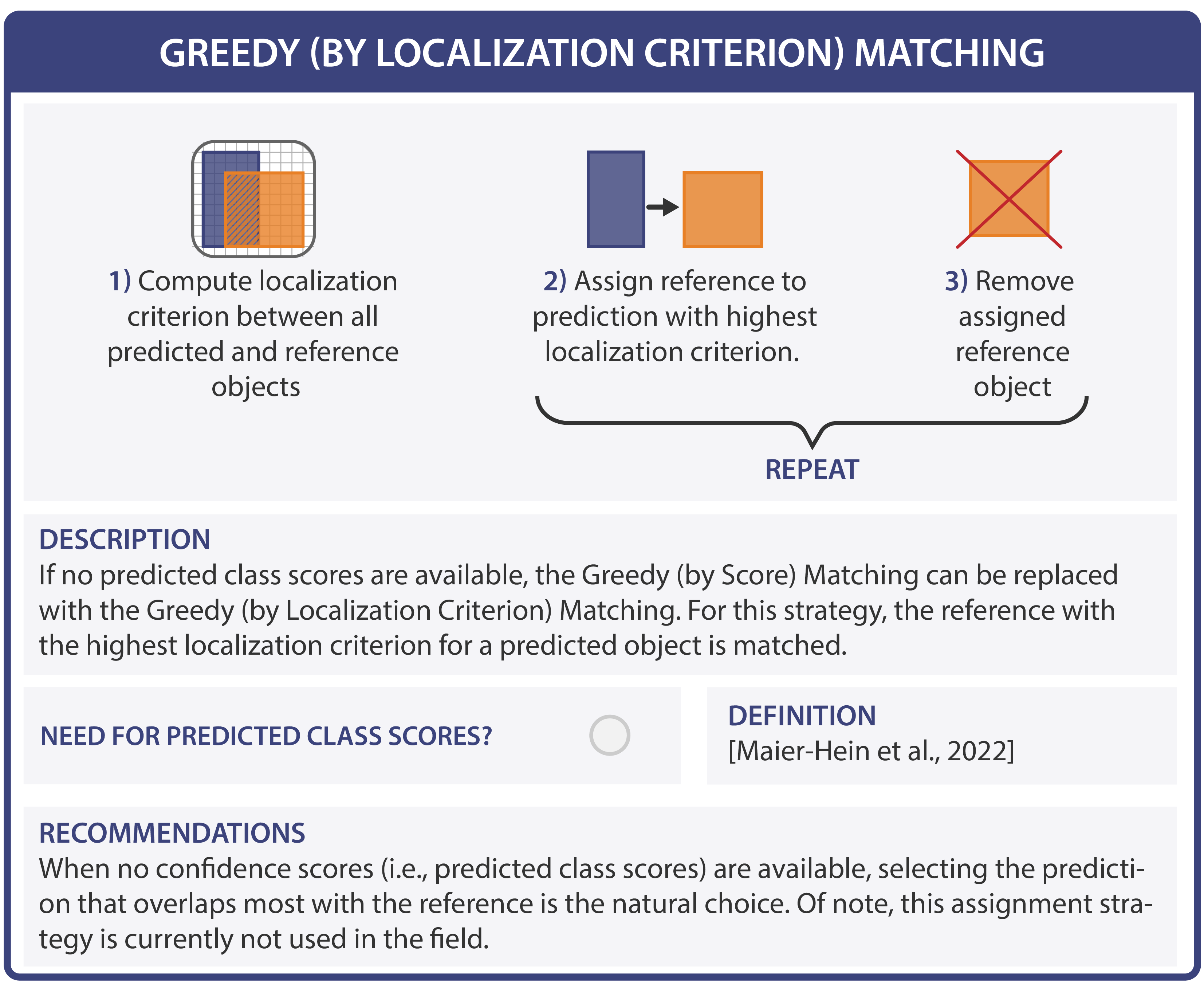}
    \caption{Cheat Sheet for the Greedy (by Localization Criterion) Matching. Reference used in the figure: Maier-Hein et al., 2022: \cite{maier2022metrics}. We recommend Greedy (by Localization Criterion) Matching as an assignment strategy in Subprocess S9 (Extended Data Fig.~\ref{fig:metric-selection:s9-assignment-strategy}).}
    \label{fig:cheat-sheet-greedy-localization}
\end{figure}

\begin{figure}[H]
    \centering
    \includegraphics[width=\textwidth]{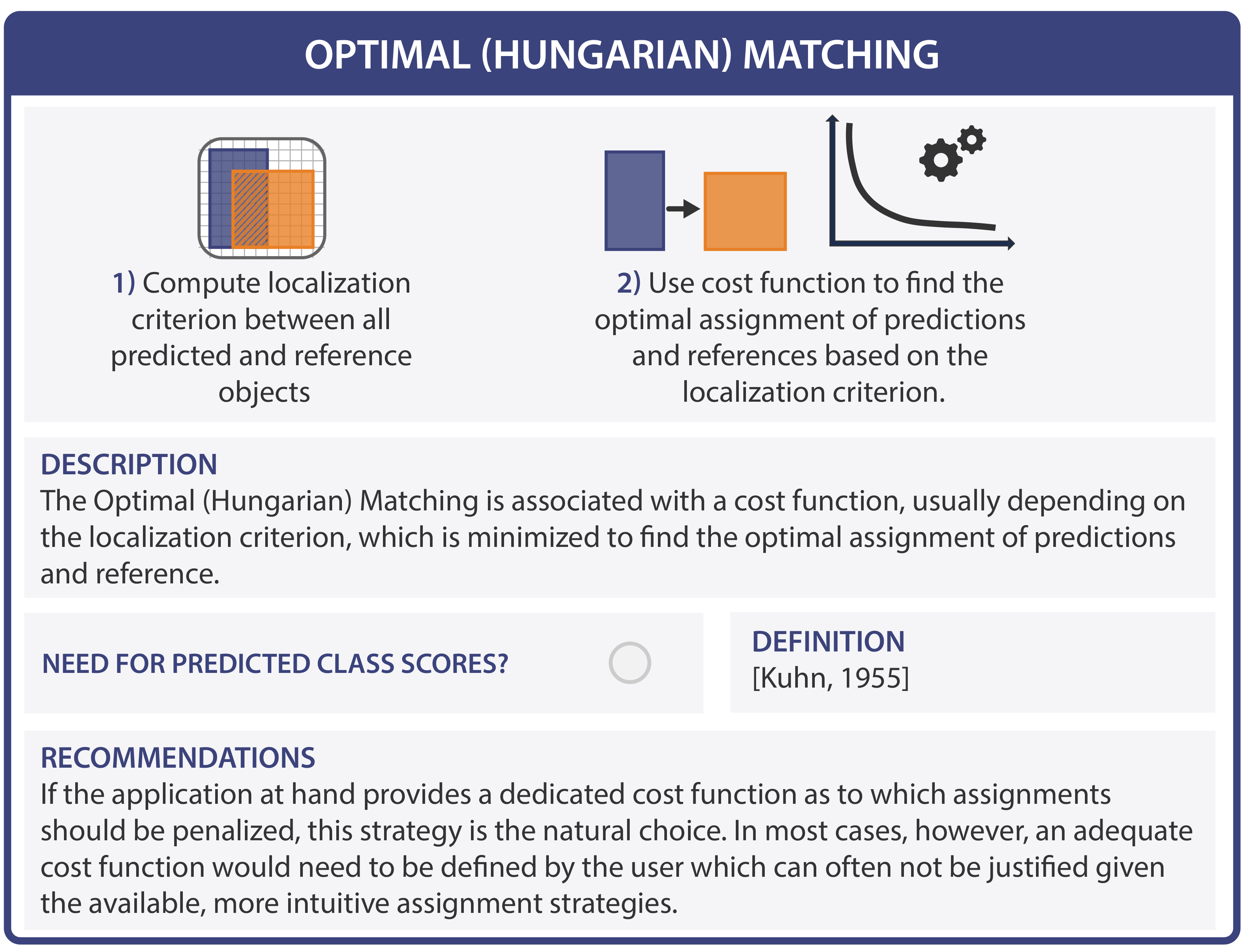}
    \caption{Cheat Sheet for the Optimal (Hungarian) Matching. Reference used in the figure: Kuhn et al., 1955: \cite{kuhn1955hungarian}. We recommend the Optimal (Hungarian) Matching as an assignment strategy in Subprocess S9 (Extended Data Fig.~\ref{fig:metric-selection:s9-assignment-strategy}).}
    \label{fig:cheat-sheet-hungarian}
\end{figure}

\begin{figure}[H]
    \centering
    \includegraphics[width=\textwidth]{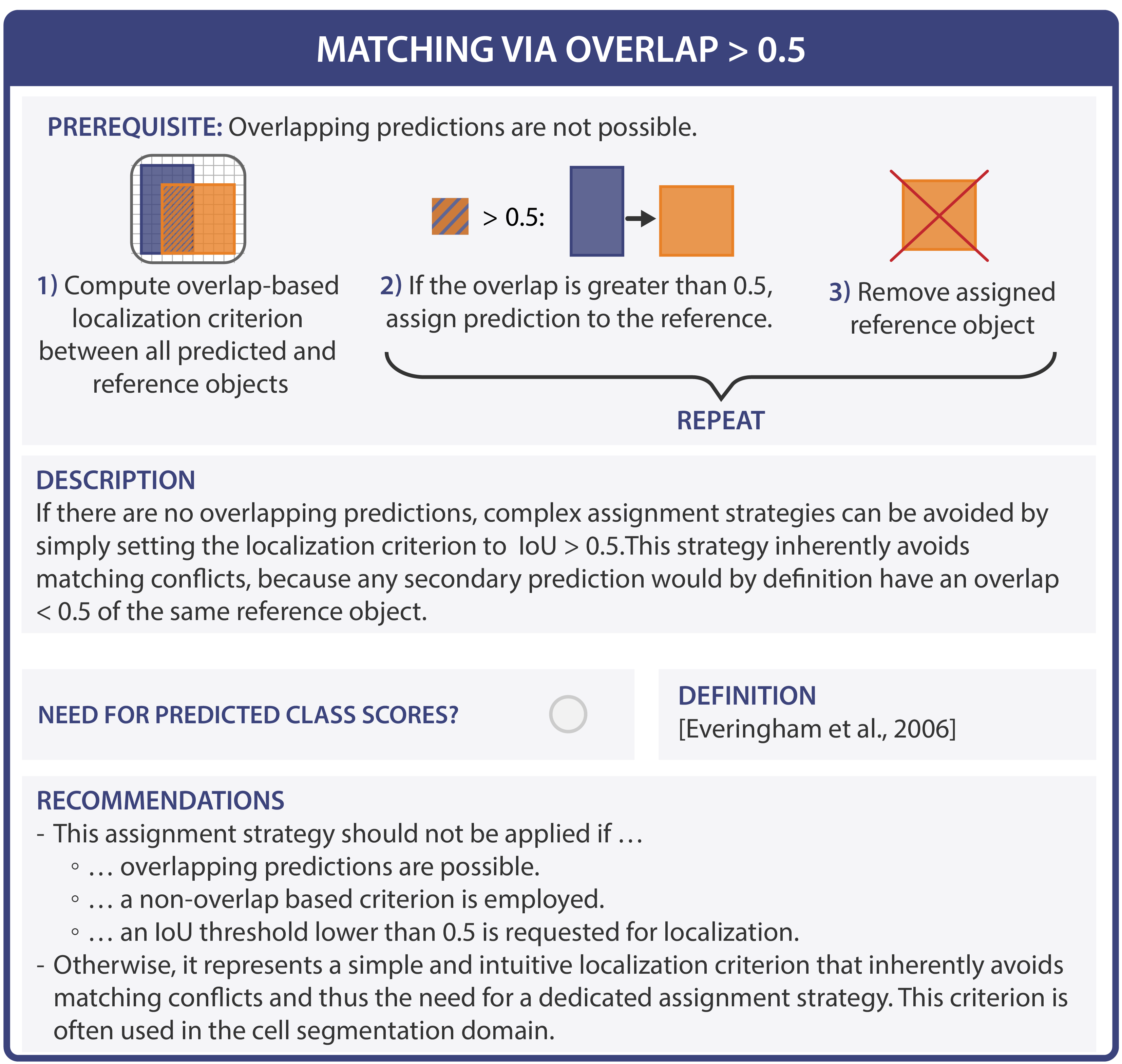}
    \caption{Cheat Sheet for the Matching via Overlap > 0.5. Reference used in the figure: Everingham et al., 2006: \cite{everingham20062005}. We recommend Matching via Overlap > 0.5 as an assignment strategy in Subprocess S9 (Extended Data Fig.~\ref{fig:metric-selection:s9-assignment-strategy}).}
    \label{fig:cheat-sheet-matching-greater}
\end{figure}

%------------------------------------
\acresetall
In this section, we present cheat sheets for the metrics deemed particularly relevant by the \textit{Metrics Reloaded} consortium. We provide a description along with the formula as well the respective value range. For every metric, we indicate further important characteristics, such as the recommended problem categories or potential prevalence dependency. Finally, we highlight our recommendations. Many of the presented metrics rely on the confusion matrix, which is illustrated in Fig.~\ref{fig:confusion-matrix}.

\newpage
\subsubsection{Discrimination metrics} 
\label{ssec:cheat-sheets-discrimination}
\hfill \\ \textbf{Counting metrics}
\FloatBarrier
\newpage

\newpage
\textbf{Multi-threshold metrics}

\newpage

\newpage
\textbf{Boundary-based metrics}

\newpage

\newpage
\subsubsection{Calibration metrics}
\label{ssec:cheat-sheets-calibration}
\hfill 
\FloatBarrier
\newpage

\newpage
\subsubsection{Localization criteria}
\label{ssec:cheat-sheets-localization}
\hfill 
\FloatBarrier

\newpage

\newpage
\subsubsection{Assignment strategies}
\label{ssec:cheat-sheets-assignment}
\hfill 
\FloatBarrier
\newpage

\setcounter{figure}{0}
\newpage

%------------------------------------
%\subsection{Instantiation of the Framework for Concrete Biomedical Problems}
%\label{app:scenarios}
%------------------------------------
1%------------------------------------
\section{Recommendations for selected use cases}
\label{app:scenarios}
%------------------------------------
\acresetall

We instantiated the framework for several biological and medical image analysis use cases. The list of use cases with a link to the figures representing the recommendations is provided below:

\subsection{Image-level classification} 
\begin{figure}[H]
    \centering
    \includegraphics[width=1\linewidth]{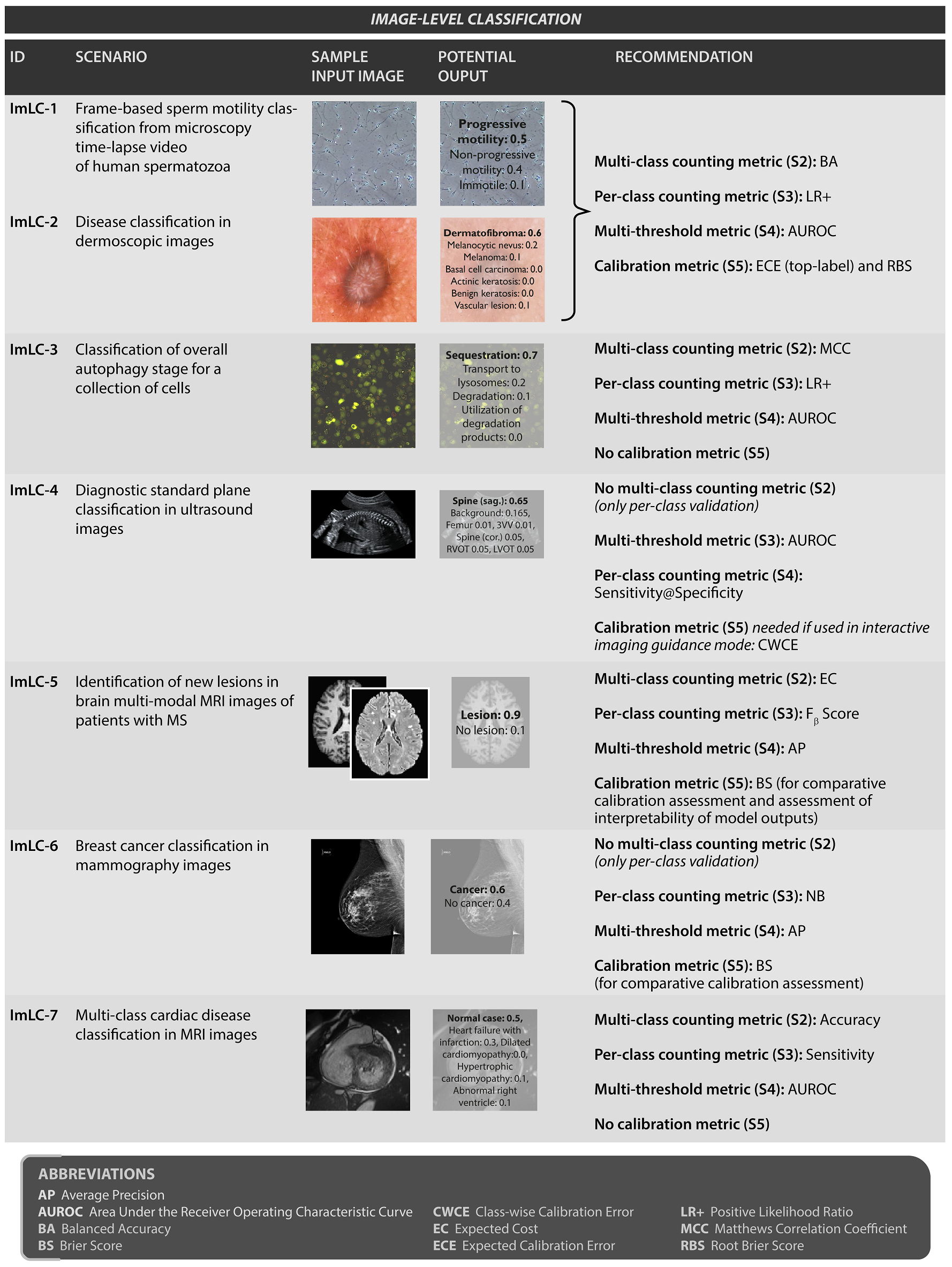}
    \caption{\textbf{Instantiation of the framework with recommendations for concrete biomedical image-level classification problems.} \textbf{(ImLC-1)} Frame-based sperm motility classification from microscopy time-lapse video of human spermatozoa \citep{haugen2019visem}. \textbf{(ImLC-2)} Disease classification in dermoscopic images \citep{codella2019skin, dermoscopedia}. \textbf{(ImLC-3)} Classification of the overall autophagy stage for a collection of cells \citep{zhang2020deepphagy, nagao2020robust}. \textbf{(ImLC-4)} Diagnostic standard plane classification in ultrasound images \citep{baumgartner2017sononet}. \textbf{(ImLC-5)} Identification of new lesions in brain multi-modal \acf{MRI} images of \ac{MS} patients \citep{kofler2022blob, commowick2018objective}. \textbf{(ImLC-6)} Breast cancer classification in mammography images \citep{le2019artificial}. \textbf{(ImLC-7)} Multi-class cardiac disease classification in \ac{MRI} images \citep{bernard2018deep}.}
    \label{fig:scenario-ImLC-overview}
\end{figure}

The following use cases have been instantiated for image-level classification problems. The resulting metric recommendations can be found in Fig.~\ref{fig:scenario-ImLC-overview}, while Figs.~\ref{fig:scenario-s2}-\ref{fig:scenario-s4} provide a detailed overview of the recommendations for the use cases in the metric selection Subprocesses S2-S5.
\begin{description}
    \item[ImLC-1] Frame-based sperm motility classification from microscopy time-lapse video of human spermatozoa \citep{haugen2019visem}
    \item[ImLC-2] Disease classification in dermoscopic images \citep{codella2019skin, dermoscopedia}
    \item[ImLC-3] Classification of the overall autophagy stage for a collection of cells \citep{zhang2020deepphagy, nagao2020robust}
    \item[ImLC-4] Diagnostic standard plane classification in ultrasound images \citep{baumgartner2017sononet}
    \item[ImLC-5] Identification of new lesions in brain multi-modal \ac{MRI} images of patients with \ac{MS} \citep{kofler2022blob, commowick2018objective}
    \item[ImLC-6] Breast cancer classification in mammography images \citep{le2019artificial}
    \item[ImLC-7] Multi-class cardiac disease classification in \ac{MRI} images \citep{bernard2018deep}
\end{description}

\newpage

% \newpage
% \begin{figure}[H]
%     \centering
%     \includegraphics[width=1\linewidth]{images/Scenarios/Scenario_ILC_sperm_skin.pdf}
%     \caption{\textbf{Instantiation of the framework for the problems of frame-based sperm motility classification based on microscopy time-lapse video containing human spermatozoa \citep{haugen2019visem} and disease classification in dermoscopic images \citep{codella2019skin, dermoscopedia}}. The upper part shows the recommendations. The middle part shows the traversal through the main path of Fig.~\ref{fig:meta-mapping} based on the generated fingerprint for the specific problem. The bottom part shows the traversal through the Subprocesses S2 (Fig.~\ref{fig:metric-selection:s2-multi-class}), S3 (Fig.~\ref{fig:metric-selection:s3-multi-threshold}) and S4 (Fig.~\ref{fig:metric-selection:s4-single-threshold}) and the resulting recommended metrics for both use cases.}
%     \label{fig:scenario-ImLC-skin-sperm-motility}
% \end{figure}

\newpage
\subsection{Semantic segmentation} 
\begin{figure}[H]
    \centering
    \includegraphics[width=1\linewidth]{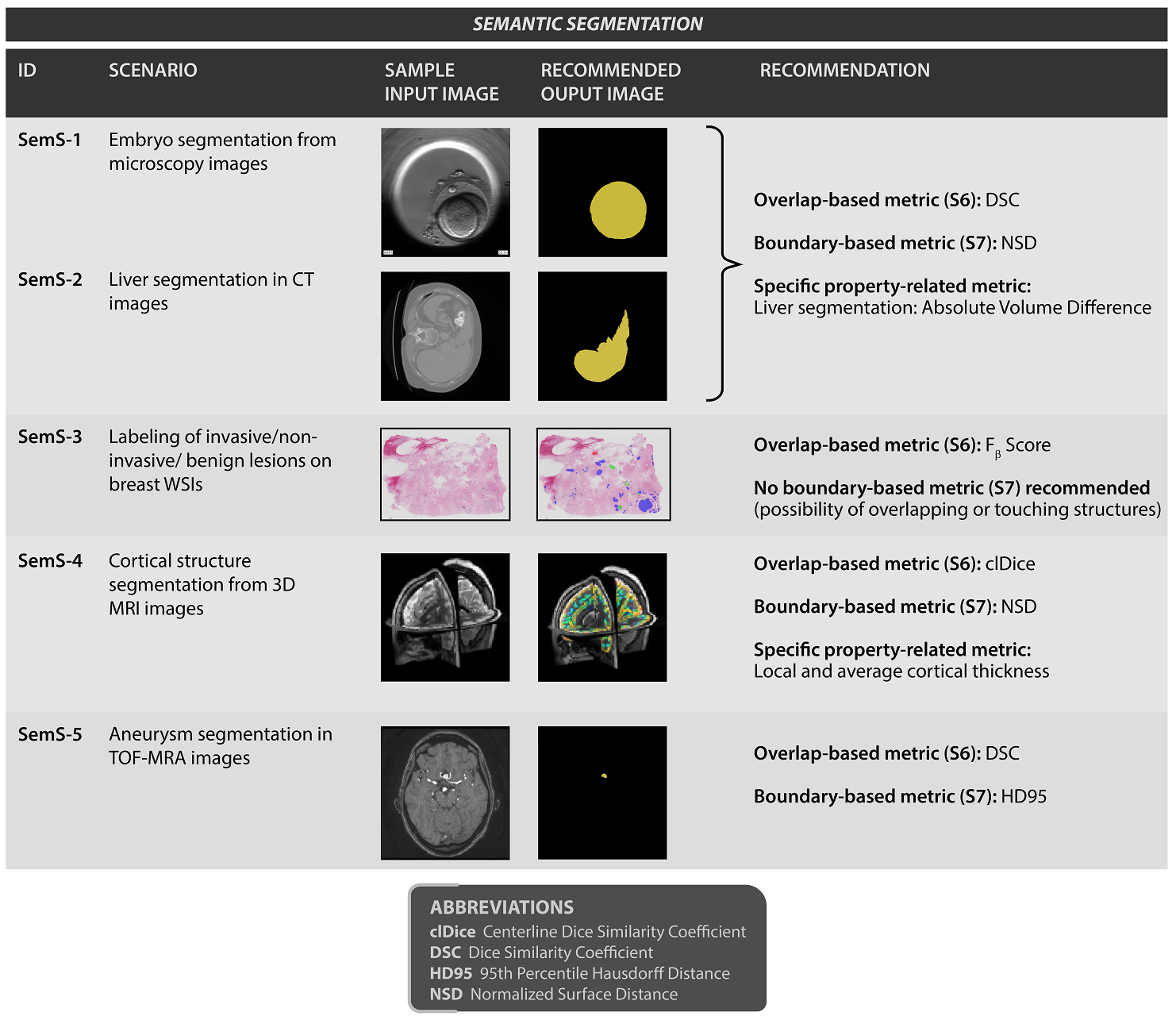}
    \caption{\textbf{Instantiation of the framework with recommendations for concrete biomedical semantic segmentation problems.} \textbf{(SemS-1)} Embryo segmentation from microscopy images \citep{targosz2021semantic}. \textbf{(SemS-2)} Liver segmentation in \acf{CT} images \citep{antonelli2021medical, simpson2019large}. \textbf{(SemS-3)} Labeling of invasive/non-invasive/benign lesions in breast \acf{WSI} \citep{aresta2019bach}. \textbf{(SemS-4)} Cortical structure segmentation from 3D \acf{MRI} images\citep{budd2020surface}. \textbf{(SemS-5)} Aneurysm segmentation in \acf{TOF-MRA} images \citep{timmins2021comparing}.}
    \label{fig:scenario-SemS-overview}
\end{figure}

The following use cases have been instantiated for semantic segmentation problems. The resulting metric recommendations can be found in Fig.~\ref{fig:scenario-SemS-overview}, while Figs.~\ref{fig:scenario-s7}-\ref{fig:scenario-s8} provide a detailed overview of the recommendations for the use cases in the metric selection Subprocesses S6 and S7.
\begin{description}
    \item[SemS-1] Embryo segmentation from microscopy images \citep{targosz2021semantic}
    \item[SemS-2] Liver segmentation in \ac{CT} images \citep{antonelli2021medical, simpson2019large}
    \item[SemS-3] Labeling of invasive/non-invasive/benign lesions in breast \ac{WSI} \citep{aresta2019bach}
    \item[SemS-4] Cortical structure segmentation from 3D \ac{MRI} images \citep{budd2020surface}
    \item[SemS-5] Aneurysm segmentation in \ac{TOF-MRA} images \citep{timmins2021comparing}
\end{description}

\newpage

% \begin{figure}[H]
%     \centering
%     \includegraphics[width=1\linewidth]{images/Scenarios/Scenario_SS_liver_cell.pdf}
%     \caption{\textbf{Instantiation of the framework for the problems of embryo segmentation from microscopy images \citep{targosz2021semantic} and liver segmentation in computed tomography images \citep{antonelli2021medical, simpson2019large}}. The upper part shows the recommendations. The middle part shows the traversal through the main path of Fig.~\ref{fig:meta-mapping} based on the generated fingerprint for the specific problem. The bottom part shows the traversal through the Subprocesses S7 (Fig.~\ref{fig:metric-selection:s7-overlap}) and S8 (Fig.~\ref{fig:metric-selection:s8-boundary}) and the resulting recommended metrics for both use cases.}
%     \label{fig:scenario-SemS-liver-lung-cell}
% \end{figure}

\newpage
\subsection{Object detection} 
\begin{figure}[H]
    \centering
    \includegraphics[width=1\linewidth]{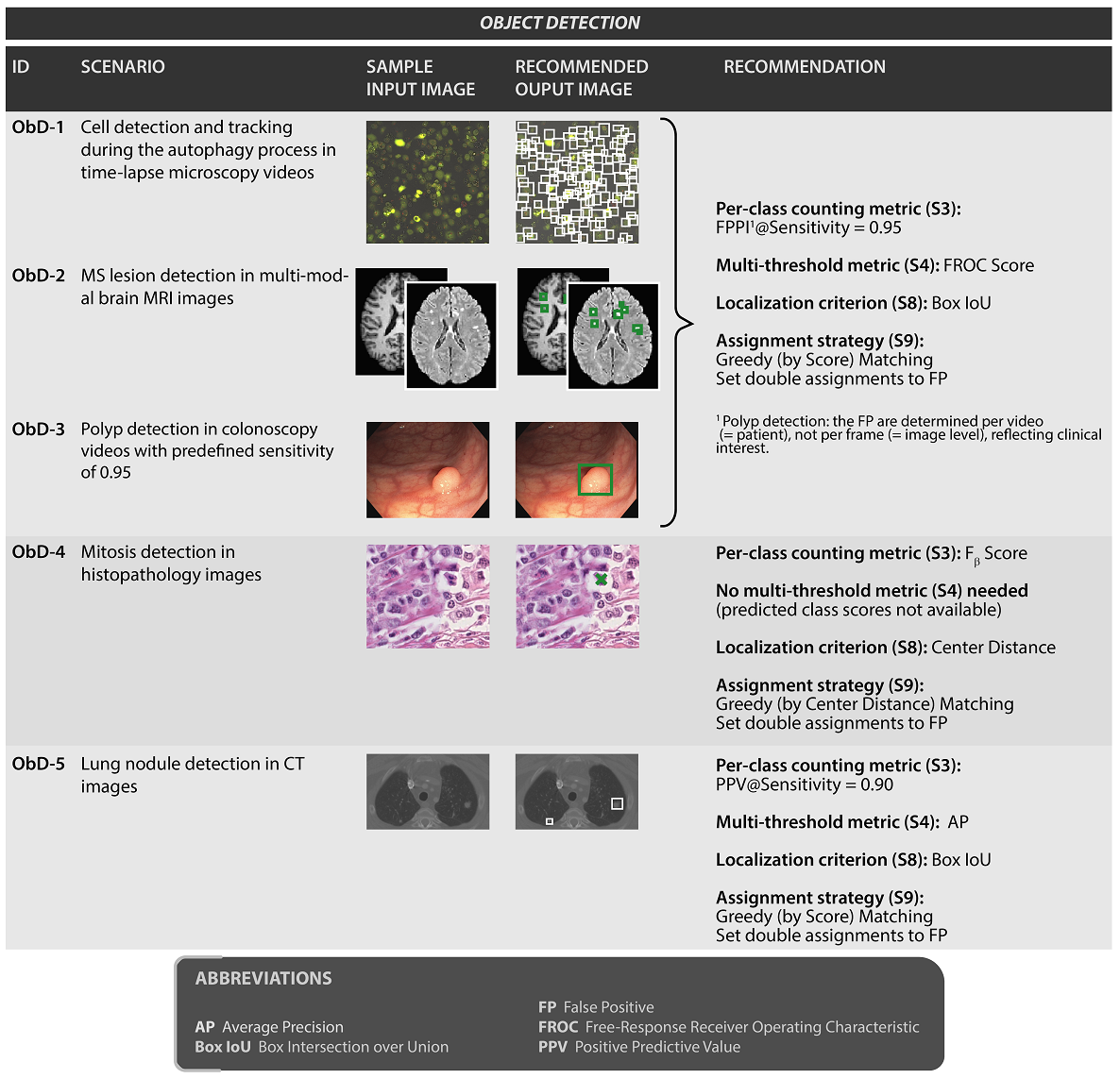}
    \caption{\textbf{Instantiation of the framework with recommendations for concrete biomedical object detection problems.} \textbf{(ObD-1)} Cell detection and tracking during the autophagy process in time-lapse microscopy \citep{zhang2020deepphagy, nagao2020robust}. \textbf{(ObD-2)} Multiple sclerosis (\ac{MS}) lesion detection in multi-modal brain \acf{MRI} images \citep{kofler2022blob, commowick2018objective}.  \textbf{(ObD-3)} Polyp detection in  colonoscopy videos with predefined sensitivity of 0.95 \citep{sanchez2019computer, bernal2019gtcreator}. \textbf{(ObD-4)} Mitosis detection in histopathology images \citep{aubreville2022mitosis}. \textbf{(ObD-5)} Lung nodule detection in \acf{CT} images \citep{armato2011lung, armato2011lungdata, clark2013cancer}.}
    \label{fig:scenario-ObD-overview}
\end{figure}

The following use cases have been instantiated for object detection problems. The resulting metric recommendations can be found in Fig.~\ref{fig:scenario-ObD-overview}, while Figs.~\ref{fig:scenario-s3}-\ref{fig:scenario-s6} provide a detailed overview of the recommendations for the use cases in the metric selection Subprocesses S3 - S4, S8 - S9.
\begin{description}
    \item[ObD-1] Cell detection and tracking during the autophagy process in time-lapse microscopy \citep{zhang2020deepphagy, nagao2020robust}
    \item[ObD-2] \ac{MS} lesion detection in multi-modal brain \ac{MRI} images \citep{kofler2022blob, commowick2018objective}
    \item[ObD-3] Polyp detection in  colonoscopy videos with predefined sensitivity of 0.95 \citep{sanchez2019computer, bernal2019gtcreator}
    \item[ObD-4] Mitosis detection in histopathology images \citep{aubreville2022mitosis}
    \item[ObD-5] Lung nodule detection in \ac{CT} images \citep{armato2011lung, armato2011lungdata, clark2013cancer}
\end{description}

\newpage

% \newpage
% \begin{figure}[H]
%     \centering
%     \includegraphics[width=1\linewidth]{images/Scenarios/Scenario_OD_cell_ms.pdf}    \caption{\textbf{Instantiation of the framework for the problems of cell detection and tracking during the autophagy process in time-lapse microscopy \citep{zhang2020deepphagy, nagao2020robust} and \ac{MS} lesion detection in multi-modal brain \ac{MRI} images \citep{kofler2022blob, commowick2018objective}}. The upper part shows the recommendations. The middle part shows the traversal through the main path of Fig.~\ref{fig:meta-mapping} based on the generated fingerprint for the specific problem. The bottom part shows the traversal through the Subprocesses S5 (Fig.~\ref{fig:metric-selection:s5-localization-criterion}), S6 (Fig.~\ref{fig:metric-selection:s6-assignment-strategy}), S3 (Fig.~\ref{fig:metric-selection:s3-multi-threshold}) and S4 (Fig.~\ref{fig:metric-selection:s4-single-threshold}) as well as the resulting recommended localization/assignment methods and metrics for both use cases.}
%     \label{fig:scenario-ObD-cell-ms-nodule}
% \end{figure}

\newpage
\subsection{Instance segmentation} 
\begin{figure}[H]
    \centering
    \includegraphics[width=1\linewidth]{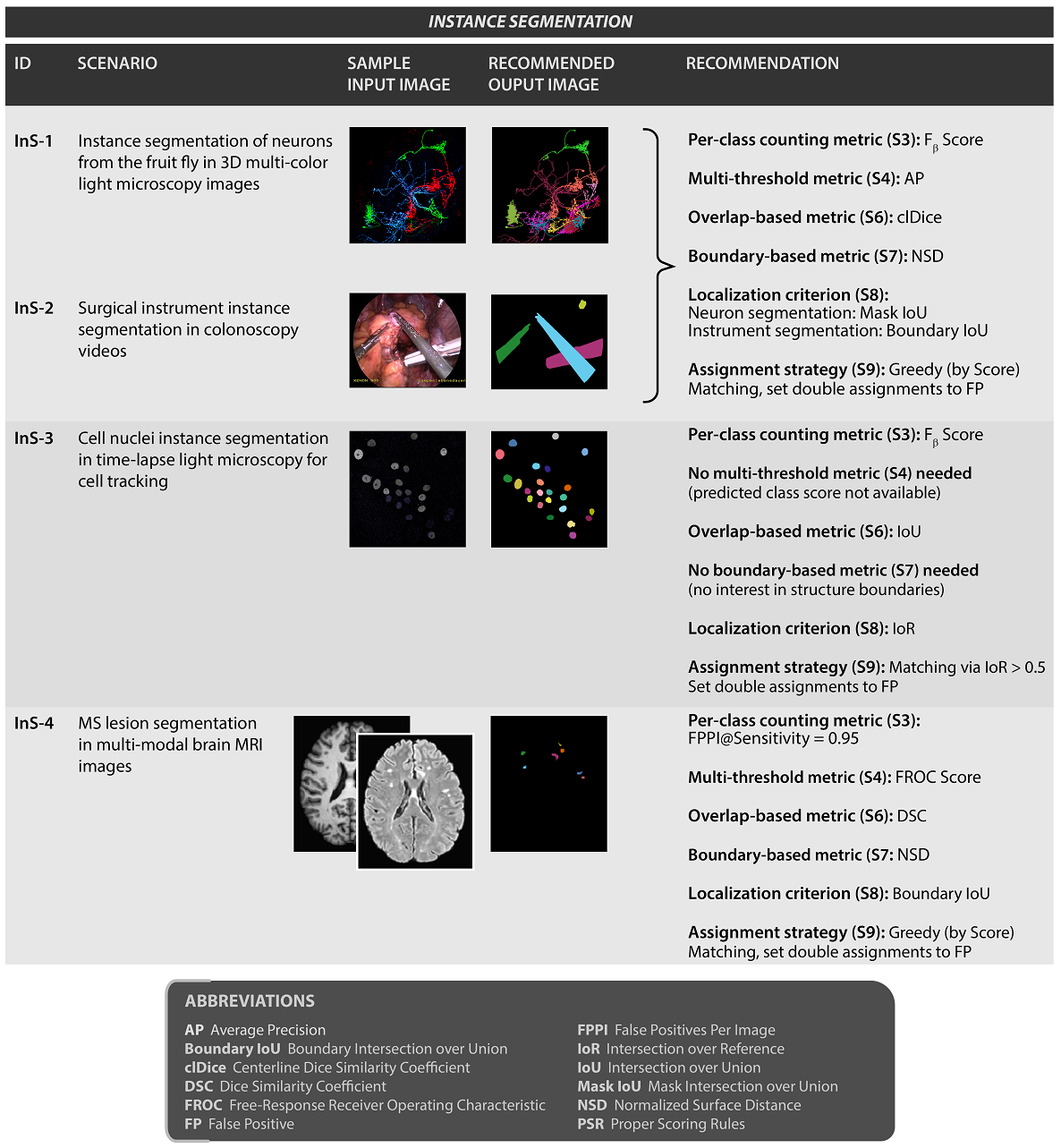}
    \caption{\textbf{Instantiation of the framework with recommendations for concrete biomedical instance segmentation problems.} \textbf{(InS-1)} Instance segmentation of neurons from the fruit fly in 3D multi-color light microscopy images \citep{tirian2017vt, mais2020patchperpix, meissner2022searchable}. \textbf{(InS-2)} Surgical instrument instance segmentation in colonoscopy videos \citep{maier2021heidelberg}. \textbf{(InS-3)} Cell nuclei instance segmentation in time-lapse light microscopy for cell tracking \citep{ulman2017objective}. \textbf{(InS-4)} Multiple sclerosis (\ac{MS}) lesion segmentation in multi-modal brain \acf{MRI} images \citep{kofler2022blob, commowick2018objective}.}
    \label{fig:scenario-InS-overview}
\end{figure}

\begin{figure}[H]
    \centering
    \includegraphics[width=1\linewidth]{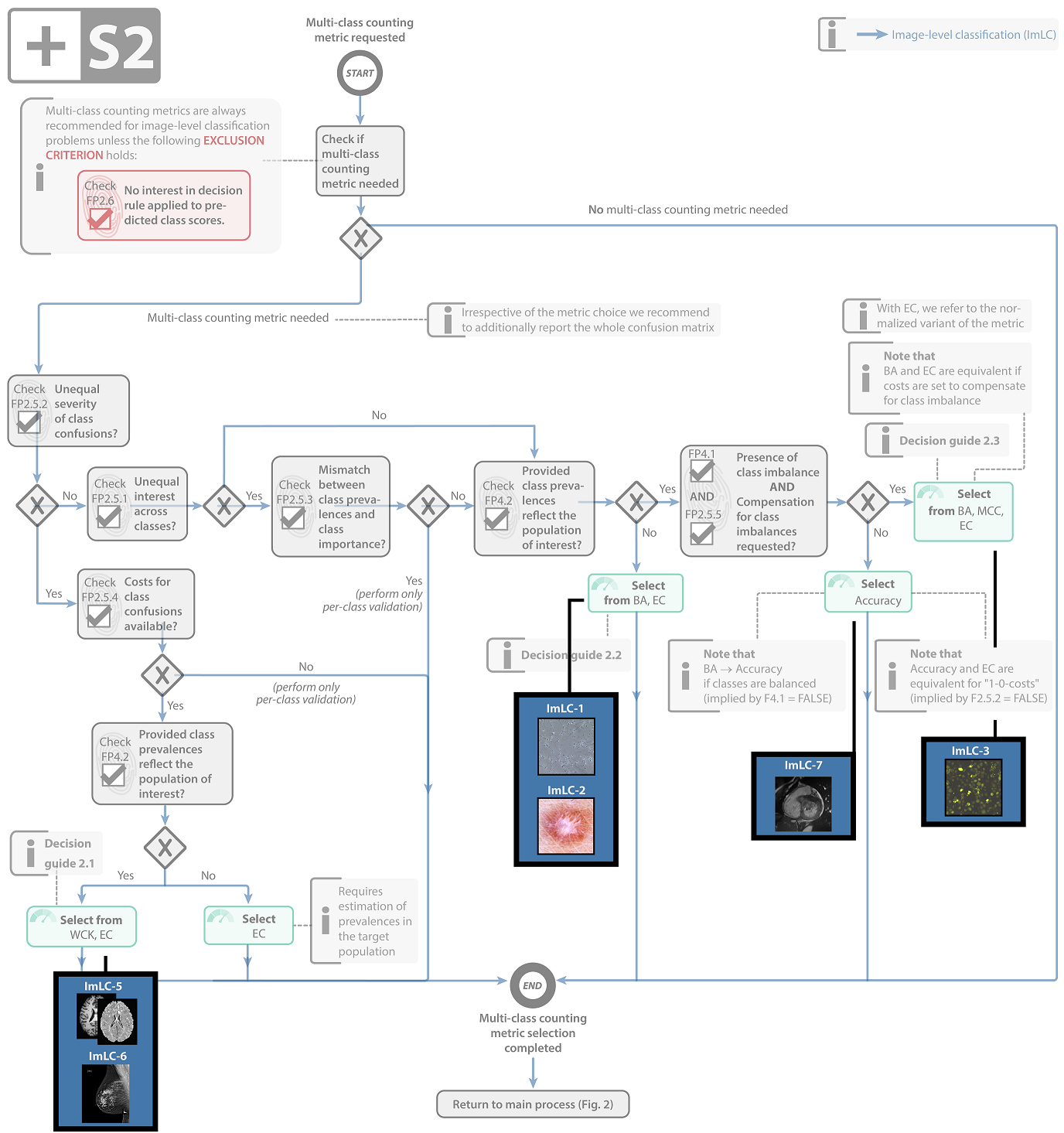}
    \caption{\textbf{Instantiation of Subprocess S2 for the selection of multi-class counting metrics with recommendations for concrete biomedical problems.} Included use cases: \textbf{(ImLC-1)} Frame-based sperm motility classification based on microscopy time-lapse video containing human spermatozoa \citep{haugen2019visem}. \textbf{(ImLC-2)} Disease classification in dermoscopic images \citep{codella2019skin, dermoscopedia}. \textbf{(ImLC-3)} Classification of the overall autophagy stage for a collection of cells \citep{zhang2020deepphagy, nagao2020robust}. \textbf{(ImLC-5)} Identification of new lesions in brain multi-modal \acf{MRI} images of \acf{MS} patients \citep{kofler2022blob, commowick2018objective}. \textbf{(ImLC-6)} Breast cancer classification in mammography images \citep{le2019artificial}. \textbf{(ImLC-7)} Multi-class cardiac disease classification in \ac{MRI} images \citep{bernard2018deep}.}
    \label{fig:scenario-s2}
\end{figure}

\begin{figure}[H]
    \centering
    \includegraphics[width=1\linewidth]{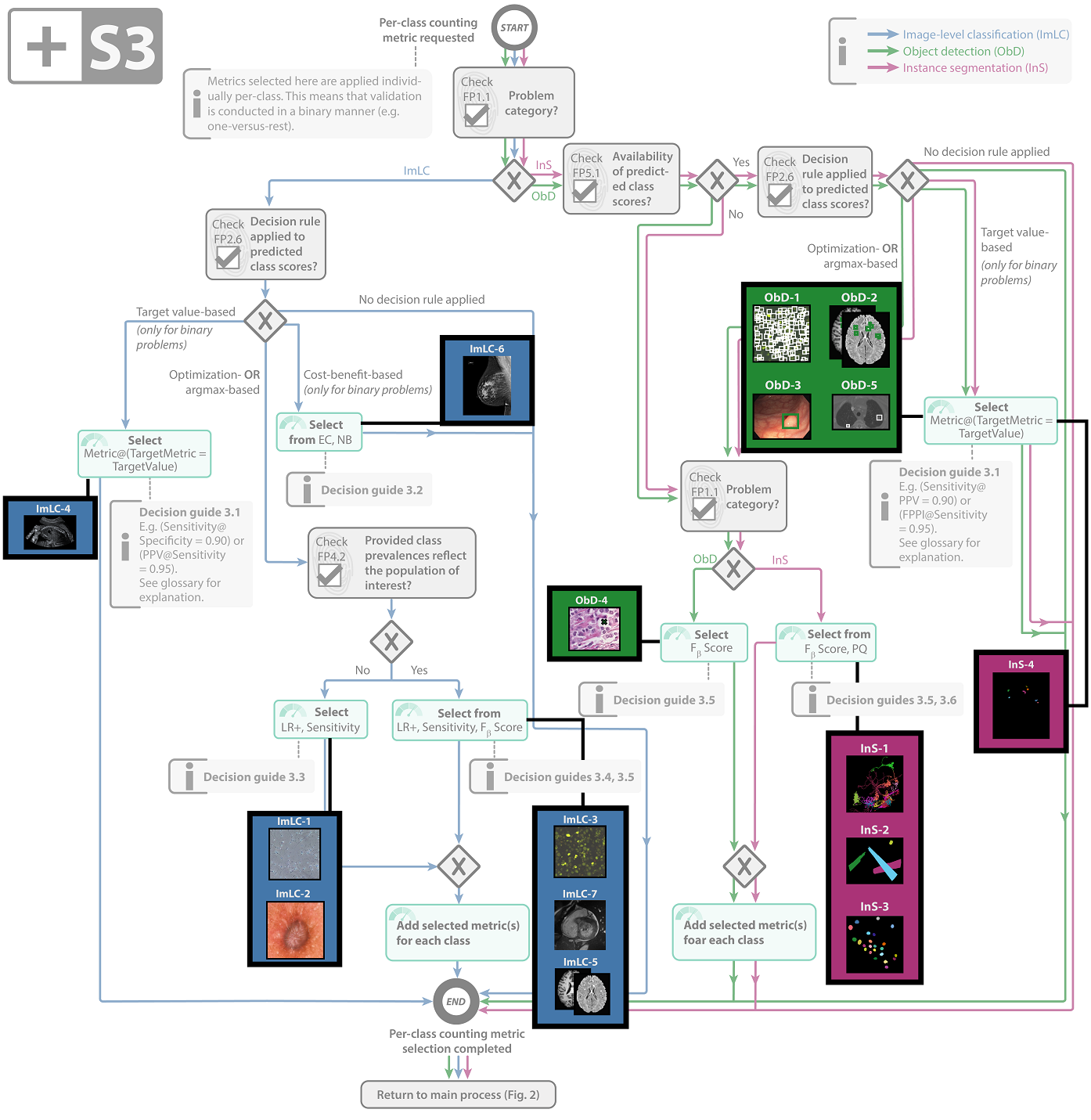}
    \caption{\textbf{Instantiation of Subprocess S3 for the selection of per-class counting metrics with recommendations for concrete biomedical problems.} Included use cases: \textbf{(ImLC-1)} Frame-based sperm motility classification from microscopy time-lapse video of human spermatozoa \citep{haugen2019visem}. \textbf{(ImLC-2)} Disease classification in dermoscopic images \citep{codella2019skin, dermoscopedia}. \textbf{(ImLC-3)} Classification of the overall autophagy stage for a collection of cells \citep{zhang2020deepphagy, nagao2020robust}. \textbf{(ImLC-4)} Diagnostic standard plane classification in ultrasound images \citep{baumgartner2017sononet}. \textbf{(ImLC-5)} Identification of new lesions in brain multi-modal \acf{MRI} images of \acf{MS} patients \citep{kofler2022blob, commowick2018objective}. \textbf{(ImLC-6)} Breast cancer classification in mammography images \citep{le2019artificial}. \textbf{(ImLC-7)} Multi-class cardiac disease classification in \acs{MRI} images \citep{bernard2018deep}. \textbf{(ObD-1)} Cell detection and tracking during the autophagy process in time-lapse microscopy \citep{zhang2020deepphagy, nagao2020robust}. \textbf{(ObD-2)} \acs{MS} lesion detection in multi-modal brain \acs{MRI} images \citep{kofler2022blob, commowick2018objective}.  \textbf{(ObD-3)} Polyp detection in  colonoscopy videos with predefined sensitivity of 0.95 \citep{sanchez2019computer, bernal2019gtcreator}. \textbf{(ObD-4)} Mitosis detection in histopathology images \citep{aubreville2022mitosis}. \textbf{(ObD-5)} Lung nodule detection in \acf{CT} images \citep{armato2011lung, armato2011lungdata, clark2013cancer}. \textbf{(InS-1)} Instance segmentation of neurons from the fruit fly in 3D multi-color light microscopy images \citep{tirian2017vt, mais2020patchperpix, meissner2022searchable}. \textbf{(InS-2)} Surgical instrument instance segmentation in colonoscopy videos \citep{maier2021heidelberg}. \textbf{(InS-3)} Cell nuclei instance segmentation in time-lapse light microscopy with a subsequent goal of cell tracking \citep{ulman2017objective}. \textbf{(InS-4)} \acs{MS} Lesion segmentation in multi-modal brain \acs{MRI} images \citep{kofler2022blob, commowick2018objective}.}
    \label{fig:scenario-s3}
\end{figure}

\begin{figure}[H]
    \centering
    \includegraphics[width=1\linewidth]{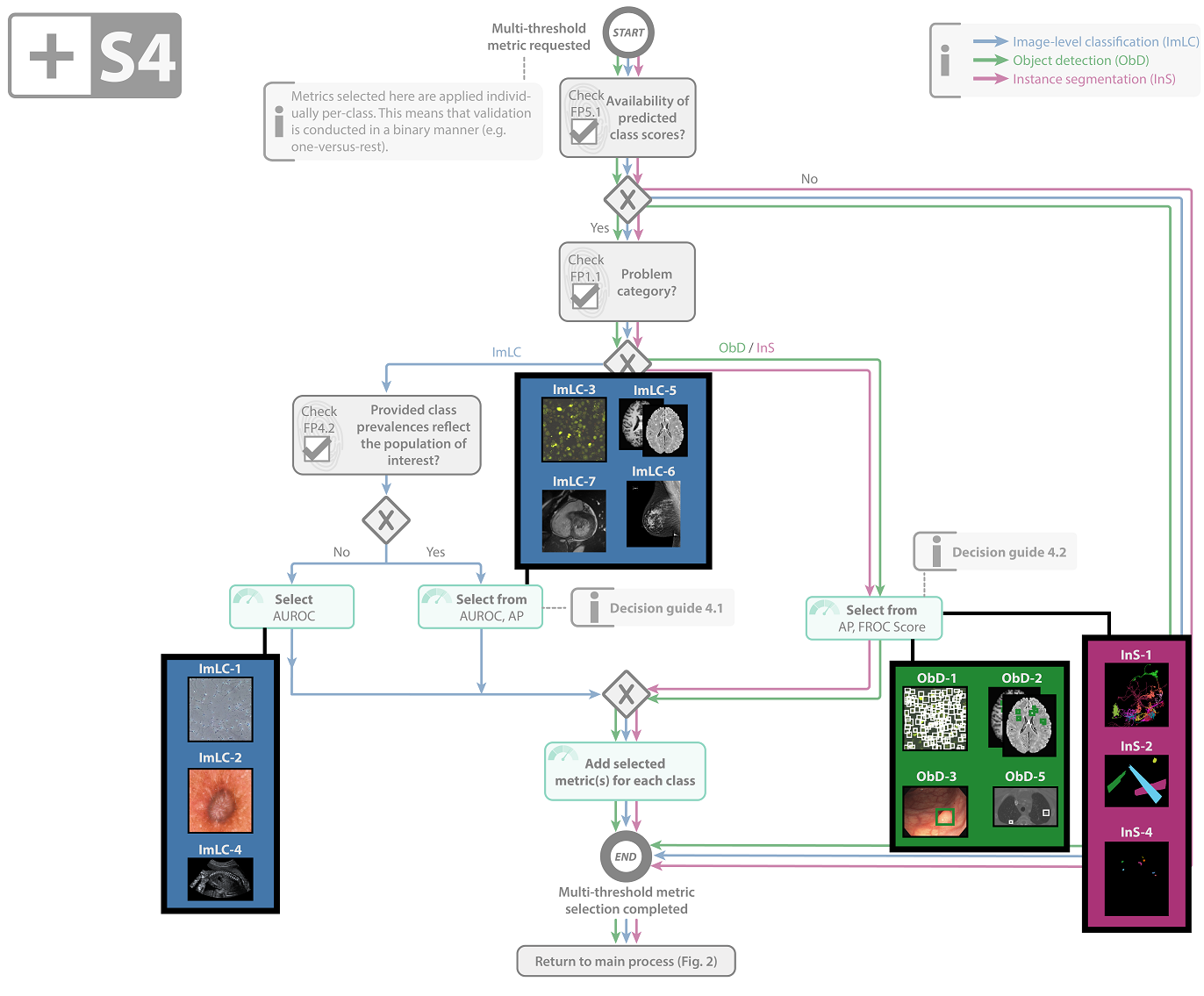}
    \caption{\textbf{Instantiation of Subprocess S4 for the selection of multi-threshold metrics with recommendations for concrete biomedical problems.} Included use cases: \textbf{(ImLC-1)} Frame-based sperm motility classification from microscopy time-lapse video of human spermatozoa \citep{haugen2019visem}. \textbf{(ImLC-2)} Disease classification in dermoscopic images \citep{codella2019skin, dermoscopedia}. \textbf{(ImLC-3)} Classification of the overall autophagy stage for a collection of cells \citep{zhang2020deepphagy, nagao2020robust}. \textbf{(ImLC-4)} Diagnostic standard plane classification in ultrasound images \citep{baumgartner2017sononet}. \textbf{(ImLC-5)} Identification of new lesions in brain multi-modal \acf{MRI} images of \acf{MS} patients  \citep{kofler2022blob, commowick2018objective}. \textbf{(ImLC-6)} Breast cancer classification in mammography images \citep{le2019artificial}. \textbf{(ImLC-7)} Multi-class cardiac disease classification in \acs{MRI} images \citep{bernard2018deep}. \textbf{(ObD-1)} Cell detection and tracking during the autophagy process in time-lapse microscopy \citep{zhang2020deepphagy, nagao2020robust}. \textbf{(ObD-2)} \acs{MS} lesion detection in multi-modal brain \acf{MRI} images \citep{kofler2022blob, commowick2018objective}.  \textbf{(ObD-3)} Polyp detection in  colonoscopy videos with predefined sensitivity of 0.95 \citep{sanchez2019computer, bernal2019gtcreator}. \textbf{(ObD-5)} Lung nodule detection in \acf{CT} images \citep{armato2011lung, armato2011lungdata, clark2013cancer}. \textbf{(InS-1)} Instance segmentation of neurons from the fruit fly in 3D multi-color light microscopy images \citep{tirian2017vt, mais2020patchperpix, meissner2022searchable}. \textbf{(InS-2)} Surgical instrument instance segmentation in colonoscopy videos \citep{maier2021heidelberg}. \textbf{(InS-4)} \acs{MS} lesion segmentation in multi-modal brain \acs{MRI} images \citep{kofler2022blob, commowick2018objective}.}
    \label{fig:scenario-s4}
\end{figure}

\begin{figure}[H]
    \centering
    \includegraphics[width=1\linewidth]{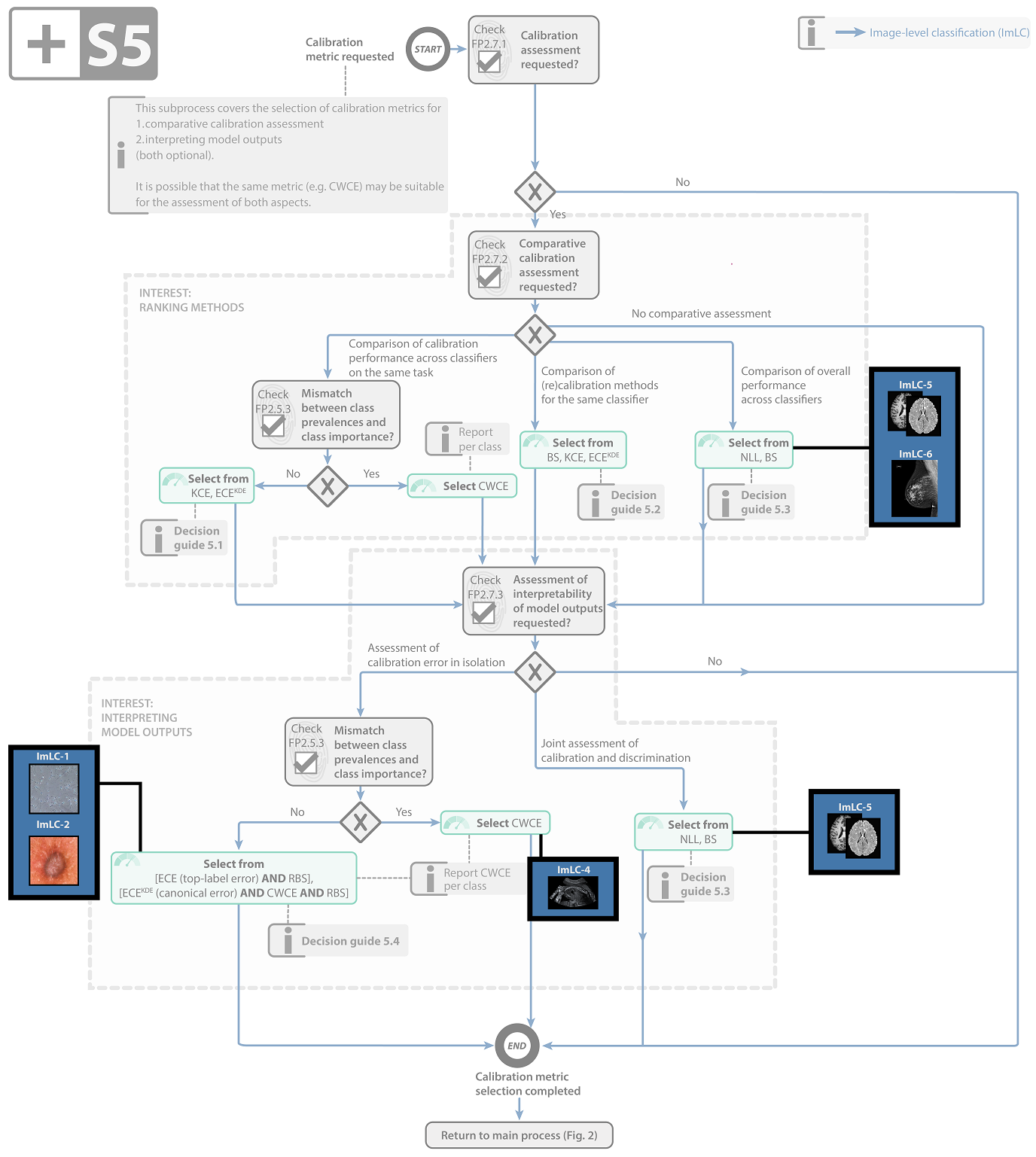}
    \caption{\textbf{Instantiation of Subprocess S5 for the selection of calibration metrics with recommendations for concrete biomedical problems.} Included use cases: \textbf{(ImLC-1)} Frame-based sperm motility classification from microscopy time-lapse video of human spermatozoa \citep{haugen2019visem}. \textbf{(ImLC-2)} Disease classification in dermoscopic images \citep{codella2019skin, dermoscopedia}. \textbf{(ImLC-4)} Diagnostic standard plane classification in ultrasound images \citep{baumgartner2017sononet}. \textbf{(ImLC-5)} Identification of new lesions in brain multi-modal \acf{MRI} images of \acf{MS} patients \citep{kofler2022blob, commowick2018objective}. \textbf{(ImLC-6)} Breast cancer classification in mammography images \citep{le2019artificial}.}
    \label{fig:scenario-s5}
\end{figure}

\begin{figure}[H]
    \centering
    \includegraphics[width=1\linewidth]{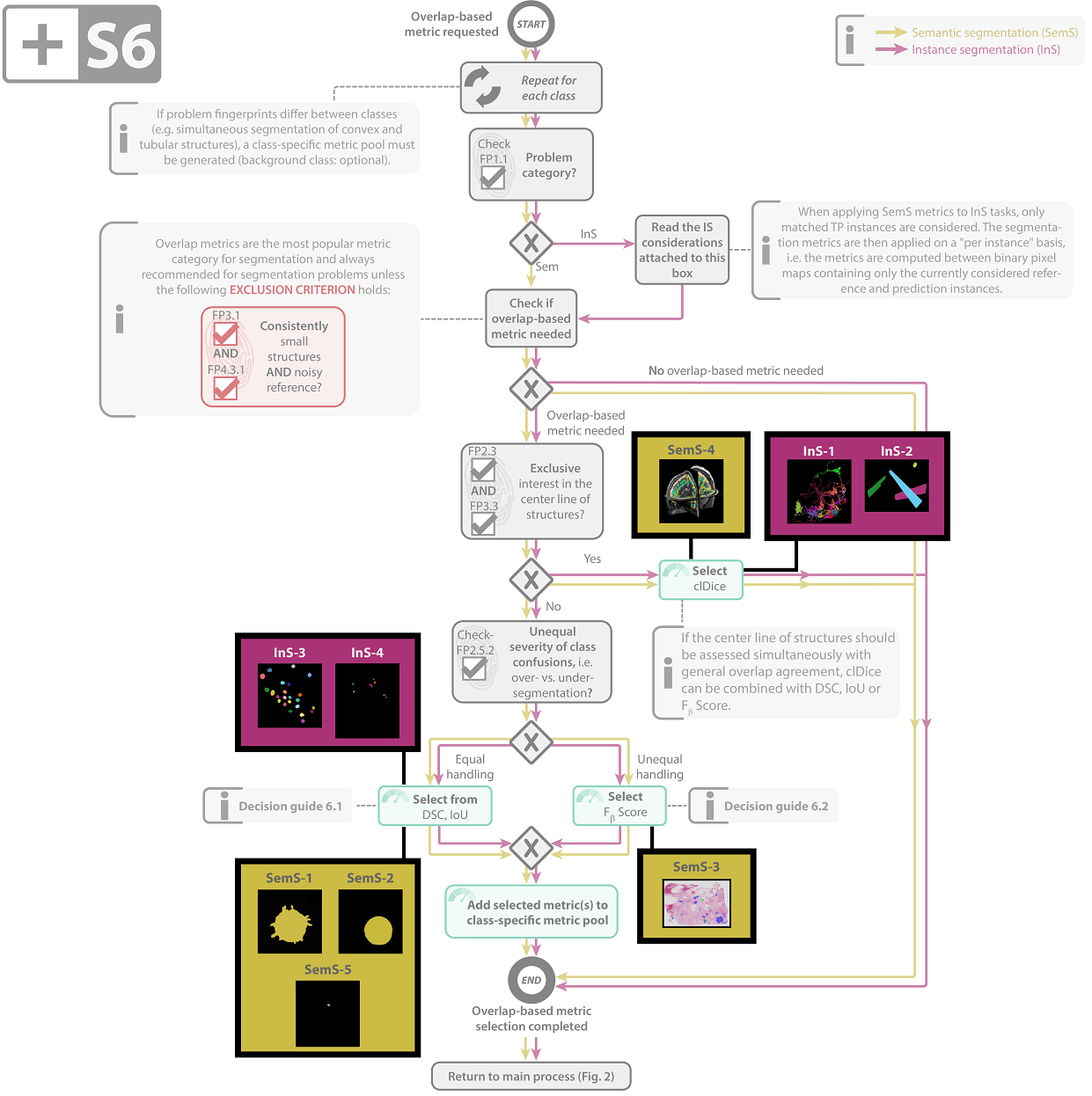}
    \caption{\textbf{Instantiation of Subprocess S6 for the selection of overlap-based metrics with recommendations for concrete biomedical problems.} \textbf{(SemS-1)} Embryo cell segmentation from microscopy images \citep{targosz2021semantic}. \textbf{(SemS-2)} Liver segmentation in \acf{CT} images \citep{antonelli2021medical, simpson2019large}. \textbf{(SemS-3)} Labeling of invasive/ non-invasive/ benign lesions in breast \acf{WSI} \citep{aresta2019bach}. \textbf{(SemS-4)} Cortical structure segmentation from 3D \acf{MRI} images\citep{budd2020surface}. \textbf{(SemS-5)} Aneurysm segmentation in \acf{TOF-MRA} images \citep{timmins2021comparing}. \textbf{(InS-1)} Instance segmentation of neurons from the fruit fly in 3D multi-color light microscopy images \citep{tirian2017vt, mais2020patchperpix, meissner2022searchable}. \textbf{(InS-2)} Surgical instrument instance segmentation in colonoscopy videos \citep{maier2021heidelberg}. \textbf{(InS-3)} Cell nuclei instance segmentation in time-lapse light microscopy for cell tracking \citep{ulman2017objective}. \textbf{(InS-4)} \acs{MS} lesion segmentation in multi-modal brain \acs{MRI} images \citep{kofler2022blob, commowick2018objective}.}
    \label{fig:scenario-s6}
\end{figure}

\begin{figure}[H]
    \centering
    \includegraphics[width=1\linewidth]{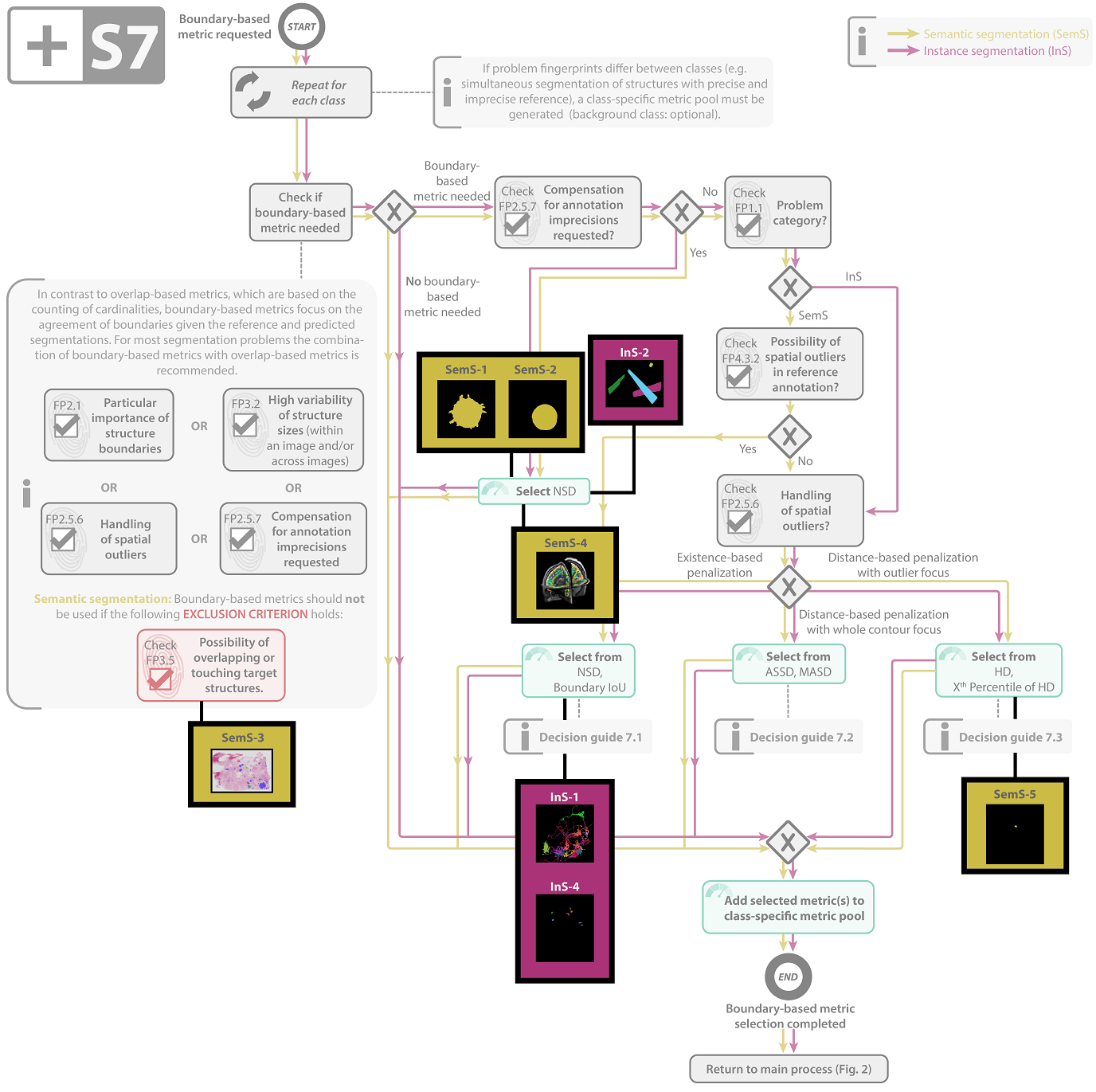}
    \caption{\textbf{Instantiation of Subprocess S7 for the selection of boundary-based metrics with recommendations for concrete biomedical problems.} \textbf{(SemS-1)} Embryo segmentation from microscopy images \citep{targosz2021semantic}. \textbf{(SemS-2)} Liver segmentation in \acf{CT} images \citep{antonelli2021medical, simpson2019large}. \textbf{(SemS-3)} Labeling of invasive/non-invasive/benign lesions in breast \acf{WSI} \citep{aresta2019bach}. \textbf{(SemS-4)} Cortical structure segmentation from 3D \acf{MRI} images\citep{budd2020surface}. \textbf{(SemS-5)} Aneurysm segmentation in \acf{TOF-MRA} images \citep{timmins2021comparing}. \textbf{(InS-1)} Instance segmentation of neurons from the fruit fly in 3D multi-color light microscopy images \citep{tirian2017vt, mais2020patchperpix, meissner2022searchable}. \textbf{(InS-2)} Surgical instrument instance segmentation in colonoscopy videos \citep{maier2021heidelberg}. \textbf{(InS-3)} Cell nuclei instance segmentation in time-lapse light microscopy for cell tracking \citep{ulman2017objective}. \textbf{(InS-4)} \acs{MS} lesion segmentation in multi-modal brain \acs{MRI} images \citep{kofler2022blob, commowick2018objective}.}
    \label{fig:scenario-s7}
\end{figure}

\begin{figure}[H]
    \centering
    \includegraphics[width=1\linewidth]{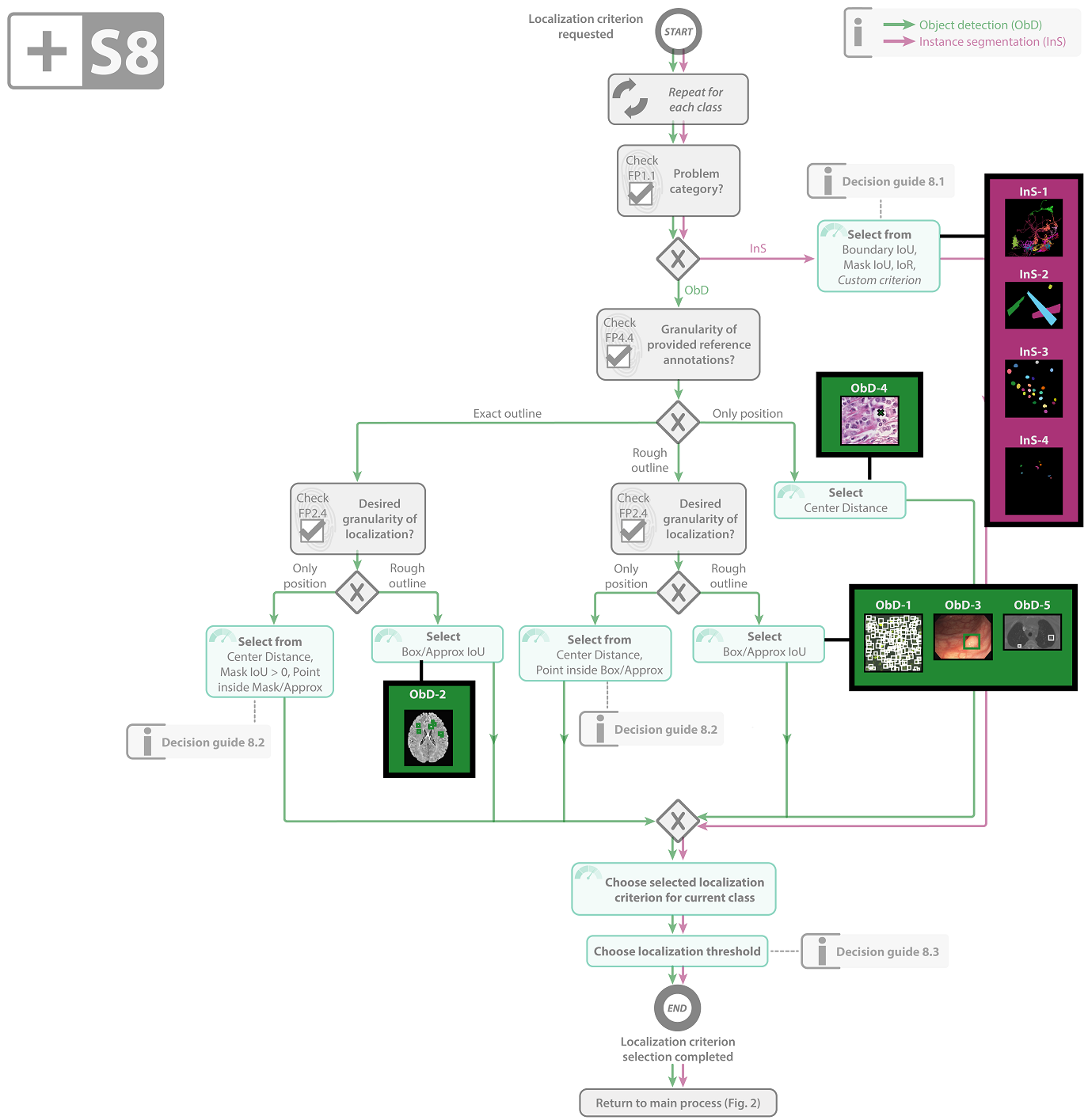}
    \caption{\textbf{Instantiation of Subprocess S8 for the selection of localization criteria with recommendations for concrete biomedical problems.} \textbf{(ObD-1)} Cell detection and tracking during the autophagy process in time-lapse microscopy \citep{zhang2020deepphagy, nagao2020robust}. \textbf{(ObD-2)} \acf{MS} lesion detection in multi-modal brain \acf{MRI} images \citep{kofler2022blob, commowick2018objective}.  \textbf{(ObD-3)} Polyp detection in  colonoscopy videos with predefined sensitivity of 0.9 \citep{sanchez2019computer, bernal2019gtcreator}. \textbf{(ObD-4)} Mitosis detection in histopathology images \citep{aubreville2022mitosis}. \textbf{(ObD-5)} Lung nodule detection in \acf{CT} images \citep{armato2011lung, armato2011lungdata, clark2013cancer}. \textbf{(InS-1)} Instance segmentation of neurons from the fruit fly in 3D multi-color light microscopy images \citep{tirian2017vt, mais2020patchperpix, meissner2022searchable}. \textbf{(InS-2)} Surgical instrument instance segmentation in colonoscopy videos \citep{maier2021heidelberg}. \textbf{(InS-3)} Cell nuclei instance segmentation in time-lapse light microscopy for cell tracking \citep{ulman2017objective}. \textbf{(InS-4)} \acs{MS} lesion segmentation in multi-modal brain \acs{MRI} images \citep{kofler2022blob, commowick2018objective}.}
    \label{fig:scenario-s8}
\end{figure}

\begin{figure}[H]
    \centering
    \includegraphics[width=1\linewidth]{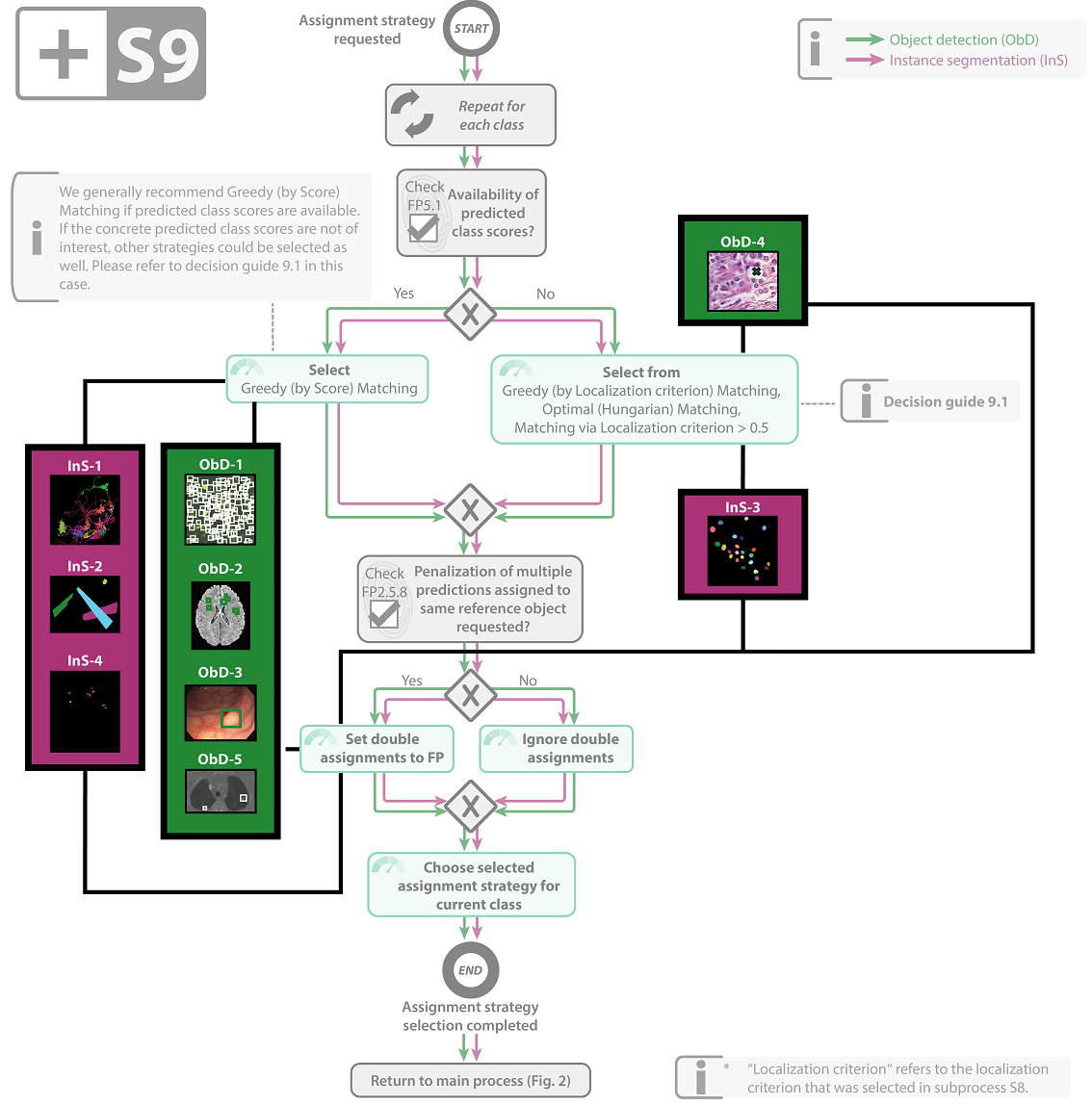}
    \caption{\textbf{Instantiation of Subprocess S9 for the selection of assignment strategies with recommendations for concrete biomedical problems.} \textbf{(ObD-1)} Cell detection and tracking during the autophagy process in time-lapse microscopy \citep{zhang2020deepphagy, nagao2020robust}. \textbf{(ObD-2)} \acf{MS} Lesion detection in multi-modal brain \acf{MRI} images \citep{kofler2022blob, commowick2018objective}.  \textbf{(ObD-3)} Polyp detection in  colonoscopy videos with predefined sensitivity of 0.95 \citep{sanchez2019computer, bernal2019gtcreator}. \textbf{(ObD-4)} Mitosis detection in histopathology images \citep{aubreville2022mitosis}. \textbf{(ObD-5)} Lung nodule detection in \acf{CT} images \citep{armato2011lung, armato2011lungdata, clark2013cancer}. \textbf{(InS-1)} Instance segmentation of neurons from the fruit fly in 3D multi-color light microscopy images \citep{tirian2017vt, mais2020patchperpix, meissner2022searchable}. \textbf{(InS-2)} Surgical instrument instance segmentation in colonoscopy videos \citep{maier2021heidelberg}. \textbf{(InS-3)} Cell nuclei instance segmentation in time-lapse light microscopy for cell tracking \citep{ulman2017objective}. \textbf{(InS-4)} \acs{MS} lesion segmentation in multi-modal brain \acs{MRI} images \citep{kofler2022blob, commowick2018objective}.}
    \label{fig:scenario-s9}
\end{figure}

The following use cases have been instantiated for instance segmentation problems. The resulting metric recommendations can be found in Fig.~\ref{fig:scenario-InS-overview}, while Figs.~\ref{fig:scenario-s3}-\ref{fig:scenario-s8} provide a detailed overview of the recommendations for the use cases in the metric selection Subprocesses S3 - S4, S6 - S9.
\begin{description}
    \item[InS-1] Instance segmentation of neurons from the fruit fly in 3D multi-color light microscopy images \citep{tirian2017vt, mais2020patchperpix, meissner2022searchable}
    \item[InS-2] Surgical instrument instance segmentation in colonoscopy videos \citep{maier2021heidelberg}
    \item[InS-3] Cell nuclei instance segmentation in time-lapse light microscopy for  cell tracking \citep{ulman2017objective}
    \item[InS-4] \ac{MS} lesion segmentation in multi-modal brain \ac{MRI} images \citep{kofler2022blob, commowick2018objective}
\end{description}

\newpage

\newpage
\textbf{Recommendations shown within the metric selection Subprocesses S2-S9}

\newpage

\newpage

\section{Terminology and Notation}
\label{app:terminology-notation}
\setcounter{figure}{0}

\subsection{Symbol References}
\label{app:symbols}

\begin{figure}[H]
    \centering
    \includegraphics[width=0.9\linewidth]{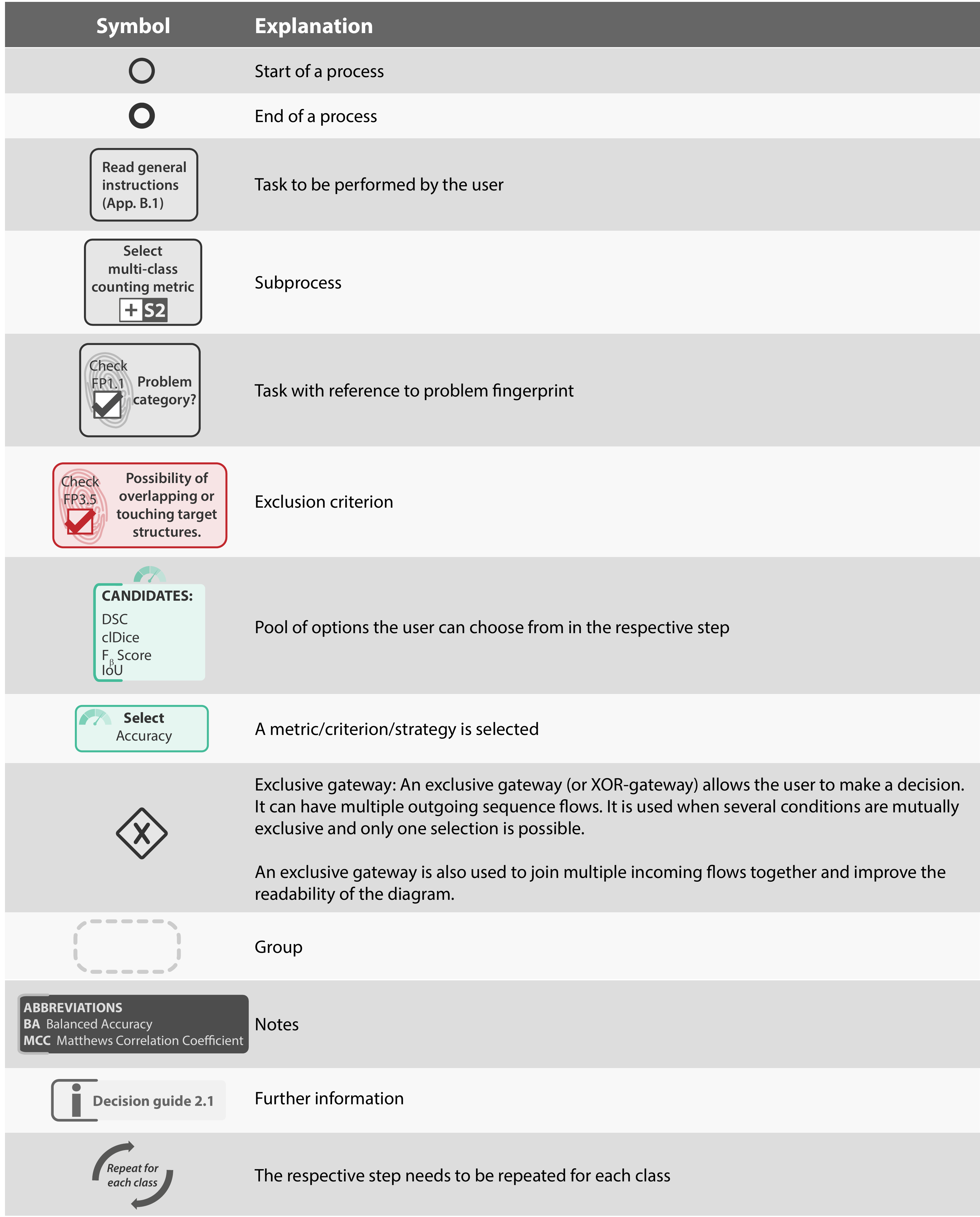}
    \caption{\textbf{Overview of symbols used in the process diagrams.} The notation used in the process diagrams originates from \acf{BPMN}.}
    \label{fig:process-diagram-symbols}
\end{figure}

%------------------------------------
%\subsection{Expected formats of reference and algorithm output}
%\label{app:formats}
%------------------------------------
\newpage
%------------------------------------
\subsection{Expected formats of reference and algorithm output}
\label{app:formats}
%------------------------------------

\begin{description}
    \item[Image-level Classification:] The metric mapping expects the following format for image-level classification with $C$ classes: For each image $I$ there is a reference annotation $y_I$ that either indicates the class for the image ($y_I \in \{1, ..., C\}$), or, in the case of multi-label classification, indicates presence for each class ($y_I \in \{0, 1\}^C$). If the algorithm does not provide predicted class scores (FP5.1 = FALSE), the algorithm output should be provided in an identical format. Otherwise, for each image $I$, the continuous class scores for each of the classes ($\hat{y}_I \in [0, 1]^C$), indicating the predicted class probability, should be provided. 
    \item[Semantic Segmentation] We assume the reference annotation and the algorithm output to be in the same coordinate system with identical spacing. The metric mapping expects the following format for semantic segmentation with $C$ classes: For each pixel $P$ there is a reference annotation $y_P$ that either assigns a single class to $P$ ($y_P \in \{1, ..., C\}$) or, in case of possible multiple labels per pixel, indicates assignment for each class ($y_P \in \{0, 1\}^C$). As for the algorithm output, for each pixel $P$ there is expected to be either a single prediction ($\hat{y}_P \in \{1, ..., C\}$) or, in case of multiple possible labels per pixel, a prediction for each class ($\hat{y}_P \in \{0, 1\}^C$). Some segmentation metrics require structure boundaries. For each class, boundaries are expected to be provided as a list of boundary pixels for both the reference and the prediction.
    %are defined only on the border pixels of the reference and predicted segmentation. Border pixels are defined as the set of all pixels of class $c$ that have an immediate neighbor not containing the class $c$.
    \item[Object detection:] The metric mapping expects the following format for object detection with $C$ classes: For each object $O$ the reference consists of a tuple $(y_O, l_O)$, where $y_O \in \{1, .., C\}$ indicates the class of the object and $l_O$ is some location information (box, center point, radius, etc.). The algorithm output for an object prediction $O$ is expected to comprise a tuple $(\hat{y}_O, \hat{l}_O)$ as well, where $\hat{y}_O$ indicates a single predicted class ($\hat{y}_O \in \{1, .., C\}$) optionally accompanied by an associated predicted class score ($\hat{c}_O \in [0, 1]$). Note that methods usually provide a predicted class score for the background class as well, but this score is typically discarded in validation as there are no "background objects"~\cite{everingham2015pascal}. See FP5.1 in case no predicted class score is provided. $\hat{l}_O$ is expected to provide location information about the prediction in a similar format as the reference (box, center point, radius, etc.). In case reference objects are represented by rough outlines (FP4.3) we assume that the chosen shapes (e.g., bounding box or ellipsoid) represent the underlying object adequately. Particular attention needs to be given to this aspect if objects feature a tubular shape (FP3.3) or can potentially appear disconnected (FP3.6). 
     \item[Instance Segmentation] The metric mapping expects the following format for instance segmentation with $C$ classes: For each object $O$ the reference consists of a tuple $(y_O, m_O)$, where $y_O \in \{1, .., C\}$ indicates the class of the object and $m_O \in \{0, 1\}^{H \times W}$ is a binary pixel map per instance matching the size of the image (height $H$ and width $W$) and indicating pixel-wise location. The algorithm output for an object prediction $O$ is expected to comprise a tuple $(\hat{y}_O, \hat{m}_O)$, where, similarly to object detection, $\hat{y}_O$ indicates a single predicted class ($\hat{y}_O \in \{1, .., C\}$) optionally accompanied by an associated predicted class score ($\hat{c}_O \in [0, 1]$). $\hat{m}_O$ denotes a binary pixel map per instance analogously to  $m_O$. For both the reference and the predictions, structure boundaries should be provided as a list of boundary pixels separately for each instance. Note that annotations from semantic segmentation (not distinguishing instances of the same class) can be transformed to the instance segmentation format via connected component analysis (in case of purely non-touching and connected instances).
\end{description}
In case the provided reference annotations deviate from the expected format, matching can be achieved via various measures (e.g., aggregation of pixel-level reference to required image-level reference). 

%\newpage
\subsection{Acronyms}
\label{app:acronyms}

\newpage

%------------------------------------
\subsection{Glossary}
\label{app:terminology}
%------------------------------------
\begin{itemize}
    \item \textbf{Bounding box:}
    A bounding box is a rectangle, typically the smallest possible, drawn around and completely surrounding an object to be detected.
    \item \textbf{Calibration plot:} A calibration plot, also referred to as reliability diagram, is a visualization of the calibration ability of a model's outputs (see e.g.,~\cite{guoCalibrationModernNeural2017}). Specifically, the diagram allows to diagnose a model's general bias towards "overconfident" or "underconfident" predictions by visualizing the deviation from perfect calibration (diagonal line in the plot) for different output scores. The diagram also acts as the basis for further diagnostic measurements such as the calibration slope.   
    \item \textbf{Challenge:}
    A challenge is an international competition, commonly hosted by individual researchers, an institute, or a professional society, that aims to comparatively assess the performance of competing algorithms on an identical data set, and thus serves to validate them. This validation is a crucial step towards the translation of an algorithm into practice.
    \item \textbf{Classification task:}
    A classification task is the task of giving categorical labels to an image or parts thereof. We distinguish classification at different scales, e.g., at image level, pixel level or object level.
    \item \textbf{Confidence:} See Predicted class scores.
    \item \textbf{Continuous class scores:} See Predicted class scores.
    \item \textbf{Decision rule:} A rule transforming continuous predicted class scores into discrete classification decisions. This rule amounts to setting a simple cutoff value in binary classification problems but is more complex to define in multi-class problems (for more information see Suppl. Note~\ref{app:general-instructions}).
    \item \textbf{Evaluation:} See Validation.
    \item \textbf{Hierarchical structure of classes/data:}
   A hierarchical structure of classes/data is present when classes or data are dependent on each other or paired, e.g., when data have been derived from the same patient, or from the same center. It requires interpretation and statistical efforts different from those suitable for independent data.
    \item \textbf{Hyperparameter:}
    A hyperparameter is a parameter whose value is optimized to control the training of an algorithm. In contrast to other parameters, it is not derived through the training process itself, but rather set before the training procedure.
    \item \textbf{Inference:} In the context of ML, inference denotes the processing of data by an algorithm to produce the desired output.
    \item \textbf{Instance:}
    An instance refers to a dedicated object, structure or entity in an image, such as an individual cell, tumor or medical instrument.
    \item \textbf{Image-level classification:}
    Image-level classification is the assignment of one or multiple category labels to an entire image, as detailed in Suppl. Note~\ref{app:general-instructions}. 
    \item \textbf{Instance segmentation:}
    Detection and delineation of each distinct object of a particular class in an image, as detailed in Suppl. Note~\ref{app:general-instructions}.
    \item \textbf{Instantiation:} Instantiation here refers to the act of creating a specific application case of a general principle/framework.
    \item \textbf{Macro/micro averaging:} Macro averaging is the process of computing a metric (e.g., Sensitivity) for each class and subsequently averaging the metric scores. Micro averaging is the process of aggregating an average metric score over all classes.
    \item \textbf{Meta-information:}
    Meta-information refers to data about an image that is not explicitly contained within the image, e.g., \ac{PHI} data about the patient in radiology images.
    \item \textbf{Metric:}
    Metrics are the measures according to which performance of algorithms is quantified and validated. Depending on the domain-specific validation goal and property of interest, we distinguish between different types of metrics, e.g., reference-based vs. non-reference based (see Reference/Reference-based metrics). Metrics can further be subdivided into different families based on their mathematical properties.
     \item \textbf{Metric@(TargetMetric = TargetValue):} (e.g., Specificity@(Sensitivity = 0.95)): Once a cutoff value for the predicted class probabilities has been set in such a way that the target metric value is achieved (here: target metric Sensitivity with a target value of 0.95), other metric values (here: Specificity) are obtained from the corresponding fixed confusion matrix. In the example, this yields the Specificity at the predefined Sensitivity level.
    \item \textbf{Object detection:}
   Detection and localization of structures of one or multiple categories in an image, as detailed in Suppl. Note~\ref{app:general-instructions}.
    \item \textbf{(Output) Calibration:} In application scenarios that involve interpreting the raw algorithm output (specifically the  predicted class scores), output calibration can be used to obtain a reliable measure of confidence associated with the decision (see description of FP2.7 in Suppl. Note.~\ref{app:fingerprint-category}).
    \item \textbf{Image-level classification:} Assignment of one or multiple category labels to the entire image or fixed regions/predefined locations within an image. 
      \item \textbf{Instance Segmentation:} Localization and delineation of each distinct structure of a particular class. It can be regarded as delivering the tasks of object detection and semantic segmentation at the same time. In contrast to object detection, instance segmentation also involves the accurate marking of the structure boundary. In contrast to semantic segmentation, it distinguishes different structures of the same class.
     \item \textbf{Object detection:} Localization and categorization of an unknown number of structures, as detailed in Suppl. Note~\ref{app:general-instructions}  
    \item \textbf{Precision:}
    Precision is a term used differently in different scientific communities. In the medical community, for example, it commonly refers to the confidence of an output. Here, we use the term to denote the \ac{PPV}.
    \item \textbf{Predicted class scores:} Modern neural network-based approaches usually output predicted class scores (also referred to as continuous class scores, confidence scores or pseudo-probabilities) between 0 and 1 for every image/object/pixel and class, indicating the probability of the image/object/pixel belonging to a specific class.
    \item \textbf{Prediction:}
    Prediction refers to the output of an algorithm. It is not used in the temporal sense in this paper.
    \item \textbf{Problem category:}
    Biomedical image analysis problems can be subdivided into problem categories according to the procedures performed. The category a problem falls into informs the appropriate choice of metrics. In this paper, we focus on four problem categories: Image-level classification, Semantic Segmentation, Object Detection, and Instance Segmentation. 
    \item \textbf{Pseudo-probabilities:} See Predicted class scores.
    \item \textbf{Reference/Reference-based metrics:} We assume that the validation process is based on the comparison of the algorithm output and a \textbf{reference} (sometimes called \textbf{gold standard}), which is assumed to be close or equal to the correct result -- the (often forever unknown) \textbf{ground truth}. In terms of metrics, we distinguish between \textit{reference-based metrics}~\cite{jannin2006model}, which use the image-based reference, and \textit{non-reference-based metrics} that assess complementary properties, such as runtime, memory consumption, or carbon footprint.
    \item \textbf{Reliability diagram:} See calibration plot.
    \item \textbf{Semantic segmentation:}
   Assignment of one or multiple category labels to each pixel in an image, as detailed in Suppl. Note~\ref{app:general-instructions}.
    \item \textbf{Structure instance:} See Instance.
    \item \textbf{Training/Test case:}
    The data sets used in the process of algorithm development and validation comprise training/test cases. A case refers to the data (typically an n-dimensional image, possibly enhanced with clinical context information) that is required for an algorithm to produce one result (e.g., a segmentation or classification). A training case refers to a data set that includes reference annotations and is thus used for training an algorithm. A test case refers to a data set that is used for performance assessment
     \item \textbf{Type 1 and Type 2 error:} A type 1 error is a \ac{FP} result, e.g., a false detection of something that is not present. A type 2 error is a \ac{FN} result, e.g., a non-detection of something that is present.
    \item \textbf{Validation:}
    Validation is the process of assessing that the validated algorithm is effectively doing what it is expected to do and what it was developed for, for example that a segmentation method is actually segmenting. Evaluation is the process of assessing that the algorithm is valuable, i.e., that it brings quantifiable added value for the clinical user in a dedicated clinical context \cite{iso2000quality}. 
\end{itemize}
\newpage

% \section{Full Author Affiliations}
% \label{app:authors}
% \printauthors

% \renewcommand{\bibsection}{\section*{References}}
% \addcontentsline{toc}{chapter}{\protect\numberline{}References}

%%

\end{document}